\documentclass{article}

\usepackage[nonatbib,final]{neurips_2025}

\usepackage[table,xcdraw, dvipsnames]{xcolor}
\usepackage{multirow}
\usepackage[utf8]{inputenc} 
\usepackage[T1]{fontenc}    
\usepackage[hyphens]{url} 
\usepackage[
  colorlinks=true,
  allcolors=blue,
  pdfborder={0 0 0}  
]{hyperref} 
\usepackage{booktabs}       
\usepackage{amsfonts}       
\usepackage{nicefrac}       
\usepackage{microtype}      
\usepackage{graphicx}
\usepackage{amsmath, amsfonts, amssymb, amsthm}
\usepackage{mathtools}
\usepackage{bbm}
\usepackage[numbers,sort,compress]{natbib}
\usepackage{wrapfig}

\usepackage{placeins}   
\usepackage{needspace}  
\usepackage{float}      
\usepackage{adjustbox}  
\usepackage{caption}    

\usepackage{subcaption}
\usepackage{listings}
\usepackage{array}        
\usepackage[english]{babel}
\usepackage{enumitem}

\usepackage{amsmath,amsfonts,bm}

\newcommand{\newterm}[1]{{\bf #1}}

\def\eqref#1{equation~\ref{#1}}

\def\1{\bm{1}}

\def\vz{{\bm{z}}}

\DeclareMathAlphabet{\mathsfit}{\encodingdefault}{\sfdefault}{m}{sl}
\SetMathAlphabet{\mathsfit}{bold}{\encodingdefault}{\sfdefault}{bx}{n}

\def\sZ{{\mathbb{Z}}}

\DeclareMathOperator*{\argmax}{arg\,max}

\usepackage{amsthm}

\theoremstyle{plain}

\theoremstyle{definition}
\newtheorem{definition}{Definition}[section]

\theoremstyle{definition}

\theoremstyle{definition}

\theoremstyle{remark}

\theoremstyle{plain}

\usepackage{longtable,booktabs}
\usepackage{makecell}
\usepackage{titletoc}

\newcommand{\custompar}[1]{\vspace{.1cm} \noindent{\bf #1.}\:}

\definecolor{seabornbluemid}{HTML}{08519C}
\definecolor{seabornorangemid}{HTML}{F16913}
\definecolor{seaborngreymid}{HTML}{737373}
\definecolor{seabornredmid}{HTML}{EF3B2C}

\newcommand{\llmtext}[1]{\texttt{\footnotesize{"#1"}}}

\newenvironment{customquote}
  {\list{}{\leftmargin=.35cm \rightmargin=.35cm}  
   \item\relax
   \footnotesize    
    } 
  {\endlist}

\makeatletter
\@removefromreset{definition}{section}
\makeatother

\title{Extracting memorized pieces of (copyrighted) books \\ from open-weight language models}

\author{%
  A. Feder Cooper\thanks{Corresponding authors: \texttt{afedercooper@gmail.com}, \texttt{mlemley@law.stanford.edu}} \\
  Stanford University \\ Yale University \\
  \And 
  Mark A. Lemley\\
  Stanford University\\
  \And
  Allison Casasola\\
  Stanford University\\
  \AND
  Ahmed Ahmed\\
  Stanford University\\
  \And
  Aaron Gokaslan \\
  Cornell University \\
  \And 
  Amy B. Cyphert\\
  West Virginia University\\
  \AND 
  Christopher De Sa\\
  Cornell University\\
  \And 
  Daniel E. Ho\\
  Stanford University\\
  \And
  Percy Liang\\
  Stanford University
}

\begin{document}

\maketitle

\vspace{-.2cm}
\begin{abstract}
\vspace{-.1cm}
Plaintiffs and defendants in copyright lawsuits make sweeping, opposing claims about the extent to which large language models (LLMs) memorize protected expression from books in their training data. 
We show that these polarized positions dramatically oversimplify the relationship between memorization and copyright. 
To do so, we develop a technique to measure memorization of books, which we apply  to $200$ books and $14$ open-weight LLMs. 
Through over $3000$ experiments, we show that memorization varies both by model and book. 
With respect to our specific measurement methodology, most LLMs do not memorize most books either in whole or in part, but there are notable exceptions. 
For instance, \textsc{Llama 3.1 70B} entirely memorizes some books, like \emph{Harry Potter and the Sorcerer's Stone}; 
memorization is so extensive that one can deterministically extract the whole book almost verbatim using the book's first few words as an initial prompt. 
We discuss why our results have significant implications for copyright cases, though not ones that unambiguously favor either side.\looseness=-1 
\end{abstract}
\vspace{-.2cm}

\setcounter{footnote}{0}

\section{Introduction}\label{sec:intro}

In the dozens of copyright lawsuits over training large language models (LLMs), both sides tend to present opposing, simplified interpretations of the underlying technology. 
Plaintiffs---such as book authors---say that LLMs are just giant copy machines that store (infringing) copies of their specific works and recombine them in their outputs~\citep{kadrey}. 
Defendants---typically tech companies---claim that LLMs merely contain non-expressive ``statistical correlations''~\citep{concordanthropic}, and therefore do not copy any particular creative work.
We show that the situation is more complicated than either side suggests.\looseness=-1 

During training, an LLM learns probabilistic relationships across the training dataset as a whole.
This informal intuition aligns with the non-expressive ``statistical correlations'' that defendants describe as being encoded within an LLM's weights. 
However, sometimes, for some training data, these learned patterns are not so high-level or abstract; 
rather, they are highly specific~\citep{cooper2024files}. 
For reasons that are not fully understood, certain training data are not so transformed by training:
they are ``memorized'' by the model---encoded in some form inside the LLM's weights~\citep{carlini2023quantifying, lee2023talkin, cooper2024files, carlini2025blog}. 
Such memorized content can sometimes be ``extracted'' later; 
it can be reproduced in an LLM's outputs at generation time~\citep{carlini2021extracting}.\looseness=-1

Models do not memorize all of their training data. 
An LLM may therefore memorize parts of some plaintiffs' copyrighted works, but not others'. 
When such memorization occurs, it involves copying that may have consequences for copyright. 
Extraction generates a ``copy'' of training data, but it also demonstrates that training data is memorized inside the model itself.
If that model is a ``copy'' (in a technical sense that copyright cares about), this also has important implications. 
Notably, a model---not just extracted training data---could be deemed an infringing ``copy'' of the training data it has memorized~\citep{cooper2024files, lee2023talkin}. 
Copyright law offers the destruction of infringing materials as a remedy.
So, just as courts have ordered the destruction of bootleg DVDs, they could order the destruction of infringing models~\citep{lee2023talkin, pamscience, townsend2024deletion}.\looseness=-1

When seeking such remedies, plaintiffs often cite technical papers that demonstrate memorization by extracting training data from open-weight models and production systems. 
But this reasoning is flawed for a key reason: 
prior research does not quantify the kinds of information most relevant to a copyright infringement claim~\citep{cooper2024files}.
This work typically shows that on \emph{average} LLMs memorize some portion of their training data; 
in contrast, copyright generally evaluates infringement with respect to \emph{specific} works.\looseness=-1

To address this gap, we propose (Section~\ref{sec:book-procedure}) and validate (Section~\ref{sec:validity}) an  extraction procedure that surfaces the extent to which specific open-weight LLMs memorize specific books. 
The procedure uses a sliding window to reveal how much and where memorization occurs within a book. 
For our main experiments, we run this procedure on $14$ open-weight LLMs and $200$ books from the \texttt{Books3} dataset, a (now notorious) 
torrented corpus of nearly $200{,}000$ books~\citep{gao2020pile, reisnerbooks3, lee2023talkin}. 
\texttt{Books3} is known to be included in the training data of several LLMs~\citep{biderman2023pythia, touvron2023llamaopenefficientfoundation, llama2, llama3}, and has been a focus of copyright  litigation~\citep{reisnerbooks3, 
kadreyamendedconsolidated}. 
With respect to our extraction methodology, most of the LLMs we test do not memorize most of the $200$ books, either in whole or in part (Section~\ref{sec:book:compare}). 
However, \textsc{Llama 3.1 70B} memorizes some books, like \emph{Harry Potter and the Sorcerer's Stone}~\citep{Harry_Potter_and_the_Sorcerer_s_Stone} (hereafter, ``\emph{Harry Potter}'') and \emph{1984}~\citep{Nineteen_Eighty-Four}, almost entirely. 
In fact, \emph{Harry Potter} is so memorized that one can deterministically extract a near-perfect copy of the book, using an initial prompt consisting of just the first few words of the first chapter (Section~\ref{sec:book:seed}). 
Last, we discuss the significant implications of our findings for copyright cases, though not ones that unambiguously favor either side. 
Our results complicate defendants' fair use story, but they also complicate plaintiffs' efforts to bring class action lawsuits (Section~\ref{sec:takeaways}).\looseness=-1

In summary, the paper proceeds as follows:\looseness=-1 
\begin{itemize}[itemsep=0cm, topsep=0cm, leftmargin=.65cm]
	\item \textbf{Section~\ref{sec:background}.}  
    We provide relevant background on memorization and extraction risk.
    \item \textbf{Section~\ref{sec:copyright}.} 
    We connect memorization and extraction risk to ``copies'' in U.S. copyright law. 
    \item \textbf{Section~\ref{sec:book-procedure}.} 
    We present our sliding-window procedure, which computes extraction risk throughout a book, and enables \emph{specific} claims about how much an LLM has memorized \emph{specific} books.\looseness=-1 
    \item \textbf{Section~\ref{sec:validity}.}
    We validate that our measurements capture true instances of extraction (and thus memorization). 
    \item \textbf{Section~\ref{sec:book:compare}.} 
    We then run experiments for $14$ LLMs on a set of $200$ books in \texttt{Books3}, which surface novel insights about how book-specific memorization varies across models and books.\looseness=-1 
    \item \textbf{Section~\ref{sec:book:seed}.} 
    We attempt to generate a book that is entirely memorized. 
    For this, we select \emph{Harry Potter} and \textsc{Llama 3.1 70B}, and we successfully extract a near-perfect copy.\looseness=-1 

    \item \textbf{Section~\ref{sec:takeaways}.} 
    We offer key takeaways for copyright law and policy. 
\end{itemize}
Despite the extent of our analysis, the main paper is relatively brief. 
The length of this document is due to an extensive appendix containing results and extended discussion for all experiments. 
We also publish a \href{https://books-memorization.github.io}{website} version of the per-book results presenetd in Appendix~\ref{app:sec:sliding-window:results}.
\vspace{-.2cm}
\section{Quantifying memorization through extraction}\label{sec:background}

Extraction and memorization are related concepts, but differ in subtle ways. 
While \newterm{extraction} refers to reconstructing specific training data in a model's  generated outputs, \newterm{memorization} is broader: 
it involves reconstructing specific training data by examining the model through any means~\citep{cooper2023report, cooper2024files}. 
Successful extraction of a sequence of training data in a model's generation is
evidence of memorization of training data \emph{inside the model}~\citep{cooper2024files, carlini2025blog, schwarzschild2024rethinking, lee2023talkin, feldman2020mem}.
This is uncontroversial in machine learning (ML). 
As Carlini explains, when a sufficiently large and unique piece of training data is extracted, ``the only possible explanation is that the model has
somewhere internally stored [that piece of training data]''~\citep{carlini2025blog}.
As Cooper and Grimmelmann note, ``in order to be able to extract memorized content from a model at generation time, that memorized content must be encoded in the model's parameters. 
There is nowhere else it could be. A model is not a magical
portal that pulls fresh information from some parallel universe into our own''~\citep[p. 25]{cooper2024files}.

LLMs do not memorize all of their training data, but they do memorize some of it. 
And so, an LLM may memorize parts of some plaintiffs' copyrighted works, but not others'. 
To show this, we first need more background on the extraction metric we use in most of our experiments (Section~\ref{sec:background:pz}), and we also provide guidance on that metric's interpretation (Section~\ref{sec:background:pz-meaning}).\looseness=-1

\subsection{Probabilistic discoverable extraction}\label{sec:background:pz}

\begin{figure*}[t]
    \vspace{-.1cm}
    \centering
    \begin{subfigure}[t]{0.5\textwidth}
    \vspace{0pt}
        \centering
        \includegraphics[width=\linewidth]{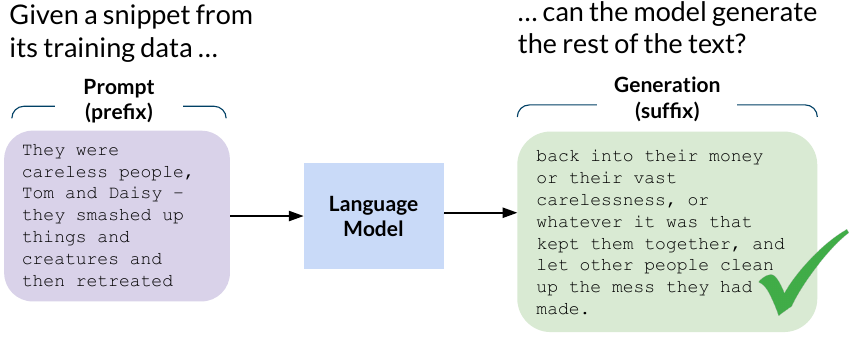}
        \label{fig:panelA}
    \end{subfigure}
    \hfill
    \begin{subfigure}[t]{0.46\textwidth}
    \vspace{0pt}
        \centering
        \includegraphics[width=\linewidth]{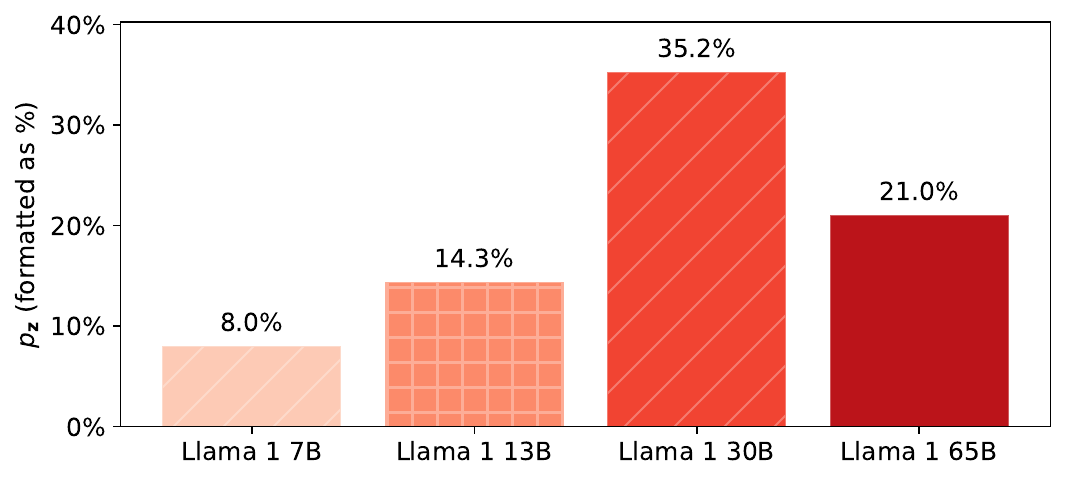}
    \end{subfigure}
    \vspace{-.2cm}
    \caption{\textbf{Quantifying memorization for a sequence drawn from \emph{The Great Gatsby}.} 
    (\textbf{left}) Discoverable extraction~\citep{lee2022dedup, carlini2023quantifying} prompts an LLM with a training-data prefix, and checks if the resulting generation exactly matches the ground-truth suffix. 
    (\textbf{right}) Probabilistic extraction~\citep{hayes2025measuringmemorizationlanguagemodels} measures the probability $p_\vz$ (\%) of generating the exact suffix given the prefix, shown here for 
    \textsc{Llama 1} models on the same quote from \emph{The Great Gatsby}. 
    \textsc{Llama 1 30B} exhibits the highest extraction risk.\looseness=-1 
    }
    \label{fig:header}
    \vspace{-.2cm}
\end{figure*}

Extraction of training data in a model's outputs is the most common form of evidence used to quantify memorization in LLMs~\citep{nasr2025scalable,hayes2025measuringmemorizationlanguagemodels,prashanth2024recite, carlini2023extracting, lee2022dedup, zhang2023counterfactual, cooper2026nv}.
The standard metric in research and frontier model release reports~\citep{reid2024gemini, gemma2, llama3} measures verbatim \newterm{discoverable extraction}: 
given a sequence of text from the training data, split it into a \newterm{prefix} and \newterm{(target) suffix}, prompt an LLM with the prefix, and deem extraction successful if the LLM reproduces the suffix \emph{exactly} (Figure~\ref{fig:header}, left; \citet{lee2022dedup,carlini2023quantifying}).  
This type of method is well-suited to \newterm{completion} models that continue the prompt text, rather than instruction-tuned \newterm{chatbots} that reply in a conversational style (Appendix~\ref{app:sec:intro}).
As is standard, we therefore only study completion LLMs, and refer to these LLMs by name, version, and size (e.g., \textsc{Llama 1 30B}).\looseness=-1 

Discoverable extraction reduces memorization to a binary (yes-or-no) outcome by issuing a single prompt to the LLM and checking whether the deterministic, \newterm{greedy-decoded} generation matches the target suffix. 
But LLMs are probabilistic: 
under realistic, \newterm{non-deterministic decoding} schemes (e.g., top-$k$ sampling), 
an LLM can produce different outputs in response to the same prompt. 
As \citet{cooper2024files} note, a copyright-relevant claim is therefore more likely to be concerned with \emph{how often} a target suffix can be extracted in practice, not just \emph{if} it can be extracted at all, i.e.,\looseness=-1
\begin{customquote}
    \vspace{-.15cm}
    a model, when (a) given a particular type of input, will (b) produce a particular type of memorized output, (c) with a particular probability. 
    That probability could be \ldots 
     1\% 
    \ldots{} [or] \ldots{}  
    35\% 
    \ldots. 
    The issue for copyright law \ldots is what to do 
    \ldots with the fact that element (c)---the probabilistic element---is inescapable \citep{cooper2024files}.\looseness=-1
    \vspace{-.15cm}
\end{customquote}

We elaborate on copyright-specific aspects of this point in Sections~\ref{sec:copyright} and~\ref{sec:takeaways}.
For now, we note that \citet{hayes2025measuringmemorizationlanguagemodels} introduce a metric called \newterm{probabilistic discoverable extraction} (hereafter, ``probabilistic extraction''), which captures this notion. 
For a training sequence $\vz$ of length $a\!+\!j$, one prompts with the $a$-token prefix $\vz_{1:a}$ and computes the probability $p_\vz\in[0,1]$ of the LLM $\theta$ generating the exact $j$-token suffix $\vz_{a+1:a+j}$ when using decoding scheme $\phi$. 
That is,
\begin{align}
\label{eq:pz}
p_\vz \triangleq 
\Pr_{\theta,\phi}\!\big(\vz_{a+1:a+j}\,\big|\,\vz_{1:a}\big) 
=\prod_{t=a+1}^{a+j} \Pr_{\theta,\phi}\!\big(\vz_t \,\big|\, \vz_{1:t-1} \big)  
= \exp\!\bigg(
  \sum_{t=a+1}^{a+j}
  \log \Pr_{\theta,\phi}\!\big(\vz_t \,\big|\, \vz_{1:t-1}\big)
\bigg). 
\end{align}
This equation simply captures the probability of the LLM generating the exact suffix $\vz_{a+1:a+j}$, obtained by multiplying the conditional probabilities of each token $z_t$ given the prefix $\vz_{1:a}$ and any earlier suffix tokens.\footnote{For numerical stability, we compute this as $\exp$ of the sum of the token conditional $\log$ probabilities.}  
The decoding scheme determines which tokens are considered at each generation step $t$, and how their 
logits are normalized into conditional probabilities (Section~\ref{sec:book-procedure:method} \& Appendix~\ref{app:sec:background:metrics}).\looseness=-1 

In Figure~\ref{fig:header} (right), we plot $p_\vz$ for a quote from \emph{The Great Gatsby} (Figure~\ref{fig:header}, left) using top-$k$ decoding (temperature $T\!=\!1$, $k\!=\!40$) with \textsc{Llama 1} models, which were trained on \emph{The Great Gatsby}~\citep{touvron2023llamaopenefficientfoundation}.
The probabilities for this sequence are enormous. 
In general, \(p_{\vz}\) is a product of many numbers between $0$ and $1$, so even relatively short suffixes should have small values---tending toward $0$ as sequence length increases.
For scale, a $50$-token suffix where each token has probability $90\%$ yields overall $p_\vz\!=\!0.9^{50}\approx\!0.5\%$.\footnote{A $100$-token suffix where each token's conditional probability is $90\%$ would yield \(p_\vz=0.9^{100}\!\approx\!0.0027\%\)---roughly $250\times$ less likely than a $50$-token sequence with the same per-token conditional probabilities.}
The exception is when the LLM assigns unusually high probability to each token in the sequence. 
For 
training data, this almost always happens only when the LLM has memorized those 
training data (Appendices~\ref{app:sec:background:metrics} \&~\ref{app:sec:validity}).

\custompar{Determining extraction success}
For traditional discoverable extraction, success means the greedy, \emph{deterministic} continuation exactly matches the target suffix. 
For probabilistic extraction (Equation~\ref{eq:pz}), we must decide how high a \emph{probability} is enough for success. 
For a training sequence 
$\vz$ and threshold $\tau_\text{min}\in(0,1]$, we deem extraction successful if \(p_{\vz} \geq \tau_{\text{min}}\): 
\begin{align} \label{eq:probsuccess:main} 
s(\vz;\tau_{\text{min}}) \;=\; \mathbf{1}\!\big[\,p_{\vz} \geq \tau_{\text{min}}\,\big]. \end{align} 
In our main experiments, we always use a (standard) suffix length of $50$ tokens, top-$k$ decoding ($T\!=\!1$, $k\!=\!40$), and a conservative $\tau_\text{min}\!=\!0.1\%$.  
(The geometric mean of the suffix tokens for this conditional probability is $(0.1\%)^{\frac{1}{50}}\!\approx\!87.1\%$.)
We discuss our choice of decoding scheme in detail in Appendix~\ref{app:sec:sliding-window:procedure}, and the validity of our choice of $\tau_\text{min}$ in Section~\ref{sec:validity}.
In practice, no repeated sampling is required to compute $p_\vz$. 
We run one \newterm{teacher-forced} forward pass through the LLM on \(\vz\), obtain the logits, apply \(\phi\) at each suffix token to get the conditional $\log$ prob, and apply Equation~\ref{eq:pz} (Appendix~\ref{app:sec:background:compute}).\looseness=-1

\subsection{Interpreting target suffix probabilities as extraction risk}\label{sec:background:pz-meaning} 

Over independent prompts using the same prefix and decoding scheme, \(p_{\vz}\) (Equation~\ref{eq:pz}) is the expected fraction of prompts that result in the LLM reproducing the verbatim target suffix. 
Therefore, \(p_{\vz}\!=\!50\%\) means the LLM will output roughly $1$ verbatim suffix every $2$ prompts with the prefix; \(p_{\vz}\!=\!0.1\%\) (our choice for $\tau_{\text{min}}$) is about $1$ in \(1000\). 
(Of course, \(p_{\vz}\!=\!0\%\) means the suffix will never be generated given the prefix, no matter how many times one were to use it as a prompt to the LLM.) 

We can therefore interpret \(p_{\vz}\) as a measure of \newterm{extraction risk}~\citep{hayes2025measuringmemorizationlanguagemodels}. 
For fixed sequence $\vz$ and fixed decoding scheme \(\phi\), this enables cross-LLM comparisons: 
for LLMs \(A\) and \(B\), if \(p_{A, \vz} > p_{B, \vz}\), 
then LLM \(A\) will in expectation reproduce the target suffix of $\vz$ more often than LLM \(B\) when prompted with that same prefix.
For instance, all of the probabilities in 
Figure~\ref{fig:header} (right) exceed $\tau_\text{min}\!=\!0.1\%$; 
these probabilities indicate that every \textsc{Llama 1} model memorized the quote.
But the quote is not as easily extractable from each LLM. 
\textsc{Llama 3.1 30B} exhibits the highest extraction risk: 
in expectation, it would take fewer than $3$ independent prompts 
(i.e., $\nicefrac{1}{0.352}\approx2.84$) with the prefix to extract the verbatim target suffix.\looseness=-1

\section{Memorization, extraction risk, and U.S. copyright}\label{sec:copyright}

The dozens of pending copyright lawsuits concerning LLMs~\citep{chatgptiseating}  generally present three interrelated issues: 
(1) whether training an LLM on copyrighted material is permitted \newterm{fair use} (i.e., limited use of the copyrighted material can, under certain circumstances, be used without permission from the copyright owner); 
(2) whether the model itself is a \newterm{copy} or \newterm{derivative work}  (technical terms in copyright law) 
of the works on which it is trained (Section~\ref{sec:takeaways}); 
and (3) whether the model reproduces copyrighted material in its outputs.
Some suits present only one issue, while others present all three questions.  
Some suits are based on content owned by a single company, while others are class action lawsuits purporting to represent all book authors.\footnote{There are also other copyright issues~\citep{lee2023talkin, pamscience, sag2023safety, goodyear2025infringement, sobelstyle, cooper2024unlearning}, for example, whether novel outputs of generative-AI systems are themselves copyrightable~\citep{lee2023talkin, uscopyrightofficereport2}.} 
These lawsuits have primarily focused on (1) training and on (3) model outputs, and the two decisions so far in the U.S. have held that training an LLM can be fair use, albeit with significant limitations~\citep{kadreyjudgment, bartzjudgment}. 
They also raise issues about the acquisition, storage, or use of \newterm{shadow libraries} of copyrighted material as training data, such as \texttt{Books3}~\citep{reisnerbooks3,lee2023talkin}.\looseness=-1  

Our work does not speak to the first issue; 
training and fair use have been discussed extensively elsewhere~\citep{lemley2021fairlearning}.
Rather, our findings affect the other two issues:
we show that copies are sometimes made in the model itself (i.e., memorized) and  those copies can sometimes be generated as outputs (i.e., extracted).
Because many LLMs are known to have been trained on \texttt{Books3}, we use books from this dataset to study memorization and extraction of copyrighted material (Sections~\ref{sec:book-procedure}--\ref{sec:book:compare}).\looseness=-1 

\textbf{Memorization and copies inside the model.}
If an LLM memorizes all or a substantial portion of a copyrighted work (near-)verbatim, the LLM itself may be or contain an infringing copy or derivative work~\citep{lee2023talkin, cooper2024files}.
This is because memorized training data are encoded inside the model (Sections~\ref{sec:intro} \&~\ref{sec:background}).
Others have argued that  encoding the work in the form of model weights satisfies the technical definition of ``copy'' in the U.S. Copyright Act~\citep{cooper2024files}.
In a technical sense, that copy may be of interest to copyright even if there is no practical way to access or extract it.
Like the copies of copyrighted works used as training data during training, that copy in model might be fair use, but the analysis of fair use would look somewhat different than \emph{internal} use of copyrighted works in training processes.  
This is particularly true for open-weight models like \textsc{DeepSeek} and \textsc{Llama}, which are not merely used internally by their developers, but are themselves shared \emph{externally} with others (Section~\ref{sec:takeaways}).\looseness=-1 

\textbf{Extracting copies in model outputs.}
Because some memorized works in the training data are extractable, LLMs can sometimes output copies of those works to users~\citep{cooper2023report, lee2023talkin, henderson2023fair, lee2024talkinshort, cyphert2024memlaw, sag2024fairness, charlesworth2025}. 
Works with higher extraction probabilities exhibit greater extraction risk; it takes fewer attempts to produce an output that is a verbatim copy of such a work (Section~\ref{sec:background} \& Appendix~\ref{app:sec:background:metrics}).
That output at generation time will be judged separately from the model itself, and is less likely to be a fair use~\citep{henderson2023fair,lee2023talkin}.\looseness=-1

Extraction is inherently probabilistic under non-deterministic decoding schemes (Section~\ref{sec:background}), so the relevant question is not only \emph{whether} a copy could ever appear for some prompt, but also \emph{how often} that copy appears in order to be ``meaningful'' ~\citep{cooper2024files}. 
When we measure suffix probabilities $p_\vz$ that are very close to $0\%$, we do not count those sequences as extracted; 
one could perhaps understand 
such generations as instances of ``a monkey at the typewriter''~\citep{borelmonkey,simpsons} (Section~\ref{sec:validity}). 
Very large $p_\vz$ (e.g., $\!>\!90\%$) indicates near-certain reproduction; 
this is clearly extraction (and therefore evidence of memorization).
However, most cases are not this clear, and it is not immediately obvious where to draw a line in the enormous range between these extremes.
To avoid false positives, we use a conservative threshold \(\tau_{\text{min}}\!=\!0.1\%\) for \emph{scientific} evidence of extraction (Sections~\ref{sec:background:pz-meaning}~\&~\ref{sec:validity}; Appendices~\ref{app:sec:background:metrics} \&~\ref{app:sec:validity}). 
But for \emph{copyright law and policy}, the practicability of extraction may require a different bar:
if reproducing a work just once would require thousands of prompts, the real-world risk may be relatively limited, even if  nonzero (Section~\ref{sec:takeaways}).\footnote{Complicating this point, our more recent work indicates that it is often possible to deterministically extract near-verbatim target suffixes that have $p_\vz \approx \tau_\text{min}$~\citep{cooper2026nv};
it takes one try, not thousands.}\looseness=-1 

Altogether, setting sensible copyright policy requires a specific understanding of both of these sets of issues. 
However, ML research does not tend to investigate these types of work-specific questions.
In the next section, we begin to address this gap by quantifying memorization of individual books.\looseness=-1 

\section{Measuring memorization of individual books}\label{sec:book-procedure}

To estimate how much a given model memorizes a given book, we introduce a sliding-window procedure that computes probabilistic extraction along the length of the entire book (Section~\ref{sec:book-procedure:method}). 
This procedure surfaces book-specific information about  memorization---information that is not readily apparent from average extraction rates over randomly sampled sequences, which prior work typically relies on to quantify memorization (Section~\ref{sec:book-procedure:averages}).
In the subsequent sections, we assess the validity of our approach (Section~\ref{sec:validity}) and demonstrate how it enables comparisons across models and books (Section~\ref{sec:book:compare}).\looseness=-1

\subsection{Sliding-window probabilistic extraction procedure}\label{sec:book-procedure:method}

\begin{figure*}[t!]
\begin{subfigure}{\linewidth}
\centering
\hspace*{0.13cm}
\includegraphics[width=\linewidth]{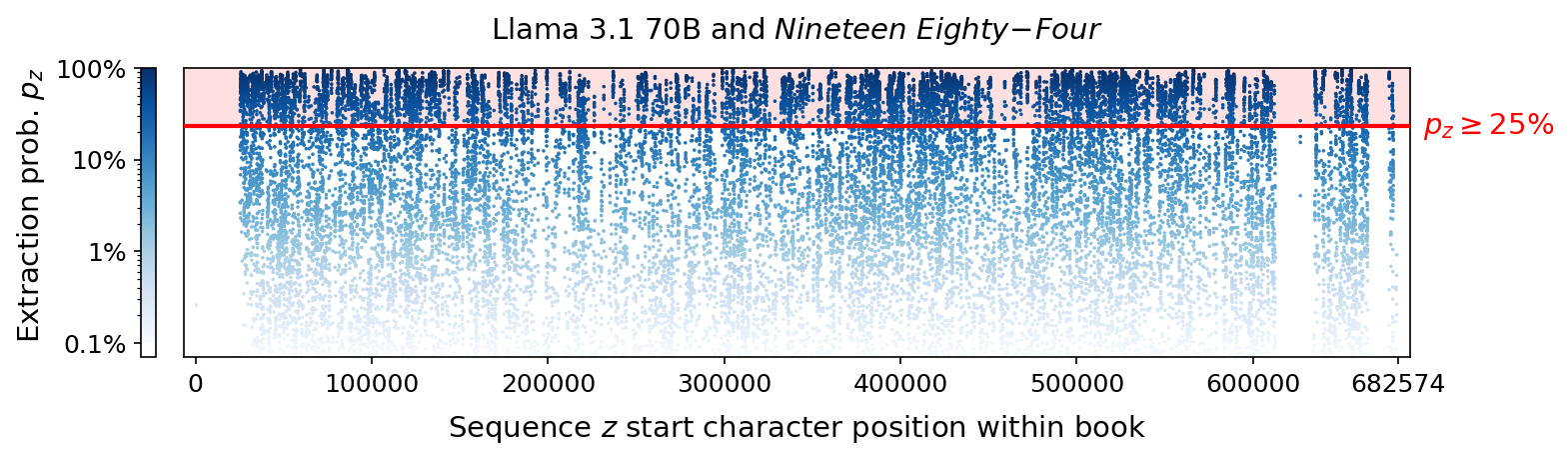}
\vspace{-.6cm}
\caption{Each dot is a sequence's extraction probability $p_\vz$ at its unique start position within the book}
\label{fig:1984:slide:main:scatter}
\end{subfigure}\\
\vspace{.5cm}
\begin{subfigure}{\linewidth}
\centering
\hspace*{-1cm}
\includegraphics[width=.95\linewidth]{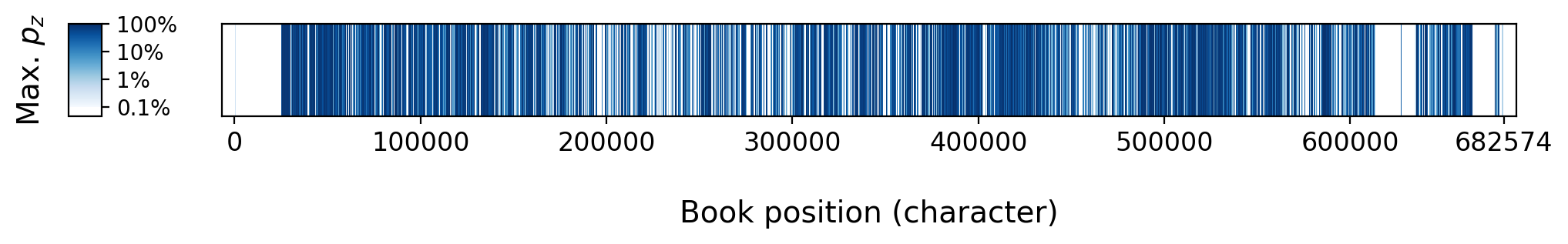}
\vspace{-.2cm}
\caption{For each character, the \emph{maximum} $p_\vz$ over all suffixes that span that character's position}
\label{fig:1984:slide:main:heatmap}
\end{subfigure}
\vspace{-.6cm}
\caption{\textbf{Visualizing our sliding-window probabilistic extraction procedure.} 
For each $100$-token window ($50$-token prefix $+$ $50$-token suffix) across \emph{1984}, 
we compute $p_\vz$ for \textsc{Llama 3.1 70B} with respect to top-$k$ decoding ($T\!=\!1$, $k\!=\!40$). 
(\textbf{a}) Scatterplot of \emph{all} extracted sequences $\vz$ shown by their unique start position in the book. 
The $100$-token sequences overlap significantly.
(\textbf{b})~Condensed heatmap view. 
At each character position, we plot the \emph{maximum} $p_\vz$ across all overlapping sequences whose suffixes cover that position. 
In both plots, colors encode $p_\vz$ on a $\log$ scale. 
We deem extraction successful if $p_\vz\!\geq\!\tau_\text{min}\!=\!0.1\%$, considering values below this  to not reflect extraction (i.e., rounding $p_\vz$ down to $0\%$).
See Appendix~\ref{app:sec:sliding-window}.\looseness=-1
}
\label{fig:1984:slide:main}
\vspace{-.4cm}
\end{figure*}

To identify regions of memorization, we develop the following ``panning for gold'' approach. 
For a given book, we start at the beginning of the text file in \texttt{Books3}. 
We take a chunk of text that is sufficiently long to contain $100$ tokens, then slide $10$ \emph{characters} forward in the book and repeat this process. 
We do this for the entire length of the book, which results in approximately $1$ sequence every $10$ characters, e.g., \emph{The Great Gatsby} has $270{,}870$ characters and so roughly $27{,}000$ sequences.
For most of our experiments, we follow the standard procedure of evaluating $100$-token sequences~\citep{carlini2023quantifying, hayes2025measuringmemorizationlanguagemodels}, which we divide into $50$-token prefix prompts and $50$-token suffixes that we attempt to extract (Equation~\ref{eq:pz}). 
Every character in the book (except for at the beginning and end) is present in multiple prefixes and suffixes.
This significant degree of overlap is deliberate. 
Since we typically do not know how open-weight models were trained, it is not exactly clear how we should break up books into token sequences when attempting extraction.
By testing various sequence starts and ends in the same area, we aim to surface as much total memorization as possible---to surface extraction ``hot-spots'' within a book (Appendix~\ref{app:sec:sliding-window:procedure}).\looseness=-1

We visualize the results of this procedure for \textsc{Llama 3.1 70B} (trained on \texttt{Books3}) on \emph{1984} (in \texttt{Books3}) using top-$k$ decoding ($T\!=\!1$, $k\!=\!40$).
Figure~\ref{fig:1984:slide:main:scatter} shows a scatterplot of every $100$-token sequence whose suffix we successfully extracted (i.e., $p_\vz\!\geq\!\tau_\text{min}\!=\!0.1\%$).
Each dot reflects a sequence's extraction probability $p_\vz$ according to its unique start character position within the book. 
White vertical gaps at a given location on the $x$-axis indicate that a sequence with that start-character position was not extracted. 
A continuous horizontal band of blue (associated with a probability $p_\vz$ on the $y$-axis) would indicate that we can extract the entire book---in pieces, with respect to $100$-token sequences---with that associated probability.\looseness=-1 

Figure~\ref{fig:1984:slide:main:heatmap} condenses the same results into a heatmap.
Unlike the scatterplot, this heatmap does not convey how many sequences are extractable at different probabilities in each area of the book.
Instead, at each character position, we plot the \emph{maximum} extraction probability $p_\vz$ across all overlapping sliding-window sequences $\vz$ whose suffixes cover that character position. 
Such heatmaps facilitate comparisons of memorization ``hot-spots'' across LLMs at the same book location (Section~\ref{sec:book:compare}). 
This heatmap shows high extraction probabilities exist throughout the entirety of \emph{1984}. 
With respect to probabilistic extraction (Equation~\ref{eq:pz}) measured with $50$-token prefixes and suffixes, the entirety of George Orwell's \emph{1984} is memorized inside \textsc{Llama 3.1 70B}'s weights.\footnote{The version of \emph{1984} in \texttt{Books3} has an editor's forward and appendix that are not memorized. 
The appendix is before selections of other writing by George Orwell, which are also highly memorized.
} 
On its own, this result does \emph{not} mean that the whole book can be extracted in one continuous segment at generation time. 
We return to reconstructing an entire book at generation time in Section~\ref{sec:book:seed}.\looseness=-1 

\subsection{Average extraction rates obscure book-specific memorization}\label{sec:book-procedure:averages}

\begin{figure*}[t]
\hspace*{-1.75cm}
\centering
\begin{minipage}[t]{0.54\textwidth}
\vspace{0pt}
\centering
\includegraphics[width=\linewidth]{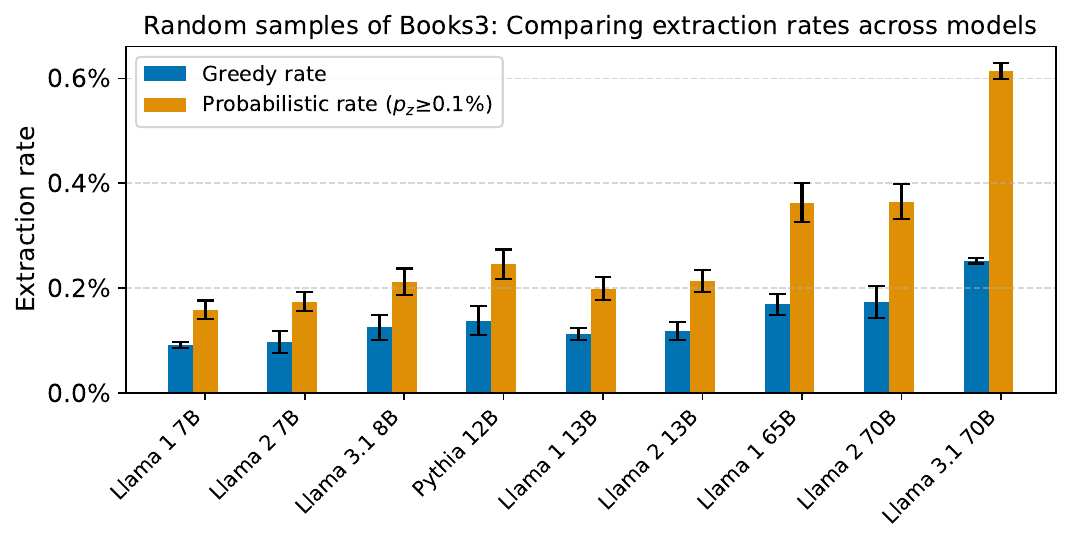}
\end{minipage}%
\hspace{.25cm}
\begin{minipage}[t]{0.45\textwidth}
\vspace{0pt}
\centering
\scriptsize
\begin{tabular}{lllll}
\toprule
\textbf{Model} & \textbf{Book} & \multicolumn{3}{c}{\textbf{\% of book extracted with $p_\vz\!\geq$}} \\
\cmidrule(lr){3-5}
 & & $\tau\!=\!1\%$ & $\tau\!=\!10\%$ & $\tau\!=\!50\%$ \\ 
\midrule
\multirow{2}{*}{\textsc{\shortstack[l]{Llama 3.1\\70B}}}  & \emph{Harry Potter}   & 90.89\% & 75.44\% & 43.26\% \\ 
                                                         & \emph{Sandman Slim}   & \;\;0.38\% & \;\;0.28\% & \;\;0.12\% \\ 
\midrule
\multirow{2}{*}{\textsc{\shortstack[l]{Llama 1\\ 65B}}}   & \emph{Harry Potter}   & 25.48\% & 15.00\% & \;\;4.40\%\\ 
                                                         & \emph{Sandman Slim}   & \;\;0.27\% & \;\;0.13\% & \;\;0.10\%\\ 
\midrule
\multirow{2}{*}{\shortstack[l]{\textsc{Pythia}\\ \textsc{12B}}} & \emph{Harry Potter} & \;\;0.40\% & \;\;0.10\% & \;\;0.08\% \\ 
                                                                & \emph{Sandman Slim} & \;\;0.34\% & \;\;0.32\% & \;\;0.26\% \\ 
\bottomrule
\end{tabular}

\end{minipage}
\vspace{-.25cm}
\caption{\textbf{Average extraction rates are low, but book-specific extraction varies widely.} 
(\textbf{left})~Comparing extraction rates (Equation \ref{eq:rate:main}) of random $100$-token sequences ($50$-token prefix $+$ $50$-token suffix) from \texttt{Books3} for different LLMs trained on \texttt{Books3}. 
Average extraction is low (Appendix~\ref{app:sec:rates}) for both greedy discoverable and probabilistic ($p_\vz\!\geq\!\tau_{\text{min}}\!=\!0.1\%$, top-$k$ with $T\!=\!1$ and $k\!=\!40$) extraction. 
(\textbf{right})~Extraction coverage (Equation~\ref{eq:cov}) for specific books reveals a more nuanced picture. 
For \emph{Harry Potter}~\citep{Harry_Potter_and_the_Sorcerer_s_Stone}  and \emph{Sandman Slim}~\citep{Sandman_Slim}, we show extraction coverage for $\tau \in \{1\%, 10\%, 50\%\}$ for $3$ models (Appendix~\ref{app:sec:sliding-window:percentage}). 
Nearly half of \emph{Harry Potter} can be extracted with respect to sequences that exhibit $p_\vz\!\geq\!\tau\!=\!50\%$.\looseness=-1}
\label{fig:rates}
\vspace{-.25cm}
\end{figure*}

Our sliding-window procedure provides a detailed, book-specific view of memorization that is distinct from prior work. 
While prior work shows that LLMs memorize certain amounts of the data they were trained on~\citep[e.g.,][]{carlini2021extracting, nasr2023scalable, hayes2025measuringmemorizationlanguagemodels, lee2022dedup, reid2024gemini, gemma2}, that work quantifies memorization through overall extraction rates---much like those in Figure~\ref{fig:rates} (left).
For some (part of a) training dataset, researchers draw (typically at random) a set $\sZ$ of sequences $\vz$ of a specified length (e.g., the first $100$ tokens).
For each sequence, they prompt with the prefix and count extraction as successful if the resulting generation matches the target suffix (Section~\ref{sec:background}). 
The extraction rate is computed as the number of attempted extractions that succeeded, relative to the total number of attempts:  
\begin{align}
\label{eq:rate:main}
\mathsf{extraction\_rate}(\sZ; s) = \frac{1}{|\sZ|}\sum_{\vz\in\sZ} \mathbf{1}[\,s(\vz)\,],
\end{align}
where $s(\vz)$ is an extraction success condition.  
As noted in Section~\ref{sec:background:pz},
for traditional discoverable extraction, the indicator in Equation~\ref{eq:rate:main} evaluates to $1$ when the greedy completion exactly matches the target suffix. 
In probabilistic extraction, the indicator evaluates to $1$ if the target suffix's probability $p_\vz$ exceeds a chosen threshold $\tau_{\text{min}}$ (Equation~\ref{eq:probsuccess:main}, Appendix~\ref{app:sec:rates}).

These reported averages are generally small, just as we observe in Figure~\ref{fig:rates} (left) for $9$ models trained on \texttt{Books3}. 
Such small rates often form the basis of defendants in copyright infringement suits calling memorization a rare ``bug''~\citep{cooper2024files,openairesponse}.
However, while low extraction rates signal that models likely do not memorize \emph{most} text in \texttt{Books3},   \emph{specific} works can still be highly memorized. 
On average, \textsc{Llama 3.1 70B} may have only memorized about $0.6\%$ of sequences in \texttt{Books3}, but concealed within that small number is that it memorized effectively \emph{all} of \emph{1984} (Figure~\ref{fig:1984:slide:main}). 

To make this observation more precise, we quantify how much of a whole book is extractable with \newterm{extraction coverage}: 
the fraction of total characters in a book that lie within at least one suffix whose extraction probability $p_\vz$ exceeds a threshold $\tau$ (Appendix~\ref{app:sec:sliding-window:percentage}).
We choose $\tau \in [\tau_{\text{min}}, 1]$ because $\tau_{\text{min}}$ is the minimum probability we consider to reflect extraction success (Sections~\ref{sec:background} \&~\ref{sec:validity}). 
More formally, let $\mathsf{span}(\vz)$ be the set of character indices spanned by a given $50$-token suffix of sequence $\vz$. 
Then\looseness=-1
\begin{align}
\label{eq:cov}
\mathsf{coverage}(\tau) = \frac{1}{L}\,\Bigl|\bigcup_{\vz:\, p_\vz \ge \tau}\, \mathsf{span}(\vz)\Bigr|, \; \text{where $L$ is the character-length of the whole book.}
\end{align}

\begin{figure*}[t]
\centering
\begin{subfigure}{0.48\linewidth}
\includegraphics[width=\linewidth]{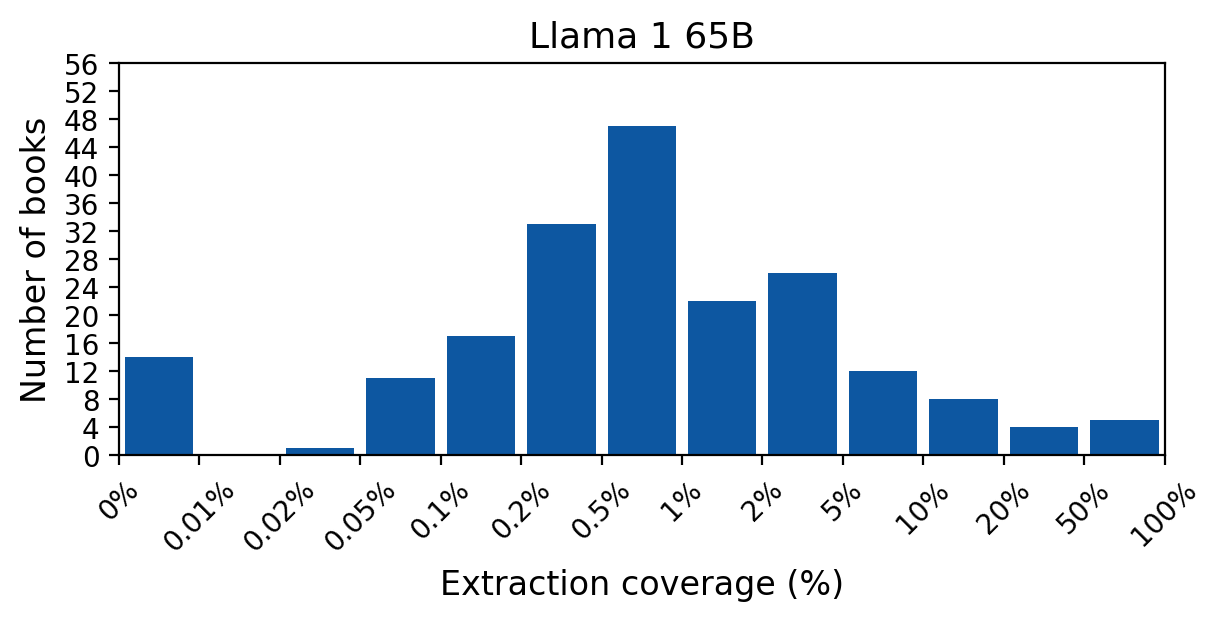}    
\end{subfigure}
\hspace{.25cm}
\begin{subfigure}{0.48\linewidth}
\includegraphics[width=\linewidth]{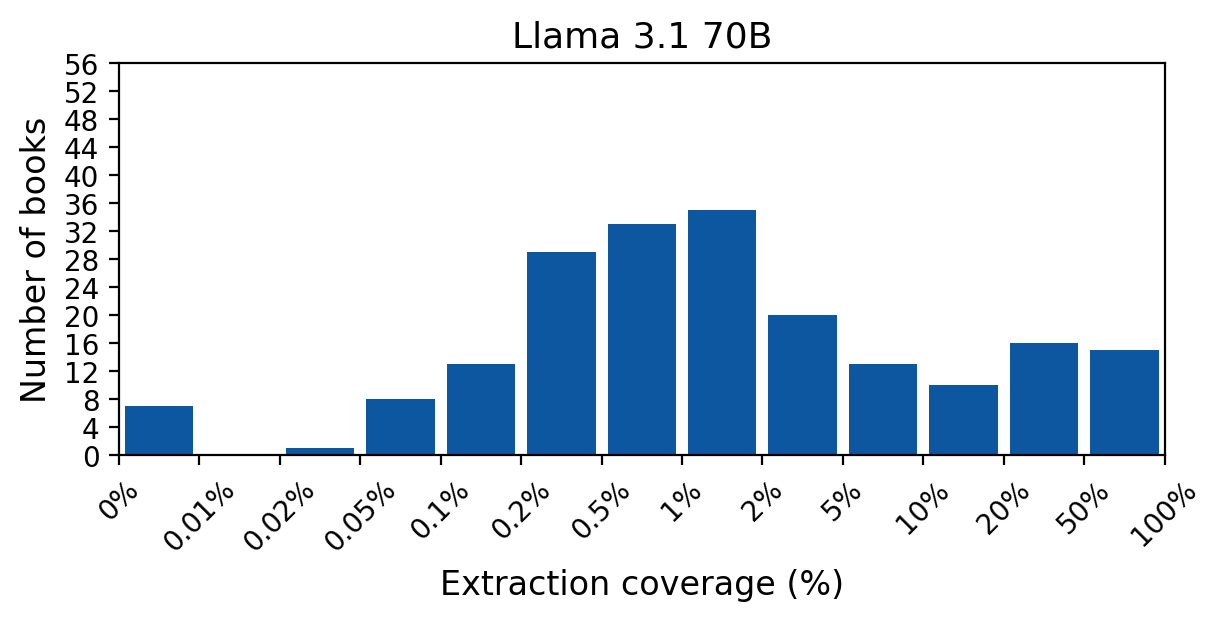}    
\end{subfigure} 
\caption{\textbf{LLM-specific distributions over extraction coverage.}
Extraction coverage (Equation~\ref{eq:cov}) differs across LLMs for the $200$ books we evaluate, illustrated for \textsc{Llama 1 65B} and \textsc{Llama 3.1 70B}.  
Results use $100$-token sequences ($50$-token prefix $+$ $50$-token suffix), top-$k$ decoding ($T\!=\!1$, $k\!=\!40$), and $\tau_\text{min}\!=\!0.1\%$ as the coverage threshold $\tau$.}
\label{fig:coverage:main}
\vspace{-.25cm}
\end{figure*}

For \textsc{Llama 3.1 70B} and \emph{1984}, Figure~\ref{fig:1984:slide:main:scatter} highlights in red the sequences used to compute Equation~\ref{eq:cov} for $p_\vz\!\geq\!\tau\!=\!25\%$. 
These sequences cover $54.39\%$ of \emph{1984}: 
for each, reproducing the verbatim $50$-token suffix would require (in expectation) only $4$ prompts with the corresponding $50$-token prefix. 
Most books and LLMs we test do not have this degree of extraction coverage.
Figure~\ref{fig:coverage:main} shows histograms over extraction coverage ($\tau=\tau_\text{min}=0.1\%$) for \textsc{Llama 1 65B} and \textsc{Llama 3.1 70B}. 
For both models, most books exhibit $<\!2\%$ coverage; 
however, many books have $>\!10\%$ coverage. 
Overall, 
\textsc{Llama 3.1 70B} exhibits higher coverage on more of the $200$ books we evaluate than any other LLM we test (Appendix~\ref{app:sec:sliding-window:percentage}).

For a more detailed comparison of book-specific coverage, in Figure~\ref{fig:rates} (right)  we show extraction coverage at various choices of $\tau$ for \emph{Harry Potter and the Sorcerer's Stone}~\citep{Harry_Potter_and_the_Sorcerer_s_Stone} and \emph{Sandman Slim}~\citep{Sandman_Slim}---one of the books by plaintiff Richard Kadrey in \emph{Kadrey et al. v. Meta, Inc.}~\citep{kadreyamendedconsolidated}. 
\emph{Sandman Slim} is hardly memorized at all. 
Extraction coverage for this book is less than $0.4\%$ for all LLMs and thresholds $\tau$.\footnote{Prior studies (and Figure~\ref{fig:rates}) show that larger LLMs exhibit higher verbatim extraction rates~\citep{carlini2023quantifying}.
This pattern does not cleanly generalize to specific works. 
At $\tau\!=\!50\%$, 
\textsc{Llama 3.1 70B} exhibits $0.12\%$ extraction coverage of \emph{Sandman Slim}, while the smaller \textsc{Pythia 12B} exhibits $0.26\%$ coverage.}
In contrast, \emph{Harry Potter}~\citep{Harry_Potter_and_the_Sorcerer_s_Stone} is highly memorized by \textsc{Llama} models  (Section~\ref{sec:book:compare}). 
For \textsc{Llama 3.1 70B}, $43.26\%$ of the book is extractable with respect to sequences exhibiting $p_\vz\!\geq\!\tau\!=\!50\%$.\footnote{
    The version we evaluate from \texttt{Book3} includes the first chapter of the second \emph{Harry Potter} book at the end. 
} 
For lower $\tau$, there is higher coverage; 
$90.89\%$ of the book is covered when $\tau\!=\!1\%$. 
While lower thresholds correspond to extracting text less reliably---i.e., requiring more prompts to the model---for an LLM, all three of these settings  reflect enormous probabilities (Section~\ref{sec:background:pz}).\looseness=-1 

\vspace{-.2cm}
\section{Validating our extraction procedure}\label{sec:validity}

Above, we emphasized that the minimum probability we count as successful extraction,  $\tau_\text{min}=0.1\%$ (Equation~\ref{eq:probsuccess:main}), is deliberately conservative.
But how can we be confident this is the case: 
that when we measure a successful extraction, we are capturing a true instance of memorization, rather than happenstance generation of a verbatim suffix (a ``monkey at the typewriter'' event, Section~\ref{sec:copyright})? 
We address this question through a set of experiments designed to validate our approach. 
We apply our sliding-window procedure with top-$k$ decoding ($T\!=\!1$, $k\!=\!40$) and $50$-token suffixes (Section~\ref{sec:book-procedure:method}), 
but we vary prefix length (up to $800$ tokens) to observe how conditioning on additional context affects our measurements.

First, we run these settings on books from \texttt{Books3} and \textsc{Llama 3.1} models, which were trained on \texttt{Books3} (Section~\ref{sec:validity:baseline}). 
As expected~\citep{carlini2023quantifying}, longer prefixes surface more memorized training data, but increasing the prefix length does not make every suffix register as extractable (Section~\ref{sec:validity:baseline}). 
Second, we run the same procedure on \emph{non}-training data, which tests whether our method spuriously registers false positives on data where memorization is impossible (Section~\ref{sec:validity:controls}). 
Unlike the experiments on training data, regardless of prefix length,  suffix probabilities remain below $\tau_\text{min}$ (with interesting exceptions\footnote{These sequences are boilerplate text (copyright notices, publisher addresses, etc.) and popular quotes---both of which are duplicated across many sources aside from \texttt{Books3}. 
    (See Appendices~\ref{app:sec:validity:membership}  \&~\ref{app:sec:book-level-discussion}.) 
}), 
demonstrating our approach does not mistakenly register non-training data as extractable.
Together, these experiments show that, 
when our measurements indicate extraction, we can be confident they are a valid reflection of memorization.\footnote{Early work on generating copyrighted text with LLMs often did not include such experiments, e.g., \citet{karamolegkou2023copyrightviolationslargelanguage}.
    Those works also tended to run experiments for which ground-truth information about the training data is unknown. 
    Without negative controls to calibrate false positives, it is difficult to interpret reported overlap-based results as valid evidence of extraction (and thus memorization)~\citep{cooper2026firstprinciples}.
}
We briefly discuss the implications of these results (Section~\ref{sec:validity:implications}), which inform the experiments we run in the following section.
Additional results and discussion can be found in Appendix~\ref{app:sec:validity}.

\begin{figure*}[t!]
\centering
\hspace*{-1cm}
\includegraphics[width=1.1\linewidth]{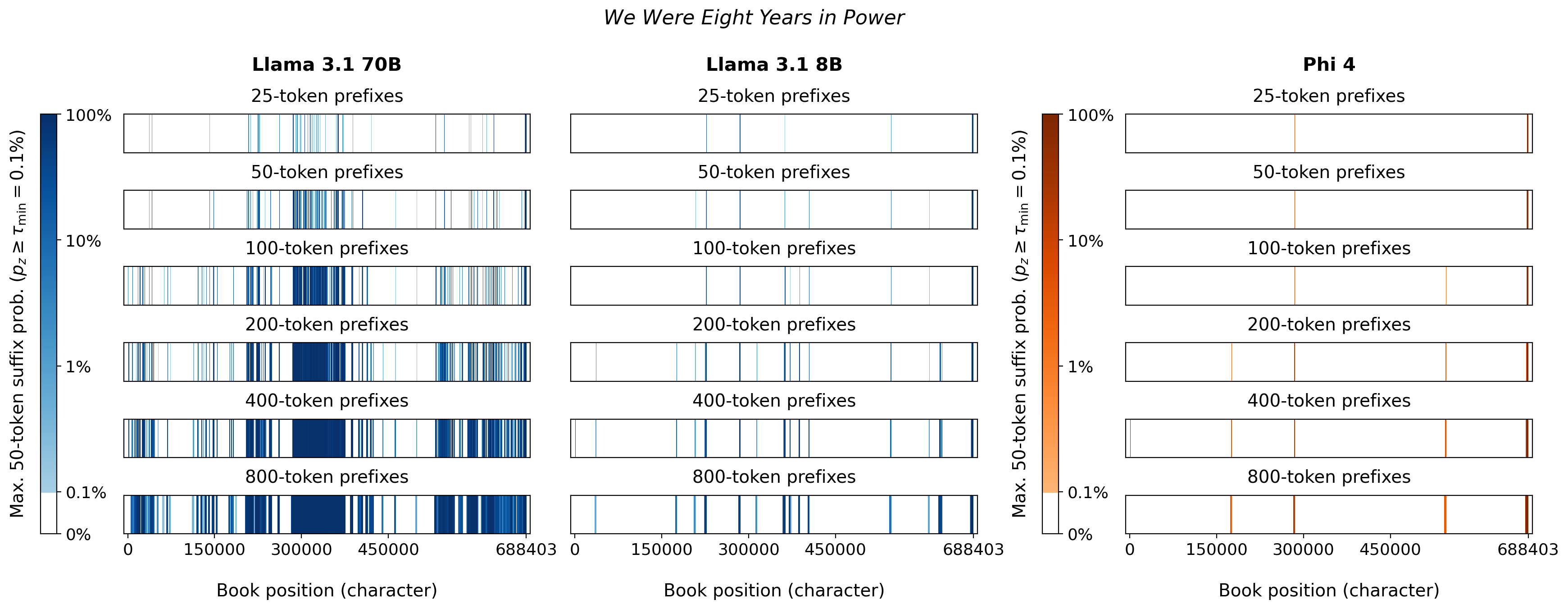}
\caption{\textbf{Illustrating validation of our extraction measurements.}
The sliding-window procedure for \textsc{Llama 3.1 70B}, \textsc{Llama 3.1 8B}, and \textsc{Phi 4} on \emph{We Were Eight Years in Power}~\citep{We_Were_Eight_Years_in_Power} (in \texttt{Books3}), using prefix lengths in $\{25, 50, 100, 200, 400, 800\}$ and $50$-token suffixes with top-$k$ decoding ($T\!=\!1$, $k\!=\!40$). 
The book is in \textsc{Llama}'s training data, but not \textsc{Phi 4}'s. 
Longer prefixes reveal more memorized sequences, though not all sequences in the book are extractable as prefix length increases. 
For \textsc{Phi 4}, the only extractable suffixes are widely duplicated quotes that likely appeared in its training data from other sources. 
These results support that, when our procedure identifies extraction, it is detecting true instances of memorization rather than chance generations.\looseness=-1
}
\label{fig:wewere:main}
\vspace{-0.4cm}
\end{figure*}

\subsection{Longer prefixes surface more memorized training data}\label{sec:validity:baseline}

When a sequence is known to be in the training data, memorization is the most plausible explanation for high-probability generation of its (sufficiently long) verbatim suffix given the corresponding prefix (Section~\ref{sec:background}). 
So, as a baseline, we apply the sliding-window procedure to this setting (Figure~\ref{fig:wewere:main}): 
we attempt to extract sequences in \emph{We Were Eight Years in Power}~\citep{We_Were_Eight_Years_in_Power} (a book in \texttt{Books3}) from \textsc{Llama 3.1 70B} and \textsc{Llama 3.1 8B} (LLMs trained on \texttt{Books3}). 
Longer prefixes reveal additional memorized sequences. 
This is what \citet{carlini2023quantifying} call the \newterm{discoverability phenomenon}:
providing additional context (a longer ground-truth prefix) can raise the conditional probability of the target suffix enough that it becomes extractable ($p_\vz \!\geq\! \tau_\text{min}$). 
Nevertheless, for both LLMs many sequences remain unextractable: 
regardless of prefix length, their suffix probabilities $p_\vz$ never surpass $\tau_{\text{min}}$. 
For instance, \textsc{Llama 3.1 70B} memorizes most of the book, yet consistent sections remain unextractable across all prefix lengths (see white gaps in Figure~\ref{fig:wewere:main}, such as from ${\sim}75$K--$120$K characters). 
The same is even clearer for \textsc{Llama 3.1 8B}: 
being smaller, it memorizes significantly less~\citep{carlini2023quantifying, hayes2025measuringmemorizationlanguagemodels, carlini2021extracting}. 
Longer prefixes do surface more memorized training data, but most of the book remains unextractable.
(Our results suggest that longer prefixes beyond $800$ tokens would enable extraction of additional sequences, but the gains diminish; not every sequence is extractable. 
See Figure~\ref{fig:prefix-coverage} and Appendix~\ref{app:sec:validity:baseline}.)\looseness=-1 

These results are consistent with prior work: 
high-quality models do not---indeed cannot~\cite{lee2022dedup, carlini2021extracting, mahdavi2024memorizationcapacitymultiheadattention, collins2017capacitytrainabilityrecurrentneural}---memorize all of the training data in their massive training datasets. 
Even as prefix length increases, our measurements do not suddenly capture extraction success for everything; 
they capture success for sequences that have been memorized. 
We observe the same for \textsc{Llama 3.1 70B} on other books in its training data (Figure~\ref{fig:prefix-coverage}). 
Some appear effectively not memorized at all, with only limited increases in extraction success at longer prefix lengths (Appendix~\ref{app:sec:validity:baseline}).

\subsection{Measurements on non-training data do not mistakenly register extraction}\label{sec:validity:controls}

Using the above results as a basis for comparison, we next test our method in settings where memorization is impossible. 
This is the role of \newterm{negative controls}:
experiments in conditions where the outcome of interest should not occur. 
If (false) positives appear in such a setting, they indicate that the measurement procedure is capturing spurious signal.
In our context, running the sliding-window procedure on \emph{non}-training data provides this sanity check.  
In general (with important exceptions noted below), observing suffix probabilities above $\tau_{\text{min}}$ would indicate false positives---i.e., would count successful extractions where none should exist.
The absence of such cases supports that the extractions we do count are valid.\looseness=-1

\begin{figure}[t]
    \centering
    \includegraphics[width=0.9\linewidth]{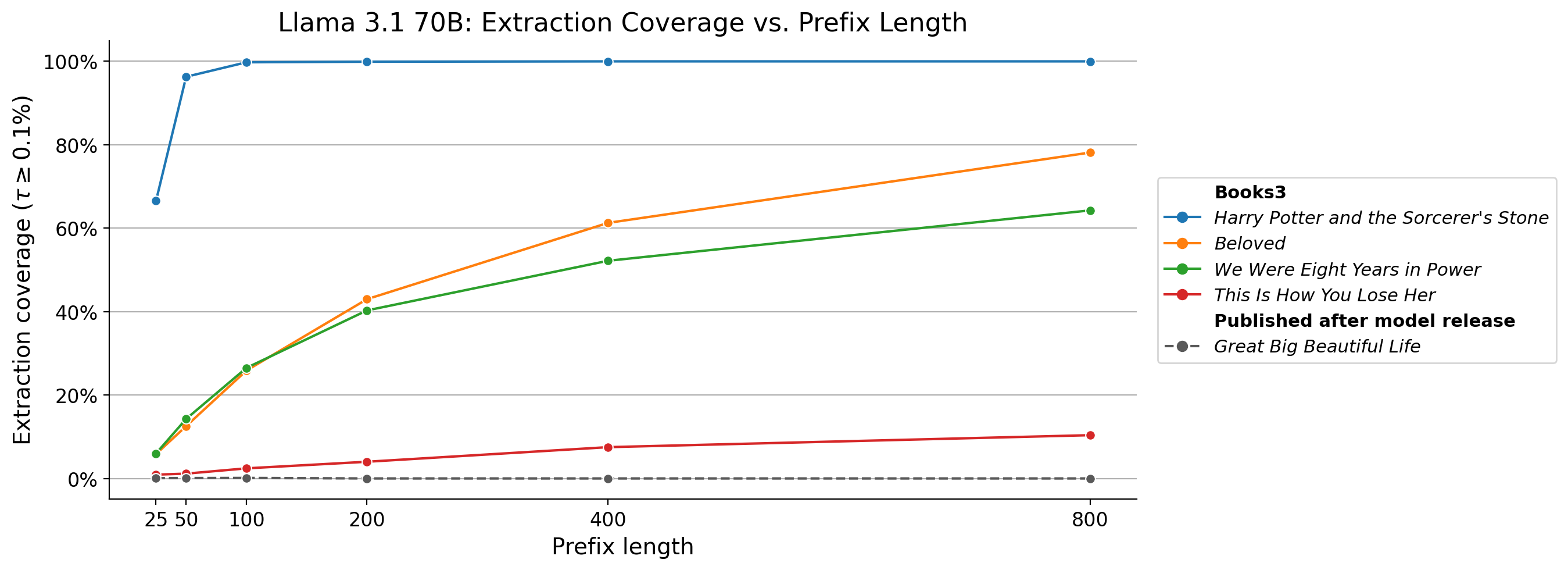}
    \caption{\textbf{Extraction coverage by prefix length.}
    Plotting extraction coverage for $\tau\!=\!\tau_\text{min}\!=\!0.1\%$ (Equation~\ref{eq:cov}) for $5$ books we test with varying prefix lengths and \textsc{Llama 3.1 70B}. 
    $4$ of these books are in \texttt{Books3} and thus in \textsc{Llama}'s training data.
    The remaining book, \emph{Great Big Beautiful Life}~\citep{greatbig}, is used in our negative controls on non-training data; 
    extraction coverage is $0\%$ (aside from the copyright notice) regardless of prefix length. 
    As we increase prefix length, extraction coverage for each book in the training data begins to level off.
    While \emph{Harry Potter}~\citep{Harry_Potter_and_the_Sorcerer_s_Stone} is entirely memorized, it does not appear to be the case that \emph{Beloved}~\citep{Beloved}, \emph{We Were Eight Years in Power}~\citep{We_Were_Eight_Years_in_Power}, or \emph{This Is How You Lose Her}~\citep{This_Is_How_You_Lose_Her} are entirely extractable with respect to $50$-token suffixes and even longer prefixes.\looseness=-1}
    \vspace{-.4cm}
    \label{fig:prefix-coverage}
\end{figure}

We run two sets of negative controls.
First, we apply our procedure to in-copyright \texttt{Books3} books with \textsc{Phi 4}~\citep{phi4}---a model that was \emph{not} trained on whole copyrighted books. 
Second, we apply the sliding-window procedure to \textsc{Llama} models on $4$ books published in 2025---after the July 2024 cutoff for \textsc{Llama 3.1} training. 
In heatmaps, we distinguish these conditions by color: 
\textcolor{seabornbluemid}{blue} for models trained on \texttt{Books3} on \texttt{Books3} text and
\textcolor{seabornorangemid}{orange} for \textsc{Phi 4} on \texttt{Books3} text. 

\custompar{Negative control 1: An LLM not trained on copyrighted books $\times$ copyrighted books in \texttt{Books3}} 
We run the same experiment as in Section~\ref{sec:validity:baseline} on \textsc{Phi 4}, which was not trained on whole in-copyright books like \emph{We Were Eight Years in Power}~\citep{We_Were_Eight_Years_in_Power}.
Unlike with \textsc{Llama 3.1} models, we do not observe extraction success for unique text from this book:
increasing the context length does not 
push the conditional probability of non-training-data suffixes above $\tau_{\text{min}}$. 
We do observe extraction signal, but it only concentrates in four small areas that do not appear to be false positives. 
Instead, they plausibly reflect memorized text: 
the suffixes our procedure counts as extracted are famous quotes that are widely duplicated on the Internet, and thus very likely to have been included in \textsc{Phi 4}'s training data via sources other than this copyrighted book~\citep{phi4}. 
For example, at ${\sim}280$K characters, we extract passages from the King James Bible~\citep{deuteronomy}; 
at ${\sim}546$K characters, we extract a quote from Barack Obama's 2004 Democratic National Convention speech~\citep{obama2004}. 
Notably, \textsc{Llama 3.1} models also exhibit high extraction probabilities in these same regions. 
While they were trained on this book, it is plausible that duplicates from other sources contributed to the memorization and extractability of these sequences~\citep{lee2022dedup}.\looseness=-1

\custompar{Negative control 2: LLMs trained on \texttt{Books3} $\times$ books published after their training} 
We apply the same procedure to \textsc{Llama} models on books published in 2025---after the July 2024 cutoff for \textsc{Llama 3.1} training. 
In Figure~\ref{fig:prefix-coverage}, 
we show a subset of the results for \emph{Great Big Beautiful Life}~\citep{greatbig}. 
Apart from the (highly duplicated, boilerplate) 
copyright notice and popular quotes, there are no suffix probabilities that we would count as successful extraction.
Not only does $p_\vz$ never surpass $\tau_\text{min}\!=\!0.1\%\!=\!10^{-1}\%$, there are no sequences with $p_\vz \geq 10^{-13}\%$. 
For $200$-, $400$-, and $800$-token prefixes, exactly $1$ sequence (out of ${\sim}62{,}000$) has $p_\vz \geq 10^{-14}\%$. 
This reflects chance generation of non-training data---effectively a ``monkey at the typewriter'' event. 
But importantly, this is not a sequence that our method counts as extractable: 
the suffix probability is $13$ orders of magnitude smaller than our setting for $\tau_\text{min}$.
With a single attempt, the odds of generating the verbatim suffix are $\nicefrac{1}{p_\vz}\!\approx\!\nicefrac{1}{10^{-14}\%}$, or $1$ in $10$ quadrillion (Appendix~\ref{app:sec:validity:cutoff}).
To give a sense of just how small this number is, the chance of winning the Powerball lottery jackpot in a single draw is about $1$ in $292$ million---roughly $34$ million times more likely.\looseness=-1


\subsection{Implications for measuring probabilistic extraction on  books}\label{sec:validity:implications}

Three important observations follow from these results and related experiments in Appendix~\ref{app:sec:validity:controls};
we highlight them here and defer additional discussion to Appendix~\ref{app:sec:validity}.
First, the negative controls provide strong evidence that our sliding-window probabilistic extraction procedure is valid. 
Second, they indicate that our choice of $\tau_\text{min}$ for counting extraction success is indeed conservative. 
Given the very low probabilities we observe for generating non-training data, we could reasonably set $\tau_\text{min}$ below $0.1\%$. 
As such, our results almost certainly under-count memorization, and should be read as a lower bound. 
This is deliberate: 
we prefer to under-count memorization (i.e., set a higher $\tau_\text{min}$) rather than risk over-counting it~\citep{carlini2022membershipinferenceattacksprinciples, hayes2025strongmia}. 
To lower $\tau_\text{min}$ and remain confident about validity, we would need to run a much larger set of negative controls. 
Part of the reason is that, as others have noted~\cite{cooper2024files}, LLMs are not actually like monkeys randomly outputting tokens. 
They have learned to generate structured, grammatical sentences, so the distribution of plausible outputs is far smaller than random chance would suggest (Appendix~\ref{app:sec:validity:monkey}). 
Exploring what this implies for setting $\tau_\text{min}$ more precisely is an important question that we explore in follow-on work~\citep{cooper2026firstprinciples}. 

Third, having validated our procedure, we are justified in extending our experiments beyond LLMs like \textsc{Llama}, where we know that \texttt{Books3} was in the training data. 
If suffix probabilities surpass $\tau_\text{min}$, it is reasonable to interpret these as true extractions and thus evidence of memorization (Appendix~\ref{app:sec:validity:membership}). 
This remains the case regardless of whether the LLM was literally trained on \texttt{Books3}, or whether the memorized training data originate from other overlapping sources---for example, highly duplicated famous quotations, as with \textsc{Phi 4} in Figure~\ref{fig:wewere:main} (Appendix~\ref{app:sec:validity:membership}). 
On this basis, in the following section, we expand our analysis to include $3$ additional  models for which it is unknown if \texttt{Books3} was included in their training data.\looseness=-1

\vspace{-.2cm}
\section{Comparing memorization across books and models}\label{sec:book:compare}

We now present the main results of applying the sliding-window procedure. 
We describe the experimental setup (Section~\ref{sec:book:compare:setup}) and highlight the types of comparisons our approach facilitates across both books and models (Section~\ref{sec:book:compare:results}). 

\subsection{Experimental setup}\label{sec:book:compare:setup}

We use the same setup as in Section~\ref{sec:book-procedure:method}:
the sliding-window procedure with top-$k$ decoding ($T\!=\!1$, $k\!=\!40$) for 
$50$-token prefixes and $50$-token suffixes.
We run the procedure on $14$ models, which we organize into three groups: 
({a}) LLMs trained on \texttt{Books3} (\textsc{Llama}~\citep{touvron2023llamaopenefficientfoundation, llama2, llama3} 
and \textsc{Pythia}~\citep{biderman2023pythia}, 
\textcolor{seabornbluemid}{blue}); 
({b}) LLMs that were very likely trained on (at least parts of) books also contained in \texttt{Books3}, based on measurements of $p_\vz\!\geq\!\tau_\text{min}\!=\!0.1\%$ for unique content from those books (\textsc{DeepSeek v1}, \textsc{Qwen 2.5}, and \textsc{Gemma 2}, \textcolor{seabornredmid}{red}); 
and ({c}) an LLM not trained on whole copyrighted books (\textsc{Phi 4}, \textcolor{seabornorangemid}{orange}).
The code for our experiments can be found \href{https://github.com/pasta41/probabilistic-extraction-toolkit}{here}.\looseness=-1 

Of the ${\sim}200{,}000$ books in \texttt{Books3}, we test $200$ books (roughly $0.1\%$)---the same books for which we analyze extraction coverage in Figure~\ref{fig:coverage:main}.
We randomly sample $100$ books; the other $100$ include works associated with 
plaintiffs in the amended class action complaint of \emph{Kadrey et al. v. Meta, Inc.}~\citep[pp. 4-5]{kadreyamendedconsolidated},\footnote{Post-dating our initial work, that suit was  decided in favor of the defendants, Meta~\citep{kadreyjudgment}.} 
as well as a range of generally popular~\citep[e.g.,][]{Harry_Potter_and_the_Sorcerer_s_Stone, The_Hobbit, The_Myth_of_Sisyphus, Catch-22} and more obscure~\citep[e.g.,][]{The_Future_of_the_Internet_and_How_to_Stop_It, Dante_and_the_Origins_of_Italian_Literary_Culture} books.
We suspected that some popular books would exhibit higher degrees of memorization (e.g., due to duplicated text from other sources).  
This is why we did not only select books at random;
for part of our sample, we deliberately chose titles intended to capture variation across \texttt{Books3}.
Further details on these decisions are in Appendix~\ref{app:sec:sliding-window:setup}.

In all, we ran over $3000$ experiments, amounting to $394$ GPU days. 
Nevertheless, our results should not be read as a complete or generalized account of memorization across \texttt{Books3}.
We provide complete per-book results in Appendix~\ref{app:sec:sliding-window:results}, and also publish a \href{https://books-memorization.github.io}{website} version.

\subsection{Key patterns in our results}\label{sec:book:compare:results}

We focus here on only a few representative books and on the first two categories of models---(a) models known to be trained on \texttt{Books3} and ({b}) models where it is not disclosed whether \texttt{Books3} was included in the training data. 
Extraction probabilities $p_\vz$ can be compared across LLMs, allowing us to evaluate which models are more or less at risk of outputting verbatim $50$-token suffixes.
We perform these comparisons using heatmaps, as they visualize how memorization varies at specific book locations.\footnote{Sequences $\vz$ for which we compute $p_\vz$ may differ, as LLMs have different token vocabularies and tokenizers that which segment text differently. 
    Character locations are fixed across all experiments with the same book.\looseness=-1
}
This enables drawing conclusions about degrees of memorization: 
how memorization varies for a specific book across models and across books for a given model.
We omit heatmaps for \textsc{Llama 3 70B}, as they resemble those for \textsc{Llama 3.1 70B}.\looseness=-1 

\begin{figure*}[t]
\centering
    \begin{subfigure}{\textwidth}
        \includegraphics[width=\linewidth]{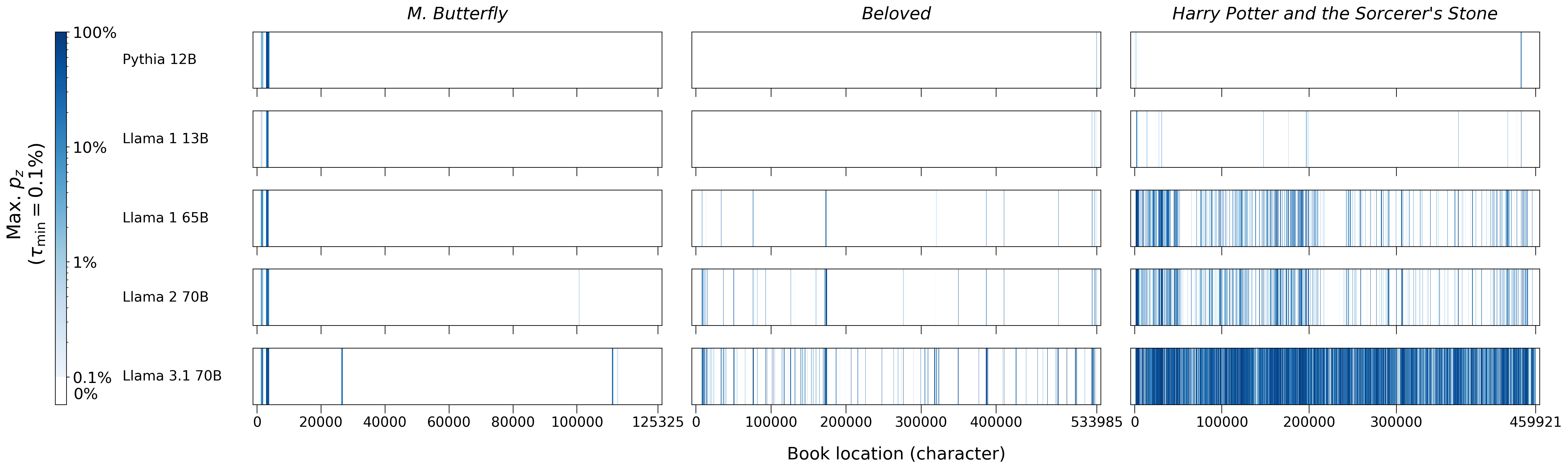}
        \caption{LLMs known to be trained on \texttt{Books3}}
        \label{fig:comparisons:heatmaps:in-books3}
    \end{subfigure}
    \begin{subfigure}{\textwidth}
        \includegraphics[width=\linewidth]{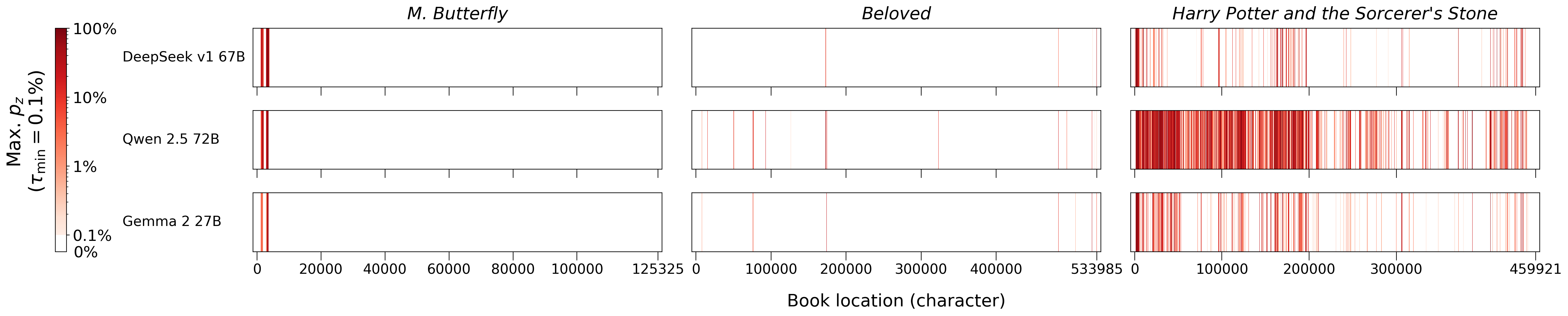}
        \caption{LLMs we conclude were trained on (at least parts of) some books that are also contained in \texttt{Books3}}
        \label{fig:comparisons:heatmaps:unsure-books3}
    \end{subfigure}
    \caption{\textbf{Memorization varies widely across books and models (non-random books).}
    We apply the sliding-window procedure with top-$k$ decoding ($T\!=\!1$, $k\!=\!40$) and $100$-token sequences ($50$-token prefixes $+$ $50$-token suffixes) for LLMs  
    (\textbf{a}) trained on \texttt{Books3} (\textcolor{seabornbluemid}{blue}) and (\textbf{b}) where it is unknown if \texttt{Books3} was included in the training data (\textcolor{seabornredmid}{red}).
    With respect to these settings, 
    \textsc{Llama 3.1 70B} memorizes effectively all of \emph{Harry Potter and the Sorcerer's Stone}~\citep{Harry_Potter_and_the_Sorcerer_s_Stone}: extraction coverage is ${\approx}96.3\%$.\looseness=-1}
    \label{fig:comparisons:heatmaps}
    \vspace{-.5cm}
\end{figure*}

\begin{figure*}[t]
\centering
    \begin{subfigure}{\textwidth}
        \includegraphics[width=\linewidth]{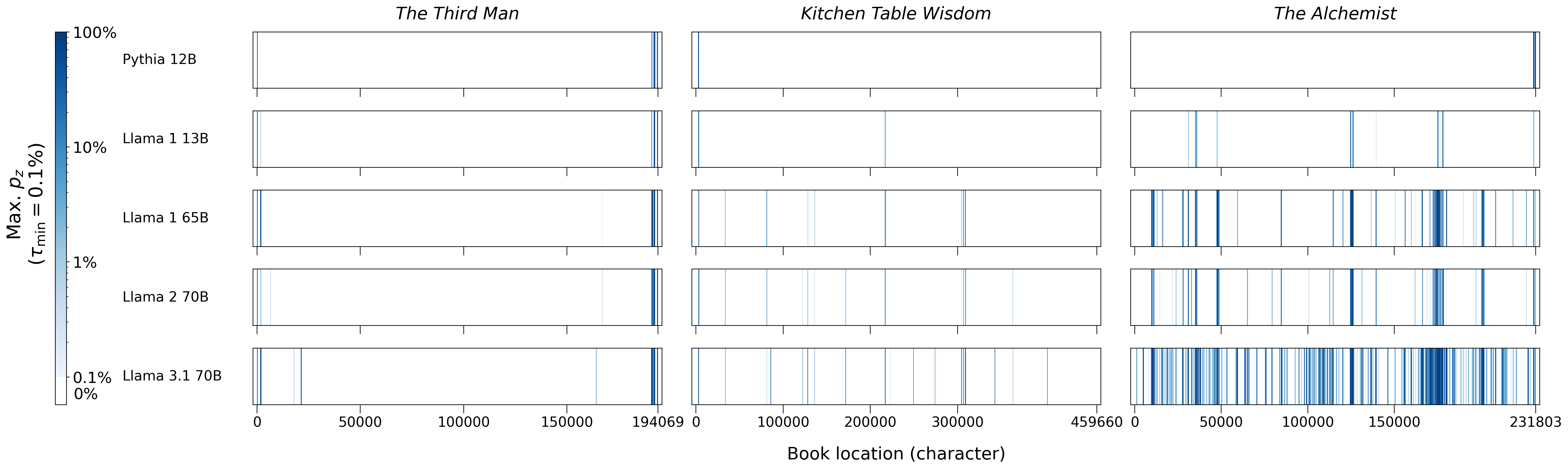}
        \caption{LLMs known to be trained on \texttt{Books3}}
        \label{fig:comparisons:heatmaps:in-books3-random}
    \end{subfigure}
    \begin{subfigure}{\textwidth}
        \includegraphics[width=\linewidth]{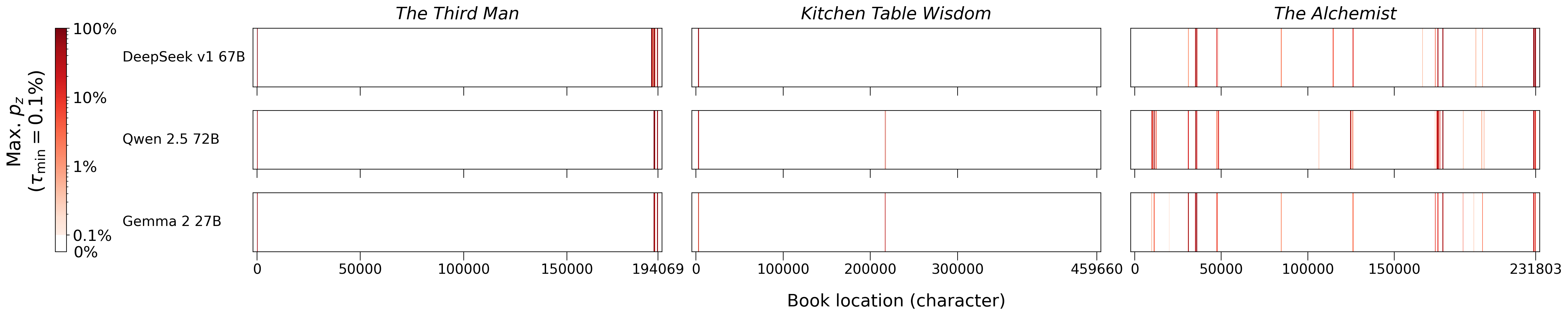}
        \caption{LLMs we conclude were trained on (at least parts of) some books that are also contained in \texttt{Books3}}
        \label{fig:comparisons:heatmaps:unsure-books3-random}
    \end{subfigure}
    \caption{\textbf{Memorization varies widely across books and models (random books).}
    We apply the sliding-window procedure with top-$k$ decoding ($T\!=\!1$, $k\!=\!40$) and $100$-token sequences ($50$-token prefixes $+$ $50$-token suffixes) for LLMs  
    (\textbf{a}) trained on \texttt{Books3} (\textcolor{seabornbluemid}{blue}) and (\textbf{b}) where it is unknown if \texttt{Books3} was included in the training data (\textcolor{seabornredmid}{red}).
    With respect to these settings, 
    \textsc{Llama 3.1 70B} memorizes a substantial portion of \emph{The Alchemist}~\citep{The_Alchemist}: extraction coverage is ${\approx}49.1\%$.\looseness=-1}
    \label{fig:comparisons:heatmaps-random}
    \vspace{-.5cm}
\end{figure*}

We show a selection of our results on $200$ books and $14$ models in Figures~\ref{fig:comparisons:heatmaps} and~\ref{fig:comparisons:heatmaps-random}. 
The $3$ books in Figure~\ref{fig:comparisons:heatmaps} are from the $100$ books we manually selected; 
the $3$ in Figure~\ref{fig:comparisons:heatmaps-random} were sampled randomly. 
Overall, we observe three high-level patterns of memorization in our results;
an example of each pattern is provided in each figure.  
First, there are books like \emph{M. Butterfly}~\citep{M_Butterfly} (Figure~\ref{fig:comparisons:heatmaps}) and \emph{The Third Man}~\citep{The_Third_Man} (Figure~\ref{fig:comparisons:heatmaps-random}), for which we observe effectively no memorization for any LLM with $50$-token prefixes.
Aside from the copyright notice, we extract little (if any) text from these books across LLMs of different sizes trained by different organizations.
Most of the $200$ books we test with $50$-token prefixes resemble this pattern.
However, for a given book, we are often (but not always) able to extract some substantive sequences from \textsc{Llama 3} and \textsc{3.1 70B}.\looseness=-1 

Second, books like \emph{Beloved}~\cite{Beloved} (Figure~\ref{fig:comparisons:heatmaps}) and \emph{Kitchen Table Wisdom} (Figure~\ref{fig:comparisons:heatmaps-random}) represent a more intermediate case.  
For these books, memorization is generally low for most LLMs with respect to $50$-token prefixes.
However, \textsc{Llama 3} and \textsc{Llama 3.1 70B} memorize a considerable number of sequences, and increasing prefix length can sometimes reveal significant amounts of memorization (Appendix~\ref{app:sec:validity:baseline}).
For \emph{Beloved}, $800$-token prefixes reveal extraction coverage of nearly $80\%$ from \textsc{Llama 3.1 70B}, compared to approximately $12.5\%$ coverage for $50$-token prefixes (Figure~\ref{fig:prefix-coverage}). 
Other \textsc{Llama} models memorize significantly less, and category (b) models (\textsc{DeepSeek v1 67B}, \textsc{Qwen 2.5 72B}, and \textsc{Gemma 2 27B}) memorize even less. 
For \emph{Beloved}, the $7$ largest models we test---category (b) and all large \textsc{Llama} models---memorize the same passages at ${\sim}170$K~\citep{belovedquote1, belovedquote3} and ${\sim}480$K characters~\citep{belovedquote2}.
Both regions contain famous quotes from the book, which we confirmed are highly duplicated on the Internet. 
In general, we find that specific passages from a given book are memorized across models trained by different organizations---Meta (\textsc{Llama}), Alibaba (\textsc{Qwen}), and Google DeepMind (\textsc{Gemma}).
As with \textsc{Phi 4} (Section~\ref{sec:validity:controls}), models that otherwise exhibit low memorization (except for such passages) may not have been trained on the whole book; 
rather, they may have memorized these passages due to their inclusion and duplication in other sources (Appendix~\ref{app:sec:validity:membership}).\looseness=-1  

Third, some books exhibit enormous degrees of memorization for the $7$ largest models, with lesser amounts of memorization in smaller models like \textsc{Llama 1 13B}. 
These are often popular books like \emph{Harry Potter and the Sorcerer's Stone}~\citep{Harry_Potter_and_the_Sorcerer_s_Stone} (Figure~\ref{fig:comparisons:heatmaps}), \emph{1984}~\citep{Nineteen_Eighty-Four} (Figure~\ref{fig:1984:slide:main}), and \emph{The Alchemist}~\cite{The_Alchemist} (Figure~\ref{fig:comparisons:heatmaps-random}). 
While memorization of such books tends to be significant for larger models, it is especially high for \textsc{Llama 3} and \textsc{3.1 70B}, which appear to memorize the majority of several books---some in their entirety~\citep[e.g.,][]{Harry_Potter_and_the_Chamber_of_Secrets, The_Great_Gatsby,Ulysses, The_Hobbit}.
For instance, with $50$-token prefixes, \textsc{Llama 3.1 70B} exhibits extraction coverage of approximately $49.1\%$ for \emph{The Alchemist}, while the similarly sized \textsc{DeepSeek v1 67B}, \textsc{Qwen 2.5 72B}  exhibit approximately $3.9\%$ and $5.8\%$, respectively (Appendix~\ref{app:sec:coverage-results}).\looseness=-1

In general, small models generally memorize less than larger ones, but there is variation across books. 
For example, \textsc{Llama 1 13B} memorizes more of \emph{Harry Potter} than \emph{M. Butterfly} and \emph{Beloved}; 
\textsc{Llama 3.1 8B} memorizes portions of \emph{We Were Eight Years in Power}~\citep{We_Were_Eight_Years_in_Power} (Figure~\ref{fig:wewere:main}), but less from other books (Appendix~\ref{app:sec:validity:baseline}). 
Variation is more pronounced in larger models, as is clear from extraction coverage distributions (Section~\ref{sec:book-procedure:averages}, Appendix~\ref{app:sec:sliding-window:percentage}).
Many results resemble those in Figure~\ref{fig:comparisons:heatmaps} for \emph{M. Butterfly}, as is the case for most of the plaintiff-authored books in \emph{Kadrey et al. v. Meta, Inc.}
\citep{kadreyamendedconsolidated} (with exceptions like \textsc{Llama 3.1 70B} for \emph{We Were Eight Years in Power}, see Figure~\ref{fig:wewere:main}).
However, for \textsc{Llama 3} and \textsc{Llama 3.1 70B}, many books yield results similar to those for \emph{Beloved} (e.g., \emph{The Handmaid's Tale}~\citep{The_Handmaid_s_Tale}, \emph{The Da Vinci Code}~\citep{The_Da_Vinci_Code}, \emph{The Myth of Sisyphus}~\citep{The_Myth_of_Sisyphus}, see Appendix~\ref{app:sec:sliding-window:results});
and, in extreme cases like \emph{Harry Potter}~\citep{Harry_Potter_and_the_Sorcerer_s_Stone}, entire books are memorized in the model. 
\textsc{Llama 1 65B}, \textsc{Llama 2 70B}, and category (b) models also exhibit high memorization for some, but not all, of these books.\looseness=-1 

Altogether, these patterns underscore the importance of training choices in shaping memorization.
Models do not memorize all of their training data~\citep{lee2022dedup, carlini2021extracting, mahdavi2024memorizationcapacitymultiheadattention, collins2017capacitytrainabilityrecurrentneural}. 
The mere existence of particular sequences within a training dataset does not imply those sequences will be memorized~\citep{lee2022dedup, lee2023explainers, lee2023talkin, lee2024talkinshort}.
For example, we observe that similarly sized LLMs trained by different organizations display different degrees of memorization of the same book---even when the evidence strongly suggests the entire book (e.g., \emph{Harry Potter}) was included in their respective training datasets (Appendix~\ref{app:sec:validity:membership}). 
This is also evident with respect to \textsc{Llama} models, which were all trained by the same organization (Meta) on \texttt{Books3}. 
All \textsc{Llama} models (by size) memorize more than the other models we evaluate. 
But notably, within size classes, newer versions typically memorize more from specific books than earlier ones:
\textsc{Llama 3} models appear to have memorized more from books than \textsc{Llama 2} models, which in turn memorize more than \textsc{Llama 1}.\looseness=-1

\vspace{-.3cm}
\section{Recovering \emph{Harry Potter and the Sorcerer's Stone} with one seed prompt}\label{sec:book:seed}
\vspace{-.2cm}

Our main experiments estimate how much of a given book is memorized within an LLM's weights, with respect to $50$-token prefixes and  suffixes. 
These results do not necessarily imply that one would be able to extract large, contiguous portions of a book all at once during generation. 
However, for \textsc{Llama 3.1 70B} on some books like \emph{Harry Potter} and \emph{1984} (Figures~\ref{fig:1984:slide:main}~\&~\ref{fig:comparisons:heatmaps}), our results show that there exist extraordinarily high extraction probabilities throughout. 
These probabilities approach or reach $100\%$ when we increase the prefix length (Figure~\ref{fig:prefix-coverage} \& Appendix~\ref{app:sec:validity:baseline}).
This degree of memorization suggests one could possibly recover the entire book near-verbatim, starting from a single short prompt of ground-truth text to seed the generation process.
Using beam search as the decoding method---which deterministically approximates the highest probability sequence under the model for a given configuration---\textsc{Llama 3.1 70B} could reasonably continue the seed prompt to complete the book.\looseness=-1

\custompar{A simple recovery procedure}
We test this idea with about $100$ lines of boilerplate generation code using HuggingFace APIs for \emph{Harry Potter}~\citep{Harry_Potter_and_the_Sorcerer_s_Stone}.
In essence, this procedure is standard autoregressive generation, where we discard the oldest tokens in the context as new ones are generated.
We begin with an $s$-token seed prompt (e.g., $s\!=\!60$, the first line of \emph{Harry Potter}), and set a maximum sliding context length of $n$ tokens.  
The LLM generates $m$ tokens at a time with beam search, until reaching $n$ tokens.
We then remove the first $m$ tokens from the $n$-length context and append the generated $m$ tokens, creating a new $n$-token prompt for the next generation step. 
We repeat this process for the length of the book.  
After the first $n$ tokens, the clipped, $n$-length context window progresses beyond the initial seed prompt.
All subsequent prompts do not contain \emph{any} ground-truth text drawn from the book;  
they consist entirely of generated text from prior iterations.\looseness=-1

\begin{figure*}[t!]
\centering
\begin{subfigure}[t]{0.48\textwidth}
\vspace{-.4cm}
    \centering
    \vspace*{0pt}
    \includegraphics[width=\linewidth]{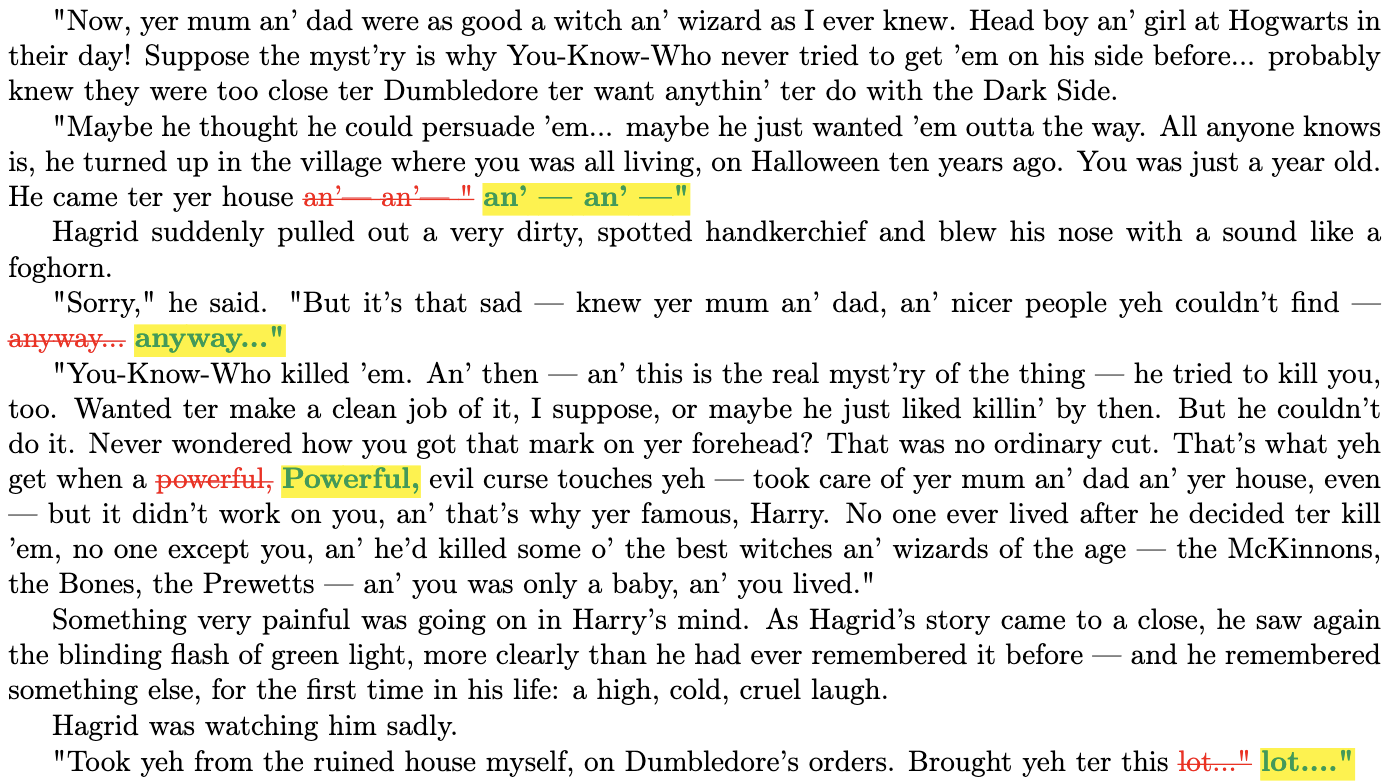}
\end{subfigure}
\hfill
\begin{subfigure}[t]{0.48\textwidth}
\vspace{-.4cm}
   \centering
   \vspace*{0pt}
   \includegraphics[width=\linewidth]{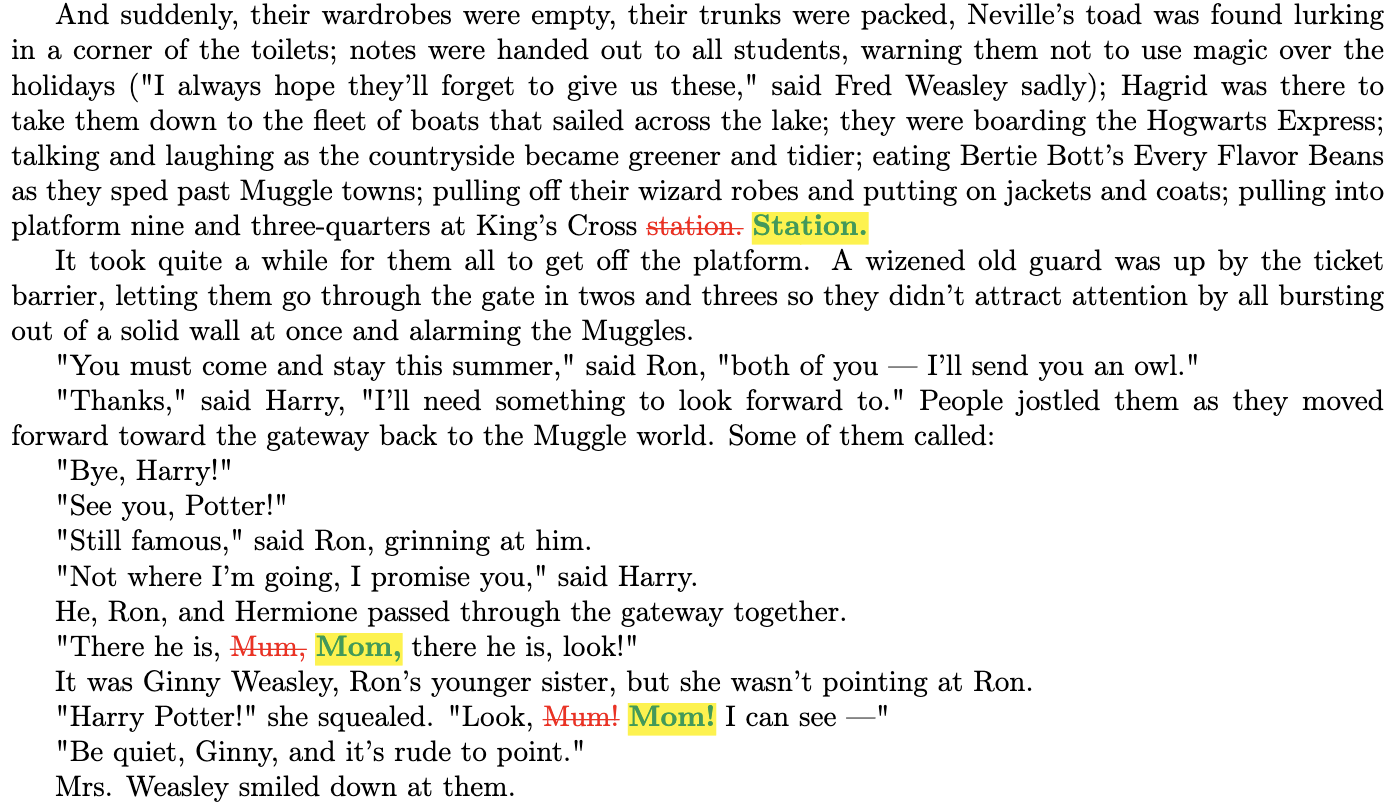}
\end{subfigure}\\ 
\vspace{.2cm}
\begin{subfigure}[t]{0.48\textwidth}
    \centering
    \vspace*{0pt}
    \includegraphics[width=\linewidth]{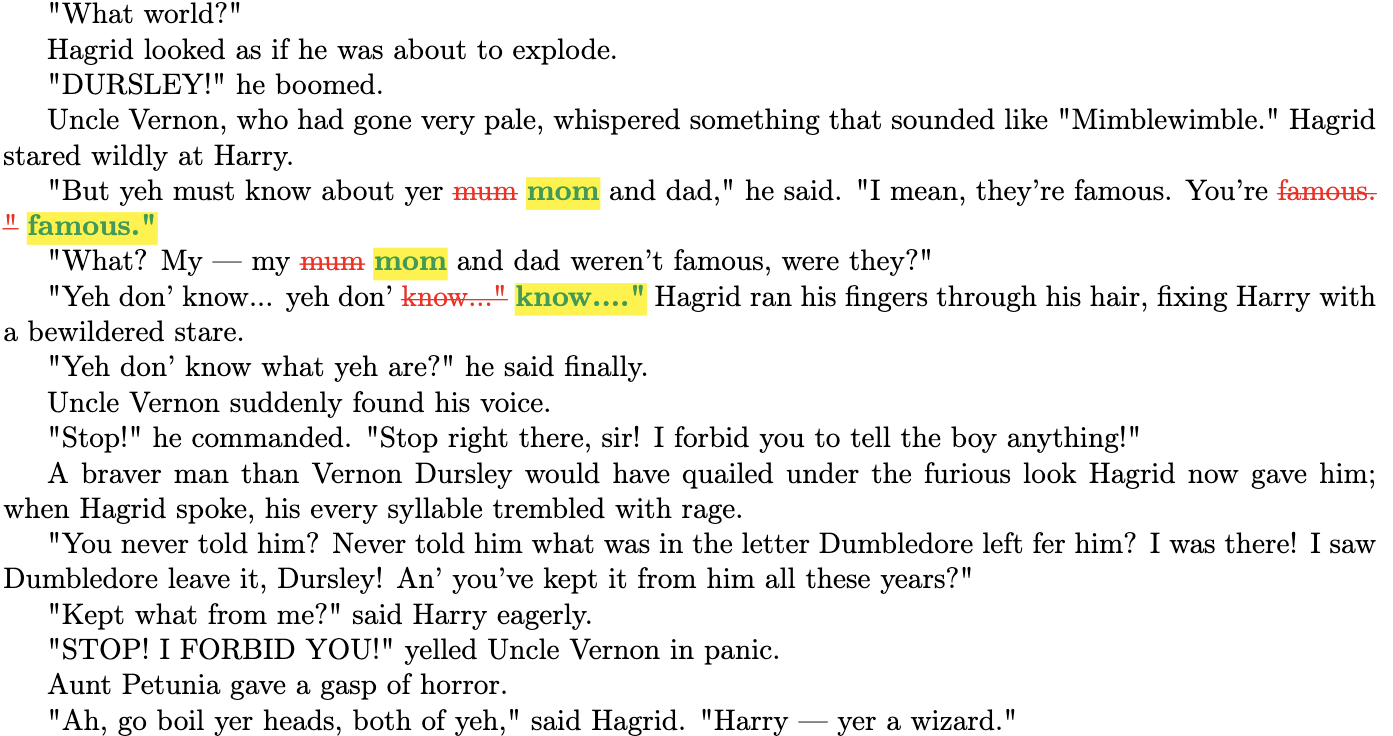}
\end{subfigure}
\hfill
\begin{subfigure}[t]{0.48\textwidth}
   \centering
   \vspace*{0pt}
   \includegraphics[width=\linewidth]{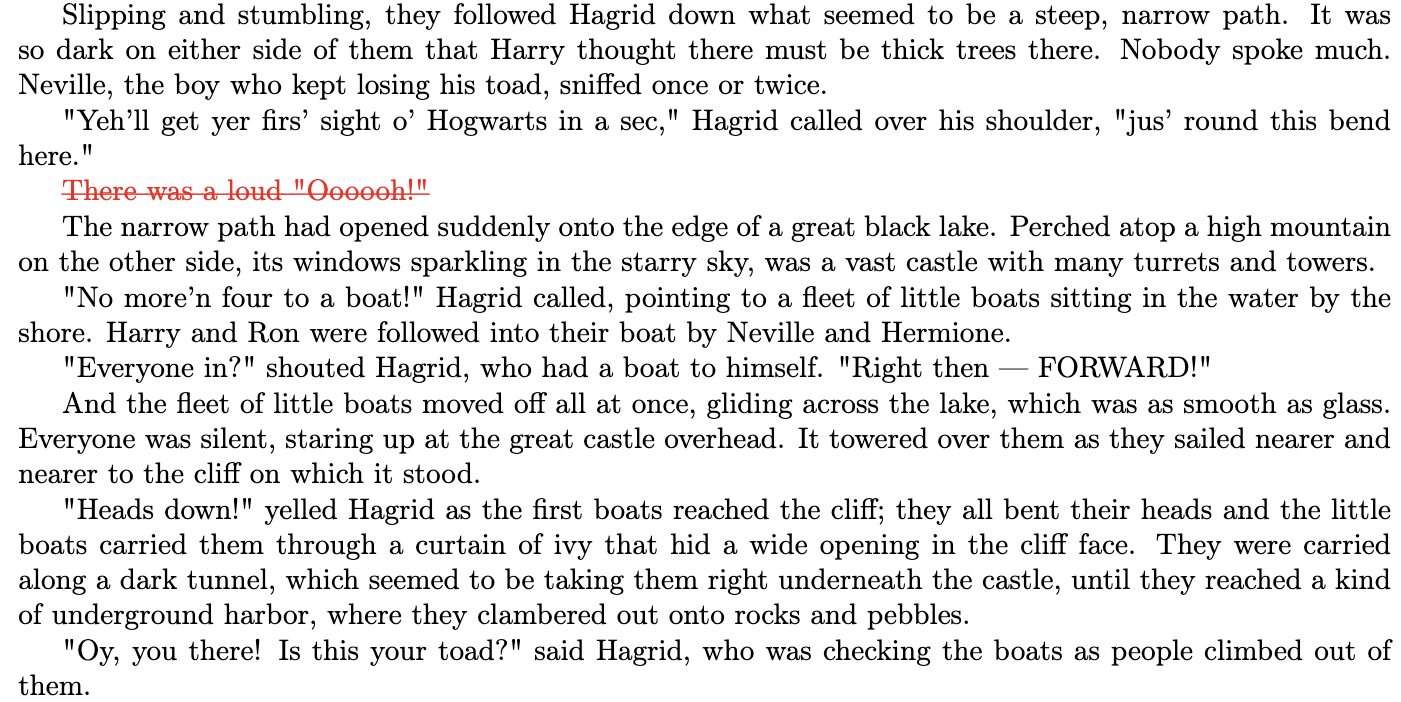}
\end{subfigure} 
\caption{\textbf{Visual diffs of long-form extracted text.}
Four samples of the diff between the ground-truth text of \emph{Harry Potter} from \texttt{Books3} and the text we generated using \textsc{Llama 3.1 70B}. 
Text crossed out in red indicates ground-truth text from the book absent in the generation.
Text highlighted in yellow is present in the generation, but absent from the ground-truth book. 
Note that this visual diffing procedure is sensitive to benign formatting differences.
We select these four samples to highlight types of variations we observe between the ground-truth book and the generation.
In general, the large majority of the diff exhibits no differences (i.e., is black-and-white). 
See Appendix~\ref{app:sec:reconstruct:method}.}
\label{fig:diff}
\vspace*{-.4cm}
\end{figure*}

On our hardware, after testing just a handful of configurations for beam search and a few different short 
and seed prompts, we \emph{deterministically}\footnote{Beam search is a deterministic decoding method that explores multiple candidate sequences in parallel to approximate the highest-probability continuation under the model. 
    At each generation step, it keeps the top-$k$ highest-probability sequences so far. 
    At the final step, it outputs the sequence with the highest probability.\looseness=-1
} generated a near-exact copy of the entire 304-page book.
Figure~\ref{fig:diff} shows partial results using a seed prompt of \llmtext{Mr.\ and Mrs.\ D} ($s{=}6$ tokens, drawn from the first line of the first chapter, prepended with the beginning-of-sequence token), 
beam width of $8$, maximum context length $n\!=\!3000$, and generation length $m\!=\!50$ tokens at each step. 
In practice, \textsc{Llama 3.1 70B} tended to predict end-of-sequence (\texttt{EOS}) tokens at the ends of chapters. 
To address this, we removed the \texttt{EOS} token from the generated $m$ tokens before appending them to the context, replacing it with the tokens for \texttt{"CHAPTER \{c+1\}"} (with \texttt{\{c+1\}} spelled out in words, e.g., when $c\!=\!2$, we insert \texttt{"CHAPTER THREE"}). 
With this simple adjustment, the model successfully continued to generate the book. 
Because beam search is a deterministic decoding algorithm, our results are reproducible.
We discuss the stability of this result and future work beyond this ``existence proof'' in Appendix~\ref{app:sec:reconstruct}.\looseness=-1

\custompar{Assessing the fidelity of the extracted book}
Qualitatively, the generated text is nearly identical to the ground-truth book.
Differences predominantly involve small formatting inconsistencies: 
white space, capitalization, use of underscores (\texttt{\_}) to indicate italics, etc. 
The \texttt{Books3} version of \emph{Harry Potter} uses British spelling (e.g., \texttt{"Mum"} instead of \texttt{"Mom"}).
Very occasionally, when there is only a single line in a paragraph, the LLM skips that line during generation. 
We also quantitatively assess the similarity of the two texts. 
We apply minimal text normalization,\footnote{We remove \texttt{\_} (used to signify italics in the \texttt{Books3} version) and align ellipses to \texttt{"..."} (which appear as \texttt{". . ."} in \texttt{Books3}).\looseness=-1} split both documents on whitespace characters, and then measure similarity on the resulting word lists with a greedy approximation of longest common substring.
This yields a score of $0.992$, where $1$ would indicate a perfect match between the ground-truth book and the generation (Appendix~\ref{app:sec:reconstruct}).\looseness=-1

\vspace{-.2cm}
\section{Takeaways for copyright law and policy}\label{sec:takeaways}
\vspace{-.1cm}

Our experiments show that the extent of memorization varies by model, model family, model size, the book tested, and even within different parts of an individual book  (Section~\ref{sec:book-procedure}--\ref{sec:book:compare}).
These results complicate the narratives both plaintiffs and defendants typically present in copyright cases when describing how LLMs work. 
They support the positions of plaintiffs in some respects and of defendants in others. 
We see three primary implications for copyright disputes: 
LLMs could be derivative works of the books they memorize (Section~\ref{sec:takeaways:existence}); 
in some cases, it may be practical to generate infringing copies (Section~\ref{sec:takeaways:outputs});
and, class actions are ill-suited for copyright claims about memorization (Section~\ref{sec:takeaways:varies}).\looseness=-1

\vspace{-.1cm}
\subsection{LLMs could be derivative works of the books they memorize}\label{sec:takeaways:existence}
\vspace{-.1cm}

In the case of some models, there is quite a lot of memorization of some books, though most books are not memorized at all---with respect to our specific extraction methodology---either in whole or in part. 
Such evidence of memorization in the model matters for the dispute over whether models themselves are derivative works  (Section~\ref{sec:copyright}), an argument that courts have thus far not been receptive to~\citep{bartzjudgment, kadreyjudgment}.\looseness=-1 

A work is not a derivative work unless it is ``substantially similar'' in significant part to the original work~\citep[e.g.,][]{litchfield}. 
While prior debates over whether models can qualify as derivative works have been largely theoretical~\citep{cooper2024files, lee2023talkin}, the results of our probabilistic extraction procedure offer empirical evidence that is directly relevant to this issue. 
They indicate that models are not, as plaintiffs sometimes contend, mere copies of all of the works on which they were trained. 
However, they also suggest that at least some LLMs (e.g., \textsc{Llama 3.1 70B}) may be derivative works of at least some books (e.g.,  \emph{Harry Potter and the Sorcerer's Stone})---because those LLMs have memorized a significant amount of protectable expression from those books. 
The law does not require that the entire work be included in the derivative; 
it is enough that the derivative incorporates a substantial amount of protectable expression.  
We find that this turns out to be true for some (but not all) books for some (but not all) models.\looseness=-1  

\custompar{Fair use}
The inquiry does not end here; an LLM, like the training dataset, may be protected under copyright's fair use doctrine~\citep{lemley2021fairlearning}. 
But the fair use analysis of the model itself may be different than the fair use analysis of the training dataset~\citep{lee2023talkin}.
Copyright law treats \newterm{intermediate copies}, such as those used internally within a system, differently from external-facing copies.
A sequence of copyrighted data used for training is an intermediate copy used internally by AI companies for model training, rather than sold as a product. 
Unless that sequence is extracted, it is only used internally in the course of the model producing outputs. 
Beyond extraction, individual sequences may to lesser or greater extents influence model outputs, but many of these outputs will not be substantially similar to any training data~\citep{henderson2023fair,lee2023talkin}.\looseness=-1  

A model, too, is an intermediate work in many cases.
If the model memorizes content but only generates noninfringing outputs, that is a strong argument that how the model makes use of that content is transformative and ought to be permissible fair use.
But for companies that sell or release LLMs to others under open source licenses, the model itself is the product, and sometimes one that is being sold for commercial gain. 
That does not defeat the fair use argument, but it makes the analogy to training data less precise. 
In these instances, it may be harder to rely on the cases that justify training on copyrighted data as fair use~\citep[e.g.,][]{authorsguild, sega, sonyconnectix}.\looseness=-1

\custompar{Potential consequences of a model being a copy of works it has memorized}
A finding that the model itself is a copy of some copyrighted works, and that a \newterm{distribution} of the model was thus a distribution of a copy of those works, could be dramatic for the AI industry. 
\textsc{Llama 3.1 70B} was downloaded $105{,}029$ times from HuggingFace in May 2025 (Figure~\ref{fig:hfllamadownloads})---far less from the height of its release in 2024. 
If we say conservatively that the model has been downloaded 1 million times since its release,\footnote{Meta announced in March 2025 that total downloads for all of its \textsc{Llama} models surpassed $1$ billion~\citep{metadownloads}.
    In 2024, the \emph{Times} reported that \textsc{Llama 2} models had been downloaded $180$ million times~\citep{nytllama2}.
} then those 1 million downloads could be seen as 1 million potentially infringing distributions of reproductions of \emph{Harry Potter and the Sorcerer's Stone}---as well as any other books in copyright for which a more than \emph{de minimis} amount of copyrighted expression has been memorized. 

In general, damages for copyright infringement can run very high:
\newterm{statutory damages} can be as high as \$150,000 per infringed work, and \newterm{actual damages} can be much higher. 
If the model itself is a copy and is not protected by fair use, courts might also order that it not be distributed or even that copies of the model itself be destroyed~\citep{17usc503}.
We have only studied a relatively small number of books ($200$) and models ($14$), so cannot draw strong conclusions about the extent of memorization of books in general.
In principle, if the overall quantity of memorized books in open-weight models is particularly high, it is not an exaggeration to say that the risk of damages across all of those books and of orders of destruction are an existential threat to the companies training these models.\looseness=-1

\custompar{Policy considerations}
The effect of these copyright rules may be to treat openly distributed models less favorably than models gated behind closed systems, because the distributed models contain a latent copy or derivative of some copyrighted works. 
However, there is no good policy reason to treat open-weight models more harshly in copyright law than proprietary ones. 
Policymakers may want to act to protect models from being declared unlawful copies, in order to prevent crushing financial liability on developers for providing a model that generates transformative, noninfringing outputs. 
However, a blanket protection for all open-weight models may also be unwarranted. 
The extent to which models memorize their training data is significantly influenced by choices during training, such as de-duplicating the training dataset~\citep{cooper2024files, carlini2023quantifying, lee2023talkin, lee2022dedup, hayes2025measuringmemorizationlanguagemodels}. 
Policymakers may be less inclined to protect models if the developers made training choices known to substantially increase the risk of memorization---for example, deliberately duplicating particular books.\looseness=-1 

\vspace{-.1cm}
\subsection{It may be practical to generate infringing copies in some cases}\label{sec:takeaways:outputs}
\vspace{-.1cm}

In our main experiments with the sliding-window probabilistic extraction procedure (Sections~\ref{sec:book-procedure} \& \ref{sec:book:compare}), we use relatively short $50$-token prompts drawn from books to see if it is possible for an LLM to generate verbatim the following $50$ tokens with high probability. 
As is standard in ML, we use extraction of text snippets for evidence of memorization inside the model. 
But no one would realistically use this technique to extract an entire book. 
Doing so would require access to at least half of the book to use as $50$-token prefixes in order to even attempt to produce the other half in $50$-token suffixes.
And, for sequences that are memorized, it might take in expectation up to $1/\tau_\text{min}\!=\!1{,}000$ tries to produce each verbatim suffix, with no indication of which of the many responses represents the actual text of the book.\footnote{Relatively low extraction probabilities (e.g., $p_\vz\!=\!0.1\%$) do not always mean that it would take a large number of attempts to extract the suffix. 
In some cases, other decoding schemes like beam search may make extraction of such suffixes deterministic, requiring just one attempt~\citep{cooper2026nv}.}\looseness=-1

\custompar{Traditional extraction measurements vs.\ recovery of (whole) books}
Reliable extraction of large segments of text would require other approaches, like the recovery procedure that we apply to \emph{Harry Potter and the Sorcerer's Stone} with \textsc{Llama 3.1 70B} (Section~\ref{sec:book:seed}). 
This result is an extreme case and is not representative of most of the other books we have tested (Appendix~\ref{app:sec:reconstruct:fail}).
Nevertheless, it is significant: 
in contrast to claims by some defendants~\citep{concordanthropic}, it illustrates that, for LLMs that exhibit high degrees of memorization of specific works, it may be possible to generate copies or derivative works of those works. 
General assessments of the extent of this type of behavior, and how easy it is to elicit it, require more detailed exploration. 

\custompar{Practicality of extraction and responsibility for infringement}
Even though our results suggest it may be possible to recover large segments of copyrighted text from an LLM, they do not imply that it will always be easy to do so. 
This may affect who courts view as directly liable for any output infringement~\citep{lee2023talkin}.
As a policy matter, the case for liability against the model developer is stronger if simple prompts generate infringing outputs~\citep{gemavoai}. 
The case is weaker if a user (e.g., a plaintiffs' lawyer) employs more intensive strategies to try to generate infringing material. 
In that case, the adversarial user---who perhaps runs hundreds or thousands of prompts to try to get one output that is infringing---seems more like a direct volitional actor misusing the model~\citep{costar}; 
they should likely bear some responsibility for the infringement.\footnote{Models are used \emph{at scale}---in some cases, by millions of users.
Hypothetically, a million users interacting with a particular model might independently prompt the model with the prefix of a memorized sequence. 
Even for a verbatim suffix that (in expectation) appears only once for every $1{,}000$ prompts (i.e., $p_\vz\!=\!\tau_\text{min}\!=\!0.1\%$), in expectation $1{,}000$ of these users could expect to see the verbatim memorized suffix in the output.\looseness=-1
} 
By contrast, in several cases (e.g., \emph{Harry Potter}, \emph{1984}), it does not take a thousand or even hundreds of generations to get large amounts of text; 
it takes only a handful.
And in one case that we have identified so far (\emph{Harry Potter}), it is possible to generate the whole book directly from a short starting prompt and boilerplate generation code (Section~\ref{sec:book:seed}). 
We will address this further in future work. 

\custompar{Considerations for LLM training and infringement}
The fact that extraction is probabilistic may affect resolution of the ``is training infringement'' issue. 
However, it is not determinative.
Using a work as training data does not mean that it will necessarily be memorized.
And, even if it is memorized, it may still be difficult to extract that work.
Large parts of a work would need to be extractable in order to create \newterm{superseding copies} (i.e., a copy that can stand in for/ replace the work).
Based on our current results, this seems unlikely (though possible, see Section~\ref{sec:book:seed}) in practice for most works.\looseness=-1 

Nonetheless, the fact that memorization occurs at all creates a point of distinction from~\citet{authorsguild}. 
In this case, the court held that Google's scanning of books to create a searchable database, and Google's public display of limited text snippets on Google Books, was fair use. 
The court also considered and rejected an argument that hackers could gain unauthorized access to Google's internal books database that supported the service, and thereby obtain plaintiffs' entire works~\citep[at 227-28]{804fd}.
In contrast, our results show that it is possible to recreate some of the content of some books by ``hacking'' the model itself. 
We think it unlikely as a practical matter that people will use the model in this way; 
there are easier and more effective ways to pirate a book. 
But at a minimum, it complicates the copyright fair use analysis.\looseness=-1 

\subsection{Class actions are ill-suited for copyright claims about memorization}\label{sec:takeaways:varies}

To be certified as a \newterm{class action}, plaintiffs generally have to show that \emph{all} putative class members share common legal and factual issues, as well as suffer common injuries. 
Because many of the pending cases are proceeding as class actions, plaintiffs will therefore need to demonstrate that \emph{all} book owners meet this standard so that a court can certify the class and treat their claims together~\citep{class}.\looseness=-1

As is clear from our results, this may be impossible.
Basic questions about whether a particular model actually incorporated any significant expression from any individual plaintiff's book cannot be generalized. 
Rather, it varies case by case:
the extent of memorization differs widely from model to model, and even from work to work within a model (Section~\ref{sec:book:compare}).
This makes it difficult to assess, on a class-wide basis, whether a particular model copied a particular work and whether, for that model, infringing outputs based on memorization are even possible. 
Indeed, we show that memorization of works varies for many of the actual named authors in lawsuits, and even across books for the same plaintiffs (e.g., Ta-Nehisi Coates, see Figure~\ref{fig:wewere:main};  Appendices~\ref{app:sec:sliding:The_Beautiful_Struggle}--\ref{app:sec:sliding:The_Water_Dancer}  \&~\ref{app:sec:validity:baseline}). 
Answering questions about the degree to which a specific LLM has memorized a specific work requires experimental evidence---for example, running every book through tests similar to the ones we performed here. 
Some plaintiffs may be able to show copying, but others will not. 
Courts generally deny class certification in such circumstances~\citep[e.g.,][]{walmart}.\looseness=-1 

\section{Conclusion}\label{sec:conclusion}

We develop (Section~\ref{sec:book-procedure}) and validate (Section~\ref{sec:validity}) a sliding-window procedure to estimate the extent to which open-weight LLMs memorize different books contained in the \texttt{Books3} training corpus.
Our results show that the extent of verbatim memorization of books in open-weight LLMs is far more significant than previously described.
Memorization varies widely from model to model and from book to book within each model (Section~\ref{sec:book:compare}). 
With respect to our extraction procedure, most LLMs do not memorize most books, but this is not always the case.
In extreme instances, an LLM may memorize a book so significantly that it is possible to extract the entire book near-verbatim, as we have shown with \textsc{Llama 3.1 70B} for \emph{Harry Potter and the Sorcerer's Stone} (Section~\ref{sec:book:seed}). 
Altogether, our results complicate current disputes over copyright infringement, both by rejecting simplistic claims made by both sides about how LLMs work and by demonstrating that there is no single answer to the question of how much an LLM memorizes (Section~\ref{sec:takeaways}).\looseness=-1


\section*{Author contributions}

AFC conceived of the project, wrote the code, designed and conducted the experiments, ran the analysis of the results, made the figures, and built the project website. 
MAL and ABC led the legal analysis, with support by AFC. 
AFC, MAL, and ABC drafted the manuscript. 
CDS provided technical feedback. 
CDS, MAL, PL, and DEH gave iterative feedback on visualizations. 
AC and AA helped compile books metadata and the supplementary material. 
PL, DEH, CDS, AA, and AC provided feedback on the manuscript.
DEH provided the computing resources for the experiments.
AG provided access to \texttt{Books3}.\looseness=-1 

\section*{Resources}

We release an interactive, searchable version of our main results on $200$ books and $14$ models \href{http://books-memorization.github.io}{online}, as well as a general-purpose coding suite for probabilistic extraction on \href{https://github.com/pasta41/probabilistic-extraction-toolkit}{GitHub}.

\begin{ack}
This work was supported by a Stanford HAI Seed Research Grant. 

We thank the Stanford Research Computing Center for providing computational resources and support that contributed to these research results. 

We thank Franziska Boenisch, danah boyd, Nicholas Carlini, Alexandra Chouldechova, James Grimmelmann, Adam Dziedzic, Jamie Hayes, Matthew Jagielski, Kevin Klyman, Katherine Lee, Milad Nasr, Matthew Sag, Pamela Samuelson, Florian Tramèr, Danny Wilf-Townsend, and others for helpful discussion and comments on earlier versions of this project. 
We also thank Timothy B. Lee for his extremely clear explanation of our work~\citep{lee2025arstechnica}, which helped us simplify the presentation of the math in this paper. 

Ahmed Ahmed acknowledges generous support from a Knight-Hennessy Fellowship, an NSF Graduate Research Fellowship, and a Georgetown Foundation Research Grant.

A. Feder Cooper is an Assistant Professor of Computer Science at Yale University, a Research Scientist at AVERI, and a Postdoctoral Affiliate at Stanford University in Percy Liang's group in the Department of Computer Science and Daniel E. Ho's group at Stanford Law School. Until December 2025, Cooper was a full-time employee of Microsoft, working as a postdoctoral researcher within Microsoft Research. 
Nothing in this paper should be attributed to Microsoft or AVERI.

Mark A. Lemley is a full-time professor at Stanford Law School. He also has a consulting practice as a partner at Lex Lumina LLP. 
As part of that practice he has represented AI companies in copyright litigation. 
Nothing in this paper should be attributed to those companies.

Percy Liang is a full-time professor at Stanford, and he is also a co-founder and consultant at Together AI, Simile AI, and Taiku AI; and also an advisor to Virtue AI and Humanitas AI. 
Nothing in this paper should be attributed to those companies.

Christopher De Sa is a full-time professor at Cornell, and he is also a consultant at Together AI, and an advisor at Sambanova Systems and Inception Labs.
Nothing in this paper should be attributed to those companies.
\end{ack}

\bibliography{references}
\bibliographystyle{plainnat}


\appendix
\newpage
\appendix

\section*{Appendix Table of Contents}



\addcontentsline{toc}{section}{Appendix}

\begin{itemize}[labelwidth=!, labelsep=1em, leftmargin=*, align=left]
    \item[\textbf{\ref{app:sec:background}}] \textbf{Additional notes on memorization and probabilistic extraction} \dotfill \pageref{app:sec:background}

    \begin{itemize}[labelwidth=!, labelsep=1em, leftmargin=*, align=left]
        \item[\textbf{\ref{app:sec:intro}}] \quad Additional notes on the ``careless people'' example \dotfill \pageref{app:sec:intro}
        
        \item[\textbf{\ref{app:sec:background:metrics}}]\quad Metrics \dotfill \pageref{app:sec:background:metrics}
        
        \item[\textbf{\ref{app:sec:background:compute}}]\quad Computing sequence probabilities in one forward pass \dotfill \pageref{app:sec:background:compute}
      \end{itemize}   

    \item[\textbf{\ref{app:sec:replication}}] \textbf{Testing our measurement pipeline} \dotfill \pageref{app:sec:replication}
    \item[\textbf{\ref{app:sec:rates}}]  \textbf{Experiments on extraction rates} \dotfill \pageref{app:sec:rates}

    \begin{itemize}[labelwidth=!, labelsep=1em, leftmargin=*, align=left]
        \item[\textbf{\ref{app:sec:rates:setup}}] \quad Setup \dotfill \pageref{app:sec:rates:setup}
        
        \item[\textbf{\ref{app:sec:rates:sample}}]\quad Experimental procedure \dotfill \pageref{app:sec:rates:sample}
        
        \item[\textbf{\ref{app:sec:rates:takeaways}}]\quad High-level takeaways \dotfill \pageref{app:sec:rates:takeaways}
    \end{itemize}
    \item[\textbf{\ref{app:sec:sliding-window}}]\textbf{Extended results for sliding-window experiments} \dotfill \pageref{app:sec:sliding-window}

        \begin{itemize}[labelwidth=!, labelsep=1em, leftmargin=*, align=left]
            \item[\textbf{\ref{app:sec:sliding-window:setup}}] Setup \dotfill \pageref{app:sec:sliding-window:setup}
            \item[\textbf{\ref{app:sec:sliding-window:procedure}}] Sliding-window probabilistic extraction procedure \dotfill \pageref{app:sec:sliding-window:procedure}

            \item[\textbf{\ref{app:sec:sliding-window:results}}] Book-specific results \dotfill \pageref{app:sec:sliding-window:results}
                \begin{itemize}[labelwidth=!, labelsep=1em, leftmargin=*, align=left]
                    \item[\textbf{\ref{app:sec:sliding:Things_Fall_Apart}}]\quad\;\; \textit{Things Fall Apart}, \citeauthor{Things_Fall_Apart} \dotfill \pageref{app:sec:sliding:Things_Fall_Apart}

                    \item[\textbf{\ref{app:sec:sliding:The_Hitchhiker_s_Guide_to_the_Galaxy_-_Omnibus}}]\quad\;\; \textit{The Hitchhiker's Guide to the Galaxy - Omnibus}, \citeauthor{The_Hitchhiker_s_Guide_to_the_Galaxy_-_Omnibus} \dotfill \pageref{app:sec:sliding:The_Hitchhiker_s_Guide_to_the_Galaxy_-_Omnibus}

                    \item[\textbf{\ref{app:sec:sliding:Americanah}}]\quad\;\; \textit{Americanah}, \citeauthor{Americanah} \dotfill \pageref{app:sec:sliding:Americanah}
                    
                    \item[\textbf{\ref{app:sec:sliding:The_Baghdad_Clock}}]\quad\;\; \textit{The Baghdad Clock}, \citeauthor{The_Baghdad_Clock} \dotfill \pageref{app:sec:sliding:The_Baghdad_Clock}

                    \item[\textbf{\ref{app:sec:sliding:Industrial_Magic}}]\quad\;\; \textit{Industrial Magic}, \citeauthor{Industrial_Magic} \dotfill \pageref{app:sec:sliding:Industrial_Magic}

                    \item[\textbf{\ref{app:sec:sliding:Fantastic_Voyage}}]\quad\;\; \textit{Fantastic Voyage}, \citeauthor{Fantastic_Voyage} \dotfill \pageref{app:sec:sliding:Fantastic_Voyage}
                    
                    \item[\textbf{\ref{app:sec:sliding:The_Complete_Robot}}]\quad\;\; \textit{The Complete Robot}, \citeauthor{The_Complete_Robot} \dotfill \pageref{app:sec:sliding:The_Complete_Robot}

                    \item[\textbf{\ref{app:sec:sliding:The_Handmaid_s_Tale}}]\quad\;\; \textit{The Handmaid's Tale}, \citeauthor{The_Handmaid_s_Tale} \dotfill \pageref{app:sec:sliding:The_Handmaid_s_Tale}

                    \item[\textbf{\ref{app:sec:sliding:Pride_and_Prejudice}}]\quad\;\; \textit{Pride and Prejudice}, \citeauthor{Pride_and_Prejudice} \dotfill \pageref{app:sec:sliding:Pride_and_Prejudice}

                    \item[\textbf{\ref{app:sec:sliding:The_Christmas_Train}}]\quad\; \textit{The Christmas Train}, \citeauthor{The_Christmas_Train} \dotfill \pageref{app:sec:sliding:The_Christmas_Train}
                    \item[\textbf{\ref{app:sec:sliding:Notes_of_a_Native_Son}}]\quad\; \textit{Notes of a Native Son}, \citeauthor{Notes_of_a_Native_Son} \dotfill \pageref{app:sec:sliding:Notes_of_a_Native_Son}

                    \item[\textbf{\ref{app:sec:sliding:Another_Country}}]\quad\; \textit{Another Country}, \citeauthor{Another_Country} \dotfill \pageref{app:sec:sliding:Another_Country}

                    \item[\textbf{\ref{app:sec:sliding:The_Lemon_Table}}]\quad\; \textit{The Lemon Table}, \citeauthor{The_Lemon_Table} \dotfill \pageref{app:sec:sliding:The_Lemon_Table}
                    \item[\textbf{\ref{app:sec:sliding:Dante_and_the_Origins_of_Italian_Literary_Culture}}]\quad\; \textit{Dante and the Origins of Italian Literary Culture}, \citeauthor{Dante_and_the_Origins_of_Italian_Literary_Culture} \dotfill \pageref{app:sec:sliding:Dante_and_the_Origins_of_Italian_Literary_Culture}

                    \item[\textbf{\ref{app:sec:sliding:The_Parthenon}}]\quad\; \textit{The Parthenon}, \citeauthor{The_Parthenon} \dotfill \pageref{app:sec:sliding:The_Parthenon}

                    \item[\textbf{\ref{app:sec:sliding:Guam_Past_and_Present}}]\quad\; \textit{Guam: Past and Present}, \citeauthor{Guam_Past_and_Present} \dotfill \pageref{app:sec:sliding:Guam_Past_and_Present}

                    \item[\textbf{\ref{app:sec:sliding:Waiting_for_Godot}}]\quad\; \textit{Waiting for Godot}, \citeauthor{Waiting_for_Godot} \dotfill \pageref{app:sec:sliding:Waiting_for_Godot}

                    \item[\textbf{\ref{app:sec:sliding:The_Lonely_Soldier}}]\quad\; \textit{The Lonely Soldier}, \citeauthor{The_Lonely_Soldier} \dotfill \pageref{app:sec:sliding:The_Lonely_Soldier}

                    \item[\textbf{\ref{app:sec:sliding:Simple_Cakes}}]\quad\; \textit{Simple Cakes}, \citeauthor{Simple_Cakes} \dotfill \pageref{app:sec:sliding:Simple_Cakes}

                    \item[\textbf{\ref{app:sec:sliding:Paradise_Valley}}]\quad\; \textit{Paradise Valley}, \citeauthor{Paradise_Valley} \dotfill \pageref{app:sec:sliding:Paradise_Valley}

                    \item[\textbf{\ref{app:sec:sliding:The_Cat_s_Pajamas}}]\quad\; \textit{The Cat's Pajamas}, \citeauthor{The_Cat_s_Pajamas} \dotfill \pageref{app:sec:sliding:The_Cat_s_Pajamas}

                    \item[\textbf{\ref{app:sec:sliding:London_in_Chains}}]\quad\; \textit{London in Chains}, \citeauthor{London_in_Chains} \dotfill \pageref{app:sec:sliding:London_in_Chains}

                    \item[\textbf{\ref{app:sec:sliding:My_Einstein}}]\quad\; \textit{My Einstein}, \citeauthor{My_Einstein} \dotfill \pageref{app:sec:sliding:My_Einstein}

                    \item[\textbf{\ref{app:sec:sliding:The_Da_Vinci_Code}}]\quad\; \textit{The Da Vinci Code}, \citeauthor{The_Da_Vinci_Code} \dotfill \pageref{app:sec:sliding:The_Da_Vinci_Code}

                    \item[\textbf{\ref{app:sec:sliding:Live_and_Learn}}]\quad\; \textit{Live and Learn}, \citeauthor{Live_and_Learn} \dotfill \pageref{app:sec:sliding:Live_and_Learn}

                    \item[\textbf{\ref{app:sec:sliding:Knowing_Your_Value}}]\quad\; \textit{Knowing Your Value}, \citeauthor{Knowing_Your_Value} \dotfill \pageref{app:sec:sliding:Knowing_Your_Value}

                    \item[\textbf{\ref{app:sec:sliding:The_Myth_of_Sisyphus}}]\quad\; \textit{The Myth of Sisyphus}, \citeauthor{The_Myth_of_Sisyphus} \dotfill \pageref{app:sec:sliding:The_Myth_of_Sisyphus}

                    \item[\textbf{\ref{app:sec:sliding:Alice_s_Adventures_in_Wonderland}}]\quad\; \textit{Alice's Adventures in Wonderland}, \citeauthor{Alice_s_Adventures_in_Wonderland} \dotfill \pageref{app:sec:sliding:Alice_s_Adventures_in_Wonderland}

                    \item[\textbf{\ref{app:sec:sliding:The_Infinity_Link}}]\quad\; \textit{The Infinity Link}, \citeauthor{The_Infinity_Link} \dotfill \pageref{app:sec:sliding:The_Infinity_Link}

                    \item[\textbf{\ref{app:sec:sliding:Murder_on_the_Orient_Express}}]\quad\; \textit{Murder on the Orient Express}, \citeauthor{Murder_on_the_Orient_Express} \dotfill \pageref{app:sec:sliding:Murder_on_the_Orient_Express}

                    \item[\textbf{\ref{app:sec:sliding:And_Then_There_Were_None}}]\quad\; \textit{And Then There Were None}, \citeauthor{And_Then_There_Were_None} \dotfill \pageref{app:sec:sliding:And_Then_There_Were_None}

                    \item[\textbf{\ref{app:sec:sliding:The_Beautiful_Struggle}}]\quad\; \textit{The Beautiful Struggle}, \citeauthor{The_Beautiful_Struggle} \dotfill \pageref{app:sec:sliding:The_Beautiful_Struggle}

                    \item[\textbf{\ref{app:sec:sliding:We_Were_Eight_Years_in_Power}}]\quad\; \textit{We Were Eight Years in Power}, \citeauthor{We_Were_Eight_Years_in_Power} \dotfill \pageref{app:sec:sliding:We_Were_Eight_Years_in_Power}

                    \item[\textbf{\ref{app:sec:sliding:The_Water_Dancer}}]\quad\;
                    \textit{The Water Dancer}, \citeauthor{The_Water_Dancer} \dotfill
                    \pageref{app:sec:sliding:The_Water_Dancer}

                    \item[\textbf{\ref{app:sec:sliding:The_Infernal_Machine}}]\quad\; \textit{The Infernal Machine}, \citeauthor{The_Infernal_Machine} \dotfill \pageref{app:sec:sliding:The_Infernal_Machine}

                    \item[\textbf{\ref{app:sec:sliding:The_Alchemist}}]\quad\; \textit{The Alchemist}, \citeauthor{The_Alchemist} \dotfill \pageref{app:sec:sliding:The_Alchemist}
                    
                    \item[\textbf{\ref{app:sec:sliding:Dungeons_and_Dragons_and_Philosophy}}]\quad\; \textit{Dungeons and Dragons and Philosophy}, \citeauthor{Dungeons_and_Dragons_and_Philosophy} \dotfill \pageref{app:sec:sliding:Dungeons_and_Dragons_and_Philosophy}

                    \item[\textbf{\ref{app:sec:sliding:Mark_Rothko}}]\quad\; \textit{Mark Rothko}, \citeauthor{Mark_Rothko} \dotfill \pageref{app:sec:sliding:Mark_Rothko}

                    \item[\textbf{\ref{app:sec:sliding:The_Hunger_Games}}]\quad\; \textit{The Hunger Games}, \citeauthor{The_Hunger_Games} \dotfill \pageref{app:sec:sliding:The_Hunger_Games}

                    \item[\textbf{\ref{app:sec:sliding:The_Dragon_Never_Sleeps}}]\quad\; \textit{The Dragon Never Sleeps}, \citeauthor{The_Dragon_Never_Sleeps} \dotfill \pageref{app:sec:sliding:The_Dragon_Never_Sleeps}
                    \item[\textbf{\ref{app:sec:sliding:The_7_Habits_of_Highly_Effective_People}}]\quad\; \textit{The 7 Habits of Highly Effective People}, \citeauthor{The_7_Habits_of_Highly_Effective_People} \dotfill \pageref{app:sec:sliding:The_7_Habits_of_Highly_Effective_People}

                    \item[\textbf{\ref{app:sec:sliding:Bad_Kid}}]\quad\; \textit{Bad Kid}, \citeauthor{Bad_Kid} \dotfill \pageref{app:sec:sliding:Bad_Kid}

                    \item[\textbf{\ref{app:sec:sliding:Lullaby_Town}}]\quad\; \textit{Lullaby Town}, \citeauthor{Lullaby_Town} \dotfill \pageref{app:sec:sliding:Lullaby_Town}

                    \item[\textbf{\ref{app:sec:sliding:Jurassic_Park}}]\quad\; \textit{Jurassic Park}, \citeauthor{Jurassic_Park} \dotfill \pageref{app:sec:sliding:Jurassic_Park}

                    \item[\textbf{\ref{app:sec:sliding:The_Hours}}]\quad\; \textit{The Hours}, \citeauthor{The_Hours} \dotfill \pageref{app:sec:sliding:The_Hours}

                    \item[\textbf{\ref{app:sec:sliding:Inhuman_Land}}]\quad\; \textit{Inhuman Land}, \citeauthor{Inhuman_Land} \dotfill \pageref{app:sec:sliding:Inhuman_Land}

                    \item[\textbf{\ref{app:sec:sliding:Charlie_and_the_Chocolate_Factory}}]\quad\; \textit{Charlie and the Chocolate Factory}, \citeauthor{Charlie_and_the_Chocolate_Factory} \dotfill \pageref{app:sec:sliding:Charlie_and_the_Chocolate_Factory}

                    \item[\textbf{\ref{app:sec:sliding:James_and_the_Giant_Peach}}]\quad\; \textit{James and the Giant Peach}, \citeauthor{James_and_the_Giant_Peach} \dotfill \pageref{app:sec:sliding:James_and_the_Giant_Peach}

                    \item[\textbf{\ref{app:sec:sliding:Automating_the_News}}]\quad\; \textit{Automating the News}, \citeauthor{Automating_the_News} \dotfill \pageref{app:sec:sliding:Automating_the_News}
                    
                    \item[\textbf{\ref{app:sec:sliding:Drown}}]\quad\; \textit{Drown}, \citeauthor{Drown} \dotfill \pageref{app:sec:sliding:Drown}

                    \item[\textbf{\ref{app:sec:sliding:The_Brief_Wondrous_Life_of_Oscar_Wao}}]\quad\; \textit{The Brief Wondrous Life of Oscar Wao}, \citeauthor{The_Brief_Wondrous_Life_of_Oscar_Wao} \dotfill \pageref{app:sec:sliding:The_Brief_Wondrous_Life_of_Oscar_Wao}

                    \item[\textbf{\ref{app:sec:sliding:This_Is_How_You_Lose_Her}}]\quad\; \textit{This Is How You Lose Her}, \citeauthor{This_Is_How_You_Lose_Her} \dotfill \pageref{app:sec:sliding:This_Is_How_You_Lose_Her}

                    \item[\textbf{\ref{app:sec:sliding:The_White_Album}}]\quad\; \textit{The White Album}, \citeauthor{The_White_Album} \dotfill \pageref{app:sec:sliding:The_White_Album}

                    \item[\textbf{\ref{app:sec:sliding:Down_and_Out_in_the_Magic_Kingdom}}]\quad\; \textit{Down and Out in the Magic Kingdom}, \citeauthor{Down_and_Out_in_the_Magic_Kingdom} \dotfill \pageref{app:sec:sliding:Down_and_Out_in_the_Magic_Kingdom}

                    \item[\textbf{\ref{app:sec:sliding:The_World_s_Wife}}]\quad\; \textit{The World's Wife}, \citeauthor{The_World_s_Wife} \dotfill \pageref{app:sec:sliding:The_World_s_Wife}

                    \item[\textbf{\ref{app:sec:sliding:A_Visit_from_the_Goon_Squad}}]\quad\; \textit{A Visit from the Goon Squad}, \citeauthor{A_Visit_from_the_Goon_Squad} \dotfill \pageref{app:sec:sliding:A_Visit_from_the_Goon_Squad}

                    \item[\textbf{\ref{app:sec:sliding:Invisible_Man}}]\quad\; \textit{Invisible Man}, \citeauthor{Invisible_Man} \dotfill \pageref{app:sec:sliding:Invisible_Man}

                    \item[\textbf{\ref{app:sec:sliding:We_Should_All_Be_Mirandas}}]\quad\; \textit{We Should All Be Mirandas}, \citeauthor{We_Should_All_Be_Mirandas} \dotfill \pageref{app:sec:sliding:We_Should_All_Be_Mirandas}

                    \item[\textbf{\ref{app:sec:sliding:The_Dude_Abides}}]\quad\; \textit{The Dude Abides}, \citeauthor{The_Dude_Abides} \dotfill \pageref{app:sec:sliding:The_Dude_Abides}
                    
                    \item[\textbf{\ref{app:sec:sliding:The_President_s_Vampire}}]\quad\; \textit{The President's Vampire}, \citeauthor{The_President_s_Vampire} \dotfill \pageref{app:sec:sliding:The_President_s_Vampire}
                    
                    \item[\textbf{\ref{app:sec:sliding:The_Great_Gatsby}}]\quad\; \textit{The Great Gatsby}, \citeauthor{The_Great_Gatsby} \dotfill \pageref{app:sec:sliding:The_Great_Gatsby}

                    \item[\textbf{\ref{app:sec:sliding:Gone_Girl}}]\quad\; \textit{Gone Girl}, \citeauthor{Gone_Girl} \dotfill \pageref{app:sec:sliding:Gone_Girl}

                    \item[\textbf{\ref{app:sec:sliding:British_Destroyers}}]\quad\; \textit{British Destroyers}, \citeauthor{British_Destroyers} \dotfill \pageref{app:sec:sliding:British_Destroyers}

                    \item[\textbf{\ref{app:sec:sliding:Florals_and_Botanicals}}]\quad\; \textit{Florals \& Botanicals}, \citeauthor{Florals_and_Botanicals} \dotfill \pageref{app:sec:sliding:Florals_and_Botanicals}

                    \item[\textbf{\ref{app:sec:sliding:Good_Omens}}]\quad\; \textit{Good Omens}, \citeauthor{Good_Omens} \dotfill \pageref{app:sec:sliding:Good_Omens}

                    \item[\textbf{\ref{app:sec:sliding:The_Slippery_Year}}]\quad\; \textit{The Slippery Year}, \citeauthor{The_Slippery_Year} \dotfill \pageref{app:sec:sliding:The_Slippery_Year}

                    \item[\textbf{\ref{app:sec:sliding:Sweater_Surgery}}]\quad\; \textit{Sweater Surgery}, \citeauthor{Sweater_Surgery} \dotfill \pageref{app:sec:sliding:Sweater_Surgery}

                    \item[\textbf{\ref{app:sec:sliding:Blink}}]\quad\; \textit{Blink}, \citeauthor{Blink} \dotfill \pageref{app:sec:sliding:Blink}

                    \item[\textbf{\ref{app:sec:sliding:The_Land_Before_Avocado}}]\quad\; \textit{The Land Before Avocado}, \citeauthor{The_Land_Before_Avocado} \dotfill \pageref{app:sec:sliding:The_Land_Before_Avocado}

                    \item[\textbf{\ref{app:sec:sliding:Dead_Ringers}}]\quad\; \textit{Dead Ringers}, \citeauthor{Dead_Ringers} \dotfill \pageref{app:sec:sliding:Dead_Ringers}

                    \item[\textbf{\ref{app:sec:sliding:Ararat}}]\quad\; \textit{Ararat}, \citeauthor{Ararat} \dotfill \pageref{app:sec:sliding:Ararat}
                    
                    \item[\textbf{\ref{app:sec:sliding:Lord_of_the_Flies}}]\quad\; \textit{Lord of the Flies}, \citeauthor{Lord_of_the_Flies} \dotfill \pageref{app:sec:sliding:Lord_of_the_Flies}

                    \item[\textbf{\ref{app:sec:sliding:Wizard_s_First_Rule}}]\quad\; \textit{Wizard's First Rule}, \citeauthor{Wizard_s_First_Rule} \dotfill \pageref{app:sec:sliding:Wizard_s_First_Rule}

                    \item[\textbf{\ref{app:sec:sliding:Rome_and_Jerusalem}}]\quad\; \textit{Rome and Jerusalem}, \citeauthor{Rome_and_Jerusalem} \dotfill \pageref{app:sec:sliding:Rome_and_Jerusalem}

                    \item[\textbf{\ref{app:sec:sliding:The_Fault_in_Our_Stars}}]\quad\; \textit{The Fault in Our Stars}, \citeauthor{The_Fault_in_Our_Stars} \dotfill \pageref{app:sec:sliding:The_Fault_in_Our_Stars}

                    \item[\textbf{\ref{app:sec:sliding:The_Third_Man}}]\quad\; \textit{The Third Man}, \citeauthor{The_Third_Man} \dotfill \pageref{app:sec:sliding:The_Third_Man}

                    \item[\textbf{\ref{app:sec:sliding:The_Confessions_of_Max_Tivoli}}]\quad\; \textit{The Confessions of Max Tivoli}, \citeauthor{The_Confessions_of_Max_Tivoli} \dotfill \pageref{app:sec:sliding:The_Confessions_of_Max_Tivoli}

                    \item[\textbf{\ref{app:sec:sliding:The_Fugitive}}]\quad\; \textit{The Fugitive}, \citeauthor{The_Fugitive} \dotfill \pageref{app:sec:sliding:The_Fugitive}

                    \item[\textbf{\ref{app:sec:sliding:The_Curious_Incident_of_the_Dog_in_the_Night-Time}}]\quad\; \textit{The Curious Incident of the Dog in the Night-Time}, \citeauthor{The_Curious_Incident_of_the_Dog_in_the_Night-Time} \dotfill \pageref{app:sec:sliding:The_Curious_Incident_of_the_Dog_in_the_Night-Time}
                    
                    \item[\textbf{\ref{app:sec:sliding:Migrations_to_Solitude}}]\quad\; \textit{Migrations to Solitude}, \citeauthor{Migrations_to_Solitude} \dotfill \pageref{app:sec:sliding:Migrations_to_Solitude}
                    
                    \item[\textbf{\ref{app:sec:sliding:Uncommon_Type}}]\quad\; \textit{Uncommon Type}, \citeauthor{Uncommon_Type} \dotfill \pageref{app:sec:sliding:Uncommon_Type}

                    \item[\textbf{\ref{app:sec:sliding:Buzz}}]\quad\; \textit{Buzz}, \citeauthor{Buzz} \dotfill \pageref{app:sec:sliding:Buzz}

                    \item[\textbf{\ref{app:sec:sliding:Requiem_for_the_Sun}}]\quad\; \textit{Requiem for the Sun}, \citeauthor{Requiem_for_the_Sun} \dotfill \pageref{app:sec:sliding:Requiem_for_the_Sun}

                    \item[\textbf{\ref{app:sec:sliding:Catch-22}}]\quad\; \textit{Catch-22}, \citeauthor{Catch-22} \dotfill \pageref{app:sec:sliding:Catch-22}

                    \item[\textbf{\ref{app:sec:sliding:The_Old_Man_and_the_Sea}}]\quad\; \textit{The Old Man and the Sea}, \citeauthor{The_Old_Man_and_the_Sea} \dotfill \pageref{app:sec:sliding:The_Old_Man_and_the_Sea}

                    \item[\textbf{\ref{app:sec:sliding:Life_on_Air}}]\quad\; \textit{Life on Air}, \citeauthor{Life_on_Air} \dotfill \pageref{app:sec:sliding:Life_on_Air}

                    \item[\textbf{\ref{app:sec:sliding:Great_Hair_Days}}]\quad\; \textit{Great Hair Days}, \citeauthor{Great_Hair_Days} \dotfill \pageref{app:sec:sliding:Great_Hair_Days}

                    \item[\textbf{\ref{app:sec:sliding:Graft}}]\quad\; \textit{Graft}, \citeauthor{Graft} \dotfill \pageref{app:sec:sliding:Graft}

                    \item[\textbf{\ref{app:sec:sliding:The_Outsiders}}]\quad\; \textit{The Outsiders}, \citeauthor{The_Outsiders} \dotfill \pageref{app:sec:sliding:The_Outsiders}

                    \item[\textbf{\ref{app:sec:sliding:The_Second_Summoning}}]\quad\; \textit{The Second Summoning}, \citeauthor{The_Second_Summoning} \dotfill \pageref{app:sec:sliding:The_Second_Summoning}

                    \item[\textbf{\ref{app:sec:sliding:Selected_Poems_of_Langston_Hughes}}]\quad\; \textit{Selected Poems of Langston Hughes}, \citeauthor{Selected_Poems_of_Langston_Hughes} \dotfill \pageref{app:sec:sliding:Selected_Poems_of_Langston_Hughes}

                    \item[\textbf{\ref{app:sec:sliding:M_Butterfly}}]\quad\; \textit{M. Butterfly}, \citeauthor{M_Butterfly} \dotfill \pageref{app:sec:sliding:M_Butterfly}
                    \item[\textbf{\ref{app:sec:sliding:Building_and_Operating_a_Realistic_Model_Railway}}]\quad\; \textit{Building and Operating a Realistic Model Railway}, \citeauthor{Building_and_Operating_a_Realistic_Model_Railway} \dotfill \pageref{app:sec:sliding:Building_and_Operating_a_Realistic_Model_Railway}

                    \item[\textbf{\ref{app:sec:sliding:All_the_Onions}}]\quad\; \textit{All the Onions}, \citeauthor{All_the_Onions} \dotfill \pageref{app:sec:sliding:All_the_Onions}

                    \item[\textbf{\ref{app:sec:sliding:Fifty_Shades_of_Grey}}]\quad\; \textit{Fifty Shades of Grey}, \citeauthor{Fifty_Shades_of_Grey} \dotfill \pageref{app:sec:sliding:Fifty_Shades_of_Grey}
                    
                    \item[\textbf{\ref{app:sec:sliding:The_Stone_Sky}}]\quad\; \textit{The Stone Sky}, \citeauthor{The_Stone_Sky} \dotfill \pageref{app:sec:sliding:The_Stone_Sky}

                    \item[\textbf{\ref{app:sec:sliding:Ulysses}}]\quad\; \textit{Ulysses}, \citeauthor{Ulysses} \dotfill \pageref{app:sec:sliding:Ulysses}

                    \item[\textbf{\ref{app:sec:sliding:Sandman_Slim}}]\quad\; \textit{Sandman Slim}, \citeauthor{Sandman_Slim} \dotfill \pageref{app:sec:sliding:Sandman_Slim}

                    \item[\textbf{\ref{app:sec:sliding:Ethnography_after_Antiquity}}]\quad\; \textit{Ethnography after Antiquity}, \citeauthor{Ethnography_after_Antiquity} \dotfill \pageref{app:sec:sliding:Ethnography_after_Antiquity}

                    \item[\textbf{\ref{app:sec:sliding:Who_Is_Rich}}]\quad \textit{Who Is Rich?}, \citeauthor{Who_Is_Rich} \dotfill \pageref{app:sec:sliding:Who_Is_Rich}

                    \item[\textbf{\ref{app:sec:sliding:The_Servants_of_Twilight}}]\quad \textit{The Servants of Twilight}, \citeauthor{The_Servants_of_Twilight} \dotfill \pageref{app:sec:sliding:The_Servants_of_Twilight}

                    \item[\textbf{\ref{app:sec:sliding:Tai_Chi_for_Depression}}]\quad \textit{Tai Chi for Depression}, \citeauthor{Tai_Chi_for_Depression} \dotfill \pageref{app:sec:sliding:Tai_Chi_for_Depression}

                    \item[\textbf{\ref{app:sec:sliding:The_Tide_Was_Always_High}}]\quad \textit{The Tide Was Always High}, \citeauthor{The_Tide_Was_Always_High} \dotfill \pageref{app:sec:sliding:The_Tide_Was_Always_High}
                    
                    \item[\textbf{\ref{app:sec:sliding:Girl_in_Translation}}]\quad \textit{Girl in Translation}, \citeauthor{Girl_in_Translation} \dotfill \pageref{app:sec:sliding:Girl_in_Translation}

                    \item[\textbf{\ref{app:sec:sliding:A_Wrinkle_in_Time}}]\quad \textit{A Wrinkle in Time}, \citeauthor{A_Wrinkle_in_Time} \dotfill \pageref{app:sec:sliding:A_Wrinkle_in_Time}

                    \item[\textbf{\ref{app:sec:sliding:Call_Me_Brooklyn}}]\quad \textit{Call Me Brooklyn}, \citeauthor{Call_Me_Brooklyn} \dotfill \pageref{app:sec:sliding:Call_Me_Brooklyn}

                    \item[\textbf{\ref{app:sec:sliding:Dead_Wake}}]\quad \textit{Dead Wake}, \citeauthor{Dead_Wake} \dotfill \pageref{app:sec:sliding:Dead_Wake}

                    \item[\textbf{\ref{app:sec:sliding:The_Girl_with_the_Dragon_Tattoo}}]\quad \textit{The Girl with the Dragon Tattoo}, \citeauthor{The_Girl_with_the_Dragon_Tattoo} \dotfill \pageref{app:sec:sliding:The_Girl_with_the_Dragon_Tattoo}

                    \item[\textbf{\ref{app:sec:sliding:The_Daughter_of_Odren}}]\quad \textit{The Daughter of Odren}, \citeauthor{The_Daughter_of_Odren} \dotfill \pageref{app:sec:sliding:The_Daughter_of_Odren}

                    \item[\textbf{\ref{app:sec:sliding:The_Chronicles_of_Narnia}}]\quad \textit{The Chronicles of Narnia}, \citeauthor{The_Chronicles_of_Narnia} \dotfill \pageref{app:sec:sliding:The_Chronicles_of_Narnia}

                    \item[\textbf{\ref{app:sec:sliding:After_I_m_Gone}}]\quad \textit{After I'm Gone}, \citeauthor{After_I_m_Gone} \dotfill \pageref{app:sec:sliding:After_I_m_Gone}

                    \item[\textbf{\ref{app:sec:sliding:Sunburn}}]\quad \textit{Sunburn}, \citeauthor{Sunburn} \dotfill \pageref{app:sec:sliding:Sunburn}

                    \item[\textbf{\ref{app:sec:sliding:Marvel_s_Spider-Man_Hostile_Takeover}}]\quad \textit{Marvel's Spider-Man: Hostile Takeover}, \citeauthor{Marvel_s_Spider-Man_Hostile_Takeover} \dotfill \pageref{app:sec:sliding:Marvel_s_Spider-Man_Hostile_Takeover}
                    
                    \item[\textbf{\ref{app:sec:sliding:Anastasia_on_Her_Own}}]\quad \textit{Anastasia on Her Own}, \citeauthor{Anastasia_on_Her_Own} \dotfill \pageref{app:sec:sliding:Anastasia_on_Her_Own}

                    \item[\textbf{\ref{app:sec:sliding:Nixon_in_China}}]\quad \textit{Nixon in China}, \citeauthor{Nixon_in_China} \dotfill \pageref{app:sec:sliding:Nixon_in_China}

                    \item[\textbf{\ref{app:sec:sliding:The_Doomsday_Prophecy}}]\quad \textit{The Doomsday Prophecy}, \citeauthor{The_Doomsday_Prophecy} \dotfill \pageref{app:sec:sliding:The_Doomsday_Prophecy}

                    \item[\textbf{\ref{app:sec:sliding:A_Game_of_Thrones}}]\quad \textit{A Game of Thrones}, \citeauthor{A_Game_of_Thrones} \dotfill \pageref{app:sec:sliding:A_Game_of_Thrones}

                    \item[\textbf{\ref{app:sec:sliding:Rough-Hewn_Land}}]\quad \textit{Rough-Hewn Land}, \citeauthor{Rough-Hewn_Land} \dotfill \pageref{app:sec:sliding:Rough-Hewn_Land}

                    \item[\textbf{\ref{app:sec:sliding:Twilight}}]\quad \textit{Twilight}, \citeauthor{Twilight} \dotfill \pageref{app:sec:sliding:Twilight}

                    \item[\textbf{\ref{app:sec:sliding:The_Duchess_War}}]\quad \textit{The Duchess War}, \citeauthor{The_Duchess_War} \dotfill \pageref{app:sec:sliding:The_Duchess_War}

                    \item[\textbf{\ref{app:sec:sliding:On_Liberty}}]\quad \textit{On Liberty}, \citeauthor{On_Liberty} \dotfill \pageref{app:sec:sliding:On_Liberty}

                    \item[\textbf{\ref{app:sec:sliding:Coal_Creek}}]\quad \textit{Coal Creek}, \citeauthor{Coal_Creek} \dotfill \pageref{app:sec:sliding:Coal_Creek}

                    \item[\textbf{\ref{app:sec:sliding:Winnie_the_Pooh}}]\quad \textit{Winnie the Pooh}, \citeauthor{Winnie_the_Pooh} \dotfill \pageref{app:sec:sliding:Winnie_the_Pooh}

                    \item[\textbf{\ref{app:sec:sliding:Catching_the_Sky}}]\quad \textit{Catching the Sky}, \citeauthor{Catching_the_Sky} \dotfill \pageref{app:sec:sliding:Catching_the_Sky}

                    \item[\textbf{\ref{app:sec:sliding:The_Heretic_Queen}}]\quad \textit{The Heretic Queen}, \citeauthor{The_Heretic_Queen} \dotfill \pageref{app:sec:sliding:The_Heretic_Queen}

                    \item[\textbf{\ref{app:sec:sliding:The_Gondola_Maker}}]\quad \textit{The Gondola Maker}, \citeauthor{The_Gondola_Maker} \dotfill \pageref{app:sec:sliding:The_Gondola_Maker}

                    \item[\textbf{\ref{app:sec:sliding:Songs_in_Ordinary_Time}}]\quad \textit{Songs in Ordinary Time}, \citeauthor{Songs_in_Ordinary_Time} \dotfill \pageref{app:sec:sliding:Songs_in_Ordinary_Time}
                    
                    \item[\textbf{\ref{app:sec:sliding:Beloved}}]\quad \textit{Beloved}, \citeauthor{Beloved} \dotfill \pageref{app:sec:sliding:Beloved}

                    \item[\textbf{\ref{app:sec:sliding:Norwegian_Wood}}]\quad \textit{Norwegian Wood}, \citeauthor{Norwegian_Wood} \dotfill \pageref{app:sec:sliding:Norwegian_Wood}

                    \item[\textbf{\ref{app:sec:sliding:Eat_More_Plants}}]\quad \textit{Eat More Plants}, \citeauthor{Eat_More_Plants} \dotfill \pageref{app:sec:sliding:Eat_More_Plants}

                    \item[\textbf{\ref{app:sec:sliding:Polaris}}]\quad \textit{Polaris}, \citeauthor{Polaris} \dotfill \pageref{app:sec:sliding:Polaris}

                    \item[\textbf{\ref{app:sec:sliding:Pagans}}]\quad \textit{Pagans}, \citeauthor{Pagans} \dotfill \pageref{app:sec:sliding:Pagans}

                    \item[\textbf{\ref{app:sec:sliding:Windfall}}]\quad \textit{Windfall}, \citeauthor{Windfall} \dotfill \pageref{app:sec:sliding:Windfall}

                    \item[\textbf{\ref{app:sec:sliding:The_Memory_Police}}]\quad \textit{The Memory Police}, \citeauthor{The_Memory_Police} \dotfill \pageref{app:sec:sliding:The_Memory_Police}

                    \item[\textbf{\ref{app:sec:sliding:Winter_Sisters}}]\quad \textit{Winter Sisters}, \citeauthor{Winter_Sisters} \dotfill \pageref{app:sec:sliding:Winter_Sisters}
                    
                    \item[\textbf{\ref{app:sec:sliding:Nineteen_Eighty-Four}}]\quad \textit{Nineteen Eighty Four}, \citeauthor{Nineteen_Eighty-Four} \dotfill \pageref{app:sec:sliding:Nineteen_Eighty-Four}

                    \item[\textbf{\ref{app:sec:sliding:Fight_Club}}]\quad \textit{Fight Club}, \citeauthor{Fight_Club} \dotfill \pageref{app:sec:sliding:Fight_Club}

                    \item[\textbf{\ref{app:sec:sliding:The_Complete_Joy_of_Homebrewing}}]\quad \textit{The Complete Joy of Homebrewing}, \citeauthor{The_Complete_Joy_of_Homebrewing} \dotfill \pageref{app:sec:sliding:The_Complete_Joy_of_Homebrewing}

                    \item[\textbf{\ref{app:sec:sliding:The_Cult_of_Loving_Kindness}}]\quad \textit{The Cult of Loving Kindness}, \citeauthor{The_Cult_of_Loving_Kindness} \dotfill \pageref{app:sec:sliding:The_Cult_of_Loving_Kindness}
                    
                    \item[\textbf{\ref{app:sec:sliding:Along_Came_a_Spider}}]\quad \textit{Along Came a Spider}, \citeauthor{Along_Came_a_Spider} \dotfill \pageref{app:sec:sliding:Along_Came_a_Spider}

                    \item[\textbf{\ref{app:sec:sliding:Payard_Cookies}}]\quad \textit{Payard Cookies}, \citeauthor{Payard_Cookies} \dotfill \pageref{app:sec:sliding:Payard_Cookies}

                    \item[\textbf{\ref{app:sec:sliding:Essential_Pepin_Desserts}}]\quad \textit{Essential Pepin Desserts}, \citeauthor{Essential_Pepin_Desserts} \dotfill \pageref{app:sec:sliding:Essential_Pepin_Desserts}

                    \item[\textbf{\ref{app:sec:sliding:Why_New_Orleans_Matters}}]\quad \textit{Why New Orleans Matters}, \citeauthor{Why_New_Orleans_Matters} \dotfill \pageref{app:sec:sliding:Why_New_Orleans_Matters}

                    \item[\textbf{\ref{app:sec:sliding:Enlightenment_Now}}]\quad \textit{Enlightenment Now}, \citeauthor{Enlightenment_Now} \dotfill \pageref{app:sec:sliding:Enlightenment_Now}

                    \item[\textbf{\ref{app:sec:sliding:Competitive_Strategy}}]\quad \textit{Competitive Strategy}, \citeauthor{Competitive_Strategy} \dotfill \pageref{app:sec:sliding:Competitive_Strategy}

                    \item[\textbf{\ref{app:sec:sliding:Night_Watch}}]\quad \textit{Night Watch}, \citeauthor{Night_Watch} \dotfill \pageref{app:sec:sliding:Night_Watch}

                    \item[\textbf{\ref{app:sec:sliding:The_Subtle_Knife}}]\quad \textit{The Subtle Knife}, \citeauthor{The_Subtle_Knife} \dotfill \pageref{app:sec:sliding:The_Subtle_Knife}

                    \item[\textbf{\ref{app:sec:sliding:The_Seductions_of_Darwin}}]\quad \textit{The Seductions of Darwin}, \citeauthor{The_Seductions_of_Darwin} \dotfill \pageref{app:sec:sliding:The_Seductions_of_Darwin}

                    \item[\textbf{\ref{app:sec:sliding:Kitchen_Table_Wisdom}}]\quad \textit{Kitchen Table Wisdom}, \citeauthor{Kitchen_Table_Wisdom} \dotfill \pageref{app:sec:sliding:Kitchen_Table_Wisdom}

                    \item[\textbf{\ref{app:sec:sliding:Backroads_Boss_Lady}}]\quad \textit{Backroads Boss Lady}, \citeauthor{Backroads_Boss_Lady} \dotfill \pageref{app:sec:sliding:Backroads_Boss_Lady}

                    \item[\textbf{\ref{app:sec:sliding:Soft_in_the_Head}}]\quad \textit{Soft in the Head}, \citeauthor{Soft_in_the_Head} \dotfill \pageref{app:sec:sliding:Soft_in_the_Head}

                    \item[\textbf{\ref{app:sec:sliding:The_Making_of_a_Mediterranean_Emirate}}]\quad \textit{The Making of a Mediterranean Emirate}, \citeauthor{The_Making_of_a_Mediterranean_Emirate} \dotfill \pageref{app:sec:sliding:The_Making_of_a_Mediterranean_Emirate}
                    
                    \item[\textbf{\ref{app:sec:sliding:Harry_Potter_and_the_Sorcerer_s_Stone}}]\quad \textit{Harry Potter and the Sorcerer's Stone}, \citeauthor{Harry_Potter_and_the_Sorcerer_s_Stone} \dotfill \pageref{app:sec:sliding:Harry_Potter_and_the_Sorcerer_s_Stone}
                    
                    \item[\textbf{\ref{app:sec:sliding:Harry_Potter_and_the_Chamber_of_Secrets}}]\quad \textit{Harry Potter and the Chamber of Secrets}, \citeauthor{Harry_Potter_and_the_Chamber_of_Secrets} \dotfill \pageref{app:sec:sliding:Harry_Potter_and_the_Chamber_of_Secrets}

                    \item[\textbf{\ref{app:sec:sliding:Harry_Potter_and_the_Goblet_of_Fire}}]\quad \textit{Harry Potter and the Goblet of Fire}, \citeauthor{Harry_Potter_and_the_Goblet_of_Fire} \dotfill \pageref{app:sec:sliding:Harry_Potter_and_the_Goblet_of_Fire}

                    \item[\textbf{\ref{app:sec:sliding:Harry_Potter_and_the_Deathly_Hallows}}]\quad \textit{Harry Potter and the Deathly Hallows}, \citeauthor{Harry_Potter_and_the_Deathly_Hallows} \dotfill \pageref{app:sec:sliding:Harry_Potter_and_the_Deathly_Hallows}

                    \item[\textbf{\ref{app:sec:sliding:Born_to_Walk}}]\quad \textit{Born to Walk}, \citeauthor{Born_to_Walk} \dotfill \pageref{app:sec:sliding:Born_to_Walk}

                    \item[\textbf{\ref{app:sec:sliding:The_Pretender}}]\quad \textit{The Pretender}, \citeauthor{The_Pretender} \dotfill \pageref{app:sec:sliding:The_Pretender}

                    \item[\textbf{\ref{app:sec:sliding:Toscanini}}]\quad \textit{Toscanini}, \citeauthor{Toscanini} \dotfill \pageref{app:sec:sliding:Toscanini}

                    \item[\textbf{\ref{app:sec:sliding:Cosmos}}]\quad \textit{Cosmos}, \citeauthor{Cosmos} \dotfill \pageref{app:sec:sliding:Cosmos}
                    
                    \item[\textbf{\ref{app:sec:sliding:Middle_India}}]\quad \textit{Middle India}, \citeauthor{Middle_India} \dotfill \pageref{app:sec:sliding:Middle_India}
                    
                    \item[\textbf{\ref{app:sec:sliding:The_Catcher_in_the_Rye}}]\quad \textit{The Catcher in the Rye}, \citeauthor{The_Catcher_in_the_Rye} \dotfill \pageref{app:sec:sliding:The_Catcher_in_the_Rye}

                    \item[\textbf{\ref{app:sec:sliding:Lean_In}}]\quad \textit{Lean In}, \citeauthor{Lean_In} \dotfill \pageref{app:sec:sliding:Lean_In}

                    \item[\textbf{\ref{app:sec:sliding:The_DevOps_Adoption_Playbook}}]\quad \textit{The DevOps Adoption Playbook}, \citeauthor{The_DevOps_Adoption_Playbook} \dotfill \pageref{app:sec:sliding:The_DevOps_Adoption_Playbook}

                    \item[\textbf{\ref{app:sec:sliding:Frankenstein}}]\quad \textit{Frankenstein}, \citeauthor{Frankenstein} \dotfill \pageref{app:sec:sliding:Frankenstein}
                    \item[\textbf{\ref{app:sec:sliding:Sally_Ride_America_s_First_Woman_in_Space}}]\quad \textit{Sally Ride: America's First Woman in Space}, \citeauthor{Sally_Ride_America_s_First_Woman_in_Space} \dotfill \pageref{app:sec:sliding:Sally_Ride_America_s_First_Woman_in_Space}

                    \item[\textbf{\ref{app:sec:sliding:A_Perfectly_Good_Family}}]\quad \textit{A Perfectly Good Family}, \citeauthor{A_Perfectly_Good_Family} \dotfill \pageref{app:sec:sliding:A_Perfectly_Good_Family}

                    \item[\textbf{\ref{app:sec:sliding:The_Bedwetter}}]\quad \textit{The Bedwetter}, \citeauthor{The_Bedwetter} \dotfill \pageref{app:sec:sliding:The_Bedwetter}

                    \item[\textbf{\ref{app:sec:sliding:On_the_Road_with_Bob_Dylan}}]\quad \textit{On the Road with Bob Dylan}, \citeauthor{On_the_Road_with_Bob_Dylan} \dotfill \pageref{app:sec:sliding:On_the_Road_with_Bob_Dylan}

                    \item[\textbf{\ref{app:sec:sliding:The_Night_Children}}]\quad \textit{The Night Children}, \citeauthor{The_Night_Children} \dotfill \pageref{app:sec:sliding:The_Night_Children}

                    \item[\textbf{\ref{app:sec:sliding:White_Teeth}}]\quad \textit{White Teeth}, \citeauthor{White_Teeth} \dotfill \pageref{app:sec:sliding:White_Teeth}

                    \item[\textbf{\ref{app:sec:sliding:No_Visible_Bruises}}]\quad \textit{No Visible Bruises}, \citeauthor{No_Visible_Bruises} \dotfill \pageref{app:sec:sliding:No_Visible_Bruises}

                    \item[\textbf{\ref{app:sec:sliding:The_Grapes_of_Wrath}}]\quad \textit{The Grapes of Wrath}, \citeauthor{The_Grapes_of_Wrath} \dotfill \pageref{app:sec:sliding:The_Grapes_of_Wrath}
                    
                    \item[\textbf{\ref{app:sec:sliding:Jesse_James}}]\quad \textit{Jesse James}, \citeauthor{Jesse_James} \dotfill \pageref{app:sec:sliding:Jesse_James}

                    \item[\textbf{\ref{app:sec:sliding:Zombie_Halloween}}]\quad \textit{Zombie Halloween}, \citeauthor{Zombie_Halloween} \dotfill \pageref{app:sec:sliding:Zombie_Halloween}
                    
                    \item[\textbf{\ref{app:sec:sliding:Fear_of_Music}}]\quad \textit{Fear of Music}, \citeauthor{Fear_of_Music} \dotfill \pageref{app:sec:sliding:Fear_of_Music}

                    \item[\textbf{\ref{app:sec:sliding:Pearl_Harbor}}]\quad \textit{Pearl Harbor}, \citeauthor{Pearl_Harbor} \dotfill \pageref{app:sec:sliding:Pearl_Harbor}
                    
                    \item[\textbf{\ref{app:sec:sliding:Chesapeake_Requiem}}]\quad \textit{Chesapeake Requiem}, \citeauthor{Chesapeake_Requiem} \dotfill \pageref{app:sec:sliding:Chesapeake_Requiem}
                    
                    \item[\textbf{\ref{app:sec:sliding:The_Goldfinch}}]\quad \textit{The Goldfinch}, \citeauthor{The_Goldfinch} \dotfill \pageref{app:sec:sliding:The_Goldfinch}
                    
                    \item[\textbf{\ref{app:sec:sliding:Unglued}}]\quad \textit{Unglued}, \citeauthor{Unglued} \dotfill \pageref{app:sec:sliding:Unglued}
                    
                    \item[\textbf{\ref{app:sec:sliding:Embraced}}]\quad \textit{Embraced}, \citeauthor{Embraced} \dotfill \pageref{app:sec:sliding:Embraced}

                    \item[\textbf{\ref{app:sec:sliding:Birding_with_Yeats}}]\quad \textit{Birding with Yeats}, \citeauthor{Birding_with_Yeats} \dotfill \pageref{app:sec:sliding:Birding_with_Yeats}

                    \item[\textbf{\ref{app:sec:sliding:The_Hobbit}}]\quad \textit{The Hobbit}, \citeauthor{The_Hobbit} \dotfill \pageref{app:sec:sliding:The_Hobbit}

                    \item[\textbf{\ref{app:sec:sliding:The_Fellowship_of_the_Ring}}]\quad \textit{The Fellowship of the Ring}, \citeauthor{The_Fellowship_of_the_Ring} \dotfill \pageref{app:sec:sliding:The_Fellowship_of_the_Ring}
                    
                    \item[\textbf{\ref{app:sec:sliding:Tree_and_Leaf}}]\quad \textit{Tree and Leaf}, \citeauthor{Tree_and_Leaf} \dotfill \pageref{app:sec:sliding:Tree_and_Leaf}

                    \item[\textbf{\ref{app:sec:sliding:Noodles_Every_Day}}]\quad \textit{Noodles Every Day}, \citeauthor{Noodles_Every_Day} \dotfill \pageref{app:sec:sliding:Noodles_Every_Day}

                    \item[\textbf{\ref{app:sec:sliding:Billionaire_Democracy}}]\quad \textit{Billionaire Democracy}, \citeauthor{Billionaire_Democracy} \dotfill \pageref{app:sec:sliding:Billionaire_Democracy}

                    \item[\textbf{\ref{app:sec:sliding:Portugal_s_Guerrilla_Wars_in_Africa}}]\quad \textit{Portugal's Guerrilla Wars in Africa}, \citeauthor{Portugal_s_Guerrilla_Wars_in_Africa} \dotfill \pageref{app:sec:sliding:Portugal_s_Guerrilla_Wars_in_Africa}

                    \item[\textbf{\ref{app:sec:sliding:Slaughterhouse-Five}}]\quad \textit{Slaughterhouse-Five}, \citeauthor{Slaughterhouse-Five} \dotfill \pageref{app:sec:sliding:Slaughterhouse-Five}
                    
                    \item[\textbf{\ref{app:sec:sliding:Animal_Rights}}]\quad \textit{Animal Rights}, \citeauthor{Animal_Rights} \dotfill \pageref{app:sec:sliding:Animal_Rights}
                    
                    \item[\textbf{\ref{app:sec:sliding:Men_We_Reaped}}]\quad \textit{Men We Reaped}, \citeauthor{Men_We_Reaped} \dotfill \pageref{app:sec:sliding:Men_We_Reaped}
                    
                    \item[\textbf{\ref{app:sec:sliding:Charlotte_s_Web}}]\quad \textit{Charlotte's Web}, \citeauthor{Charlotte_s_Web} \dotfill \pageref{app:sec:sliding:Charlotte_s_Web}
                    
                    \item[\textbf{\ref{app:sec:sliding:A_Return_to_Love}}]\quad \textit{A Return to Love}, \citeauthor{A_Return_to_Love} \dotfill \pageref{app:sec:sliding:A_Return_to_Love}

                    \item[\textbf{\ref{app:sec:sliding:Another_Brooklyn}}]\quad \textit{Another Brooklyn}, \citeauthor{Another_Brooklyn} \dotfill \pageref{app:sec:sliding:Another_Brooklyn}

                    \item[\textbf{\ref{app:sec:sliding:Brown_Girl_Dreaming}}]\quad \textit{Brown Girl Dreaming}, \citeauthor{Brown_Girl_Dreaming} \dotfill \pageref{app:sec:sliding:Brown_Girl_Dreaming}
                    
                    \item[\textbf{\ref{app:sec:sliding:A_Little_Life}}]\quad \textit{A Little Life}, \citeauthor{A_Little_Life} \dotfill \pageref{app:sec:sliding:A_Little_Life}

                    \item[\textbf{\ref{app:sec:sliding:The_Art_of_Bonsai}}]\quad \textit{The Art of Bonsai}, \citeauthor{The_Art_of_Bonsai} \dotfill \pageref{app:sec:sliding:The_Art_of_Bonsai}

                    \item[\textbf{\ref{app:sec:sliding:A_People_s_History_of_the_United_States}}]\quad \textit{A People's History of the United States}, \citeauthor{A_People_s_History_of_the_United_States} \dotfill \pageref{app:sec:sliding:A_People_s_History_of_the_United_States}

                    \item[\textbf{\ref{app:sec:sliding:The_Future_of_the_Internet_and_How_to_Stop_It}}]\quad \textit{The Future of the Internet and How to Stop It}, \citeauthor{The_Future_of_the_Internet_and_How_to_Stop_It} \dotfill \pageref{app:sec:sliding:The_Future_of_the_Internet_and_How_to_Stop_It}

                    \item[\textbf{\ref{app:sec:sliding:The_Book_Thief}}]\quad \textit{The Book Thief}, \citeauthor{The_Book_Thief} \dotfill \pageref{app:sec:sliding:The_Book_Thief}
        \end{itemize}

            \item[\textbf{\ref{app:sec:sliding-window:percentage}}] Estimating how much of a book is memorized by a model \dotfill \pageref{app:sec:sliding-window:percentage}
            \begin{itemize}[labelwidth=!, labelsep=1em, leftmargin=*, align=left]
                \item[\textbf{\ref{app:sec:coverage}}]\quad \; Quantifying extraction coverage \dotfill \pageref{app:sec:coverage}
                \item[\textbf{\ref{app:sec:coverage}}]\quad \; Extended extraction coverage results\dotfill \pageref{app:sec:coverage-results}
            \end{itemize}            
        \end{itemize}
    \item[\textbf{\ref{app:sec:validity}}]\textbf{Measurement validity} \dotfill \pageref{app:sec:validity}
    \begin{itemize}[labelwidth=!, labelsep=1em, leftmargin=*, align=left]
        \item[\textbf{\ref{app:sec:validity:motivations}}] \quad Motivating our validity experiments \dotfill \pageref{app:sec:validity:motivations}
        \begin{itemize}[labelwidth=!, labelsep=1em, leftmargin=*, align=left]
                \item[\textbf{\ref{app:sec:validity:motivations:extend}}]\quad\; Setting $\tau_\text{min}$ and extending to models with unknown training data \dotfill \pageref{app:sec:validity:motivations:extend}
                \item[\textbf{\ref{app:sec:validity:motivations:lit}}]\quad\; Relating greedy-decoded and probabilistic discoverable extraction \dotfill \pageref{app:sec:validity:motivations:lit}

                \item[\textbf{\ref{app:sec:validity:motivations:vary}}]\quad\; The role of prefix length in our experiments \dotfill \pageref{app:sec:validity:motivations:vary}
        \end{itemize}
        
        \item[\textbf{\ref{app:sec:validity:setup}}]\quad Setup \dotfill \pageref{app:sec:validity:setup}

        \item[\textbf{\ref{app:sec:validity:baseline}}]\quad Results of varying prefix length on known training data \dotfill \pageref{app:sec:validity:baseline}
        
        \item[\textbf{\ref{app:sec:validity:controls}}]\quad Results of negative control experiments \dotfill \pageref{app:sec:validity:controls}
         \begin{itemize}[labelwidth=!, labelsep=1em, leftmargin=*, align=left]
                \item[\textbf{\ref{app:sec:validity:phi4}}]\quad\; Results on \textsc{Phi 4} \dotfill \pageref{app:sec:validity:phi4}

                \item[\textbf{\ref{app:sec:validity:cutoff}}]\quad\; Results on books from after training cutoffs  \dotfill \pageref{app:sec:validity:cutoff}
        \end{itemize}

        \item[\textbf{\ref{app:sec:validity:monkey}}]\quad Additional notes on ``the monkey at the typewriter'' metaphor \dotfill \pageref{app:sec:validity:monkey}

        \item[\textbf{\ref{app:sec:validity:membership}}]\quad Additional notes on extraction and membership inference \dotfill \pageref{app:sec:validity:membership}
        \begin{itemize}[labelwidth=!, labelsep=1em, leftmargin=*, align=left]
                \item[\textbf{\ref{app:sec:validity:membership:inference}}]\quad\; The relationship between membership inference and extraction \dotfill \pageref{app:sec:validity:membership:inference}

                \item[\textbf{\ref{app:sec:validity:membership:known}}]\quad\; We know ground-truth membership for most of our experiments \dotfill \pageref{app:sec:validity:membership:known}

                \item[\textbf{\ref{app:sec:validity:membership:unknown}}]\quad\; Inferring membership for LLMs with undisclosed training data \dotfill \pageref{app:sec:validity:membership:unknown}

                \item[\textbf{\ref{app:sec:validity:membership:sequence}}]\quad\; We infer  membership for sequences, not whole books \dotfill \pageref{app:sec:validity:membership:sequence}

                \item[\textbf{\ref{app:sec:validity:membership:nonmember}}]\quad\; Absence of extraction does not imply anything about membership \dotfill \pageref{app:sec:validity:membership:nonmember}

                \item[\textbf{\ref{app:sec:validity:membership:copyright}}]\quad\; Extractability and copyright \dotfill \pageref{app:sec:validity:membership:copyright}
        \end{itemize}
      \end{itemize}

    \item[\textbf{\ref{app:sec:reconstruct}}]  \textbf{Recovering books with one seed prompt} \dotfill \pageref{app:sec:reconstruct}
    \begin{itemize}[labelwidth=!, labelsep=1em, leftmargin=*, align=left]
        \item[\textbf{\ref{app:sec:reconstruct:method}}] \quad Recovery method \dotfill \pageref{app:sec:reconstruct:method}

        \item[\textbf{\ref{app:sec:reconstruct:partial}}] \quad Partial recovery \dotfill \pageref{app:sec:reconstruct:partial}
        \item[\textbf{\ref{app:sec:reconstruct:fail}}]\quad Running recovery for books with low memorization signal \dotfill \pageref{app:sec:reconstruct:fail}
      \end{itemize}
      
    \item[\textbf{\ref{app:sec:book-level-discussion}}]  \textbf{Discussion of extended results} \dotfill \pageref{app:sec:book-level-discussion}
\end{itemize}

\definecolor{codegreen}{rgb}{0,0.6,0}
\definecolor{codegray}{rgb}{0.5,0.5,0.5}
\definecolor{codepurple}{rgb}{0.58,0,0.82}
\definecolor{backcolour}{rgb}{0.95,0.95,0.92}

\lstdefinestyle{mystyle}{
    backgroundcolor=\color{backcolour},   
    commentstyle=\color{codegreen},
    keywordstyle=\color{magenta},
    numberstyle=\tiny\color{codegray},
    stringstyle=\color{codepurple},
    basicstyle=\ttfamily\scriptsize,
    breakatwhitespace=false,         
    breaklines=true,                 
    captionpos=b,                    
    keepspaces=true,                 
    numbers=left,                    
    numbersep=5pt,                  
    showspaces=false,                
    showstringspaces=false,
    showtabs=false,                  
    tabsize=2
}

\clearpage
\section{Additional notes on memorization and probabilistic extraction}\label{app:sec:background}

We provide additional notes on the background material in this paper, including the ``careless people'' quote in Section~\ref{sec:background} (Appendix~\ref{app:sec:intro}), the key probabilistic extraction metric in this paper (formally called $(n,p)$-discoverable extraction, Appendix~\ref{app:sec:background:metrics}), as well as our implementation  for this metric (Appendix~\ref{app:sec:background:compute}).
Please also refer to Section~\ref{sec:background}.

\begin{figure*}[t!]
\centering
\includegraphics[width=\textwidth]{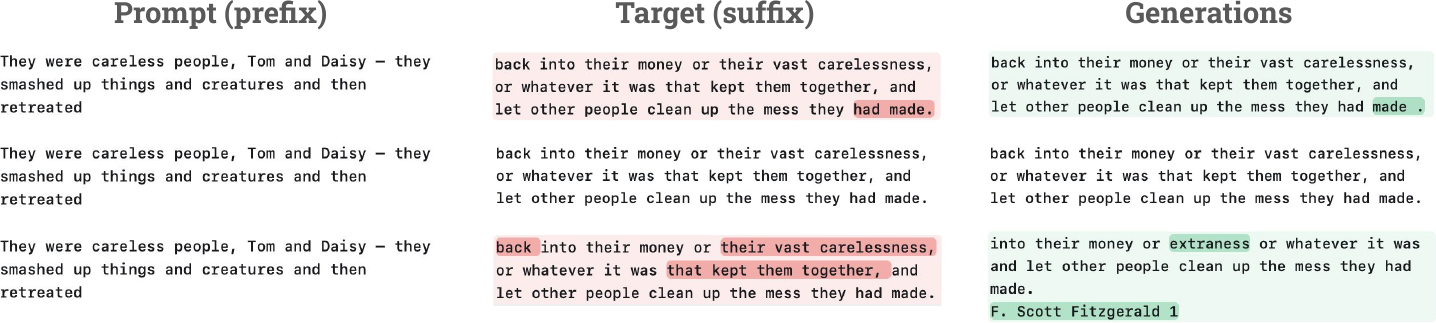}
\caption{\textbf{Illustrating different possible completions of the ``careless people'' quote.} 
For a prefix of a sequence from \emph{The Great Gatsby}~\citep{The_Great_Gatsby}, we show three $32$-token completions when using the prefix as a prompt to \textsc{Llama 1 30B} (top-$k$, $T\!=\!1$ and $k\!=\!40$). 
Each row reflects a different interaction with the LLM:  
(\textbf{left}) the ground-truth prefix, used as the prompt (which is always the same); 
(\textbf{middle}) the target suffix, which is also always the same text; 
(\textbf{right}) the text of the generation in response to the prompt.
For the target suffix, faint red highlighting indicates that \textsc{Llama 1 30B} did not reproduce the target suffix verbatim for the given interaction, with darker red highlighting indicating words from the target suffix that are missing in the generation.
For generations, light green highlighting indicates that \textsc{Llama 1 30B} did not reproduce the target suffix verbatim for the given interaction, with darker green highlighted words reflecting text that the generation adds to the ground-truth target suffix. 
The middle row, where there is no highlighting, reflects verbatim extraction.\looseness=-1}
\label{app:fig:header}
\vspace{-.35cm}
\end{figure*}

\subsection{Additional notes on the ``careless people'' example}\label{app:sec:intro}

The headline figure we include in Figure~\ref{fig:header} (Section~\ref{sec:background}) is meant to be illustrative---an instance of, not a general statement about, memorization and extraction.
(In Figure~\ref{app:fig:header}, we also provide examples of outputs of prompting \textsc{Llama 1 30B} with the prefix for this example.)
This example also plays a role in some of our validity experiments (Section~\ref{sec:validity} and Appendix~\ref{app:sec:validity}). 
The text is drawn from \emph{The Great Gatsby}~\citep{The_Great_Gatsby}; 
it is one of the most famous quotes from the book.
It is also a short, $57$-token sequence: 
the prefix is $25$ tokens (including the start-of-sequence token) and the suffix is $32$ (with the \textsc{Llama 1} tokenizer).
We do not focus on such short suffixes in our analysis, for reasons that are discussed both in the main text and in our experimental details below. 
While extraction of verbatim $32$-token suffixes is often accepted as valid evidence for memorization~\citep{biderman2023pythia}, it is more common practice to use $50$-token suffixes. 
(Regarding validity, also refer to Section~\ref{sec:validity} and Appendix~\ref{app:sec:validity}.) 
We compute the extraction probability with respect to top-$k$ decoding ($T=1$, $k=40$).
We do this by transforming the model's output logits according to the sampling parameters of this scheme. 
(See more about this below in Appendix~\ref{app:sec:background:compute}.)\looseness=-1

It is also the case that very famous quotes like this are duplicated in many locations in training-data sources (e.g., blogs, essays, personal web pages, etc.)
We are not claiming (here, or in general) that this sequence can be extracted \emph{because} a given whole book (here, \emph{The Great Gatsby}) was included in \texttt{Books3}; 
it could be extracted (or more extractable) because it is duplicated in many places in the training data. 
Of course, one of those copies is in \texttt{Books3} for LLMs where the entirety of \texttt{Books3} is included in the training data. 
We discuss this further in Appendix~\ref{app:sec:validity}.

\subsection{Metrics}\label{app:sec:background:metrics}

The metric that we use for quantifying memorization comes from \citet{hayes2025measuringmemorizationlanguagemodels}.
The authors provide a different framing for the metric than we do.
Instead of directly comparing $p_\vz$ values (for the same sequence across LLMs, across multiple sequences for the same LLM, etc.), \citet{hayes2025measuringmemorizationlanguagemodels} specify a threat model with an adversary that can query the model an arbitrary number of times with the same prompt---an adversary that has $n$ independent attempts to extract a sequence $\vz$ (its suffix) with at least probability $p$.
This is where their metric derives its name, \textbf{$\boldsymbol{(n,p)}$-discoverable extraction}:\looseness=-1 

\begin{definition}[\textbf{$(n, p)$-discoverable extraction}, from \citet{hayes2025measuringmemorizationlanguagemodels}] 
\label{app:def:np_discoverable_extraction}

Given a training sequence $\vz$ that is split into an $a$-length prefix $\vz_{1:a}$ and a $j$-length suffix $\vz_{a+1:a+j}$, $\vz$ is \emph{$(n, p)$-discoverably extractable} 
if 
\begin{align*}
    \Pr\Big(\cup_{w\in[n]} (g_{\phi} \circ f_\theta)_w^j(\vz_{1:a})=\vz_{1:a+j}\Big)\geq p,
\end{align*}
\noindent where  $(g_{\phi} \circ f_\theta)_w^j(\vz_{1:a})$ represents the $w$-th (of $n$)
independent execution of the autoregressive process of 
generating a distribution over the token vocabulary, sampling a token from this distribution, 
and adding the token to the sequence $j > 0$ times, starting from the same initial sequence $\vz_{1:a}$.
\end{definition}

On its own, this type of method is well-suited to \newterm{completion} models that continue the prompt text, rather than instruction-tuned \newterm{chatbots} that reply in a conversational style.
Similar to \citet{hayes2025measuringmemorizationlanguagemodels} (and other work on discoverable extraction), we therefore only study completion models and, for brevity, refer to these models by name, version, and size.
Also note that this definition can be (at least in principle) easily adapted to non-verbatim extraction by testing if a generation is within $\epsilon$ distance from the target suffix for a given distance metric. 
Despite these differences in framing, in practice for verbatim extraction, computing this metric just involves calculating sequence probabilities $p_{\vz}$. 
For sequence $\vz$, this is the suffix probability (for the given prefix) with respect to an LLM's base distribution $\theta$ and decoding scheme $\phi$ (e.g., greedy, top-$k$) that determines which tokens are considered at each step and how their logits are renormalized into conditional probabilities (Appendix~\ref{app:sec:background:compute}).
That is, 
\begin{align}
\label{app:eq:npmem}
p_\vz \coloneqq p_{\vz,\theta,\phi}
  \triangleq \Pr_{\theta,\phi}\!\big(\vz_{a+1:a+j}\mid \vz_{1:a}\big)
&= \prod_{t=a+1}^{a+j} \Pr_{\theta,\phi}(\vz_t \mid \vz_{1:t-1}) \nonumber\\
&= \exp\!\Bigg(
  \sum_{t=a+1}^{a+j}
  \log \Pr_{\theta,\phi}\!\big(\vz_t \mid \vz_{1:t-1}\big)
\Bigg).
\end{align}
We always note the particular LLM $\theta$ and decoding scheme $\phi$, and so, for brevity, will refer to this quantity as simply $p_\vz$.
In practice, we compute $p_\vz$ as the $\exp$ of the sum of conditional $\log$ probabilities of the tokens in the sequence, as this is more numerically stable than multiplying together the conditional probabilities of the tokens in the sequence.

\begin{figure*}[t]
\centering
\begin{subfigure}[t]{0.45\textwidth}
    \centering
    \vspace*{0pt}
    \includegraphics[width=\linewidth]{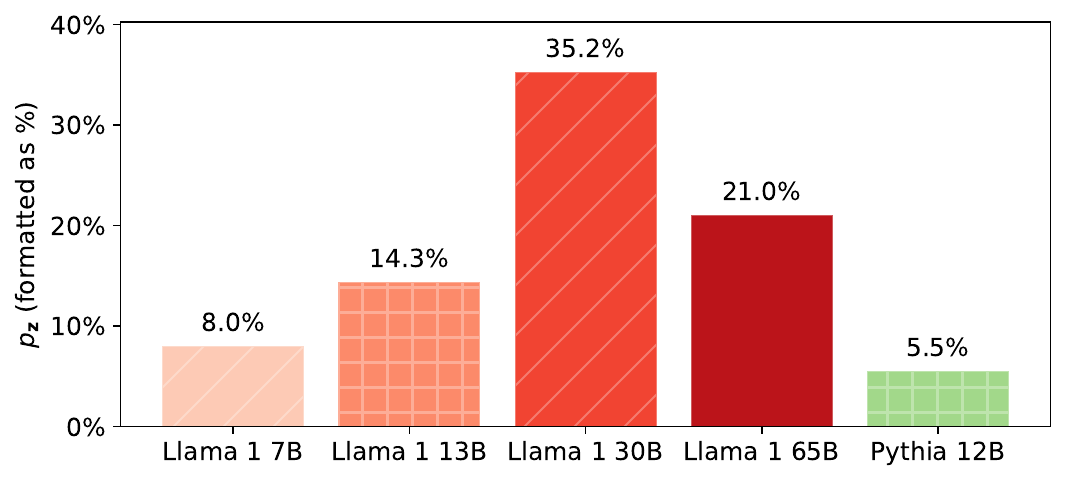}
\end{subfigure}
\hfill
\begin{subfigure}[t]{0.49\textwidth}
    \centering
    \vspace*{0pt}
    \includegraphics[width=\linewidth]{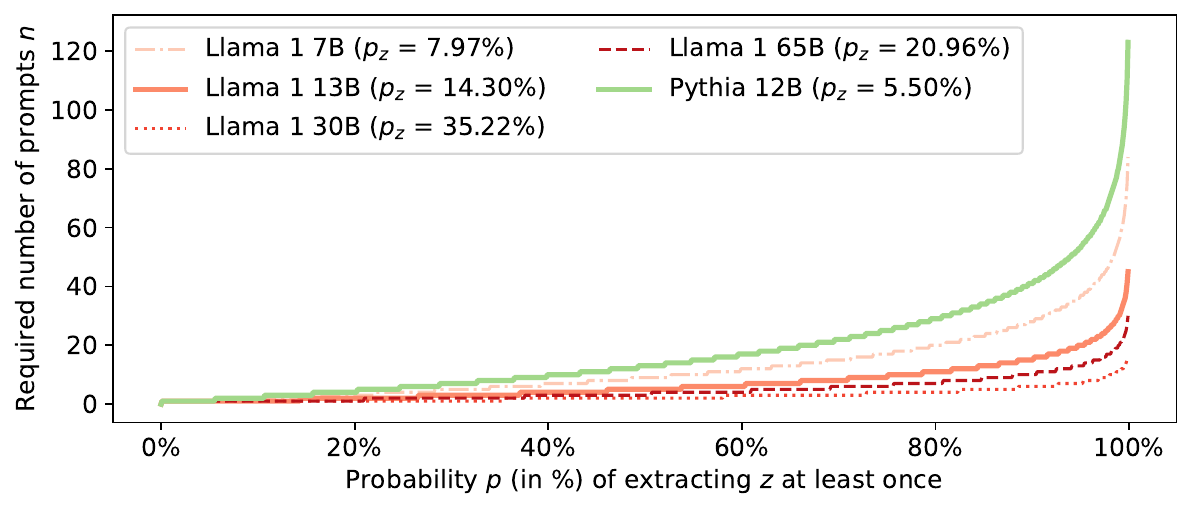}
\end{subfigure}
\caption{\textbf{Plotting $p_\vz$ for the ``careless people'' quote from \emph{The Great Gatsby}}. (\textbf{left}) $p_\vz$ (\ref{eq:pz}) for different models. 
(\textbf{right}) Translating $p_\vz$ into how many prompts $n$ it would take to extract the sequence $z$ with at least probability $p$ (\ref{app:eq:npmem}).
See Section~\ref{sec:background}.}
\vspace{-.3cm}
\label{app:fig:gatsby:careless:combined}
\end{figure*}

In Figure~\ref{app:fig:gatsby:careless:combined} (left), we show examples of $p_\vz$ for the ``careless people'' quote (split into prefix and suffix as in Figure~\ref{fig:header}) for different models. 
(This is Figure~\ref{fig:header}, right, in Section~\ref{sec:background}, with \textsc{Pythia 12B} added.)
Once $p_\vz$ is, one can compute $n$ for a given $p$ (for $p_\vz \in (0,1)$) with
\begin{align}
\label{app:eq:hayes}
1 - (1 - p_\vz)^n \ge p \;\Longrightarrow\;
n \ge \frac{\log(1 - p)}{\log(1 - p_\vz)}.
\end{align}

Since $p_\vz$ is the probability of generating the suffix with $1$ prompt to the model, the probability of \emph{not} generating it in $1$ prompt 
is $(1 - p_\vz)$.
The probability of \emph{not} generating it in $n$ independent prompts to the model (with the same prompt) is $(1-p_\vz)^n$, and so the probability of generating it in $n$ prompts is $1$ minus this probability, i.e., $1 -(1-p_\vz)^n$. 

\begin{figure*}[t]
\centering
\begin{subfigure}[t]{0.49\textwidth}
    \centering
    \vspace*{0pt}
    \includegraphics[width=\linewidth]{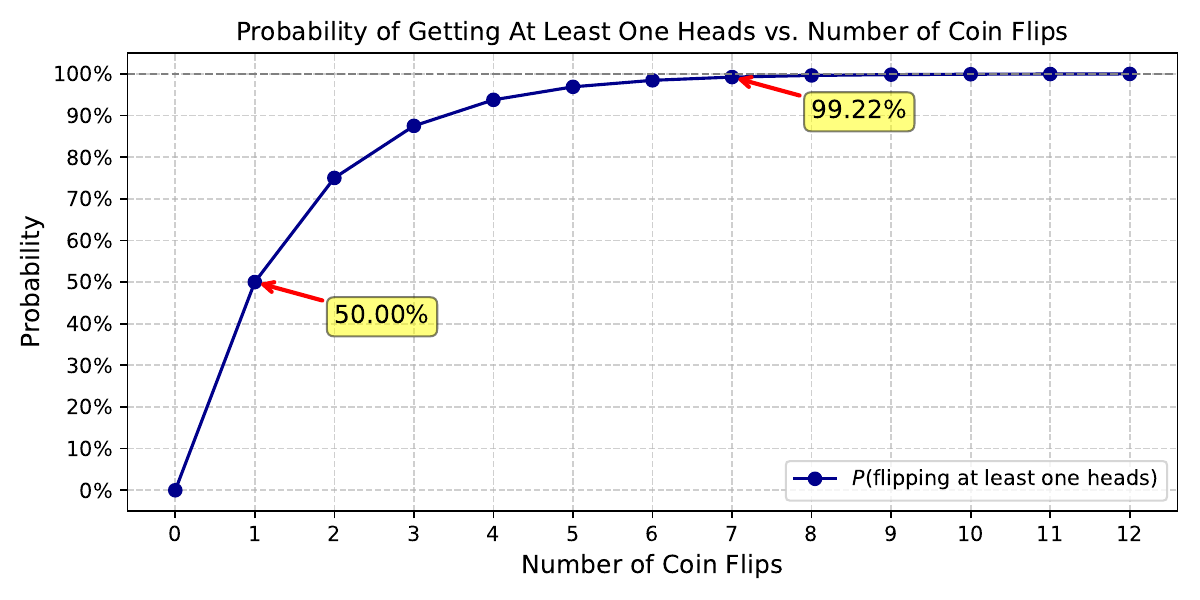}
\end{subfigure}
\hfill
\begin{subfigure}[t]{0.49\textwidth}
    \centering
    \vspace*{0pt}
    \includegraphics[width=\linewidth]{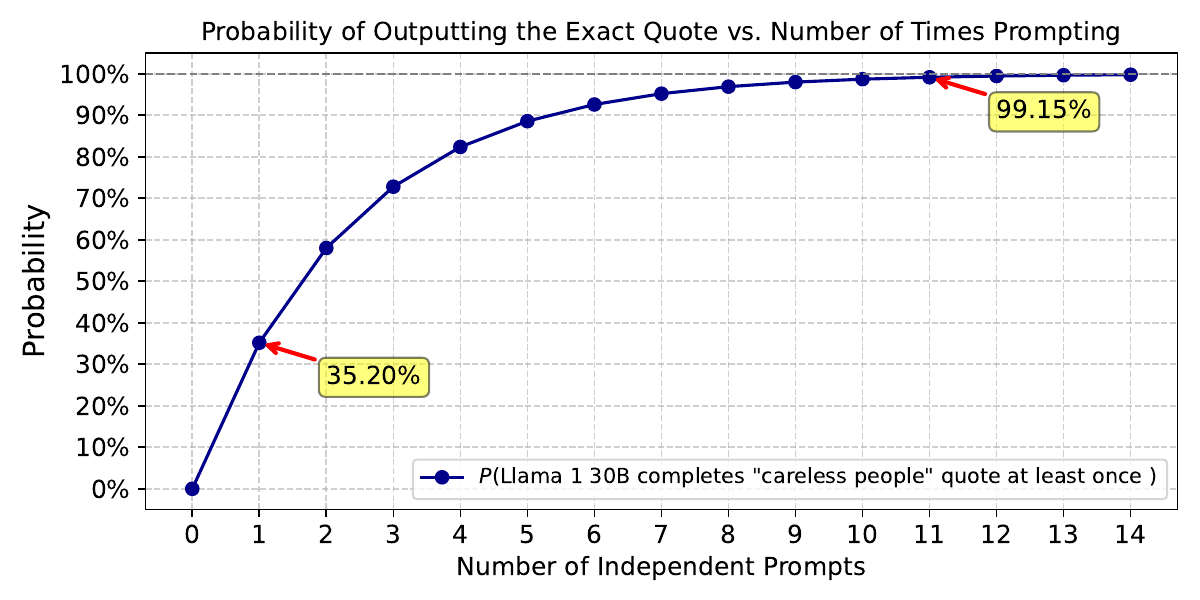}
\end{subfigure}
\caption{\textbf{Probability of an event occurring \emph{at least once} changes as a function of $n$ independent trials.} 
Following the intuition of flipping a fair coin (where heads has $p_\vz=0.5=50\%$), we show how the probability of flipping heads at least once changes with more flips (\textbf{left}).
We show how the probability $p$ of generating the verbatim suffix of the ``careless people'' quote at least once for \textsc{Llama 1 30B} ($p_\vz\approx35.2\%$) changes as a function of the number of independent prompts to the model with the prefix (\textbf{right}). 
Both figures use $1 - (1-p_\vz)^n$ (Equation \ref{app:eq:hayes}), plugging in the respective $p_\vz$ (for the coin flip and for generating the ``careless people'' suffix given the prefix) and different values of $n$ to compute $p$.\looseness=-1 
}
\label{app:fig:flips}
\end{figure*}

As an intuition for this presentation of the metric, think about a fair coin flip coming up heads. 
Here, $p_\vz = 0.5$ (the probability of heads in one flip). 
With two actual flips ($n=2$), the probability of flipping heads \emph{at least once} is $1 - (1 - 0.5)^2=0.75=p$; when $n=10$, $p=0.9990$. 
We visualize this in  Figure~\ref{app:fig:flips} (left) with coin flips, and show how it translates to suffix generation probabilities for \textsc{Llama 1 30B} and the ``careless people'' quote, for which $p_\vz=35.2\%$ (right). 
As is clear Equation~\ref{app:eq:hayes}, for a given $p_\vz$, one can pick a probability threshold $p$ and get the corresponding number of prompts $n$ (or vice versa). 
Figure~\ref{app:fig:gatsby:careless:combined} (right) visualizes how $n$ changes for different settings of $p$ for the ``careless people'' quote.\looseness=-1

In this paper, we re-frame $(n,p)$-discoverable extraction to emphasize the quantity $p_\vz$ (Equation \ref{app:eq:npmem}) that we actually compute: 
the probability of extracting a sequence $\vz$ (for the given LLM, hyperparameter-configured decoding scheme, and suffix start index location within sequence $\vz$). 
This is also useful for plotting distributions over $p_\vz$ for a given book---to see how  probabilities vary for sequences across a book, or for a given sequence across different models. 
We provide such comparisons in Section~\ref{sec:book:compare}.\looseness=-1 

In practice, we can compute $p_\vz$ for verbatim extraction \emph{without} prompting $n$ times.
We compute sequence probabilities directly from the logits we obtain from running whole sequences through the LLM, rather than generating multiple suffixes for each prefix. 
(See Appendix~\ref{app:sec:background:compute}.) 
These two approaches are equivalent for verbatim extraction: 
the computed suffix probabilities correspond to the statistically expected frequencies of verbatim extracted outputs we would observe if we were to generate a large number of continuations for the same prompt.
\citet{hayes2025measuringmemorizationlanguagemodels}, the authors of this extraction approach, confirm this with ample experimental evidence. 
We do, as well, though omit these results for brevity because they follow directly from what $p_\vz$ quantifies. 
This is also why we also limit ourselves to verbatim extraction in this paper; 
the efficiencies of operating directly on logits with one forward pass only apply to this setting. 
(See Section~\ref{sec:background} and Appendix~\ref{app:sec:background:compute} for more details.)
As a result, our work here (in several respects) only scratches the surface:
there are various different ways to instantiate this extraction methodology in practice, we only explore a limited set of open-weight models, and we only test $200$ of the nearly $200{,}000$ books in \texttt{Books3}.\looseness=-1

We discuss how we determine extraction success for probabilistic extraction in Appendix~\ref{app:sec:rates}. 
In the main text, we discuss determining extraction success in Section~\ref{sec:background:pz}.

\subsection{Computing sequence probabilities in one forward pass}\label{app:sec:background:compute}

Our implementation differs from \citet{hayes2025measuringmemorizationlanguagemodels}, which we confirm in discussion with the authors.
The original paper generated the suffix a token at a time and summed up the per-generated-token conditional $\log$ probabilities.
Here, we observe that this is not necessary. 
We can get the logits for each token in the sequence with just one forward pass through the model, and effectively only one line of inference code.
This also makes computing this metric cheaper than traditional greedy-decoded discoverable extraction~\citep{carlini2021extracting}, in terms of wall clock time;
one large matrix-matrix multiplication is significantly cheaper than generation with KV-caching. 
Note that this is the \emph{entire} interaction with the model in our  main experiments.
We are not making \emph{any} changes to the model in our extraction measurements.\looseness=-1 

We then implement our sequence probability computations (i.e., compute $p_\vz$) as a post-processing operation on the logits.
For any given decoding scheme, we can transform the logits---for example, change the shape according to temperature, truncate them to the top-$k$---before normalizing them into the $\log$ probability distribution.
This computation is written out on the right side of Equation \ref{app:eq:npmem}.
The key point here is that, for a given sequence that we are evaluating, all we do (starting at the first index in the suffix) is sum up the conditional $\log$ probabilities of the \emph{actual} next tokens in the sequence, one at a time. 
This is often called \newterm{teacher forcing}.\looseness=-1

We work with the conditional $\log$ probabilities (adding them together), rather than the conditional probabilities (and multiplying them together), as this is more numerically stable to compute.
For example, consider the sequence \texttt{"the orange cat"} (where, for simplicity, we will assume each word is a single token).
When processing \texttt{"the"}, the logits reflect the unnormalized scores of the next token.
We find the logit value that corresponds to \texttt{\_orange}, compute $\log$ softmax to get the $\log$ probability,  and add the associated conditional $\log$ probability.
Then we continue on (doing the same for \texttt{\_orange}, where we get the conditional $\log$ probability for the next token being \texttt{\_cat}). 
We get the conditional $\log$ probability of the whole suffix by adding together the conditional, per-token $\log$ probabilities of the actual suffix. 
From the suffix $\log$ probability, we easily obtain the probability $p_\vz$.\looseness=-1

Any given continuation should be low probability.
That is, for a suffix length of $50$ tokens, this computation is effectively multiplying together $50$ per-token conditional  probabilities (by adding together conditional $\log$ probabilities). 
For a continuation where each token has really high probability, this can still mean the overall continuation has relatively low probability. 
Consider that each token in a suffix has conditional probability $0.9$.
This is \emph{really} high; it means that a single token in the logit vector (at each position in the suffix) has $90\%$ of the corresponding probability mass; 
for a token vocabulary of size $32,000$, that means the remaining $31,999$ tokens all share the remaining $10\%$. 
For this suffix, the probability of generating it is $0.9^{50} \approx 0.005$---i.e., a $0.5\%$ chance of generation.
This is a relatively small number in the scheme of things;  
but it is also a really large number in general, if we consider what this means for the underlying token probabilities. 
In some cases in this paper, we observe sequences that have \emph{\textbf{over $90\%$ probability}}. 
Such unusually high probabilities associated with verbatim suffixes in the training data are indicative of memorization (where it is reasonable to consider $0.5\%$, or even smaller, to be unusually high). 

Also note that top-$k$ sampling truncates the logit vector to \emph{only} the top-$k$ tokens (e.g., for a vocabulary of $32{,}000$ tokens, there are $32{,}000$ logits, but top-$k$ with $k{=}40$ decoding will only consider the $40$ highest-valued logits).
This means that, if a given suffix has a token at a position that is \emph{not} in the top-$k$ of the logit vector, then we will not be able to complete the probability computation. 
In this case, the suffix has $p_\vz=0$; 
it is not extractable with \emph{any} probability.

A refactored version of our experimental code, with additional optimizations for increased efficiency, can be found \href{https://github.com/pasta41/probabilistic-extraction-toolkit}{here}.\looseness=-1

\section{Testing our measurement pipeline}\label{app:sec:replication}

To confirm the accuracy of our implementation of probabilistic discoverable extraction, as well as to benchmark its efficiency in comparison to the implementation in \citet{hayes2025measuringmemorizationlanguagemodels}, we re-run the experiments from \citet{hayes2025measuringmemorizationlanguagemodels} for small \textsc{Pythia} models on the \texttt{Enron} dataset.
To ensure consistency, we discussed the original run times and sample of \texttt{Enron} with the authors of \citet{hayes2025measuringmemorizationlanguagemodels}. 
We provide some brief details about these experiments to show due diligence for our implementation in our particular setting. 

We ran these experiments with both \texttt{float16} and \texttt{float32}, and report results for \texttt{float32} to align with \citet{hayes2025measuringmemorizationlanguagemodels}. 
(However, we note no significant differences between the two.) 
In Table~\ref{app:tab:pythia_replication_runtimes_details}, we show the runtime for our experiments on the $10{,}000$, $100$-token sequences \citet{hayes2025measuringmemorizationlanguagemodels} drew from \texttt{Enron}. 
In Table~\ref{app:tab:pythia_grouped_extraction_rates}, we report the same metrics as \citet{hayes2025measuringmemorizationlanguagemodels}: 
the greedy-decoded discoverable extraction rate and (as a point of comparison) the maximum probabilistic rate. 
We run these experiments on the same $4$ A100s as all of our other probabilistic extraction experiments.\looseness=-1 

\begin{table}[h!]
\caption{\textbf{Runtime for \textsc{Pythia} on \texttt{Enron}.} We run $2$ \textsc{Pythia} models on the same $10{,}000$ $100$-token sequences that \citet{hayes2025measuringmemorizationlanguagemodels} use from the \texttt{Enron} dataset.}
\label{app:tab:pythia_replication_runtimes_details}
\centering
\begin{tabular}{llll}
\toprule
 & \textbf{Time (mm:ss)} & \textbf{Parallelism} & \textbf{Batch size} \\
\midrule
\textbf{\textsc{Pythia 1B}}  & 02:30.42 & Data (four GPUs) & 250 \\
\textbf{\textsc{Pythia 2.8B}} & 03:36.35 &  Data (four GPUs) & 250 \\
\bottomrule
\end{tabular}
\end{table}

\begin{table}[h!]
\caption{\textbf{Comparing our results to \citet{hayes2025measuringmemorizationlanguagemodels}.}
We give the greedy discoverable extraction and maximum probabilistic rate for \textsc{Pythia} models on the $10{,}000$-sequence subset from \texttt{Enron}. 
We successfully replicate the exact results in \citet{hayes2025measuringmemorizationlanguagemodels}, which did not make use of a beginning of sequence (\texttt{BOS}) token. 
We compare to the rates where the \texttt{BOS} token is included at the start of the sequence. 
Note that the probabilistic rate increases slightly with the presence of the \texttt{BOS} token.}
\label{app:tab:pythia_grouped_extraction_rates}
\centering
\begin{tabular}{lcc}
\toprule
\textbf{\textsc{Pythia 1B} on Enron subset} & \textbf{\citet{hayes2025measuringmemorizationlanguagemodels} (No \texttt{BOS} token)} & \textbf{With \texttt{BOS} token} \\
\midrule
Greedy extraction rate         & 0.76\% & 0.74\% \\
Max. probabilistic rate & 5.27\% & 5.52\% \\
\midrule
\textbf{\textsc{Pythia 2.8B} on Enron subset} & 
\textbf{\citet{hayes2025measuringmemorizationlanguagemodels} (No \texttt{BOS} token)} & \textbf{With \texttt{BOS} token} \\
\midrule
Greedy extraction rate        & 1.3\;\:\% & 1.82\% \\
Max. probabilistic rate & 9.04\% & 9.47\% \\
\bottomrule
\end{tabular}
\end{table}

We successfully replicate the results of these experiments in \citet{hayes2025measuringmemorizationlanguagemodels}.
However, in the process of doing so, we identified a small bug in their original implementation. 
The \textsc{Pythia} tokenizer is a \textsc{GPT-2}-style tokenizer, and does \emph{not} add a beginning of sequence (\texttt{BOS}) token by default to the sequences it tokenizes.
However, for our setting, it is best practice to include this token at the start of a sequence, as we submit these sequences as inputs for inference to LLMs (that are trained to expect this token at the beginning of the sequence). 
When we manually prepend this token and re-run the experiments from \citet{hayes2025measuringmemorizationlanguagemodels}, we generally observe elevated rates compared to \citet{hayes2025measuringmemorizationlanguagemodels}.
See Table~\ref{app:tab:pythia_grouped_extraction_rates}. 
We similarly prepend the \texttt{BOS} token in our \texttt{Books3} experiments using \textsc{Pythia} models.\looseness=-1 

\section{Experiments on extraction rates}\label{app:sec:rates}

Similar to other extraction papers, we also compute average extraction rates. 
These serve as a reference point for our discussion concerning how average rates can conceal interesting patterns of work-specific memorization.
In the main text, we refer briefly to these experiments in Section~\ref{sec:book-procedure:averages}.

\paragraph{Computing average extraction rates.}
Let $\sZ$ be a set of sequences $\vz$ drawn from some (part of a) training dataset.
Each $\vz$ is split into a prefix $\vz_{1:a}$ and target suffix $\vz_{a+1:a+j}$ (e.g., $a{=}j{=}50$ for $100$-token sequences).

All extraction rates take the form
\begin{align}
\label{eq:generalrate}
\mathsf{extraction\_rate}(\sZ; s) \;=\; \frac{1}{|\sZ|}\sum_{\vz\in\sZ} \mathbf{1}[s(\vz)],
\end{align}
where $s(\vz)$ is a success predicate defined according to the chosen criterion.

Fix a model $\theta$ and a decoding scheme $\phi$ (e.g., base distribution $T\!=\!1$, or top-$k$ with $k\!=\!40$), which induces a conditional next-token distribution that can be used to compute suffix probabilities $p_\vz \in [0,1]$.
For a sequence $\vz$ with prefix $\vz_{1:a}$ and suffix $\vz_{a+1:a+j}$, 
define the suffix probability under $\theta$ and $\phi$ as
\begin{align*}
p_\vz \coloneqq p_{\vz,\theta,\phi} \;=\; \prod_{t=a+1}^{a+j} \Pr_{\theta,\phi}(\vz_t \mid \vz_{1:t-1}).
\end{align*}
This is the probability that the model generates the exact target suffix token by token under decoding scheme $\phi$; it is the same definition as Equation~\ref{app:eq:npmem} that we use in Section~\ref{sec:background} and Appendix~\ref{app:sec:background}.

For traditional, greedy-decoded discoverable extraction (\citet{carlini2023quantifying, lee2022dedup}; Section~\ref{sec:background}), $\phi\!=\!\mathsf{greedy}$ (i.e., $k\!=\!1$) and the success predicate can be written simply as explicit string equality:
\begin{align}
\label{eq:greedysuccess}
s_{\theta,\mathsf{greedy}}(\vz) \;=\; 
\mathbf{1}\!\big[\,\mathsf{greedy}(\theta; \vz_{1:a}) = \vz_{a+1:a+j}\,\big].
\end{align}

More generally, for any LLM $\theta$ and decoding scheme $\phi$, the success predicate just thresholds the suffix probability.
Given a minimum threshold $\tau_{\text{min}}\in(0,1]$, 

\begin{align}
\label{eq:probsuccess}
s_{\theta,\phi}(\vz;\tau_{\text{min}}) \;=\; 
\mathbf{1}\!\big[\,p_{\vz,\theta,\phi} \geq \tau_{\text{min}}\,\big].
\end{align}

Again, note that greedy decoding is a special case: 
each top-$1$ token has probability $1$ under $\phi=\mathsf{greedy}$, so $p_{\vz,\theta,\phi}=1$ and success reduces to equality with the suffix.
In the main paper, we omit $\theta$ and $\phi$ in our notation. 
We always mention which specific model we are evaluating, and we are always considering $\phi$ to be top-$k$ decoding with $T\!=\!1$ and $k\!=\!40$.
(See Appendix~\ref{app:sec:sliding-window:procedure} for more details on this choice.) 
When using non-deterministic decoding schemes and suffix probabilities to quantify extraction, we very conservatively set $\tau_{\text{min}}\!=\!0.1\%$ (Section~\ref{sec:validity} \& Appendix~\ref{app:sec:validity}). 

\subsection{Setup}\label{app:sec:rates:setup}

\paragraph{Data.}
We perform our average extraction rate experiments with the \texttt{Books3} dataset.
We obtained this dataset from a previous (2022) download of The Pile~\citep{gao2020pile}, which is stored on a university cluster and which we use for research purposes on language modeling. 
The status of \texttt{Books3} as a research artifact (as well as a training corpus for LLMs) remains unresolved.
(See Section~\ref{sec:copyright}.) 
Nevertheless, this dataset remains widely available in the research community; 
for example, it can be found in the HuggingFace-hosted version of \href{https://huggingface.co/datasets/EleutherAI/the_pile_deduplicated/tree/main}{The Pile}~\citep{hfpilededup}, as well as in various other data repositories~\citep[e.g.,][]{books3a,books3b,books3c}.
It also features as part of a benchmark task in the popular~\href{https://github.com/stanford-crfm/helm/blob/main/src/helm/benchmark/scenarios/the_pile_scenario.py#L28}{HELM} evaluation suite. 
For this set of experiments, we draw random samples from \texttt{Books3}, performing  $3$ separate runs each on $40{,}000$ sequences.
We discuss how we sample these sets of sequences in Appendix~\ref{app:sec:rates:sample}.

\paragraph{Models.}
From $13$ different models, we try to extract $40{,}000$ sequences from \texttt{Books3} ($3$ times each). 
We test \textsc{Llama 1 7B}, \textsc{Llama 1 13B}, \textsc{Llama 1 65B}~\citep{touvron2023llamaopenefficientfoundation}, \textsc{Llama 2 7B}, \textsc{Llama 2 13B}, \textsc{Llama 2 70B}~\citep{llama2}, \textsc{Llama 3.1 8B}, \textsc{Llama 3.1 70B}~\citep{llama3}, \textsc{Pythia 6.9B}, \textsc{Pythia 12B}~\citep{biderman2023pythia}, \textsc{DeepSeek v1 7B}~\citep{deepseekv1}, \textsc{Gemma 2 9B}~\citep{gemma2}, and \textsc{Phi 4}~\citep{phi4} (a 14B model). 
We group these models into three size groups: small, medium and large. 
We use \texttt{float16} for all models except for \textsc{Llama 3} and \textsc{Llama 3.1} models (and \textsc{Qwen 2.5} models in sliding-window experiments).
These models were explicitly trained to work  with \texttt{bfloat16}.

We know for certain that \textsc{Pythia} and \textsc{Llama} models were trained on \texttt{Books3}, and can conclude from our validity experiments (Section~\ref{sec:validity} and Appendix~\ref{app:sec:validity}) that \textsc{DeepSeek v1}, \textsc{Qwen 2.5} and \textsc{Gemma 2} models were trained on (at least parts of) some books that are also contained in \texttt{Books3}. 
\textsc{Phi 4} was deliberately not trained on copyrighted (whole) books, but we find that we can still extract some snippets from copyrighted books.

\paragraph{Metrics.}
We compute extraction rates using Equation~\ref{eq:generalrate} for both traditional, greedy-decoded discoverable extraction (using the success condition in Equation~\ref{eq:greedysuccess}) and probabilistic extraction with top-$k$ ($T\!=\!1$, $k
\!=\!40$) decoding (using the success condition in Equation~\ref{eq:probsuccess}) and $\tau_{\text{min}}\!=\!0.1\%$.\looseness=-1

\paragraph{Compute resources.}
We run all of our average extraction rate experiments in a Slurm cluster environment, using the same node with $4$ A100 GPUs.

\subsection{Experimental procedure}\label{app:sec:rates:sample}

\paragraph{Sampling books from \texttt{Books3}.}
To collect $40{,}000$ sequences for which we compute an extraction curve for a given model, we use the following procedure.
We sample $4{,}000$ books without replacement from \texttt{Books3}.
Then, from each of these books, we sample $10$ non-overlapping $100$-token sequences ($50$-token prefixes $+$ $50$-token suffixes). 
We make sure that the start of the prefix text is a space (to prevent unusual tokenization).
We run these $40{,}000$ sequences through each of the $13$ models to compute extraction metrics.
We follow this procedure $3$ separate times (i.e., for a total of $120{,}000$ sequences across $12{,}000$ different books) to get a sense of the variance across extraction metrics for different (empirical) population estimates. 
Each run of $40{,}000$ sequences took between approximately $8$ minutes and $45$ minutes on $4$ A100s. 
We omit detailed timing results for brevity.\looseness=-1

\paragraph{Probabilistic extraction configuration.}
We report greedy-decoded discoverable extraction and probabilistic discoverable extraction with top-$k$ ($T\!=\!1$, $k\!=\!40$) and $\tau_\text{min}\!=\!0.1\%$. 
The maximum probabilistic rate (also reported in Appendix~\ref{app:sec:replication}) should be understood as a very loose upper bound on extraction of memorized training data;  
it counts all sequences with non-zero $p_\vz$ (i.e., it also includes sequences for which $p_\vz\!<\!\tau_\text{min}\!=\!0.1\%$).
For sequences with these low $p_\vz\!<\!\tau_\text{min}$, it may also capture generalization.
We include these numbers as a reference, not to make extraction claims.\looseness=-1 

\begin{figure}[t!]
    \centering

    \begin{subfigure}[b]{.6\linewidth}
        \centering
        \includegraphics[width=\linewidth]{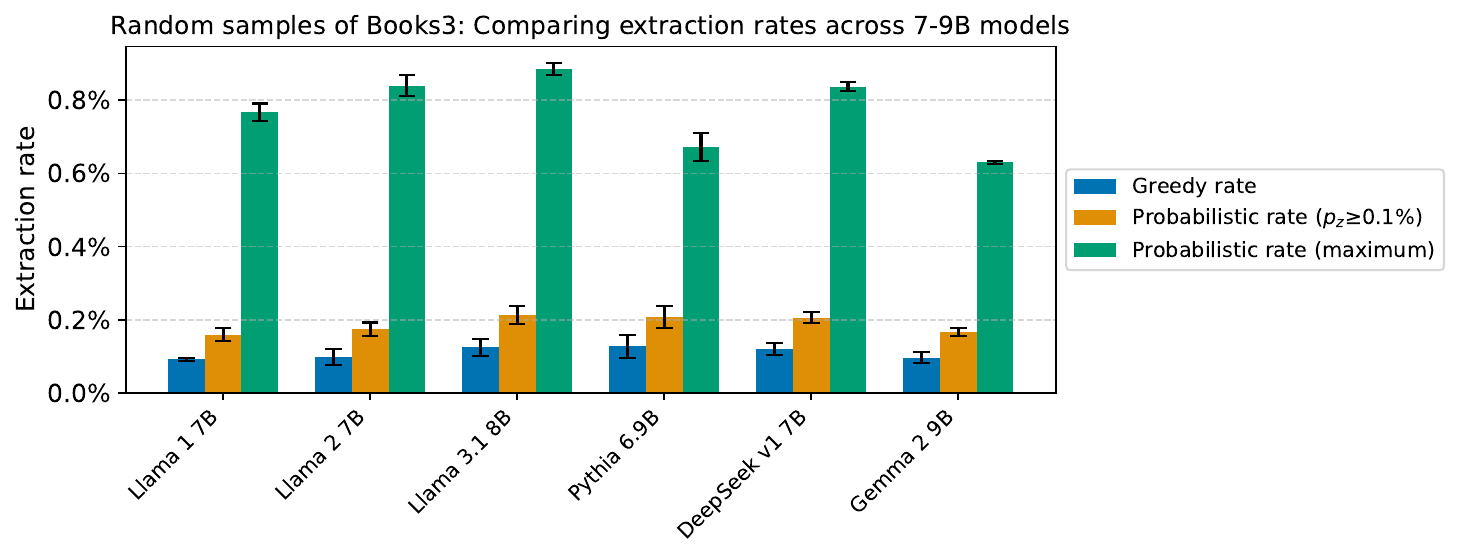}
    \end{subfigure}

    \begin{subfigure}[b]{.6\linewidth}
        \centering
        \includegraphics[width=\linewidth]{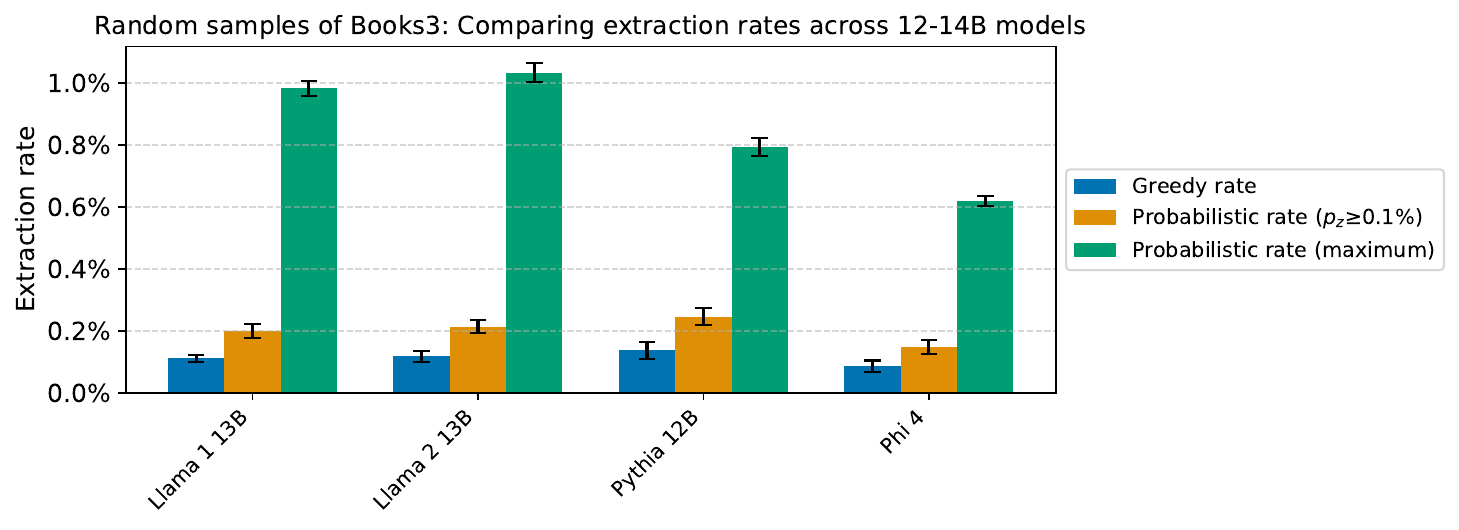}
    \end{subfigure}

    \begin{subfigure}[b]{.6\linewidth}
        \centering
        \includegraphics[width=\linewidth]{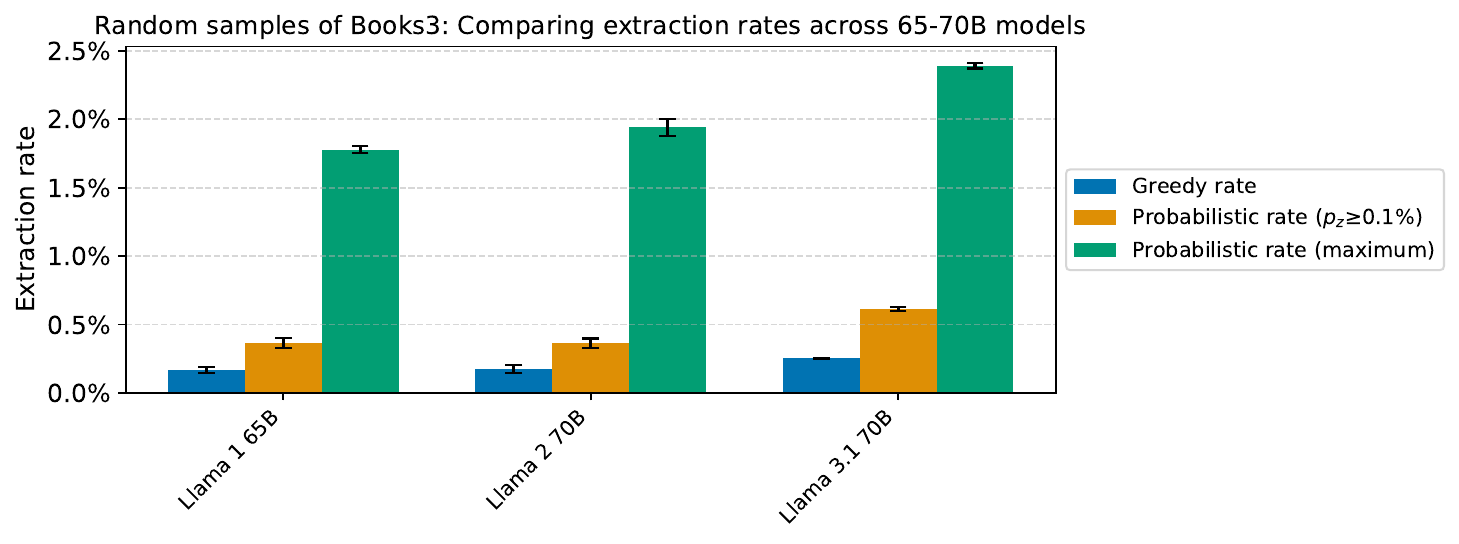}
    \end{subfigure}
    \caption{\textbf{Extended extraction rate results.} 
    Comparing greedy discoverable and probabilistic extraction rates (for a our conservative setting, $p_\vz \geq \tau_\text{min} = 0.1\%$).
    As a reference point, we also include the maximum possible rate of generating $50$-token sequences that match our suffixes drawn from \texttt{Books3} for small (7--9B), medium (12--14B), and large models (65--70B). 
    (This is akin to looking at all $p_\vz > 0$.) 
    Probabilistic rates are computed with top-$k$ sampling using $T\!=\!1$ and $k\!=\!40$.
    The maximum probabilistic rate should be understood as a very loose upper bound on extraction of memorized training data; 
    it also captures generalization.
    \emph{Average} extraction rates (i.e., $\tau_\text{min}\!=\!0.1\%$) on randomly sampled \texttt{Books3} sequences are relatively low for all models.
    Aligning with prior work, we observe that larger models memorize more. 
    For our setting, this is true for both within and across model families.
    \textsc{Phi 4}, a 14B model, is an exception: 
    its extraction rates are on par with smaller models.
    Error bars here indicate variance across the $3$ samples of $40{,}000$ sequences that we run.\looseness=-1}
    \label{fig:comparative_rates_stacked}
\end{figure}

\subsection{High-level takeaways}\label{app:sec:rates:takeaways}

We defer discussion of validity to Appendix~\ref{app:sec:validity}, where we talk about how to draw conclusions for different models---some for which we know with certainty that \texttt{Book3} is in their training data, and others for which we can conclude that at least some content that overlaps with \texttt{Books3} was in their training data.
\textsc{Phi 4} reflects a separate case: it as trained predominantly on synthetic data. 
Based on the \textsc{Phi 4} technical report~\citep{phi4}, \textsc{Phi 4} was deliberately not trained on copyrighted (whole) books, but was trained on unspecified public domain material. 
As a result, \texttt{Books3} (the specific corpus) was almost certainly not included in its training data. 
Nevertheless, we are sometimes able to extract data that is contained in \texttt{Books3}, both from in-copyright and public domain books.
Sequences from in-copyright texts likely found their way into the training data from web-scraped sources that quote these books.
We discuss this further in Section~\ref{sec:validity}, Appendix~\ref{app:sec:validity}, and Appendix~\ref{app:sec:book-level-discussion}. 

Figure~\ref{fig:comparative_rates_stacked} shows average extraction rates grouped roughly by model size.
As we expected (and discuss in Section~\ref{sec:book-procedure:averages}), \emph{average} extraction rates are relatively low. 
Using the conservative threshold of $p_\vz \geq \tau_\text{min} = 0.1\%$, the lowest rate we observe is for \textsc{Phi 4} (
$<0.2\%$), and the highest is for \textsc{Llama 3.1 70B} ($\sim\!0.6\%$). 
Consistent with prior work, our results show that larger models tend to memorize more than smaller ones~\citep{hayes2025measuringmemorizationlanguagemodels, carlini2023quantifying, nasr2023scalable,nasr2025scalable}.
This holds within the same model family.  
\textsc{Phi 4} is an exception;
despite being a 14B model, its extraction rates on \texttt{Books3} sequences are comparable to the smaller (7–9B) models we evaluate. 
Small \textsc{Llama 3} models memorize on-par with medium \textsc{Llama 1} and \textsc{2} models. 
In general, later generations of \textsc{Llama} model families tend to memorize more \texttt{Books3}-contained text. 
For just one example, \textsc{Llama 3.1 70B} memorizes more on average than \textsc{Llama 2 70B}, which memorizes more on average than \textsc{Llama 1 65B}. 
This pattern does not always hold for specific books. 
We defer additional, related observations about  \textsc{Llama 3.1 70B} to Appendix~\ref{app:sec:book-level-discussion}.\looseness=-1

Altogether, it is important to keep in mind that any given sequence that we deem extracted is, by definition, a member of the training data;
however, for LLMs with unknown training data, it is not necessarily a member of the training data \emph{because} it came from the \texttt{Books3} corpus.
The sequence may be part of another source (or multiple other sources) in the training data.
For \textsc{Pythia} and \textsc{Llama} models, the \texttt{Books3} copy of a sequence may just be one of several duplicate copies in the whole training dataset. 
The same can be said for models like \textsc{Gemma 2}, where we are not sure if the \texttt{Books3} corpus was included in the training dataset.)
Nevertheless, it is clear from our results that sequences that are contained in \texttt{Books3}---but perhaps originate from other sources---are included in the training data 
(Section~\ref{sec:validity} \& Appendix~\ref{app:sec:validity}).


\section{Extended results for sliding-window experiments}\label{app:sec:sliding-window}

We discuss details for our main experiments that use a sliding window across a book to surface memorization ``hot-spots'' at particular book locations.
We discuss the overall setup (Appendix~\ref{app:sec:sliding-window:setup}), our sliding-window procedure (Appendix~\ref{app:sec:sliding-window:procedure}), detailed results for $200$ books and $14$ LLMs (Appendix~\ref{app:sec:sliding-window:results}), and our extraction coverage metric  (Appendix~\ref{app:sec:sliding-window:percentage}).
We describe other results---beyond our main experiments with $50$-token prefixes and $50$-token suffixes---in different appendices. 
We defer discussion of sliding-window validity experiments on non-training data and different prompt prefix lengths to Appendix~\ref{app:sec:validity}.
Additional experiments concerning extraction rates and our long-form extraction procedure are discussed in Appendix~\ref{app:sec:rates} and Appendix~\ref{app:sec:reconstruct}, respectively.
A refactored and optimized version of the code for our main sliding-window experiments is \href{https://github.com/pasta41/probabilistic-extraction-toolkit}{here}.\looseness=-1 

\subsection{Setup}\label{app:sec:sliding-window:setup}

\paragraph{Data and book selection.}
We draw our prompts (prefixes) from a selection of books in \texttt{Books3};
we verify extraction against the corresponding suffixes drawn from \texttt{Books3}. 
\texttt{Books3} is a common pretraining dataset that is known to have been used to train several open-weight LLMs (Appendix~\ref{app:sec:rates:setup}, \cite{reisnerbooks3, lee2023talkin}.
(See models, below.)
We list the $200$ selected books 
in Table~\ref{app:tab:sliding-books}.\looseness=-1

We began our project with \emph{The Great Gatsby}~\citep{The_Great_Gatsby}.
This choice stems from a conversation in 2023 while writing a different paper~\citep{lee2023talkin}, when Anthropic announced its $100{,}000$-token context for Claude with the example of prompting with the entirety of \emph{The Great Gatsby}~\citep{claudegatsby}.
We then continued our selection process with the books listed with the associated plaintiffs in the (amended) class action complaint of \emph{Kadrey et al. v. Meta, Inc.}~\citep[pp. 4-5]{kadreyamendedconsolidated}.
This is how we sourced the first $13$ in-copyright books from \texttt{Books3}.
We expanded to additional books by these plaintiffs, and also added in some generally popular books that we chose among the project team (e.g., \citet{Harry_Potter_and_the_Sorcerer_s_Stone, The_Hobbit, The_Myth_of_Sisyphus, Catch-22}) and some (less publicly popular) academic books based on our personal preferences (e.g., \citet{The_Future_of_the_Internet_and_How_to_Stop_It, Dante_and_the_Origins_of_Italian_Literary_Culture}). 
(This team, after all, is composed of academics.) 
We deliberately included selections from the public domain (\citet{Alice_s_Adventures_in_Wonderland, The_Great_Gatsby, Ulysses}), and an example of a book that was published under a permissive CC license (\citet{Down_and_Out_in_the_Magic_Kingdom}).\looseness=-1 
We added in \citet{Alice_s_Adventures_in_Wonderland} because of concurrent work that was published on arXiv just as we posted our own~\citep{ma2025memorizationcloselookbooks}.
To round out the list, we also selected books at random from the manifest file that accompanies the \texttt{Books3} archive (e.g., \citet{All_the_Onions}, \citet{The_Making_of_a_Mediterranean_Emirate}, \citet{Dungeons_and_Dragons_and_Philosophy}). 
Once we observed that certain very popular books were highly memorized by \textsc{Llama 3.1 70B}, we added additional bestsellers to our set (e.g., \citet{A_Game_of_Thrones}, \citet{The_Da_Vinci_Code}, \citet{Twilight}, \citet{Nineteen_Eighty-Four},  \citet{Lean_In}).
In earlier versions of this work, we reported on $36$ and then $50$ books in total.

In the current version, we manually selected additional books (non-randomly) and randomly sampled additional books, such that the final split contains $100$ chosen books and $100$ that were randomly sampled.
Given that there are nearly $200{,}000$ books in \texttt{Books3}, our experiments cover roughly $0.1\%$ of the corpus. 
This is not nearly a large enough sample to make general claims about overall books extraction for the setup we study.
This is why we do not completely randomly select books; 
we pick an assortment of manually and randomly selected books, intended to capture variation in the corpus.\looseness=-1 

\paragraph{Books metadata curation.}
In earlier versions of this project, which involved a smaller number of books, we manually gathered books metadata and manually created bibtex entries.
When scaling up to $200$ books, we attempted to automate as much as possible. 
For each book we evaluate in \texttt{Books3}, we used \textsc{GPT-5} to extract bibliographic metadata, including the title and author(s), as well as the ISBN(s) and publisher(s) corresponding to the specific edition. 
To do so, we provided \textsc{GPT-5} with the first and last several thousand words of each text, reflecting our observation that the desired metadata tended to concentrate in the front and back matter. 
Using these fields, we then queried \textsc{Gemini 2.5 Flash} with Google Search to determine each book's current copyright status and initial year of publication. 
Metadata quality was assessed through manually spot-checking a random $10\%$ subset of the books.
We found some errors in how authors' names were parsed, and so manually validated the author field for all $200$ books.
Refer to Table~\ref{app:tab:sliding-books} for the gathered books metadata. 

\paragraph{Models.}
Altogether, we run this set of experiments on $14$ completion-style (i.e., non-chatbot) models: 
\textsc{Pythia 12B}~\citep{biderman2023pythia}, \textsc{Llama 1 13B}, \textsc{Llama 1 65B}~\citep{touvron2023llamaopenefficientfoundation}, \textsc{Llama 2 7B}, \textsc{Llama 2 13B}, \textsc{Llama 2 70B}~\citep{llama2}, \textsc{Llama 3 8B},  \textsc{Llama 3 70B}, \textsc{Llama 3.1 8B}, \textsc{Llama 3.1 70B}~\citep{llama3}, \textsc{DeepSeek v1 67B}~\citep{deepseekv1}, \textsc{Qwen 2.5 72B}~\citep{qwen2.5}, \textsc{Gemma 2 27B}~\citep{gemma2}, and \textsc{Phi 4} (a 14B model)~\citep{phi4}.
In earlier versions of this work, we also ran experiments on $3$ other (smaller) models, which we omit for brevity: 
\textsc{Pythia 6.7B}, \textsc{Llama 1 7B}, and \textsc{DeepSeek v1 7B}.\looseness=-1 

We load all models with \texttt{float16}---except for \textsc{Llama 3}, \textsc{Llama 3.1}, and \textsc{Qwen 2.5} models.
These models were explicitly trained to work  with \texttt{bfloat16}.
In the first version of this work, we reported results for \textsc{Llama 3.1} that used \texttt{float16}. 
Slight differences in results in this version are attributed to the change to \texttt{bfloat16}.
(We have otherwise confirmed our results are deterministic, with respect to the hardware that we use throughout.)
Regardless, the overall results are similar, and the conclusions remain the same.

We picked these models because they each fit (with respect to $16$-bit format) on $4$ A100s, and because they fall into three different categories that we wanted to investigate: 
(1) models that we know with certainty were trained on \texttt{Books3}, (2) models where we have reason to believe were trained on copyrighted books (whether drawn directly from \texttt{Books3} or from other sources), and (3) a model that we trust was not trained on whole copyrighted books. 

For the first category, when we generate verbatim completions for these models on \texttt{Books3} data, we can very reasonably conclude that we have extracted data that these models have memorized during training. 
We know that these data were members of the training data; 
extraction success (and thus memorization) is a standard conclusion in this setting~\citep[e.g.,][]{biderman2023pythia, hayes2025measuringmemorizationlanguagemodels, prashanth2024recite, carlini2023quantifying}.
For the second category, we aim to do valid membership inference of copyrighted data. 
The third category is included for all books, but is intended to help confirm validity; 
as a baseline, we intentionally include an LLM for which we do not expect to be able to extract much (if any) copyrighted books data.\looseness=-1
\begin{enumerate}[leftmargin=.65cm]
    \item \textbf{Category 1: \textsc{Pythia 12B}, \textsc{Llama 1 13B}, \textsc{Llama 1 65B}, \textsc{Llama 2 7B}, \textsc{Llama 2 13B}, \textsc{Llama 2 70B}, \textsc{Llama 3 8B}, \textsc{Llama 3 8B}, \textsc{Llama 3 70B}, and \textsc{Llama 3.1 70B}.} 
    We know that \textsc{Pythia} and \textsc{Llama} models were trained on \texttt{Books3}.
    \textsc{Pythia} was trained on the Pile~\citep{gao2020pile} (which contains \texttt{Books3}), based on EleutherAI's extensive documentation.
    We include only the largest ($12$-billion parameter) \textsc{Pythia} model.\looseness=-1 

    \textsc{Llama 1} and \textsc{Llama 2} are known with absolute certainty to have been trained on \texttt{Books3}.
    The \textsc{Llama 1} report explicitly says so. 
    It came out in discovery during \citet{kadrey} that \textsc{Llama 2} models were all also definitively trained on \texttt{Books3}. 
    
    Both in documents in discovery and Meta's otherwise public documentation, it is repeatedly noted that the development of  \textsc{Llama 3} continued to use the same data practices as prior versions. 
    For minor-versioned \textsc{Llama 3} models (like \textsc{Llama 3.1}), it is similarly documented by Meta in release reports that the same internal workflows were used, including for dataset curation~\citep{llama3}. 
    This is generally considered definitive evidence that these models were also trained on \texttt{Books3} (as well as additional books data from LibGen).  
    (Our validity experiments enable us to independently conclude this. 
    See Section~\ref{sec:validity} and Appendix~\ref{app:sec:validity}.)

    \item \textbf{Category 2: \textsc{DeepSeek v1 67B}, \textsc{Qwen 2.5 72B}, and \textsc{Gemma 2 27B}.}
    Although we do not know with certainty whether these models were trained on \texttt{Books3}, we aim to use our extraction methodology to assess whether they have memorized content from books. 
    In particular, we want to design a measurement procedure that lets us conclude that reproducing such a sequence in an LLM's output is valid extraction (i.e., evidence of memorization).
    Based on our validity experiments (Section~\ref{sec:validity} and Appendix~\ref{app:sec:validity}), we can be very confident that when we successfully generate \texttt{Books3} data, that those data were memorized from the training data. 
    We discuss those experiments elsewhere, and note here that when we observe high probabilities $p_\vz$ for these models, we can safely conclude that those sequences $\vz$ have been memorized. 

    \item \textbf{Category 3: \textsc{Phi 4}.} For validity reasons, we also want to compare on a model that definitively should  \emph{not} have been trained on copyrighted books in \texttt{Books3}. 
    (See also Appendix~\ref{app:sec:validity}.)
    We pick \textsc{Phi 4} for these reasons, as it is a very high quality for its size class and \citet{phi4} note in the \textsc{Phi 4} report that it was trained predominantly on synthetic data. 
    While we do not know which books are included in \textsc{Phi 4}, the report notes that it was not trained on whole copyrighted books.
    Public domain data may have been included. 
    This makes \textsc{Phi 4} and imperfect negative control, but we include it anyway as a basis for comparison. 
    \textsc{Phi 4} is similar in size to \textsc{Pythia 12B}, \textsc{Llama 1 13B}, and \textsc{Llama 2 13B}, which facilitates cross-model comparisons of extraction risk. (This is because the amount of memorization, reasonably, also varies according to model size; see also \citet{carlini2023quantifying}.) 
\end{enumerate}

\paragraph{Compute resources.}
We run all of our sliding-window experiments in a Slurm cluster environment, using the same node with $4$ A100 GPUs.
All told, this set of experiments took $394$ GPU days. 

\rowcolors{2}{gray!15}{white}
\setlength{\tabcolsep}{2pt}

\newcolumntype{L}[1]{>{\raggedright\arraybackslash\scriptsize}p{#1}}
\newcolumntype{C}[1]{>{\centering\arraybackslash\scriptsize}p{#1}}
\newcolumntype{T}[1]{>{\raggedright\arraybackslash\scriptsize}p{#1}}

\begin{center}
{\scriptsize
\begin{longtable}{C{0.35cm} L{2cm} L{3.25cm} C{0.7cm} C{1.2cm} C{1.1cm} T{4.5cm}}
\caption{\textbf{Books drawn from \texttt{Books3}.}
The $200$ books from \texttt{Books3} that we evaluate with our sliding-window probabilistic extraction procedure.
See Sections~\ref{sec:book-procedure} and~\ref{sec:book:compare}, as well as Appendices~\ref{app:sec:sliding-window:setup} \&~\ref{app:sec:sliding-window:procedure}.}
\label{app:tab:sliding-books}\\

\toprule
 & \textbf{Author} & \textbf{Title} & \textbf{Year} & \textbf{Status} & \textbf{Random} & \textbf{\texttt{Books3} path} \\
\midrule
\endfirsthead

\multicolumn{7}{l}{\footnotesize\emph{Table \thetable\ (continued). The 200 books drawn from \texttt{Books3}.}}\\
\toprule
 & \textbf{Author} & \textbf{Title} & \textbf{Year} & \textbf{Status} & \textbf{Random} & \textbf{\texttt{Books3} path} \\
\midrule
\endhead

\midrule
\multicolumn{7}{r}{\footnotesize\emph{Continued on next page}}\\
\endfoot

\bottomrule
\endlastfoot
1 & Achebe, Chinua & \textit{Things Fall Apart}~\citep{Things_Fall_Apart} & 1958 & \textcopyright & False & \ttfamily https:/\allowbreak /\allowbreak the-\allowbreak eye.\allowbreak eu/\allowbreak public/\allowbreak Books/\allowbreak Bibliotik/\allowbreak C/\allowbreak Chinua\%20\allowbreak Achebe\%20\allowbreak -\allowbreak \%20\allowbreak Things\%20\allowbreak Fall\%20\allowbreak Apart.\allowbreak epub \\
2 & Adams, Douglas & \textit{The Hitchhiker's Guide to the Galaxy - Omnibus}~\citep{The_Hitchhiker_s_Guide_to_the_Galaxy_-_Omnibus} & 1979 & \textcopyright & False & \ttfamily https:/\allowbreak /\allowbreak the-\allowbreak eye.\allowbreak eu/\allowbreak public/\allowbreak Books/\allowbreak Bibliotik/\allowbreak T/\allowbreak The\%20\allowbreak Hitchhiker\%27s\%20\allowbreak Guide\%20\allowbreak to\%20\allowbreak the\%20\allowbreak Galaxy\%20\allowbreak Omnibus\%20\allowbreak A\%20\allowbreak Trilogy\%20\allowbreak in\%20\allowbreak Five\%20\allowbreak Parts\%20\allowbreak -\allowbreak \%20\allowbreak Douglas\%20\allowbreak Adams.\allowbreak epub \\
3 & Adichie, Chimamanda Ngozi & \textit{Americanah}~\citep{Americanah} & 2013 & \textcopyright & False & \ttfamily https:/\allowbreak /\allowbreak the-\allowbreak eye.\allowbreak eu/\allowbreak public/\allowbreak Books/\allowbreak Bibliotik/\allowbreak A/\allowbreak Americanah-\allowbreak Adichie.\allowbreak epub \\
4 & Al Rawi, Shahad & \textit{The Baghdad Clock}~\citep{The_Baghdad_Clock} & 2016 & \textcopyright & True & \ttfamily https:/\allowbreak /\allowbreak the-\allowbreak eye.\allowbreak eu/\allowbreak public/\allowbreak Books/\allowbreak Bibliotik/\allowbreak T/\allowbreak The\%20\allowbreak Baghdad\%20\allowbreak Clock\%20\allowbreak -\allowbreak \%20\allowbreak Shahad\%20\allowbreak Al\%20\allowbreak Rawi.\allowbreak epub \\
\end{longtable}
}
\end{center}
\normalsize
\rowcolors{2}{white}{white}

\subsection{Sliding-window probabilistic extraction procedure}\label{app:sec:sliding-window:procedure}

In this appendix, we provide more details on our main extraction experiments (Section~\ref{sec:book-procedure}). 
These experiments take the approach of constructing $100$-token sequences $\vz$ out of \texttt{Books3} books, which we attempt to extract. 
These sequences are constructed using a sliding window across the length of an entire book, and we run these experiments for each of the $200$ books in Table~\ref{app:tab:sliding-books}. 
We use a very short, $10$-character shift for our sliding window. 
This is deliberate;
it is  meant to help us, effectively, ``pan for gold''---to identify regions within specific books where there are high-probability stretches of memorized content. 
The approach that we take reveals the position (in character space) in each book where these regions occur (and if they occur at all). 
We explain the intention behind these experiments, and how the associated results should be interpreted.\looseness=-1 

\paragraph{Sequence length.}
We use $100$-token sequences $\vz$, with a prefix (prompt) length of $50$ tokens and a (target) suffix length of $50$ tokens.
We pick $100$ tokens and a $50$/$50$ split because this is accepted in the literature as a reasonable minimum for being confident that extracted suffixes are reflective of memorization~\citep{carlini2023quantifying, carlini2021extracting, lee2022dedup, hayes2025measuringmemorizationlanguagemodels}. 
$50$ tokens for the prefix is the standard choice for $50$-token suffixes, though is somewhat arbitrary.
Of course, this prefix length uses an amount of context that is equivalent to the output, and so is compelling in this respect. 
However, when we are confident that our measurements are capturing valid instances of extraction, using longer prefixes can help \emph{discover} more underlying memorization. 
(Indeed, this is part of where \emph{discoverable} extraction gets its name; 
\citet{carlini2023quantifying} demonstrate the ``discoverability phenomenon'' with respect to showing how extraction success can increase with longer-context prompts.) 
We discuss this further (and include experiments with varied prefix lengths) in Section~\ref{sec:validity} and Appendix~\ref{app:sec:validity}.
Other work uses shorter sequences (e.g., $64$ tokens and a $32$/$32$ token split; see ~\citet{biderman2023pythia}).\looseness=-1 

\paragraph{Sliding-window sequence chunking.}
For a given book, we start at the beginning of the text file in \texttt{Books3}. 
We sample a chunk of text that is sufficiently long to contain $100$ tokens of corresponding tokenized text: 
to do this, we take a chunk of $800$ characters, tokenize, take the first $100$ tokens of the resulting tokenized sequence as the sequence, and discard the rest.
(Each $100$ token sequence is typically, but not always, reflective of $300$-$400$ characters of text.
Rare words and formatting make this vary considerably.
This is why we use a character-chunk size of $800$---to make sure we have sufficient head-room to always end up with a $100$ token sequence.)
We then shift $10$ \emph{characters} forward in the book text and repeat this process.
We do this for the entire length of the book, which results in approximately one sequence every $10$ characters, i.e., $\texttt{len(book\_characters)} / 10$ total sequences $\vz$, for which we compute $p_\vz$. 
For example, \emph{The Great Gatsby} has $270{,}870$ characters, which results in roughly $27{,}000$ sequences, for which we compute $p_\vz$ (Equation~\ref{app:eq:npmem}).
(See Section~\ref{sec:background} and Appendix~\ref{app:sec:background:metrics}.)\looseness=-1 

This means that the $100$-token sequences overlap significantly. 
(Generally speaking, $100$ tokens covers $300$-$400$ characters, so shifting only $10$ characters means there is a lot of overlap.) 
This is deliberate. 
We typically do not know how different open-weight models were trained; 
\emph{a priori}, it is not clear exactly where we should begin sequences (or where they should end). 
With this approach, for extractable content, we expect to surface high-probability \emph{regions} of memorized content, which we can then explore in more detail and more precisely in follow-up experiments. 
We picked a sliding window of $10$ characters by running initial experiments with $1$-, $10$-, $50$-, and $100$-character windows.
$10$ characters exhibited a small loss in extraction signal (at $10\times$ cheaper cost than $1$ character).\looseness=-1 


\paragraph{Decoding configuration}
Similar to  \citet{hayes2025measuringmemorizationlanguagemodels}, after some initial tests with different decoding schemes, we use temperature $T\!=\!1$ for top-$k$ decoding with  $k\!=\!40$.
Setting $T\!=\!1$ makes the most sense for studying memorization, as this reflects the LLM's base probability distribution. 
Since extraction is only successful when the underlying training sequence's logits reflect very high probabilities (Section~\ref{sec:background}), it makes sense to clip the full-vocabulary logit distribution to reflect some number of the top-highest probability tokens per step. 
This is the same intuition behind using greedy decoding for discoverable extraction---but with a twist. 
Greedy decoding deterministically picks the single-most-likely token per step. 
(This is equivalent to top-$k$ decoding with $k\!=\!1$.)
As the results in \citet{hayes2025measuringmemorizationlanguagemodels} show, this can be a suboptimal decoding strategy for measuring extraction, for several reasons:
\begin{itemize}[leftmargin=.65cm]
    \item Greedy decoding is rarely used with LLMs in practice. 
    Non-deterministic sampling schemes are  more reflective of realistic conditions. 
    Because greedy decoding is deterministic, the same prompt always results in the same output. 
    As a result, discoverable extraction provides (in this sense) $1$-bit of information with respect to extraction: 
    was it possible to extract the sequence?
    Using non-deterministic decoding schemes, we can get more detailed information about how extraction risk varies across sequences.  
    (See Section~\ref{sec:background:pz-meaning} \& Appendix~\ref{app:sec:background}.)
    \item While greedy decoding picks the \emph{locally} highest probability token per step, it is not guaranteed to result in the \emph{globally} highest probability sequence (of the given length) under the model. 
    Being able to explore more of the model's distribution with non-deterministic decoding schemes can surface sequences (of the same given length) that are higher probability. 
    Such higher probability sequences can match the target suffix, where the greedy-decoded suffix does not (i.e., fails to extract that suffix).
    \item More generally, this allows for a bit more ``wiggle room.'' 
    By being able to pick beyond just the top-$1$ token per step, the decoding process take locally suboptimal steps, which can still result in high-probability sequences that (even if slightly lower probability than the greedy sequence, with respect to the normalized top-$k$ distribution) reflect extraction.
    While we do not investigate this in detail in this work, we observe that (when this happens) the greedy-decoded suffix is a near-miss for the verbatim suffix (it is near-verbatim). 
    It may be reasonable under more relaxed (near-verbatim) metrics for extraction to have counted this greedy suffix as extracted, but that is not the approach we (or \citet{hayes2025measuringmemorizationlanguagemodels}) take, and so these suffixes would be missed. 
\end{itemize}

The above reasons also reflect the general motivations for using probabilistic extraction  as our metric  (Sections~\ref{sec:background} \&~\ref{sec:copyright}; Appendix~\ref{app:sec:background}). 
Similar motivations for useful extraction measurements are also noted in early extraction papers (e.g., \citet{carlini2021extracting}, \citet{carlini2023quantifying}); 
however, this work typically opts to use greedy decoding based on the perception that it is cheaper to do so at scale. 
This work does not claim that greedy decoding is the best or only way to measure extraction, and in fact notes~\citep{carlini2021extracting}---and in select cases shows~\citep{carlini2023quantifying}---that other (more expensive) decoding schemes are able to extract more training data.\looseness=-1 

Further, it turns out that it is actually \emph{cheaper} to compute probabilistic extraction (with any top-$k$ decoding configuration, including greedy/ $k\!=\!1$), as we compute extraction probabilities as a post-processing operation directly off of the logits obtained from \emph{inference}, rather than (as has been done in prior work~\citep{carlini2021extracting, carlini2023quantifying}) actually generating (greedy) completions.
We can compute verbatim extraction probabilities with a single forward pass through the model, which has reduced overhead compared to greedy generation (Appendix~\ref{app:sec:background:compute}). 

Note that because our implementation of probabilistic  extraction makes computing $\log$ probabilities a post-processing operation on the output model logits (Appendix~\ref{app:sec:background}), we also collect results for greedy-decoded discoverable extraction by setting $k\!=\!1$, incurring negligible cost to do so. 
For brevity, we do not include detailed results for this setting.
They under-count valid extraction, aligning with our findings in Section~\ref{sec:book-procedure:averages} and Appendix~\ref{app:sec:rates}, as well as \citet{hayes2025measuringmemorizationlanguagemodels}.  

\subsection{Book-specific results}\label{app:sec:sliding-window:results}

For each of the $200$ books we test, we include a separate subsection with the following information for $14$ models (documented in Appendix~\ref{app:sec:sliding-window:setup}):
\begin{itemize}[leftmargin=0.65cm]
    \item \textbf{Heatmaps of extraction probabilities by location.}  
    We include heatmaps (as in Figure~\ref{fig:1984:slide:main:heatmap} and Section~\ref{sec:book:compare}). 
    At each character, the heatmap plots the maximum extraction probability among overlapping $50$-token suffixes spanning that character. 
    These visualizations highlight ``hot-spots'' of high-probability extraction and show how such regions emerge and taper off within a book.

    \item \textbf{Histograms of extraction probabilities.}  
    We plot the distribution over suffix extraction probabilities across all overlapping sequences from the sliding-window procedure (with a $10$-character stride).  
    These histograms illustrate how many sequences are extractable for a given model and how their extraction probabilities vary.
\end{itemize}

For histograms and heatmaps, results are presented in three groups corresponding to the categories of models described in Appendix~\ref{app:sec:sliding-window:setup}: 
models known with certainty to have been trained on \texttt{Books3} (\textcolor{seabornbluemid}{blue}), 
models that were likely trained on books also present in \texttt{Books3} (\textcolor{seabornredmid}{red}), 
and one model (\textsc{Phi 4}) that, based on documentation, was not trained on whole copyrighted books (\textcolor{seabornorangemid}{orange}). 
See Section~\ref{sec:validity} and Appendix~\ref{app:sec:validity} for further discussion about how our results for all three categories reflect valid extraction.\looseness=-1

We release an interactive, searchable version of these results \href{http://books-memorization.github.io}{online}.

\clearpage
\subsubsection{\textit{Things Fall Apart}, \citeauthor{Things_Fall_Apart}}\label{app:sec:sliding:Things_Fall_Apart}
\vspace{-.2cm}
\begin{figure}[h]
  \centering
  \begin{minipage}[t]{0.53\textwidth}
    \centering
    \vspace{0cm}
    \includegraphics[width=\linewidth]{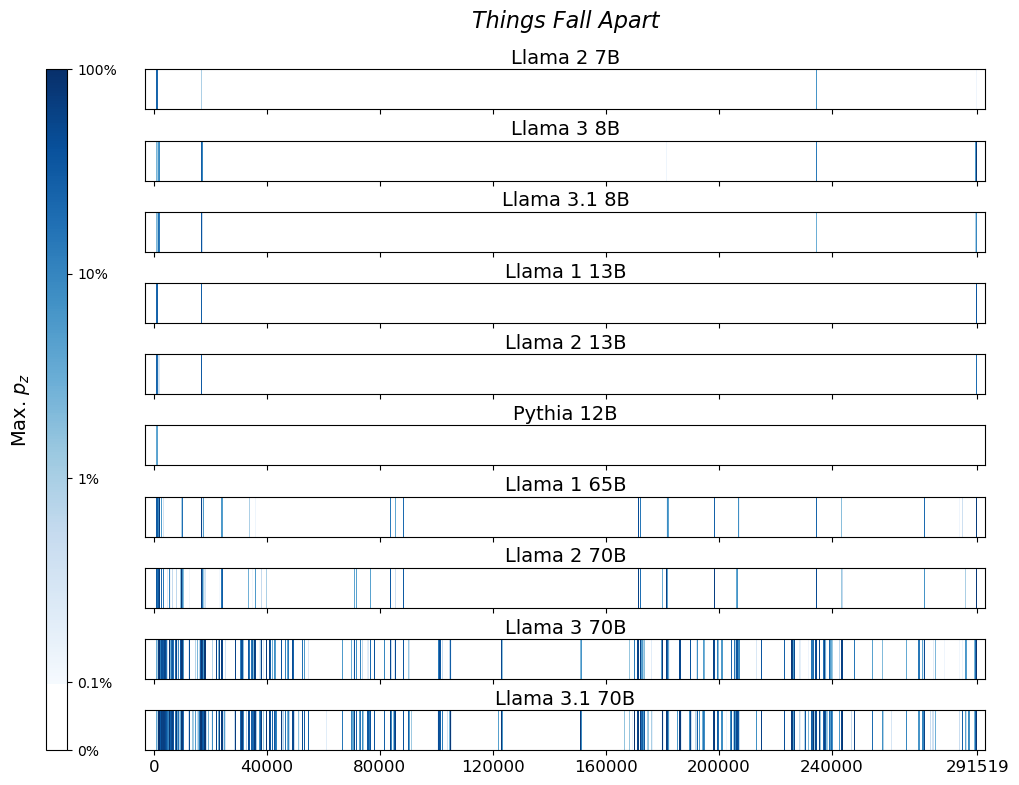}
    \includegraphics[width=\linewidth]{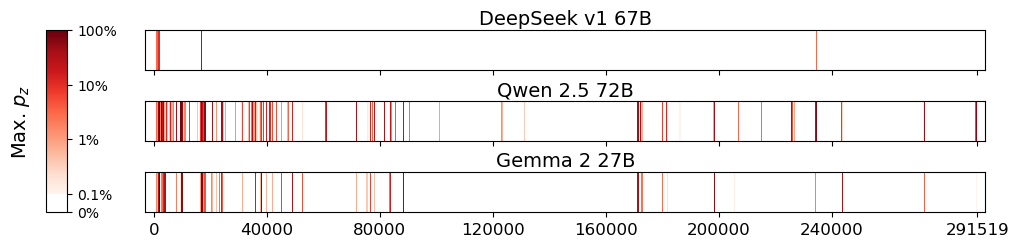}
    \includegraphics[width=\linewidth]{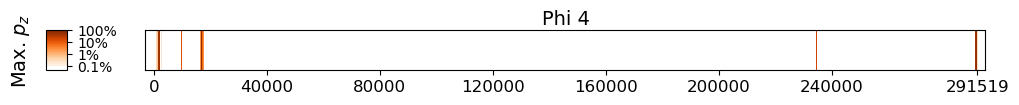}
  \end{minipage}
  \hfill
  \begin{minipage}[t]{0.45\textwidth}
    \centering
    \vspace{0cm}
    \includegraphics[width=\linewidth]{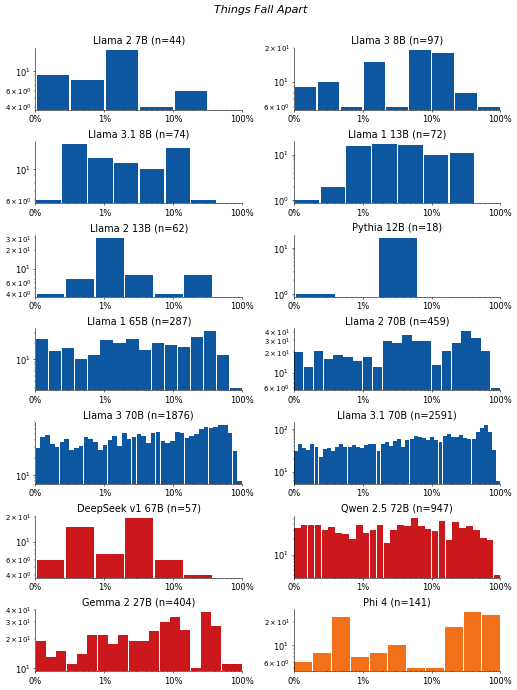}
  \end{minipage}
  \vspace{-.2cm}
  \caption{
    \textbf{\textit{Things Fall Apart}, \citeauthor{Things_Fall_Apart}.}
    For $14$ LLMs,
    (\textbf{left}) heatmaps for the sliding-window procedure and
    (\textbf{right}) corresponding distributions over suffix extraction probabilities
    ($\tau_\text{min}=0.1\%$).
  }
  \label{fig:slidingwindow:Things_Fall_Apart}
\end{figure}
\FloatBarrier

\subsubsection{\textit{The Hitchhiker's Guide to the Galaxy - Omnibus}, \citeauthor{The_Hitchhiker_s_Guide_to_the_Galaxy_-_Omnibus}}\label{app:sec:sliding:The_Hitchhiker_s_Guide_to_the_Galaxy_-_Omnibus}
\vspace{-.2cm}
\begin{figure}[h]
  \centering
  \begin{minipage}[t]{0.53\textwidth}
    \centering
    \vspace{0cm}
    \includegraphics[width=\linewidth]{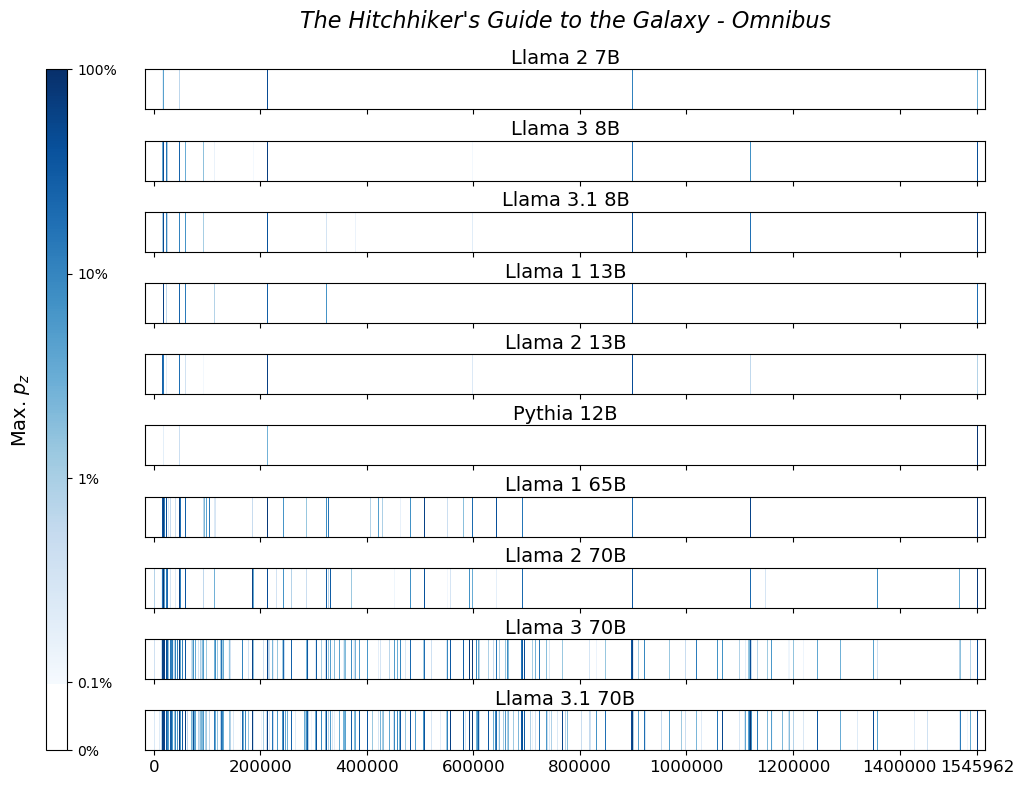}
    \includegraphics[width=\linewidth]{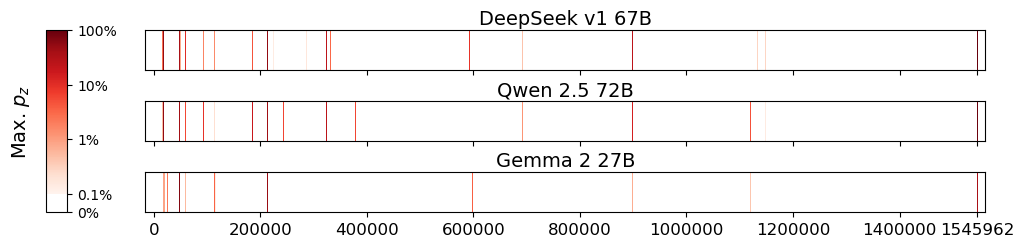}
    \includegraphics[width=\linewidth]{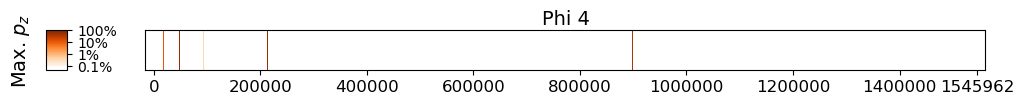}
  \end{minipage}
  \hfill
  \begin{minipage}[t]{0.45\textwidth}
    \centering
    \vspace{0cm}
    \includegraphics[width=\linewidth]{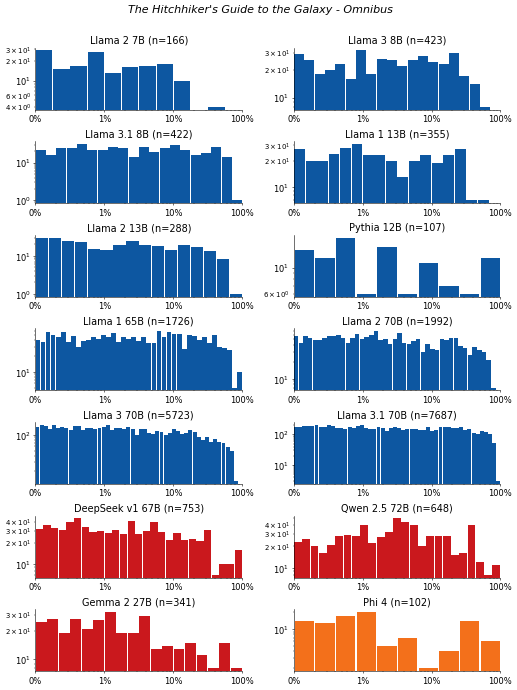}
  \end{minipage}
  \vspace{-.2cm}
  \caption{
    \textbf{\textit{The Hitchhiker's Guide to the Galaxy - Omnibus}, \citeauthor{The_Hitchhiker_s_Guide_to_the_Galaxy_-_Omnibus}.}
    For $14$ LLMs,
    (\textbf{left}) heatmaps for the sliding-window procedure and
    (\textbf{right}) corresponding distributions over suffix extraction probabilities
    ($\tau_\text{min}=0.1\%$).
  }
  \label{fig:slidingwindow:The_Hitchhiker_s_Guide_to_the_Galaxy_-_Omnibus}
\end{figure}
\FloatBarrier

\clearpage
\subsubsection{\textit{Americanah}, \citeauthor{Americanah}}\label{app:sec:sliding:Americanah}
\vspace{-.2cm}
\begin{figure}[h]
  \centering
  \begin{minipage}[t]{0.53\textwidth}
    \centering
    \vspace{0cm}
    \includegraphics[width=\linewidth]{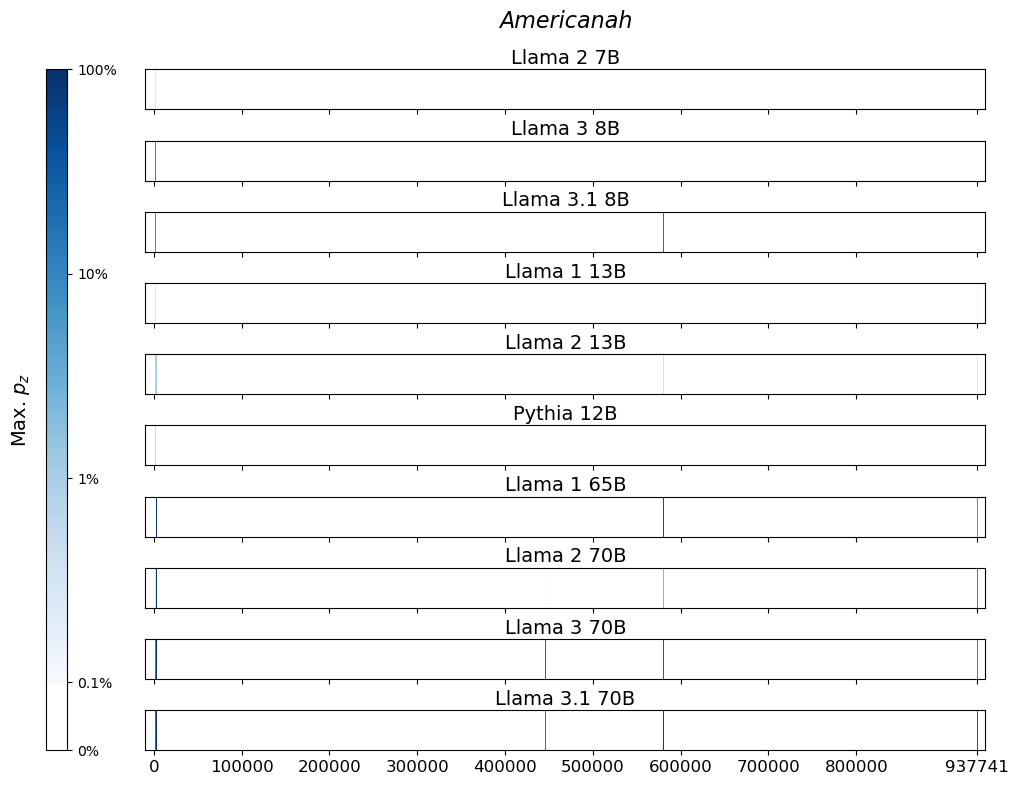}
    \includegraphics[width=\linewidth]{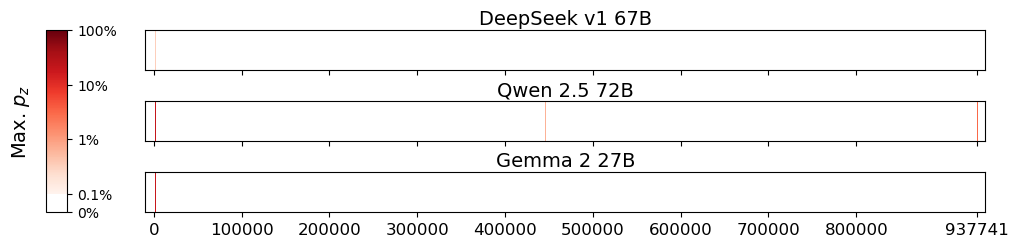}
    \includegraphics[width=\linewidth]{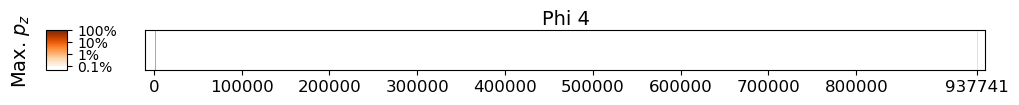}
  \end{minipage}
  \hfill
  \begin{minipage}[t]{0.45\textwidth}
    \centering
    \vspace{0cm}
    \includegraphics[width=\linewidth]{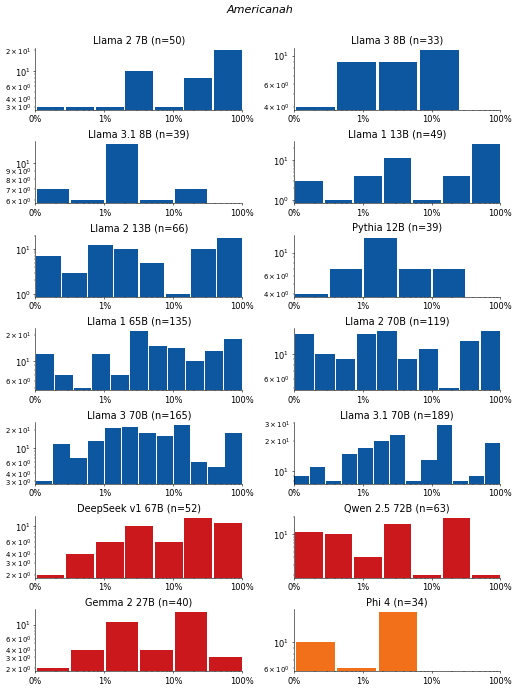}
  \end{minipage}
  \vspace{-.2cm}
  \caption{
    \textbf{\textit{Americanah}, \citeauthor{Americanah}.}
    For $14$ LLMs,
    (\textbf{left}) heatmaps for the sliding-window procedure and
    (\textbf{right}) corresponding distributions over suffix extraction probabilities
    ($\tau_\text{min}=0.1\%$).
  }
  \label{fig:slidingwindow:Americanah}
\end{figure}
\FloatBarrier

\subsubsection{\textit{The Baghdad Clock}, \citeauthor{The_Baghdad_Clock}}\label{app:sec:sliding:The_Baghdad_Clock}
\vspace{-.2cm}
\begin{figure}[h]
  \centering
  \begin{minipage}[t]{0.53\textwidth}
    \centering
    \vspace{0cm}
    \includegraphics[width=\linewidth]{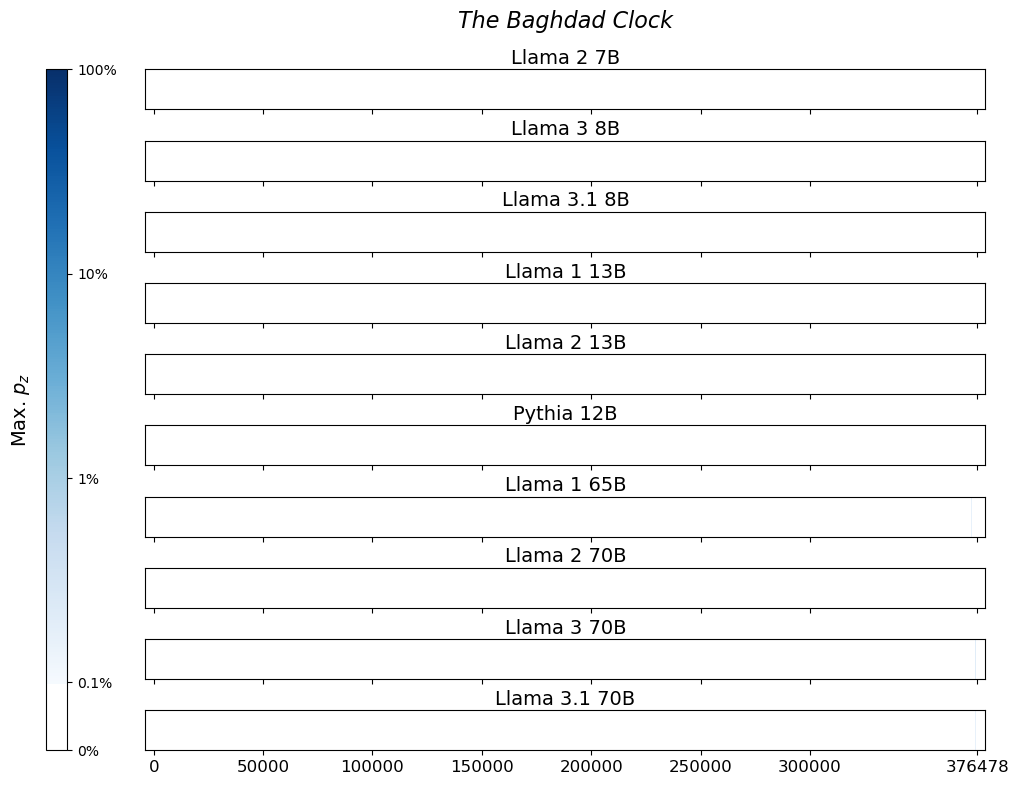}
    \includegraphics[width=\linewidth]{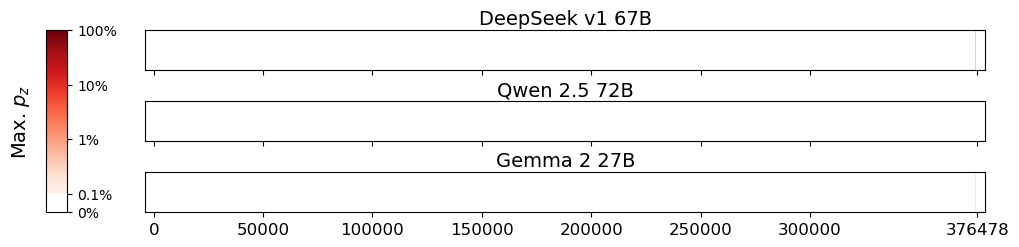}
    \includegraphics[width=\linewidth]{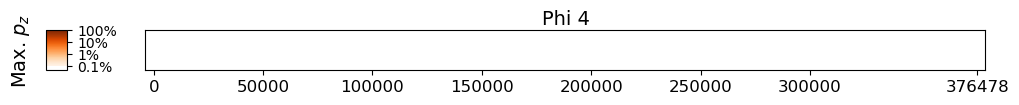}
  \end{minipage}
  \hfill
  \begin{minipage}[t]{0.45\textwidth}
    \centering
    \vspace{0cm}
    \includegraphics[width=\linewidth]{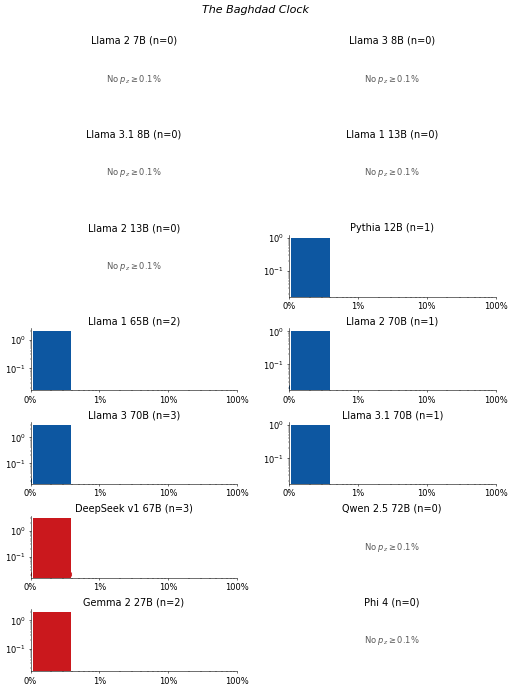}
  \end{minipage}
  \vspace{-.2cm}
  \caption{
    \textbf{\textit{The Baghdad Clock}, \citeauthor{The_Baghdad_Clock}.}
    For $14$ LLMs,
    (\textbf{left}) heatmaps for the sliding-window procedure and
    (\textbf{right}) corresponding distributions over suffix extraction probabilities
    ($\tau_\text{min}=0.1\%$).
  }
  \label{fig:slidingwindow:The_Baghdad_Clock}
\end{figure}
\FloatBarrier

\clearpage
\subsubsection{\textit{Industrial Magic}, \citeauthor{Industrial_Magic}}\label{app:sec:sliding:Industrial_Magic}
\vspace{-.2cm}
\begin{figure}[h]
  \centering
  \begin{minipage}[t]{0.53\textwidth}
    \centering
    \vspace{0cm}
    \includegraphics[width=\linewidth]{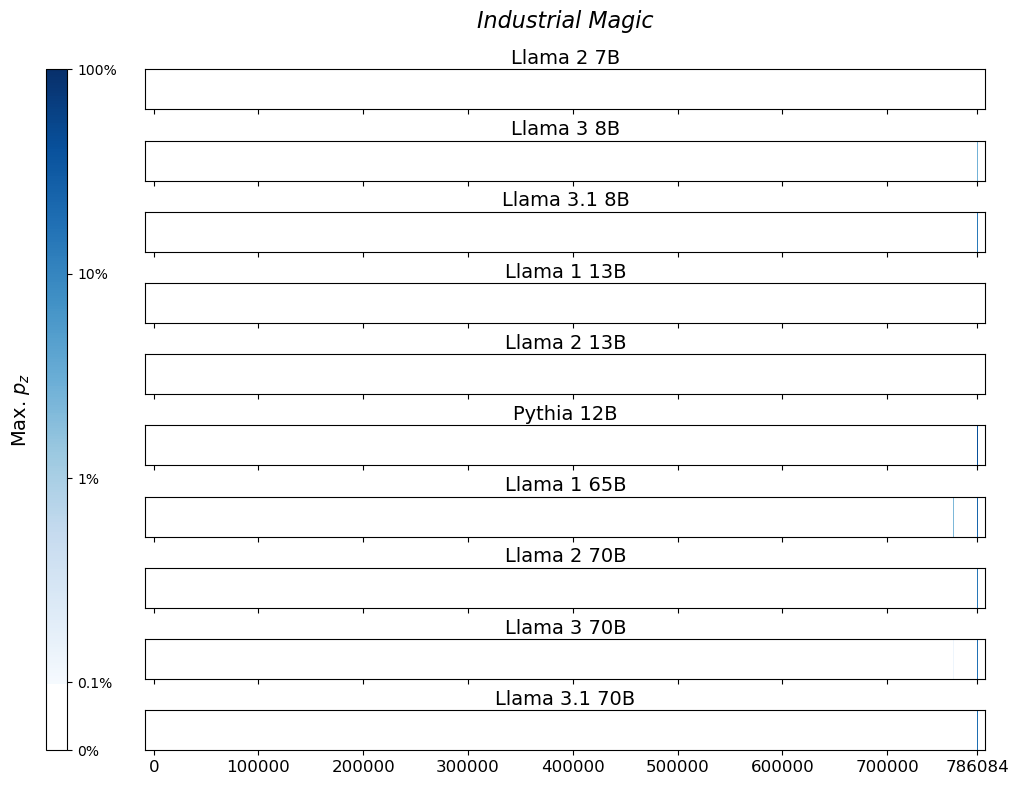}
    \includegraphics[width=\linewidth]{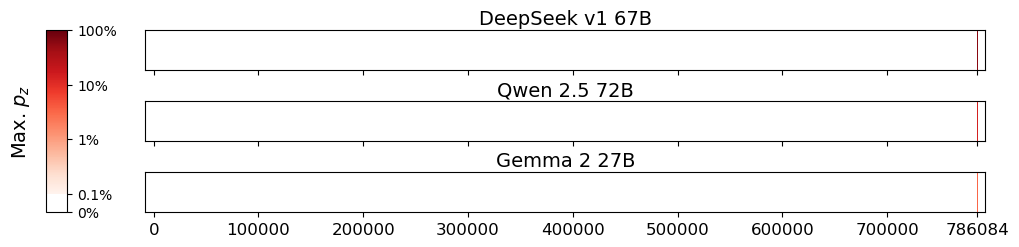}
    \includegraphics[width=\linewidth]{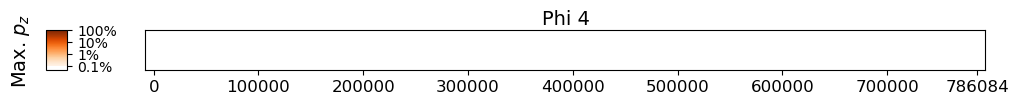}
  \end{minipage}
  \hfill
  \begin{minipage}[t]{0.45\textwidth}
    \centering
    \vspace{0cm}
    \includegraphics[width=\linewidth]{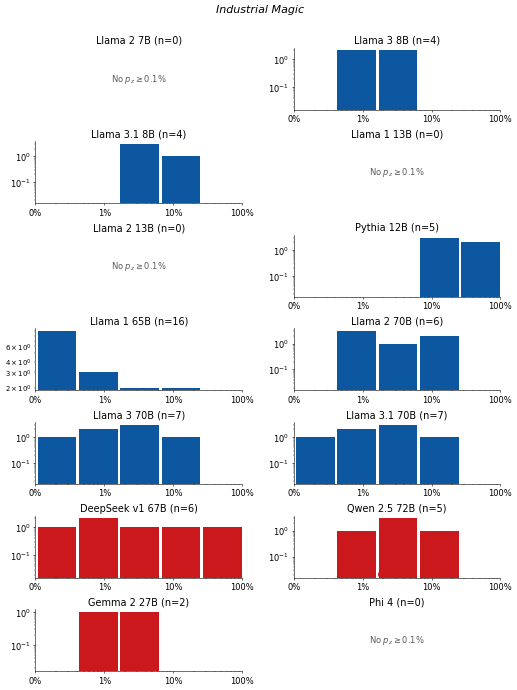}
  \end{minipage}
  \vspace{-.2cm}
  \caption{
    \textbf{\textit{Industrial Magic}, \citeauthor{Industrial_Magic}.}
    For $14$ LLMs,
    (\textbf{left}) heatmaps for the sliding-window procedure and
    (\textbf{right}) corresponding distributions over suffix extraction probabilities
    ($\tau_\text{min}=0.1\%$).
  }
  \label{fig:slidingwindow:Industrial_Magic}
\end{figure}
\FloatBarrier

\subsubsection{\textit{Fantastic Voyage}, \citeauthor{Fantastic_Voyage}}\label{app:sec:sliding:Fantastic_Voyage}
\begin{figure}[h]
  \vspace{-.2cm}
  \centering
  \begin{minipage}[t]{0.53\textwidth}
    \centering
    \vspace{0cm}
    \includegraphics[width=\linewidth]{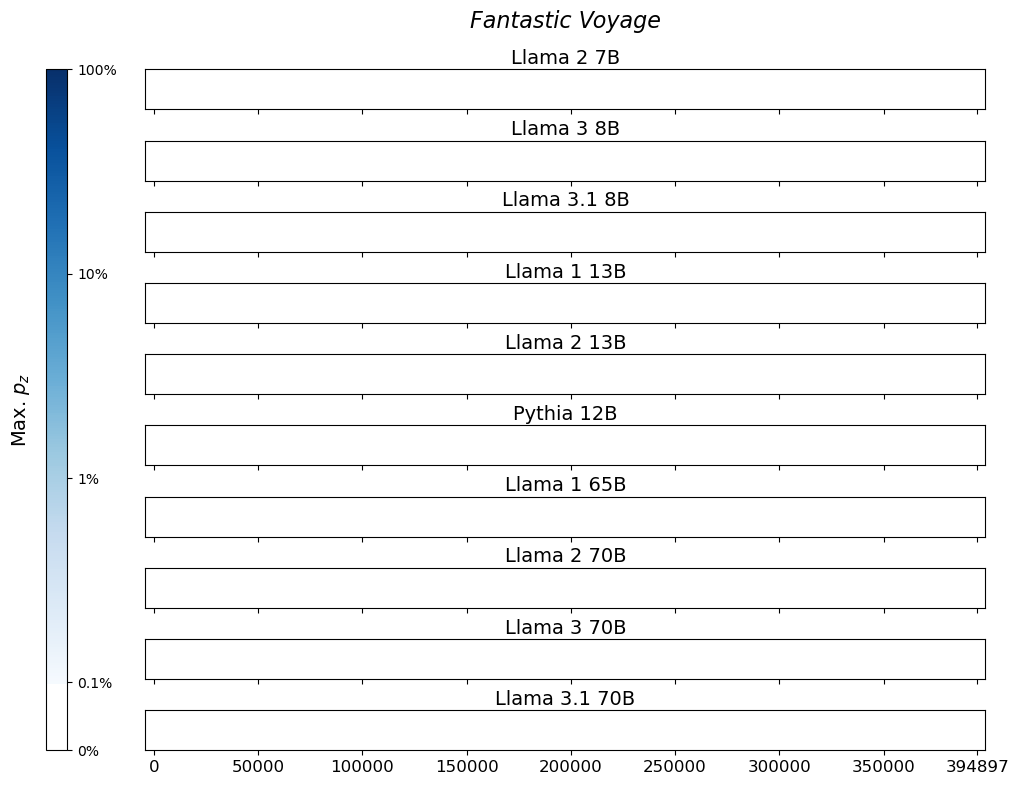}
    \includegraphics[width=\linewidth]{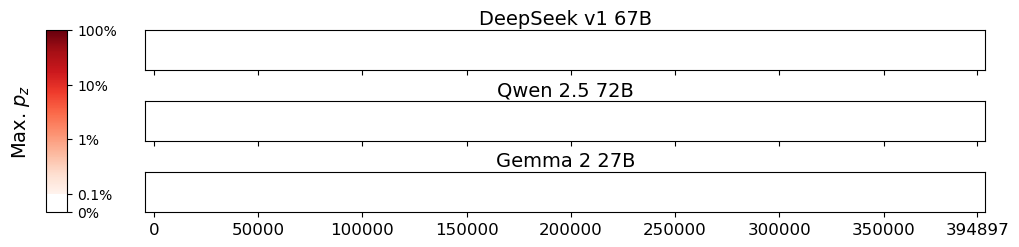}
    \includegraphics[width=\linewidth]{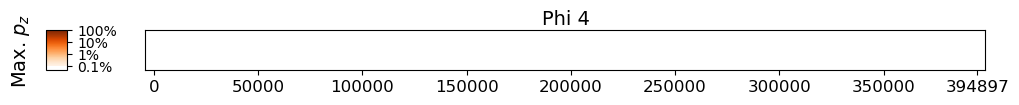}
  \end{minipage}
  \hfill
  \begin{minipage}[t]{0.45\textwidth}
    \centering
    \vspace{0cm}
    \includegraphics[width=\linewidth]{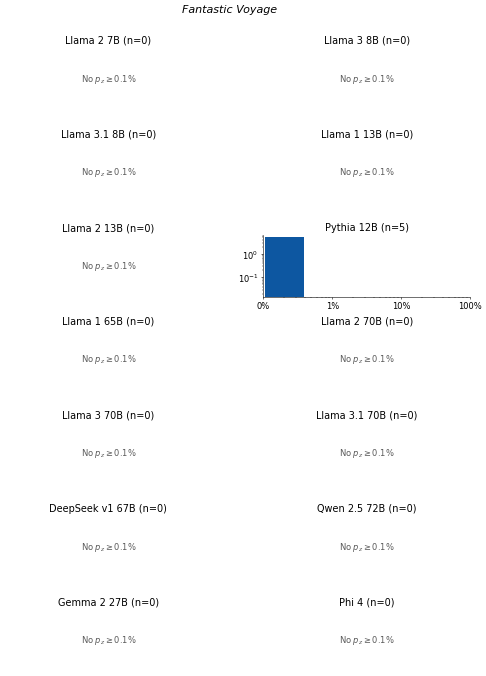}
  \end{minipage}
  \vspace{-.2cm}
  \caption{
    \textbf{\textit{Fantastic Voyage}, \citeauthor{Fantastic_Voyage}.}
    For $14$ LLMs,
    (\textbf{left}) heatmaps for the sliding-window procedure and
    (\textbf{right}) corresponding distributions over suffix extraction probabilities
    ($\tau_\text{min}=0.1\%$).
  }
  \label{fig:slidingwindow:Fantastic_Voyage}
\end{figure}
\FloatBarrier

\clearpage
\subsubsection{\textit{The Complete Robot}, \citeauthor{The_Complete_Robot}}\label{app:sec:sliding:The_Complete_Robot}
\vspace{-.2cm}
\begin{figure}[h]
  \centering
  \begin{minipage}[t]{0.53\textwidth}
    \centering
    \vspace{0cm}
    \includegraphics[width=\linewidth]{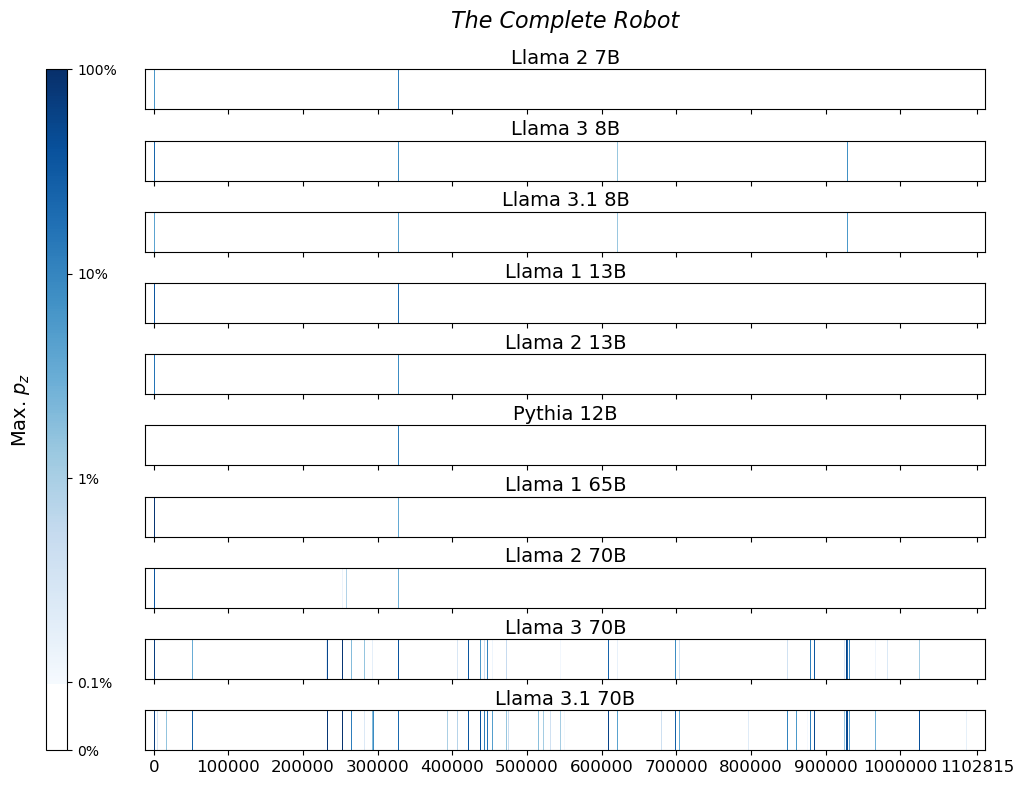}
    \includegraphics[width=\linewidth]{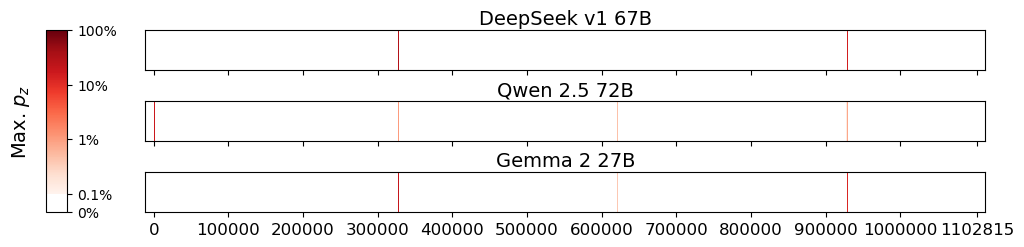}
    \includegraphics[width=\linewidth]{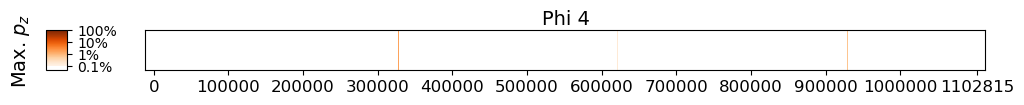}
  \end{minipage}
  \hfill
  \begin{minipage}[t]{0.45\textwidth}
    \centering
    \vspace{0cm}
    \includegraphics[width=\linewidth]{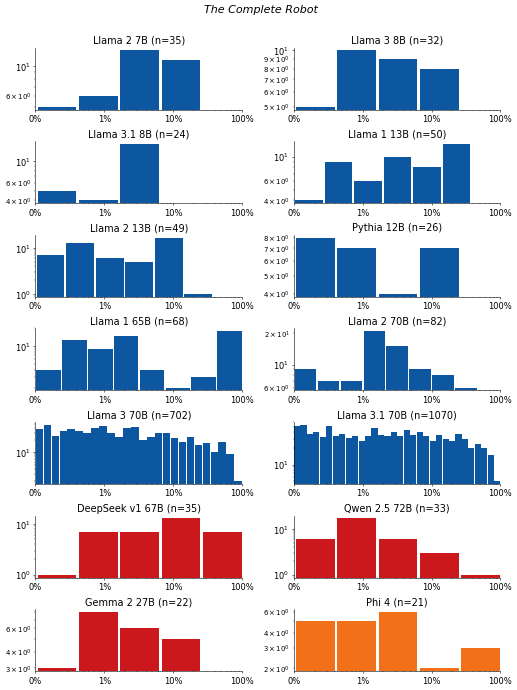}
  \end{minipage}
  \vspace{-.2cm}
  \caption{
    \textbf{\textit{The Complete Robot}, \citeauthor{The_Complete_Robot}.}
    For $14$ LLMs,
    (\textbf{left}) heatmaps for the sliding-window procedure and
    (\textbf{right}) corresponding distributions over suffix extraction probabilities
    ($\tau_\text{min}=0.1\%$).
  }
  \label{fig:slidingwindow:The_Complete_Robot}
\end{figure}
\FloatBarrier

\subsubsection{\textit{The Handmaid's Tale}, \citeauthor{The_Handmaid_s_Tale}}\label{app:sec:sliding:The_Handmaid_s_Tale}
\vspace{-.2cm}
\begin{figure}[h]
  \centering
  \begin{minipage}[t]{0.53\textwidth}
    \centering
    \vspace{0cm}
    \includegraphics[width=\linewidth]{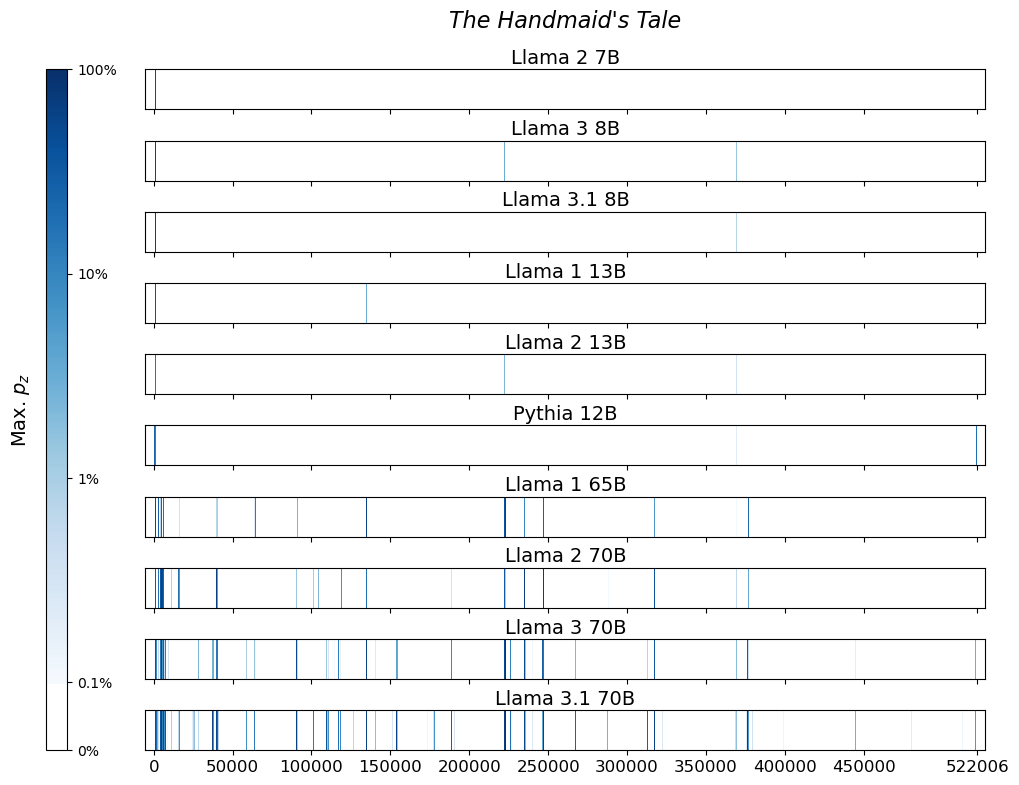}
    \includegraphics[width=\linewidth]{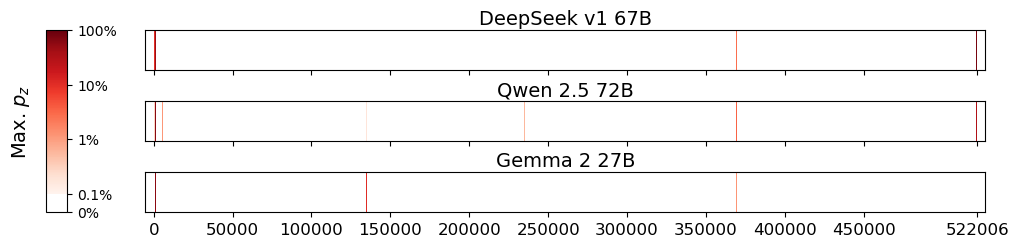}
    \includegraphics[width=\linewidth]{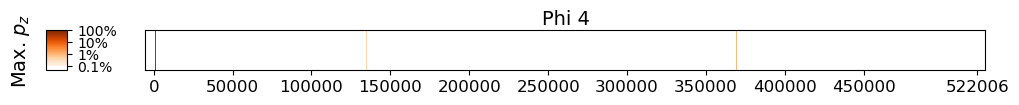}
  \end{minipage}
  \hfill
  \begin{minipage}[t]{0.45\textwidth}
    \centering
    \vspace{0cm}
    \includegraphics[width=\linewidth]{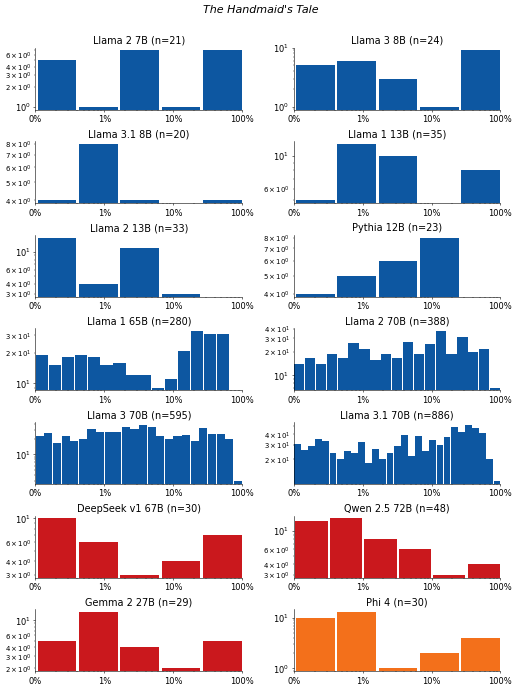}
  \end{minipage}
  \vspace{-.2cm}
  \caption{
    \textbf{\textit{The Handmaid's Tale}, \citeauthor{The_Handmaid_s_Tale}.}
    For $14$ LLMs,
    (\textbf{left}) heatmaps for the sliding-window procedure and
    (\textbf{right}) corresponding distributions over suffix extraction probabilities
    ($\tau_\text{min}=0.1\%$).
  }
  \label{fig:slidingwindow:The_Handmaid_s_Tale}
\end{figure}
\FloatBarrier

\clearpage
\subsubsection{\textit{Pride and Prejudice}, \citeauthor{Pride_and_Prejudice}}\label{app:sec:sliding:Pride_and_Prejudice}
\vspace{-.2cm}
\begin{figure}[h]
  \centering
  \begin{minipage}[t]{0.53\textwidth}
    \centering
    \vspace{0cm}
    \includegraphics[width=\linewidth]{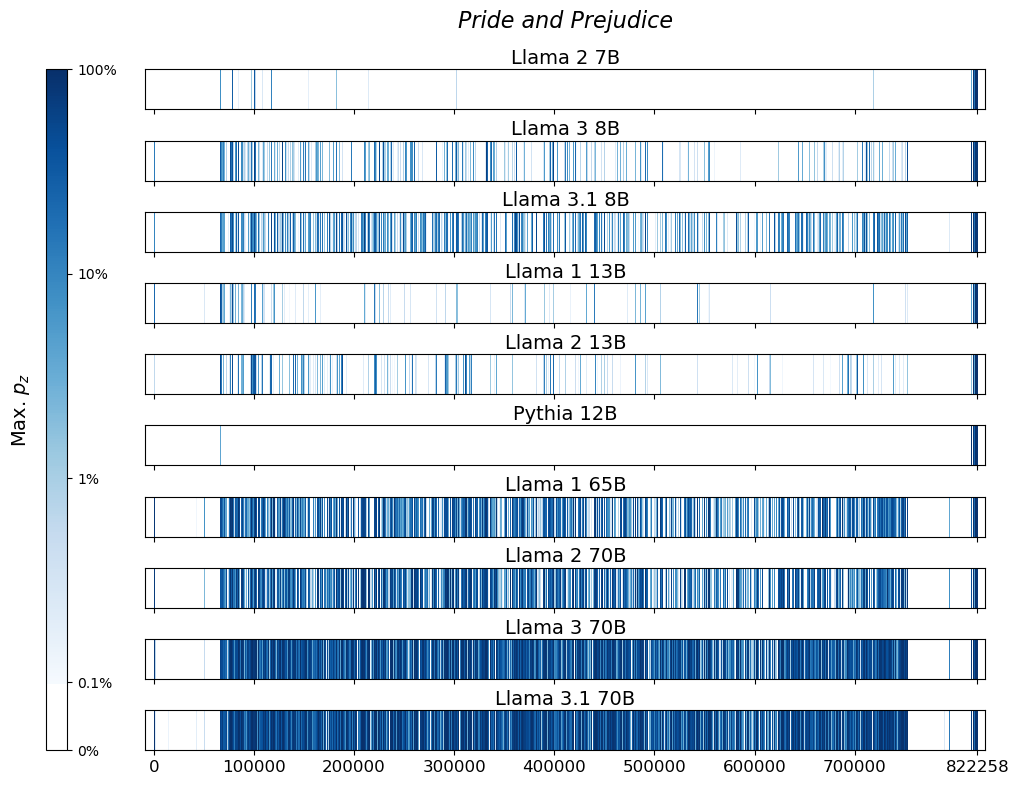}
    \includegraphics[width=\linewidth]{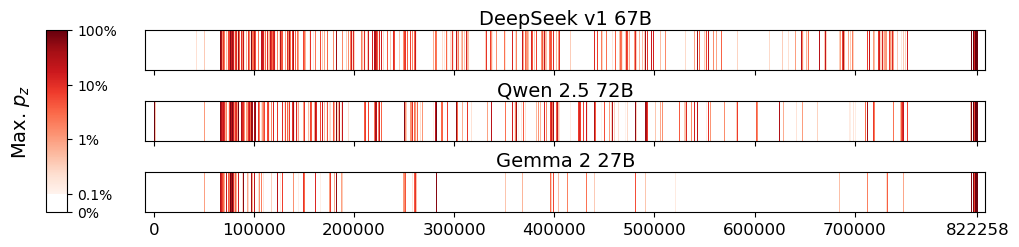}
    \includegraphics[width=\linewidth]{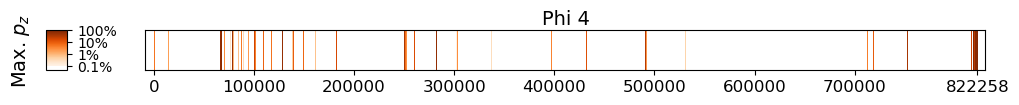}
  \end{minipage}
  \hfill
  \begin{minipage}[t]{0.45\textwidth}
    \centering
    \vspace{0cm}
    \includegraphics[width=\linewidth]{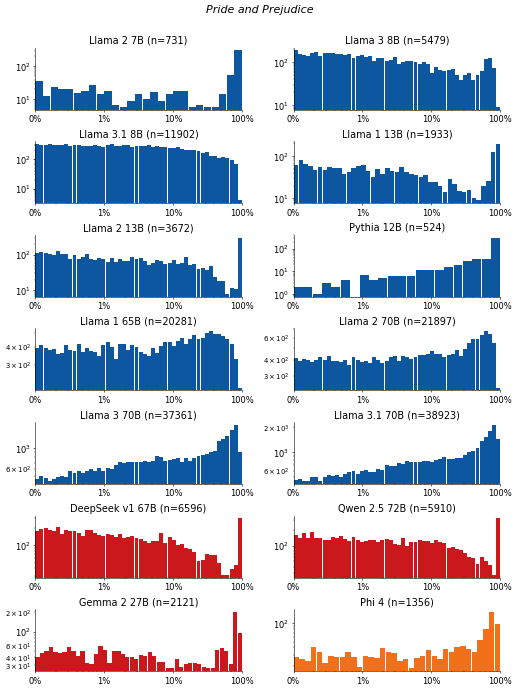}
  \end{minipage}
  \vspace{-.2cm}
  \caption{
    \textbf{\textit{Pride and Prejudice}, \citeauthor{Pride_and_Prejudice}.}
    For $14$ LLMs,
    (\textbf{left}) heatmaps for the sliding-window procedure and
    (\textbf{right}) corresponding distributions over suffix extraction probabilities
    ($\tau_\text{min}=0.1\%$).
  }
  \label{fig:slidingwindow:Pride_and_Prejudice}
\end{figure}
\FloatBarrier

\subsubsection{\textit{The Christmas Train}, \citeauthor{The_Christmas_Train}}\label{app:sec:sliding:The_Christmas_Train}
\vspace{-.2cm}
\begin{figure}[h]
  \centering
  \begin{minipage}[t]{0.53\textwidth}
    \centering
    \vspace{0cm}
    \includegraphics[width=\linewidth]{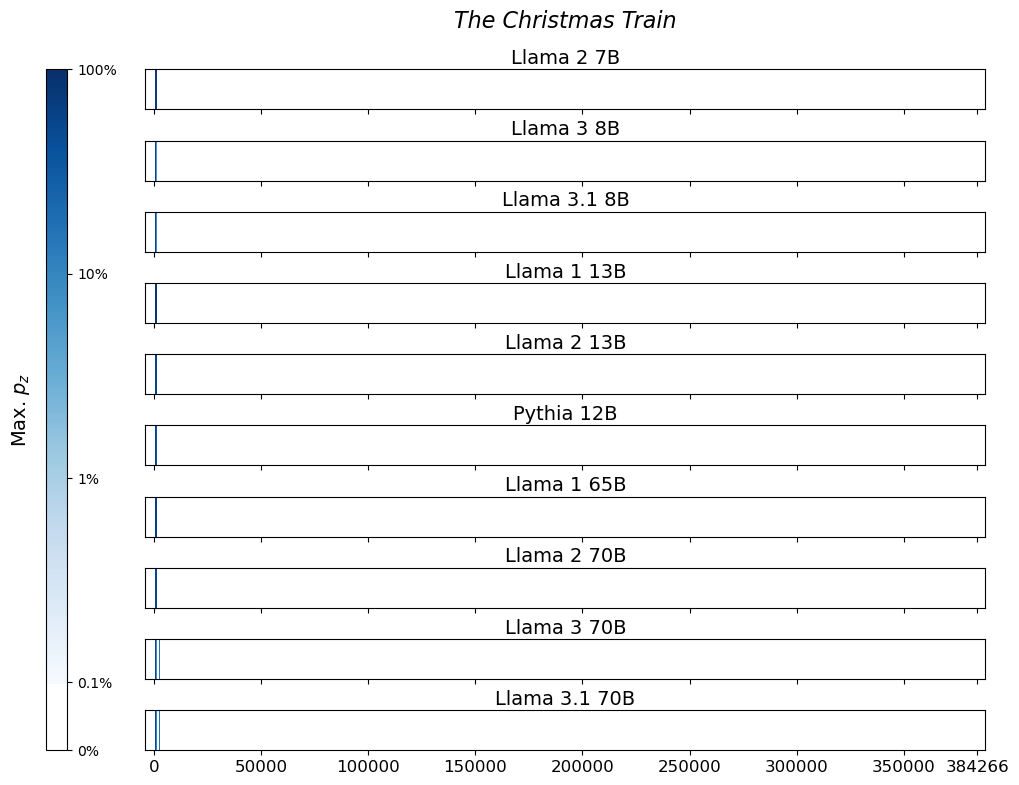}
    \includegraphics[width=\linewidth]{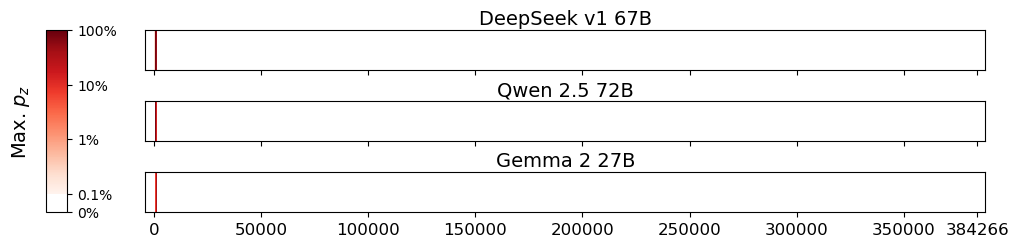}
    \includegraphics[width=\linewidth]{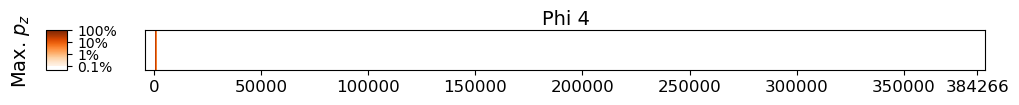}
  \end{minipage}
  \hfill
  \begin{minipage}[t]{0.45\textwidth}
    \centering
    \vspace{0cm}
    \includegraphics[width=\linewidth]{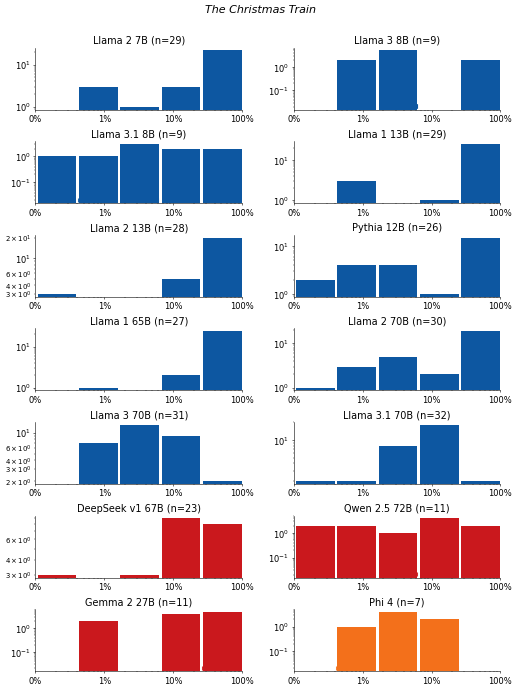}
  \end{minipage}
  \vspace{-.2cm}
  \caption{
    \textbf{\textit{The Christmas Train}, \citeauthor{The_Christmas_Train}.}
    For $14$ LLMs,
    (\textbf{left}) heatmaps for the sliding-window procedure and
    (\textbf{right}) corresponding distributions over suffix extraction probabilities
    ($\tau_\text{min}=0.1\%$).
  }
  \label{fig:slidingwindow:The_Christmas_Train}
\end{figure}
\FloatBarrier

\clearpage
\subsubsection{\textit{Notes of a Native Son}, \citeauthor{Notes_of_a_Native_Son}}\label{app:sec:sliding:Notes_of_a_Native_Son}
\vspace{-.2cm}
\begin{figure}[h]
  \centering
  \begin{minipage}[t]{0.53\textwidth}
    \centering
    \vspace{0cm}
    \includegraphics[width=\linewidth]{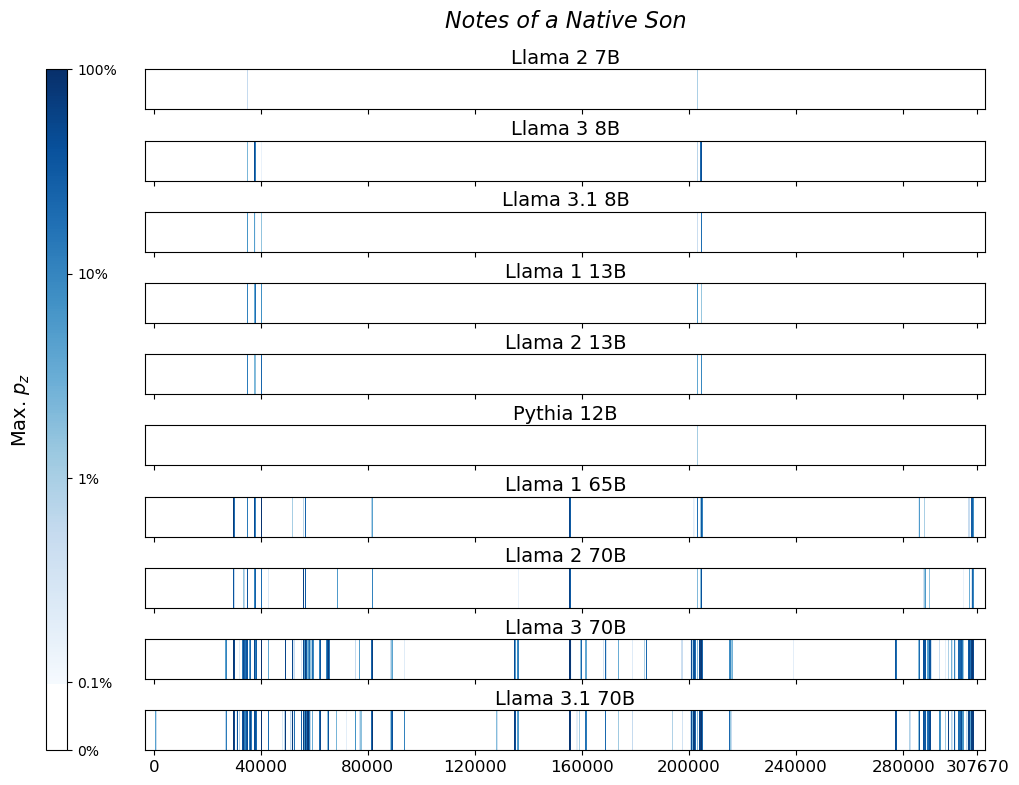}
    \includegraphics[width=\linewidth]{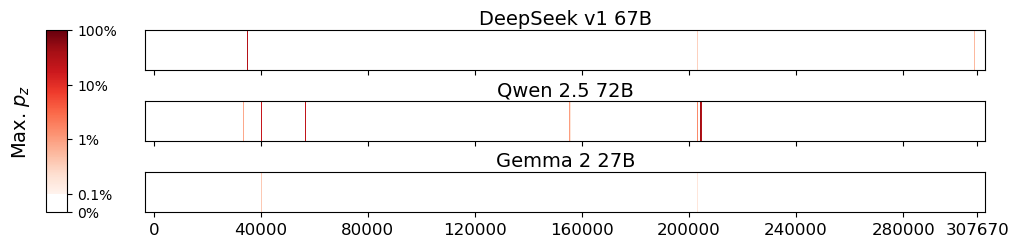}
    \includegraphics[width=\linewidth]{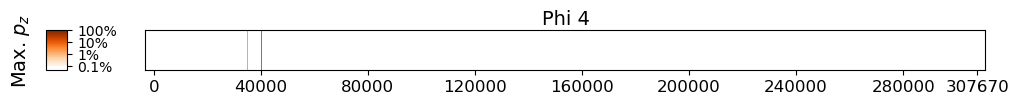}
  \end{minipage}
  \hfill
  \begin{minipage}[t]{0.45\textwidth}
    \centering
    \vspace{0cm}
    \includegraphics[width=\linewidth]{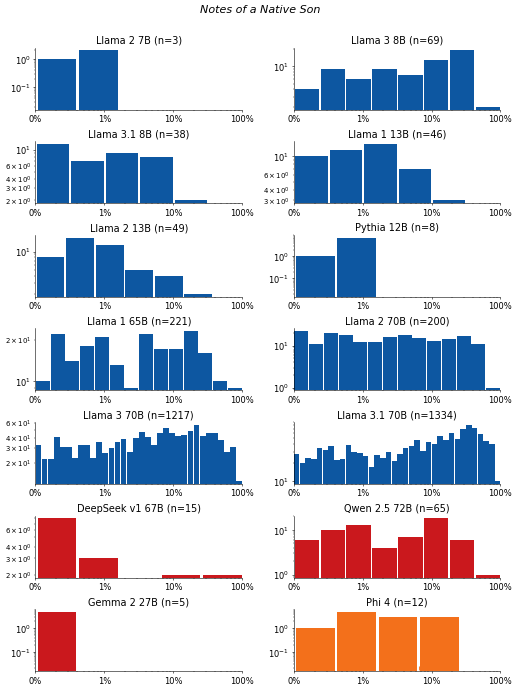}
  \end{minipage}
  \vspace{-.2cm}
  \caption{
    \textbf{\textit{Notes of a Native Son}, \citeauthor{Notes_of_a_Native_Son}.}
    For $14$ LLMs,
    (\textbf{left}) heatmaps for the sliding-window procedure and
    (\textbf{right}) corresponding distributions over suffix extraction probabilities
    ($\tau_\text{min}=0.1\%$).
  }
  \label{fig:slidingwindow:Notes_of_a_Native_Son}
\end{figure}
\FloatBarrier

\subsubsection{\textit{Another Country}, \citeauthor{Another_Country}}\label{app:sec:sliding:Another_Country}
\vspace{-.2cm}
\begin{figure}[h]
  \centering
  \begin{minipage}[t]{0.53\textwidth}
    \centering
    \vspace{0cm}
    \includegraphics[width=\linewidth]{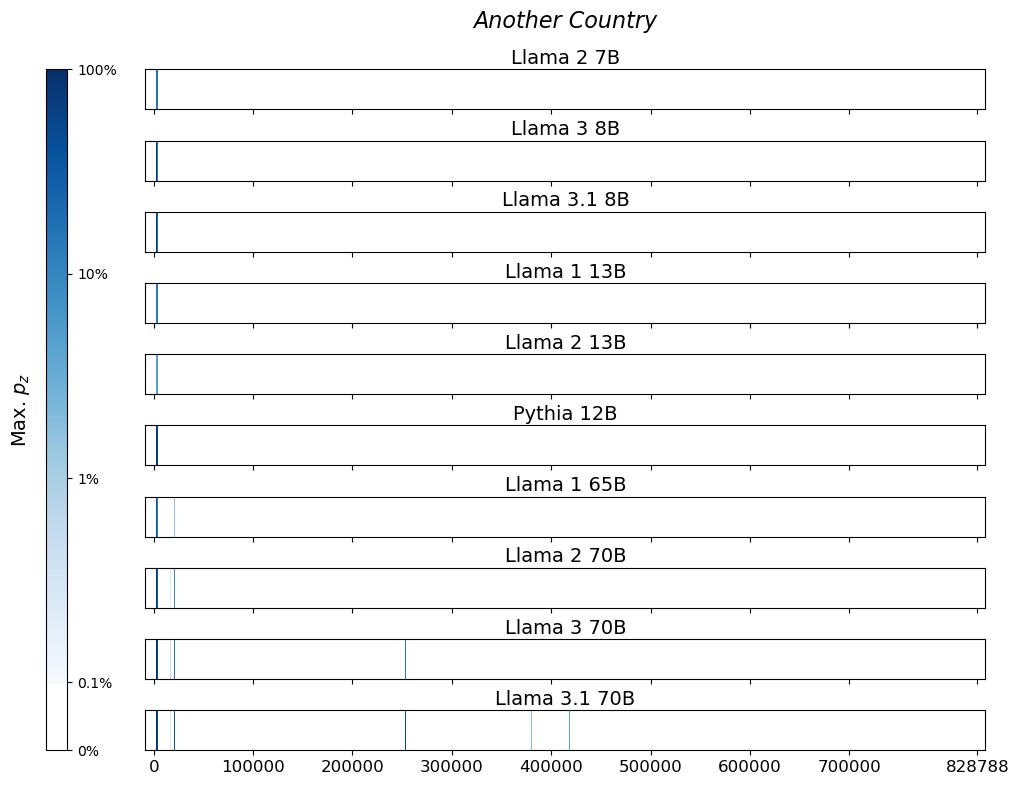}
    \includegraphics[width=\linewidth]{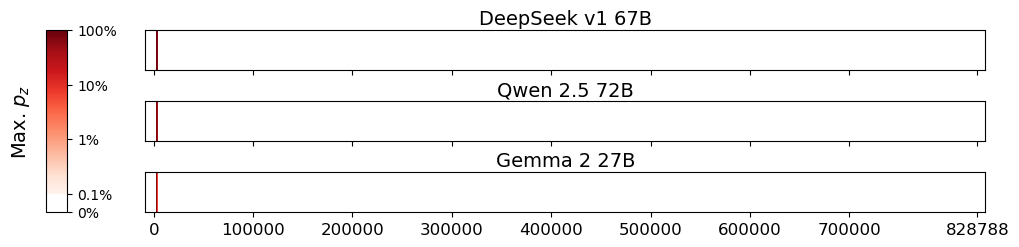}
    \includegraphics[width=\linewidth]{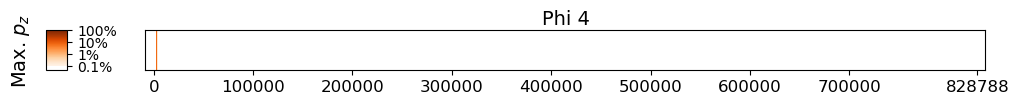}
  \end{minipage}
  \hfill
  \begin{minipage}[t]{0.45\textwidth}
    \centering
    \vspace{0cm}
    \includegraphics[width=\linewidth]{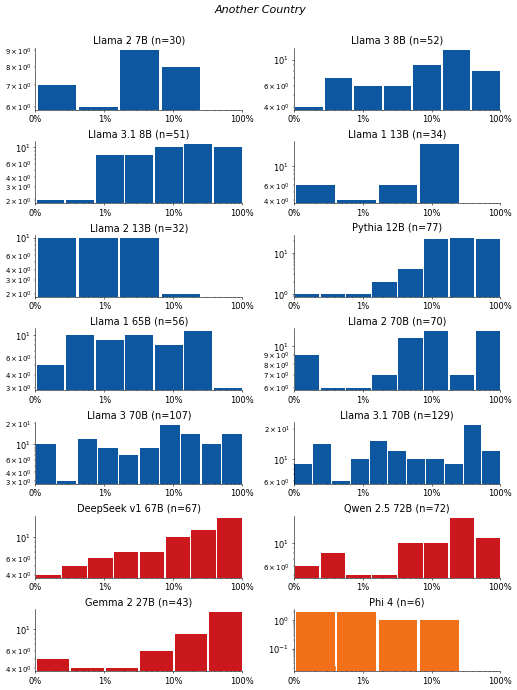}
  \end{minipage}
  \vspace{-.2cm}
  \caption{
    \textbf{\textit{Another Country}, \citeauthor{Another_Country}.}
    For $14$ LLMs,
    (\textbf{left}) heatmaps for the sliding-window procedure and
    (\textbf{right}) corresponding distributions over suffix extraction probabilities
    ($\tau_\text{min}=0.1\%$).
  }
  \label{fig:slidingwindow:Another_Country}
\end{figure}
\FloatBarrier

\clearpage
\subsubsection{\textit{The Lemon Table}, \citeauthor{The_Lemon_Table}}\label{app:sec:sliding:The_Lemon_Table}
\vspace{-.2cm}
\begin{figure}[h]
  \centering
  \begin{minipage}[t]{0.53\textwidth}
    \centering
    \vspace{0cm}
    \includegraphics[width=\linewidth]{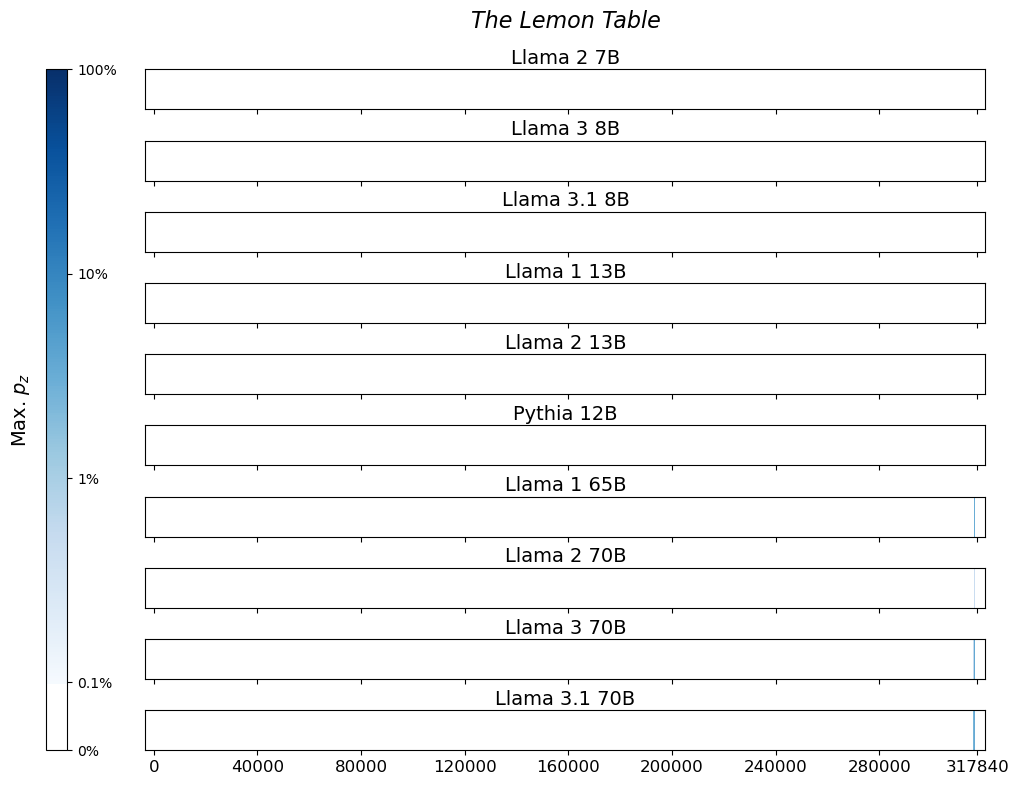}
    \includegraphics[width=\linewidth]{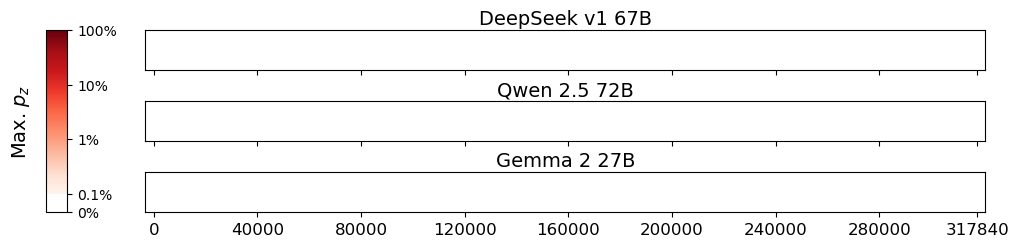}
    \includegraphics[width=\linewidth]{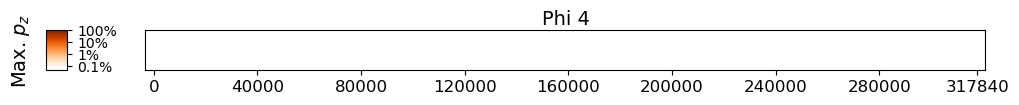}
  \end{minipage}
  \hfill
  \begin{minipage}[t]{0.45\textwidth}
    \centering
    \vspace{0cm}
    \includegraphics[width=\linewidth]{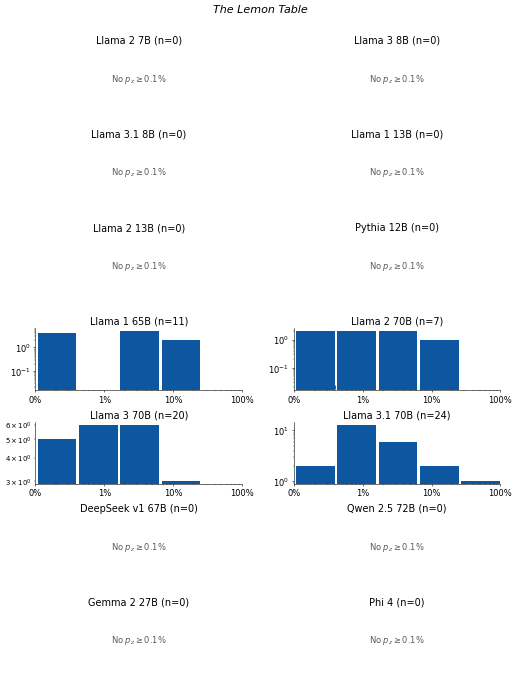}
  \end{minipage}
  \vspace{-.2cm}
  \caption{
    \textbf{\textit{The Lemon Table}, \citeauthor{The_Lemon_Table}.}
    For $14$ LLMs,
    (\textbf{left}) heatmaps for the sliding-window procedure and
    (\textbf{right}) corresponding distributions over suffix extraction probabilities
    ($\tau_\text{min}=0.1\%$).
  }
  \label{fig:slidingwindow:The_Lemon_Table}
\end{figure}
\FloatBarrier

\subsubsection{\textit{Dante and the Origins of Italian Literary Culture}, \citeauthor{Dante_and_the_Origins_of_Italian_Literary_Culture}}\label{app:sec:sliding:Dante_and_the_Origins_of_Italian_Literary_Culture}
\begin{figure}[h]
  \vspace{-.2cm}
  \centering
  \begin{minipage}[t]{0.53\textwidth}
    \centering
    \vspace{0cm}
    \includegraphics[width=\linewidth]{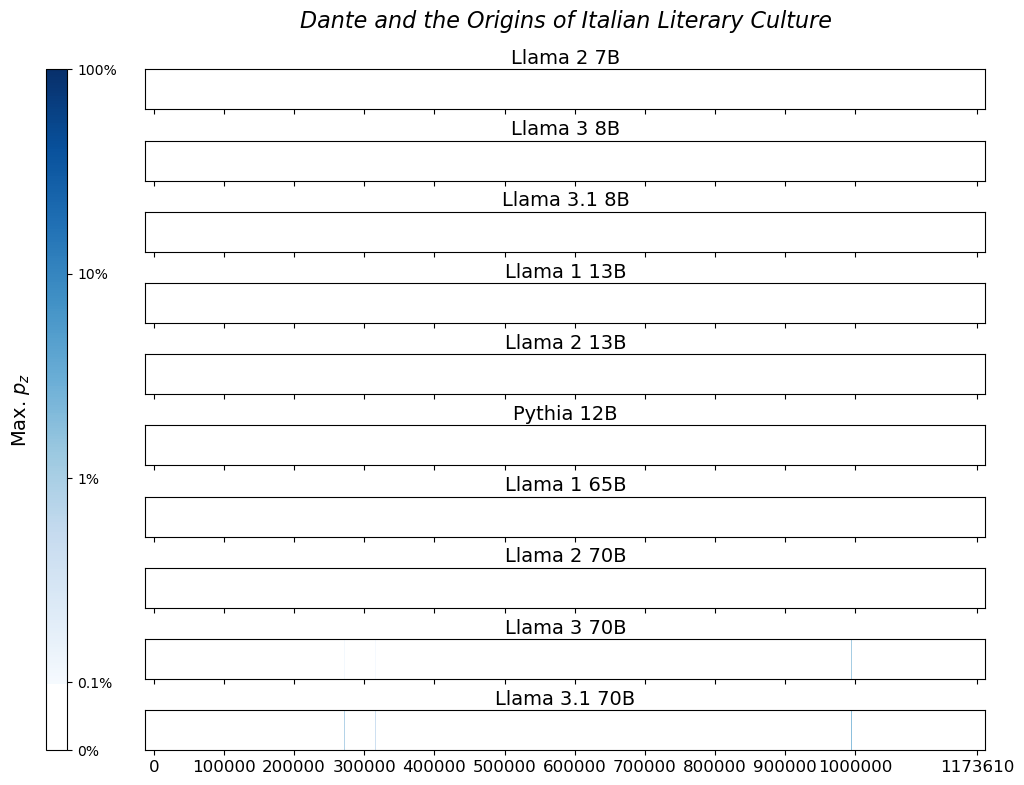}
    \includegraphics[width=\linewidth]{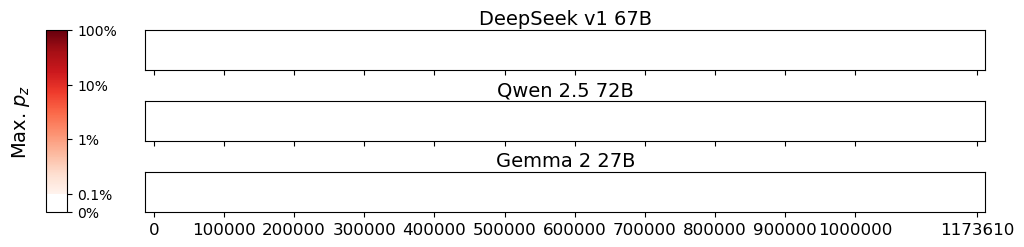}
    \includegraphics[width=\linewidth]{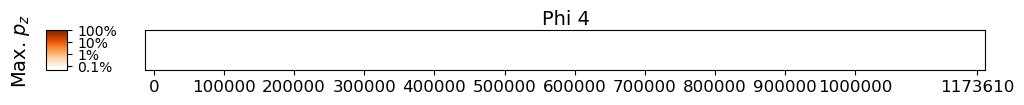}
  \end{minipage}
  \hfill
  \begin{minipage}[t]{0.45\textwidth}
    \centering
    \vspace{0cm}
    \includegraphics[width=\linewidth]{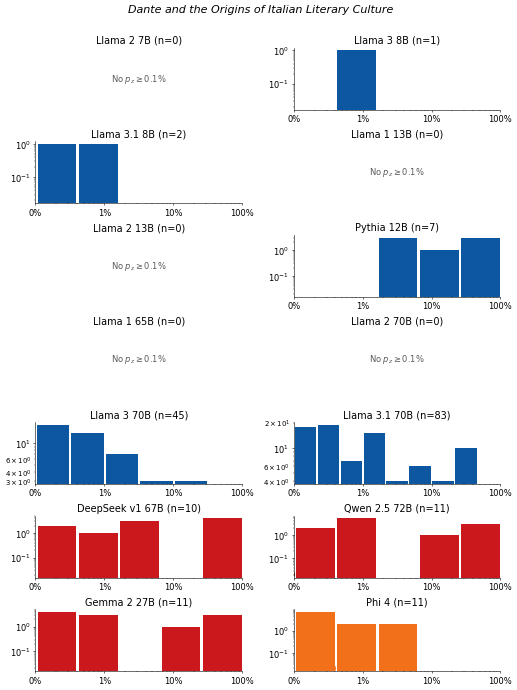}
  \end{minipage}
  \vspace{-.2cm}
  \caption{
    \textbf{\textit{Dante and the Origins of Italian Literary Culture}, \citeauthor{Dante_and_the_Origins_of_Italian_Literary_Culture}.}
    For $14$ LLMs,
    (\textbf{left}) heatmaps for the sliding-window procedure and
    (\textbf{right}) corresponding distributions over suffix extraction probabilities
    ($\tau_\text{min}=0.1\%$).
  }
  \label{fig:slidingwindow:Dante_and_the_Origins_of_Italian_Literary_Culture}
\end{figure}
\FloatBarrier

\clearpage
\subsubsection{\textit{The Parthenon}, \citeauthor{The_Parthenon}}\label{app:sec:sliding:The_Parthenon}
\vspace{-.2cm}
\begin{figure}[h]
  \centering
  \begin{minipage}[t]{0.53\textwidth}
    \centering
    \vspace{0cm}
    \includegraphics[width=\linewidth]{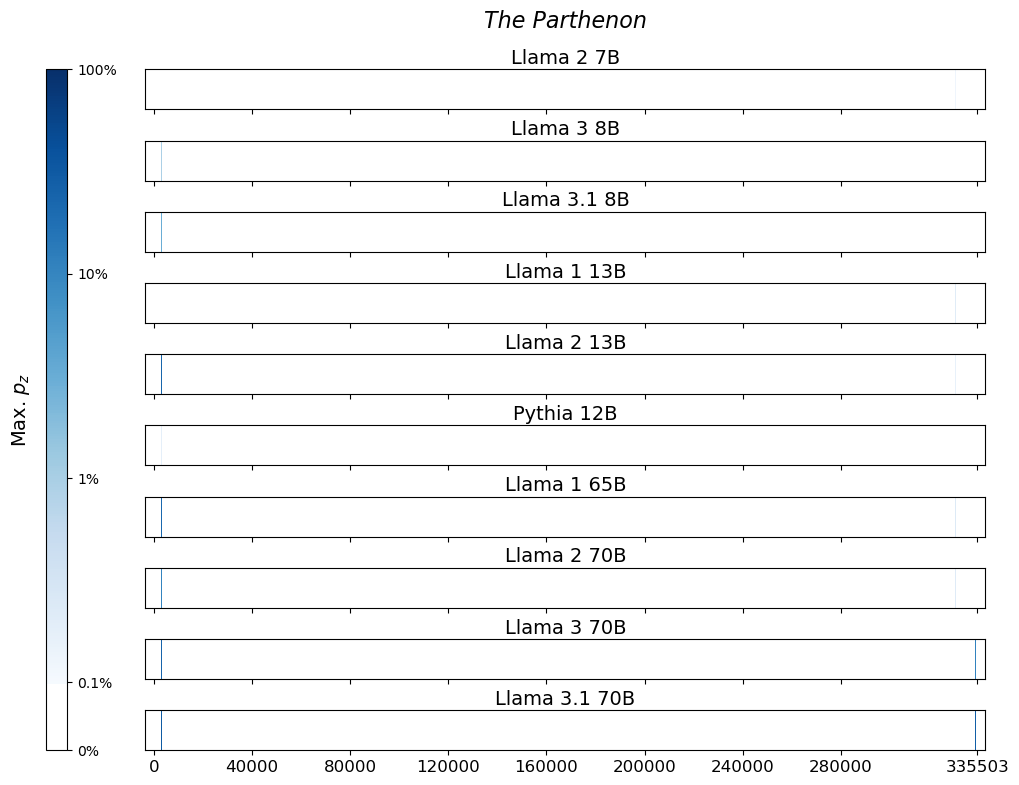}
    \includegraphics[width=\linewidth]{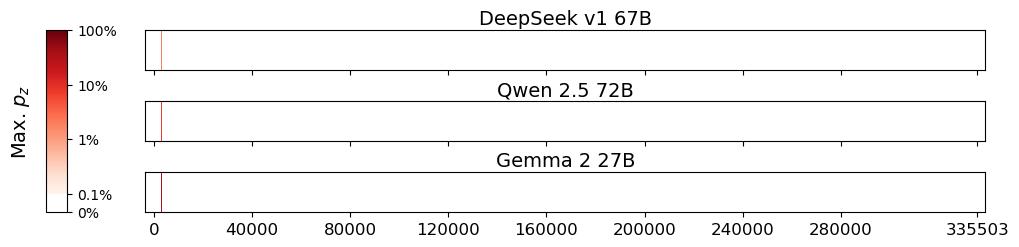}
    \includegraphics[width=\linewidth]{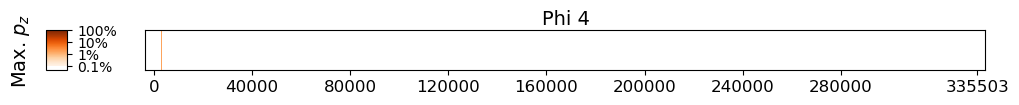}
  \end{minipage}
  \hfill
  \begin{minipage}[t]{0.45\textwidth}
    \centering
    \vspace{0cm}
    \includegraphics[width=\linewidth]{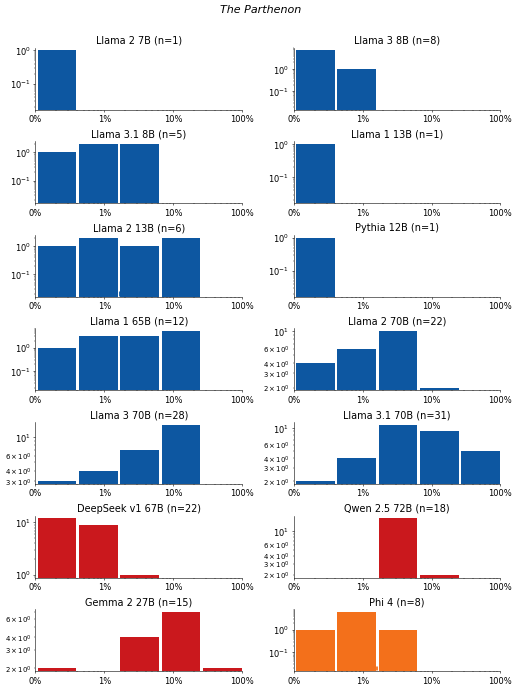}
  \end{minipage}
  \vspace{-.2cm}
  \caption{
    \textbf{\textit{The Parthenon}, \citeauthor{The_Parthenon}.}
    For $14$ LLMs,
    (\textbf{left}) heatmaps for the sliding-window procedure and
    (\textbf{right}) corresponding distributions over suffix extraction probabilities
    ($\tau_\text{min}=0.1\%$).
  }
  \label{fig:slidingwindow:The_Parthenon}
\end{figure}
\FloatBarrier

\subsubsection{\textit{Guam: Past and Present}, \citeauthor{Guam_Past_and_Present}}\label{app:sec:sliding:Guam_Past_and_Present}
\vspace{-.2cm}
\begin{figure}[h]
  \centering
  \begin{minipage}[t]{0.53\textwidth}
    \centering
    \vspace{0cm}
    \includegraphics[width=\linewidth]{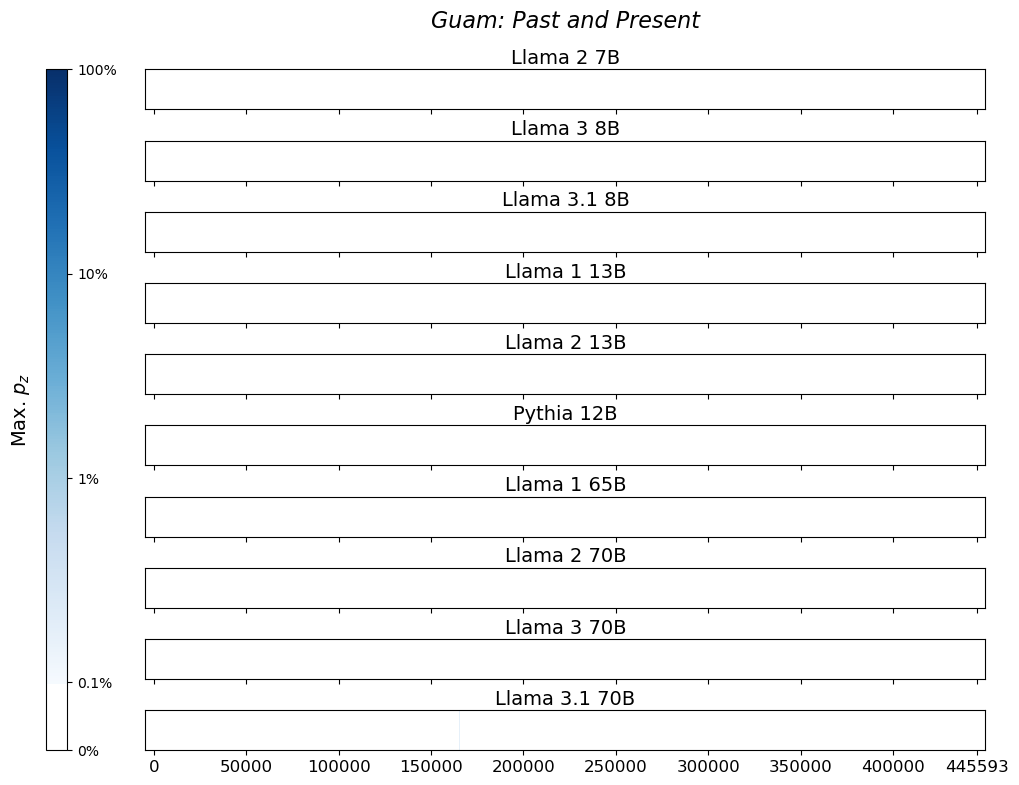}
    \includegraphics[width=\linewidth]{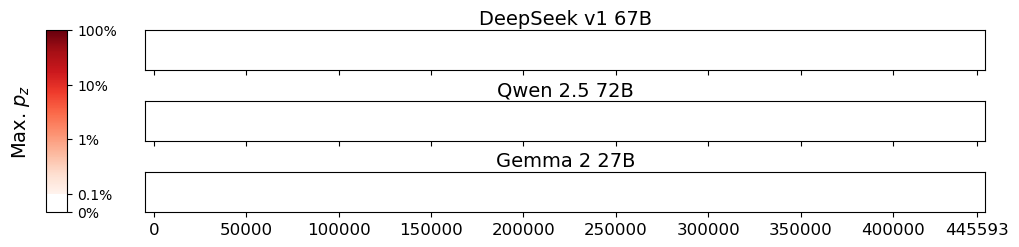}
    \includegraphics[width=\linewidth]{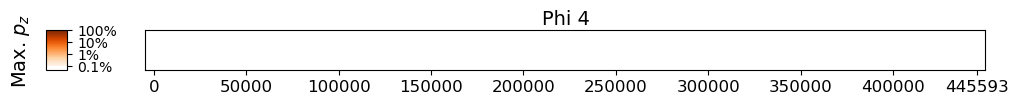}
  \end{minipage}
  \hfill
  \begin{minipage}[t]{0.45\textwidth}
    \centering
    \vspace{0cm}
    \includegraphics[width=\linewidth]{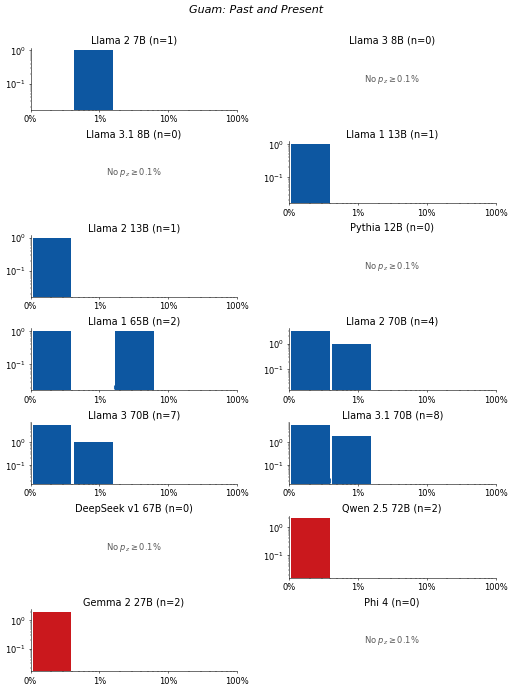}
  \end{minipage}
  \vspace{-.2cm}
  \caption{
    \textbf{\textit{Guam: Past and Present}, \citeauthor{Guam_Past_and_Present}.}
    For $14$ LLMs,
    (\textbf{left}) heatmaps for the sliding-window procedure and
    (\textbf{right}) corresponding distributions over suffix extraction probabilities
    ($\tau_\text{min}=0.1\%$).
  }
  \label{fig:slidingwindow:Guam_Past_and_Present}
\end{figure}
\FloatBarrier

\clearpage
\subsubsection{\textit{Waiting for Godot}, \citeauthor{Waiting_for_Godot}}\label{app:sec:sliding:Waiting_for_Godot}
\vspace{-.2cm}
\begin{figure}[h]
  \centering
  \begin{minipage}[t]{0.53\textwidth}
    \centering
    \vspace{0cm}
    \includegraphics[width=\linewidth]{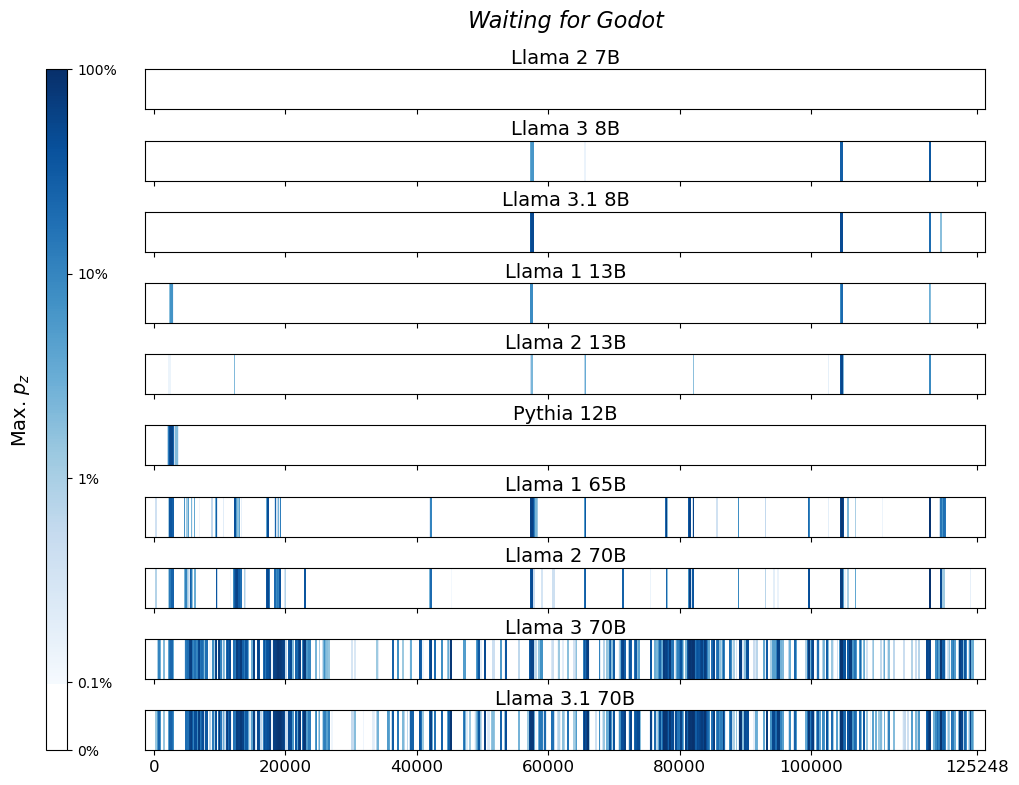}
    \includegraphics[width=\linewidth]{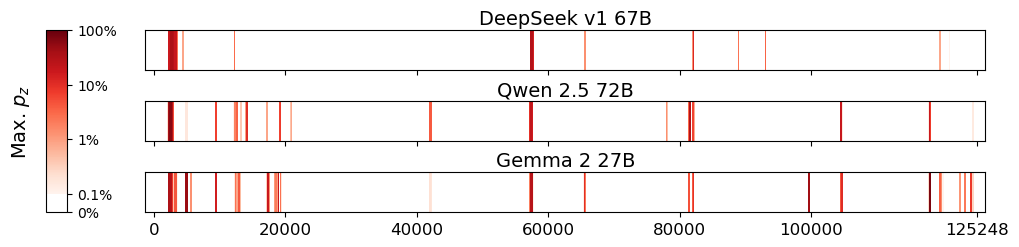}
    \includegraphics[width=\linewidth]{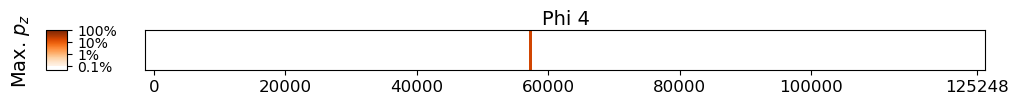}
  \end{minipage}
  \hfill
  \begin{minipage}[t]{0.45\textwidth}
    \centering
    \vspace{0cm}
    \includegraphics[width=\linewidth]{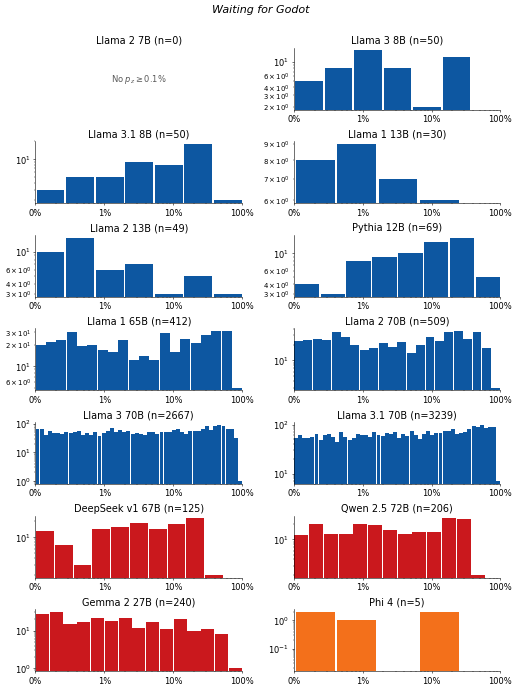}
  \end{minipage}
  \vspace{-.2cm}
  \caption{
    \textbf{\textit{Waiting for Godot}, \citeauthor{Waiting_for_Godot}.}
    For $14$ LLMs,
    (\textbf{left}) heatmaps for the sliding-window procedure and
    (\textbf{right}) corresponding distributions over suffix extraction probabilities
    ($\tau_\text{min}=0.1\%$).
  }
  \label{fig:slidingwindow:Waiting_for_Godot}
\end{figure}
\FloatBarrier

\subsubsection{\textit{The Lonely Soldier}, \citeauthor{The_Lonely_Soldier}}\label{app:sec:sliding:The_Lonely_Soldier}
\vspace{-.2cm}
\begin{figure}[h]
  \centering
  \begin{minipage}[t]{0.53\textwidth}
    \centering
    \vspace{0cm}
    \includegraphics[width=\linewidth]{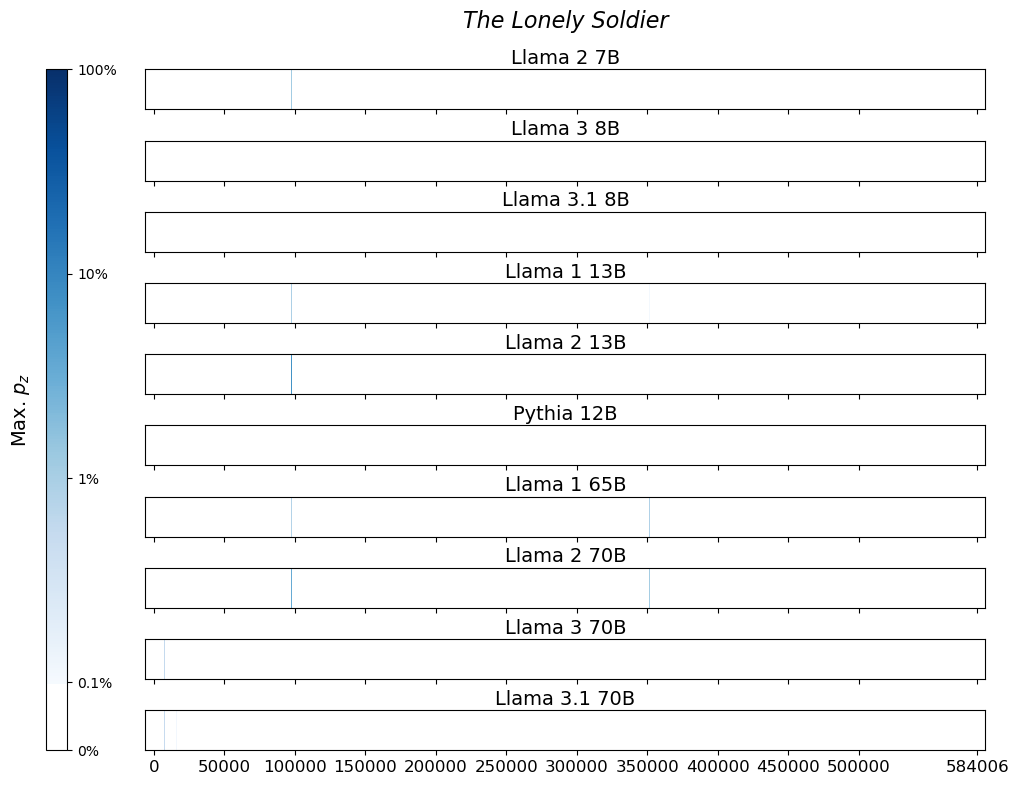}
    \includegraphics[width=\linewidth]{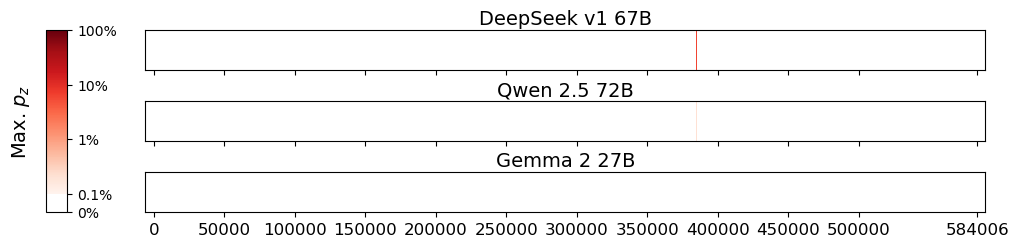}
    \includegraphics[width=\linewidth]{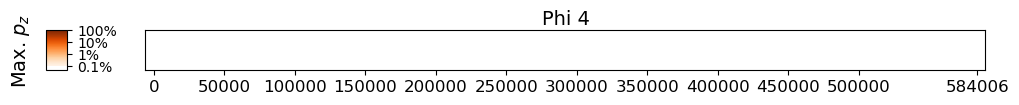}
  \end{minipage}
  \hfill
  \begin{minipage}[t]{0.45\textwidth}
    \centering
    \vspace{0cm}
    \includegraphics[width=\linewidth]{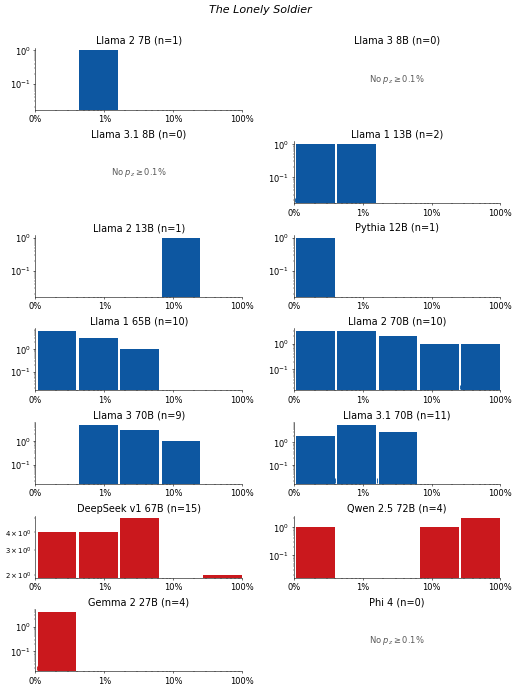}
  \end{minipage}
  \vspace{-.2cm}
  \caption{
    \textbf{\textit{The Lonely Soldier}, \citeauthor{The_Lonely_Soldier}.}
    For $14$ LLMs,
    (\textbf{left}) heatmaps for the sliding-window procedure and
    (\textbf{right}) corresponding distributions over suffix extraction probabilities
    ($\tau_\text{min}=0.1\%$).
  }
  \label{fig:slidingwindow:The_Lonely_Soldier}
\end{figure}
\FloatBarrier

\clearpage
\subsubsection{\textit{Simple Cakes}, \citeauthor{Simple_Cakes}}\label{app:sec:sliding:Simple_Cakes}
\vspace{-.2cm}
\begin{figure}[h]
  \centering
  \begin{minipage}[t]{0.53\textwidth}
    \centering
    \vspace{0cm}
    \includegraphics[width=\linewidth]{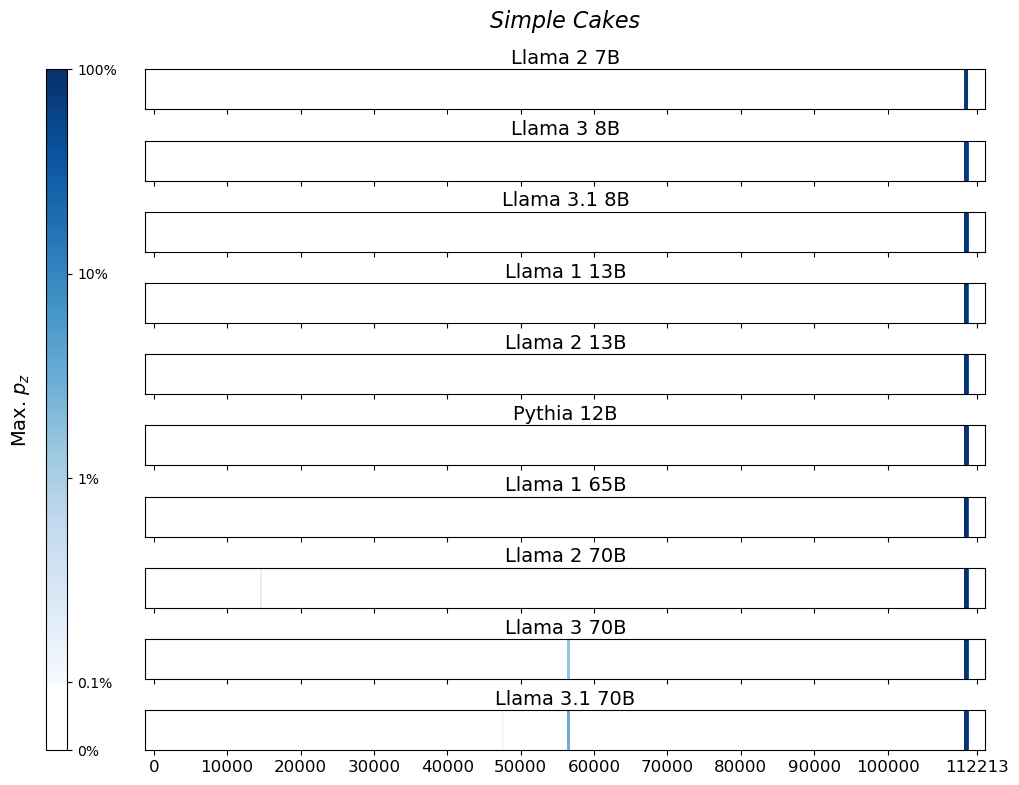}
    \includegraphics[width=\linewidth]{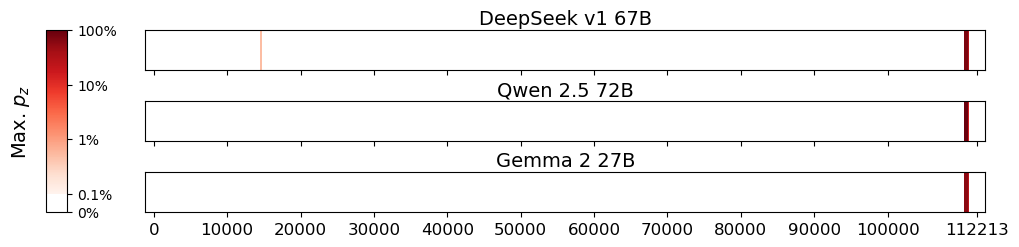}
    \includegraphics[width=\linewidth]{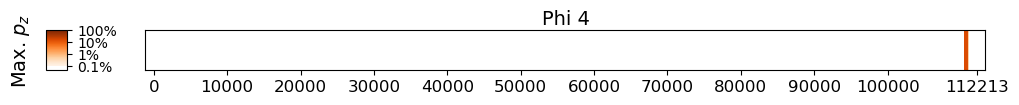}
  \end{minipage}
  \hfill
  \begin{minipage}[t]{0.45\textwidth}
    \centering
    \vspace{0cm}
    \includegraphics[width=\linewidth]{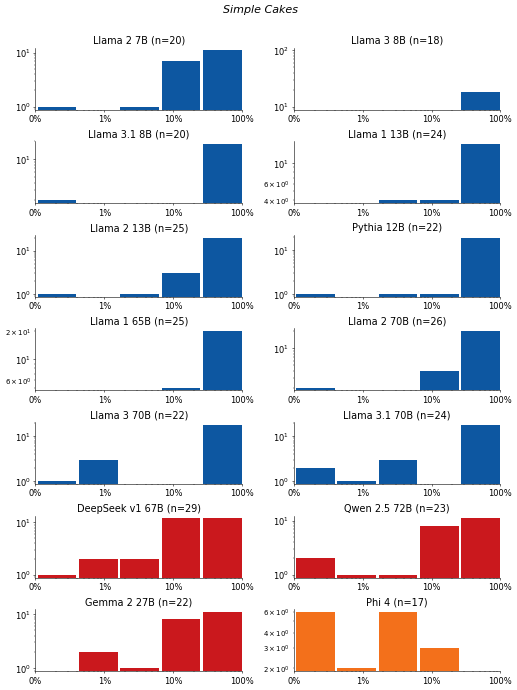}
  \end{minipage}
  \vspace{-.2cm}
  \caption{
    \textbf{\textit{Simple Cakes}, \citeauthor{Simple_Cakes}.}
    For $14$ LLMs,
    (\textbf{left}) heatmaps for the sliding-window procedure and
    (\textbf{right}) corresponding distributions over suffix extraction probabilities
    ($\tau_\text{min}=0.1\%$).
  }
  \label{fig:slidingwindow:Simple_Cakes}
\end{figure}
\FloatBarrier

\subsubsection{\textit{Paradise Valley}, \citeauthor{Paradise_Valley}}\label{app:sec:sliding:Paradise_Valley}
\vspace{-.2cm}
\begin{figure}[h]
  \centering
  \begin{minipage}[t]{0.53\textwidth}
    \centering
    \vspace{0cm}
    \includegraphics[width=\linewidth]{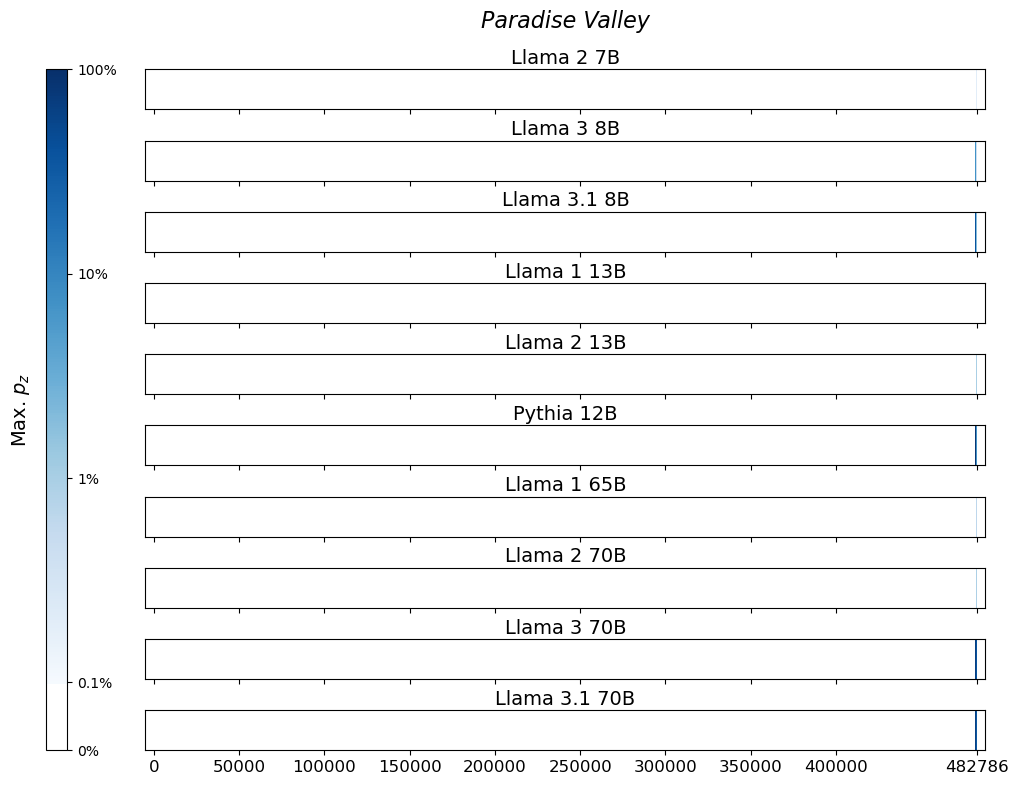}
    \includegraphics[width=\linewidth]{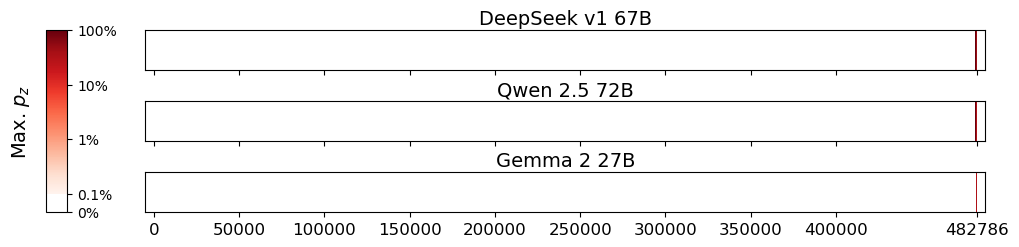}
    \includegraphics[width=\linewidth]{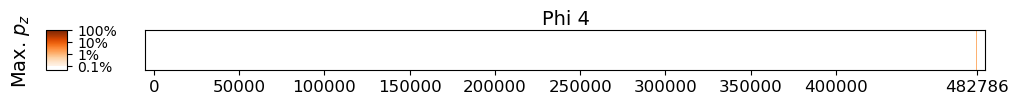}
  \end{minipage}
  \hfill
  \begin{minipage}[t]{0.45\textwidth}
    \centering
    \vspace{0cm}
    \includegraphics[width=\linewidth]{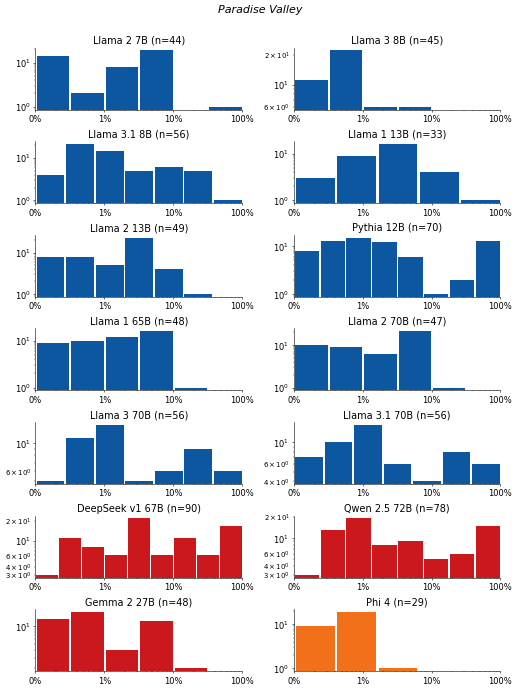}
  \end{minipage}
  \vspace{-.2cm}
  \caption{
    \textbf{\textit{Paradise Valley}, \citeauthor{Paradise_Valley}.}
    For $14$ LLMs,
    (\textbf{left}) heatmaps for the sliding-window procedure and
    (\textbf{right}) corresponding distributions over suffix extraction probabilities
    ($\tau_\text{min}=0.1\%$).
  }
  \label{fig:slidingwindow:Paradise_Valley}
\end{figure}
\FloatBarrier

\clearpage
\subsubsection{\textit{The Cat's Pajamas}, \citeauthor{The_Cat_s_Pajamas}}\label{app:sec:sliding:The_Cat_s_Pajamas}
\vspace{-.2cm}
\begin{figure}[h]
  \centering
  \begin{minipage}[t]{0.53\textwidth}
    \centering
    \vspace{0cm}
    \includegraphics[width=\linewidth]{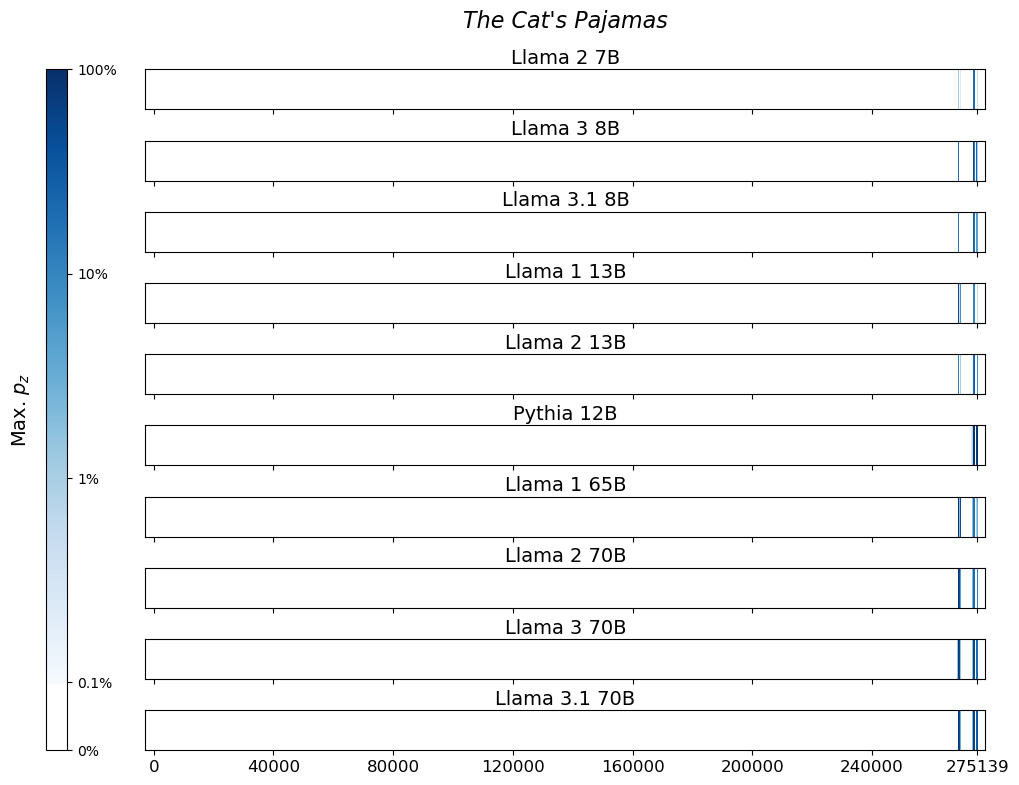}
    \includegraphics[width=\linewidth]{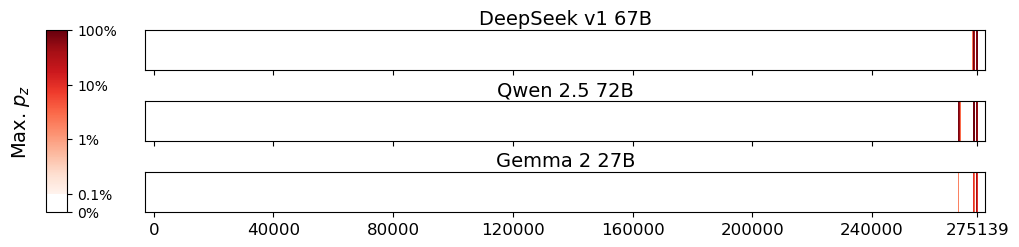}
    \includegraphics[width=\linewidth]{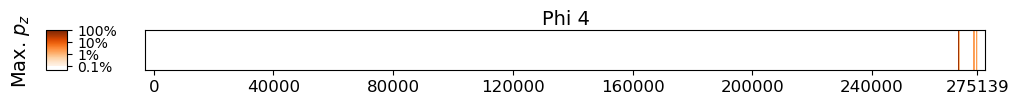}
  \end{minipage}
  \hfill
  \begin{minipage}[t]{0.45\textwidth}
    \centering
    \vspace{0cm}
    \includegraphics[width=\linewidth]{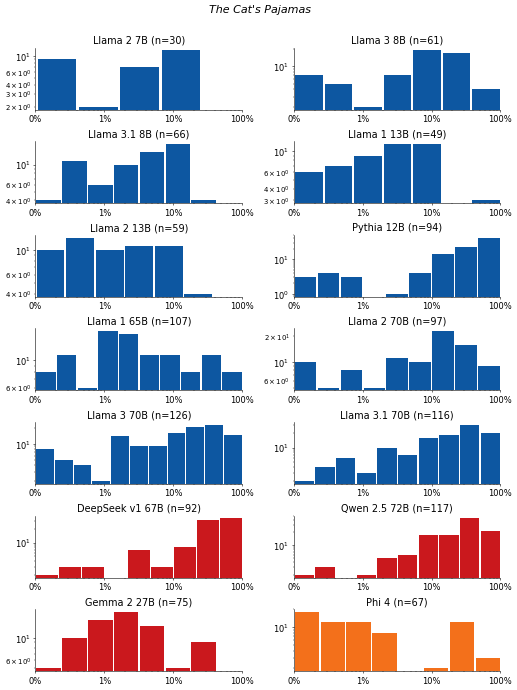}
  \end{minipage}
  \vspace{-.2cm}
  \caption{
    \textbf{\textit{The Cat's Pajamas}, \citeauthor{The_Cat_s_Pajamas}.}
    For $14$ LLMs,
    (\textbf{left}) heatmaps for the sliding-window procedure and
    (\textbf{right}) corresponding distributions over suffix extraction probabilities
    ($\tau_\text{min}=0.1\%$).
  }
  \label{fig:slidingwindow:The_Cat_s_Pajamas}
\end{figure}
\FloatBarrier

\subsubsection{\textit{London in Chains}, \citeauthor{London_in_Chains}}\label{app:sec:sliding:London_in_Chains}
\vspace{-.2cm}
\begin{figure}[h]
  \centering
  \begin{minipage}[t]{0.53\textwidth}
    \centering
    \vspace{0cm}
    \includegraphics[width=\linewidth]{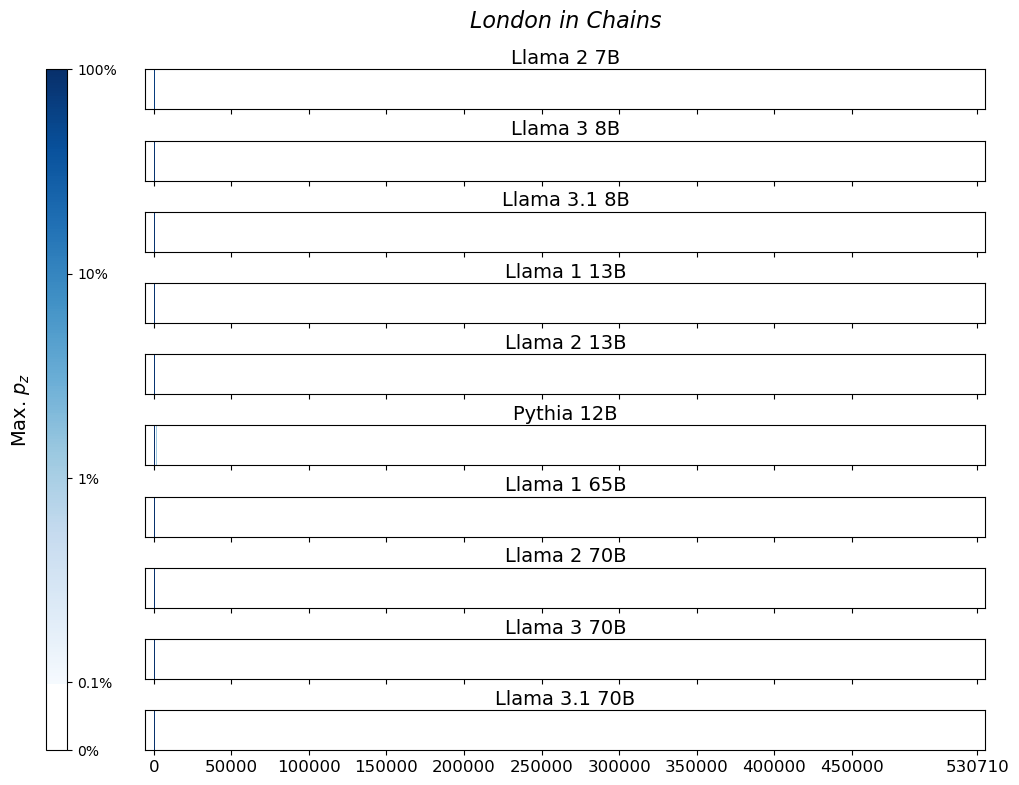}
    \includegraphics[width=\linewidth]{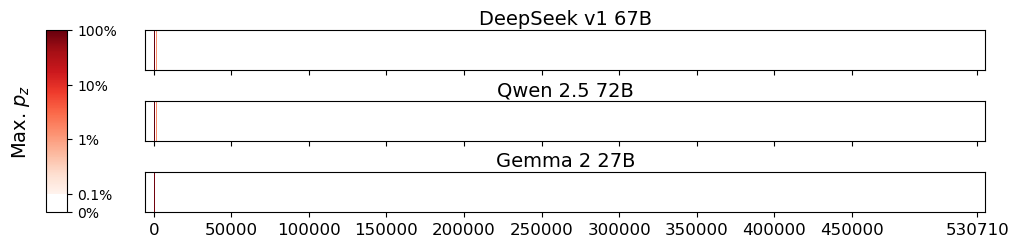}
    \includegraphics[width=\linewidth]{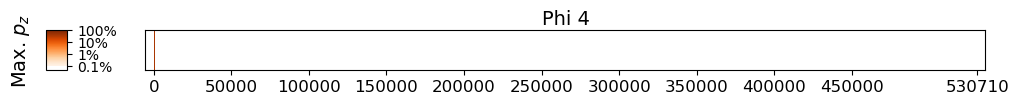}
  \end{minipage}
  \hfill
  \begin{minipage}[t]{0.45\textwidth}
    \centering
    \vspace{0cm}
    \includegraphics[width=\linewidth]{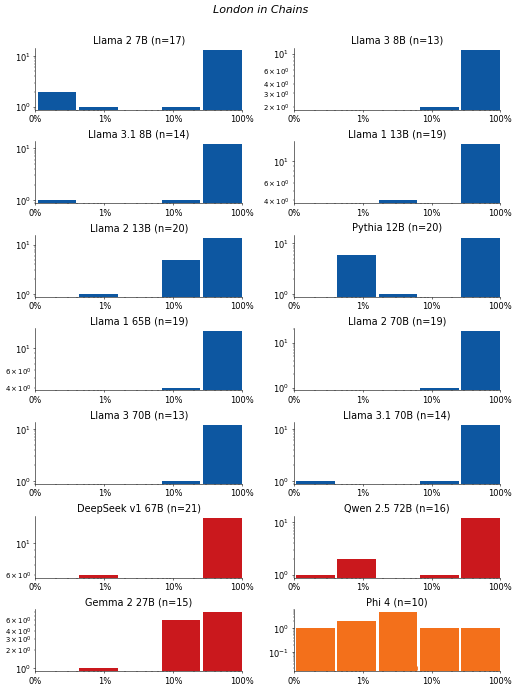}
  \end{minipage}
  \vspace{-.2cm}
  \caption{
    \textbf{\textit{London in Chains}, \citeauthor{London_in_Chains}.}
    For $14$ LLMs,
    (\textbf{left}) heatmaps for the sliding-window procedure and
    (\textbf{right}) corresponding distributions over suffix extraction probabilities
    ($\tau_\text{min}=0.1\%$).
  }
  \label{fig:slidingwindow:London_in_Chains}
\end{figure}
\FloatBarrier

\clearpage
\subsubsection{\textit{My Einstein}, \citeauthor{My_Einstein}}\label{app:sec:sliding:My_Einstein}
\vspace{-.2cm}
\begin{figure}[h]
  \centering
  \begin{minipage}[t]{0.53\textwidth}
    \centering
    \vspace{0cm}
    \includegraphics[width=\linewidth]{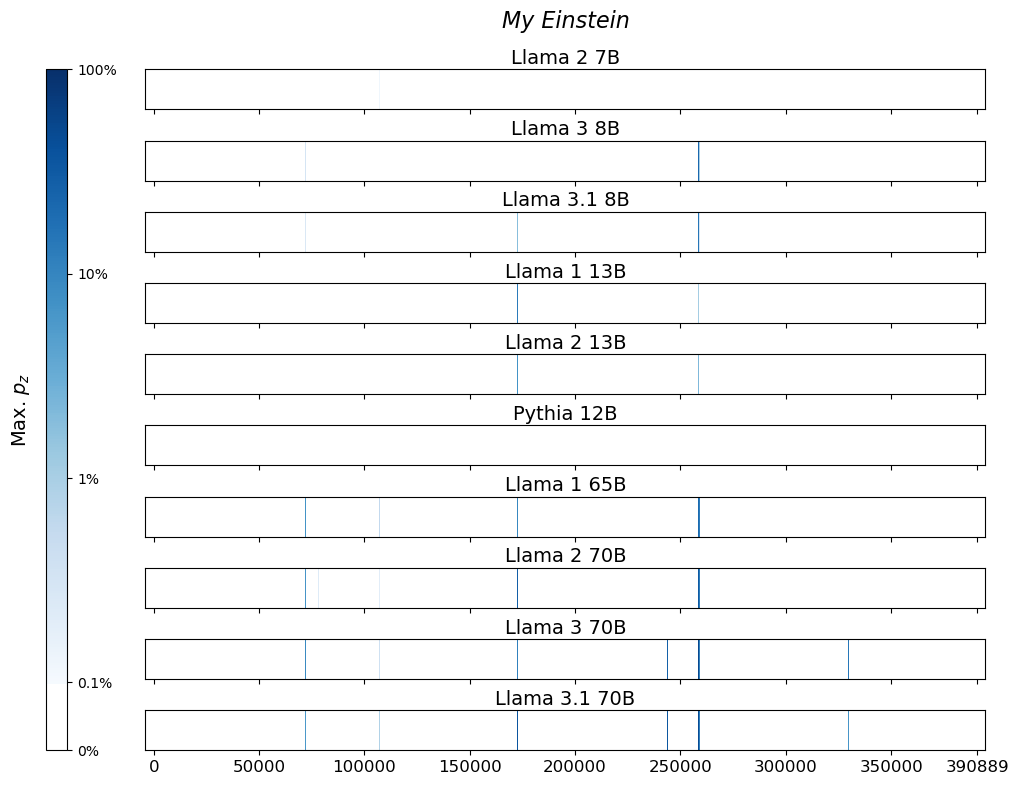}
    \includegraphics[width=\linewidth]{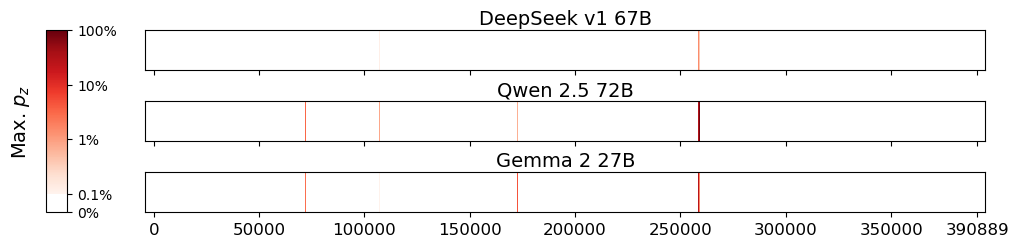}
    \includegraphics[width=\linewidth]{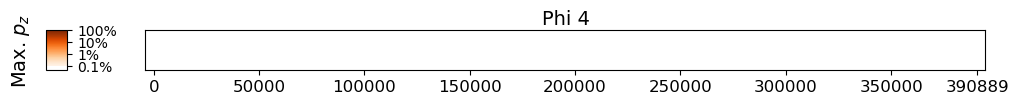}
  \end{minipage}
  \hfill
  \begin{minipage}[t]{0.45\textwidth}
    \centering
    \vspace{0cm}
    \includegraphics[width=\linewidth]{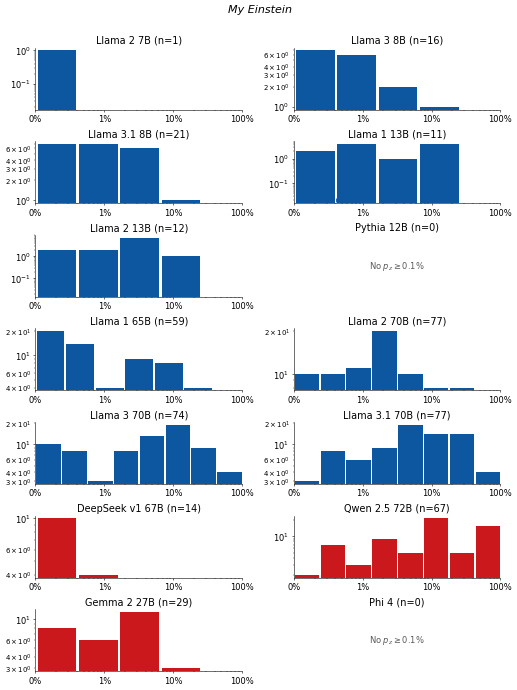}
  \end{minipage}
  \vspace{-.2cm}
  \caption{
    \textbf{\textit{My Einstein}, \citeauthor{My_Einstein}.}
    For $14$ LLMs,
    (\textbf{left}) heatmaps for the sliding-window procedure and
    (\textbf{right}) corresponding distributions over suffix extraction probabilities
    ($\tau_\text{min}=0.1\%$).
  }
  \label{fig:slidingwindow:My_Einstein}
\end{figure}
\FloatBarrier

\subsubsection{\textit{The Da Vinci Code}, \citeauthor{The_Da_Vinci_Code}}\label{app:sec:sliding:The_Da_Vinci_Code}
\vspace{-.2cm}
\begin{figure}[h]
  \centering
  \begin{minipage}[t]{0.53\textwidth}
    \centering
    \vspace{0cm}
    \includegraphics[width=\linewidth]{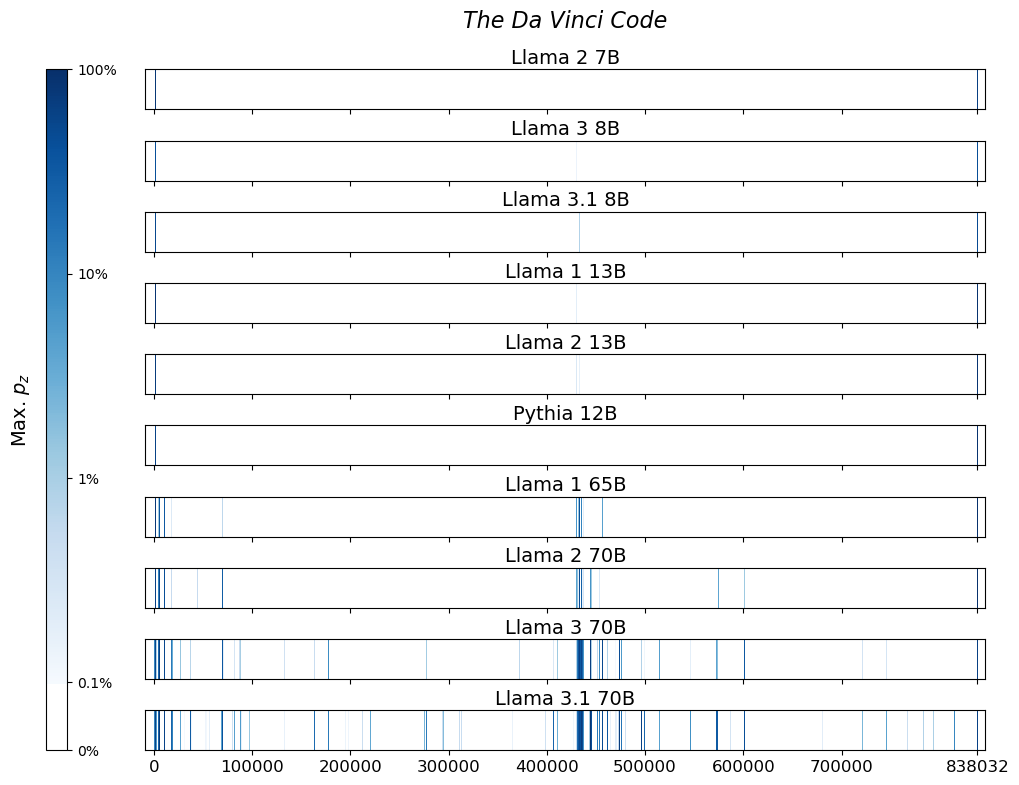}
    \includegraphics[width=\linewidth]{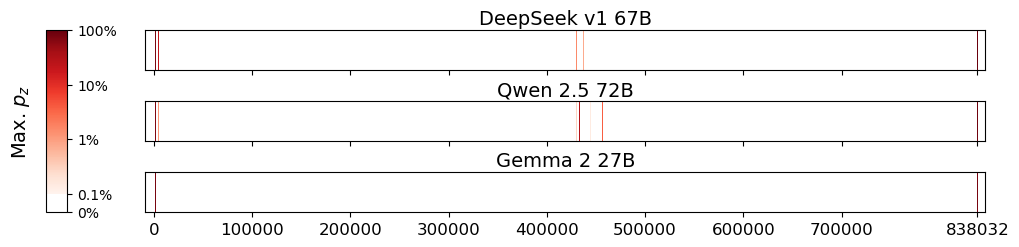}
    \includegraphics[width=\linewidth]{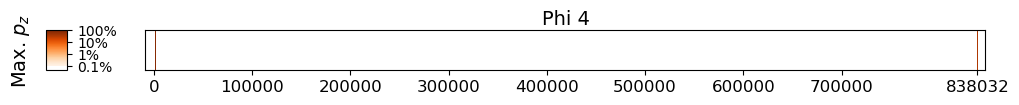}
  \end{minipage}
  \hfill
  \begin{minipage}[t]{0.45\textwidth}
    \centering
    \vspace{0cm}
    \includegraphics[width=\linewidth]{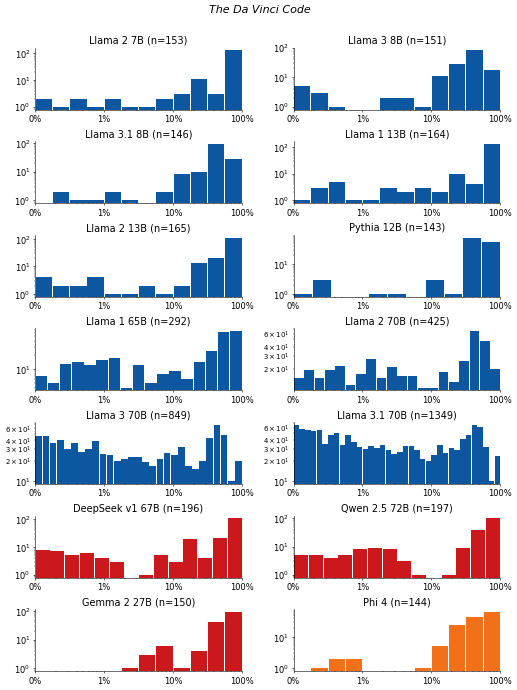}
  \end{minipage}
  \vspace{-.2cm}
  \caption{
    \textbf{\textit{The Da Vinci Code}, \citeauthor{The_Da_Vinci_Code}.}
    For $14$ LLMs,
    (\textbf{left}) heatmaps for the sliding-window procedure and
    (\textbf{right}) corresponding distributions over suffix extraction probabilities
    ($\tau_\text{min}=0.1\%$).
  }
  \label{fig:slidingwindow:The_Da_Vinci_Code}
\end{figure}
\FloatBarrier

\clearpage
\subsubsection{\textit{Live and Learn}, \citeauthor{Live_and_Learn}}\label{app:sec:sliding:Live_and_Learn}
\vspace{-.2cm}
\begin{figure}[h]
  \centering
  \begin{minipage}[t]{0.53\textwidth}
    \centering
    \vspace{0cm}
    \includegraphics[width=\linewidth]{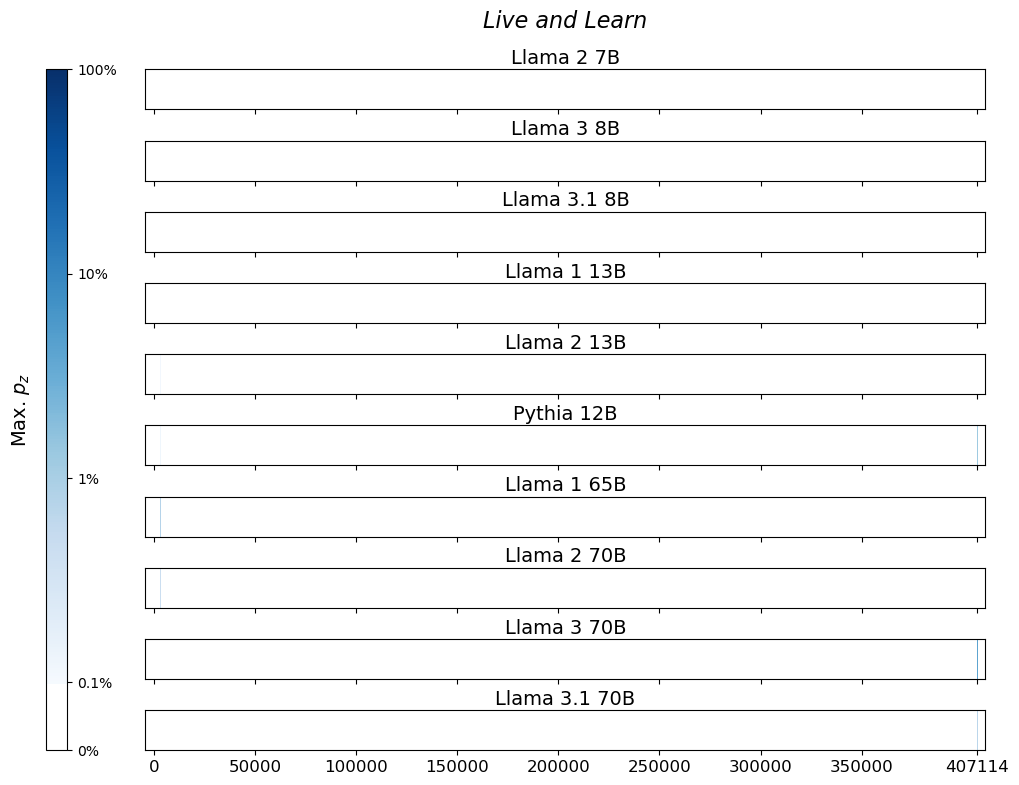}
    \includegraphics[width=\linewidth]{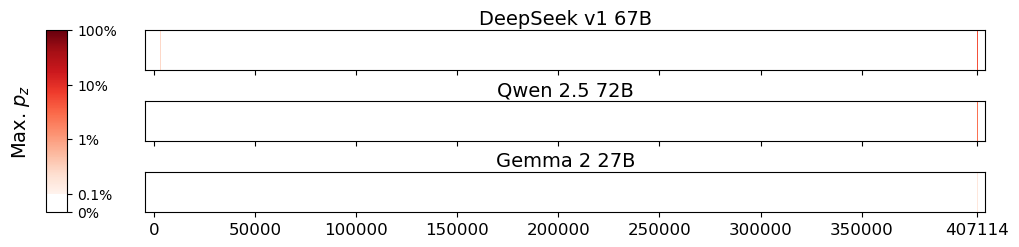}
    \includegraphics[width=\linewidth]{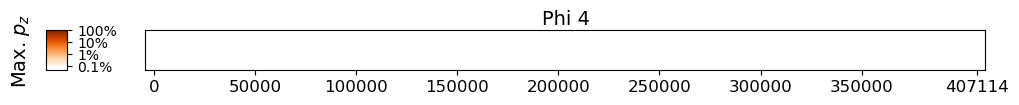}
  \end{minipage}
  \hfill
  \begin{minipage}[t]{0.45\textwidth}
    \centering
    \vspace{0cm}
    \includegraphics[width=\linewidth]{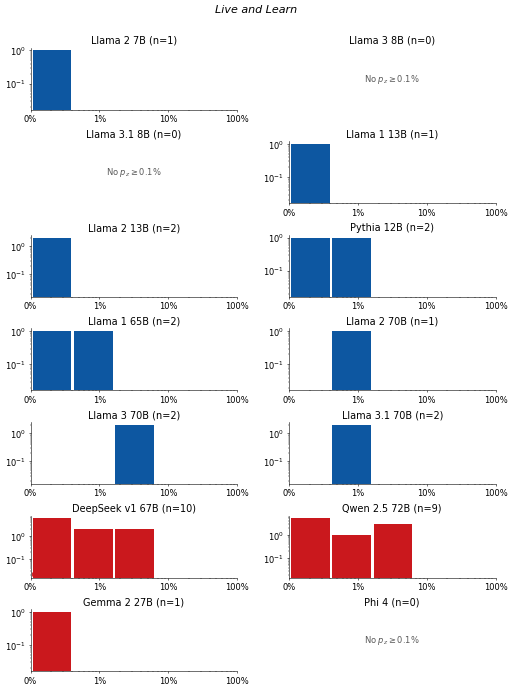}
  \end{minipage}
  \vspace{-.2cm}
  \caption{
    \textbf{\textit{Live and Learn}, \citeauthor{Live_and_Learn}.}
    For $14$ LLMs,
    (\textbf{left}) heatmaps for the sliding-window procedure and
    (\textbf{right}) corresponding distributions over suffix extraction probabilities
    ($\tau_\text{min}=0.1\%$).
  }
  \label{fig:slidingwindow:Live_and_Learn}
\end{figure}
\FloatBarrier

\subsubsection{\textit{Knowing Your Value}, \citeauthor{Knowing_Your_Value}}\label{app:sec:sliding:Knowing_Your_Value}
\vspace{-.2cm}
\begin{figure}[h]
  \centering
  \begin{minipage}[t]{0.53\textwidth}
    \centering
    \vspace{0cm}
    \includegraphics[width=\linewidth]{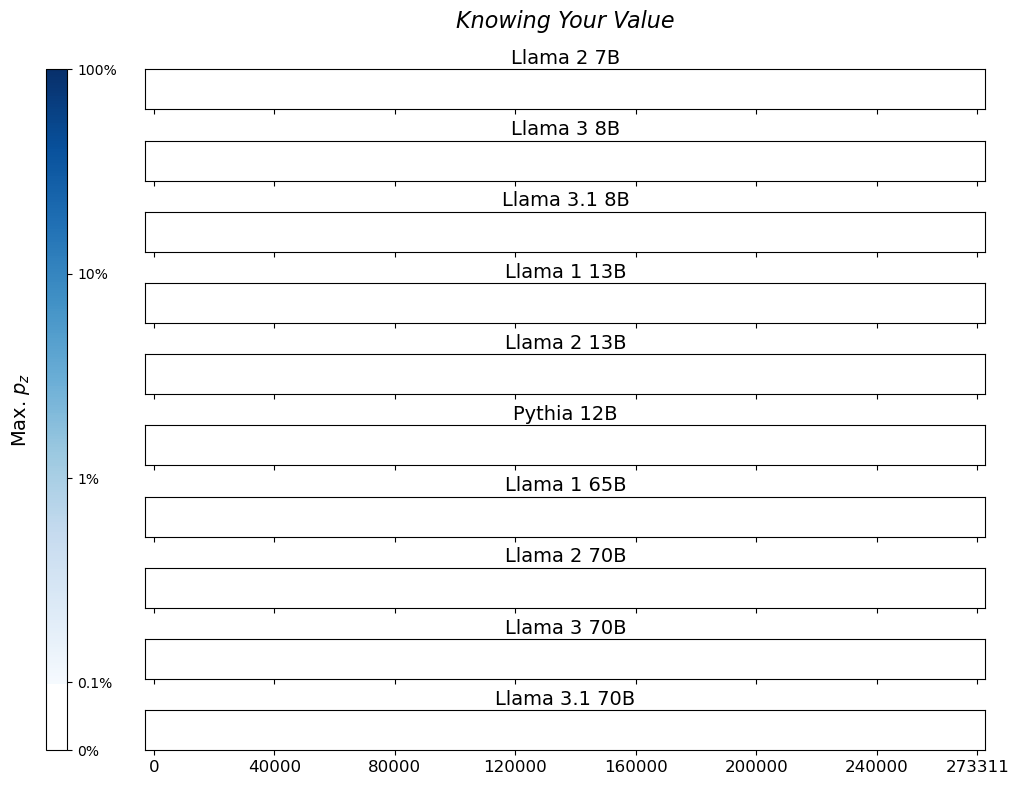}
    \includegraphics[width=\linewidth]{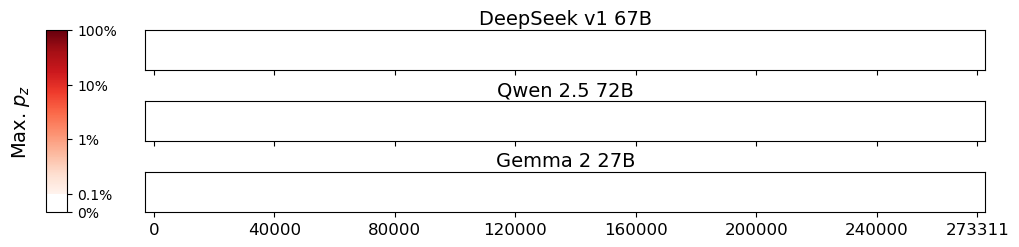}
    \includegraphics[width=\linewidth]{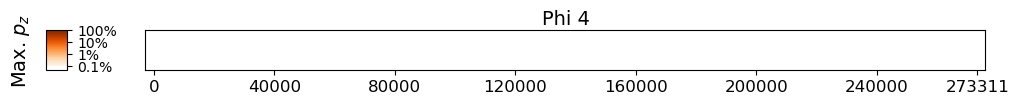}
  \end{minipage}
  \hfill
  \begin{minipage}[t]{0.45\textwidth}
    \centering
    \vspace{0cm}
    \includegraphics[width=\linewidth]{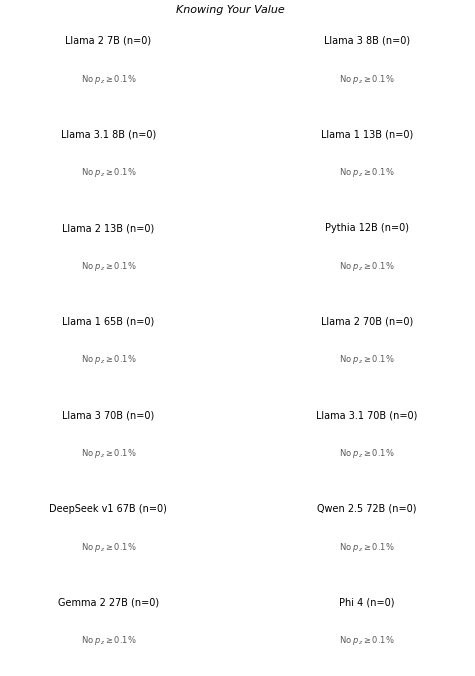}
  \end{minipage}
  \vspace{-.2cm}
  \caption{
    \textbf{\textit{Knowing Your Value}, \citeauthor{Knowing_Your_Value}.}
    For $14$ LLMs,
    (\textbf{left}) heatmaps for the sliding-window procedure and
    (\textbf{right}) corresponding distributions over suffix extraction probabilities
    ($\tau_\text{min}=0.1\%$).
  }
  \label{fig:slidingwindow:Knowing_Your_Value}
\end{figure}
\FloatBarrier

\clearpage
\subsubsection{\textit{The Myth of Sisyphus}, \citeauthor{The_Myth_of_Sisyphus}}\label{app:sec:sliding:The_Myth_of_Sisyphus}
\vspace{-.2cm}
\begin{figure}[h]
  \centering
  \begin{minipage}[t]{0.53\textwidth}
    \centering
    \vspace{0cm}
    \includegraphics[width=\linewidth]{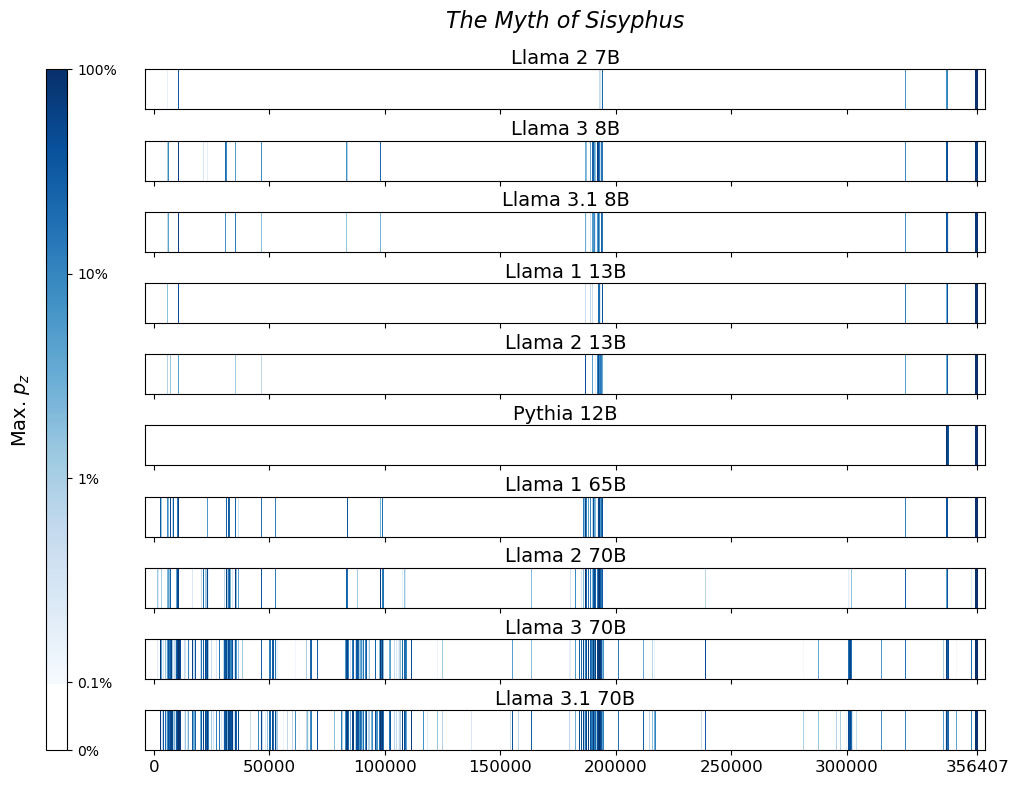}
    \includegraphics[width=\linewidth]{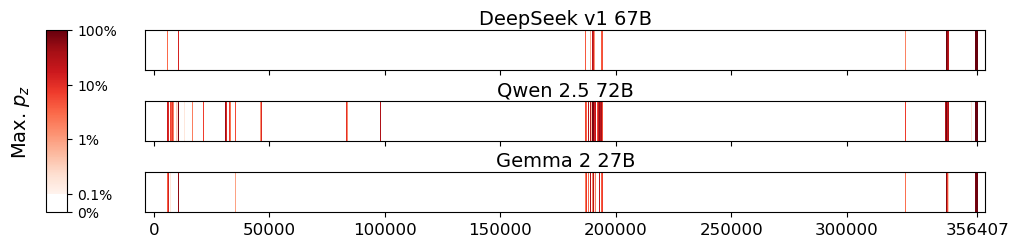}
    \includegraphics[width=\linewidth]{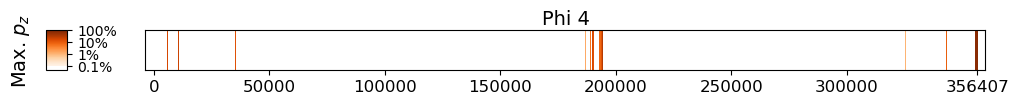}
  \end{minipage}
  \hfill
  \begin{minipage}[t]{0.45\textwidth}
    \centering
    \vspace{0cm}
    \includegraphics[width=\linewidth]{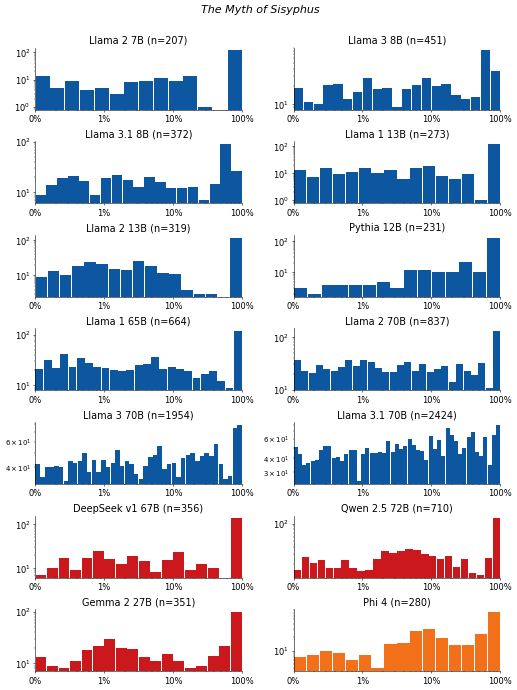}
  \end{minipage}
  \vspace{-.2cm}
  \caption{
    \textbf{\textit{The Myth of Sisyphus}, \citeauthor{The_Myth_of_Sisyphus}.}
    For $14$ LLMs,
    (\textbf{left}) heatmaps for the sliding-window procedure and
    (\textbf{right}) corresponding distributions over suffix extraction probabilities
    ($\tau_\text{min}=0.1\%$).
  }
  \label{fig:slidingwindow:The_Myth_of_Sisyphus}
\end{figure}
\FloatBarrier

\subsubsection{\textit{Alice's Adventures in Wonderland}, \citeauthor{Alice_s_Adventures_in_Wonderland}}\label{app:sec:sliding:Alice_s_Adventures_in_Wonderland}
\vspace{-.2cm}
\begin{figure}[h]
  \centering
  \begin{minipage}[t]{0.53\textwidth}
    \centering
    \vspace{0cm}
    \includegraphics[width=\linewidth]{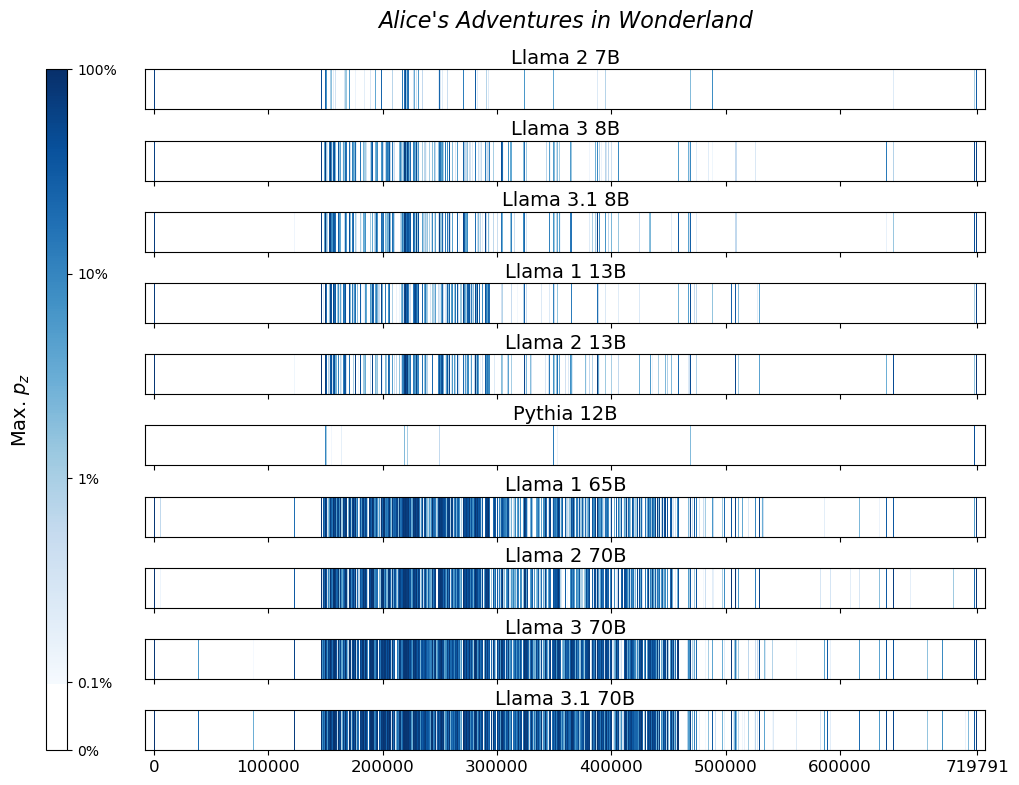}
    \includegraphics[width=\linewidth]{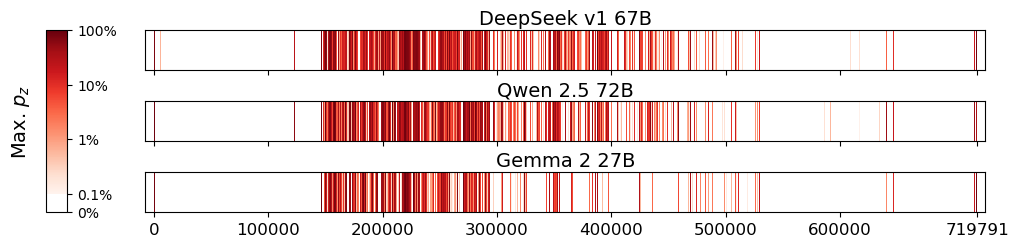}
    \includegraphics[width=\linewidth]{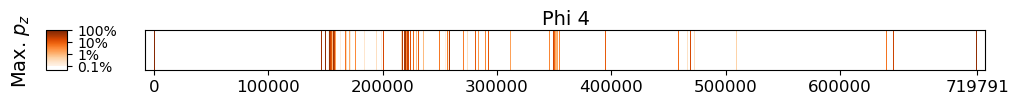}
  \end{minipage}
  \hfill
  \begin{minipage}[t]{0.45\textwidth}
    \centering
    \vspace{0cm}
    \includegraphics[width=\linewidth]{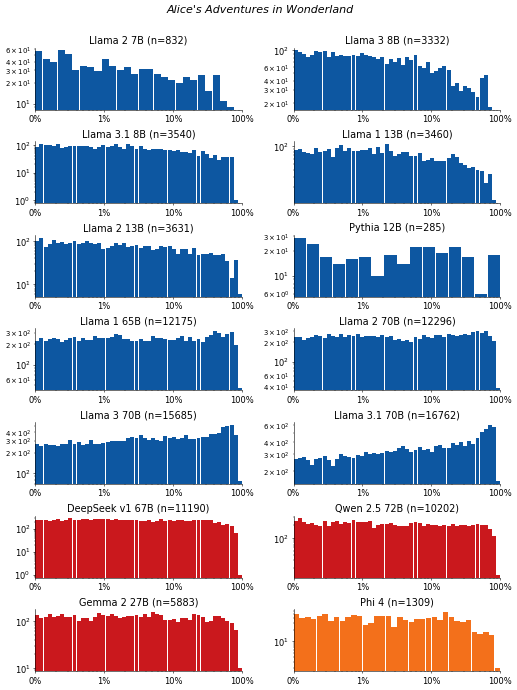}
  \end{minipage}
  \vspace{-.2cm}
  \caption{
    \textbf{\textit{Alice's Adventures in Wonderland}, \citeauthor{Alice_s_Adventures_in_Wonderland}.}
    For $14$ LLMs,
    (\textbf{left}) heatmaps for the sliding-window procedure and
    (\textbf{right}) corresponding distributions over suffix extraction probabilities
    ($\tau_\text{min}=0.1\%$).
  }
  \label{fig:slidingwindow:Alice_s_Adventures_in_Wonderland}
\end{figure}
\FloatBarrier

\clearpage
\subsubsection{\textit{The Infinity Link}, \citeauthor{The_Infinity_Link}}\label{app:sec:sliding:The_Infinity_Link}
\vspace{-.2cm}
\begin{figure}[h]
  \centering
  \begin{minipage}[t]{0.53\textwidth}
    \centering
    \vspace{0cm}
    \includegraphics[width=\linewidth]{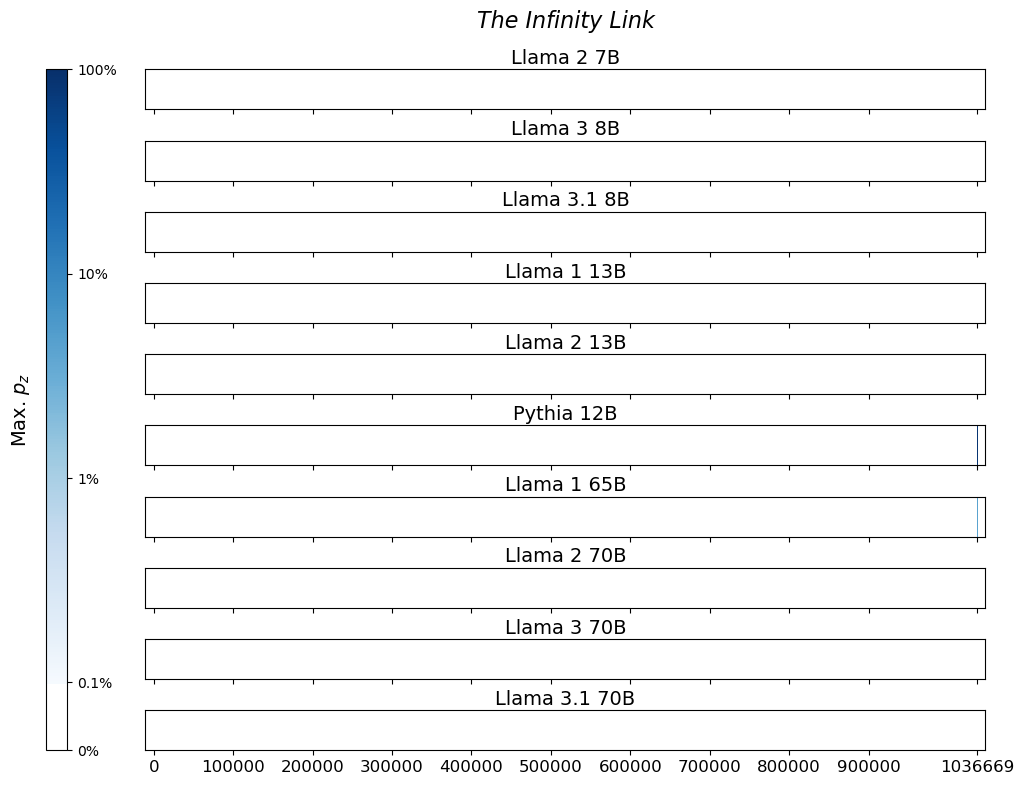}
    \includegraphics[width=\linewidth]{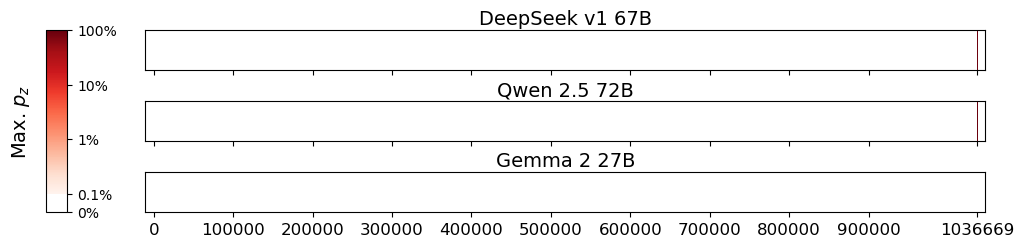}
    \includegraphics[width=\linewidth]{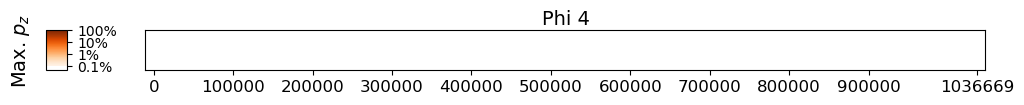}
  \end{minipage}
  \hfill
  \begin{minipage}[t]{0.45\textwidth}
    \centering
    \vspace{0cm}
    \includegraphics[width=\linewidth]{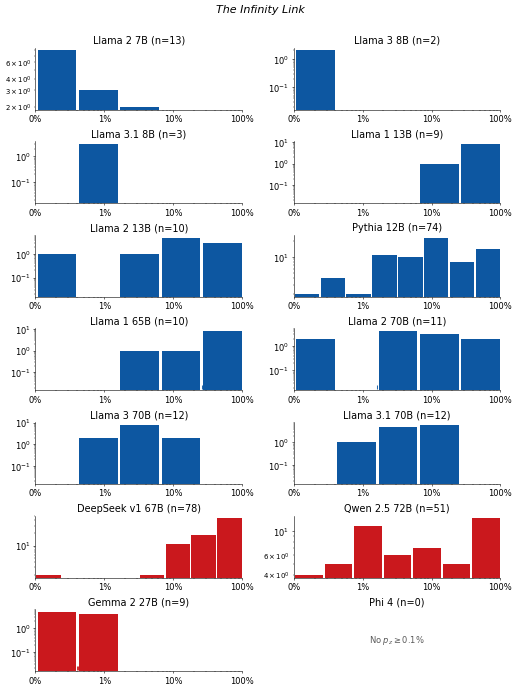}
  \end{minipage}
  \vspace{-.2cm}
  \caption{
    \textbf{\textit{The Infinity Link}, \citeauthor{The_Infinity_Link}.}
    For $14$ LLMs,
    (\textbf{left}) heatmaps for the sliding-window procedure and
    (\textbf{right}) corresponding distributions over suffix extraction probabilities
    ($\tau_\text{min}=0.1\%$).
  }
  \label{fig:slidingwindow:The_Infinity_Link}
\end{figure}
\FloatBarrier

\subsubsection{\textit{Murder on the Orient Express}, \citeauthor{Murder_on_the_Orient_Express}}\label{app:sec:sliding:Murder_on_the_Orient_Express}
\vspace{-.2cm}
\begin{figure}[h]
  \centering
  \begin{minipage}[t]{0.53\textwidth}
    \centering
    \vspace{0cm}
    \includegraphics[width=\linewidth]{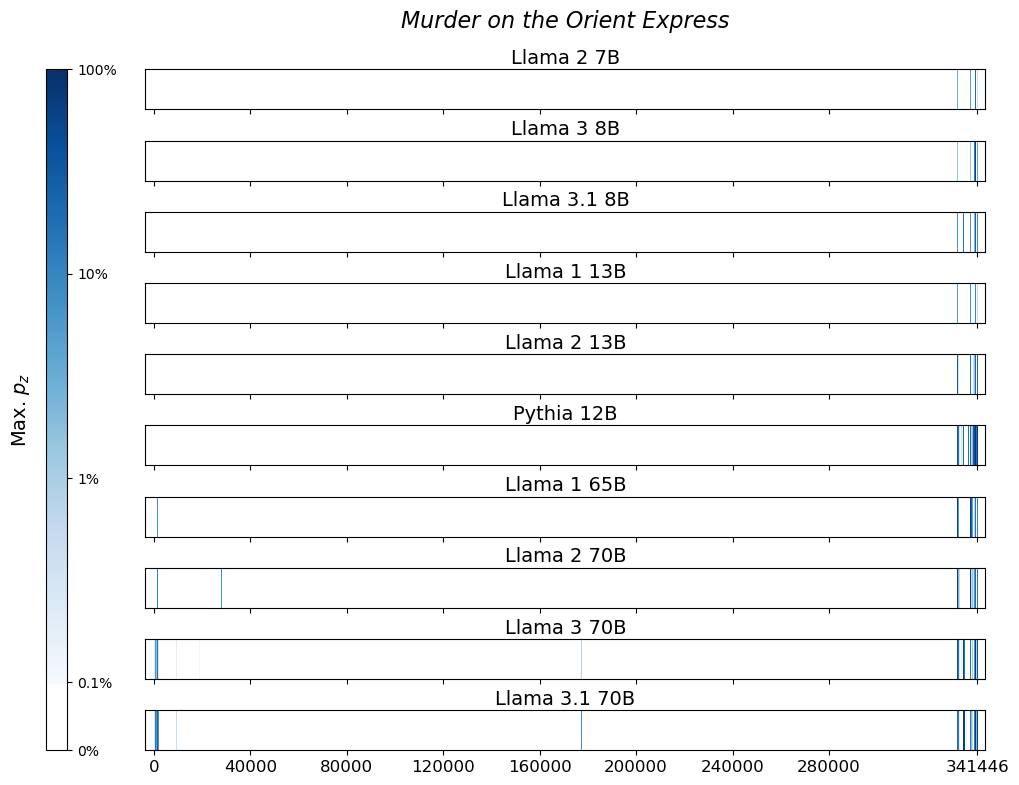}
    \includegraphics[width=\linewidth]{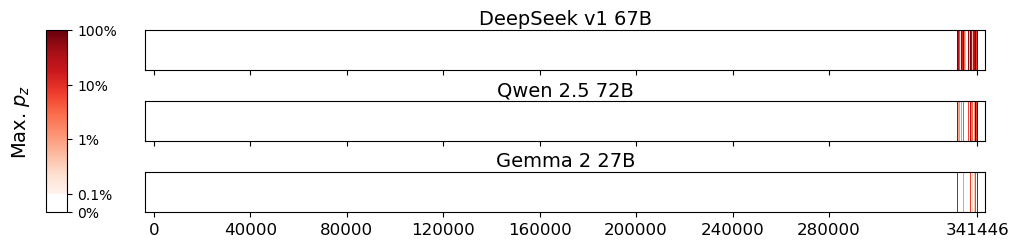}
    \includegraphics[width=\linewidth]{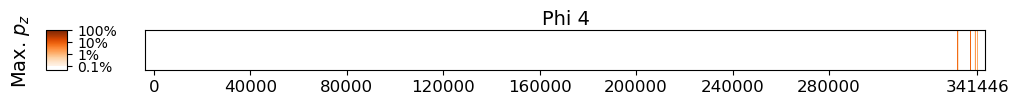}
  \end{minipage}
  \hfill
  \begin{minipage}[t]{0.45\textwidth}
    \centering
    \vspace{0cm}
    \includegraphics[width=\linewidth]{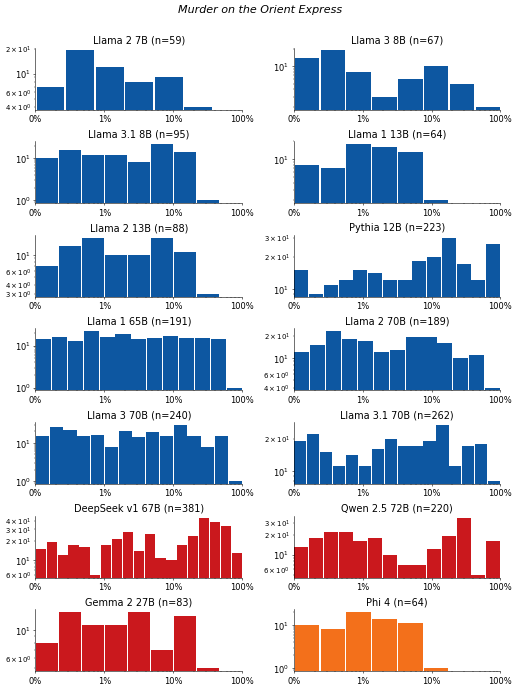}
  \end{minipage}
  \vspace{-.2cm}
  \caption{
    \textbf{\textit{Murder on the Orient Express}, \citeauthor{Murder_on_the_Orient_Express}.}
    For $14$ LLMs,
    (\textbf{left}) heatmaps for the sliding-window procedure and
    (\textbf{right}) corresponding distributions over suffix extraction probabilities
    ($\tau_\text{min}=0.1\%$).
  }
  \label{fig:slidingwindow:Murder_on_the_Orient_Express}
\end{figure}
\FloatBarrier

\clearpage
\subsubsection{\textit{And Then There Were None}, \citeauthor{And_Then_There_Were_None}}\label{app:sec:sliding:And_Then_There_Were_None}
\begin{figure}[h]
  \vspace{-.2cm}
  \centering
  \begin{minipage}[t]{0.53\textwidth}
    \centering
    \vspace{0cm}
    \includegraphics[width=\linewidth]{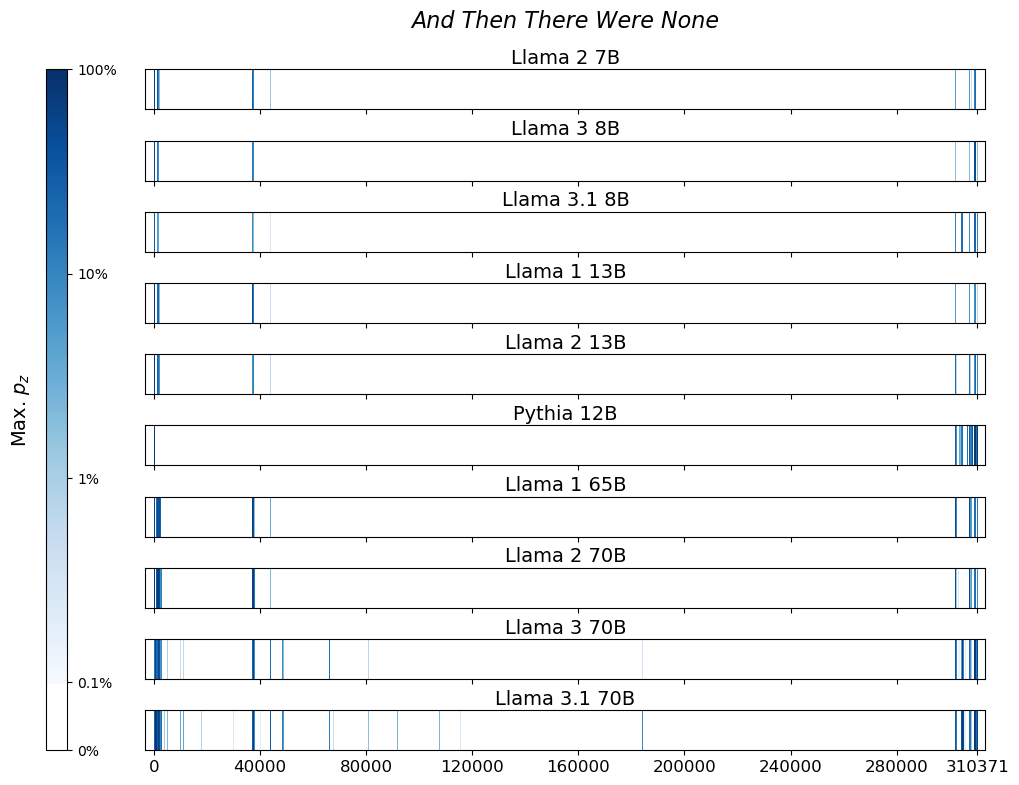}
    \includegraphics[width=\linewidth]{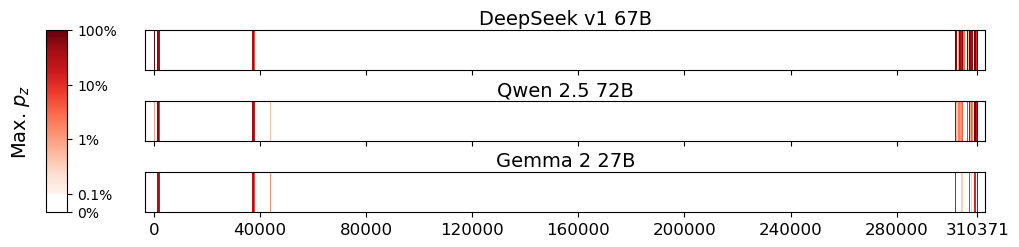}
    \includegraphics[width=\linewidth]{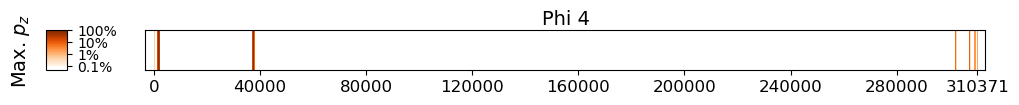}
  \end{minipage}
  \hfill
  \begin{minipage}[t]{0.45\textwidth}
    \centering
    \vspace{0cm}
    \includegraphics[width=\linewidth]{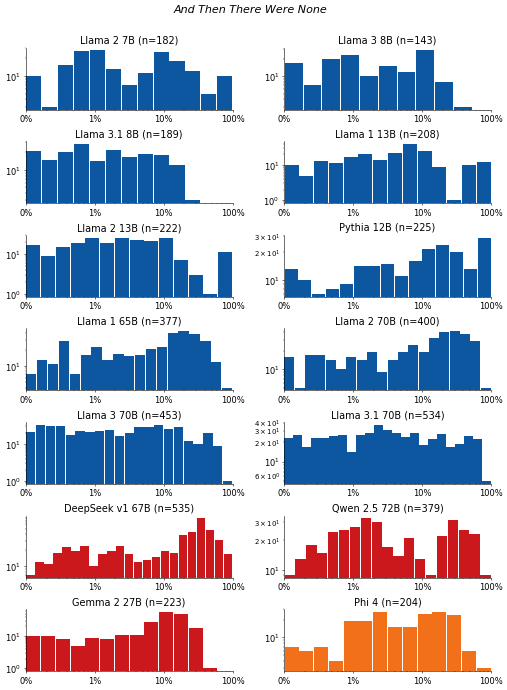}
  \end{minipage}
  \vspace{-.2cm}
  \caption{
    \textbf{\textit{And Then There Were None}, \citeauthor{And_Then_There_Were_None}.}
    For $14$ LLMs,
    (\textbf{left}) heatmaps for the sliding-window procedure and
    (\textbf{right}) corresponding distributions over suffix extraction probabilities
    ($\tau_\text{min}=0.1\%$).
  }
  \label{fig:slidingwindow:And_Then_There_Were_None}
\end{figure}
\FloatBarrier

\subsubsection{\textit{The Beautiful Struggle}, \citeauthor{The_Beautiful_Struggle}}\label{app:sec:sliding:The_Beautiful_Struggle}
\begin{figure}[h]
  \vspace{-.2cm}
  \centering
  \begin{minipage}[t]{0.53\textwidth}
    \centering
    \vspace{0cm}
    \includegraphics[width=\linewidth]{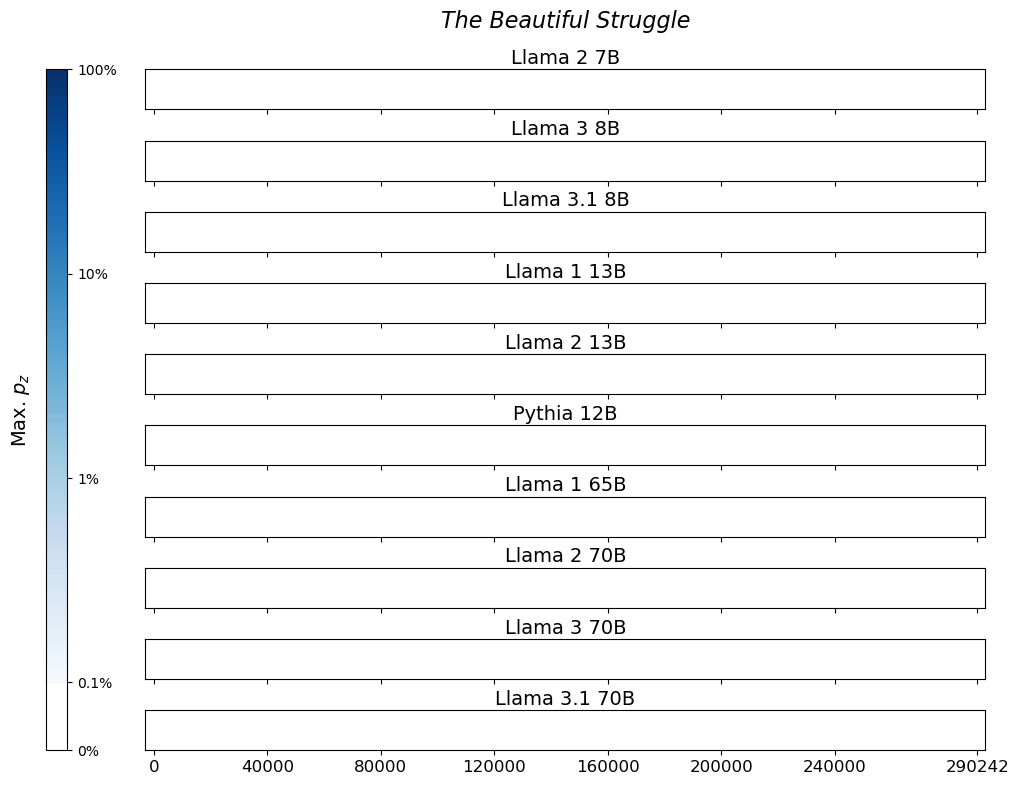}
    \includegraphics[width=\linewidth]{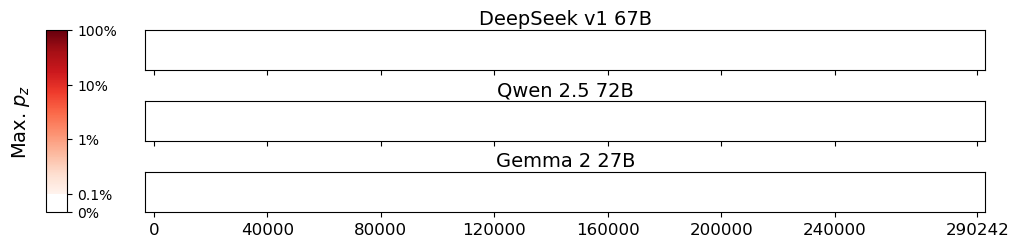}
    \includegraphics[width=\linewidth]{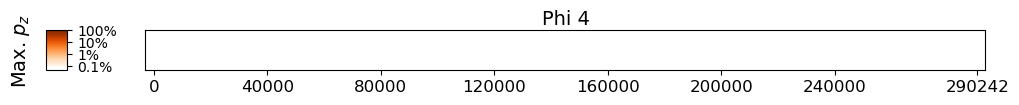}
  \end{minipage}
  \hfill
  \begin{minipage}[t]{0.45\textwidth}
    \centering
    \vspace{0cm}
    \includegraphics[width=\linewidth]{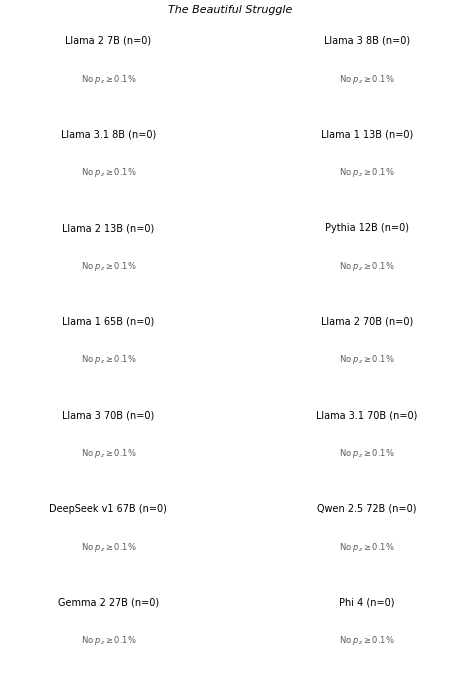}
  \end{minipage}
  \vspace{-.2cm}
  \caption{
    \textbf{\textit{The Beautiful Struggle}, \citeauthor{The_Beautiful_Struggle}.}
    For $14$ LLMs,
    (\textbf{left}) heatmaps for the sliding-window procedure and
    (\textbf{right}) corresponding distributions over suffix extraction probabilities
    ($\tau_\text{min}=0.1\%$).
  }
  \label{fig:slidingwindow:The_Beautiful_Struggle}
\end{figure}
\FloatBarrier

\clearpage
\subsubsection{\textit{We Were Eight Years in Power}, \citeauthor{We_Were_Eight_Years_in_Power}}\label{app:sec:sliding:We_Were_Eight_Years_in_Power}
\vspace{-.2cm}
\begin{figure}[h]
  \centering
  \begin{minipage}[t]{0.53\textwidth}
    \centering
    \vspace{0cm}
    \includegraphics[width=\linewidth]{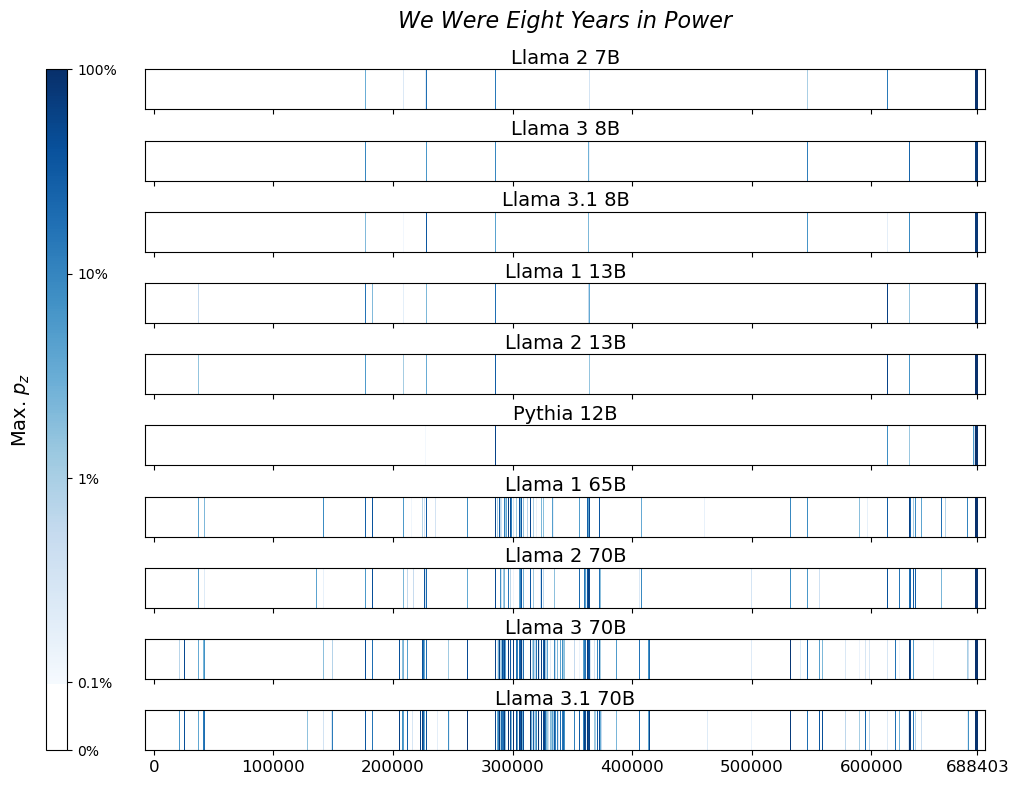}
    \includegraphics[width=\linewidth]{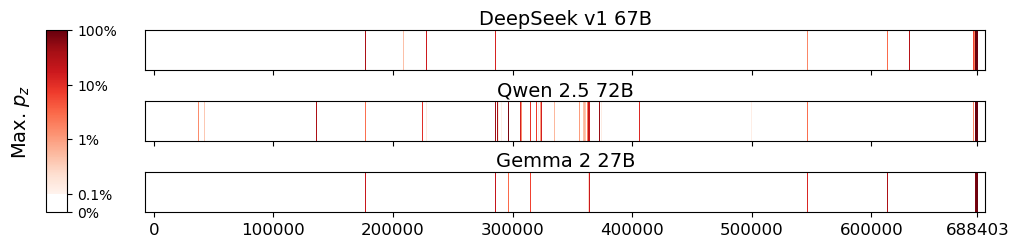}
    \includegraphics[width=\linewidth]{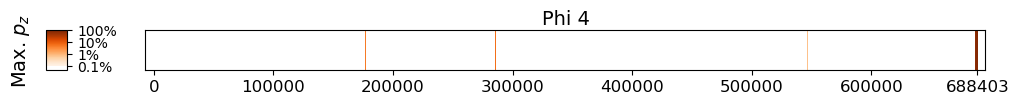}
  \end{minipage}
  \hfill
  \begin{minipage}[t]{0.45\textwidth}
    \centering
    \vspace{0cm}
    \includegraphics[width=\linewidth]{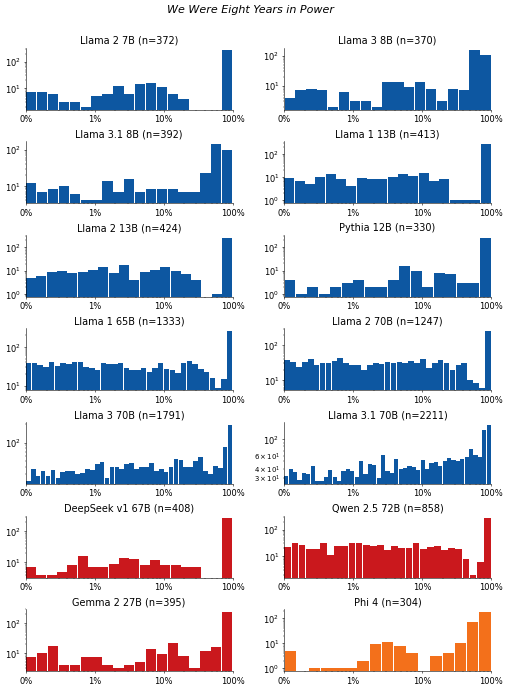}
  \end{minipage}
  \vspace{-.2cm}
  \caption{
    \textbf{\textit{We Were Eight Years in Power}, \citeauthor{We_Were_Eight_Years_in_Power}.}
    For $14$ LLMs,
    (\textbf{left}) heatmaps for the sliding-window procedure and
    (\textbf{right}) corresponding distributions over suffix extraction probabilities
    ($\tau_\text{min}=0.1\%$).
  }
  \label{fig:slidingwindow:We_Were_Eight_Years_in_Power}
\end{figure}
\FloatBarrier

\subsubsection{\textit{The Water Dancer}, \citeauthor{The_Water_Dancer}}\label{app:sec:sliding:The_Water_Dancer}
\vspace{-.2cm}
\begin{figure}[h]
  \centering
  \begin{minipage}[t]{0.53\textwidth}
    \centering
    \vspace{0cm}
    \includegraphics[width=\linewidth]{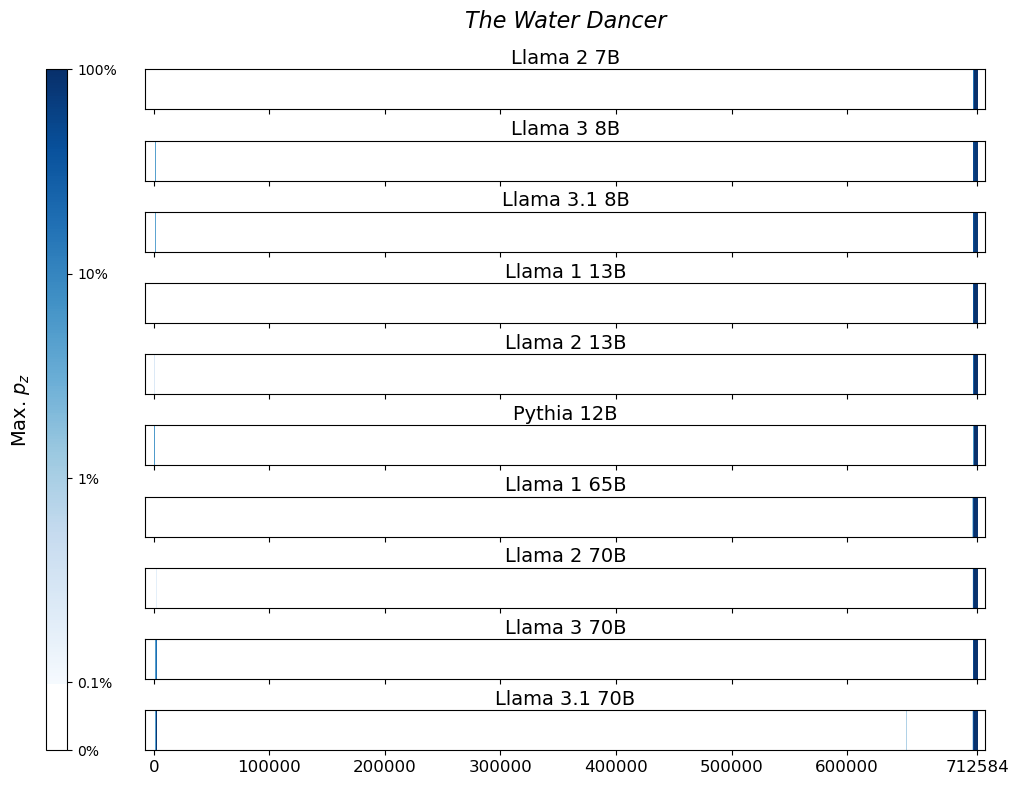}
    \includegraphics[width=\linewidth]{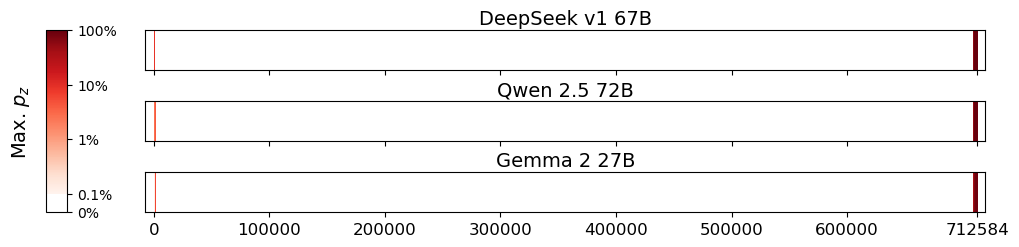}
    \includegraphics[width=\linewidth]{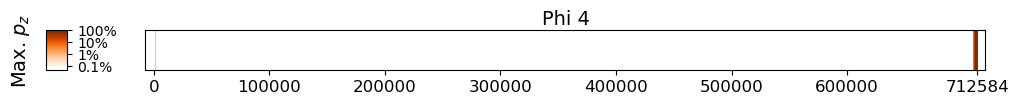}
  \end{minipage}
  \hfill
  \begin{minipage}[t]{0.45\textwidth}
    \centering
    \vspace{0cm}
    \includegraphics[width=\linewidth]{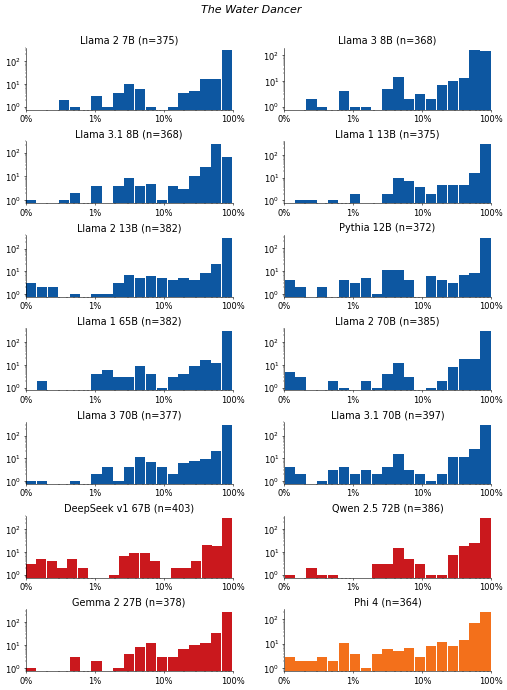}
  \end{minipage}
  \vspace{-.2cm}
  \caption{
    \textbf{\textit{The Water Dancer}, \citeauthor{The_Water_Dancer}.}
    For $14$ LLMs,
    (\textbf{left}) heatmaps for the sliding-window procedure and
    (\textbf{right}) corresponding distributions over suffix extraction probabilities
    ($\tau_\text{min}=0.1\%$).
  }
  \label{fig:slidingwindow:The_Water_Dancer}
\end{figure}
\FloatBarrier

\clearpage
\subsubsection{\textit{The Infernal Machine}, \citeauthor{The_Infernal_Machine}}\label{app:sec:sliding:The_Infernal_Machine}
\vspace{-.2cm}
\begin{figure}[h]
  \centering
  \begin{minipage}[t]{0.53\textwidth}
    \centering
    \vspace{0cm}
    \includegraphics[width=\linewidth]{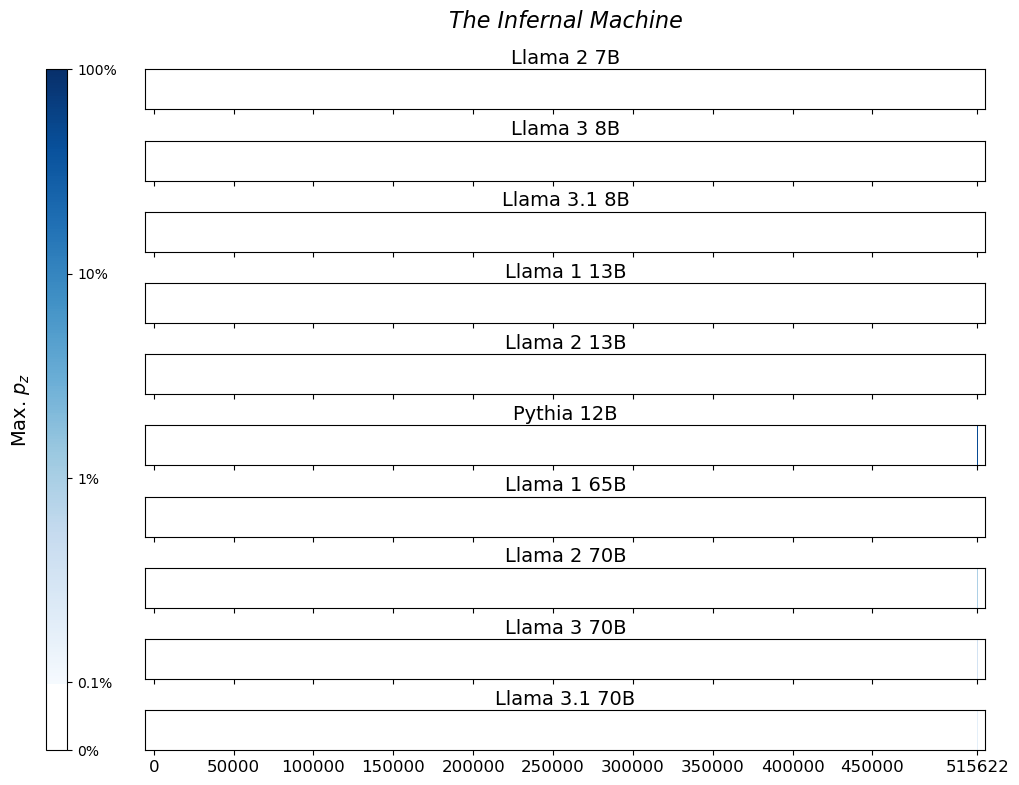}
    \includegraphics[width=\linewidth]{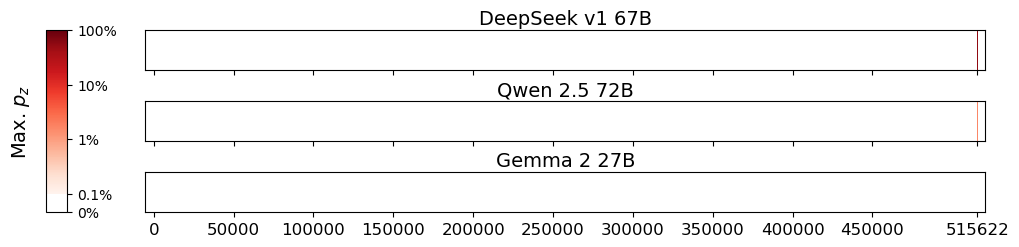}
    \includegraphics[width=\linewidth]{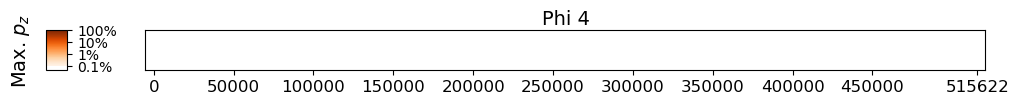}
  \end{minipage}
  \hfill
  \begin{minipage}[t]{0.45\textwidth}
    \centering
    \vspace{0cm}
    \includegraphics[width=\linewidth]{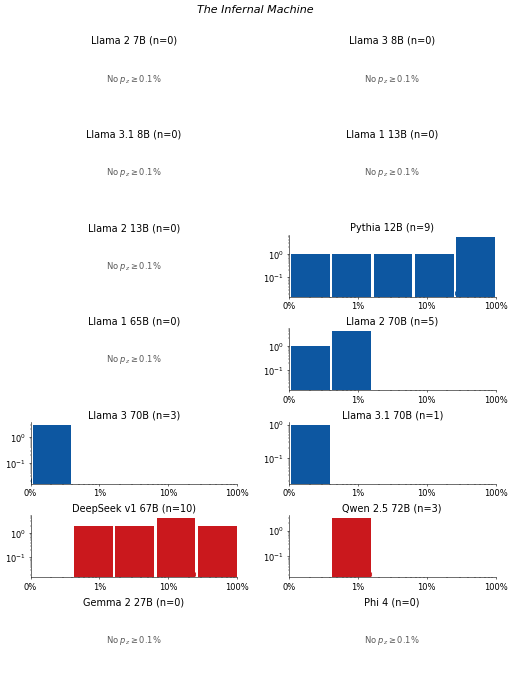}
  \end{minipage}
  \vspace{-.2cm}
  \caption{
    \textbf{\textit{The Infernal Machine}, \citeauthor{The_Infernal_Machine}.}
    For $14$ LLMs,
    (\textbf{left}) heatmaps for the sliding-window procedure and
    (\textbf{right}) corresponding distributions over suffix extraction probabilities
    ($\tau_\text{min}=0.1\%$).
  }
  \label{fig:slidingwindow:The_Infernal_Machine}
\end{figure}
\FloatBarrier

\subsubsection{\textit{The Alchemist}, \citeauthor{The_Alchemist}}\label{app:sec:sliding:The_Alchemist}
\vspace{-.2cm}
\begin{figure}[h]
  \centering
  \begin{minipage}[t]{0.53\textwidth}
    \centering
    \vspace{0cm}
    \includegraphics[width=\linewidth]{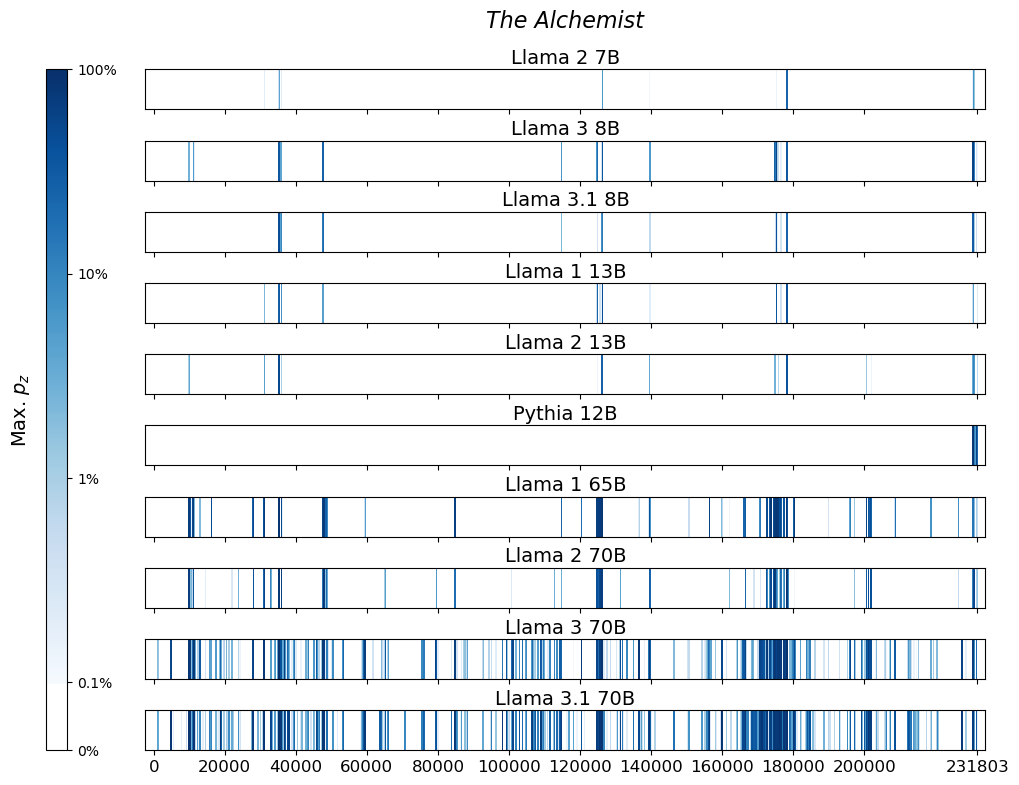}
    \includegraphics[width=\linewidth]{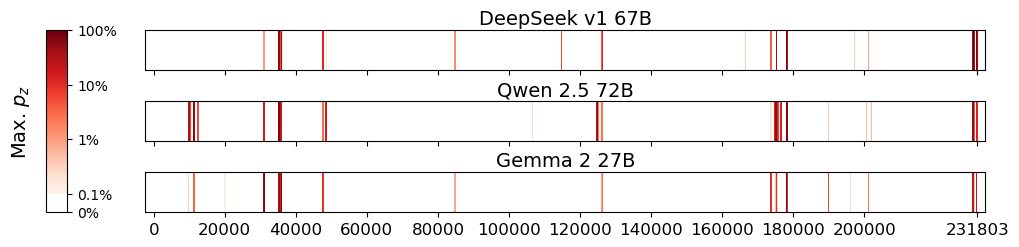}
    \includegraphics[width=\linewidth]{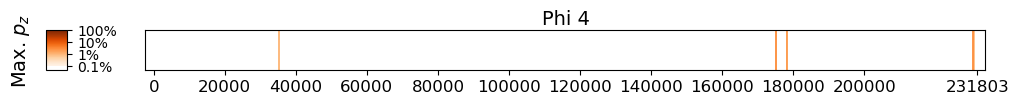}
  \end{minipage}
  \hfill
  \begin{minipage}[t]{0.45\textwidth}
    \centering
    \vspace{0cm}
    \includegraphics[width=\linewidth]{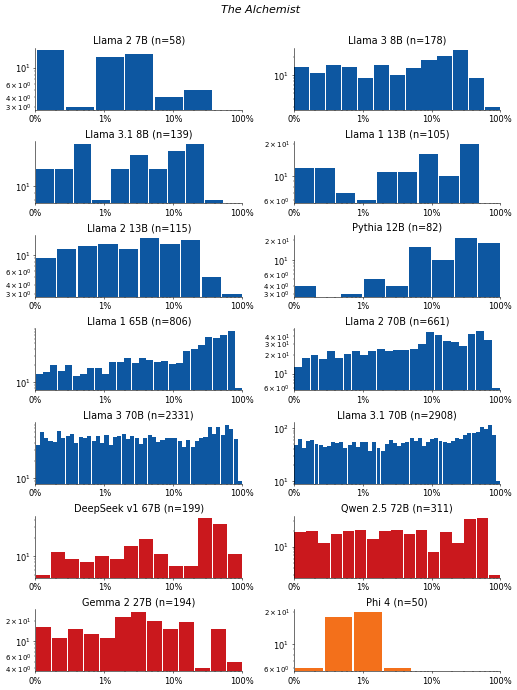}
  \end{minipage}
  \vspace{-.2cm}
  \caption{
    \textbf{\textit{The Alchemist}, \citeauthor{The_Alchemist}.}
    For $14$ LLMs,
    (\textbf{left}) heatmaps for the sliding-window procedure and
    (\textbf{right}) corresponding distributions over suffix extraction probabilities
    ($\tau_\text{min}=0.1\%$).
  }
  \label{fig:slidingwindow:The_Alchemist}
\end{figure}
\FloatBarrier

\clearpage
\subsubsection{\textit{Dungeons and Dragons and Philosophy}, \citeauthor{Dungeons_and_Dragons_and_Philosophy}}\label{app:sec:sliding:Dungeons_and_Dragons_and_Philosophy}
\begin{figure}[h]
  \vspace{-.2cm}
  \centering
  \begin{minipage}[t]{0.53\textwidth}
    \centering
    \vspace{0cm}
    \includegraphics[width=\linewidth]{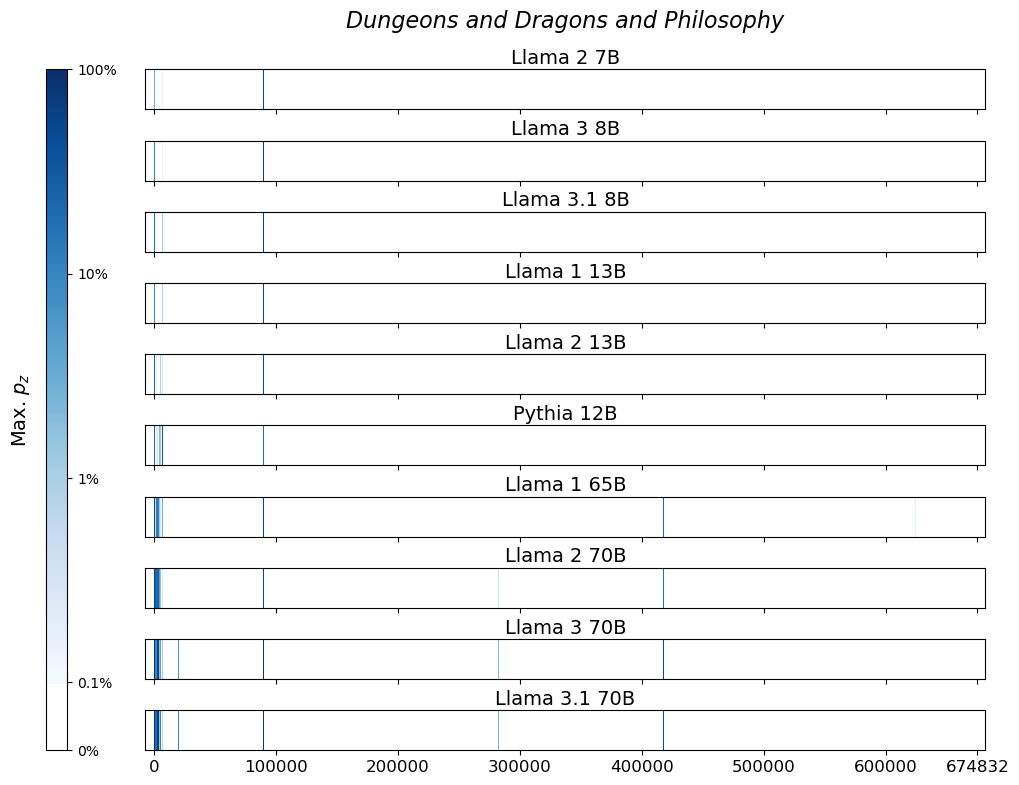}
    \includegraphics[width=\linewidth]{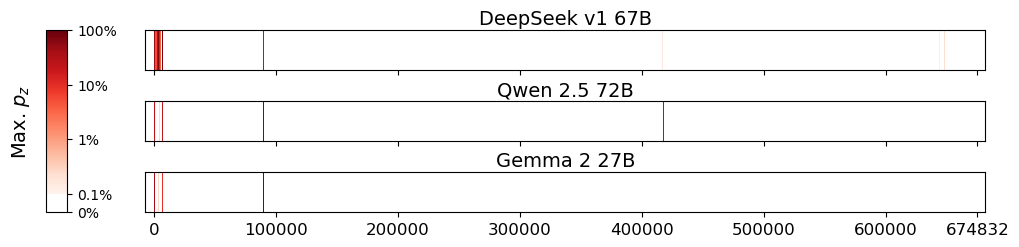}
    \includegraphics[width=\linewidth]{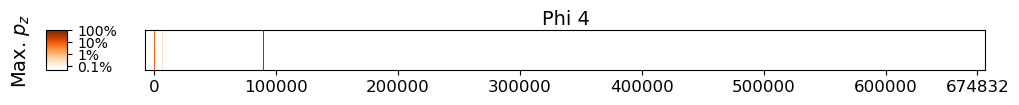}
  \end{minipage}
  \hfill
  \begin{minipage}[t]{0.45\textwidth}
    \centering
    \vspace{0cm}
    \includegraphics[width=\linewidth]{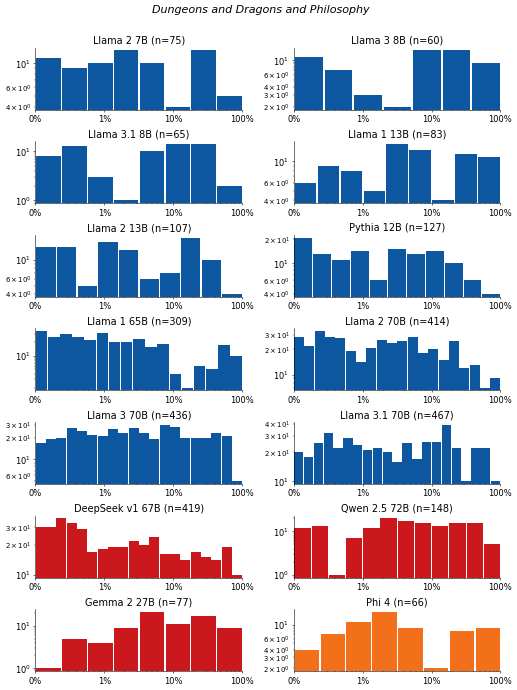}
  \end{minipage}
  \vspace{-.2cm}
  \caption{
    \textbf{\textit{Dungeons and Dragons and Philosophy}, \citeauthor{Dungeons_and_Dragons_and_Philosophy}.}
    For $14$ LLMs,
    (\textbf{left}) heatmaps for the sliding-window procedure and
    (\textbf{right}) corresponding distributions over suffix extraction probabilities
    ($\tau_\text{min}=0.1\%$).
  }
  \label{fig:slidingwindow:Dungeons_and_Dragons_and_Philosophy}
\end{figure}
\FloatBarrier

\subsubsection{\textit{Mark Rothko}, \citeauthor{Mark_Rothko}}\label{app:sec:sliding:Mark_Rothko}
\vspace{-.2cm}
\begin{figure}[h]
  \centering
  \begin{minipage}[t]{0.53\textwidth}
    \centering
    \vspace{0cm}
    \includegraphics[width=\linewidth]{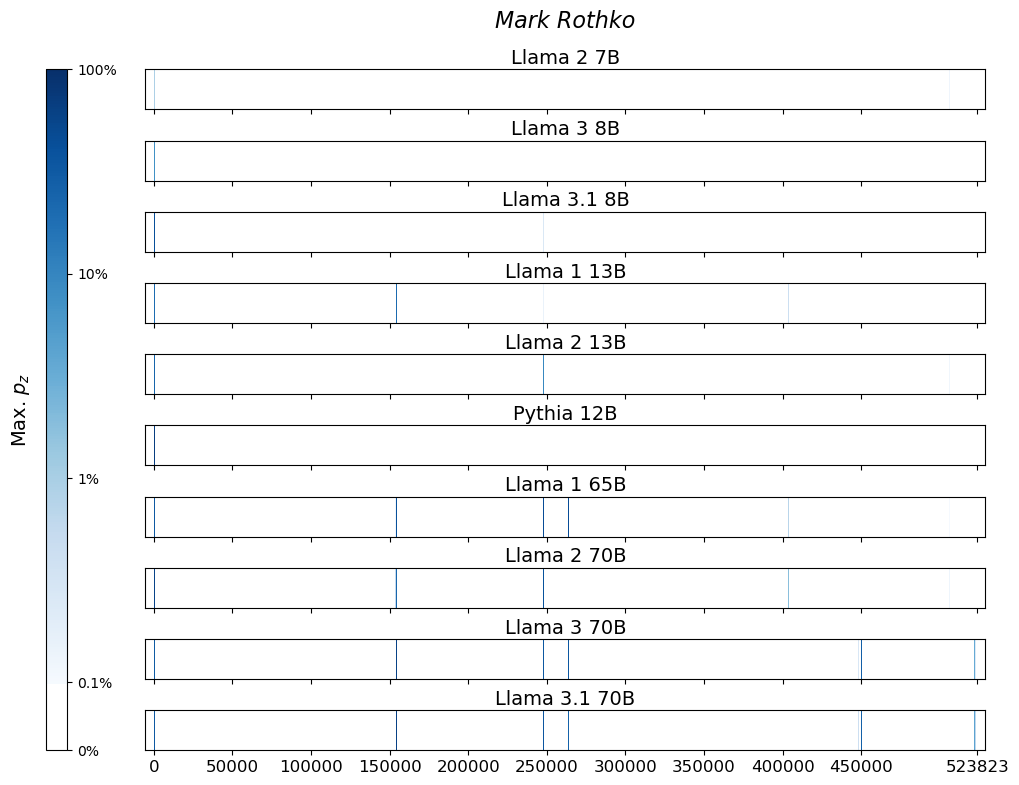}
    \includegraphics[width=\linewidth]{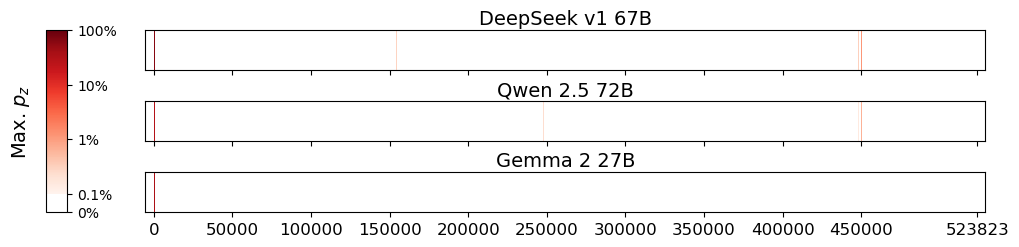}
    \includegraphics[width=\linewidth]{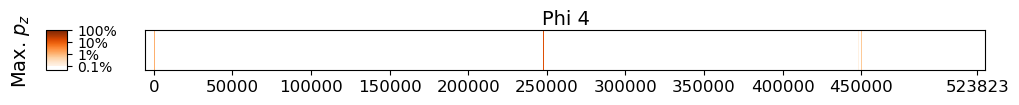}
  \end{minipage}
  \hfill
  \begin{minipage}[t]{0.45\textwidth}
    \centering
    \vspace{0cm}
    \includegraphics[width=\linewidth]{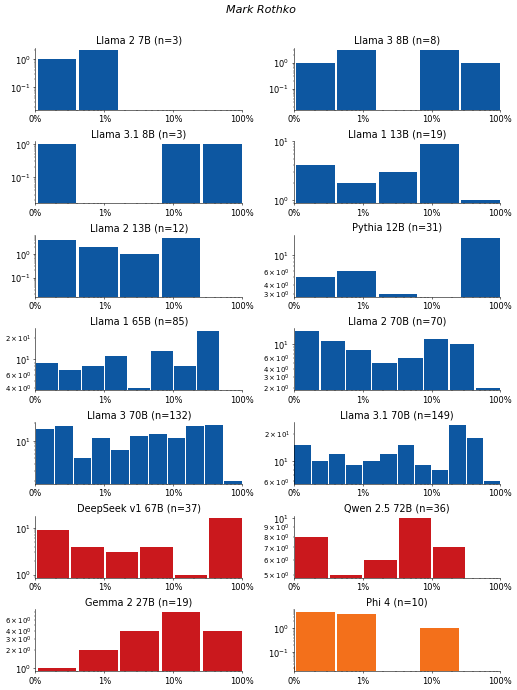}
  \end{minipage}
  \vspace{-.2cm}
  \caption{
    \textbf{\textit{Mark Rothko}, \citeauthor{Mark_Rothko}.}
    For $14$ LLMs,
    (\textbf{left}) heatmaps for the sliding-window procedure and
    (\textbf{right}) corresponding distributions over suffix extraction probabilities
    ($\tau_\text{min}=0.1\%$).
  }
  \label{fig:slidingwindow:Mark_Rothko}
\end{figure}
\FloatBarrier

\clearpage
\subsubsection{\textit{The Hunger Games}, \citeauthor{The_Hunger_Games}}\label{app:sec:sliding:The_Hunger_Games}
\vspace{-.2cm}
\begin{figure}[h]
  \centering
  \begin{minipage}[t]{0.53\textwidth}
    \centering
    \vspace{0cm}
    \includegraphics[width=\linewidth]{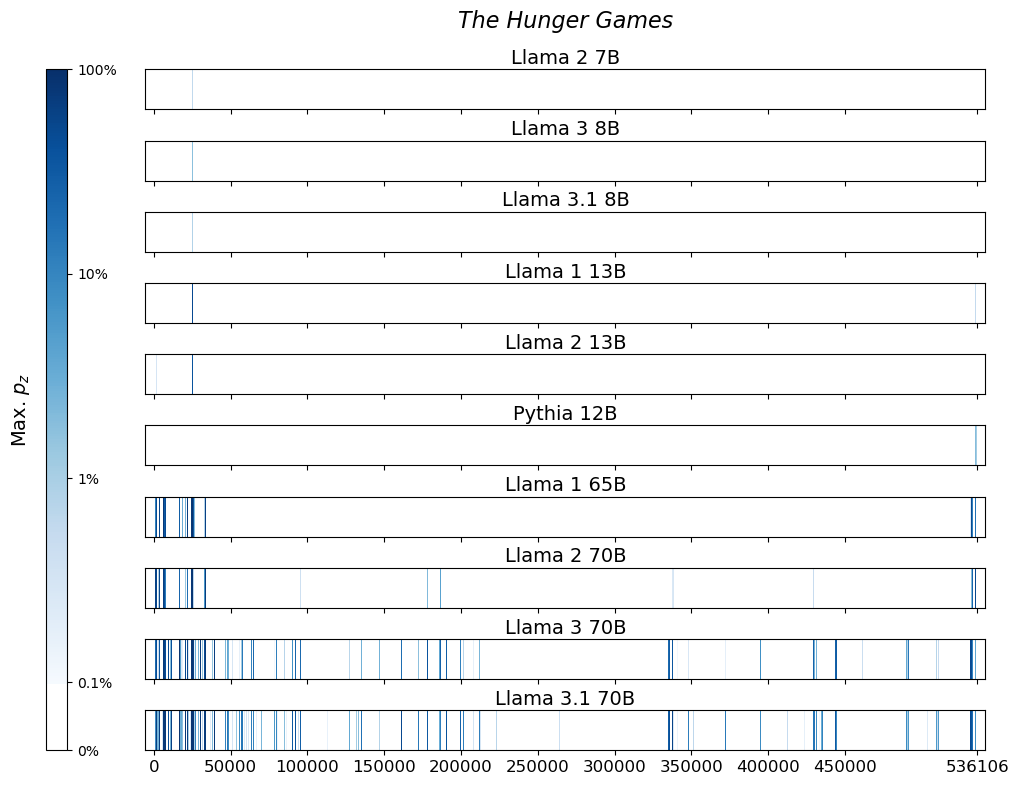}
    \includegraphics[width=\linewidth]{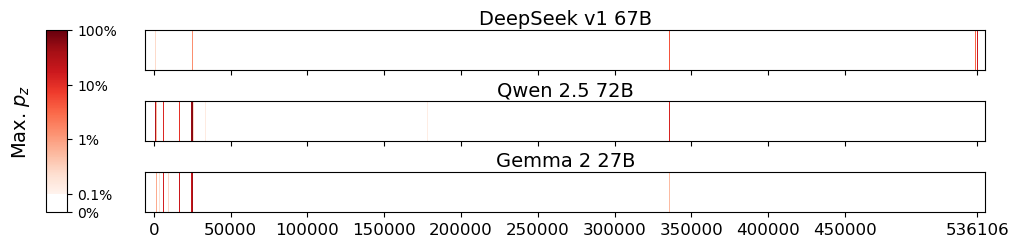}
    \includegraphics[width=\linewidth]{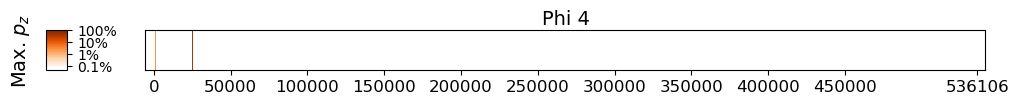}
  \end{minipage}
  \hfill
  \begin{minipage}[t]{0.45\textwidth}
    \centering
    \vspace{0cm}
    \includegraphics[width=\linewidth]{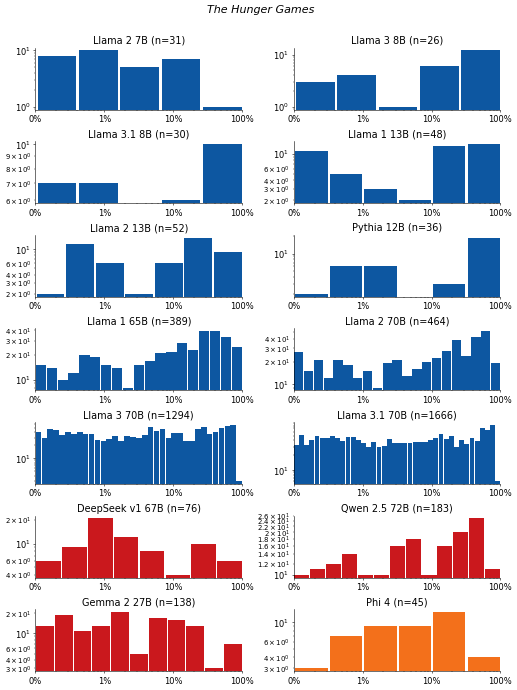}
  \end{minipage}
  \vspace{-.2cm}
  \caption{
    \textbf{\textit{The Hunger Games}, \citeauthor{The_Hunger_Games}.}
    For $14$ LLMs,
    (\textbf{left}) heatmaps for the sliding-window procedure and
    (\textbf{right}) corresponding distributions over suffix extraction probabilities
    ($\tau_\text{min}=0.1\%$).
  }
  \label{fig:slidingwindow:The_Hunger_Games}
\end{figure}
\FloatBarrier

\subsubsection{\textit{The Dragon Never Sleeps}, \citeauthor{The_Dragon_Never_Sleeps}}\label{app:sec:sliding:The_Dragon_Never_Sleeps}
\vspace{-.2cm}
\begin{figure}[h]
  \centering
  \begin{minipage}[t]{0.53\textwidth}
    \centering
    \vspace{0cm}
    \includegraphics[width=\linewidth]{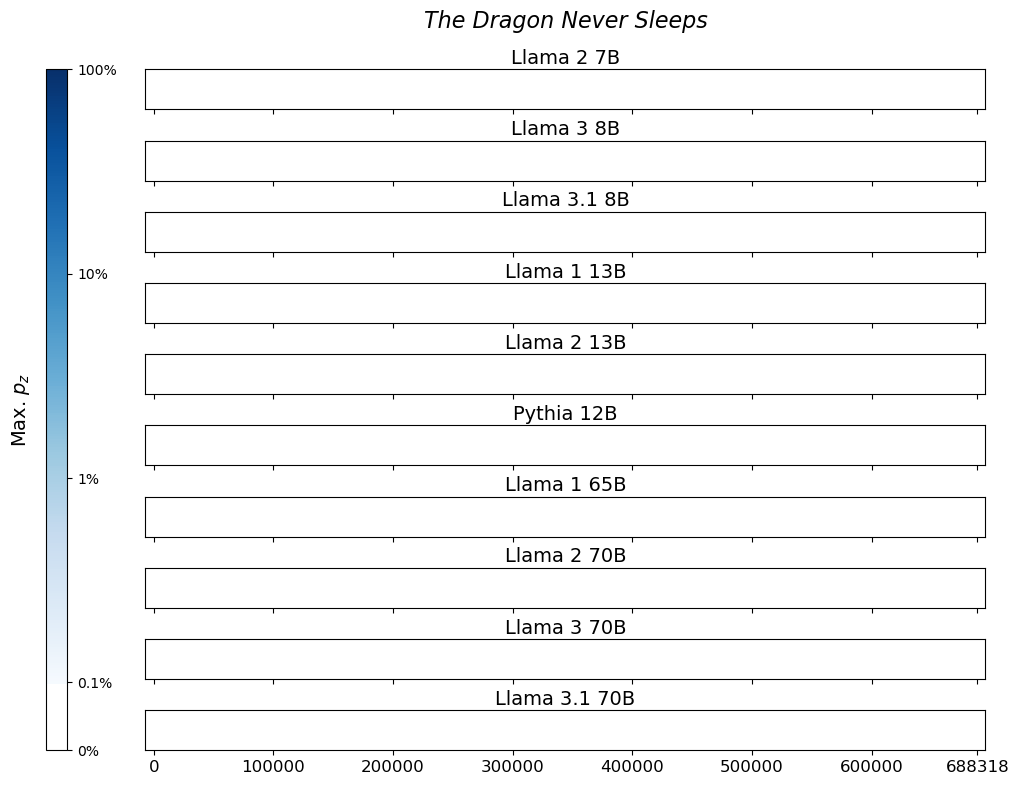}
    \includegraphics[width=\linewidth]{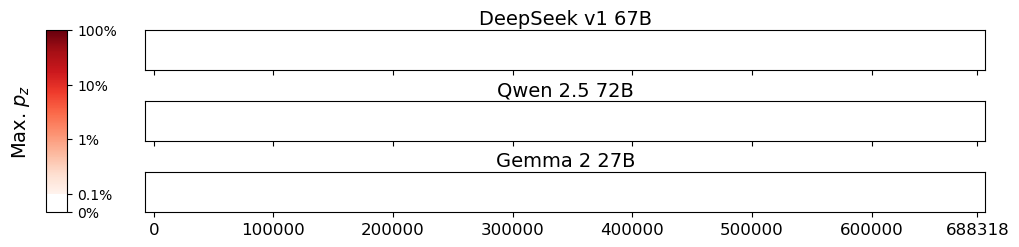}
    \includegraphics[width=\linewidth]{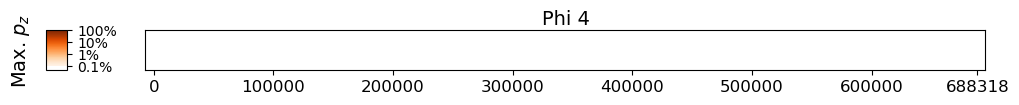}
  \end{minipage}
  \hfill
  \begin{minipage}[t]{0.45\textwidth}
    \centering
    \vspace{0cm}
    \includegraphics[width=\linewidth]{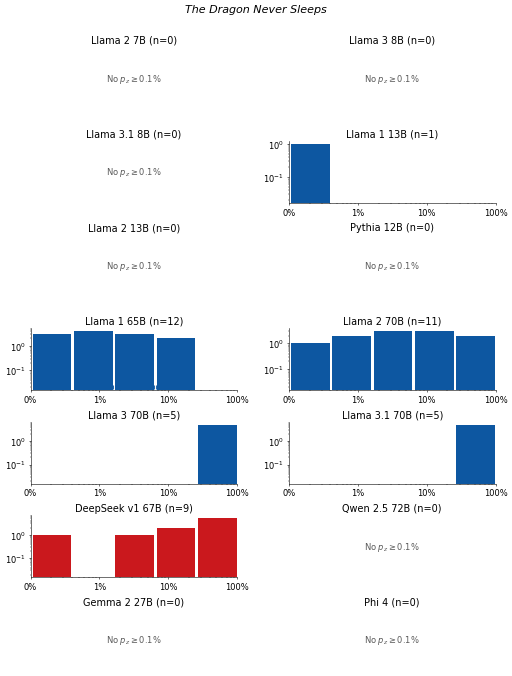}
  \end{minipage}
  \vspace{-.2cm}
  \caption{
    \textbf{\textit{The Dragon Never Sleeps}, \citeauthor{The_Dragon_Never_Sleeps}.}
    For $14$ LLMs,
    (\textbf{left}) heatmaps for the sliding-window procedure and
    (\textbf{right}) corresponding distributions over suffix extraction probabilities
    ($\tau_\text{min}=0.1\%$).
  }
  \label{fig:slidingwindow:The_Dragon_Never_Sleeps}
\end{figure}
\FloatBarrier

\clearpage
\subsubsection{\textit{The 7 Habits of Highly Effective People}, \citeauthor{The_7_Habits_of_Highly_Effective_People}}\label{app:sec:sliding:The_7_Habits_of_Highly_Effective_People}
\begin{figure}[h]
  \vspace{-.2cm}
  \centering
  \begin{minipage}[t]{0.53\textwidth}
    \centering
    \vspace{0cm}
    \includegraphics[width=\linewidth]{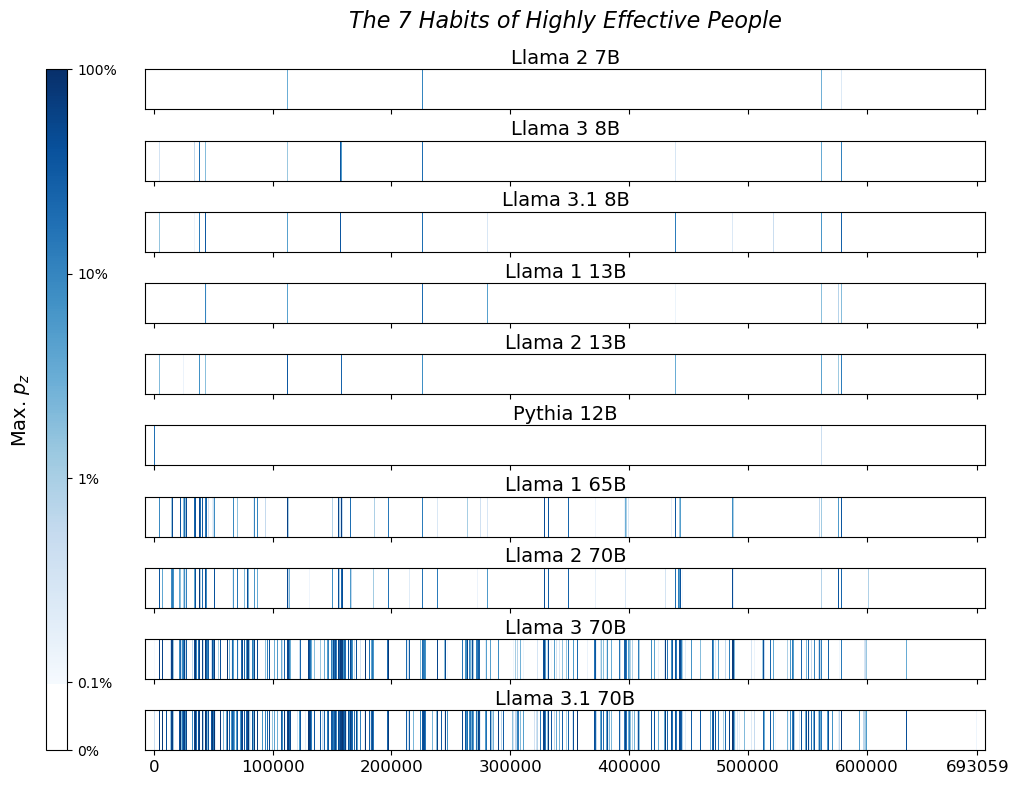}
    \includegraphics[width=\linewidth]{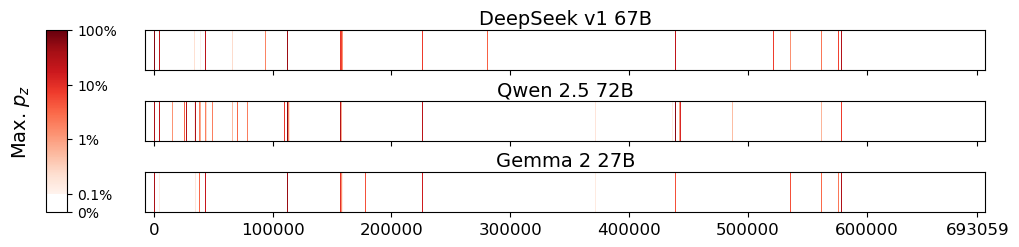}
    \includegraphics[width=\linewidth]{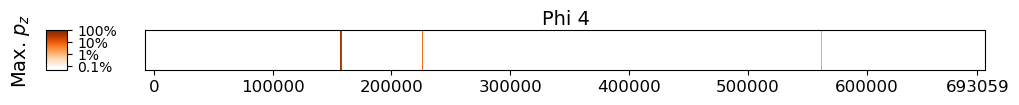}
  \end{minipage}
  \hfill
  \begin{minipage}[t]{0.45\textwidth}
    \centering
    \vspace{0cm}
    \includegraphics[width=\linewidth]{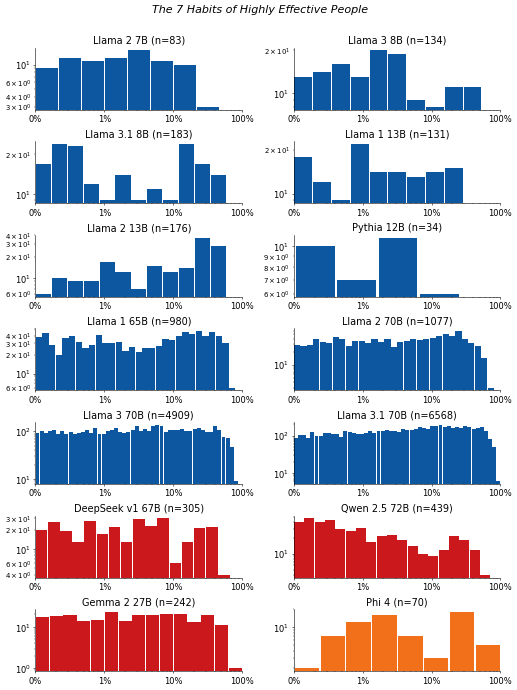}
  \end{minipage}
  \vspace{-.2cm}
  \caption{
    \textbf{\textit{The 7 Habits of Highly Effective People}, \citeauthor{The_7_Habits_of_Highly_Effective_People}.}
    For $14$ LLMs,
    (\textbf{left}) heatmaps for the sliding-window procedure and
    (\textbf{right}) corresponding distributions over suffix extraction probabilities
    ($\tau_\text{min}=0.1\%$).
  }
  \label{fig:slidingwindow:The_7_Habits_of_Highly_Effective_People}
\end{figure}
\FloatBarrier

\subsubsection{\textit{Bad Kid}, \citeauthor{Bad_Kid}}\label{app:sec:sliding:Bad_Kid}
\vspace{-.2cm}
\begin{figure}[h]
  \centering
  \begin{minipage}[t]{0.53\textwidth}
    \centering
    \vspace{0cm}
    \includegraphics[width=\linewidth]{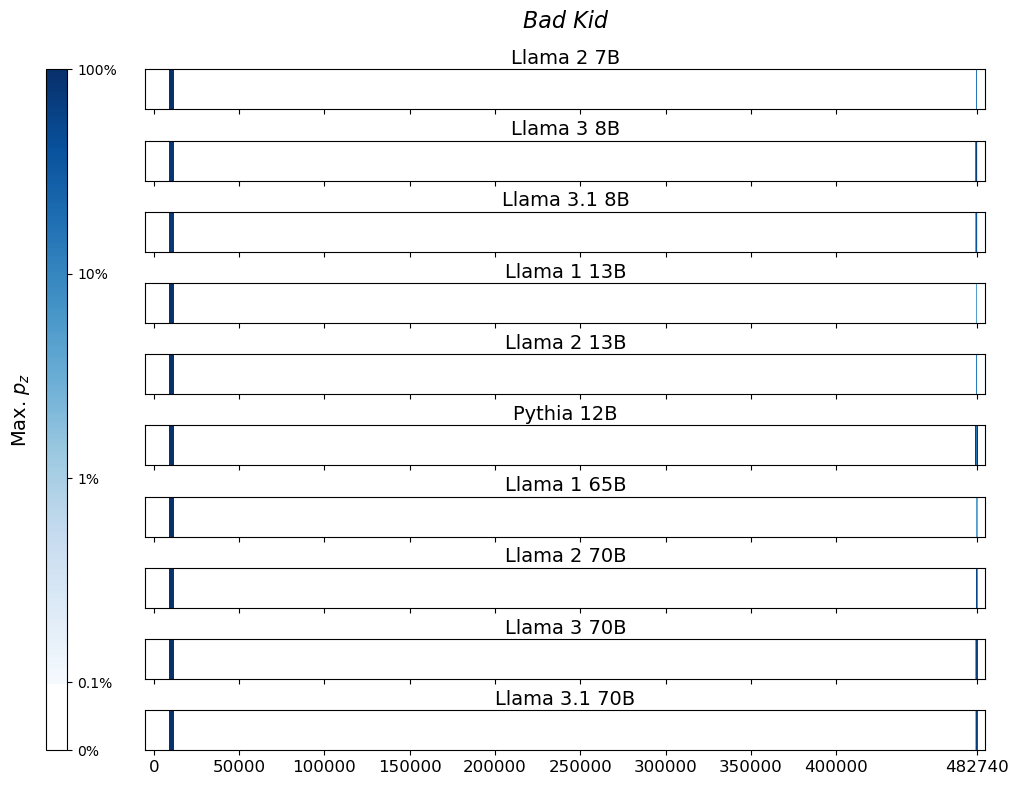}
    \includegraphics[width=\linewidth]{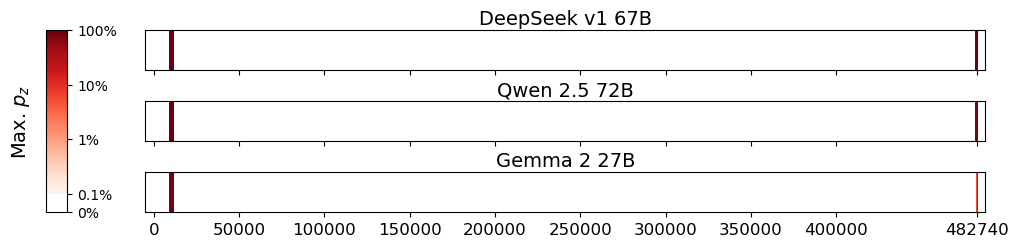}
    \includegraphics[width=\linewidth]{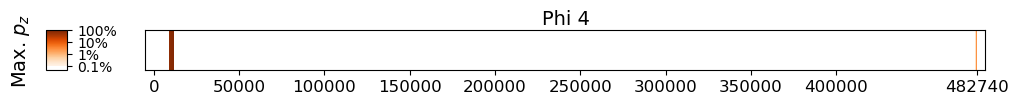}
  \end{minipage}
  \hfill
  \begin{minipage}[t]{0.45\textwidth}
    \centering
    \vspace{0cm}
    \includegraphics[width=\linewidth]{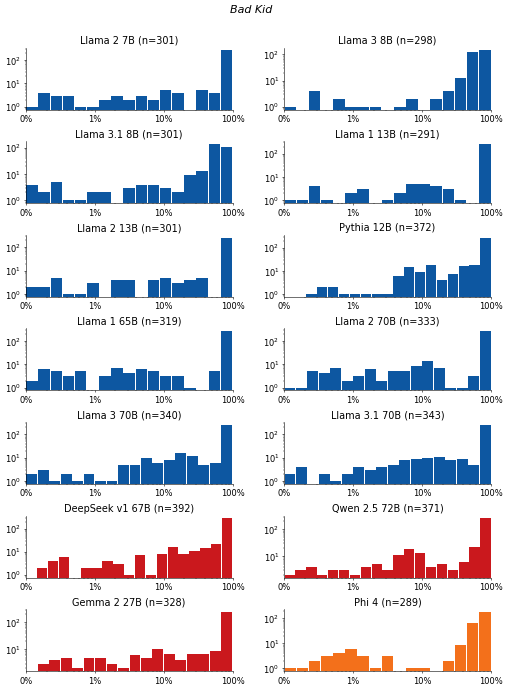}
  \end{minipage}
  \vspace{-.2cm}
  \caption{
    \textbf{\textit{Bad Kid}, \citeauthor{Bad_Kid}.}
    For $14$ LLMs,
    (\textbf{left}) heatmaps for the sliding-window procedure and
    (\textbf{right}) corresponding distributions over suffix extraction probabilities
    ($\tau_\text{min}=0.1\%$).
  }
  \label{fig:slidingwindow:Bad_Kid}
\end{figure}
\FloatBarrier

\clearpage
\subsubsection{\textit{Lullaby Town}, \citeauthor{Lullaby_Town}}\label{app:sec:sliding:Lullaby_Town}
\vspace{-.2cm}
\begin{figure}[h]
  \centering
  \begin{minipage}[t]{0.53\textwidth}
    \centering
    \vspace{0cm}
    \includegraphics[width=\linewidth]{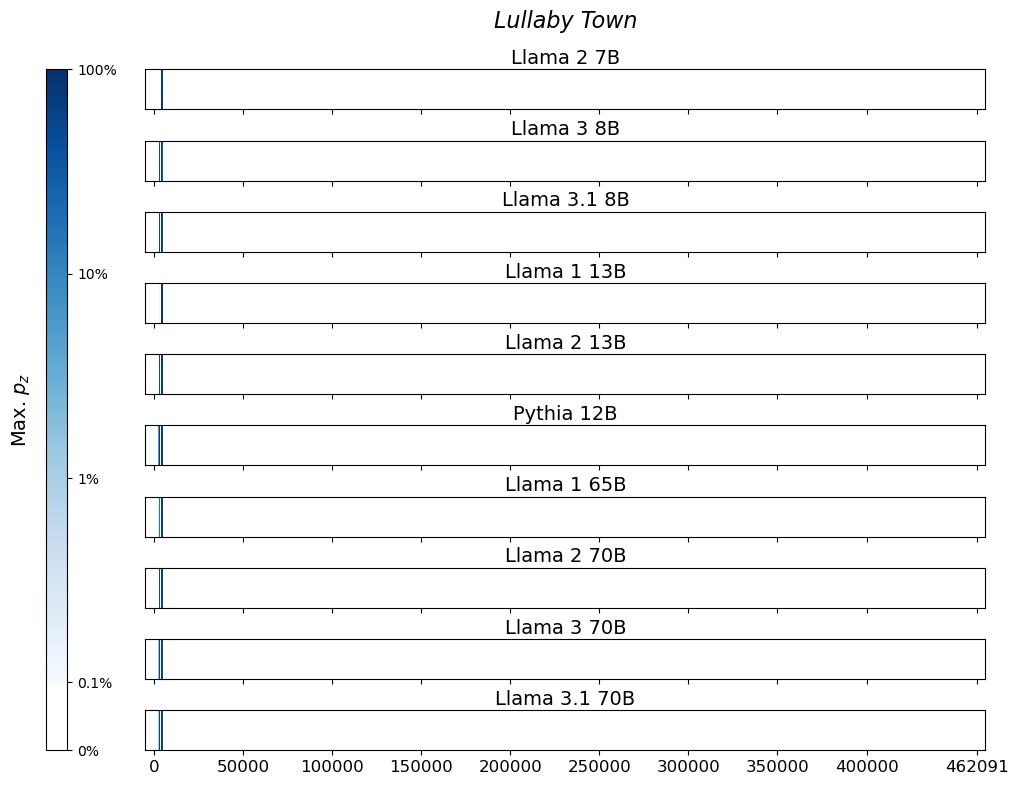}
    \includegraphics[width=\linewidth]{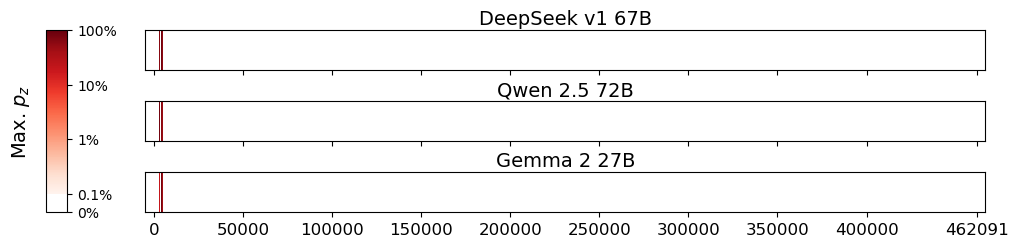}
    \includegraphics[width=\linewidth]{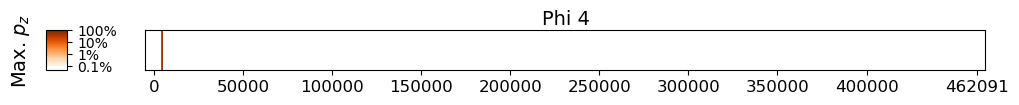}
  \end{minipage}
  \hfill
  \begin{minipage}[t]{0.45\textwidth}
    \centering
    \vspace{0cm}
    \includegraphics[width=\linewidth]{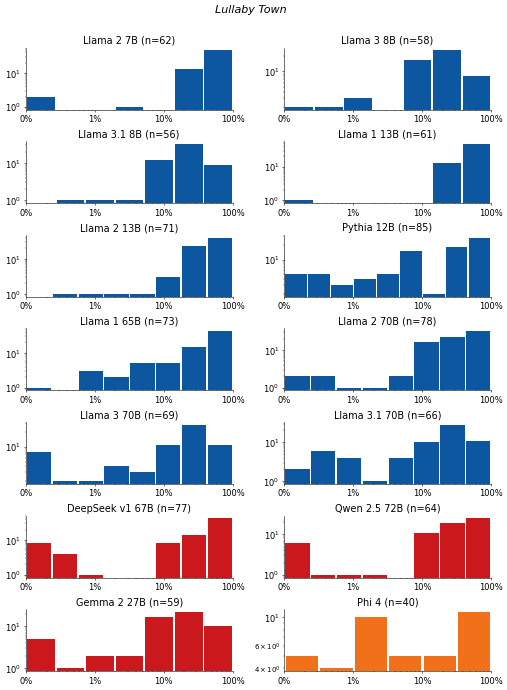}
  \end{minipage}
  \vspace{-.2cm}
  \caption{
    \textbf{\textit{Lullaby Town}, \citeauthor{Lullaby_Town}.}
    For $14$ LLMs,
    (\textbf{left}) heatmaps for the sliding-window procedure and
    (\textbf{right}) corresponding distributions over suffix extraction probabilities
    ($\tau_\text{min}=0.1\%$).
  }
  \label{fig:slidingwindow:Lullaby_Town}
\end{figure}
\FloatBarrier

\subsubsection{\textit{Jurassic Park}, \citeauthor{Jurassic_Park}}\label{app:sec:sliding:Jurassic_Park}
\vspace{-.2cm}
\begin{figure}[h]
  \centering
  \begin{minipage}[t]{0.53\textwidth}
    \centering
    \vspace{0cm}
    \includegraphics[width=\linewidth]{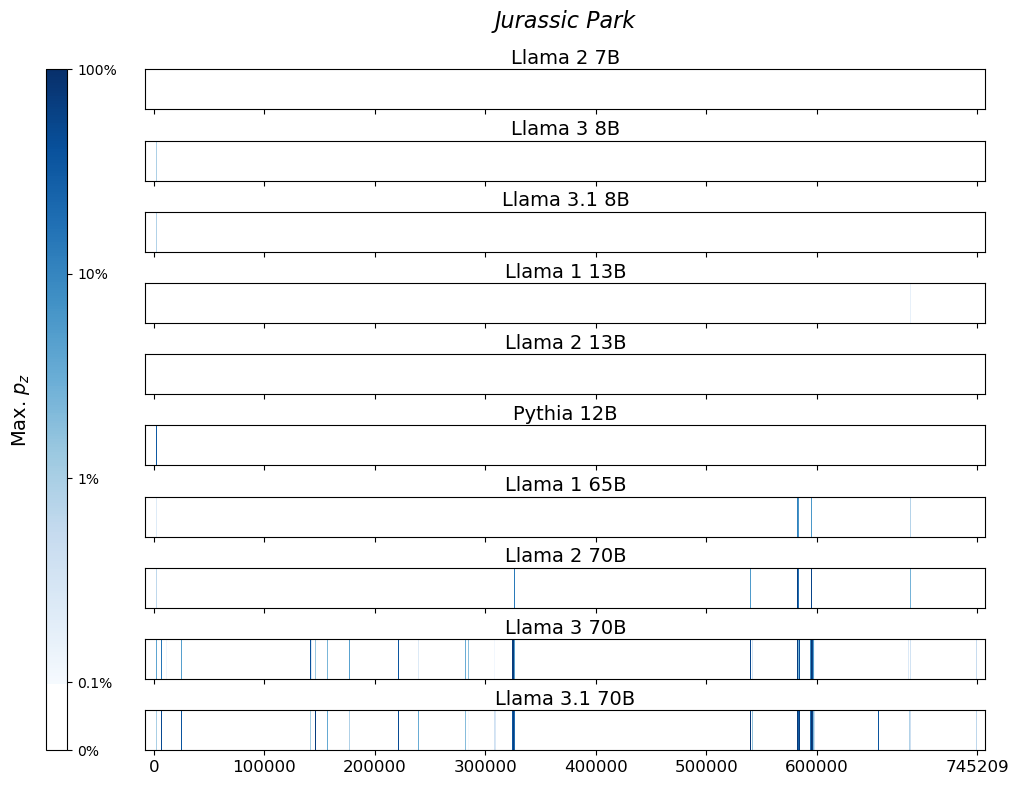}
    \includegraphics[width=\linewidth]{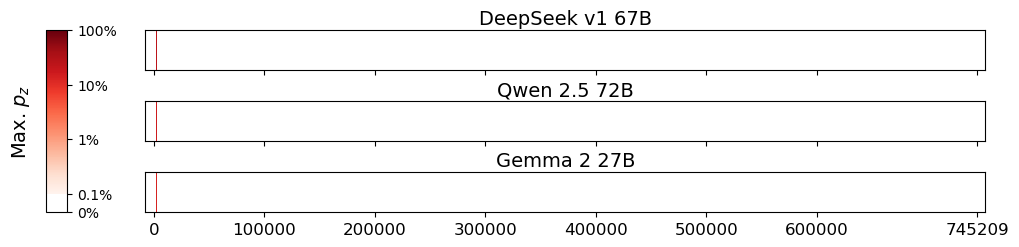}
    \includegraphics[width=\linewidth]{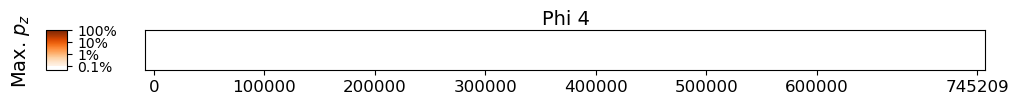}
  \end{minipage}
  \hfill
  \begin{minipage}[t]{0.45\textwidth}
    \centering
    \vspace{0cm}
    \includegraphics[width=\linewidth]{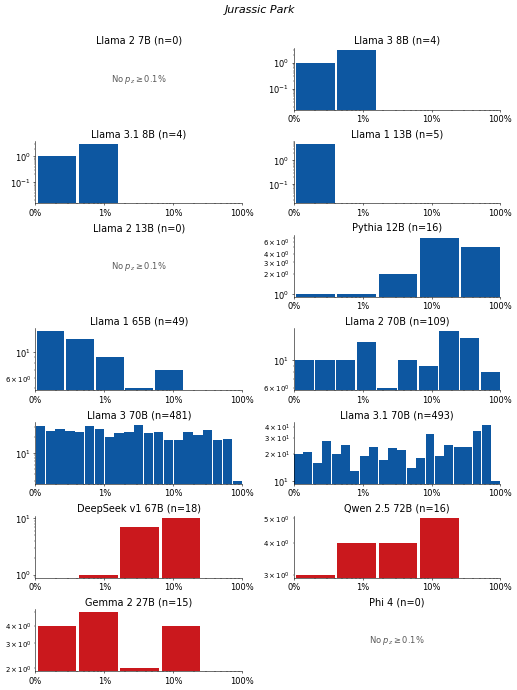}
  \end{minipage}
  \vspace{-.2cm}
  \caption{
    \textbf{\textit{Jurassic Park}, \citeauthor{Jurassic_Park}.}
    For $14$ LLMs,
    (\textbf{left}) heatmaps for the sliding-window procedure and
    (\textbf{right}) corresponding distributions over suffix extraction probabilities
    ($\tau_\text{min}=0.1\%$).
  }
  \label{fig:slidingwindow:Jurassic_Park}
\end{figure}
\FloatBarrier

\clearpage
\subsubsection{\textit{The Hours}, \citeauthor{The_Hours}}\label{app:sec:sliding:The_Hours}
\vspace{-.2cm}
\begin{figure}[h]
  \centering
  \begin{minipage}[t]{0.53\textwidth}
    \centering
    \vspace{0cm}
    \includegraphics[width=\linewidth]{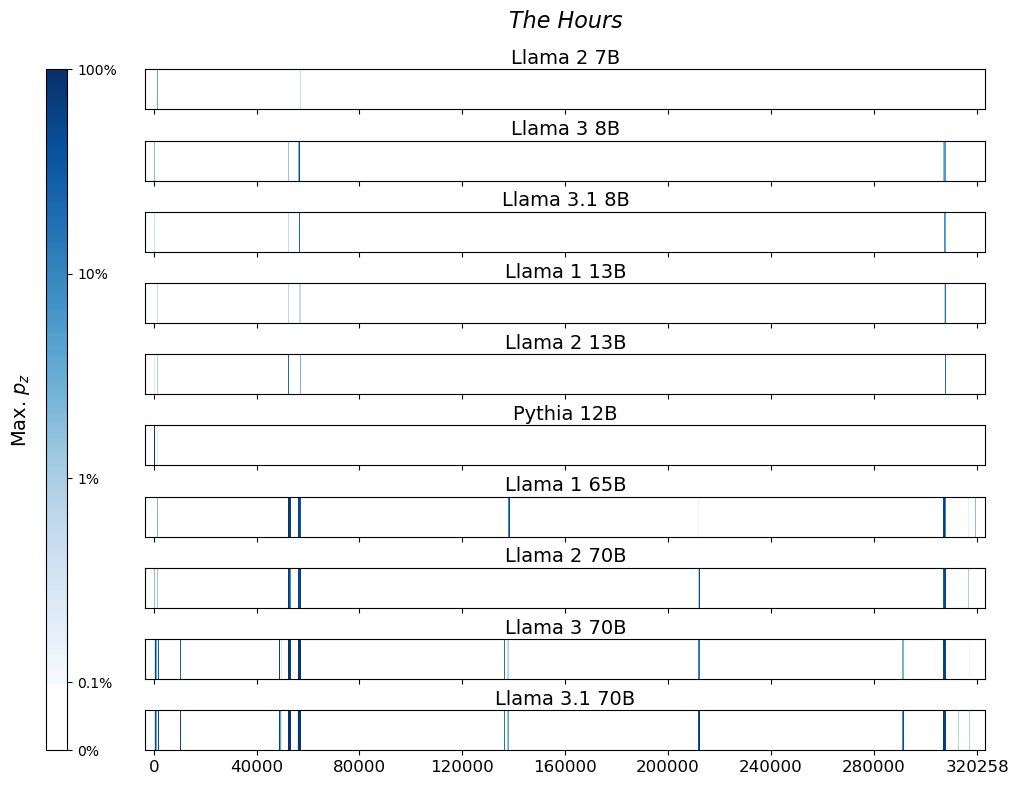}
    \includegraphics[width=\linewidth]{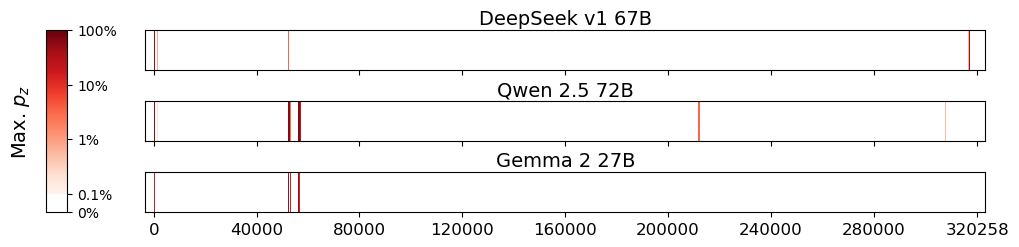}
    \includegraphics[width=\linewidth]{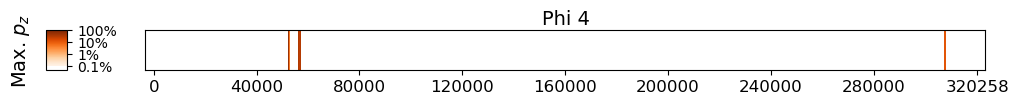}
  \end{minipage}
  \hfill
  \begin{minipage}[t]{0.45\textwidth}
    \centering
    \vspace{0cm}
    \includegraphics[width=\linewidth]{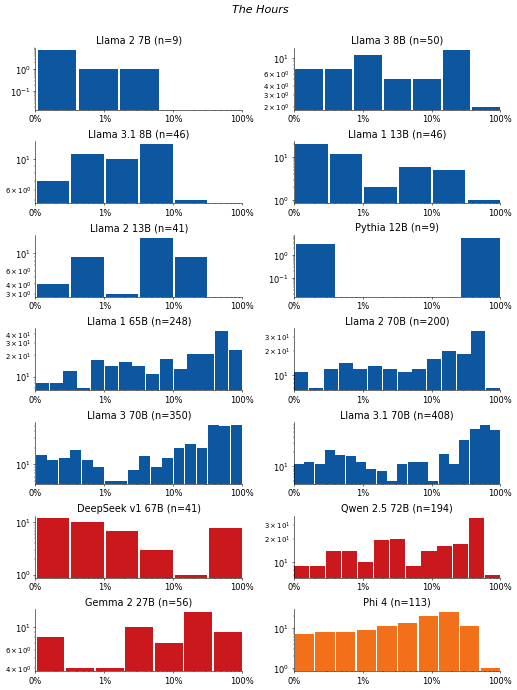}
  \end{minipage}
  \vspace{-.2cm}
  \caption{
    \textbf{\textit{The Hours}, \citeauthor{The_Hours}.}
    For $14$ LLMs,
    (\textbf{left}) heatmaps for the sliding-window procedure and
    (\textbf{right}) corresponding distributions over suffix extraction probabilities
    ($\tau_\text{min}=0.1\%$).
  }
  \label{fig:slidingwindow:The_Hours}
\end{figure}
\FloatBarrier

\subsubsection{\textit{Inhuman Land}, \citeauthor{Inhuman_Land}}\label{app:sec:sliding:Inhuman_Land}
\vspace{-.2cm}
\begin{figure}[h]
  \centering
  \begin{minipage}[t]{0.53\textwidth}
    \centering
    \vspace{0cm}
    \includegraphics[width=\linewidth]{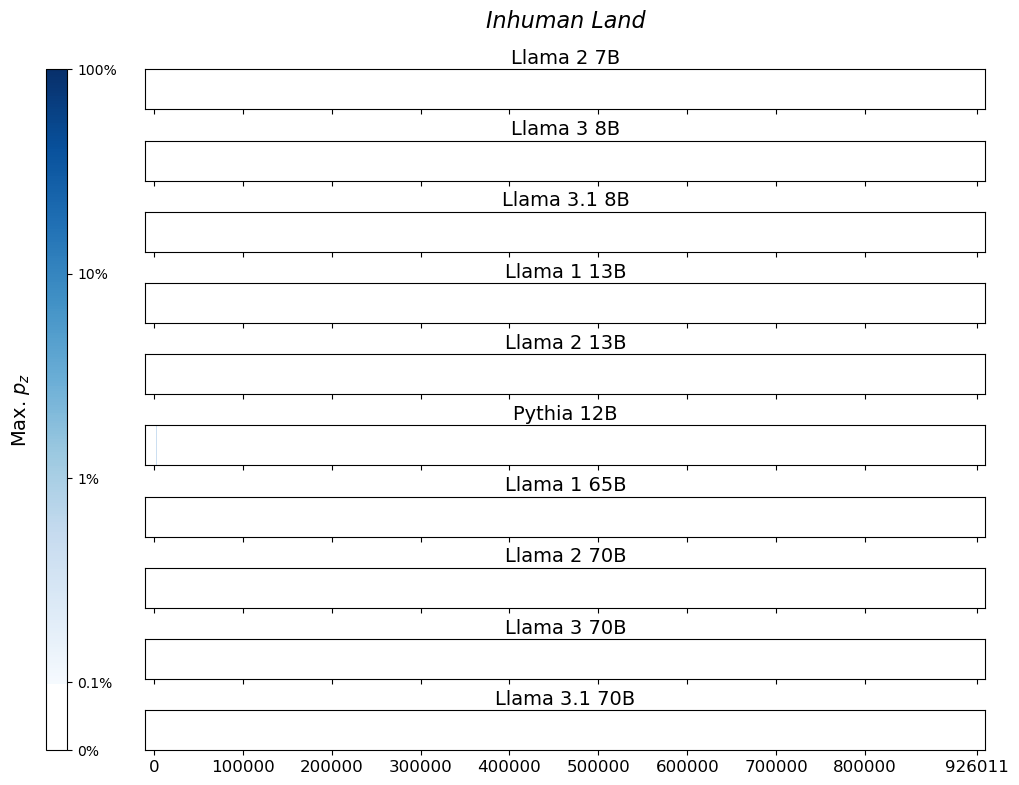}
    \includegraphics[width=\linewidth]{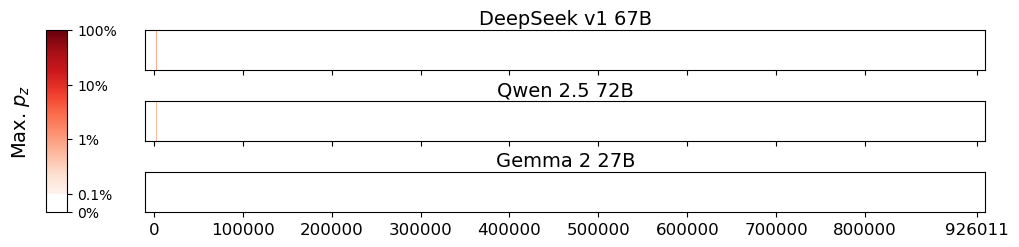}
    \includegraphics[width=\linewidth]{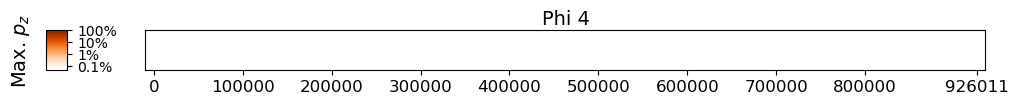}
  \end{minipage}
  \hfill
  \begin{minipage}[t]{0.45\textwidth}
    \centering
    \vspace{0cm}
    \includegraphics[width=\linewidth]{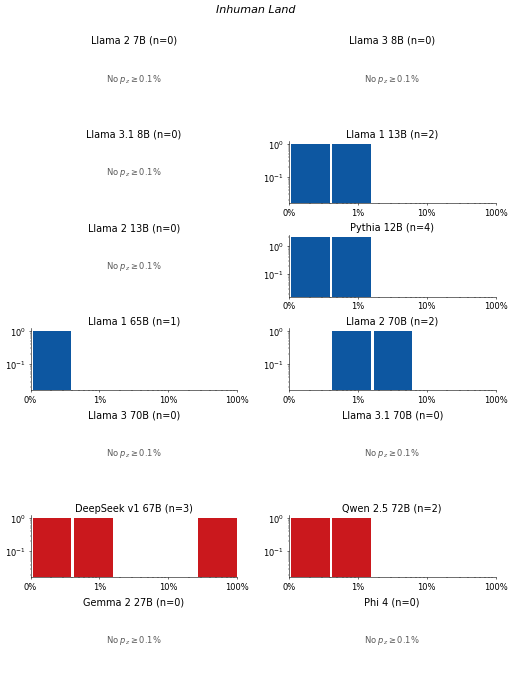}
  \end{minipage}
  \vspace{-.2cm}
  \caption{
    \textbf{\textit{Inhuman Land}, \citeauthor{Inhuman_Land}.}
    For $14$ LLMs,
    (\textbf{left}) heatmaps for the sliding-window procedure and
    (\textbf{right}) corresponding distributions over suffix extraction probabilities
    ($\tau_\text{min}=0.1\%$).
  }
  \label{fig:slidingwindow:Inhuman_Land}
\end{figure}
\FloatBarrier

\clearpage
\subsubsection{\textit{Charlie and the Chocolate Factory}, \citeauthor{Charlie_and_the_Chocolate_Factory}}\label{app:sec:sliding:Charlie_and_the_Chocolate_Factory}
\begin{figure}[h]
  \vspace{-.2cm}
  \centering
  \begin{minipage}[t]{0.53\textwidth}
    \centering
    \vspace{0cm}
    \includegraphics[width=\linewidth]{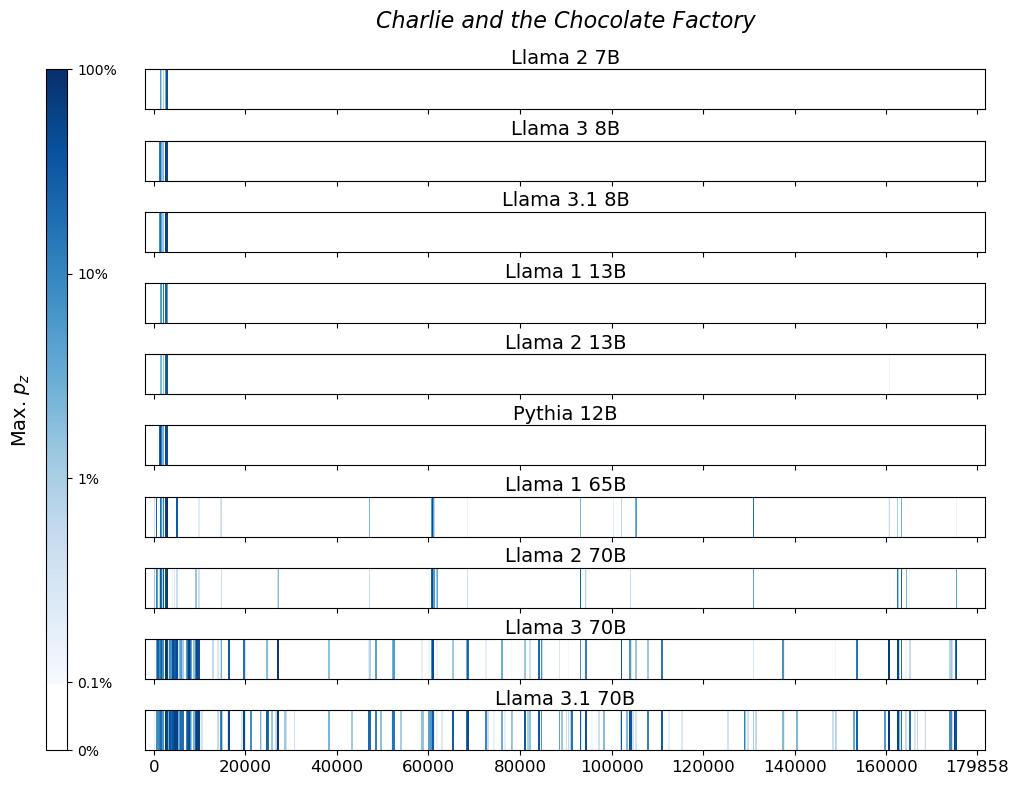}
    \includegraphics[width=\linewidth]{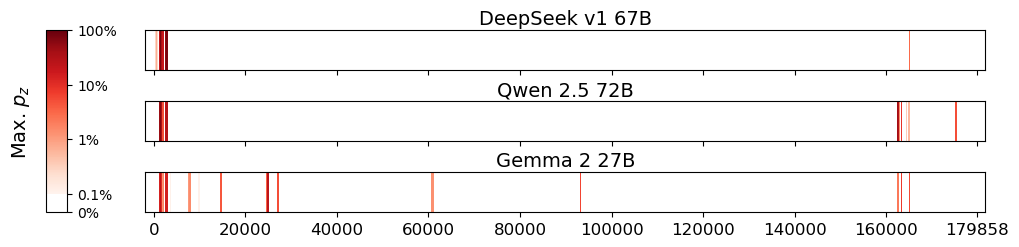}
    \includegraphics[width=\linewidth]{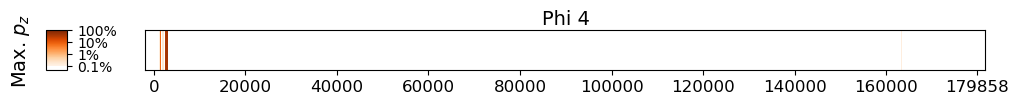}
  \end{minipage}
  \hfill
  \begin{minipage}[t]{0.45\textwidth}
    \centering
    \vspace{0cm}
    \includegraphics[width=\linewidth]{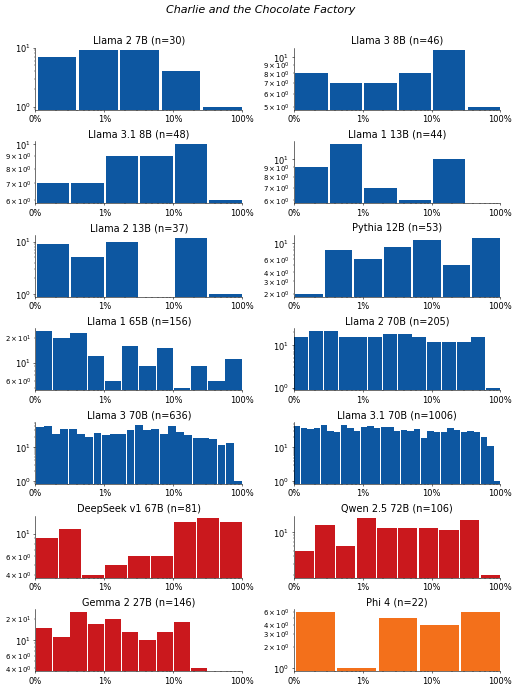}
  \end{minipage}
  \vspace{-.2cm}
  \caption{
    \textbf{\textit{Charlie and the Chocolate Factory}, \citeauthor{Charlie_and_the_Chocolate_Factory}.}
    For $14$ LLMs,
    (\textbf{left}) heatmaps for the sliding-window procedure and
    (\textbf{right}) corresponding distributions over suffix extraction probabilities
    ($\tau_\text{min}=0.1\%$).
  }
  \label{fig:slidingwindow:Charlie_and_the_Chocolate_Factory}
\end{figure}
\FloatBarrier

\subsubsection{\textit{James and the Giant Peach}, \citeauthor{James_and_the_Giant_Peach}}\label{app:sec:sliding:James_and_the_Giant_Peach}
\vspace{-.2cm}
\begin{figure}[h]
  \centering
  \begin{minipage}[t]{0.53\textwidth}
    \centering
    \vspace{0cm}
    \includegraphics[width=\linewidth]{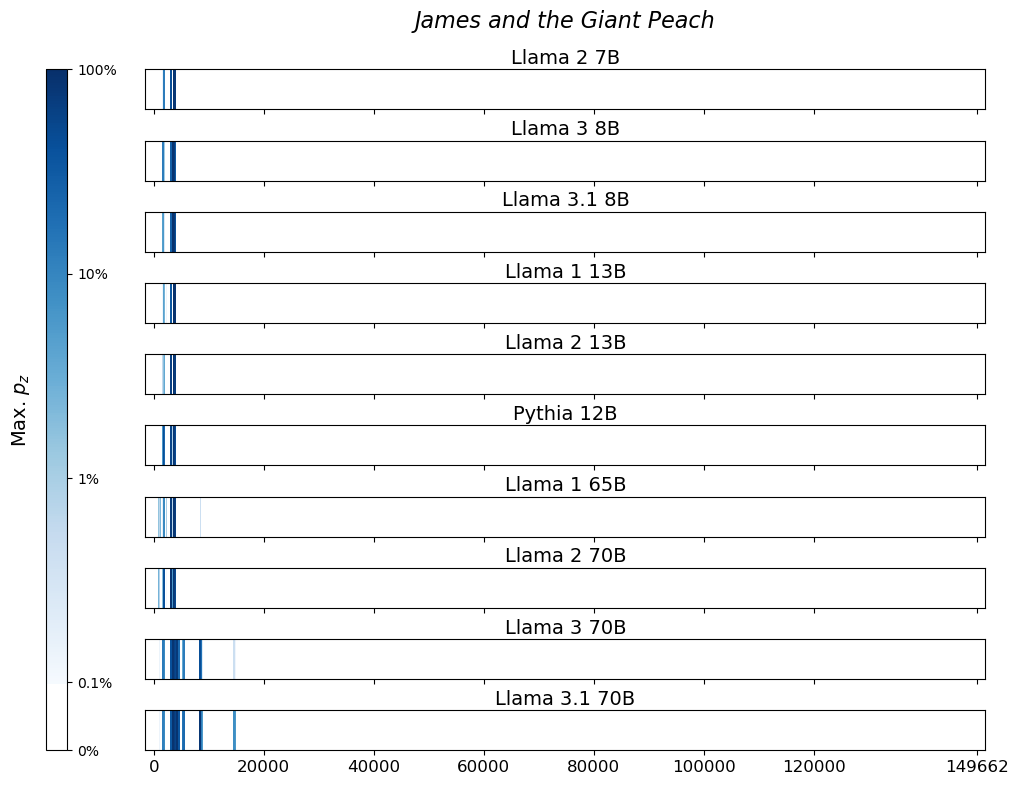}
    \includegraphics[width=\linewidth]{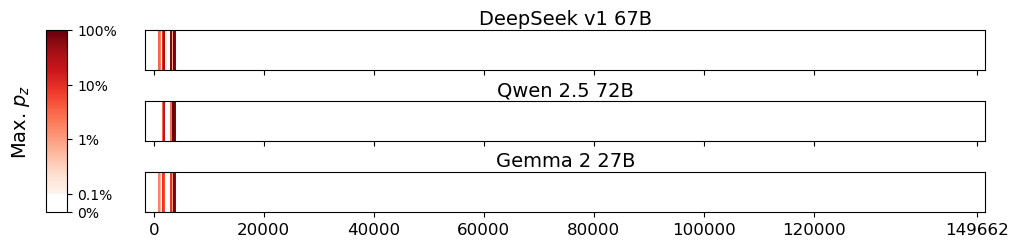}
    \includegraphics[width=\linewidth]{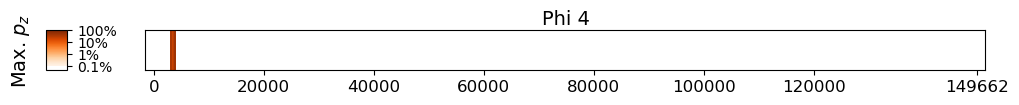}
  \end{minipage}
  \hfill
  \begin{minipage}[t]{0.45\textwidth}
    \centering
    \vspace{0cm}
    \includegraphics[width=\linewidth]{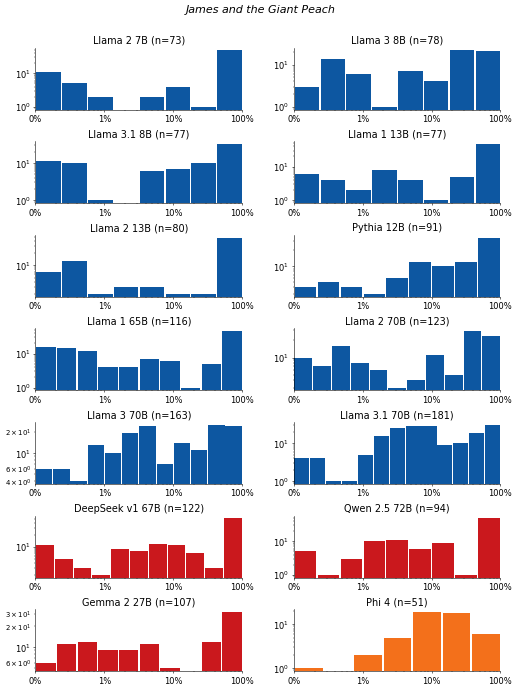}
  \end{minipage}
  \vspace{-.2cm}
  \caption{
    \textbf{\textit{James and the Giant Peach}, \citeauthor{James_and_the_Giant_Peach}.}
    For $14$ LLMs,
    (\textbf{left}) heatmaps for the sliding-window procedure and
    (\textbf{right}) corresponding distributions over suffix extraction probabilities
    ($\tau_\text{min}=0.1\%$).
  }
  \label{fig:slidingwindow:James_and_the_Giant_Peach}
\end{figure}
\FloatBarrier

\clearpage
\subsubsection{\textit{Automating the News}, \citeauthor{Automating_the_News}}\label{app:sec:sliding:Automating_the_News}
\vspace{-.2cm}
\begin{figure}[h]
  \centering
  \begin{minipage}[t]{0.53\textwidth}
    \centering
    \vspace{0cm}
    \includegraphics[width=\linewidth]{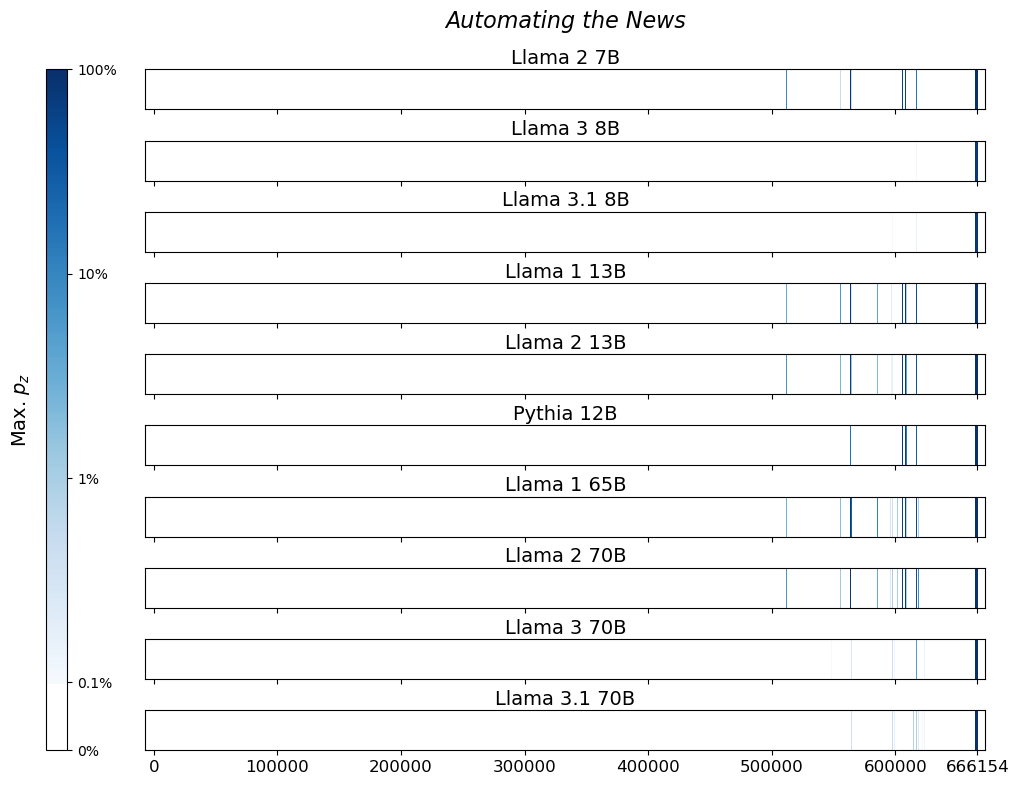}
    \includegraphics[width=\linewidth]{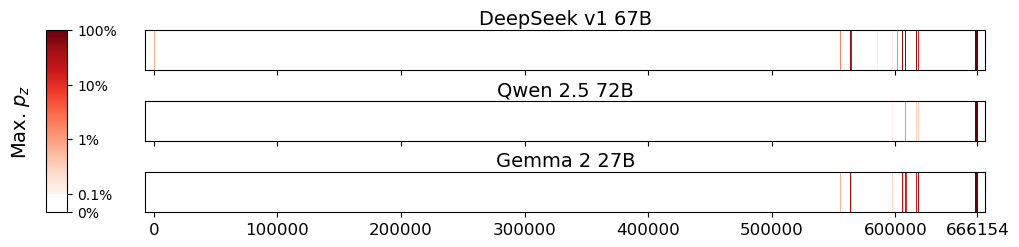}
    \includegraphics[width=\linewidth]{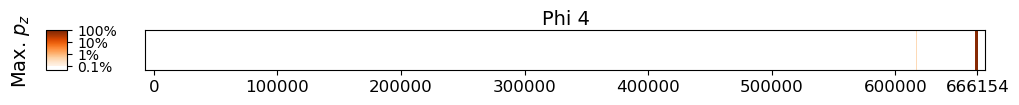}
  \end{minipage}
  \hfill
  \begin{minipage}[t]{0.45\textwidth}
    \centering
    \vspace{0cm}
    \includegraphics[width=\linewidth]{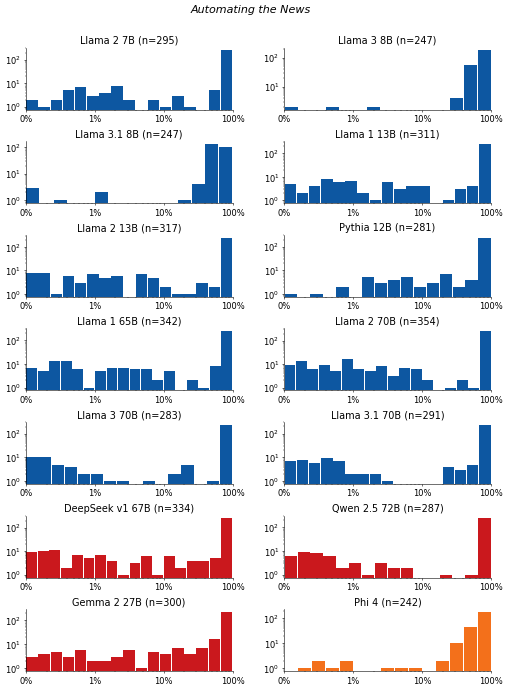}
  \end{minipage}
  \vspace{-.2cm}
  \caption{
    \textbf{\textit{Automating the News}, \citeauthor{Automating_the_News}.}
    For $14$ LLMs,
    (\textbf{left}) heatmaps for the sliding-window procedure and
    (\textbf{right}) corresponding distributions over suffix extraction probabilities
    ($\tau_\text{min}=0.1\%$).
  }
  \label{fig:slidingwindow:Automating_the_News}
\end{figure}
\FloatBarrier

\subsubsection{\textit{Drown}, \citeauthor{Drown}}\label{app:sec:sliding:Drown}
\vspace{-.2cm}
\begin{figure}[h]
  \centering
  \begin{minipage}[t]{0.53\textwidth}
    \centering
    \vspace{0cm}
    \includegraphics[width=\linewidth]{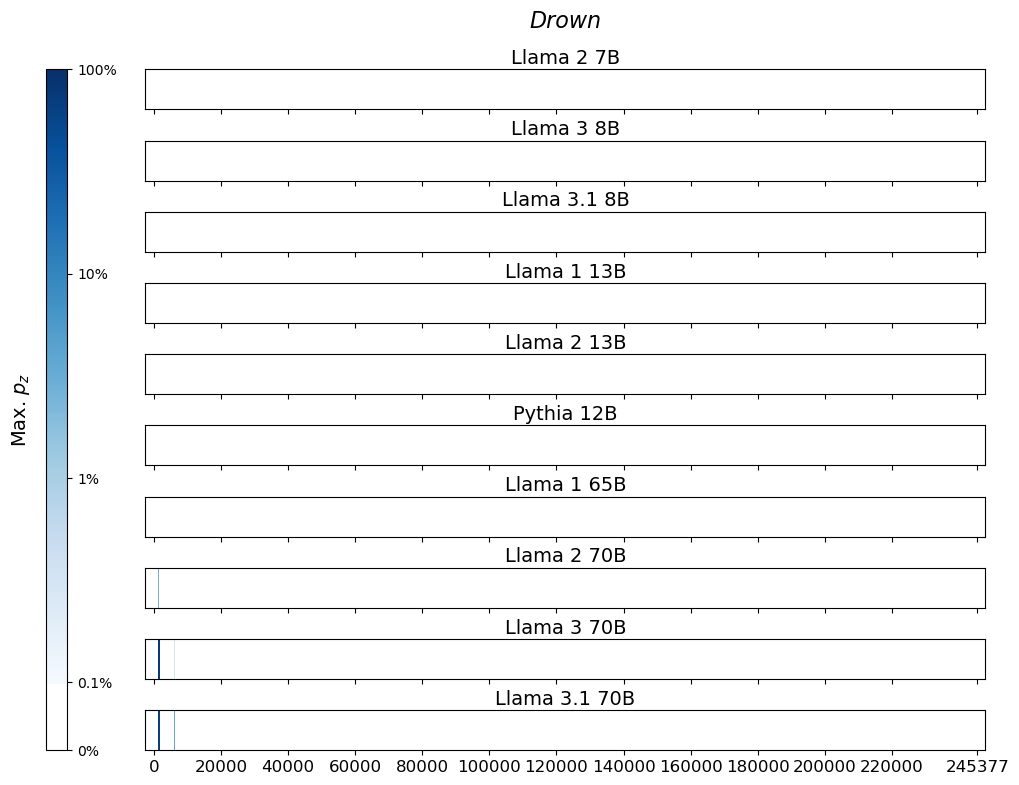}
    \includegraphics[width=\linewidth]{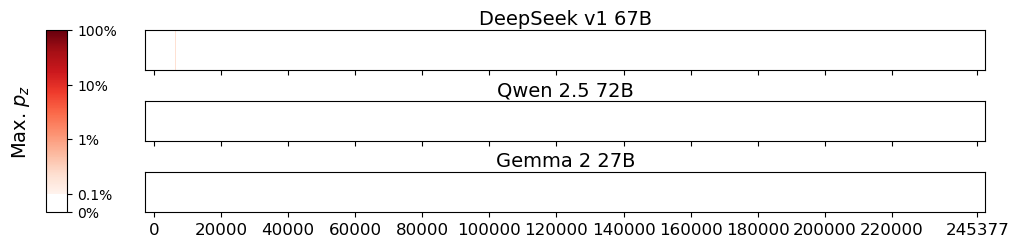}
    \includegraphics[width=\linewidth]{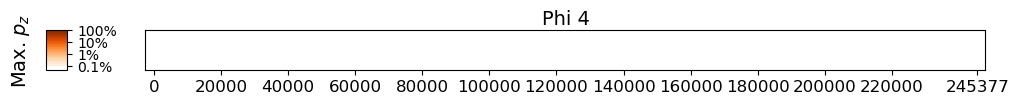}
  \end{minipage}
  \hfill
  \begin{minipage}[t]{0.45\textwidth}
    \centering
    \vspace{0cm}
    \includegraphics[width=\linewidth]{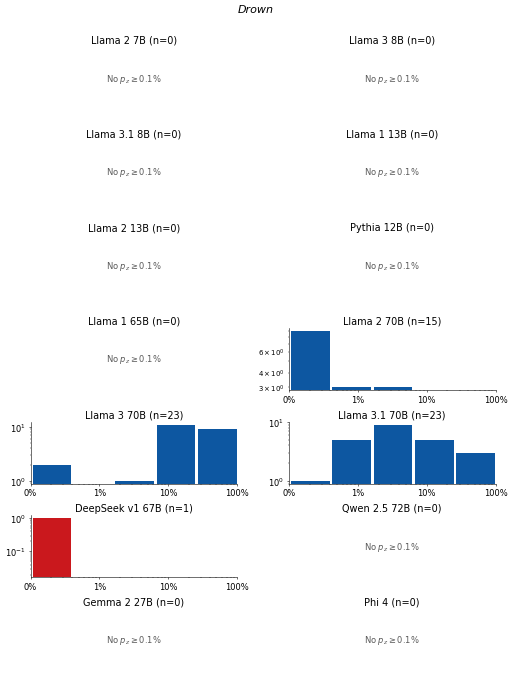}
  \end{minipage}
  \vspace{-.2cm}
  \caption{
    \textbf{\textit{Drown}, \citeauthor{Drown}.}
    For $14$ LLMs,
    (\textbf{left}) heatmaps for the sliding-window procedure and
    (\textbf{right}) corresponding distributions over suffix extraction probabilities
    ($\tau_\text{min}=0.1\%$).
  }
  \label{fig:slidingwindow:Drown}
\end{figure}
\FloatBarrier

\clearpage
\subsubsection{\textit{The Brief Wondrous Life of Oscar Wao}, \citeauthor{The_Brief_Wondrous_Life_of_Oscar_Wao}}\label{app:sec:sliding:The_Brief_Wondrous_Life_of_Oscar_Wao}
\vspace{-.2cm}
\begin{figure}[h]
  \centering
  \begin{minipage}[t]{0.53\textwidth}
    \centering
    \vspace{0cm}
    \includegraphics[width=\linewidth]{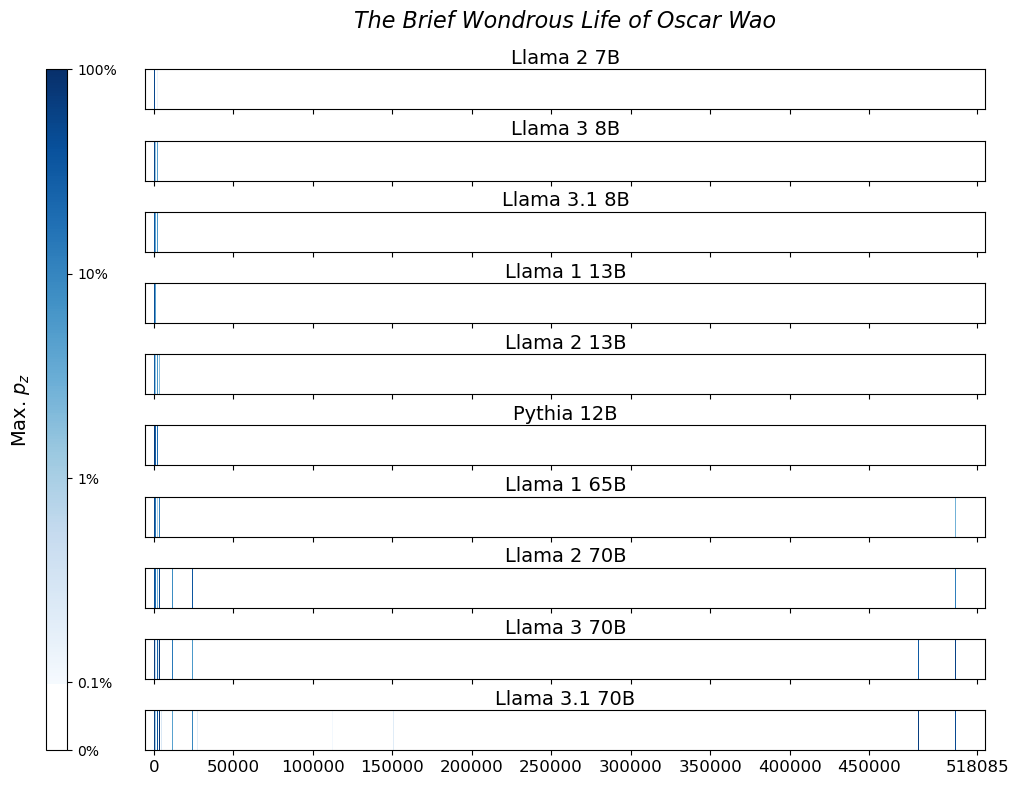}
    \includegraphics[width=\linewidth]{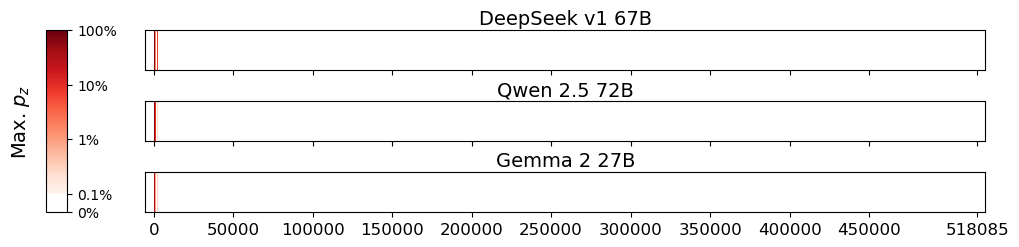}
    \includegraphics[width=\linewidth]{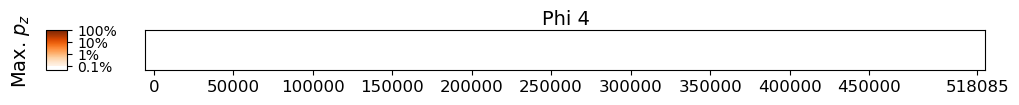}
  \end{minipage}
  \hfill
  \begin{minipage}[t]{0.45\textwidth}
    \centering
    \vspace{0cm}
    \includegraphics[width=\linewidth]{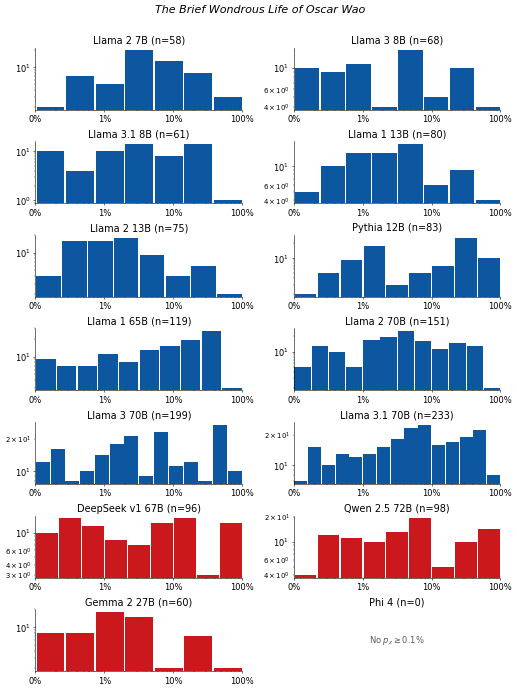}
  \end{minipage}
  \vspace{-.2cm}
  \caption{
    \textbf{\textit{The Brief Wondrous Life of Oscar Wao}, \citeauthor{The_Brief_Wondrous_Life_of_Oscar_Wao}.}
    For $14$ LLMs,
    (\textbf{left}) heatmaps for the sliding-window procedure and
    (\textbf{right}) corresponding distributions over suffix extraction probabilities
    ($\tau_\text{min}=0.1\%$).
  }
  \label{fig:slidingwindow:The_Brief_Wondrous_Life_of_Oscar_Wao}
\end{figure}
\FloatBarrier

\subsubsection{\textit{This Is How You Lose Her}, \citeauthor{This_Is_How_You_Lose_Her}}\label{app:sec:sliding:This_Is_How_You_Lose_Her}
\vspace{-.2cm}
\begin{figure}[h]
  \centering
  \begin{minipage}[t]{0.53\textwidth}
    \centering
    \vspace{0cm}
    \includegraphics[width=\linewidth]{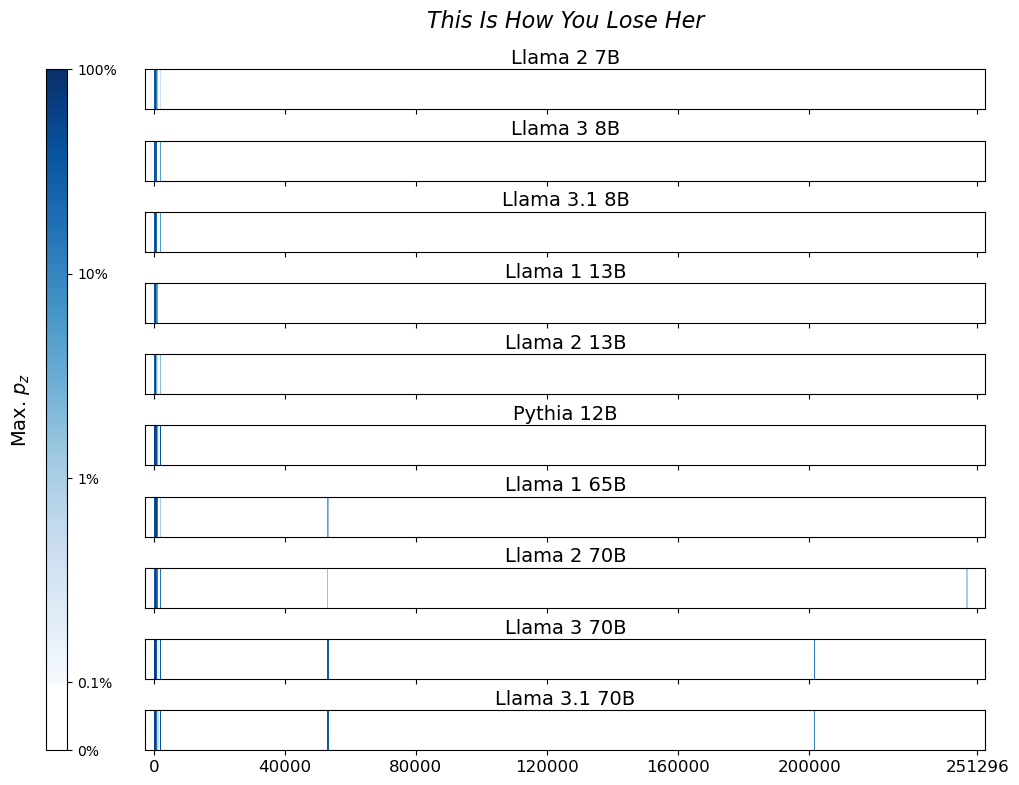}
    \includegraphics[width=\linewidth]{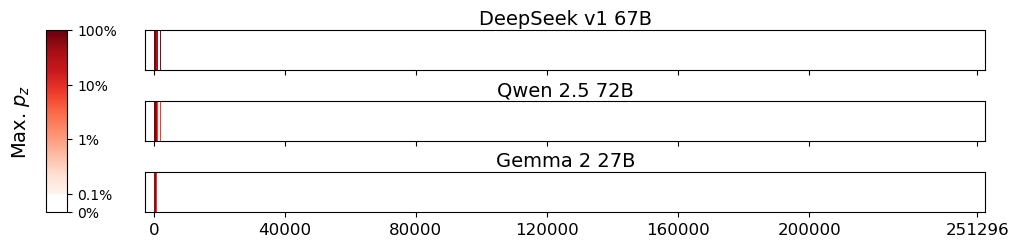}
    \includegraphics[width=\linewidth]{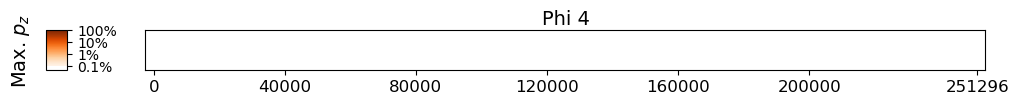}
  \end{minipage}
  \hfill
  \begin{minipage}[t]{0.45\textwidth}
    \centering
    \vspace{0cm}
    \includegraphics[width=\linewidth]{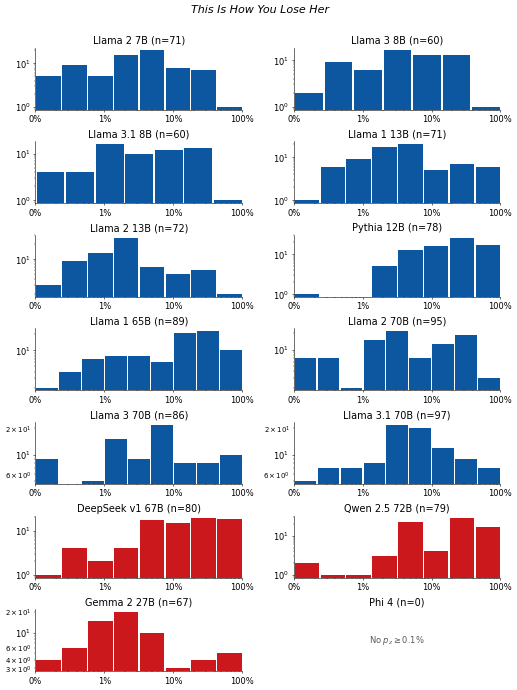}
  \end{minipage}
  \vspace{-.2cm}
  \caption{
    \textbf{\textit{This Is How You Lose Her}, \citeauthor{This_Is_How_You_Lose_Her}.}
    For $14$ LLMs,
    (\textbf{left}) heatmaps for the sliding-window procedure and
    (\textbf{right}) corresponding distributions over suffix extraction probabilities
    ($\tau_\text{min}=0.1\%$).
  }
  \label{fig:slidingwindow:This_Is_How_You_Lose_Her}
\end{figure}
\FloatBarrier

\clearpage
\subsubsection{\textit{The White Album}, \citeauthor{The_White_Album}}\label{app:sec:sliding:The_White_Album}
\vspace{-.2cm}
\begin{figure}[h]
  \centering
  \begin{minipage}[t]{0.53\textwidth}
    \centering
    \vspace{0cm}
    \includegraphics[width=\linewidth]{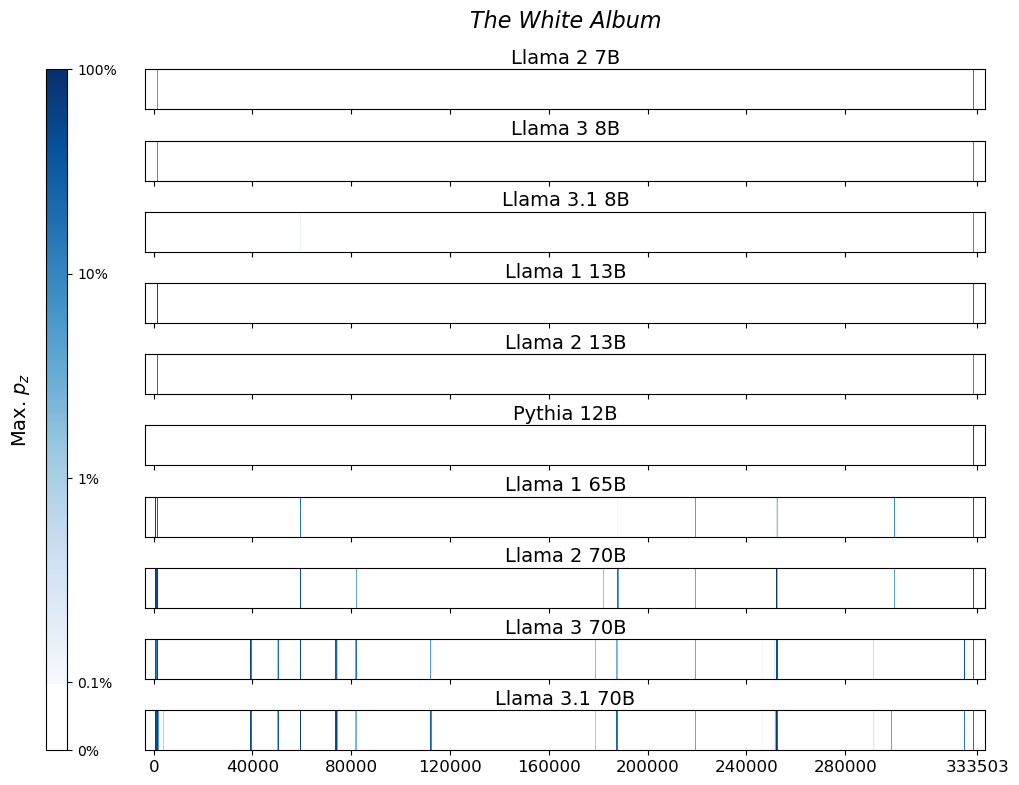}
    \includegraphics[width=\linewidth]{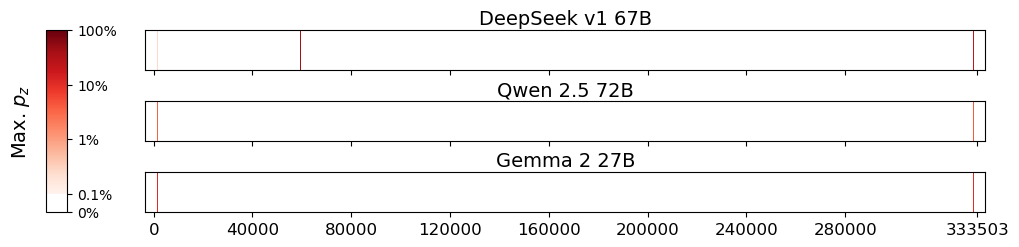}
    \includegraphics[width=\linewidth]{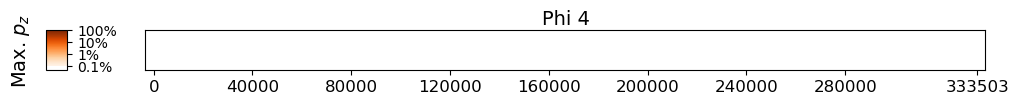}
  \end{minipage}
  \hfill
  \begin{minipage}[t]{0.45\textwidth}
    \centering
    \vspace{0cm}
    \includegraphics[width=\linewidth]{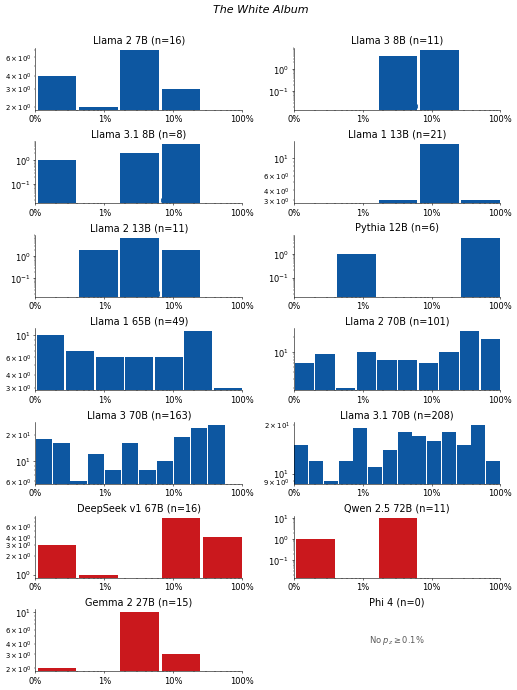}
  \end{minipage}
  \vspace{-.2cm}
  \caption{
    \textbf{\textit{The White Album}, \citeauthor{The_White_Album}.}
    For $14$ LLMs,
    (\textbf{left}) heatmaps for the sliding-window procedure and
    (\textbf{right}) corresponding distributions over suffix extraction probabilities
    ($\tau_\text{min}=0.1\%$).
  }
  \label{fig:slidingwindow:The_White_Album}
\end{figure}
\FloatBarrier

\subsubsection{\textit{Down and Out in the Magic Kingdom}, \citeauthor{Down_and_Out_in_the_Magic_Kingdom}}\label{app:sec:sliding:Down_and_Out_in_the_Magic_Kingdom}
\vspace{-.2cm}
\begin{figure}[h]
  \centering
  \begin{minipage}[t]{0.53\textwidth}
    \centering
    \vspace{0cm}
    \includegraphics[width=\linewidth]{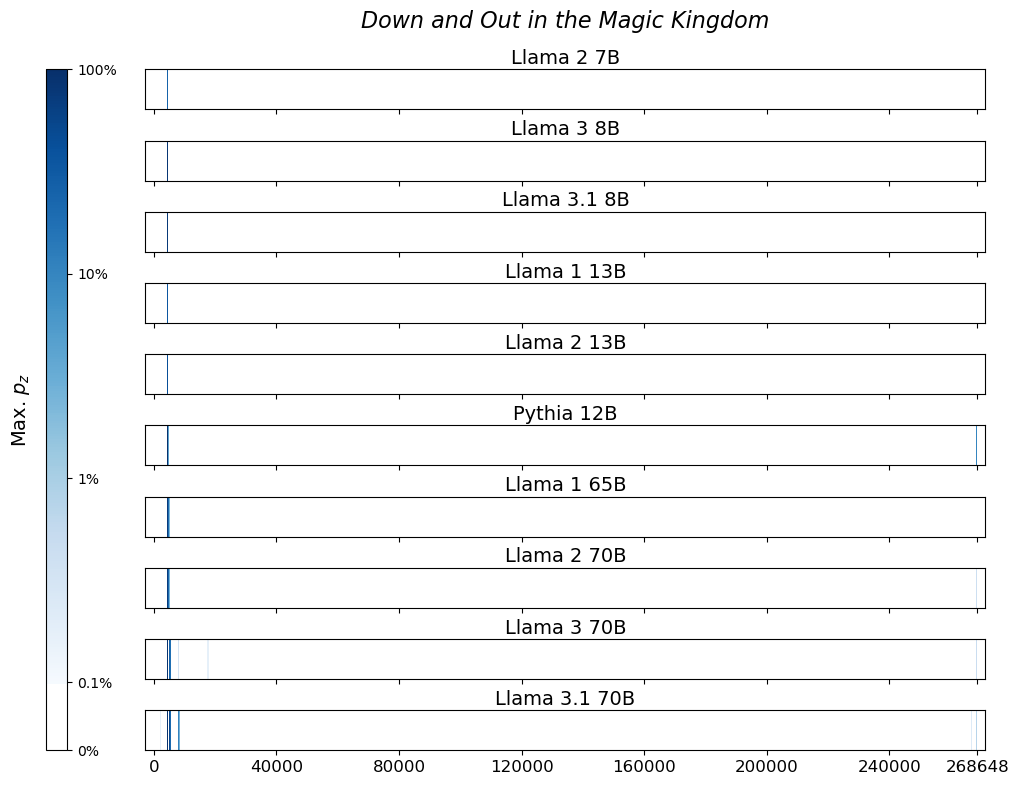}
    \includegraphics[width=\linewidth]{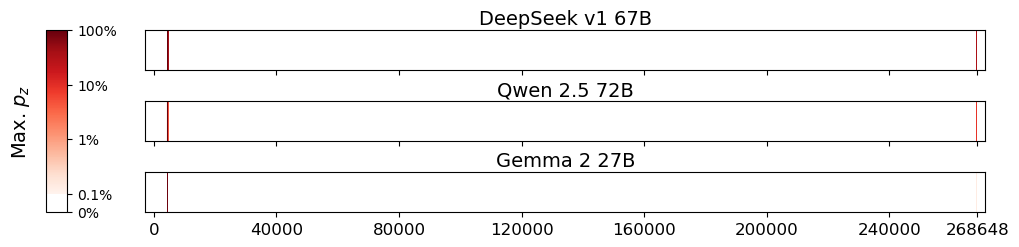}
    \includegraphics[width=\linewidth]{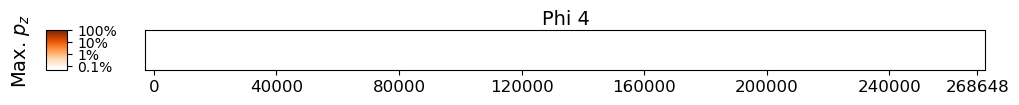}
  \end{minipage}
  \hfill
  \begin{minipage}[t]{0.45\textwidth}
    \centering
    \vspace{0cm}
    \includegraphics[width=\linewidth]{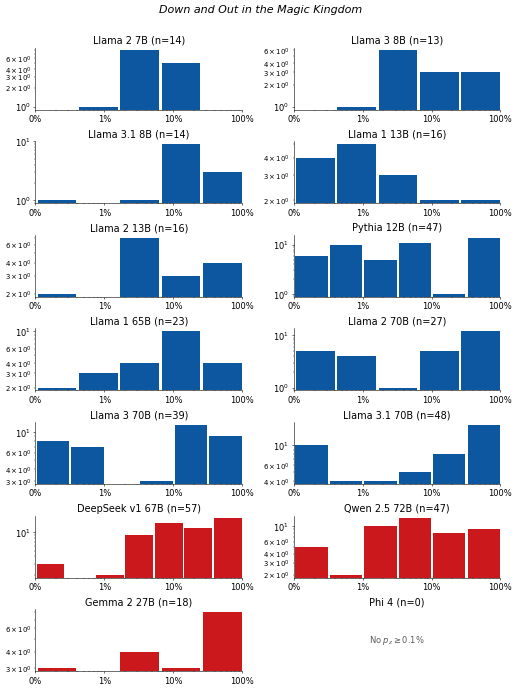}
  \end{minipage}
  \vspace{-.2cm}
  \caption{
    \textbf{\textit{Down and Out in the Magic Kingdom}, \citeauthor{Down_and_Out_in_the_Magic_Kingdom}.}
    For $14$ LLMs,
    (\textbf{left}) heatmaps for the sliding-window procedure and
    (\textbf{right}) corresponding distributions over suffix extraction probabilities
    ($\tau_\text{min}=0.1\%$).
  }
  \label{fig:slidingwindow:Down_and_Out_in_the_Magic_Kingdom}
\end{figure}
\FloatBarrier

\clearpage
\subsubsection{\textit{The World's Wife}, \citeauthor{The_World_s_Wife}}\label{app:sec:sliding:The_World_s_Wife}
\vspace{-.2cm}
\begin{figure}[h]
  \centering
  \begin{minipage}[t]{0.53\textwidth}
    \centering
    \vspace{0cm}
    \includegraphics[width=\linewidth]{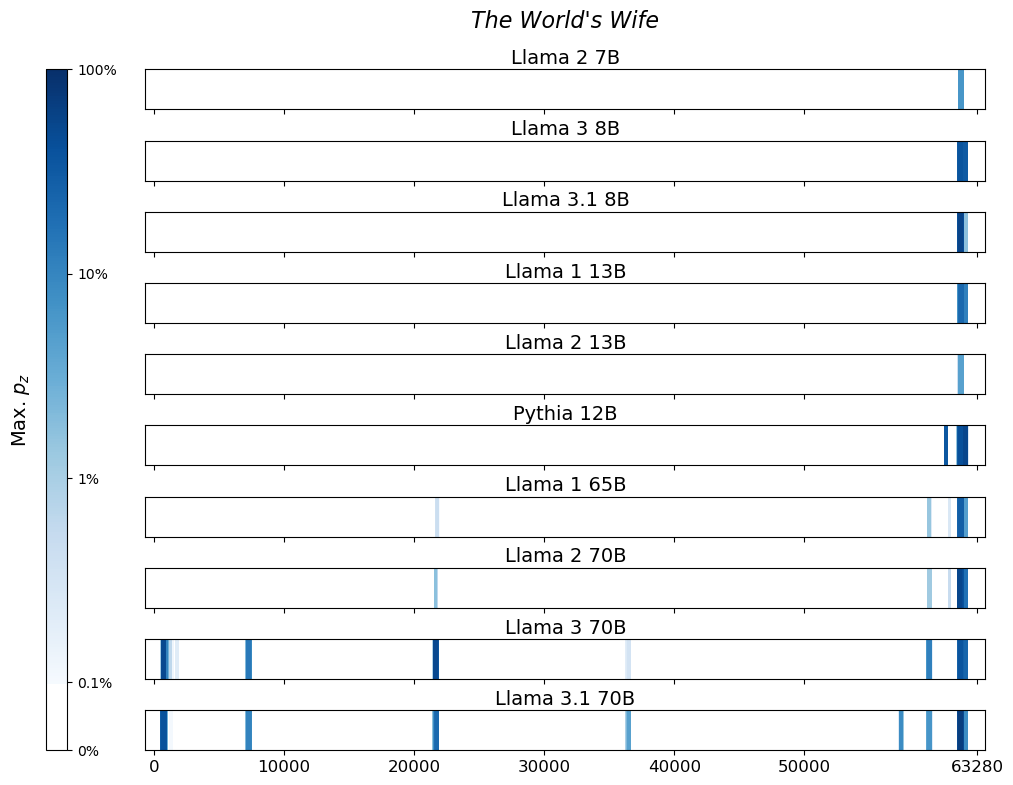}
    \includegraphics[width=\linewidth]{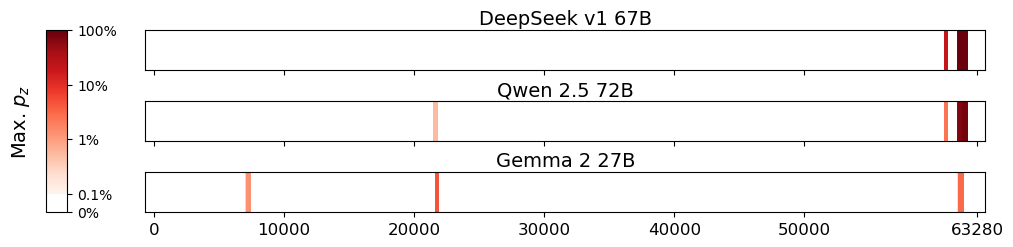}
    \includegraphics[width=\linewidth]{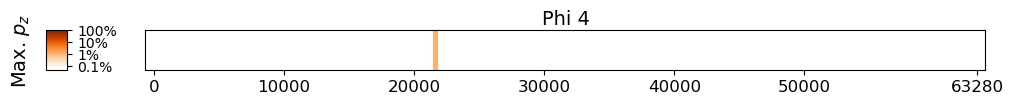}
  \end{minipage}
  \hfill
  \begin{minipage}[t]{0.45\textwidth}
    \centering
    \vspace{0cm}
    \includegraphics[width=\linewidth]{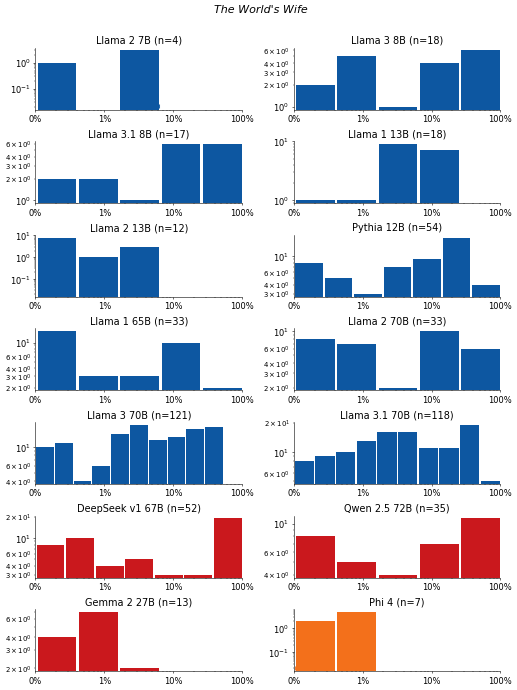}
  \end{minipage}
  \vspace{-.2cm}
  \caption{
    \textbf{\textit{The World's Wife}, \citeauthor{The_World_s_Wife}.}
    For $14$ LLMs,
    (\textbf{left}) heatmaps for the sliding-window procedure and
    (\textbf{right}) corresponding distributions over suffix extraction probabilities
    ($\tau_\text{min}=0.1\%$).
  }
  \label{fig:slidingwindow:The_World_s_Wife}
\end{figure}
\FloatBarrier

\subsubsection{\textit{A Visit from the Goon Squad}, \citeauthor{A_Visit_from_the_Goon_Squad}}\label{app:sec:sliding:A_Visit_from_the_Goon_Squad}
\vspace{-.2cm}
\begin{figure}[h]
  \centering
  \begin{minipage}[t]{0.53\textwidth}
    \centering
    \vspace{0cm}
    \includegraphics[width=\linewidth]{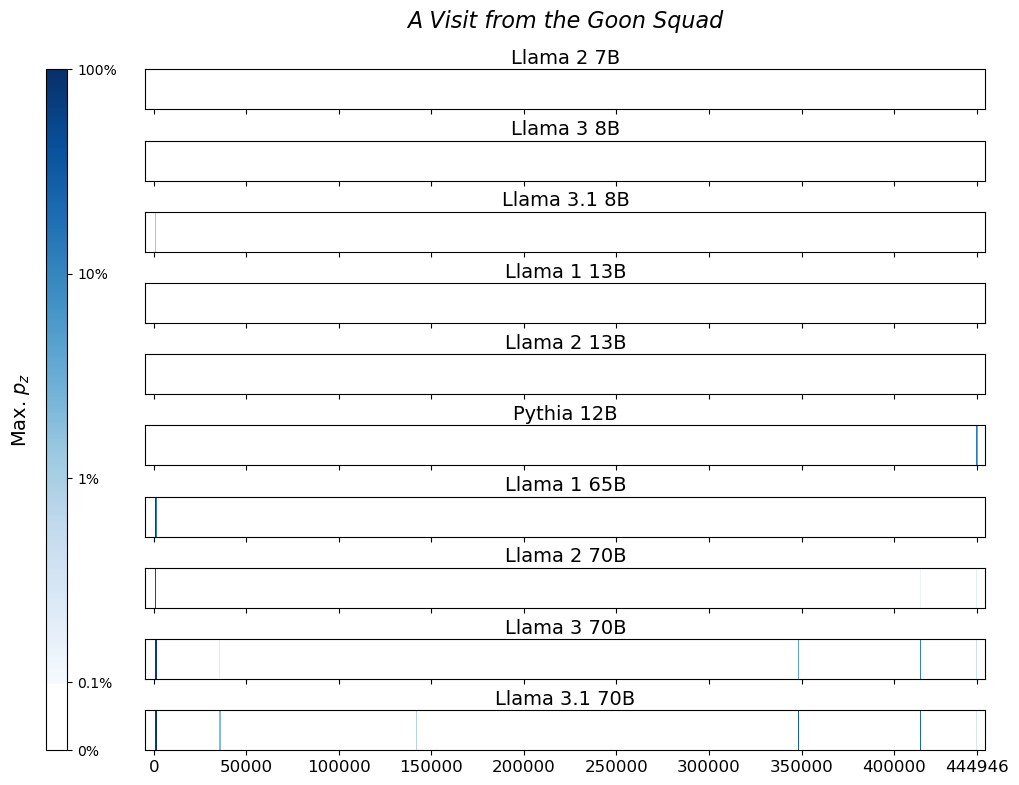}
    \includegraphics[width=\linewidth]{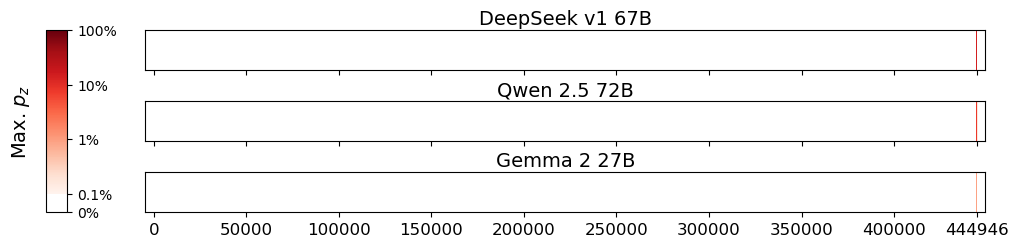}
    \includegraphics[width=\linewidth]{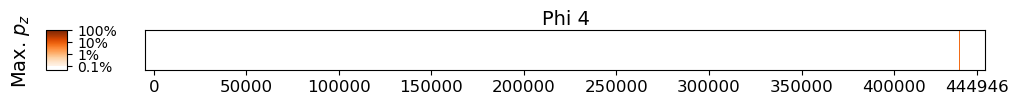}
  \end{minipage}
  \hfill
  \begin{minipage}[t]{0.45\textwidth}
    \centering
    \vspace{0cm}
    \includegraphics[width=\linewidth]{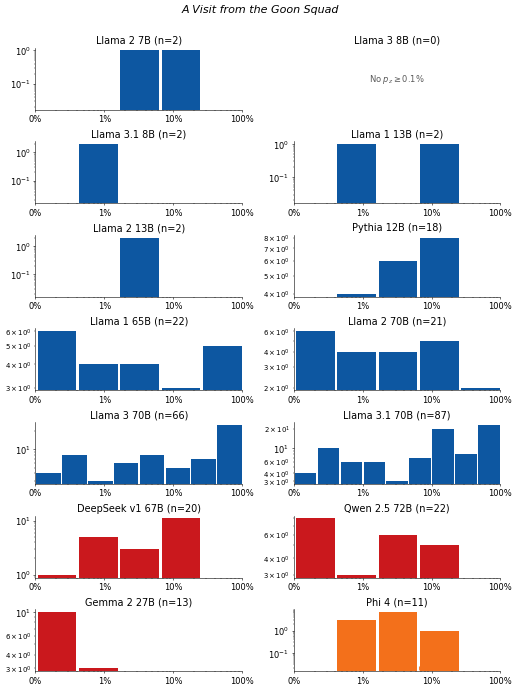}
  \end{minipage}
  \vspace{-.2cm}
  \caption{
    \textbf{\textit{A Visit from the Goon Squad}, \citeauthor{A_Visit_from_the_Goon_Squad}.}
    For $14$ LLMs,
    (\textbf{left}) heatmaps for the sliding-window procedure and
    (\textbf{right}) corresponding distributions over suffix extraction probabilities
    ($\tau_\text{min}=0.1\%$).
  }
  \label{fig:slidingwindow:A_Visit_from_the_Goon_Squad}
\end{figure}
\FloatBarrier

\clearpage
\subsubsection{\textit{Invisible Man}, \citeauthor{Invisible_Man}}\label{app:sec:sliding:Invisible_Man}
\vspace{-.2cm}
\begin{figure}[h]
  \centering
  \begin{minipage}[t]{0.53\textwidth}
    \centering
    \vspace{0cm}
    \includegraphics[width=\linewidth]{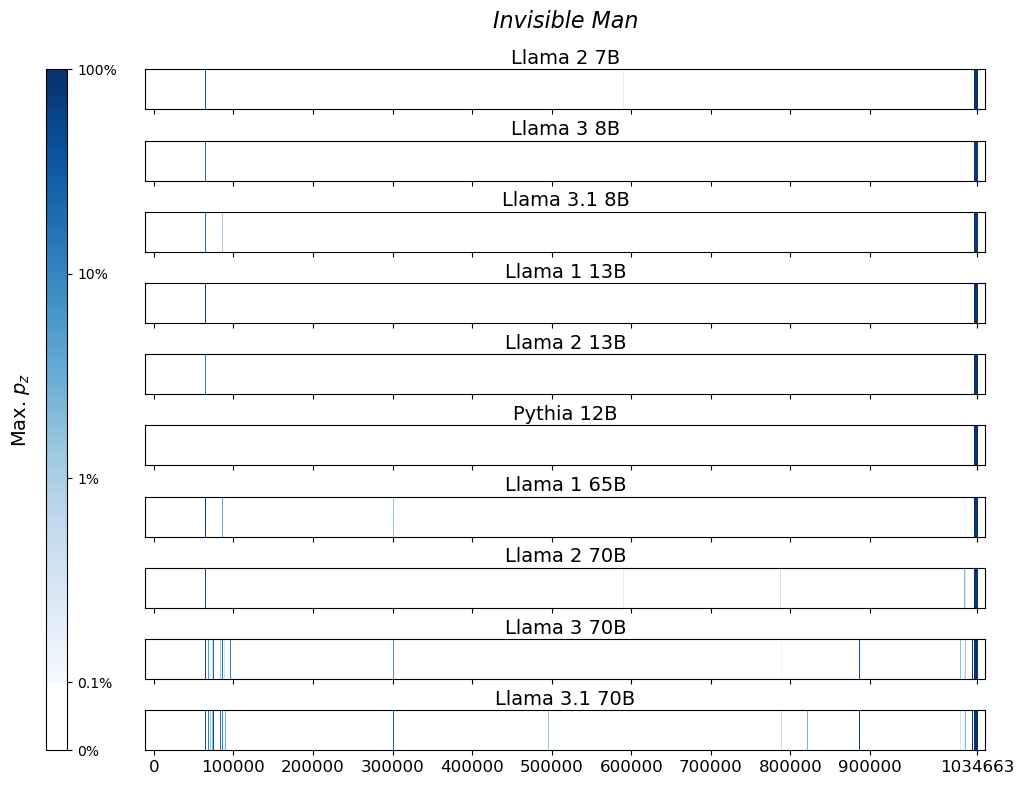}
    \includegraphics[width=\linewidth]{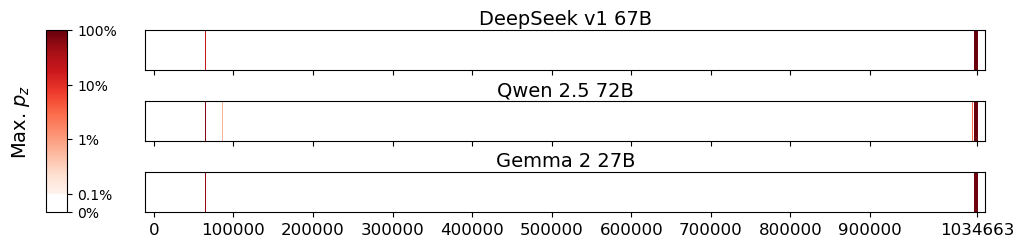}
    \includegraphics[width=\linewidth]{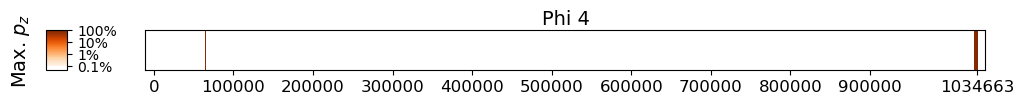}
  \end{minipage}
  \hfill
  \begin{minipage}[t]{0.45\textwidth}
    \centering
    \vspace{0cm}
    \includegraphics[width=\linewidth]{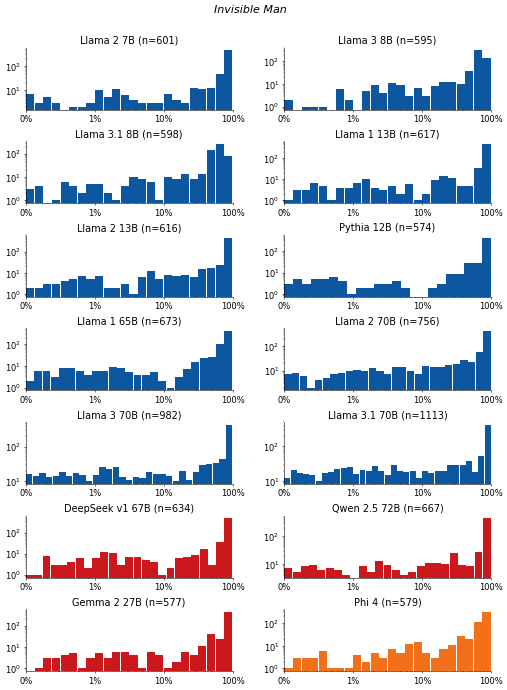}
  \end{minipage}
  \vspace{-.2cm}
  \caption{
    \textbf{\textit{Invisible Man}, \citeauthor{Invisible_Man}.}
    For $14$ LLMs,
    (\textbf{left}) heatmaps for the sliding-window procedure and
    (\textbf{right}) corresponding distributions over suffix extraction probabilities
    ($\tau_\text{min}=0.1\%$).
  }
  \label{fig:slidingwindow:Invisible_Man}
\end{figure}
\FloatBarrier

\subsubsection{\textit{We Should All Be Mirandas}, \citeauthor{We_Should_All_Be_Mirandas}}\label{app:sec:sliding:We_Should_All_Be_Mirandas}
\begin{figure}[h]
  \vspace{-.2cm}
  \centering
  \begin{minipage}[t]{0.53\textwidth}
    \centering
    \vspace{0cm}
    \includegraphics[width=\linewidth]{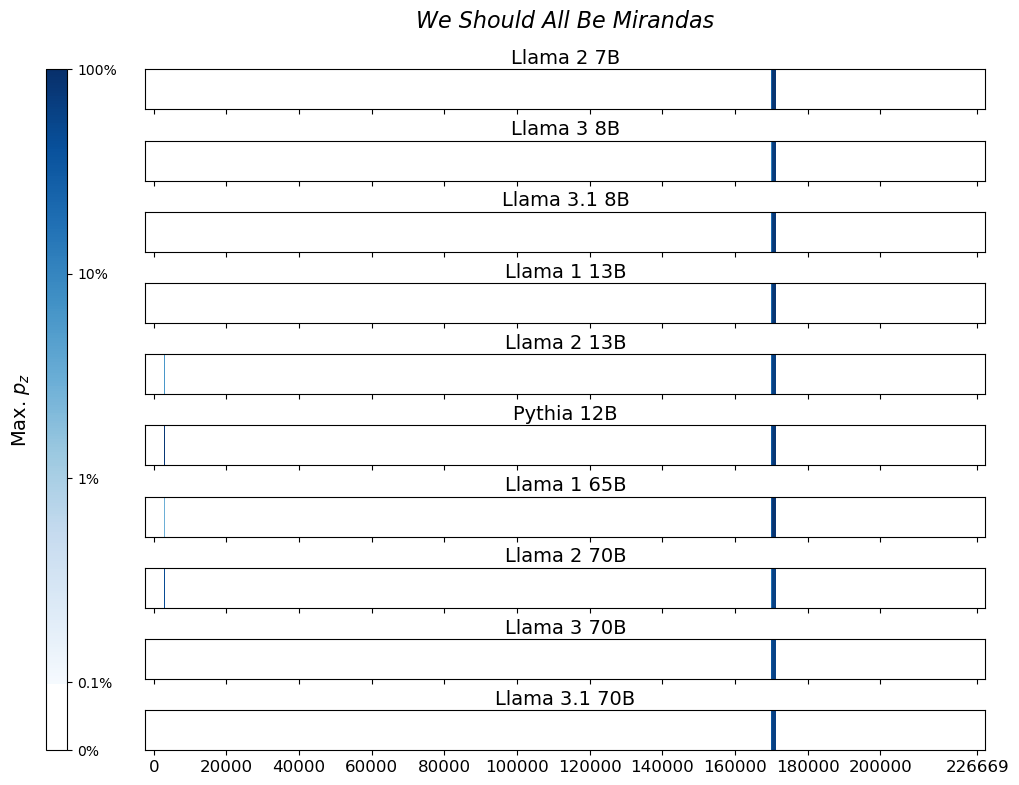}
    \includegraphics[width=\linewidth]{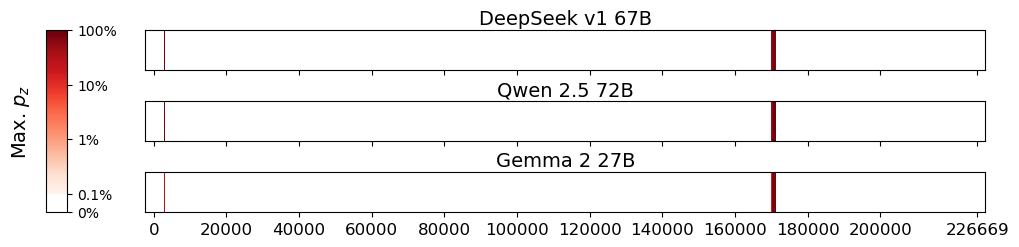}
    \includegraphics[width=\linewidth]{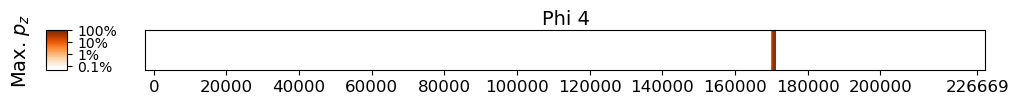}
  \end{minipage}
  \hfill
  \begin{minipage}[t]{0.45\textwidth}
    \centering
    \vspace{0cm}
    \includegraphics[width=\linewidth]{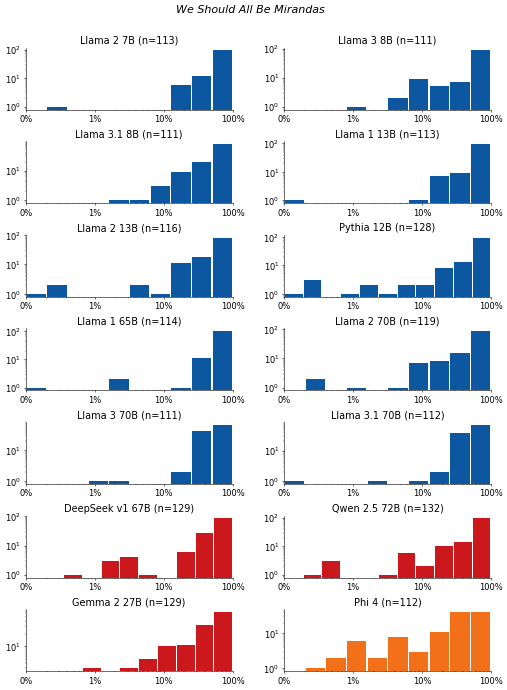}
  \end{minipage}
  \vspace{-.2cm}
  \caption{
    \textbf{\textit{We Should All Be Mirandas}, \citeauthor{We_Should_All_Be_Mirandas}.}
    For $14$ LLMs,
    (\textbf{left}) heatmaps for the sliding-window procedure and
    (\textbf{right}) corresponding distributions over suffix extraction probabilities
    ($\tau_\text{min}=0.1\%$).
  }
  \label{fig:slidingwindow:We_Should_All_Be_Mirandas}
\end{figure}
\FloatBarrier

\clearpage
\subsubsection{\textit{The Dude Abides}, \citeauthor{The_Dude_Abides}}\label{app:sec:sliding:The_Dude_Abides}
\vspace{-.2cm}
\begin{figure}[h]
  \centering
  \begin{minipage}[t]{0.53\textwidth}
    \centering
    \vspace{0cm}
    \includegraphics[width=\linewidth]{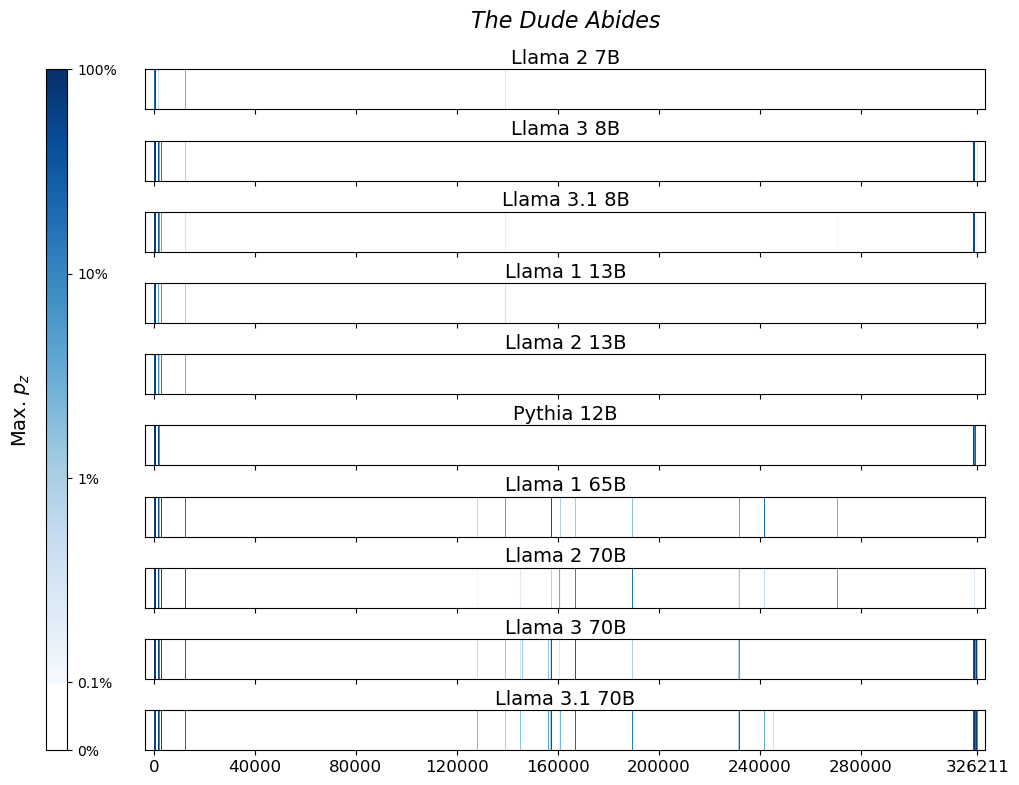}
    \includegraphics[width=\linewidth]{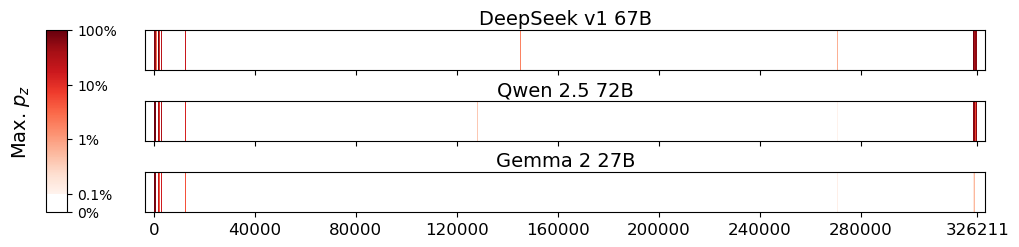}
    \includegraphics[width=\linewidth]{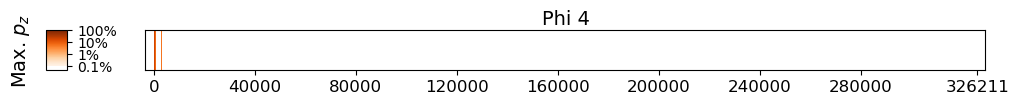}
  \end{minipage}
  \hfill
  \begin{minipage}[t]{0.45\textwidth}
    \centering
    \vspace{0cm}
    \includegraphics[width=\linewidth]{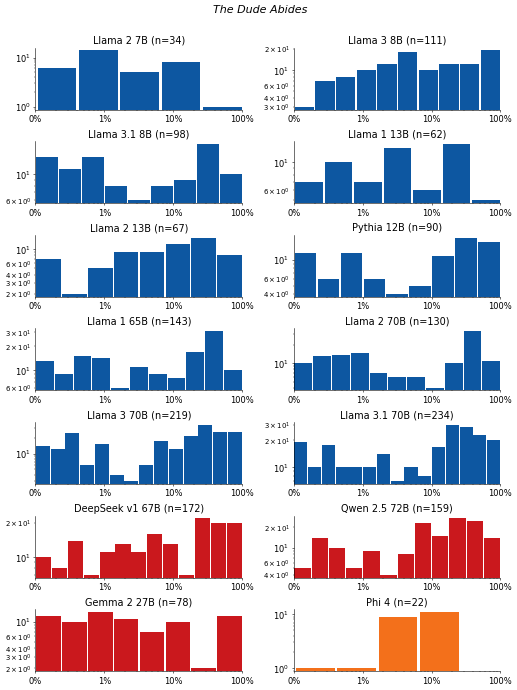}
  \end{minipage}
  \vspace{-.2cm}
  \caption{
    \textbf{\textit{The Dude Abides}, \citeauthor{The_Dude_Abides}.}
    For $14$ LLMs,
    (\textbf{left}) heatmaps for the sliding-window procedure and
    (\textbf{right}) corresponding distributions over suffix extraction probabilities
    ($\tau_\text{min}=0.1\%$).
  }
  \label{fig:slidingwindow:The_Dude_Abides}
\end{figure}
\FloatBarrier

\subsubsection{\textit{The President's Vampire}, \citeauthor{The_President_s_Vampire}}\label{app:sec:sliding:The_President_s_Vampire}
\begin{figure}[h]
  \vspace{-.2cm}
  \centering
  \begin{minipage}[t]{0.53\textwidth}
    \centering
    \vspace{0cm}
    \includegraphics[width=\linewidth]{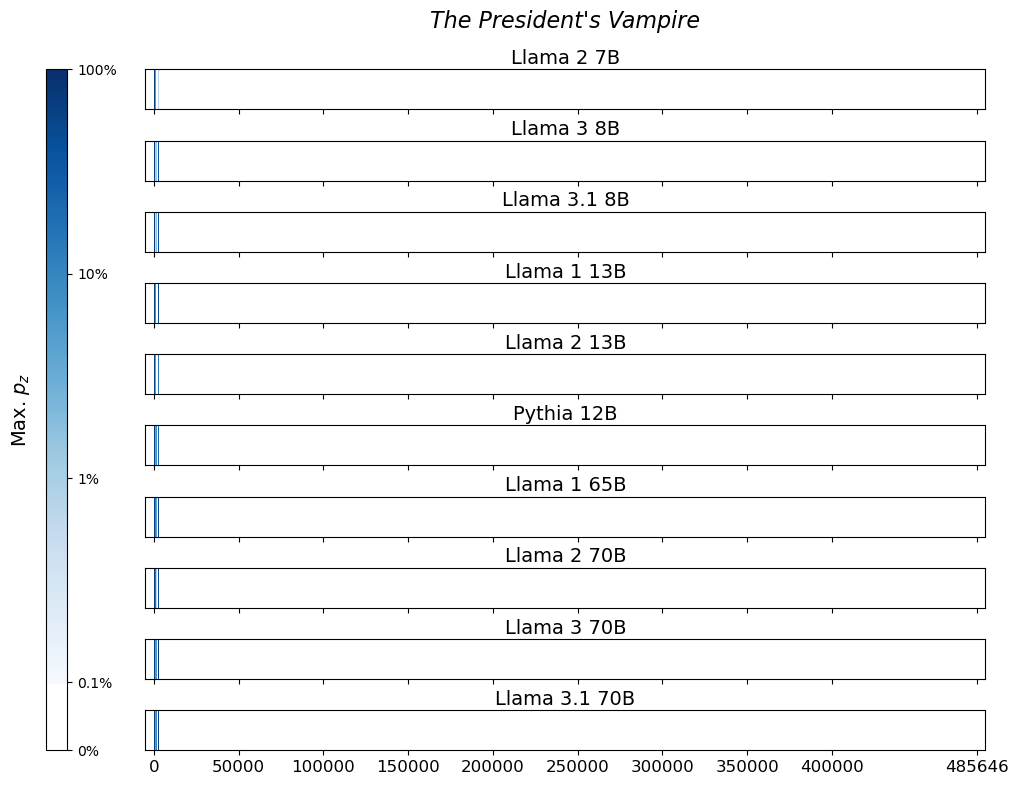}
    \includegraphics[width=\linewidth]{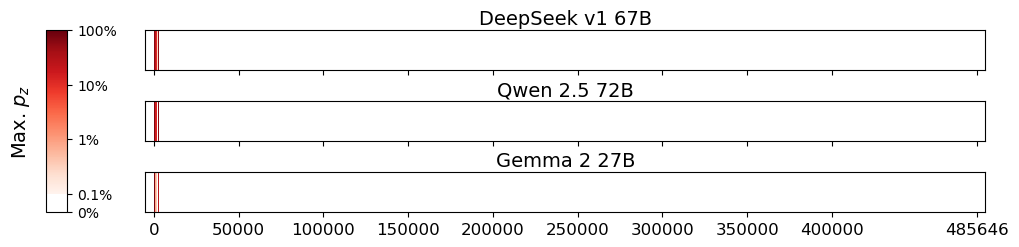}
    \includegraphics[width=\linewidth]{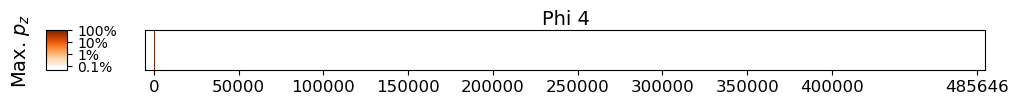}
  \end{minipage}
  \hfill
  \begin{minipage}[t]{0.45\textwidth}
    \centering
    \vspace{0cm}
    \includegraphics[width=\linewidth]{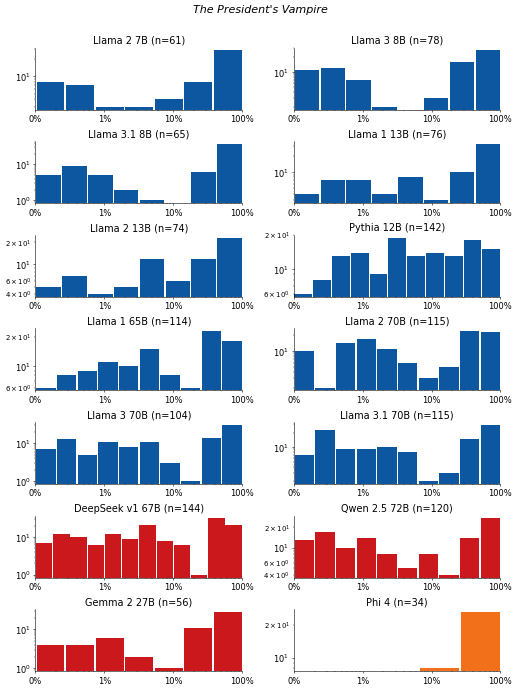}
  \end{minipage}
  \vspace{-.2cm}
  \caption{
    \textbf{\textit{The President's Vampire}, \citeauthor{The_President_s_Vampire}.}
    For $14$ LLMs,
    (\textbf{left}) heatmaps for the sliding-window procedure and
    (\textbf{right}) corresponding distributions over suffix extraction probabilities
    ($\tau_\text{min}=0.1\%$).
  }
  \label{fig:slidingwindow:The_President_s_Vampire}
\end{figure}
\FloatBarrier

\clearpage
\subsubsection{\textit{The Great Gatsby}, \citeauthor{The_Great_Gatsby}}\label{app:sec:sliding:The_Great_Gatsby}
\vspace{-.2cm}
\begin{figure}[h]
  \centering
  \begin{minipage}[t]{0.53\textwidth}
    \centering
    \vspace{0cm}
    \includegraphics[width=\linewidth]{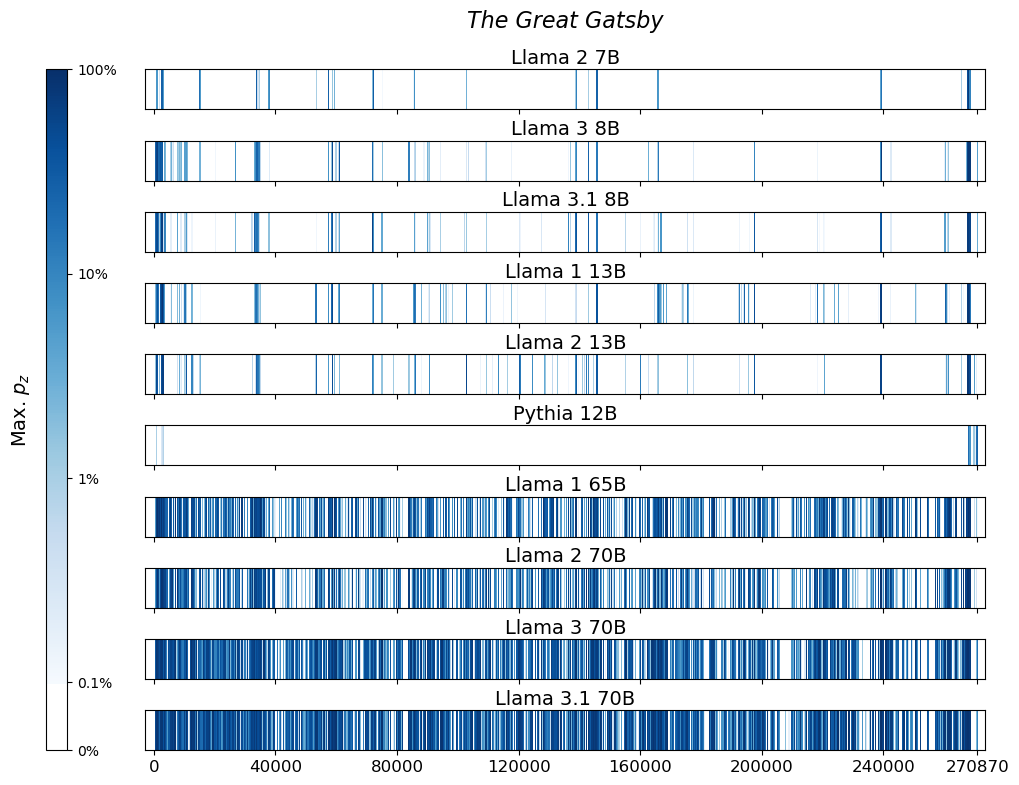}
    \includegraphics[width=\linewidth]{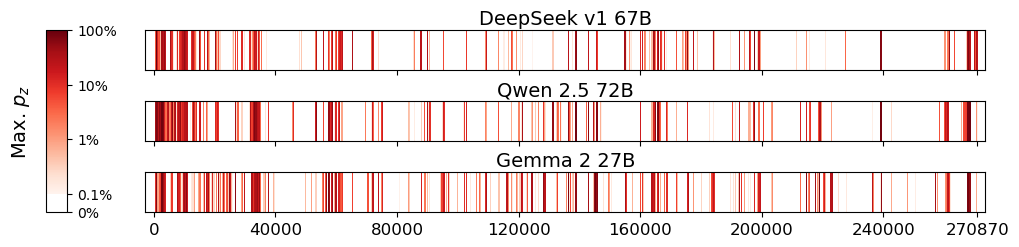}
    \includegraphics[width=\linewidth]{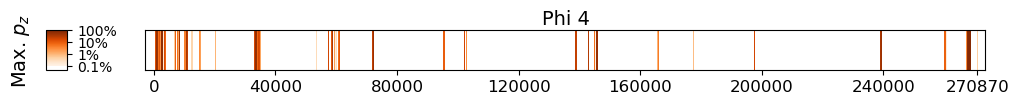}
  \end{minipage}
  \hfill
  \begin{minipage}[t]{0.45\textwidth}
    \centering
    \vspace{0cm}
    \includegraphics[width=\linewidth]{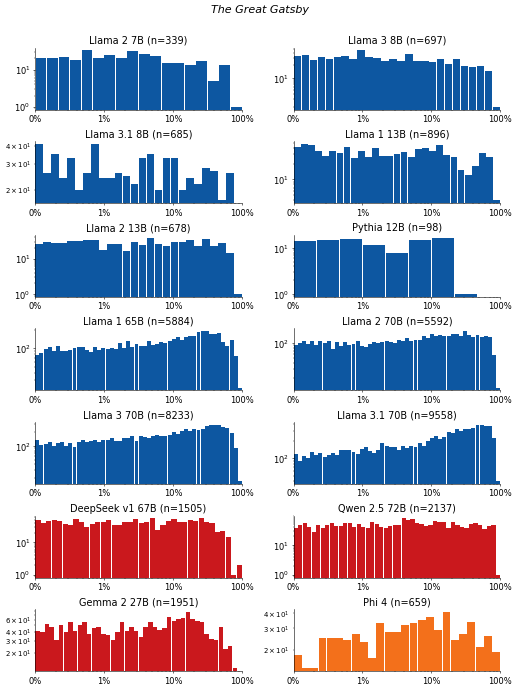}
  \end{minipage}
  \vspace{-.2cm}
  \caption{
    \textbf{\textit{The Great Gatsby}, \citeauthor{The_Great_Gatsby}.}
    For $14$ LLMs,
    (\textbf{left}) heatmaps for the sliding-window procedure and
    (\textbf{right}) corresponding distributions over suffix extraction probabilities
    ($\tau_\text{min}=0.1\%$).
  }
  \label{fig:slidingwindow:The_Great_Gatsby}
\end{figure}
\FloatBarrier

\subsubsection{\textit{Gone Girl}, \citeauthor{Gone_Girl}}\label{app:sec:sliding:Gone_Girl}
\vspace{-.2cm}
\begin{figure}[h]
  \centering
  \begin{minipage}[t]{0.53\textwidth}
    \centering
    \vspace{0cm}
    \includegraphics[width=\linewidth]{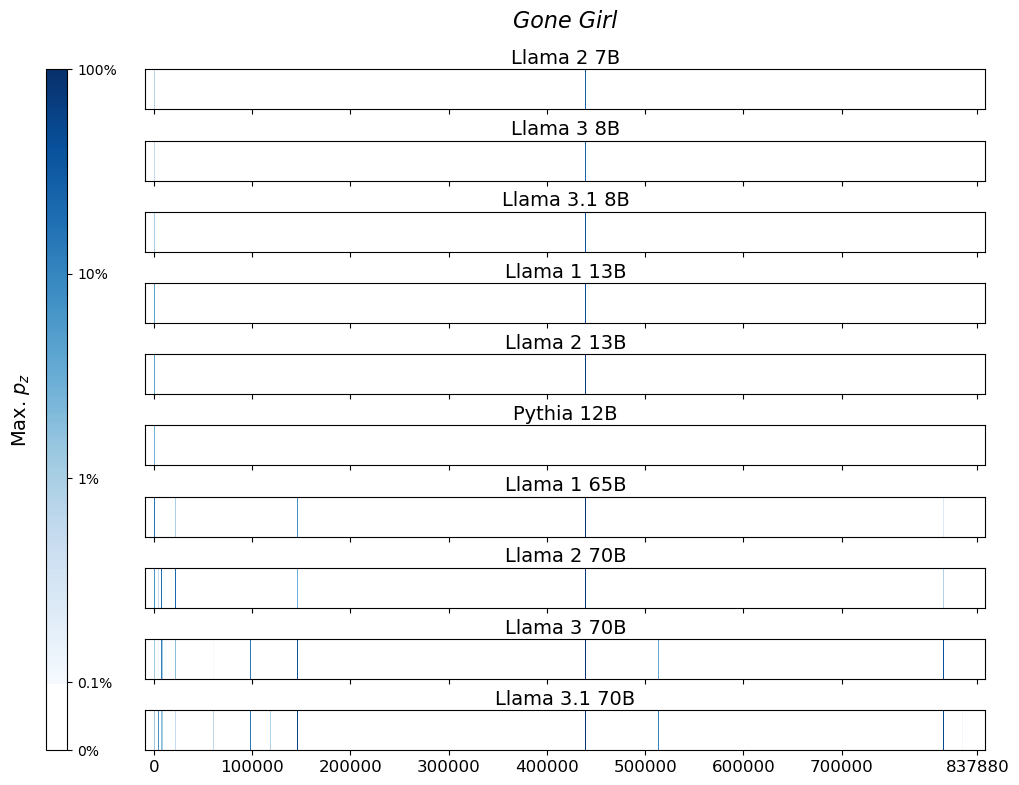}
    \includegraphics[width=\linewidth]{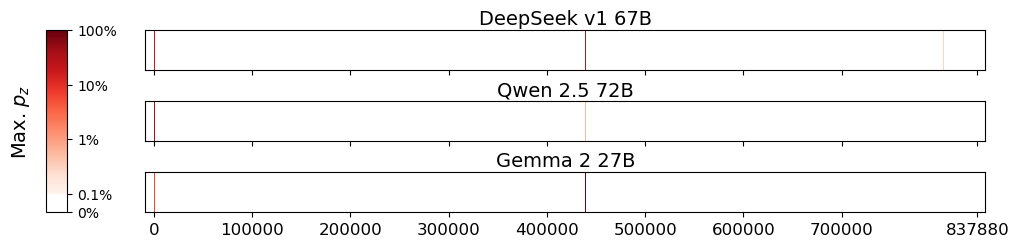}
    \includegraphics[width=\linewidth]{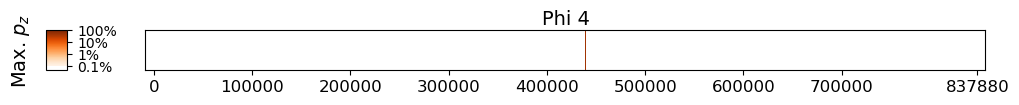}
  \end{minipage}
  \hfill
  \begin{minipage}[t]{0.45\textwidth}
    \centering
    \vspace{0cm}
    \includegraphics[width=\linewidth]{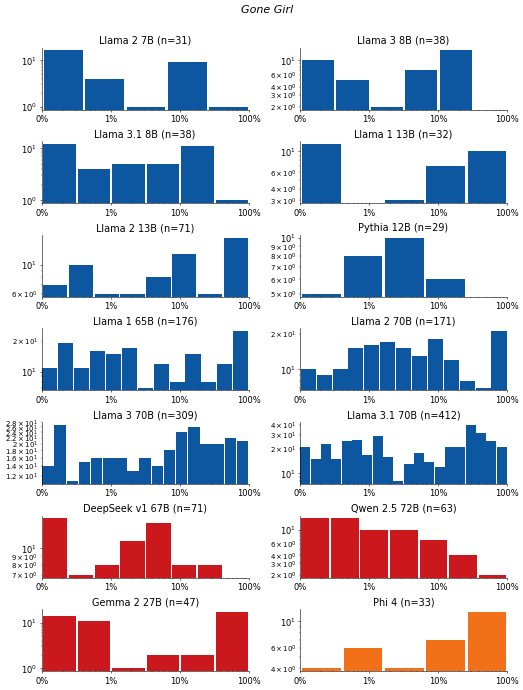}
  \end{minipage}
  \vspace{-.2cm}
  \caption{
    \textbf{\textit{Gone Girl}, \citeauthor{Gone_Girl}.}
    For $14$ LLMs,
    (\textbf{left}) heatmaps for the sliding-window procedure and
    (\textbf{right}) corresponding distributions over suffix extraction probabilities
    ($\tau_\text{min}=0.1\%$).
  }
  \label{fig:slidingwindow:Gone_Girl}
\end{figure}
\FloatBarrier

\clearpage
\subsubsection{\textit{British Destroyers}, \citeauthor{British_Destroyers}}\label{app:sec:sliding:British_Destroyers}
\vspace{-.2cm}
\begin{figure}[h]
  \centering
  \begin{minipage}[t]{0.53\textwidth}
    \centering
    \vspace{0cm}
    \includegraphics[width=\linewidth]{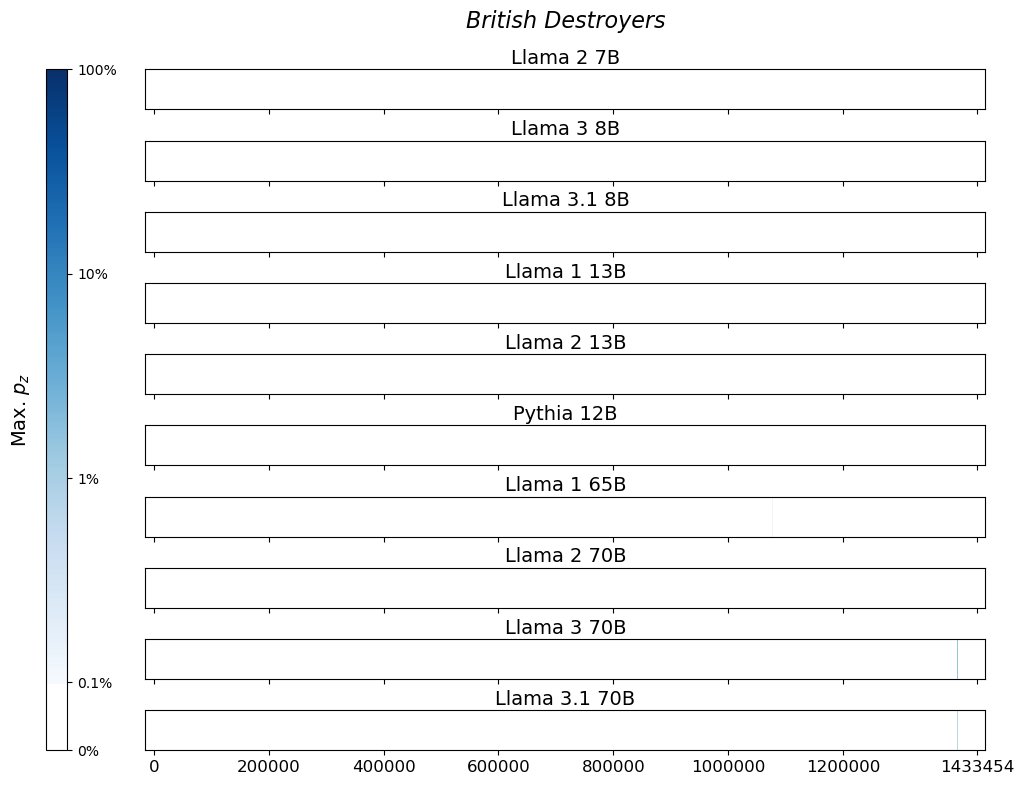}
    \includegraphics[width=\linewidth]{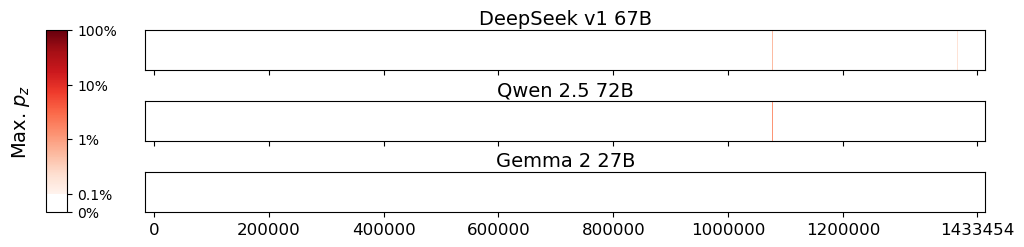}
    \includegraphics[width=\linewidth]{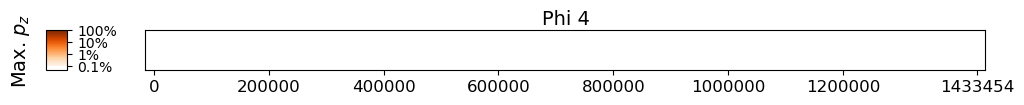}
  \end{minipage}
  \hfill
  \begin{minipage}[t]{0.45\textwidth}
    \centering
    \vspace{0cm}
    \includegraphics[width=\linewidth]{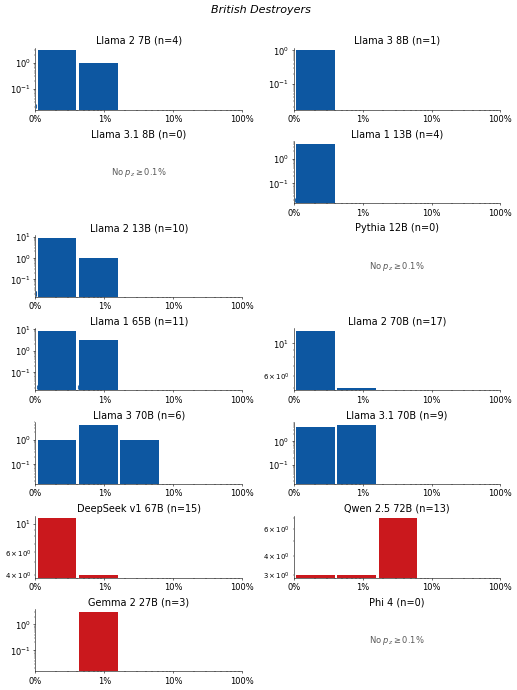}
  \end{minipage}
  \vspace{-.2cm}
  \caption{
    \textbf{\textit{British Destroyers}, \citeauthor{British_Destroyers}.}
    For $14$ LLMs,
    (\textbf{left}) heatmaps for the sliding-window procedure and
    (\textbf{right}) corresponding distributions over suffix extraction probabilities
    ($\tau_\text{min}=0.1\%$).
  }
  \label{fig:slidingwindow:British_Destroyers}
\end{figure}
\FloatBarrier

\subsubsection{\textit{Florals \& Botanicals}, \citeauthor{Florals_and_Botanicals}}\label{app:sec:sliding:Florals_and_Botanicals}
\begin{figure}[h]
  \vspace{-.2cm}
  \centering
  \begin{minipage}[t]{0.53\textwidth}
    \centering
    \vspace{0cm}
    \includegraphics[width=\linewidth]{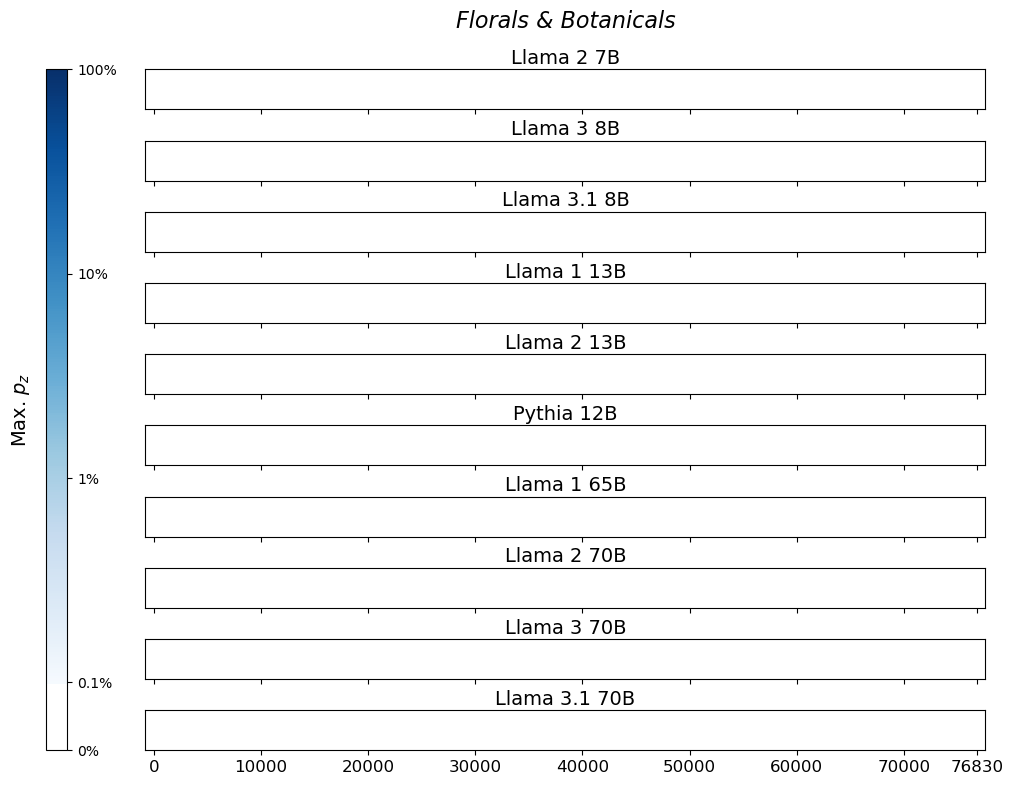}
    \includegraphics[width=\linewidth]{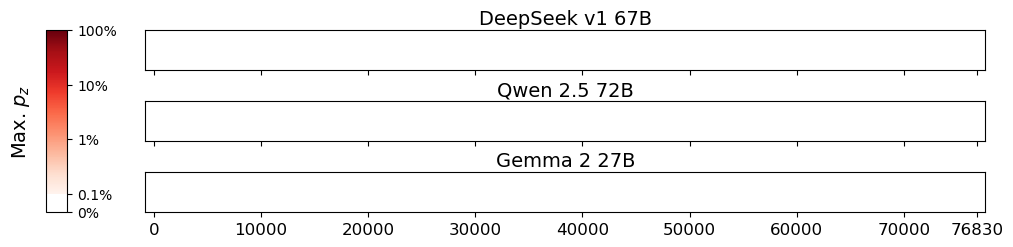}
    \includegraphics[width=\linewidth]{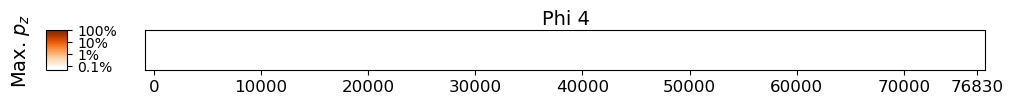}
  \end{minipage}
  \hfill
  \begin{minipage}[t]{0.45\textwidth}
    \centering
    \vspace{0cm}
    \includegraphics[width=\linewidth]{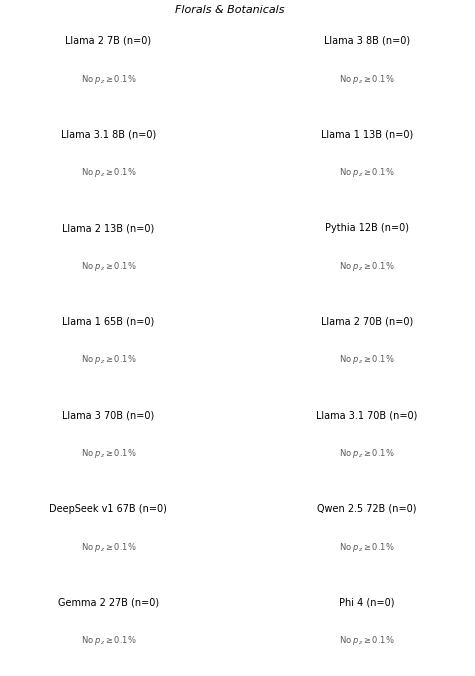}
  \end{minipage}
  \vspace{-.2cm}
  \caption{
    \textbf{\textit{Florals \& Botanicals}, \citeauthor{Florals_and_Botanicals}.}
    For $14$ LLMs,
    (\textbf{left}) heatmaps for the sliding-window procedure and
    (\textbf{right}) corresponding distributions over suffix extraction probabilities
    ($\tau_\text{min}=0.1\%$).
  }
  \label{fig:slidingwindow:Florals_and_Botanicals}
\end{figure}
\FloatBarrier

\clearpage
\subsubsection{\textit{Good Omens}, \citeauthor{Good_Omens}}\label{app:sec:sliding:Good_Omens}
\vspace{-.2cm}
\begin{figure}[h]
  \centering
  \begin{minipage}[t]{0.53\textwidth}
    \centering
    \vspace{0cm}
    \includegraphics[width=\linewidth]{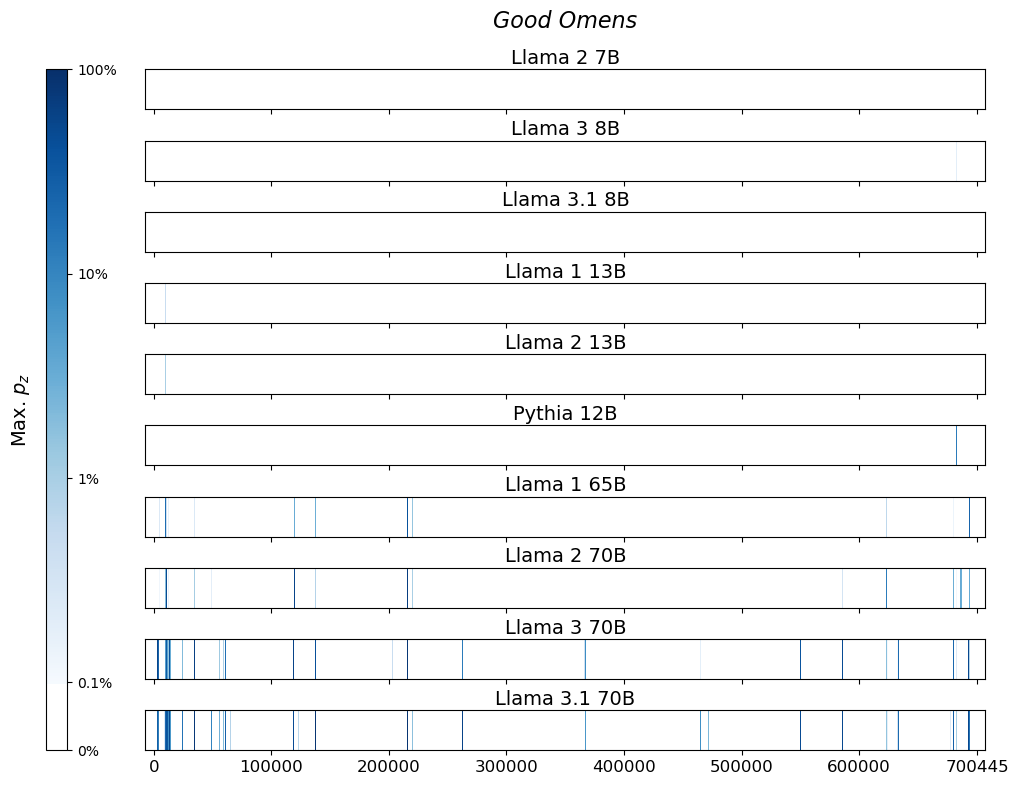}
    \includegraphics[width=\linewidth]{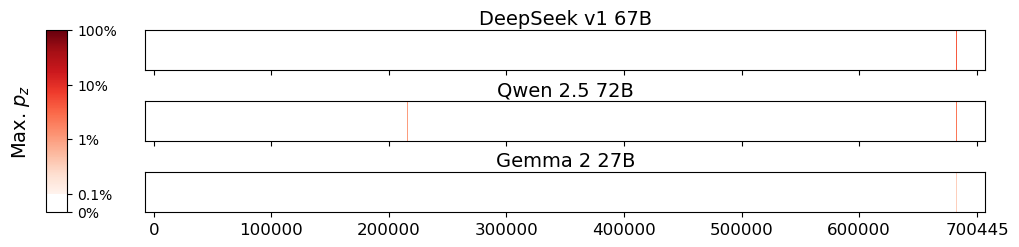}
    \includegraphics[width=\linewidth]{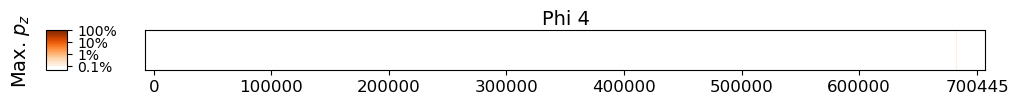}
  \end{minipage}
  \hfill
  \begin{minipage}[t]{0.45\textwidth}
    \centering
    \vspace{0cm}
    \includegraphics[width=\linewidth]{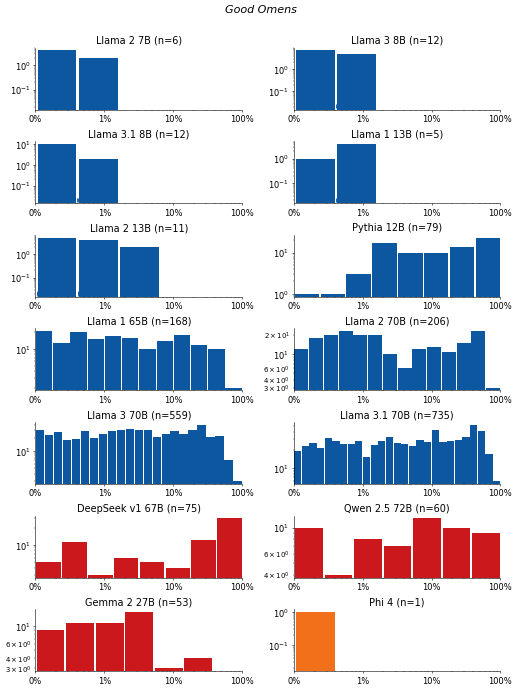}
  \end{minipage}
  \vspace{-.2cm}
  \caption{
    \textbf{\textit{Good Omens}, \citeauthor{Good_Omens}.}
    For $14$ LLMs,
    (\textbf{left}) heatmaps for the sliding-window procedure and
    (\textbf{right}) corresponding distributions over suffix extraction probabilities
    ($\tau_\text{min}=0.1\%$).
  }
  \label{fig:slidingwindow:Good_Omens}
\end{figure}
\FloatBarrier

\subsubsection{\textit{The Slippery Year}, \citeauthor{The_Slippery_Year}}\label{app:sec:sliding:The_Slippery_Year}
\vspace{-.2cm}
\begin{figure}[h]
  \centering
  \begin{minipage}[t]{0.53\textwidth}
    \centering
    \vspace{0cm}
    \includegraphics[width=\linewidth]{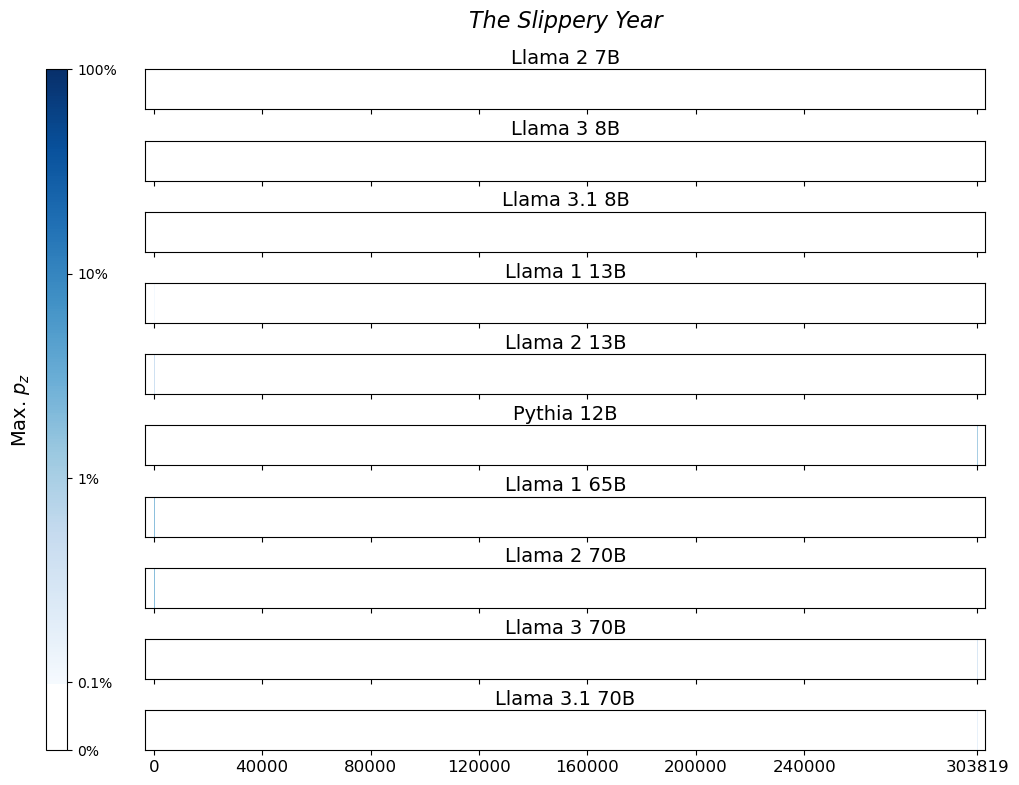}
    \includegraphics[width=\linewidth]{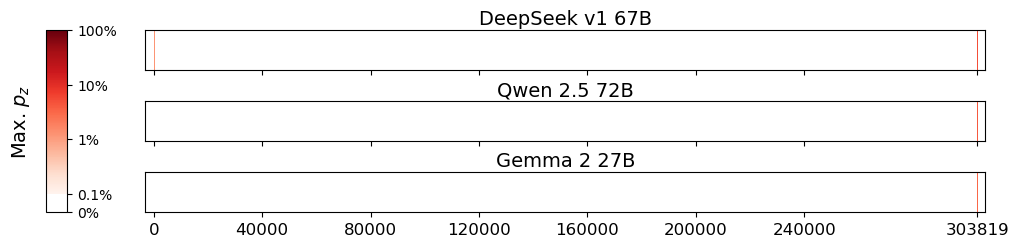}
    \includegraphics[width=\linewidth]{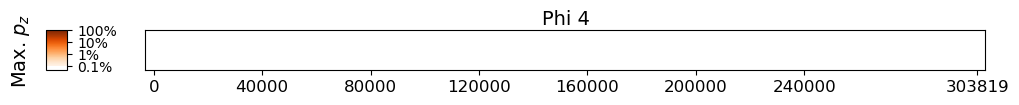}
  \end{minipage}
  \hfill
  \begin{minipage}[t]{0.45\textwidth}
    \centering
    \vspace{0cm}
    \includegraphics[width=\linewidth]{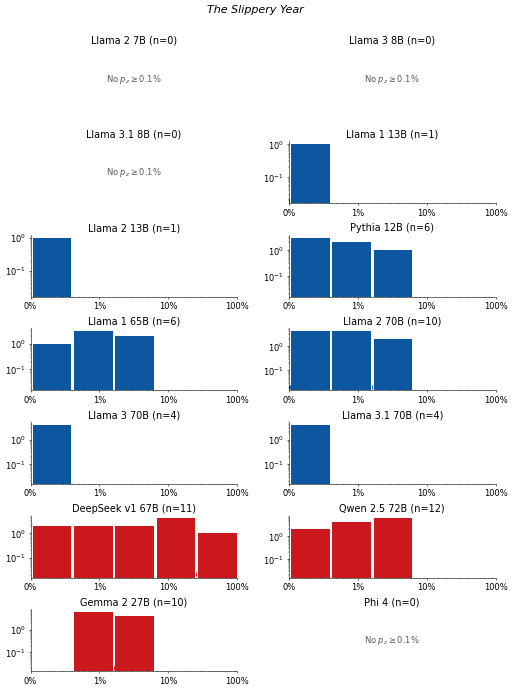}
  \end{minipage}
  \vspace{-.2cm}
  \caption{
    \textbf{\textit{The Slippery Year}, \citeauthor{The_Slippery_Year}.}
    For $14$ LLMs,
    (\textbf{left}) heatmaps for the sliding-window procedure and
    (\textbf{right}) corresponding distributions over suffix extraction probabilities
    ($\tau_\text{min}=0.1\%$).
  }
  \label{fig:slidingwindow:The_Slippery_Year}
\end{figure}
\FloatBarrier

\clearpage
\subsubsection{\textit{Sweater Surgery}, \citeauthor{Sweater_Surgery}}\label{app:sec:sliding:Sweater_Surgery}
\vspace{-.2cm}
\begin{figure}[h]
  \centering
  \begin{minipage}[t]{0.53\textwidth}
    \centering
    \vspace{0cm}
    \includegraphics[width=\linewidth]{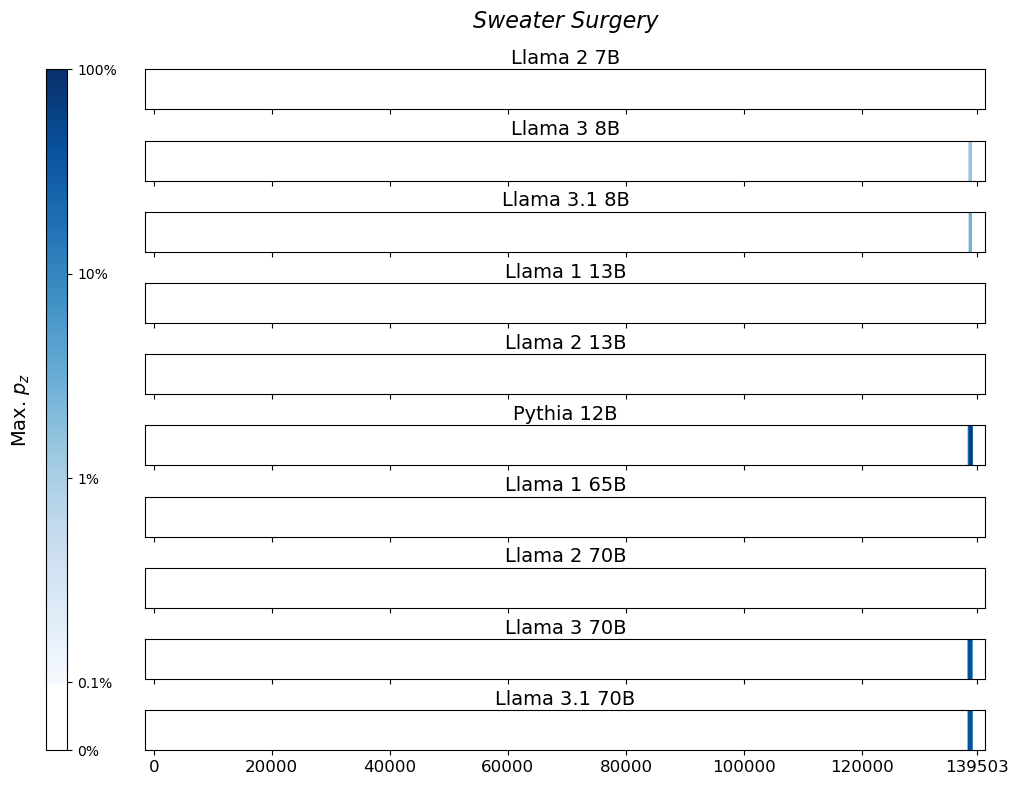}
    \includegraphics[width=\linewidth]{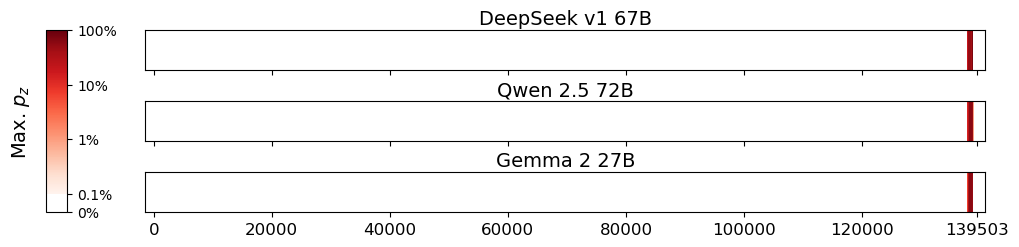}
    \includegraphics[width=\linewidth]{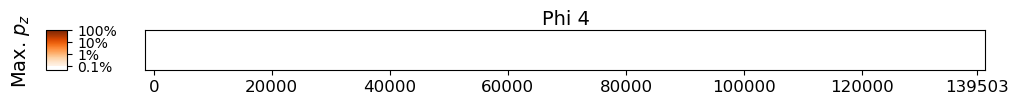}
  \end{minipage}
  \hfill
  \begin{minipage}[t]{0.45\textwidth}
    \centering
    \vspace{0cm}
    \includegraphics[width=\linewidth]{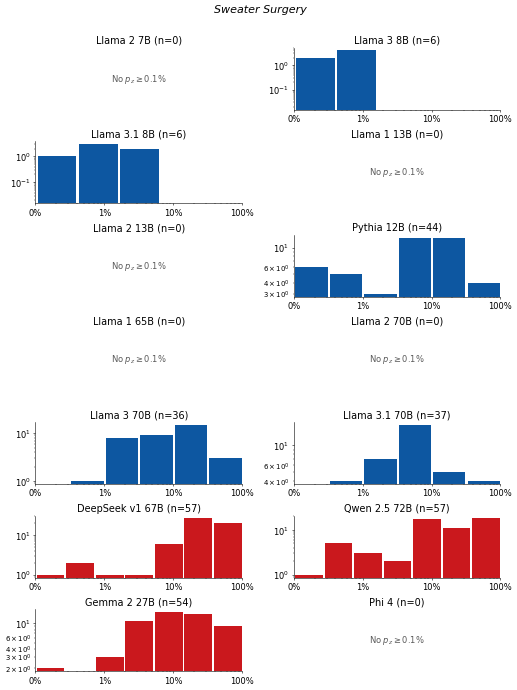}
  \end{minipage}
  \vspace{-.2cm}
  \caption{
    \textbf{\textit{Sweater Surgery}, \citeauthor{Sweater_Surgery}.}
    For $14$ LLMs,
    (\textbf{left}) heatmaps for the sliding-window procedure and
    (\textbf{right}) corresponding distributions over suffix extraction probabilities
    ($\tau_\text{min}=0.1\%$).
  }
  \label{fig:slidingwindow:Sweater_Surgery}
\end{figure}
\FloatBarrier

\subsubsection{\textit{Blink}, \citeauthor{Blink}}\label{app:sec:sliding:Blink}
\vspace{-.2cm}
\begin{figure}[h]
  \centering
  \begin{minipage}[t]{0.53\textwidth}
    \centering
    \vspace{0cm}
    \includegraphics[width=\linewidth]{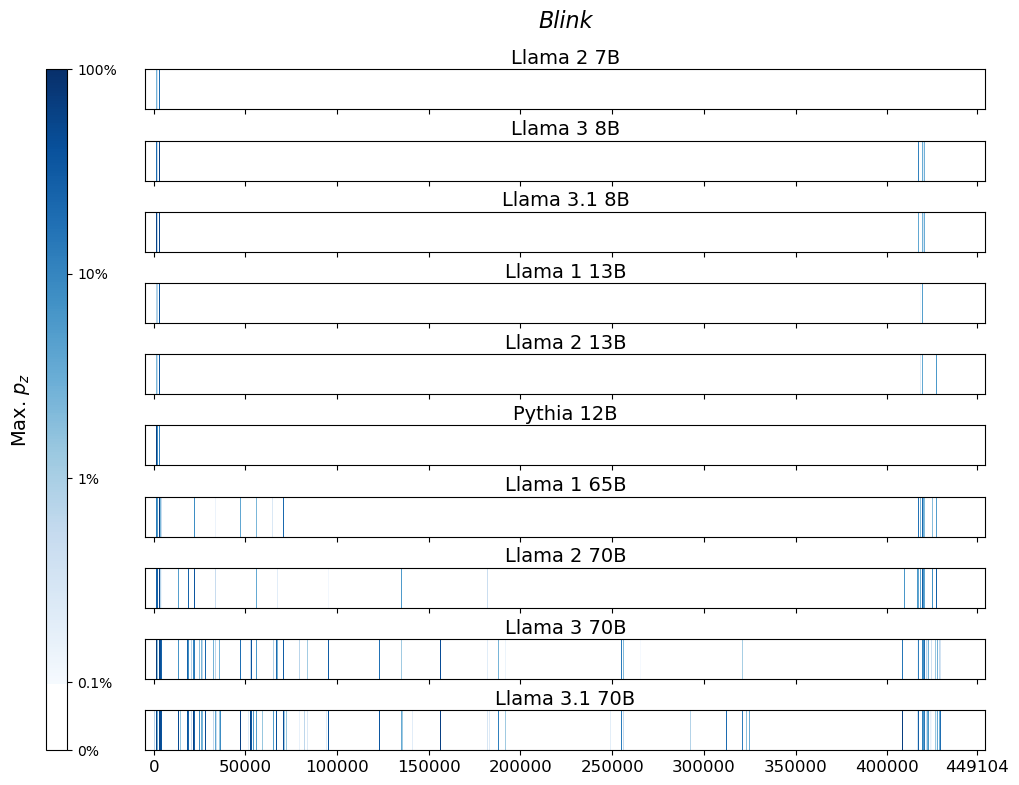}
    \includegraphics[width=\linewidth]{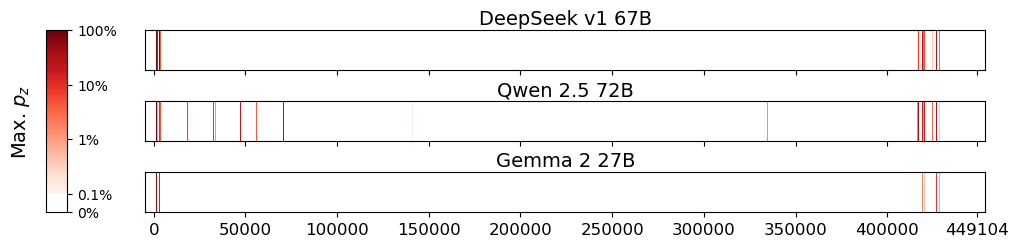}
    \includegraphics[width=\linewidth]{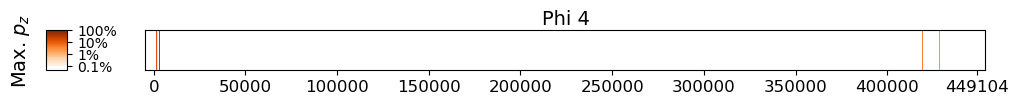}
  \end{minipage}
  \hfill
  \begin{minipage}[t]{0.45\textwidth}
    \centering
    \vspace{0cm}
    \includegraphics[width=\linewidth]{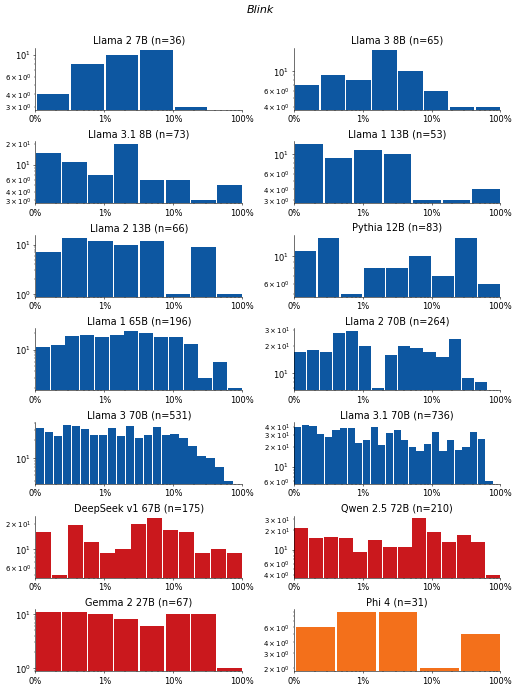}
  \end{minipage}
  \vspace{-.2cm}
  \caption{
    \textbf{\textit{Blink}, \citeauthor{Blink}.}
    For $14$ LLMs,
    (\textbf{left}) heatmaps for the sliding-window procedure and
    (\textbf{right}) corresponding distributions over suffix extraction probabilities
    ($\tau_\text{min}=0.1\%$).
  }
  \label{fig:slidingwindow:Blink}
\end{figure}
\FloatBarrier

\clearpage
\subsubsection{\textit{The Land Before Avocado}, \citeauthor{The_Land_Before_Avocado}}\label{app:sec:sliding:The_Land_Before_Avocado}
\vspace{-.2cm}
\begin{figure}[h]
  \centering
  \begin{minipage}[t]{0.53\textwidth}
    \centering
    \vspace{0cm}
    \includegraphics[width=\linewidth]{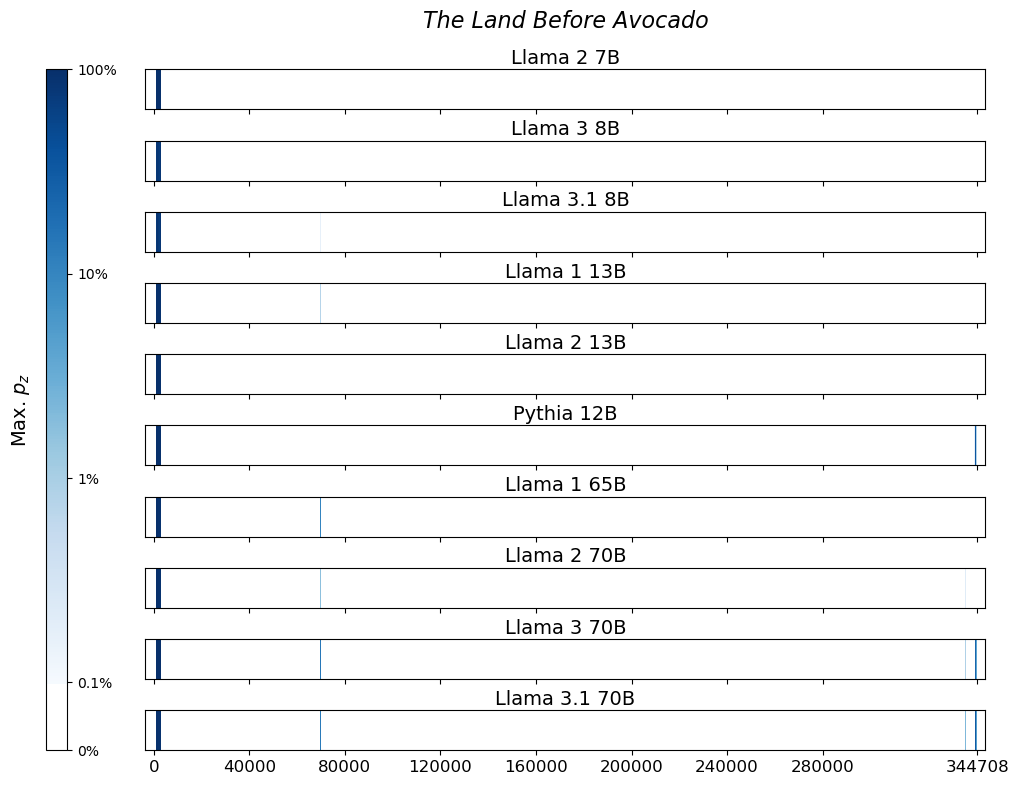}
    \includegraphics[width=\linewidth]{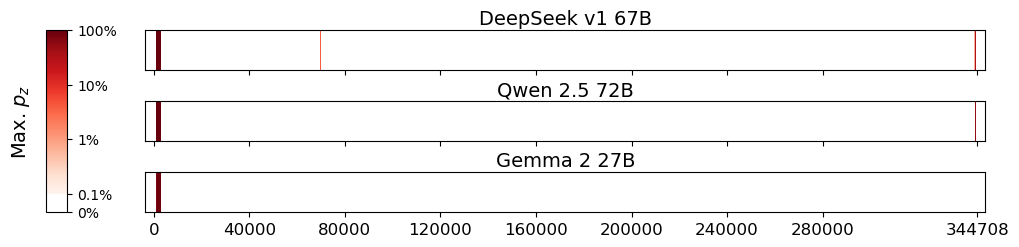}
    \includegraphics[width=\linewidth]{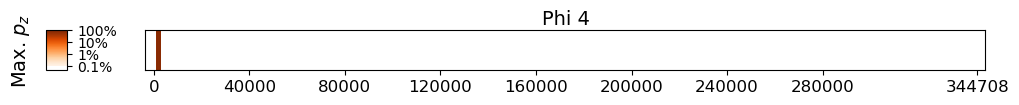}
  \end{minipage}
  \hfill
  \begin{minipage}[t]{0.45\textwidth}
    \centering
    \vspace{0cm}
    \includegraphics[width=\linewidth]{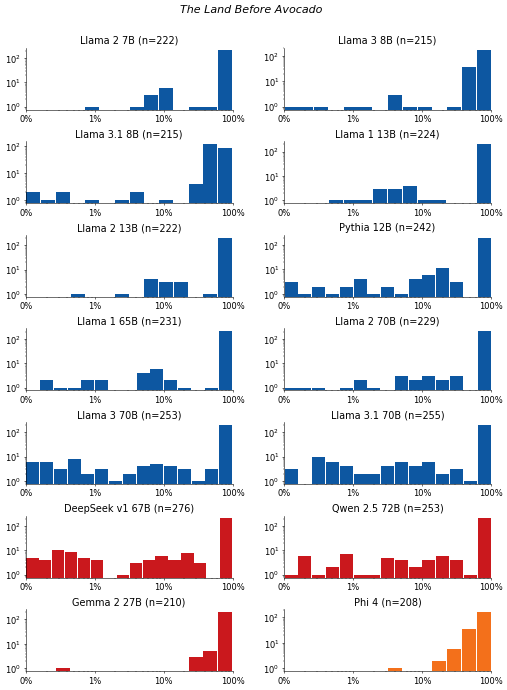}
  \end{minipage}
  \vspace{-.2cm}
  \caption{
    \textbf{\textit{The Land Before Avocado}, \citeauthor{The_Land_Before_Avocado}.}
    For $14$ LLMs,
    (\textbf{left}) heatmaps for the sliding-window procedure and
    (\textbf{right}) corresponding distributions over suffix extraction probabilities
    ($\tau_\text{min}=0.1\%$).
  }
  \label{fig:slidingwindow:The_Land_Before_Avocado}
\end{figure}
\FloatBarrier

\subsubsection{\textit{Dead Ringers}, \citeauthor{Dead_Ringers}}\label{app:sec:sliding:Dead_Ringers}
\vspace{-.2cm}
\begin{figure}[h]
  \centering
  \begin{minipage}[t]{0.53\textwidth}
    \centering
    \vspace{0cm}
    \includegraphics[width=\linewidth]{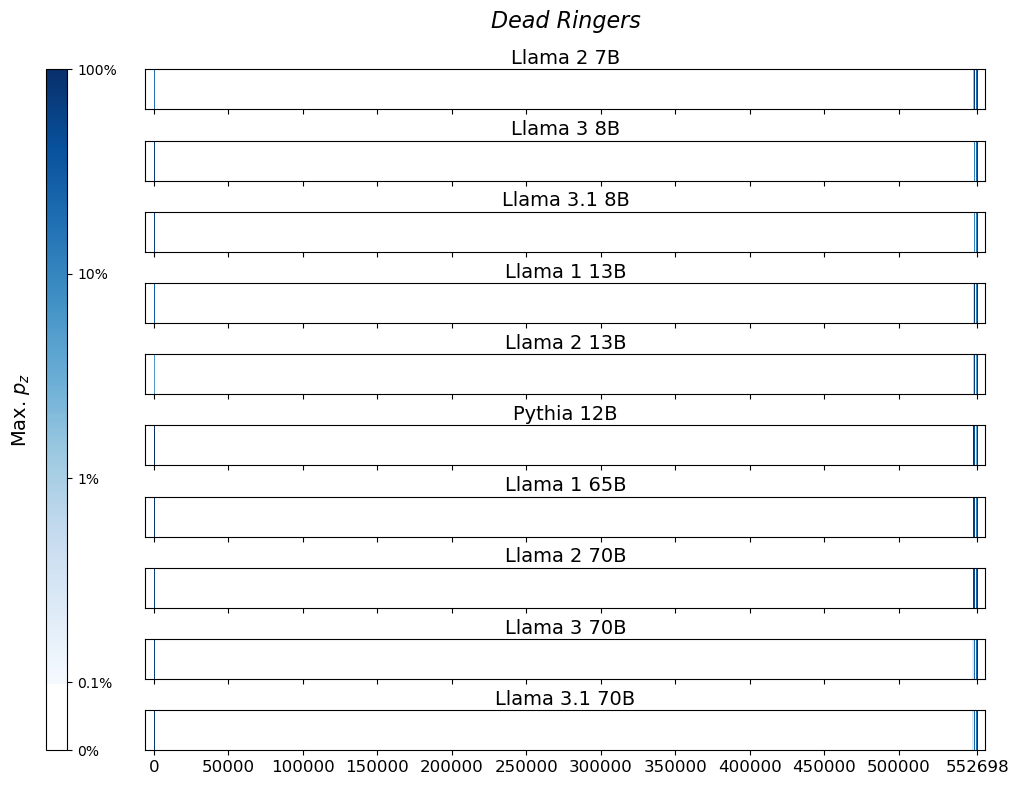}
    \includegraphics[width=\linewidth]{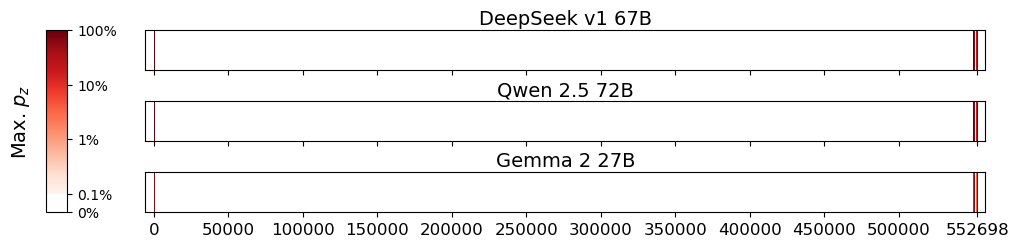}
    \includegraphics[width=\linewidth]{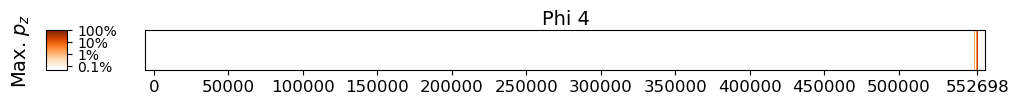}
  \end{minipage}
  \hfill
  \begin{minipage}[t]{0.45\textwidth}
    \centering
    \vspace{0cm}
    \includegraphics[width=\linewidth]{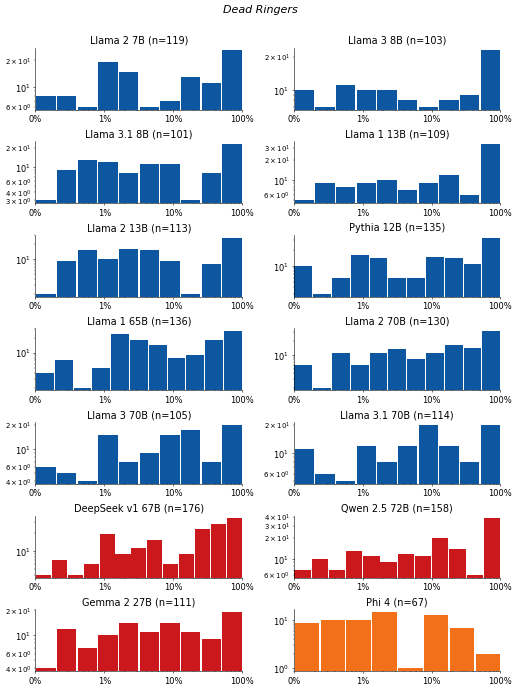}
  \end{minipage}
  \vspace{-.2cm}
  \caption{
    \textbf{\textit{Dead Ringers}, \citeauthor{Dead_Ringers}.}
    For $14$ LLMs,
    (\textbf{left}) heatmaps for the sliding-window procedure and
    (\textbf{right}) corresponding distributions over suffix extraction probabilities
    ($\tau_\text{min}=0.1\%$).
  }
  \label{fig:slidingwindow:Dead_Ringers}
\end{figure}
\FloatBarrier

\clearpage
\subsubsection{\textit{Ararat}, \citeauthor{Ararat}}\label{app:sec:sliding:Ararat}
\vspace{-.2cm}
\begin{figure}[h]
  \centering
  \begin{minipage}[t]{0.53\textwidth}
    \centering
    \vspace{0cm}
    \includegraphics[width=\linewidth]{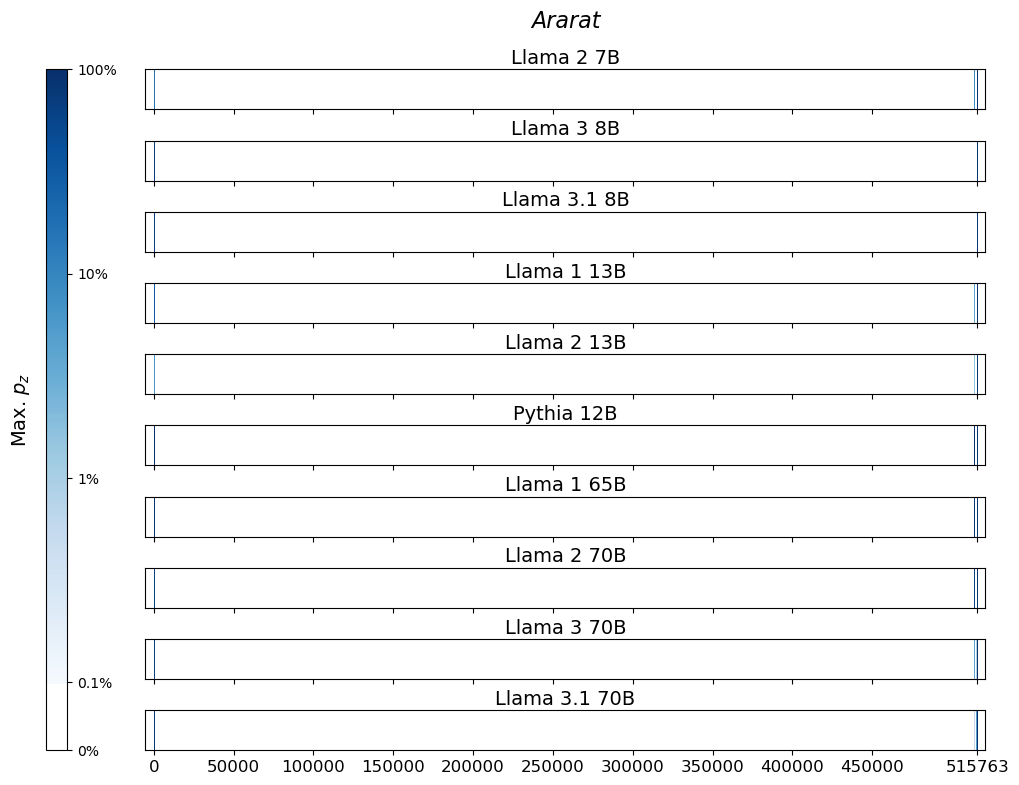}
    \includegraphics[width=\linewidth]{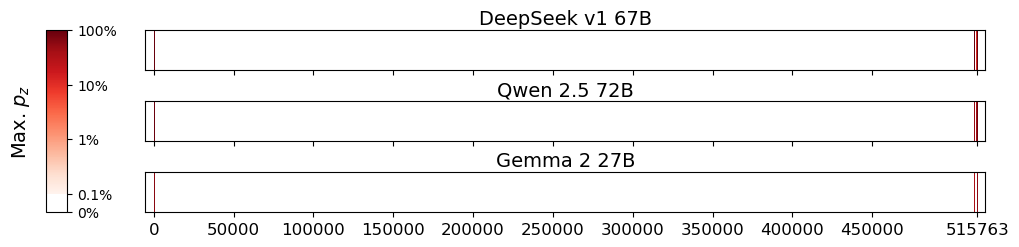}
    \includegraphics[width=\linewidth]{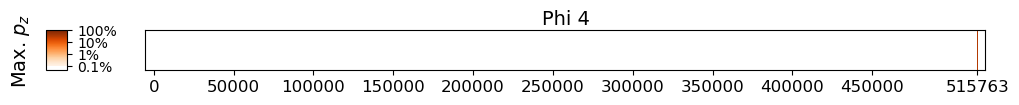}
  \end{minipage}
  \hfill
  \begin{minipage}[t]{0.45\textwidth}
    \centering
    \vspace{0cm}
    \includegraphics[width=\linewidth]{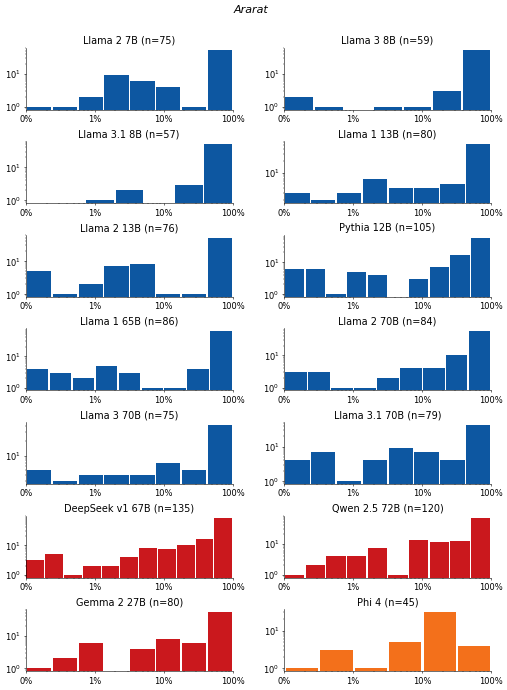}
  \end{minipage}
  \vspace{-.2cm}
  \caption{
    \textbf{\textit{Ararat}, \citeauthor{Ararat}.}
    For $14$ LLMs,
    (\textbf{left}) heatmaps for the sliding-window procedure and
    (\textbf{right}) corresponding distributions over suffix extraction probabilities
    ($\tau_\text{min}=0.1\%$).
  }
  \label{fig:slidingwindow:Ararat}
\end{figure}
\FloatBarrier

\subsubsection{\textit{Lord of the Flies}, \citeauthor{Lord_of_the_Flies}}\label{app:sec:sliding:Lord_of_the_Flies}
\vspace{-.2cm}
\begin{figure}[h]
  \centering
  \begin{minipage}[t]{0.53\textwidth}
    \centering
    \vspace{0cm}
    \includegraphics[width=\linewidth]{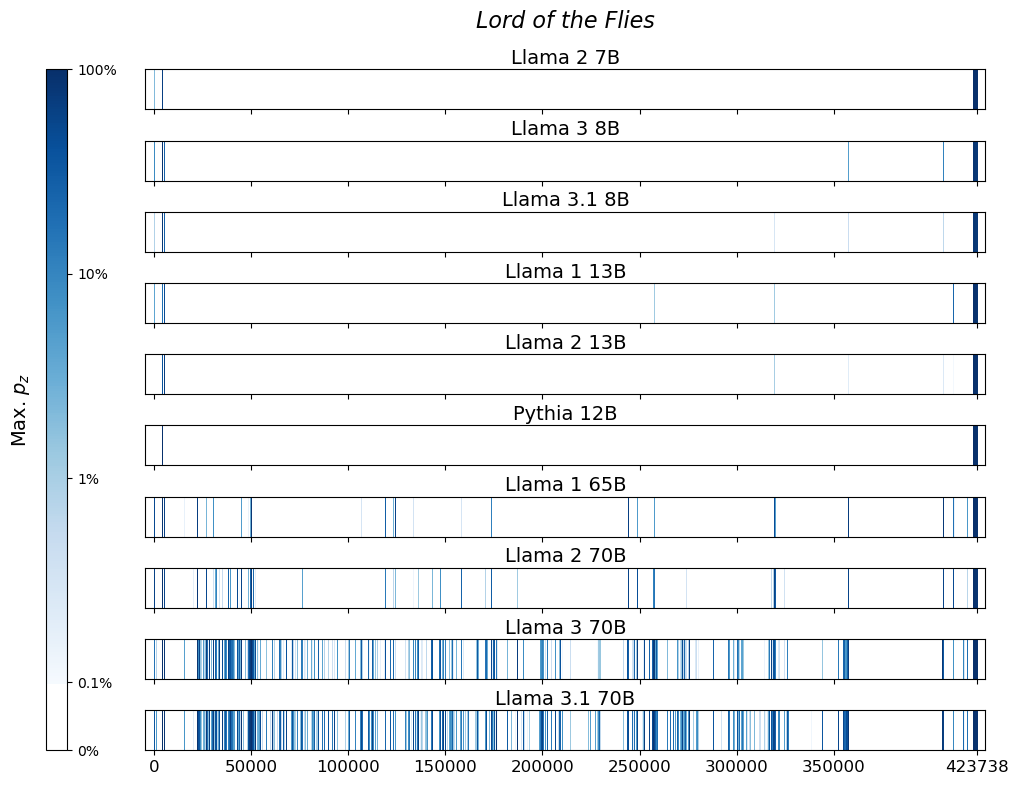}
    \includegraphics[width=\linewidth]{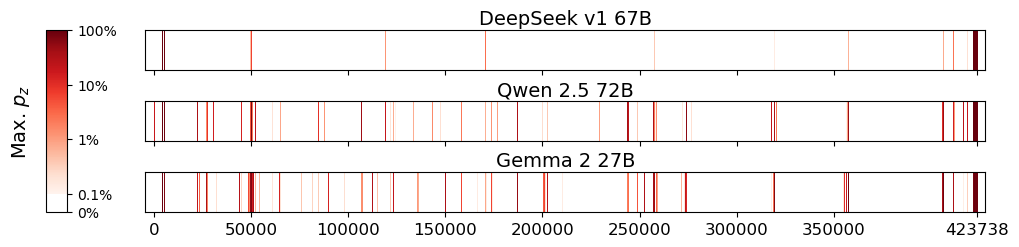}
    \includegraphics[width=\linewidth]{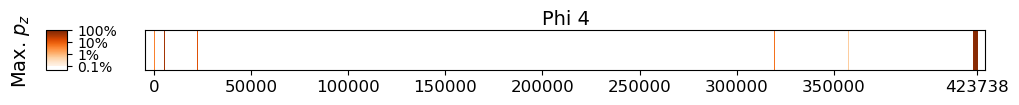}
  \end{minipage}
  \hfill
  \begin{minipage}[t]{0.45\textwidth}
    \centering
    \vspace{0cm}
    \includegraphics[width=\linewidth]{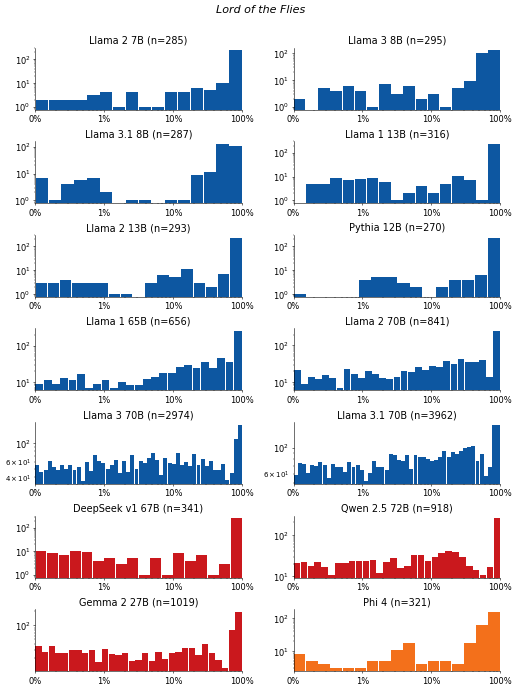}
  \end{minipage}
  \vspace{-.2cm}
  \caption{
    \textbf{\textit{Lord of the Flies}, \citeauthor{Lord_of_the_Flies}.}
    For $14$ LLMs,
    (\textbf{left}) heatmaps for the sliding-window procedure and
    (\textbf{right}) corresponding distributions over suffix extraction probabilities
    ($\tau_\text{min}=0.1\%$).
  }
  \label{fig:slidingwindow:Lord_of_the_Flies}
\end{figure}
\FloatBarrier

\clearpage
\subsubsection{\textit{Wizard's First Rule}, \citeauthor{Wizard_s_First_Rule}}\label{app:sec:sliding:Wizard_s_First_Rule}
\vspace{-.2cm}
\begin{figure}[h]
  \centering
  \begin{minipage}[t]{0.53\textwidth}
    \centering
    \vspace{0cm}
    \includegraphics[width=\linewidth]{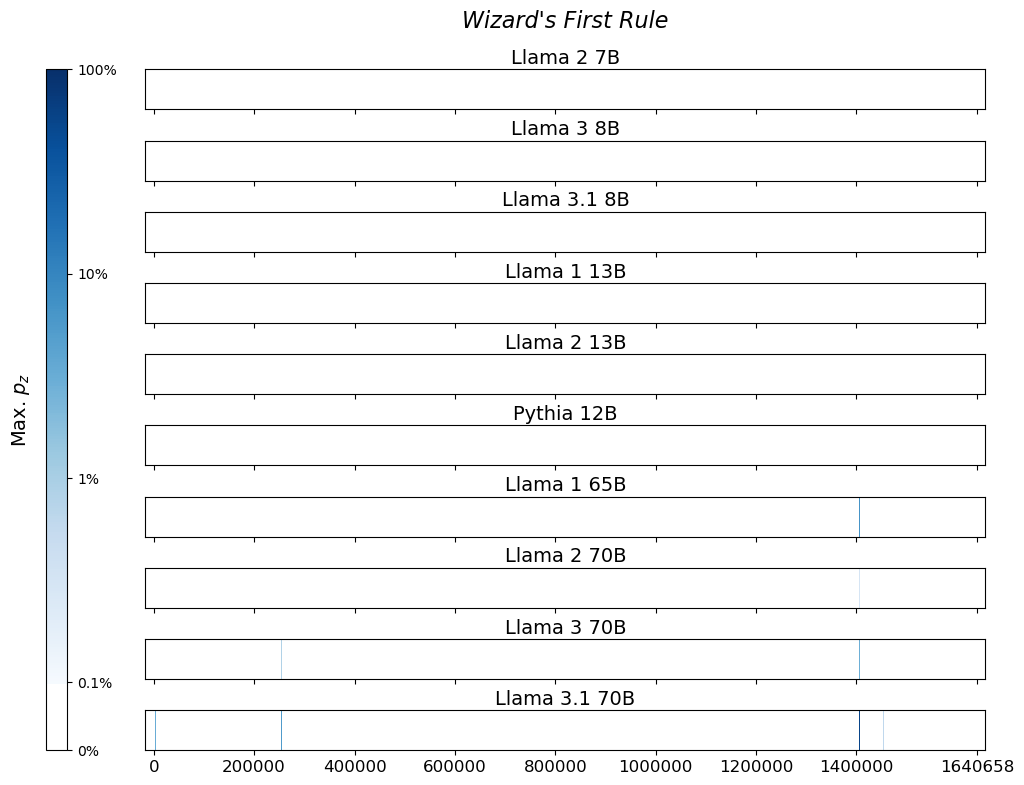}
    \includegraphics[width=\linewidth]{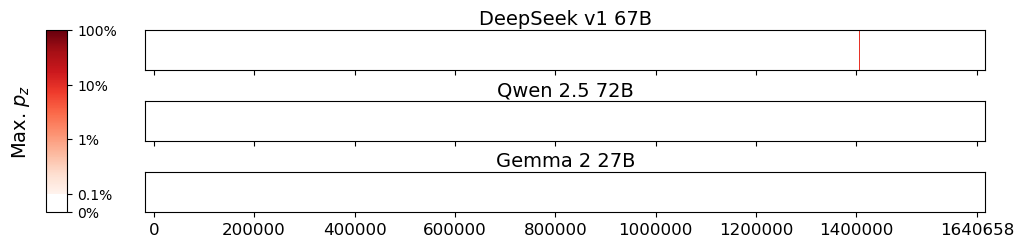}
    \includegraphics[width=\linewidth]{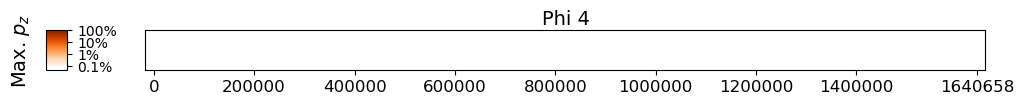}
  \end{minipage}
  \hfill
  \begin{minipage}[t]{0.45\textwidth}
    \centering
    \vspace{0cm}
    \includegraphics[width=\linewidth]{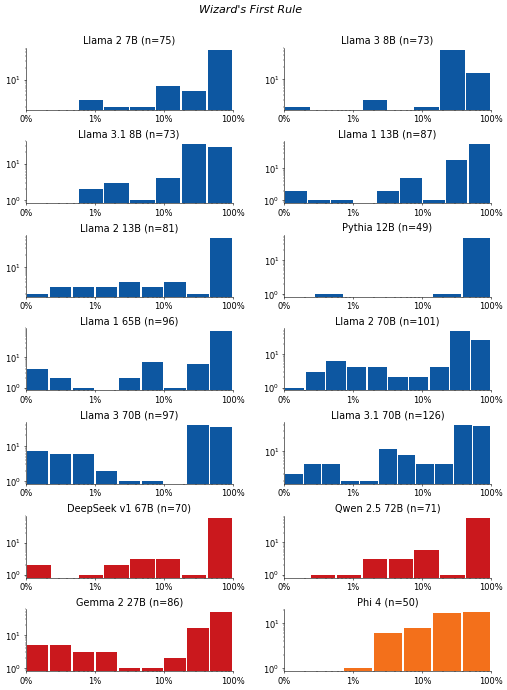}
  \end{minipage}
  \vspace{-.2cm}
  \caption{
    \textbf{\textit{Wizard's First Rule}, \citeauthor{Wizard_s_First_Rule}.}
    For $14$ LLMs,
    (\textbf{left}) heatmaps for the sliding-window procedure and
    (\textbf{right}) corresponding distributions over suffix extraction probabilities
    ($\tau_\text{min}=0.1\%$).
  }
  \label{fig:slidingwindow:Wizard_s_First_Rule}
\end{figure}
\FloatBarrier

\subsubsection{\textit{Rome and Jerusalem}, \citeauthor{Rome_and_Jerusalem}}\label{app:sec:sliding:Rome_and_Jerusalem}
\vspace{-.2cm}
\begin{figure}[h]
  \centering
  \begin{minipage}[t]{0.53\textwidth}
    \centering
    \vspace{0cm}
    \includegraphics[width=\linewidth]{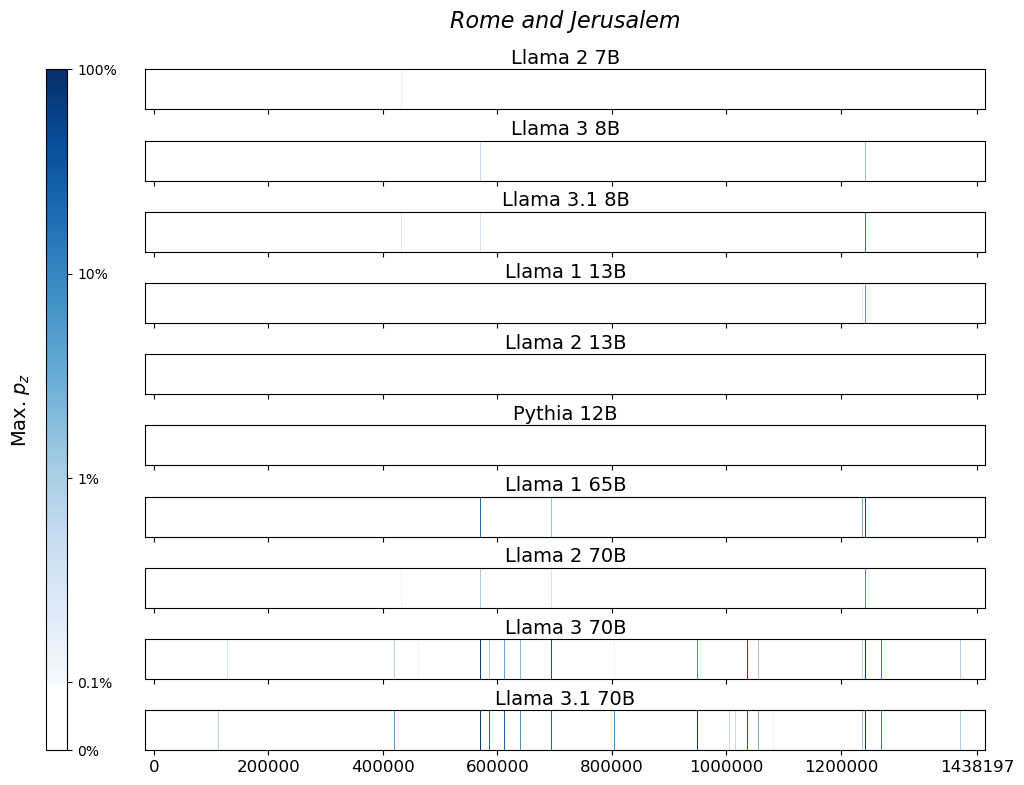}
    \includegraphics[width=\linewidth]{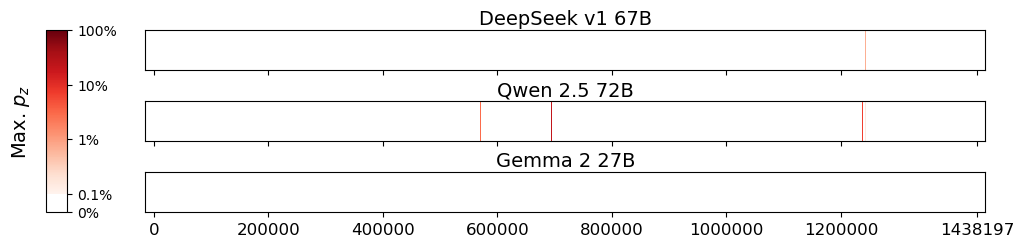}
    \includegraphics[width=\linewidth]{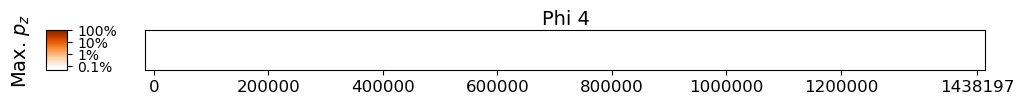}
  \end{minipage}
  \hfill
  \begin{minipage}[t]{0.45\textwidth}
    \centering
    \vspace{0cm}
    \includegraphics[width=\linewidth]{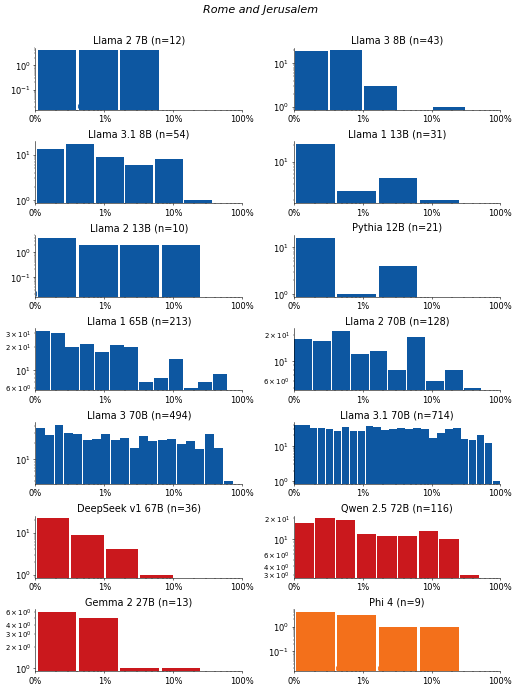}
  \end{minipage}
  \vspace{-.2cm}
  \caption{
    \textbf{\textit{Rome and Jerusalem}, \citeauthor{Rome_and_Jerusalem}.}
    For $14$ LLMs,
    (\textbf{left}) heatmaps for the sliding-window procedure and
    (\textbf{right}) corresponding distributions over suffix extraction probabilities
    ($\tau_\text{min}=0.1\%$).
  }
  \label{fig:slidingwindow:Rome_and_Jerusalem}
\end{figure}
\FloatBarrier

\clearpage
\subsubsection{\textit{The Fault in Our Stars}, \citeauthor{The_Fault_in_Our_Stars}}\label{app:sec:sliding:The_Fault_in_Our_Stars}
\vspace{-.2cm}
\begin{figure}[h]
  \centering
  \begin{minipage}[t]{0.53\textwidth}
    \centering
    \vspace{0cm}
    \includegraphics[width=\linewidth]{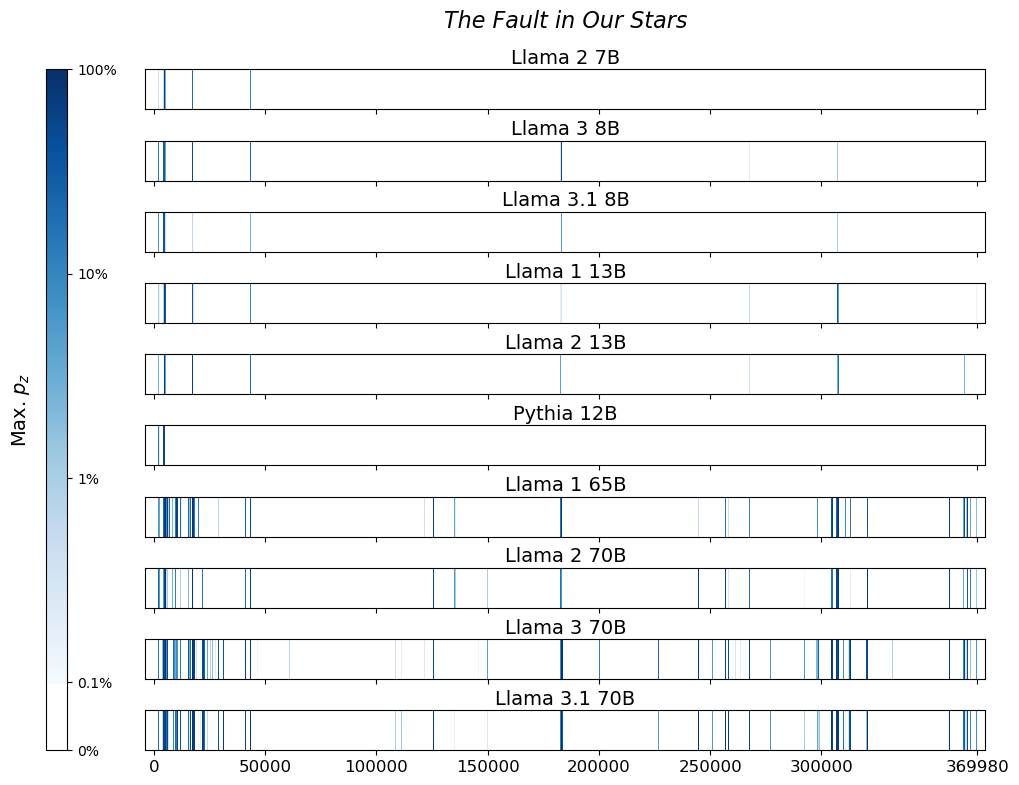}
    \includegraphics[width=\linewidth]{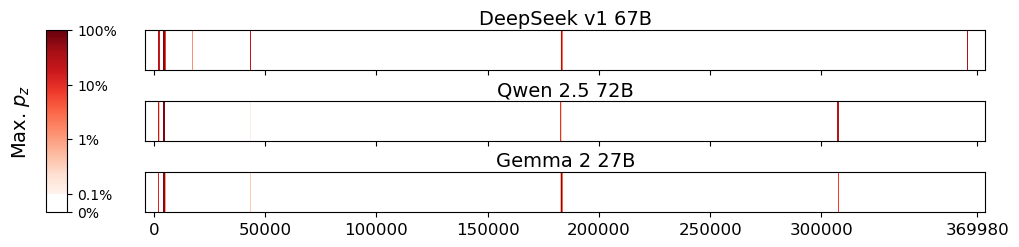}
    \includegraphics[width=\linewidth]{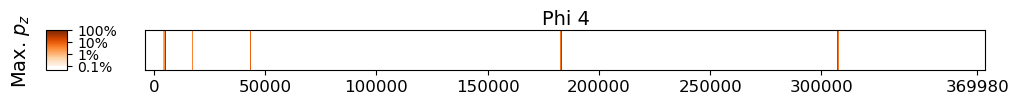}
  \end{minipage}
  \hfill
  \begin{minipage}[t]{0.45\textwidth}
    \centering
    \vspace{0cm}
    \includegraphics[width=\linewidth]{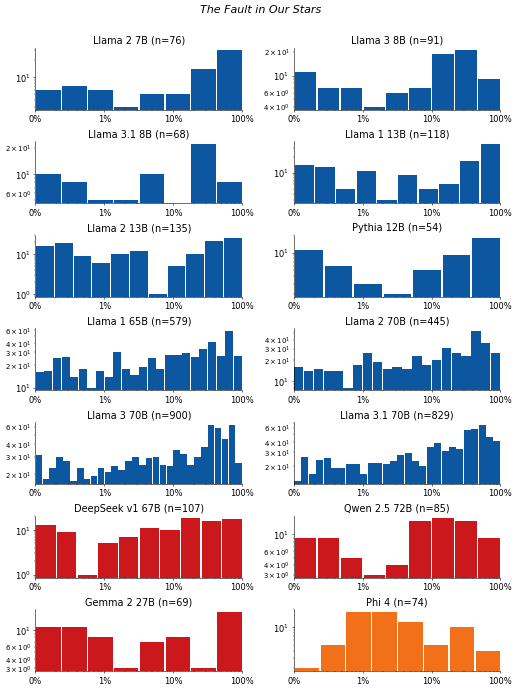}
  \end{minipage}
  \vspace{-.2cm}
  \caption{
    \textbf{\textit{The Fault in Our Stars}, \citeauthor{The_Fault_in_Our_Stars}.}
    For $14$ LLMs,
    (\textbf{left}) heatmaps for the sliding-window procedure and
    (\textbf{right}) corresponding distributions over suffix extraction probabilities
    ($\tau_\text{min}=0.1\%$).
  }
  \label{fig:slidingwindow:The_Fault_in_Our_Stars}
\end{figure}
\FloatBarrier

\subsubsection{\textit{The Third Man}, \citeauthor{The_Third_Man}}\label{app:sec:sliding:The_Third_Man}
\vspace{-.2cm}
\begin{figure}[h]
  \centering
  \begin{minipage}[t]{0.53\textwidth}
    \centering
    \vspace{0cm}
    \includegraphics[width=\linewidth]{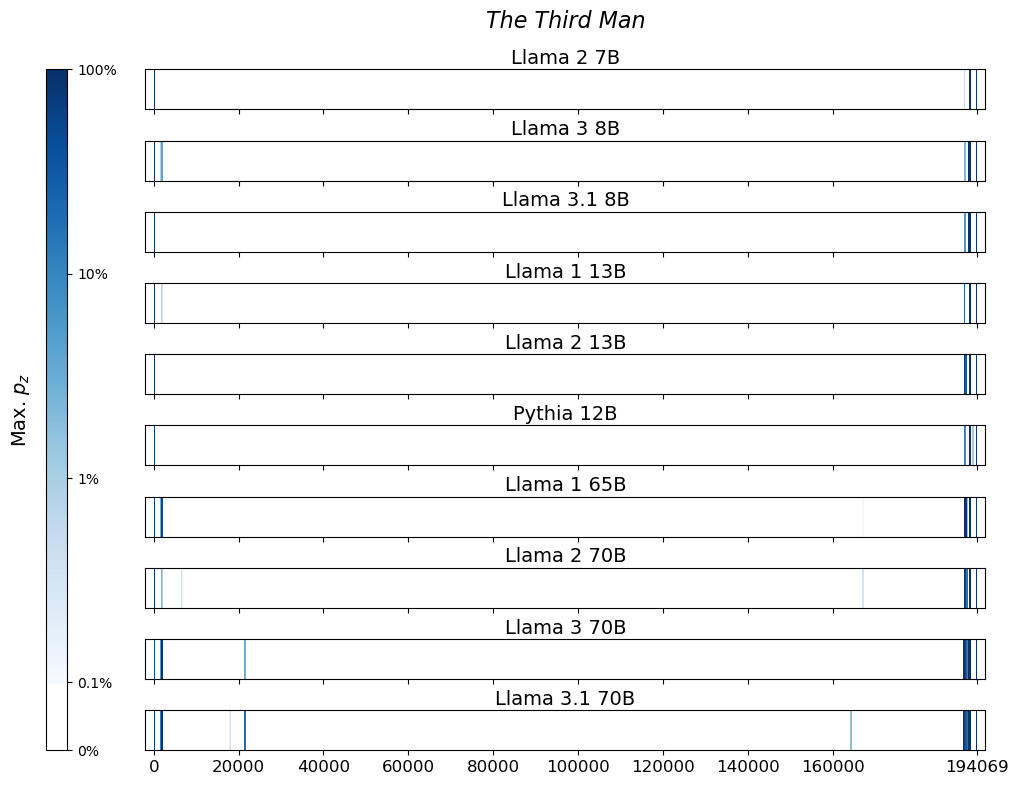}
    \includegraphics[width=\linewidth]{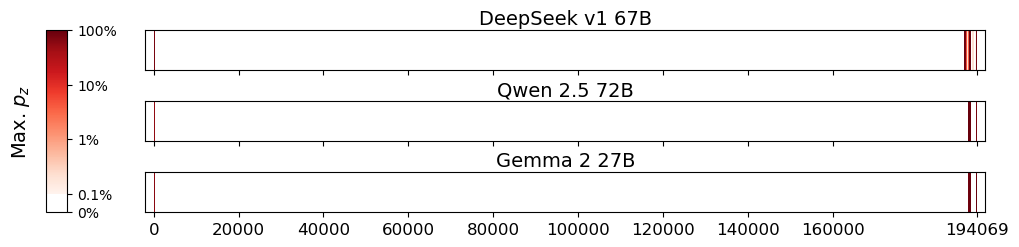}
    \includegraphics[width=\linewidth]{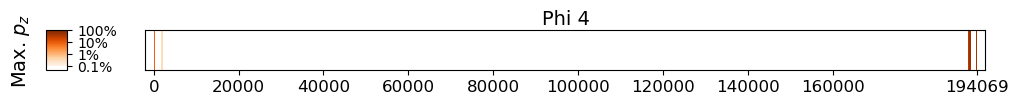}
  \end{minipage}
  \hfill
  \begin{minipage}[t]{0.45\textwidth}
    \centering
    \vspace{0cm}
    \includegraphics[width=\linewidth]{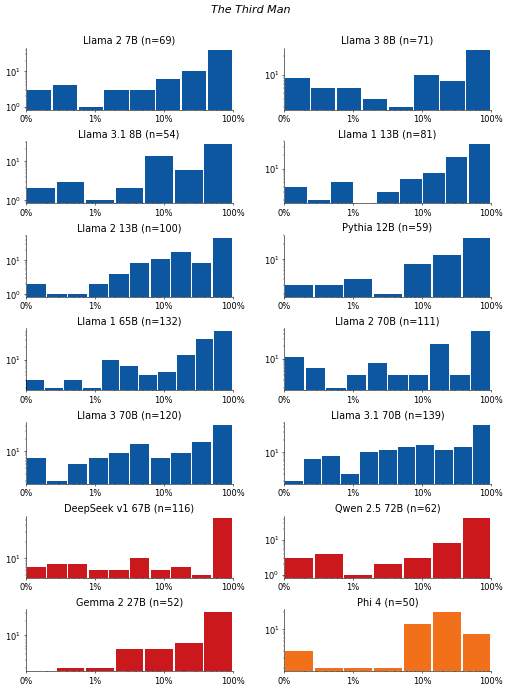}
  \end{minipage}
  \vspace{-.2cm}
  \caption{
    \textbf{\textit{The Third Man}, \citeauthor{The_Third_Man}.}
    For $14$ LLMs,
    (\textbf{left}) heatmaps for the sliding-window procedure and
    (\textbf{right}) corresponding distributions over suffix extraction probabilities
    ($\tau_\text{min}=0.1\%$).
  }
  \label{fig:slidingwindow:The_Third_Man}
\end{figure}
\FloatBarrier

\clearpage
\subsubsection{\textit{The Confessions of Max Tivoli}, \citeauthor{The_Confessions_of_Max_Tivoli}}\label{app:sec:sliding:The_Confessions_of_Max_Tivoli}
\vspace{-.2cm}
\begin{figure}[h]
  \centering
  \begin{minipage}[t]{0.53\textwidth}
    \centering
    \vspace{0cm}
    \includegraphics[width=\linewidth]{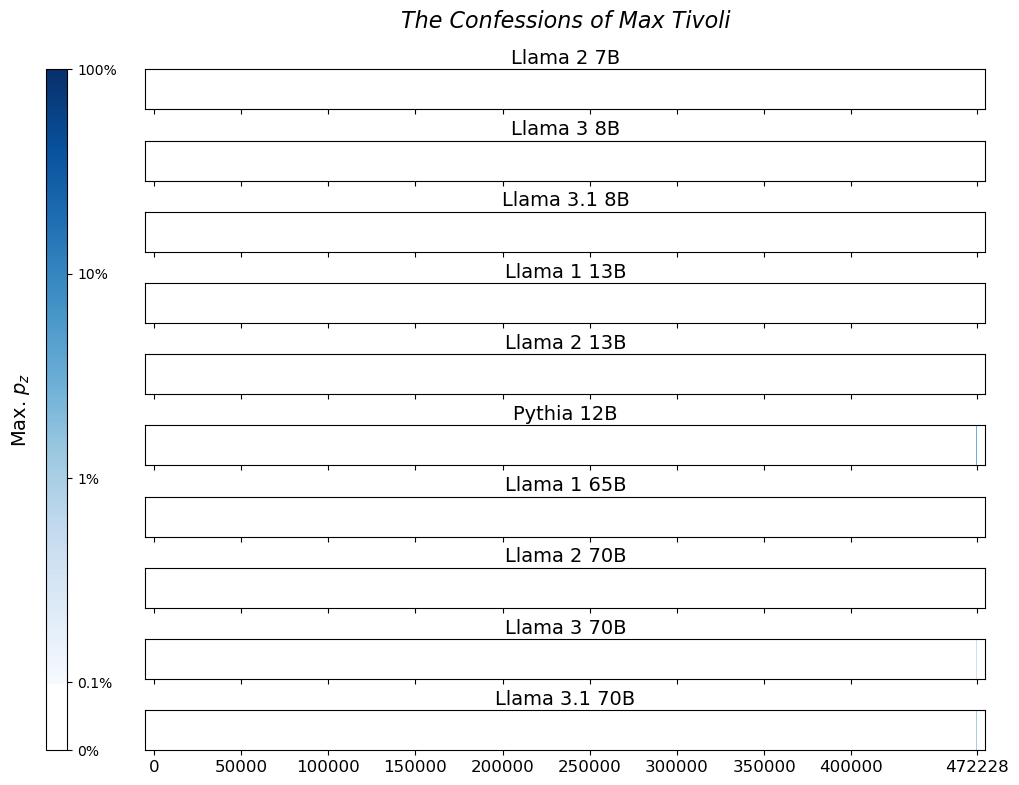}
    \includegraphics[width=\linewidth]{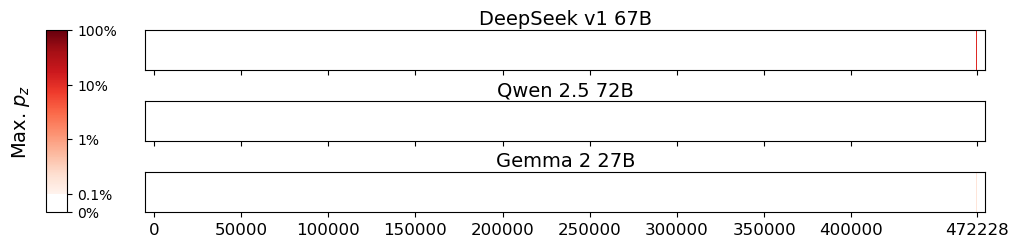}
    \includegraphics[width=\linewidth]{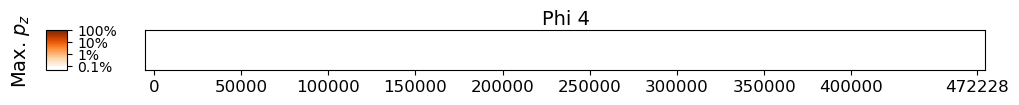}
  \end{minipage}
  \hfill
  \begin{minipage}[t]{0.45\textwidth}
    \centering
    \vspace{0cm}
    \includegraphics[width=\linewidth]{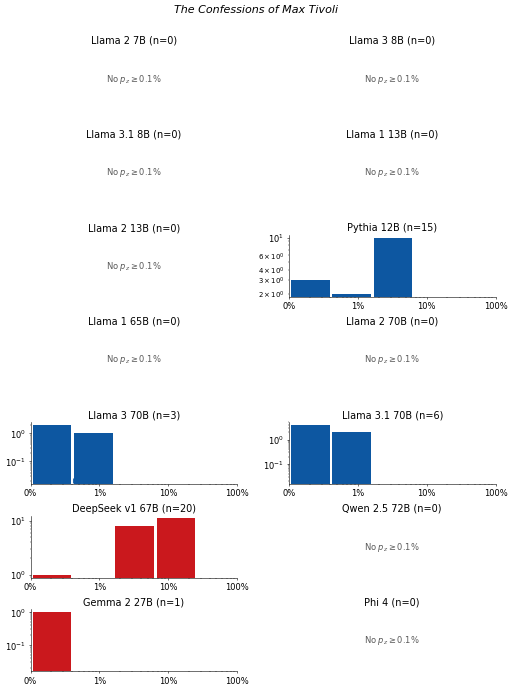}
  \end{minipage}
  \vspace{-.2cm}
  \caption{
    \textbf{\textit{The Confessions of Max Tivoli}, \citeauthor{The_Confessions_of_Max_Tivoli}.}
    For $14$ LLMs,
    (\textbf{left}) heatmaps for the sliding-window procedure and
    (\textbf{right}) corresponding distributions over suffix extraction probabilities
    ($\tau_\text{min}=0.1\%$).
  }
  \label{fig:slidingwindow:The_Confessions_of_Max_Tivoli}
\end{figure}
\FloatBarrier

\subsubsection{\textit{The Fugitive}, \citeauthor{The_Fugitive}}\label{app:sec:sliding:The_Fugitive}
\vspace{-.2cm}
\begin{figure}[h]
  \centering
  \begin{minipage}[t]{0.53\textwidth}
    \centering
    \vspace{0cm}
    \includegraphics[width=\linewidth]{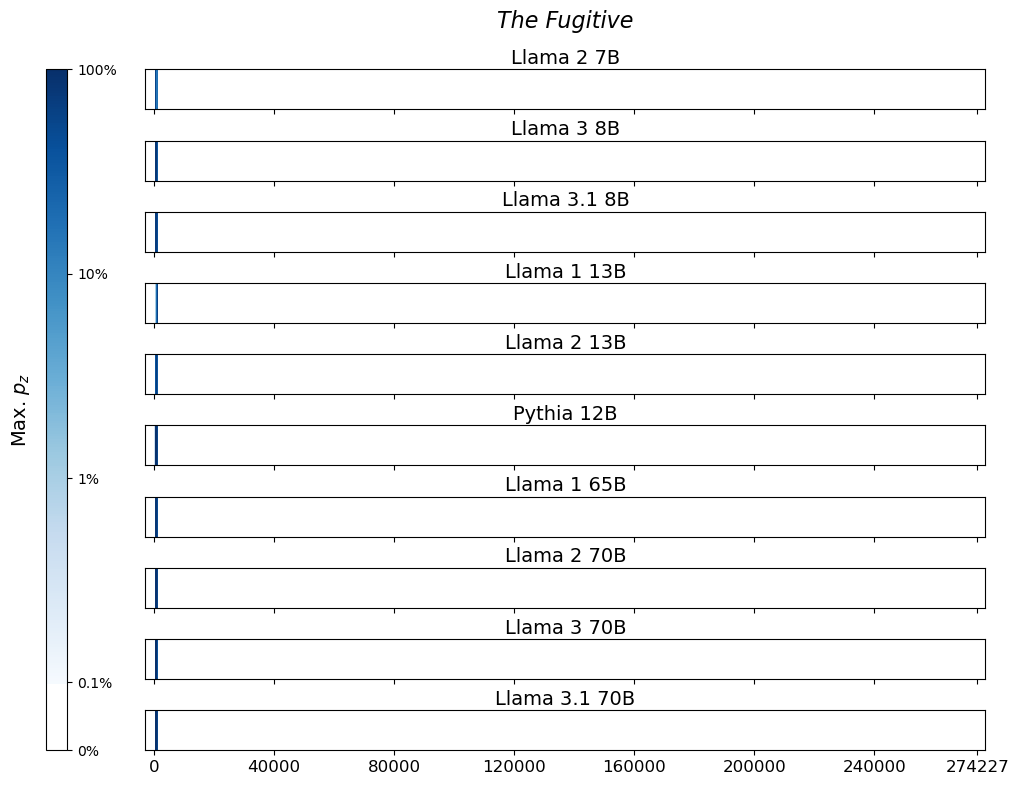}
    \includegraphics[width=\linewidth]{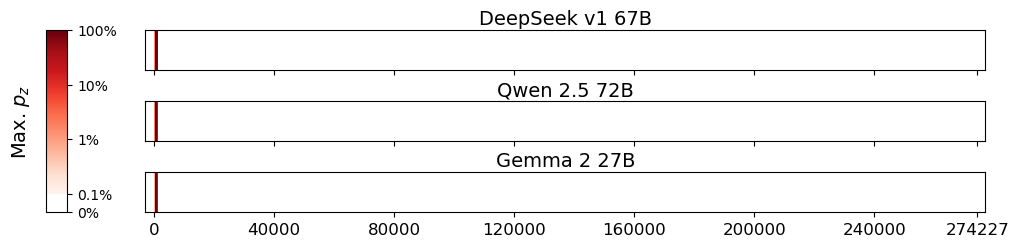}
    \includegraphics[width=\linewidth]{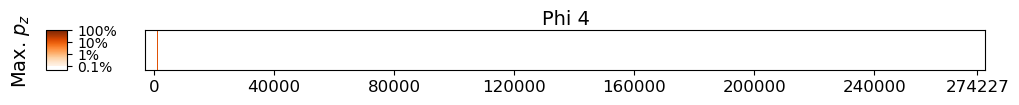}
  \end{minipage}
  \hfill
  \begin{minipage}[t]{0.45\textwidth}
    \centering
    \vspace{0cm}
    \includegraphics[width=\linewidth]{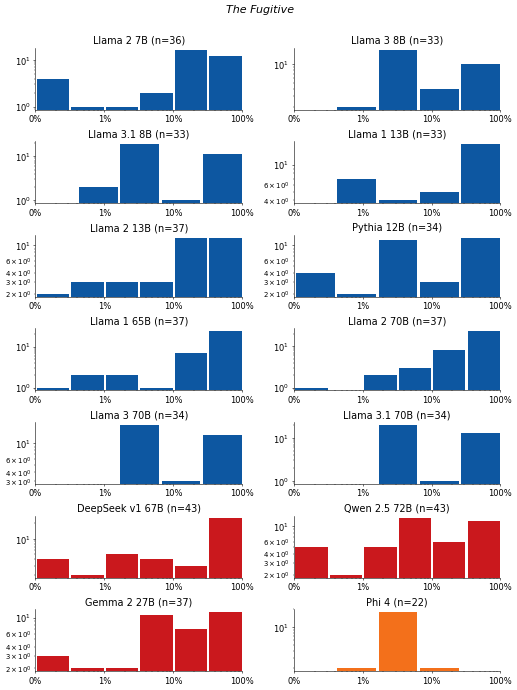}
  \end{minipage}
  \vspace{-.2cm}
  \caption{
    \textbf{\textit{The Fugitive}, \citeauthor{The_Fugitive}.}
    For $14$ LLMs,
    (\textbf{left}) heatmaps for the sliding-window procedure and
    (\textbf{right}) corresponding distributions over suffix extraction probabilities
    ($\tau_\text{min}=0.1\%$).
  }
  \label{fig:slidingwindow:The_Fugitive}
\end{figure}
\FloatBarrier

\clearpage
\subsubsection{\textit{The Curious Incident of the Dog in the Night-Time}, \citeauthor{The_Curious_Incident_of_the_Dog_in_the_Night-Time}}\label{app:sec:sliding:The_Curious_Incident_of_the_Dog_in_the_Night-Time}
\vspace{-.2cm}
\begin{figure}[h]
  \centering
  \begin{minipage}[t]{0.53\textwidth}
    \centering
    \vspace{0cm}
    \includegraphics[width=\linewidth]{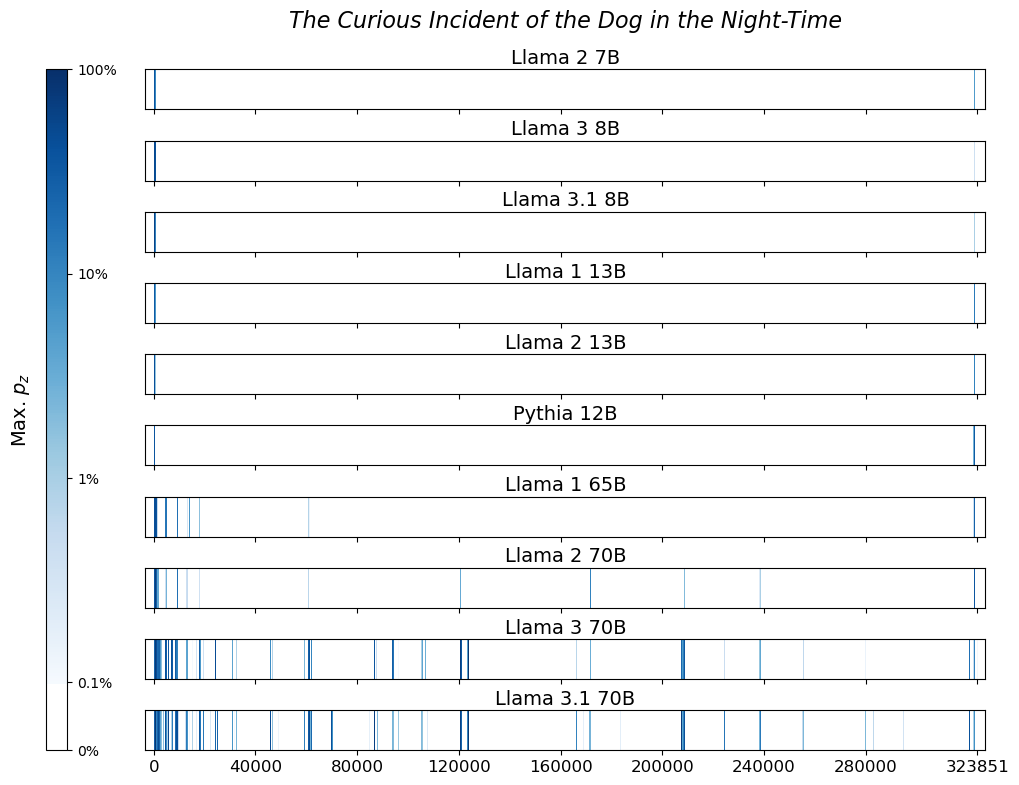}
    \includegraphics[width=\linewidth]{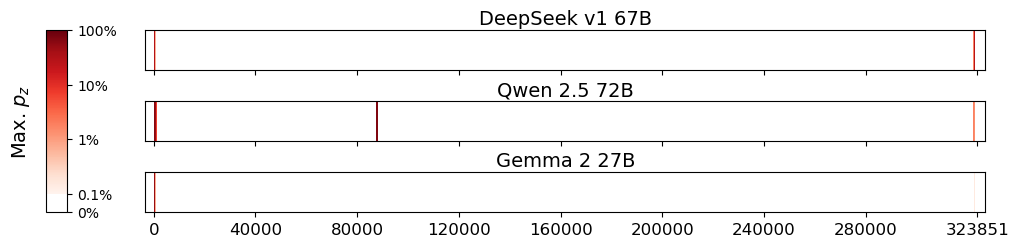}
    \includegraphics[width=\linewidth]{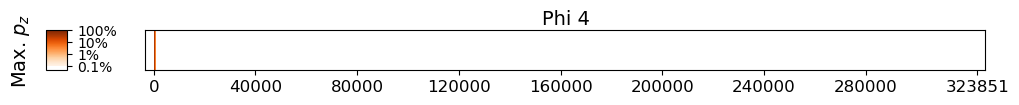}
  \end{minipage}
  \hfill
  \begin{minipage}[t]{0.45\textwidth}
    \centering
    \vspace{0cm}
    \includegraphics[width=\linewidth]{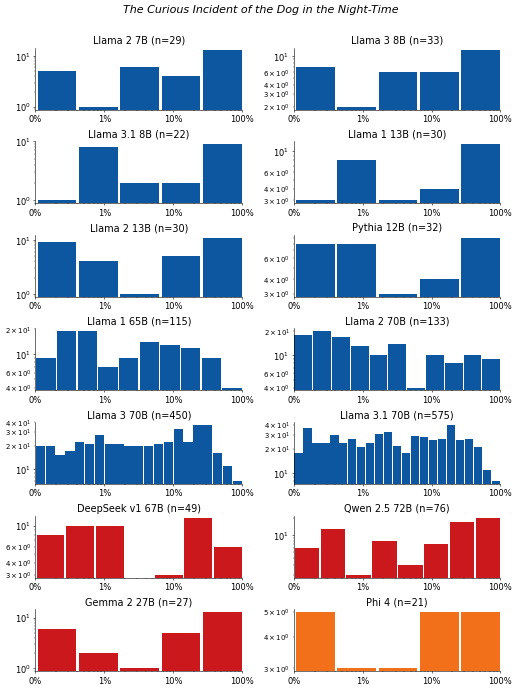}
  \end{minipage}
  \vspace{-.2cm}
  \caption{
    \textbf{\textit{The Curious Incident of the Dog in the Night-Time}, \citeauthor{The_Curious_Incident_of_the_Dog_in_the_Night-Time}.}
    For $14$ LLMs,
    (\textbf{left}) heatmaps for the sliding-window procedure and
    (\textbf{right}) corresponding distributions over suffix extraction probabilities
    ($\tau_\text{min}=0.1\%$).
  }
  \label{fig:slidingwindow:The_Curious_Incident_of_the_Dog_in_the_Night-Time}
\end{figure}
\FloatBarrier

\subsubsection{\textit{Migrations to Solitude}, \citeauthor{Migrations_to_Solitude}}\label{app:sec:sliding:Migrations_to_Solitude}
\vspace{-.2cm}
\begin{figure}[h]
  \centering
  \begin{minipage}[t]{0.53\textwidth}
    \centering
    \vspace{0cm}
    \includegraphics[width=\linewidth]{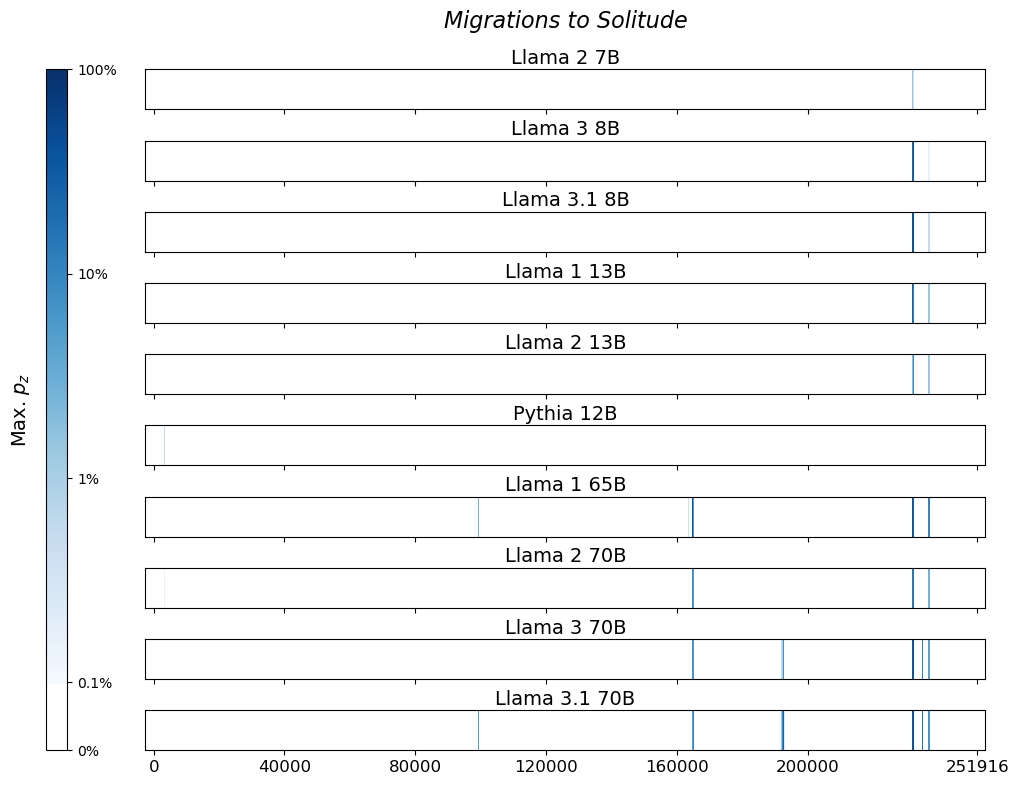}
    \includegraphics[width=\linewidth]{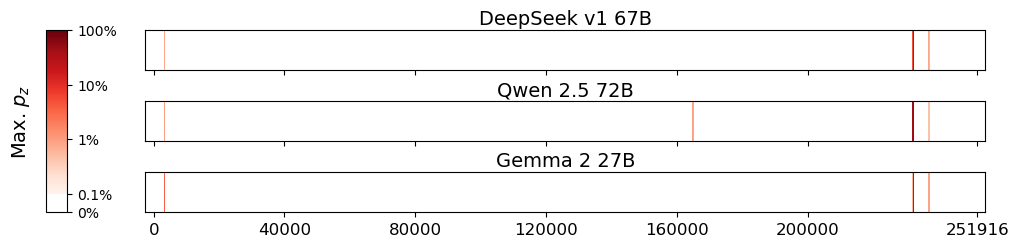}
    \includegraphics[width=\linewidth]{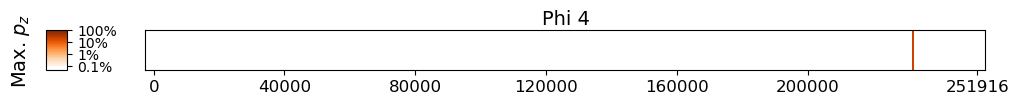}
  \end{minipage}
  \hfill
  \begin{minipage}[t]{0.45\textwidth}
    \centering
    \vspace{0cm}
    \includegraphics[width=\linewidth]{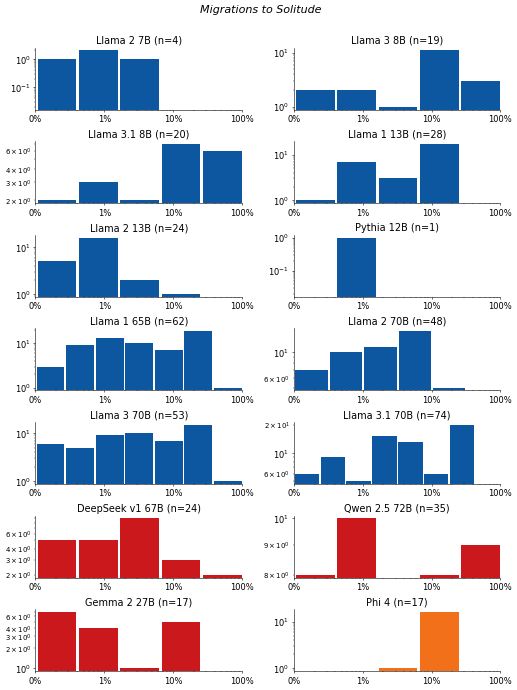}
  \end{minipage}
  \vspace{-.2cm}
  \caption{
    \textbf{\textit{Migrations to Solitude}, \citeauthor{Migrations_to_Solitude}.}
    For $14$ LLMs,
    (\textbf{left}) heatmaps for the sliding-window procedure and
    (\textbf{right}) corresponding distributions over suffix extraction probabilities
    ($\tau_\text{min}=0.1\%$).
  }
  \label{fig:slidingwindow:Migrations_to_Solitude}
\end{figure}
\FloatBarrier

\clearpage
\subsubsection{\textit{Uncommon Type}, \citeauthor{Uncommon_Type}}\label{app:sec:sliding:Uncommon_Type}
\vspace{-.2cm}
\begin{figure}[h]
  \centering
  \begin{minipage}[t]{0.53\textwidth}
    \centering
    \vspace{0cm}
    \includegraphics[width=\linewidth]{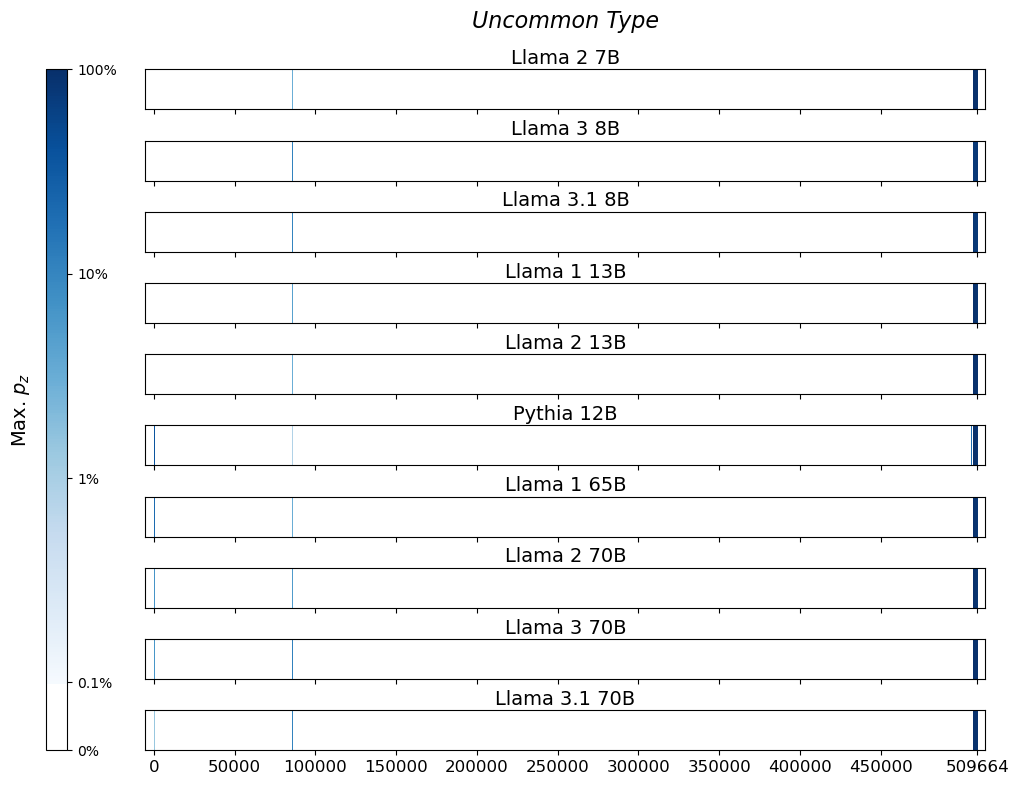}
    \includegraphics[width=\linewidth]{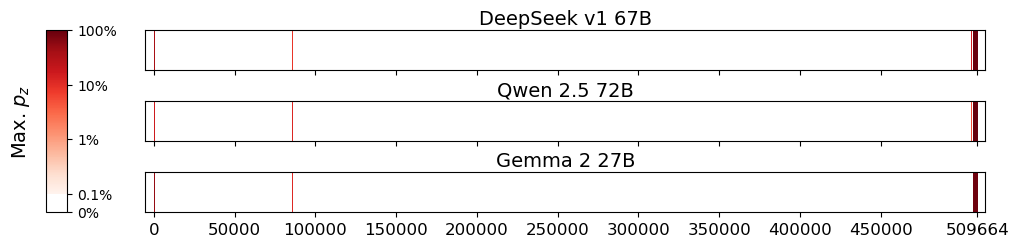}
    \includegraphics[width=\linewidth]{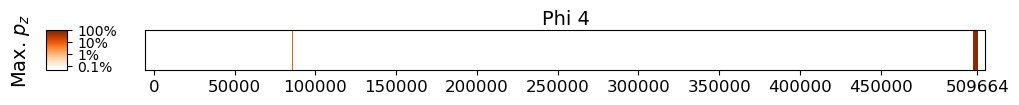}
  \end{minipage}
  \hfill
  \begin{minipage}[t]{0.45\textwidth}
    \centering
    \vspace{0cm}
    \includegraphics[width=\linewidth]{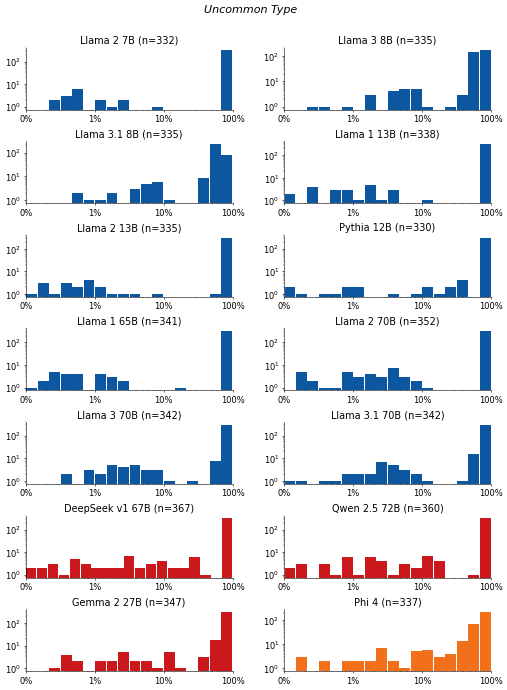}
  \end{minipage}
  \vspace{-.2cm}
  \caption{
    \textbf{\textit{Uncommon Type}, \citeauthor{Uncommon_Type}.}
    For $14$ LLMs,
    (\textbf{left}) heatmaps for the sliding-window procedure and
    (\textbf{right}) corresponding distributions over suffix extraction probabilities
    ($\tau_\text{min}=0.1\%$).
  }
  \label{fig:slidingwindow:Uncommon_Type}
\end{figure}
\FloatBarrier

\subsubsection{\textit{Buzz}, \citeauthor{Buzz}}\label{app:sec:sliding:Buzz}
\vspace{-.2cm}
\begin{figure}[h]
  \centering
  \begin{minipage}[t]{0.53\textwidth}
    \centering
    \vspace{0cm}
    \includegraphics[width=\linewidth]{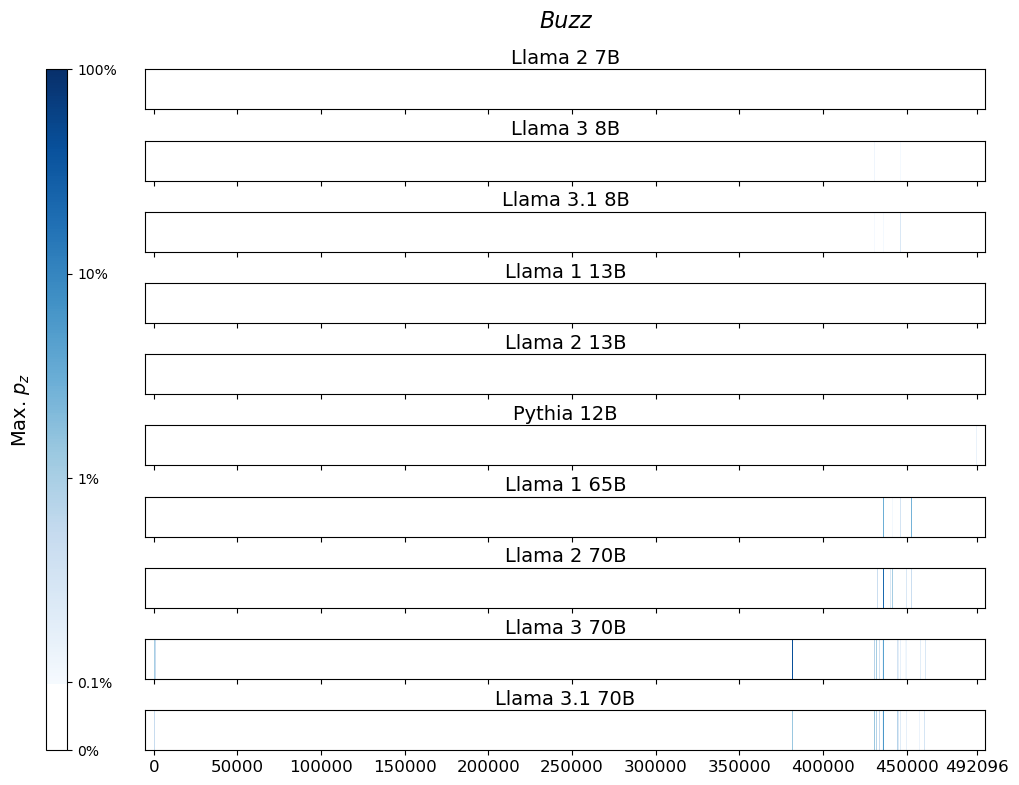}
    \includegraphics[width=\linewidth]{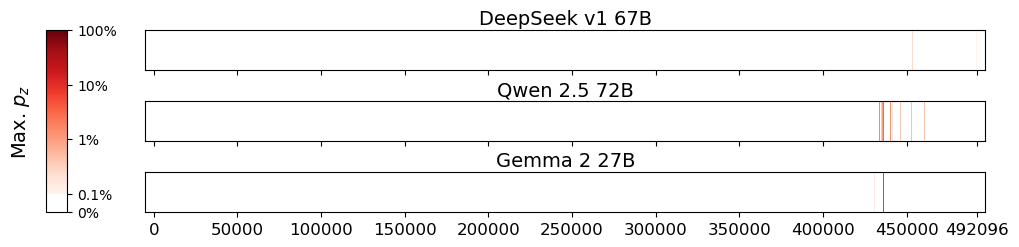}
    \includegraphics[width=\linewidth]{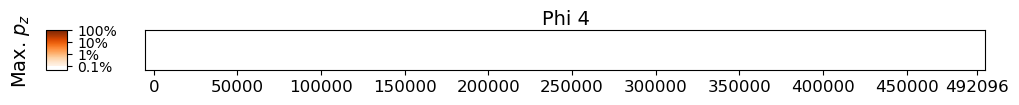}
  \end{minipage}
  \hfill
  \begin{minipage}[t]{0.45\textwidth}
    \centering
    \vspace{0cm}
    \includegraphics[width=\linewidth]{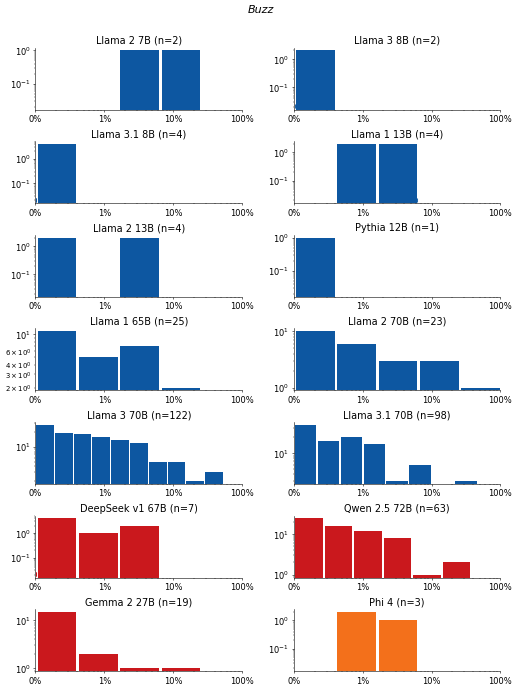}
  \end{minipage}
  \vspace{-.2cm}
  \caption{
    \textbf{\textit{Buzz}, \citeauthor{Buzz}.}
    For $14$ LLMs,
    (\textbf{left}) heatmaps for the sliding-window procedure and
    (\textbf{right}) corresponding distributions over suffix extraction probabilities
    ($\tau_\text{min}=0.1\%$).
  }
  \label{fig:slidingwindow:Buzz}
\end{figure}
\FloatBarrier

\clearpage
\subsubsection{\textit{Requiem for the Sun}, \citeauthor{Requiem_for_the_Sun}}\label{app:sec:sliding:Requiem_for_the_Sun}
\vspace{-.2cm}
\begin{figure}[h]
  \centering
  \begin{minipage}[t]{0.53\textwidth}
    \centering
    \vspace{0cm}
    \includegraphics[width=\linewidth]{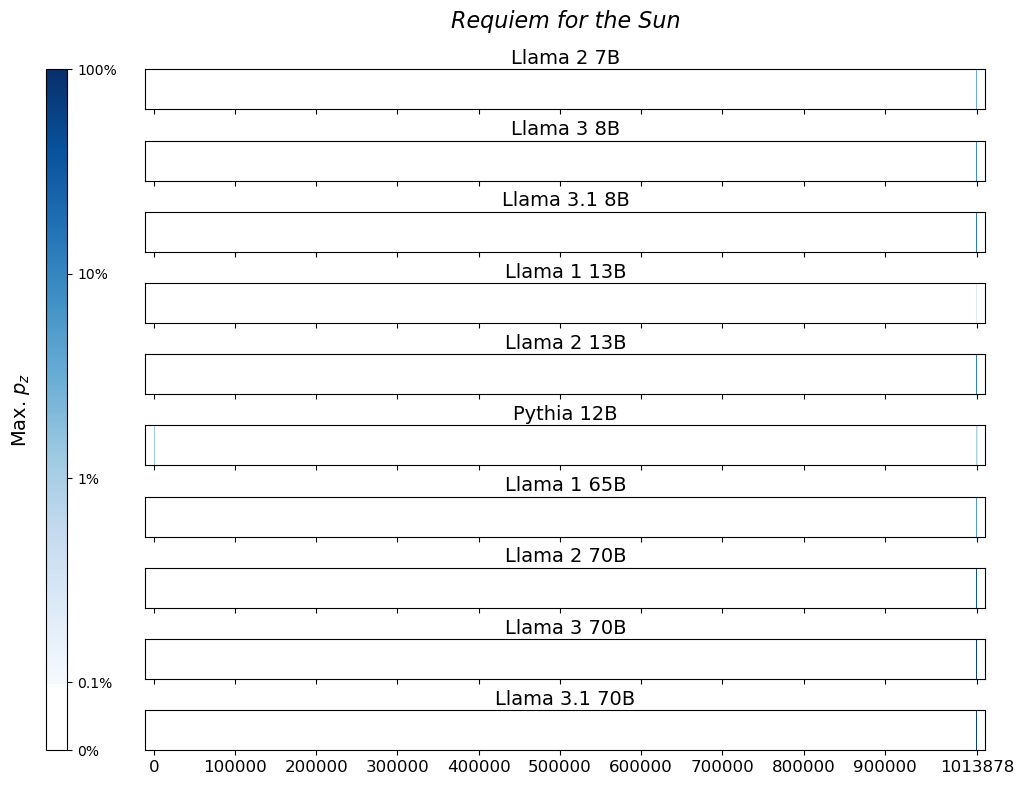}
    \includegraphics[width=\linewidth]{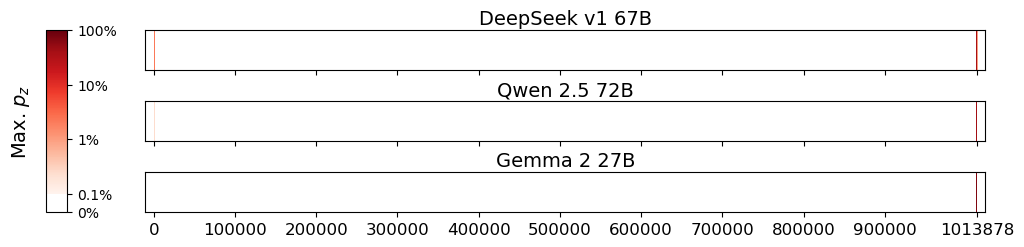}
    \includegraphics[width=\linewidth]{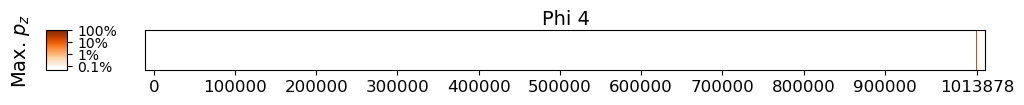}
  \end{minipage}
  \hfill
  \begin{minipage}[t]{0.45\textwidth}
    \centering
    \vspace{0cm}
    \includegraphics[width=\linewidth]{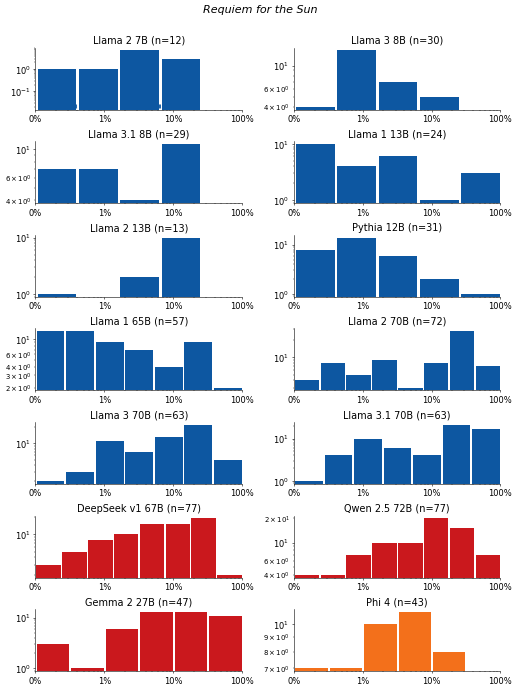}
  \end{minipage}
  \vspace{-.2cm}
  \caption{
    \textbf{\textit{Requiem for the Sun}, \citeauthor{Requiem_for_the_Sun}.}
    For $14$ LLMs,
    (\textbf{left}) heatmaps for the sliding-window procedure and
    (\textbf{right}) corresponding distributions over suffix extraction probabilities
    ($\tau_\text{min}=0.1\%$).
  }
  \label{fig:slidingwindow:Requiem_for_the_Sun}
\end{figure}
\FloatBarrier

\subsubsection{\textit{Catch-22}, \citeauthor{Catch-22}}\label{app:sec:sliding:Catch-22}
\vspace{-.2cm}
\begin{figure}[h]
  \centering
  \begin{minipage}[t]{0.53\textwidth}
    \centering
    \vspace{0cm}
    \includegraphics[width=\linewidth]{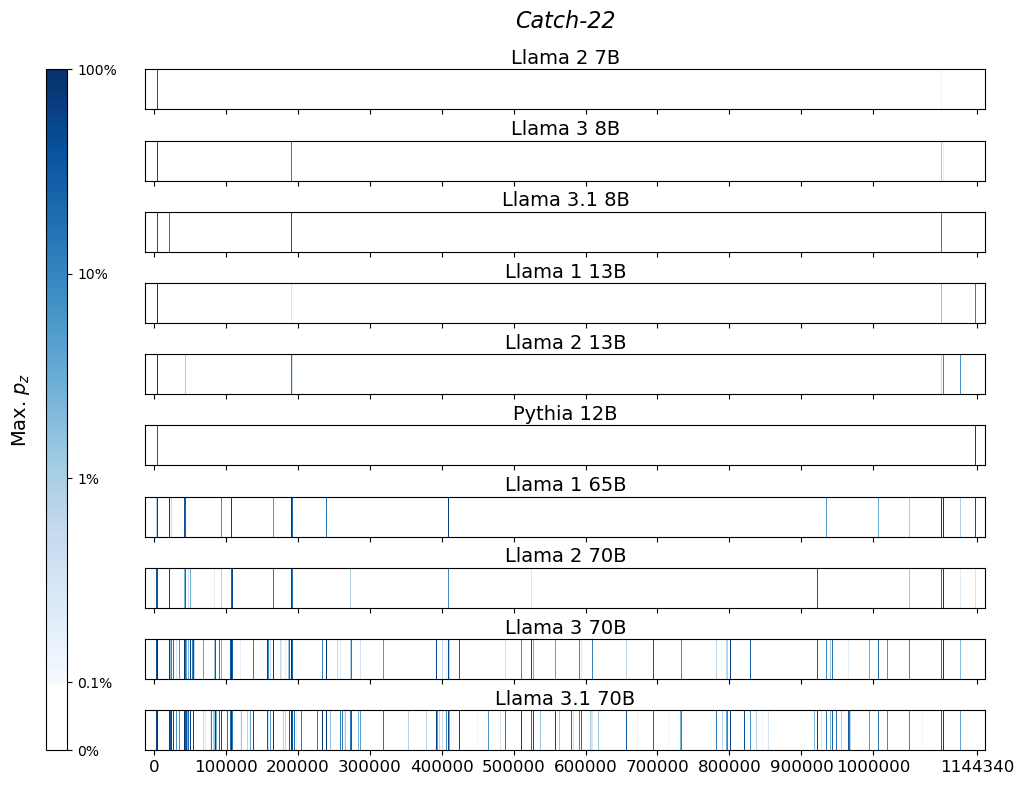}
    \includegraphics[width=\linewidth]{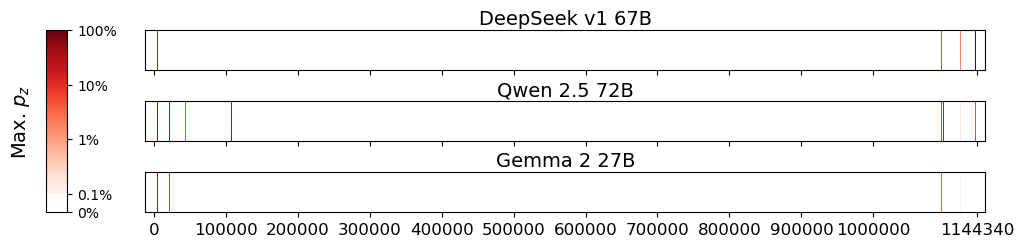}
    \includegraphics[width=\linewidth]{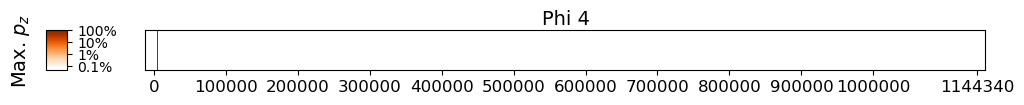}
  \end{minipage}
  \hfill
  \begin{minipage}[t]{0.45\textwidth}
    \centering
    \vspace{0cm}
    \includegraphics[width=\linewidth]{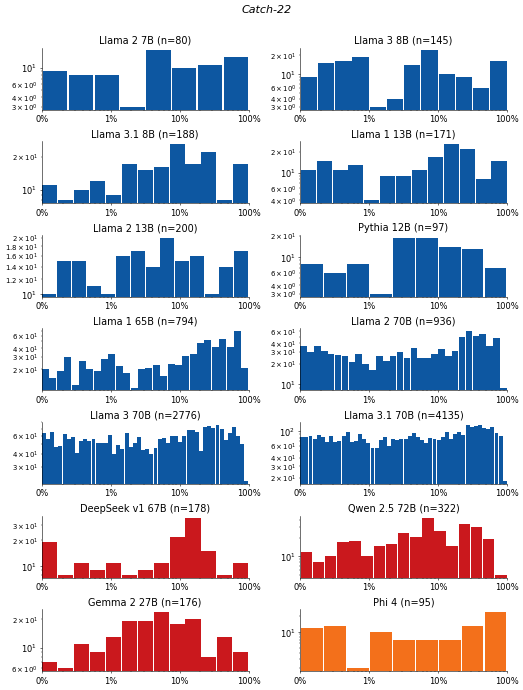}
  \end{minipage}
  \vspace{-.2cm}
  \caption{
    \textbf{\textit{Catch-22}, \citeauthor{Catch-22}.}
    For $14$ LLMs,
    (\textbf{left}) heatmaps for the sliding-window procedure and
    (\textbf{right}) corresponding distributions over suffix extraction probabilities
    ($\tau_\text{min}=0.1\%$).
  }
  \label{fig:slidingwindow:Catch-22}
\end{figure}
\FloatBarrier

\clearpage
\subsubsection{\textit{The Old Man and the Sea}, \citeauthor{The_Old_Man_and_the_Sea}}\label{app:sec:sliding:The_Old_Man_and_the_Sea}
\vspace{-.2cm}
\begin{figure}[h]
  \centering
  \begin{minipage}[t]{0.53\textwidth}
    \centering
    \vspace{0cm}
    \includegraphics[width=\linewidth]{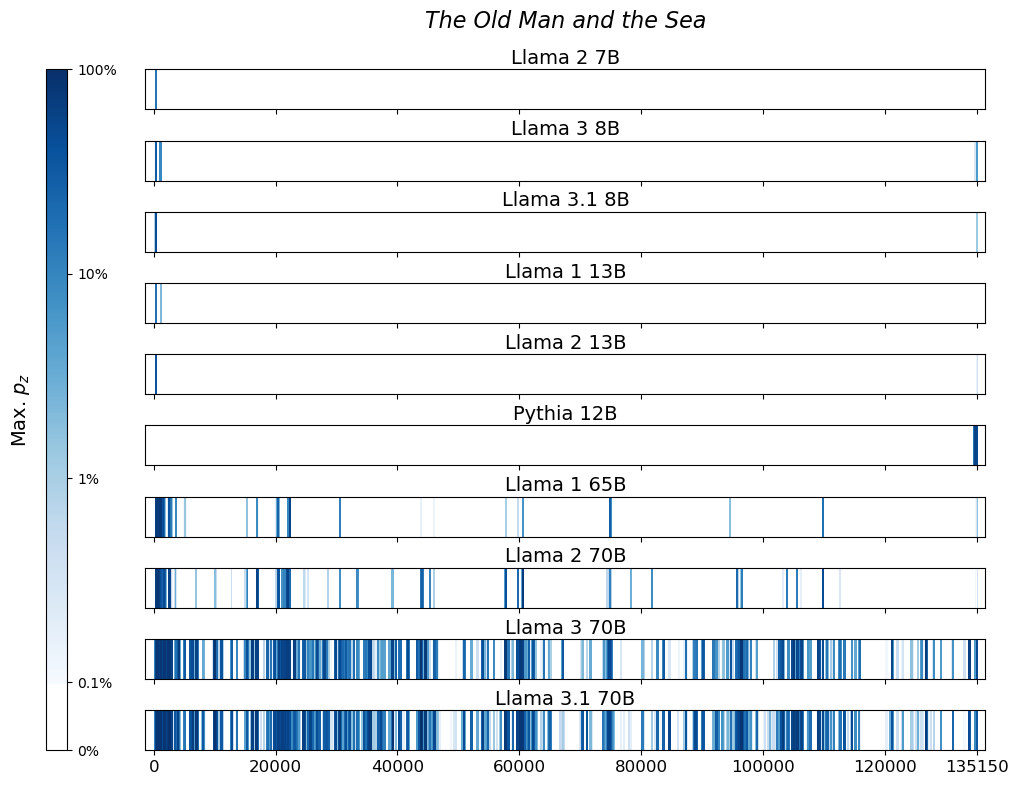}
    \includegraphics[width=\linewidth]{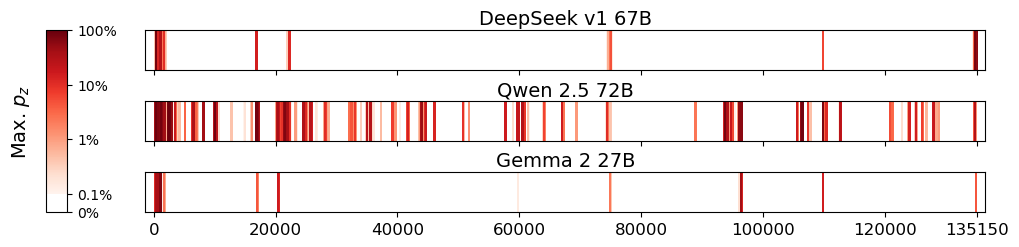}
    \includegraphics[width=\linewidth]{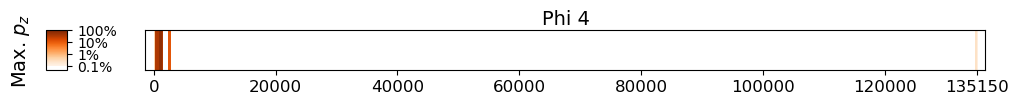}
  \end{minipage}
  \hfill
  \begin{minipage}[t]{0.45\textwidth}
    \centering
    \vspace{0cm}
    \includegraphics[width=\linewidth]{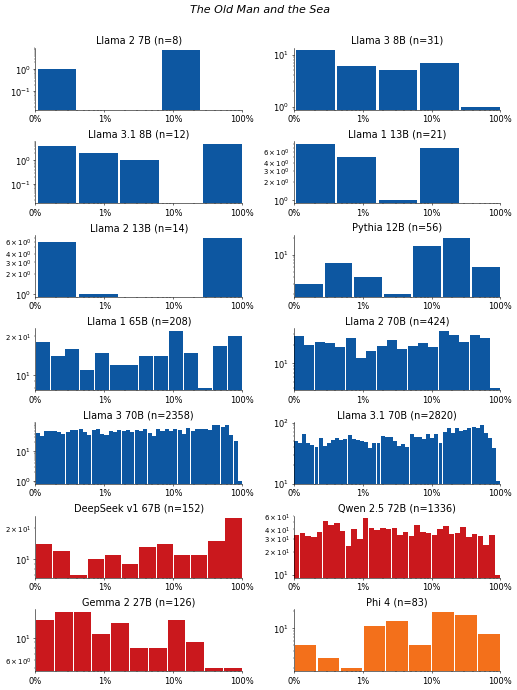}
  \end{minipage}
  \vspace{-.2cm}
  \caption{
    \textbf{\textit{The Old Man and the Sea}, \citeauthor{The_Old_Man_and_the_Sea}.}
    For $14$ LLMs,
    (\textbf{left}) heatmaps for the sliding-window procedure and
    (\textbf{right}) corresponding distributions over suffix extraction probabilities
    ($\tau_\text{min}=0.1\%$).
  }
  \label{fig:slidingwindow:The_Old_Man_and_the_Sea}
\end{figure}
\FloatBarrier

\subsubsection{\textit{Life on Air}, \citeauthor{Life_on_Air}}\label{app:sec:sliding:Life_on_Air}
\vspace{-.2cm}
\begin{figure}[h]
  \centering
  \begin{minipage}[t]{0.53\textwidth}
    \centering
    \vspace{0cm}
    \includegraphics[width=\linewidth]{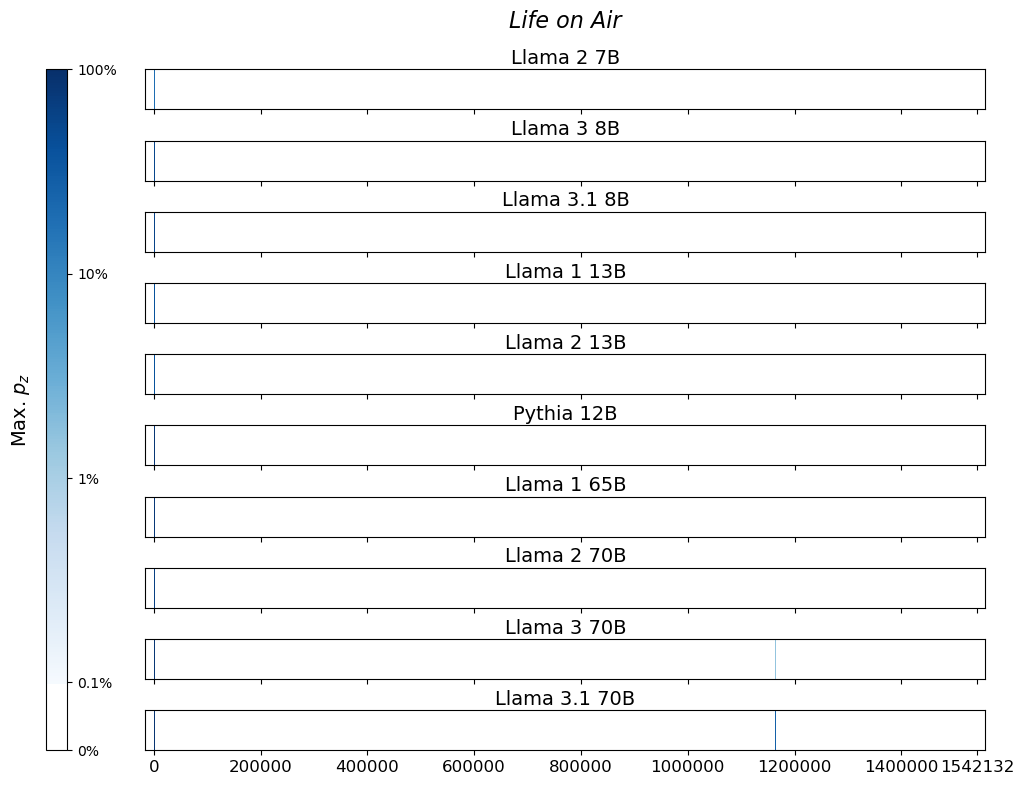}
    \includegraphics[width=\linewidth]{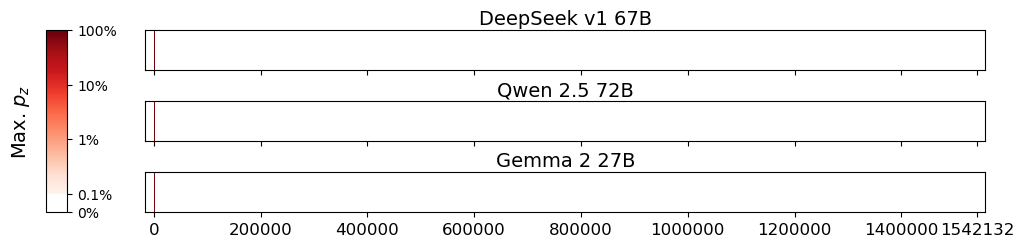}
    \includegraphics[width=\linewidth]{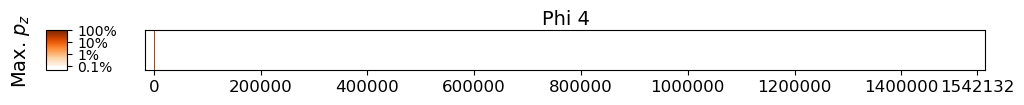}
  \end{minipage}
  \hfill
  \begin{minipage}[t]{0.45\textwidth}
    \centering
    \vspace{0cm}
    \includegraphics[width=\linewidth]{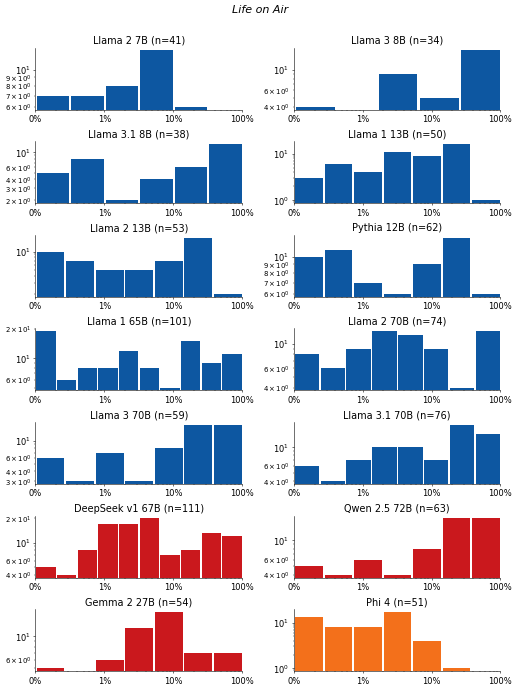}
  \end{minipage}
  \vspace{-.2cm}
  \caption{
    \textbf{\textit{Life on Air}, \citeauthor{Life_on_Air}.}
    For $14$ LLMs,
    (\textbf{left}) heatmaps for the sliding-window procedure and
    (\textbf{right}) corresponding distributions over suffix extraction probabilities
    ($\tau_\text{min}=0.1\%$).
  }
  \label{fig:slidingwindow:Life_on_Air}
\end{figure}
\FloatBarrier

\clearpage
\subsubsection{\textit{Great Hair Days}, \citeauthor{Great_Hair_Days}}\label{app:sec:sliding:Great_Hair_Days}
\vspace{-.2cm}
\begin{figure}[h]
  \centering
  \begin{minipage}[t]{0.53\textwidth}
    \centering
    \vspace{0cm}
    \includegraphics[width=\linewidth]{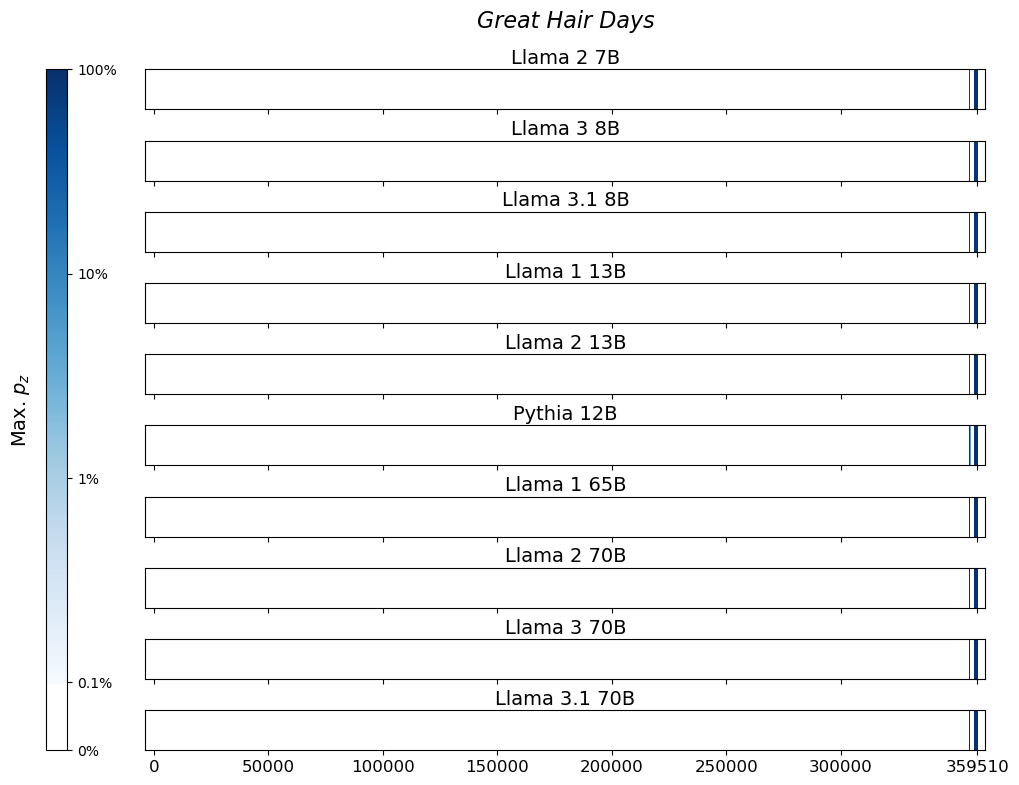}
    \includegraphics[width=\linewidth]{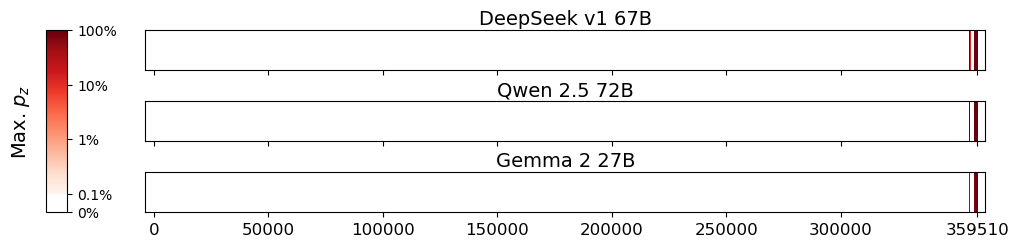}
    \includegraphics[width=\linewidth]{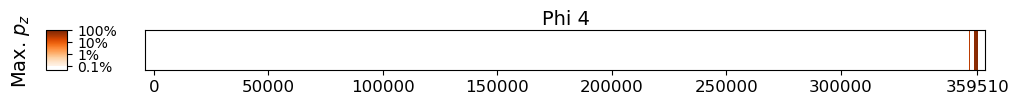}
  \end{minipage}
  \hfill
  \begin{minipage}[t]{0.45\textwidth}
    \centering
    \vspace{0cm}
    \includegraphics[width=\linewidth]{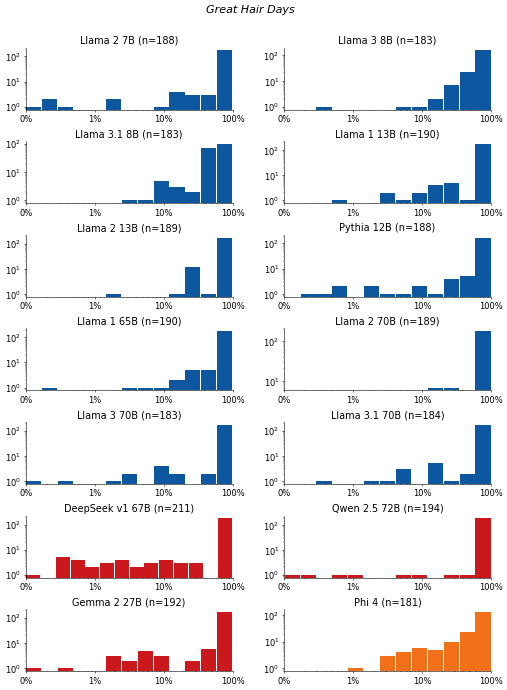}
  \end{minipage}
  \vspace{-.2cm}
  \caption{
    \textbf{\textit{Great Hair Days}, \citeauthor{Great_Hair_Days}.}
    For $14$ LLMs,
    (\textbf{left}) heatmaps for the sliding-window procedure and
    (\textbf{right}) corresponding distributions over suffix extraction probabilities
    ($\tau_\text{min}=0.1\%$).
  }
  \label{fig:slidingwindow:Great_Hair_Days}
\end{figure}
\FloatBarrier

\subsubsection{\textit{Graft}, \citeauthor{Graft}}\label{app:sec:sliding:Graft}
\vspace{-.2cm}
\begin{figure}[h]
  \centering
  \begin{minipage}[t]{0.53\textwidth}
    \centering
    \vspace{0cm}
    \includegraphics[width=\linewidth]{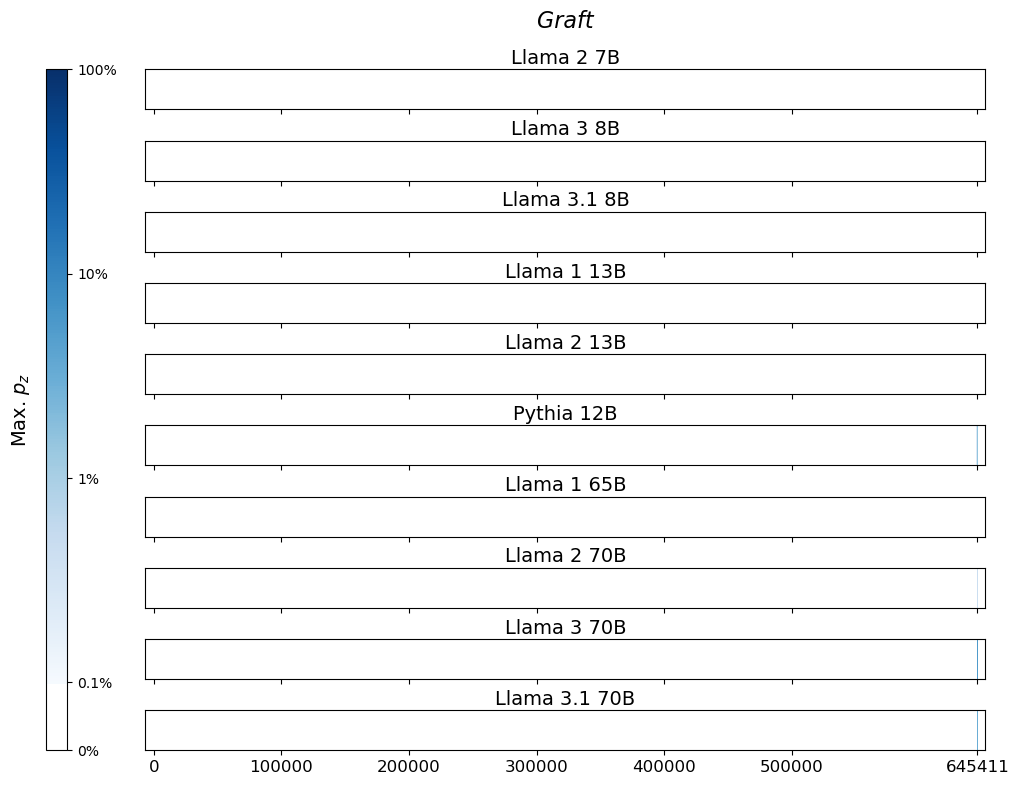}
    \includegraphics[width=\linewidth]{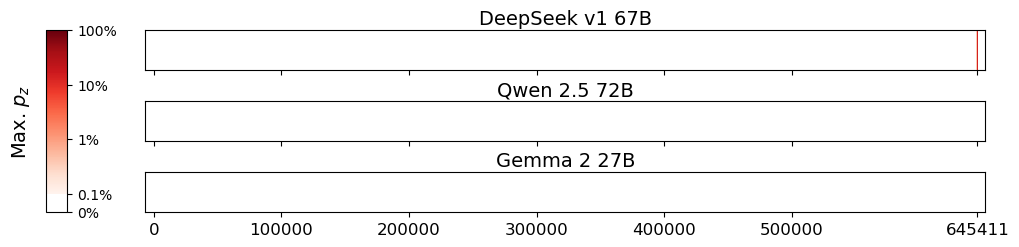}
    \includegraphics[width=\linewidth]{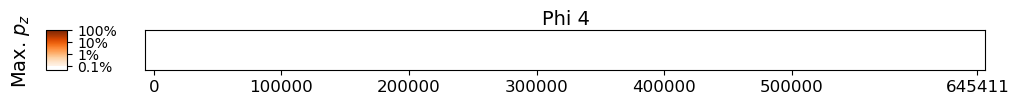}
  \end{minipage}
  \hfill
  \begin{minipage}[t]{0.45\textwidth}
    \centering
    \vspace{0cm}
    \includegraphics[width=\linewidth]{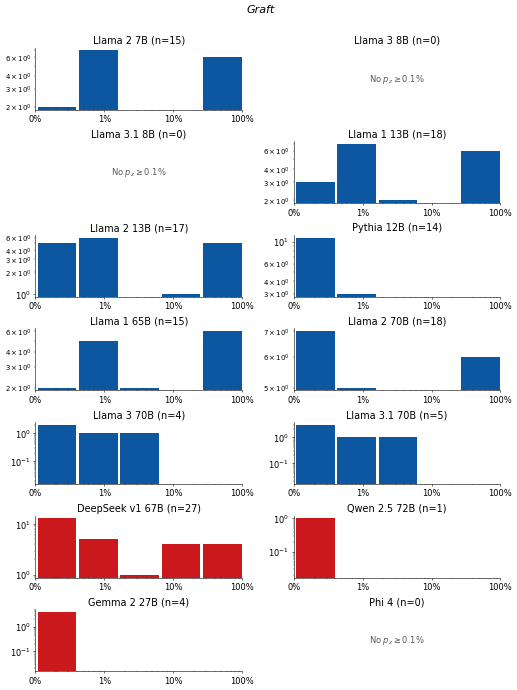}
  \end{minipage}
  \vspace{-.2cm}
  \caption{
    \textbf{\textit{Graft}, \citeauthor{Graft}.}
    For $14$ LLMs,
    (\textbf{left}) heatmaps for the sliding-window procedure and
    (\textbf{right}) corresponding distributions over suffix extraction probabilities
    ($\tau_\text{min}=0.1\%$).
  }
  \label{fig:slidingwindow:Graft}
\end{figure}
\FloatBarrier

\clearpage
\subsubsection{\textit{The Outsiders}, \citeauthor{The_Outsiders}}\label{app:sec:sliding:The_Outsiders}
\vspace{-.2cm}
\begin{figure}[h]
  \centering
  \begin{minipage}[t]{0.53\textwidth}
    \centering
    \vspace{0cm}
    \includegraphics[width=\linewidth]{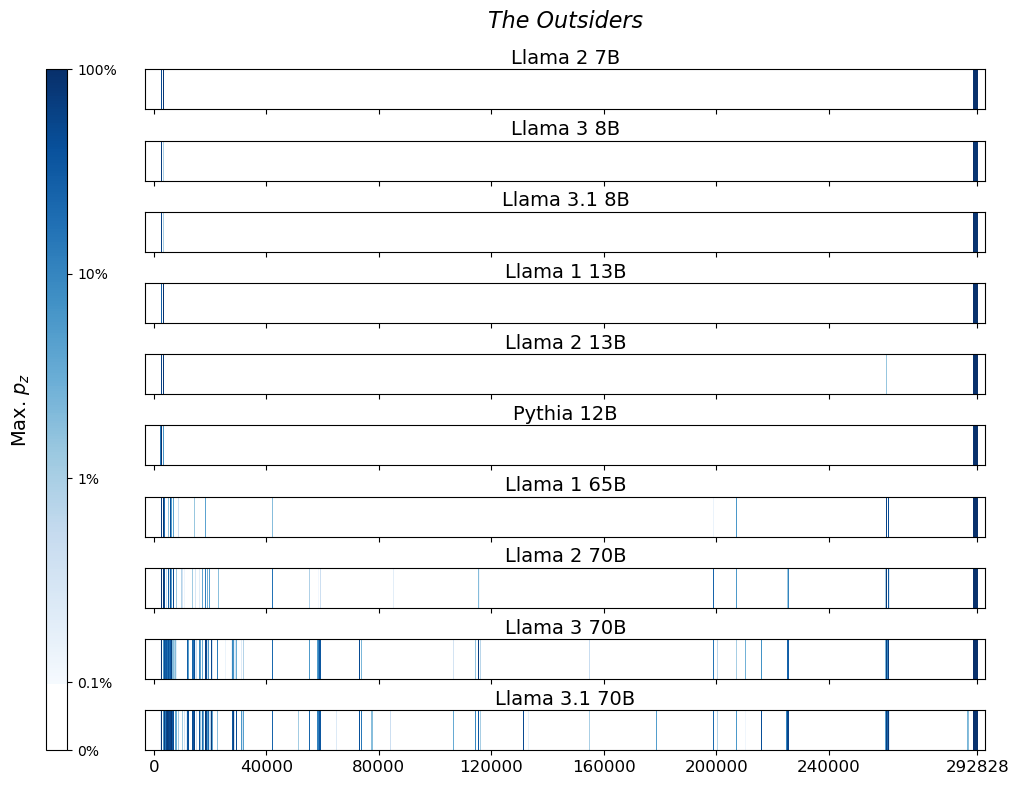}
    \includegraphics[width=\linewidth]{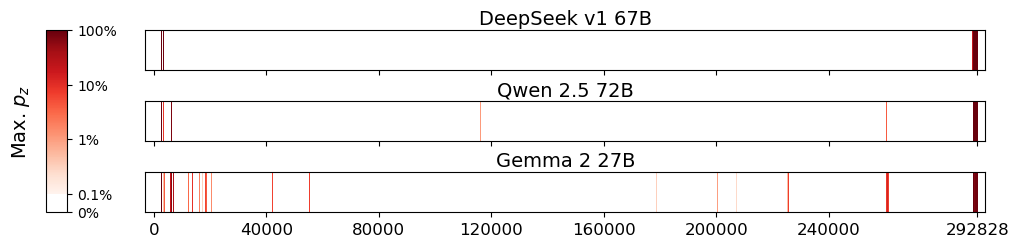}
    \includegraphics[width=\linewidth]{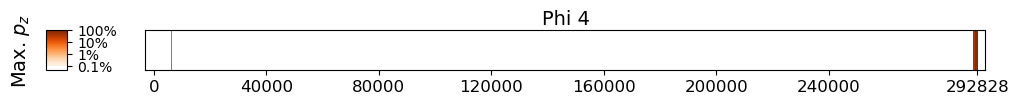}
  \end{minipage}
  \hfill
  \begin{minipage}[t]{0.45\textwidth}
    \centering
    \vspace{0cm}
    \includegraphics[width=\linewidth]{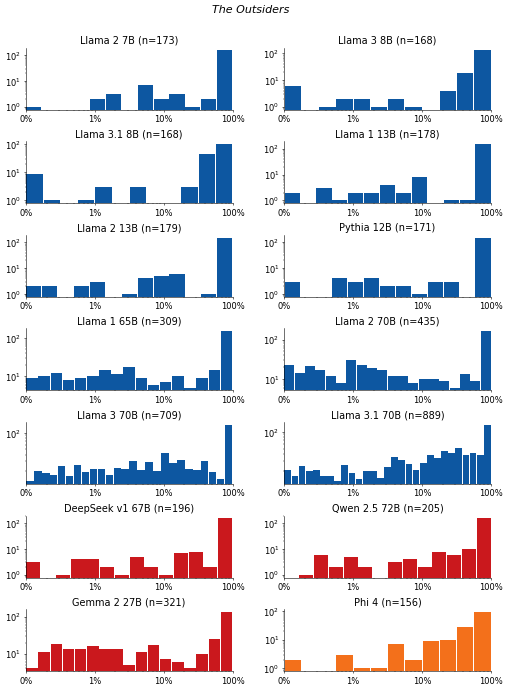}
  \end{minipage}
  \vspace{-.2cm}
  \caption{
    \textbf{\textit{The Outsiders}, \citeauthor{The_Outsiders}.}
    For $14$ LLMs,
    (\textbf{left}) heatmaps for the sliding-window procedure and
    (\textbf{right}) corresponding distributions over suffix extraction probabilities
    ($\tau_\text{min}=0.1\%$).
  }
  \label{fig:slidingwindow:The_Outsiders}
\end{figure}
\FloatBarrier

\subsubsection{\textit{The Second Summoning}, \citeauthor{The_Second_Summoning}}\label{app:sec:sliding:The_Second_Summoning}
\vspace{-.2cm}
\begin{figure}[h]
  \centering
  \begin{minipage}[t]{0.53\textwidth}
    \centering
    \vspace{0cm}
    \includegraphics[width=\linewidth]{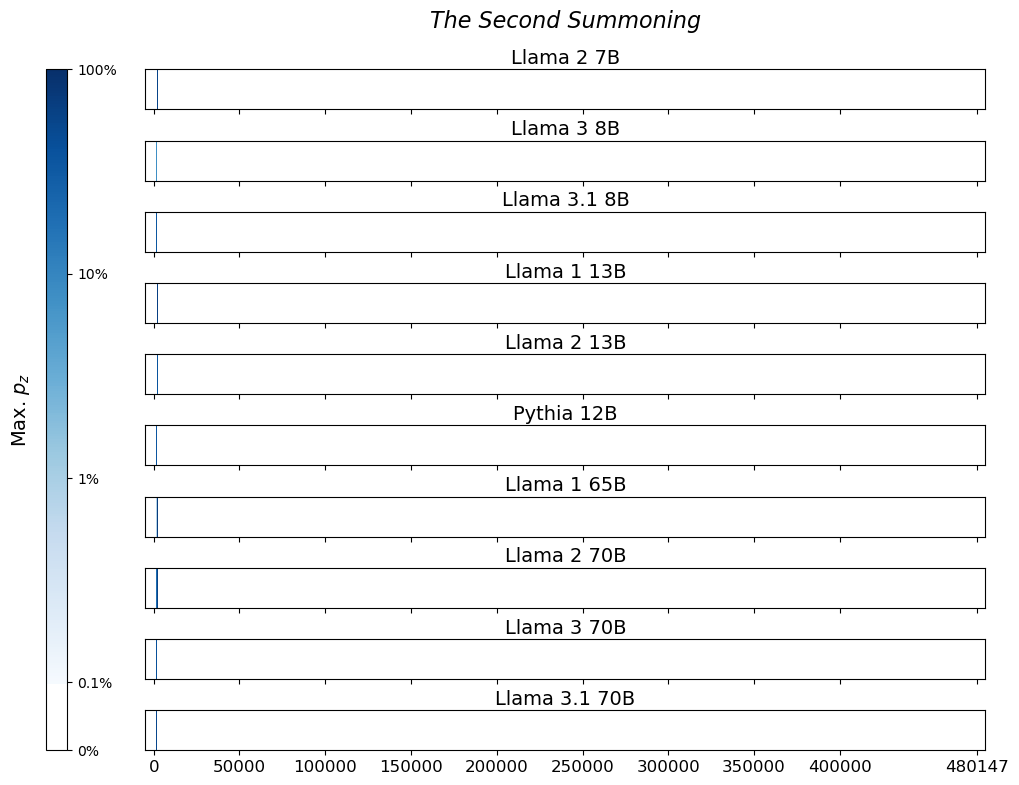}
    \includegraphics[width=\linewidth]{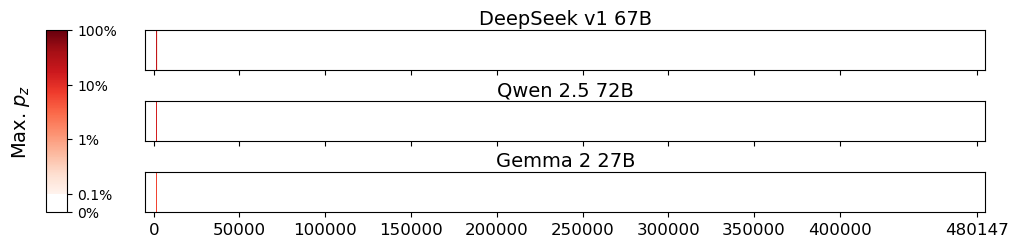}
    \includegraphics[width=\linewidth]{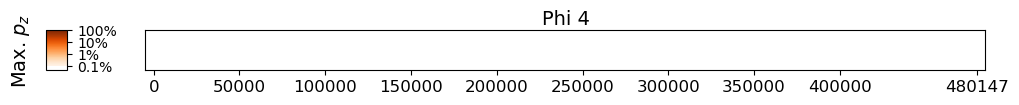}
  \end{minipage}
  \hfill
  \begin{minipage}[t]{0.45\textwidth}
    \centering
    \vspace{0cm}
    \includegraphics[width=\linewidth]{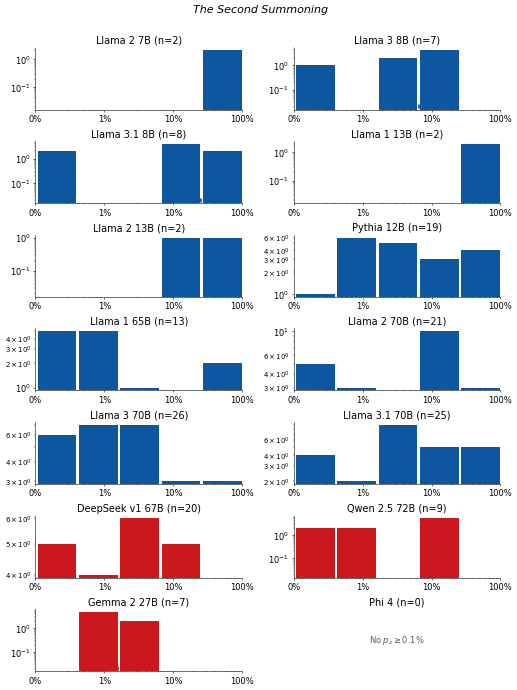}
  \end{minipage}
  \vspace{-.2cm}
  \caption{
    \textbf{\textit{The Second Summoning}, \citeauthor{The_Second_Summoning}.}
    For $14$ LLMs,
    (\textbf{left}) heatmaps for the sliding-window procedure and
    (\textbf{right}) corresponding distributions over suffix extraction probabilities
    ($\tau_\text{min}=0.1\%$).
  }
  \label{fig:slidingwindow:The_Second_Summoning}
\end{figure}
\FloatBarrier

\clearpage
\subsubsection{\textit{Selected Poems of Langston Hughes}, \citeauthor{Selected_Poems_of_Langston_Hughes}}\label{app:sec:sliding:Selected_Poems_of_Langston_Hughes}
\vspace{-.2cm}
\begin{figure}[h]
  \centering
  \begin{minipage}[t]{0.53\textwidth}
    \centering
    \vspace{0cm}
    \includegraphics[width=\linewidth]{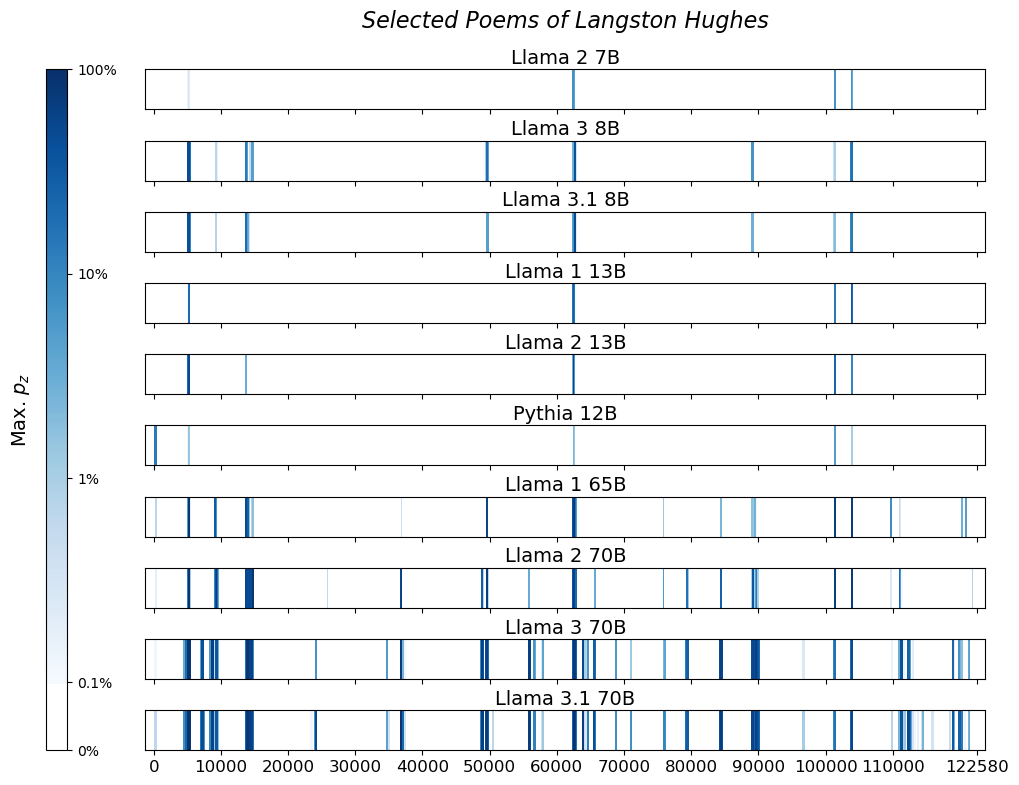}
    \includegraphics[width=\linewidth]{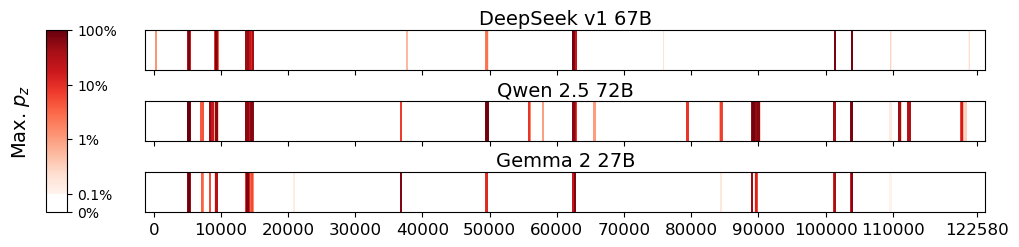}
    \includegraphics[width=\linewidth]{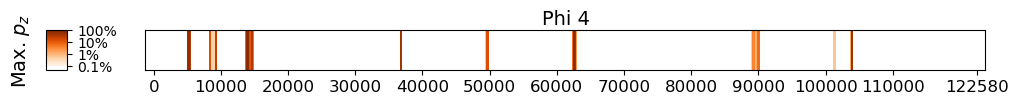}
  \end{minipage}
  \hfill
  \begin{minipage}[t]{0.45\textwidth}
    \centering
    \vspace{0cm}
    \includegraphics[width=\linewidth]{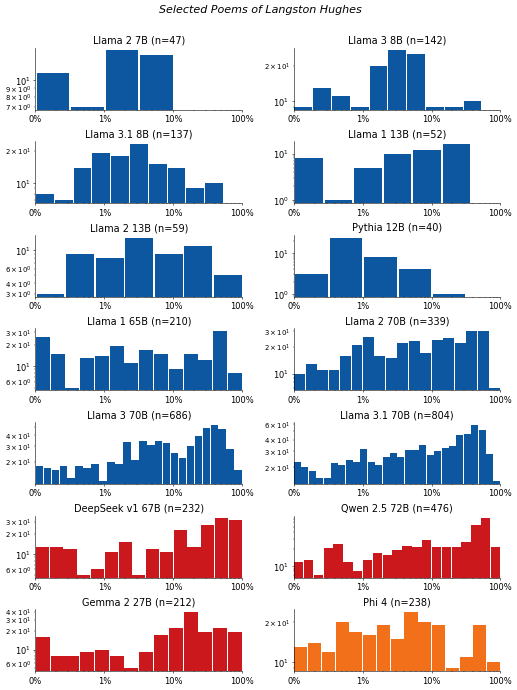}
  \end{minipage}
  \vspace{-.2cm}
  \caption{
    \textbf{\textit{Selected Poems of Langston Hughes}, \citeauthor{Selected_Poems_of_Langston_Hughes}.}
    For $14$ LLMs,
    (\textbf{left}) heatmaps for the sliding-window procedure and
    (\textbf{right}) corresponding distributions over suffix extraction probabilities
    ($\tau_\text{min}=0.1\%$).
  }
  \label{fig:slidingwindow:Selected_Poems_of_Langston_Hughes}
\end{figure}
\FloatBarrier

\subsubsection{\textit{M. Butterfly}, \citeauthor{M_Butterfly}}\label{app:sec:sliding:M_Butterfly}
\vspace{-.2cm}
\begin{figure}[h]
  \centering
  \begin{minipage}[t]{0.53\textwidth}
    \centering
    \vspace{0cm}
    \includegraphics[width=\linewidth]{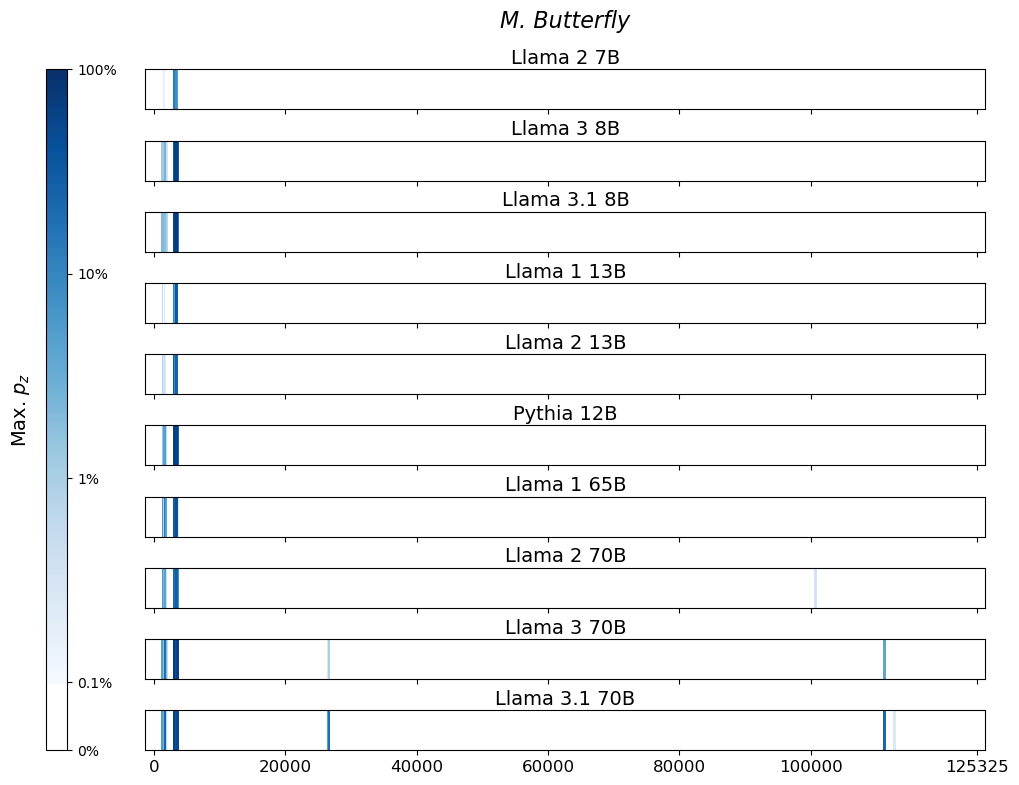}
    \includegraphics[width=\linewidth]{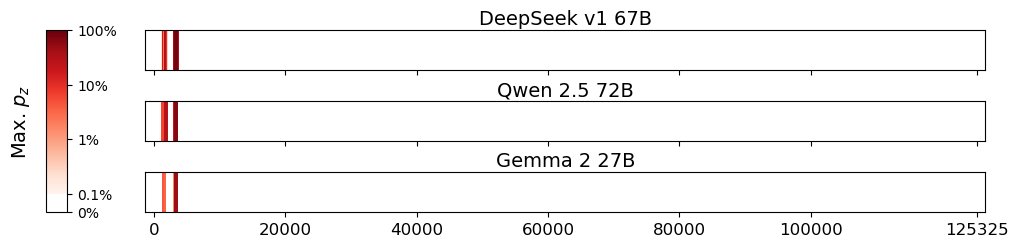}
    \includegraphics[width=\linewidth]{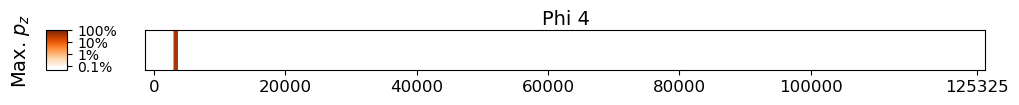}
  \end{minipage}
  \hfill
  \begin{minipage}[t]{0.45\textwidth}
    \centering
    \vspace{0cm}
    \includegraphics[width=\linewidth]{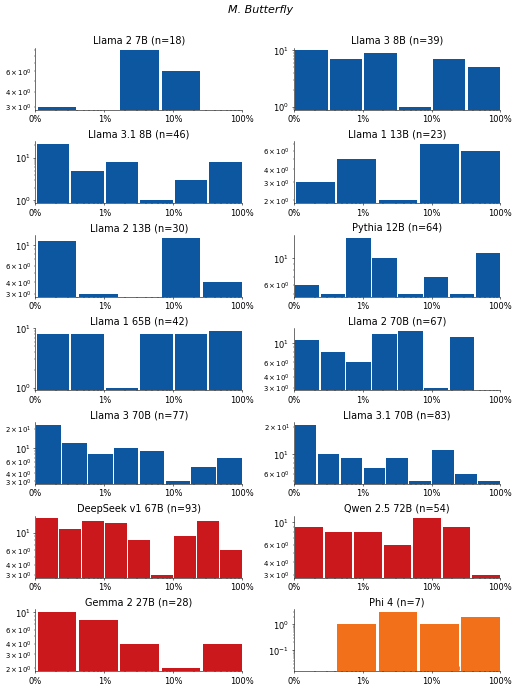}
  \end{minipage}
  \vspace{-.2cm}
  \caption{
    \textbf{\textit{M. Butterfly}, \citeauthor{M_Butterfly}.}
    For $14$ LLMs,
    (\textbf{left}) heatmaps for the sliding-window procedure and
    (\textbf{right}) corresponding distributions over suffix extraction probabilities
    ($\tau_\text{min}=0.1\%$).
  }
  \label{fig:slidingwindow:M_Butterfly}
\end{figure}
\FloatBarrier

\clearpage
\subsubsection{\textit{Building and Operating a Realistic Model Railway}, \citeauthor{Building_and_Operating_a_Realistic_Model_Railway}}\label{app:sec:sliding:Building_and_Operating_a_Realistic_Model_Railway}
\vspace{-.2cm}
\begin{figure}[h]
  \centering
  \begin{minipage}[t]{0.53\textwidth}
    \centering
    \vspace{0cm}
    \includegraphics[width=\linewidth]{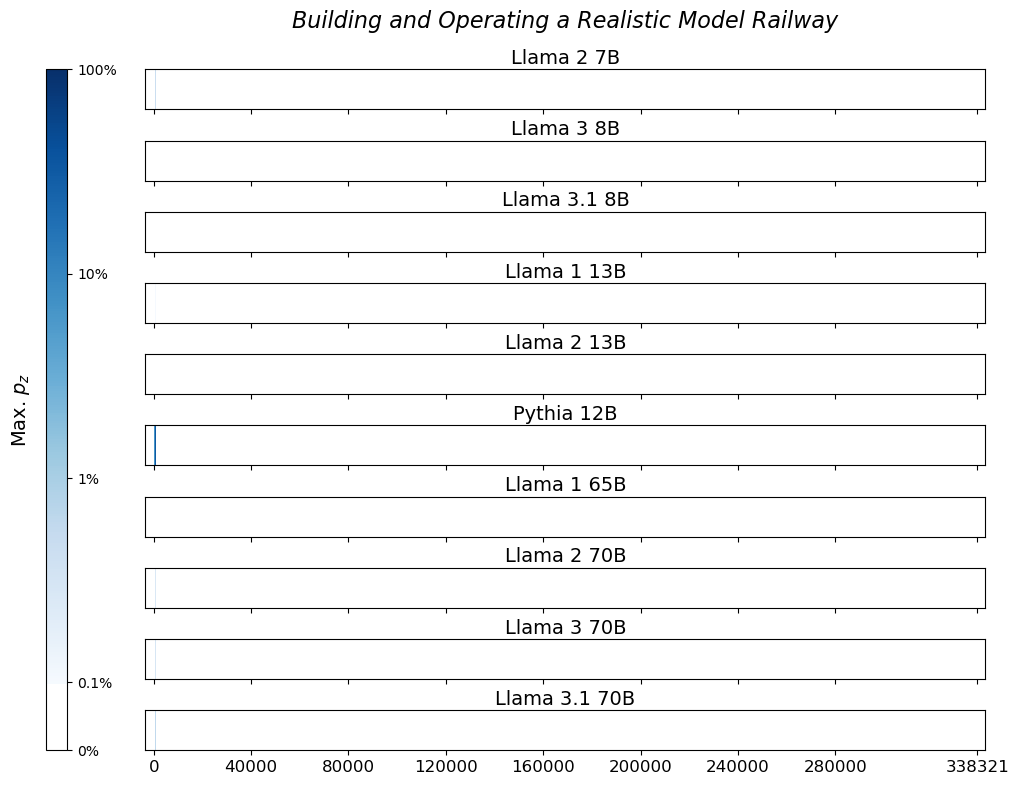}
    \includegraphics[width=\linewidth]{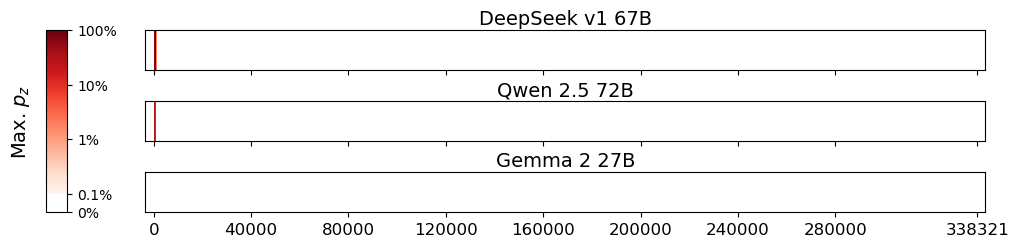}
    \includegraphics[width=\linewidth]{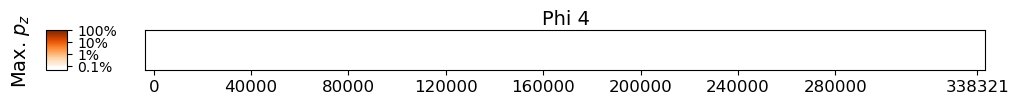}
  \end{minipage}
  \hfill
  \begin{minipage}[t]{0.45\textwidth}
    \centering
    \vspace{0cm}
    \includegraphics[width=\linewidth]{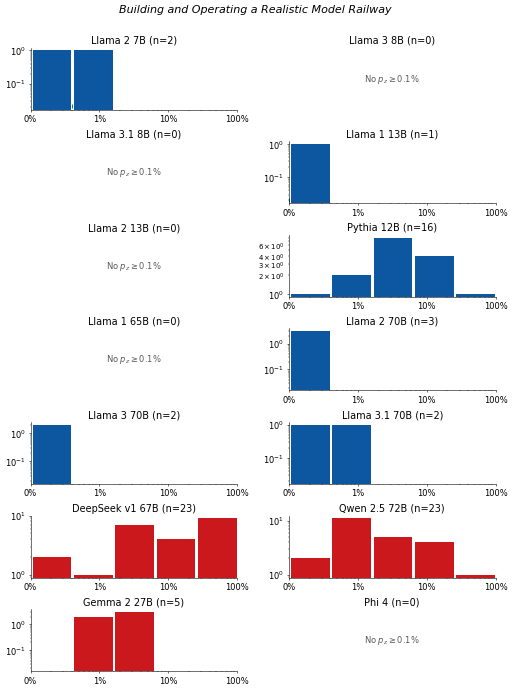}
  \end{minipage}
  \vspace{-.2cm}
  \caption{
    \textbf{\textit{Building and Operating a Realistic Model Railway}, \citeauthor{Building_and_Operating_a_Realistic_Model_Railway}.}
    For $14$ LLMs,
    (\textbf{left}) heatmaps for the sliding-window procedure and
    (\textbf{right}) corresponding distributions over suffix extraction probabilities
    ($\tau_\text{min}=0.1\%$).
  }
  \label{fig:slidingwindow:Building_and_Operating_a_Realistic_Model_Railway}
\end{figure}
\FloatBarrier

\subsubsection{\textit{All the Onions}, \citeauthor{All_the_Onions}}\label{app:sec:sliding:All_the_Onions}
\vspace{-.2cm}
\begin{figure}[h]
  \centering
  \begin{minipage}[t]{0.53\textwidth}
    \centering
    \vspace{0cm}
    \includegraphics[width=\linewidth]{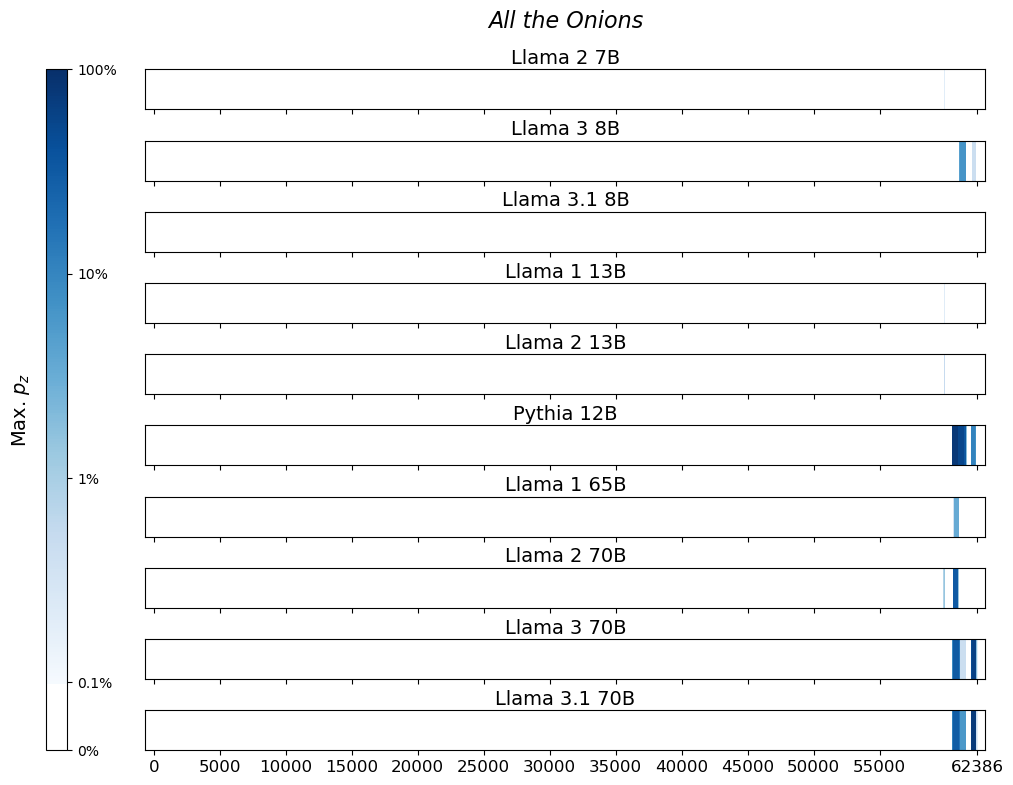}
    \includegraphics[width=\linewidth]{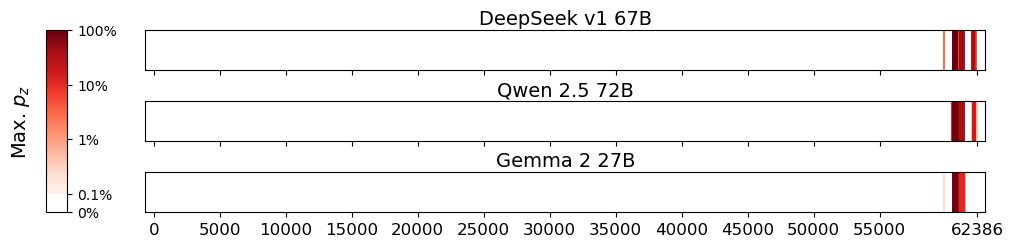}
    \includegraphics[width=\linewidth]{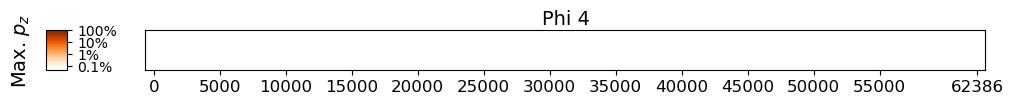}
  \end{minipage}
  \hfill
  \begin{minipage}[t]{0.45\textwidth}
    \centering
    \vspace{0cm}
    \includegraphics[width=\linewidth]{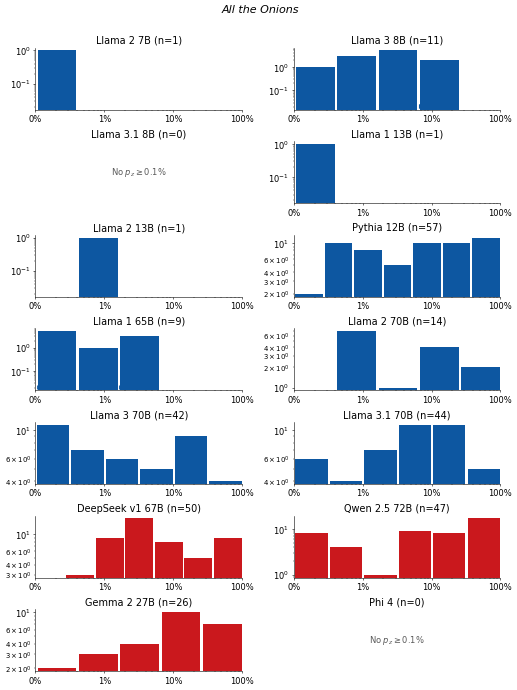}
  \end{minipage}
  \vspace{-.2cm}
  \caption{
    \textbf{\textit{All the Onions}, \citeauthor{All_the_Onions}.}
    For $14$ LLMs,
    (\textbf{left}) heatmaps for the sliding-window procedure and
    (\textbf{right}) corresponding distributions over suffix extraction probabilities
    ($\tau_\text{min}=0.1\%$).
  }
  \label{fig:slidingwindow:All_the_Onions}
\end{figure}
\FloatBarrier

\clearpage
\subsubsection{\textit{Fifty Shades of Grey}, \citeauthor{Fifty_Shades_of_Grey}}\label{app:sec:sliding:Fifty_Shades_of_Grey}
\vspace{-.2cm}
\begin{figure}[h]
  \centering
  \begin{minipage}[t]{0.53\textwidth}
    \centering
    \vspace{0cm}
    \includegraphics[width=\linewidth]{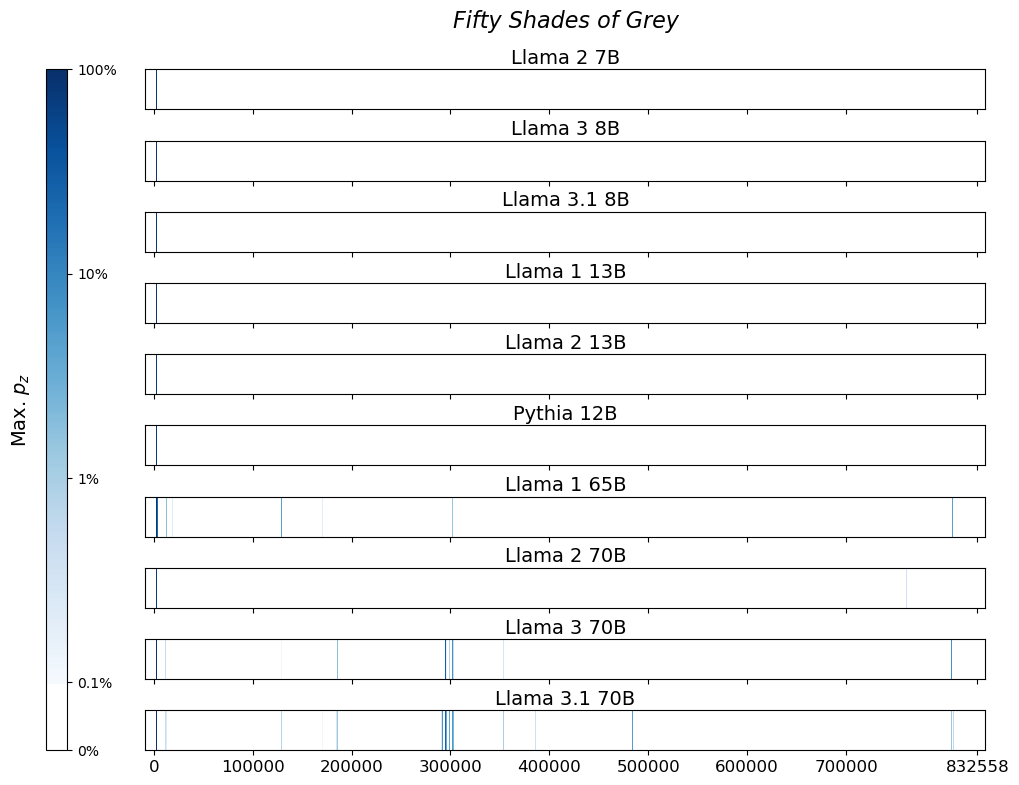}
    \includegraphics[width=\linewidth]{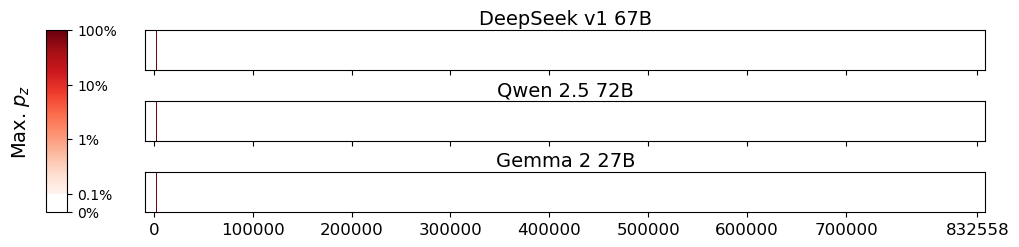}
    \includegraphics[width=\linewidth]{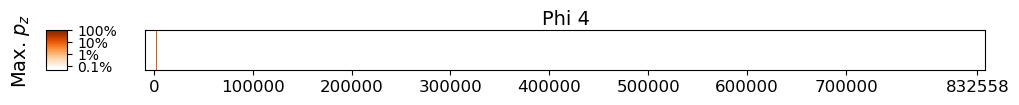}
  \end{minipage}
  \hfill
  \begin{minipage}[t]{0.45\textwidth}
    \centering
    \vspace{0cm}
    \includegraphics[width=\linewidth]{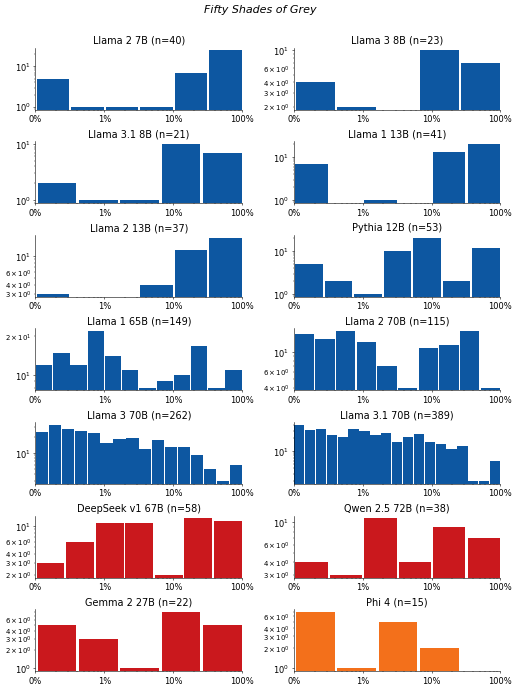}
  \end{minipage}
  \vspace{-.2cm}
  \caption{
    \textbf{\textit{Fifty Shades of Grey}, \citeauthor{Fifty_Shades_of_Grey}.}
    For $14$ LLMs,
    (\textbf{left}) heatmaps for the sliding-window procedure and
    (\textbf{right}) corresponding distributions over suffix extraction probabilities
    ($\tau_\text{min}=0.1\%$).
  }
  \label{fig:slidingwindow:Fifty_Shades_of_Grey}
\end{figure}
\FloatBarrier

\subsubsection{\textit{The Stone Sky}, \citeauthor{The_Stone_Sky}}\label{app:sec:sliding:The_Stone_Sky}
\vspace{-.2cm}
\begin{figure}[h]
  \centering
  \begin{minipage}[t]{0.53\textwidth}
    \centering
    \vspace{0cm}
    \includegraphics[width=\linewidth]{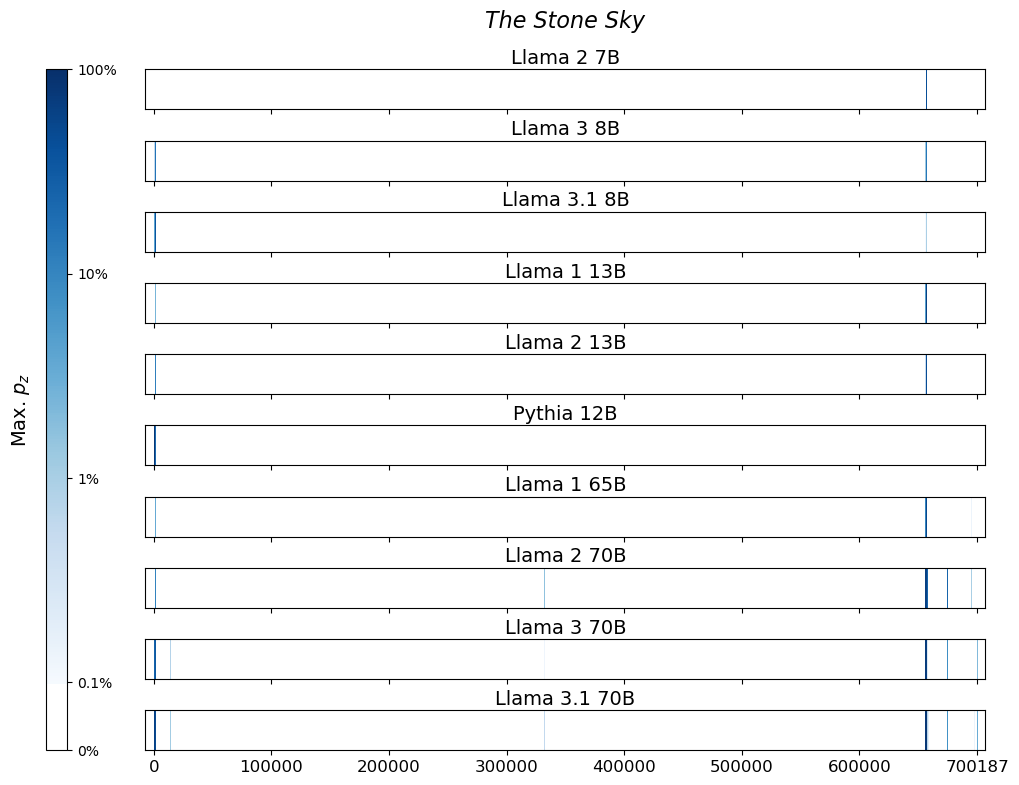}
    \includegraphics[width=\linewidth]{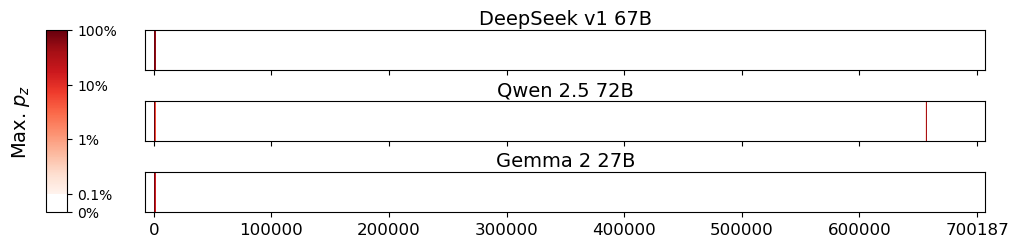}
    \includegraphics[width=\linewidth]{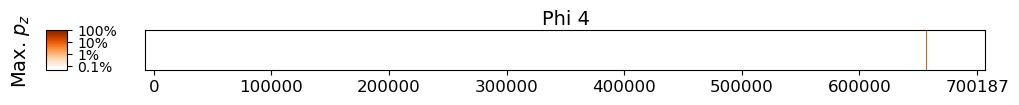}
  \end{minipage}
  \hfill
  \begin{minipage}[t]{0.45\textwidth}
    \centering
    \vspace{0cm}
    \includegraphics[width=\linewidth]{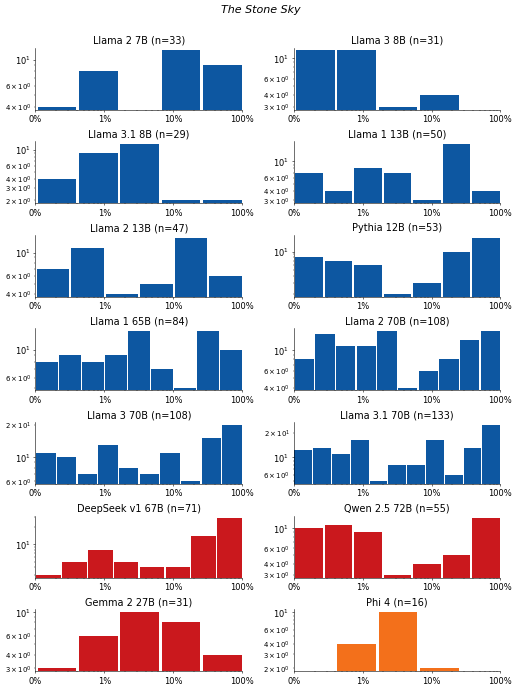}
  \end{minipage}
  \vspace{-.2cm}
  \caption{
    \textbf{\textit{The Stone Sky}, \citeauthor{The_Stone_Sky}.}
    For $14$ LLMs,
    (\textbf{left}) heatmaps for the sliding-window procedure and
    (\textbf{right}) corresponding distributions over suffix extraction probabilities
    ($\tau_\text{min}=0.1\%$).
  }
  \label{fig:slidingwindow:The_Stone_Sky}
\end{figure}
\FloatBarrier

\clearpage
\subsubsection{\textit{Ulysses}, \citeauthor{Ulysses}}\label{app:sec:sliding:Ulysses}
\vspace{-.2cm}
\begin{figure}[h]
  \centering
  \begin{minipage}[t]{0.53\textwidth}
    \centering
    \vspace{0cm}
    \includegraphics[width=\linewidth]{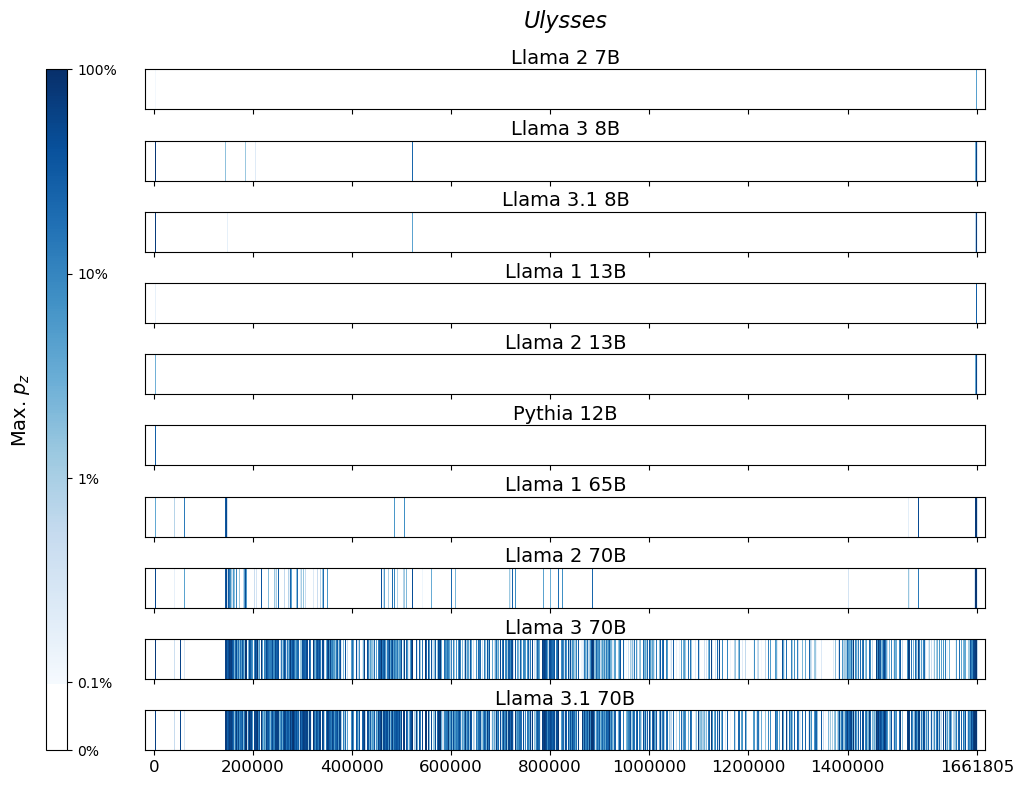}
    \includegraphics[width=\linewidth]{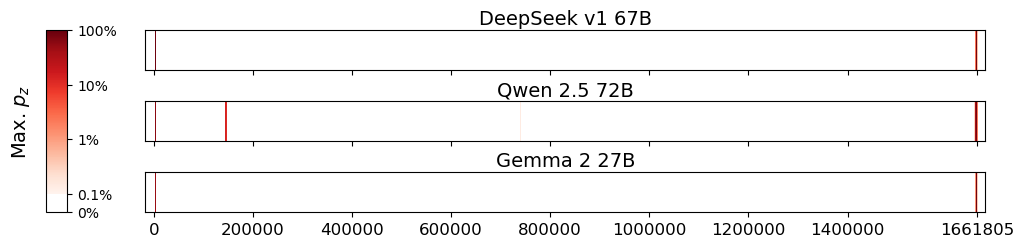}
    \includegraphics[width=\linewidth]{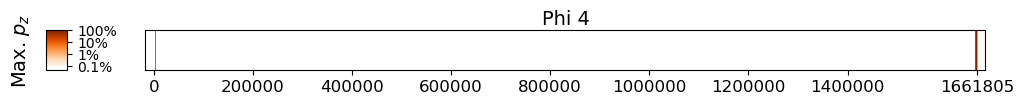}
  \end{minipage}
  \hfill
  \begin{minipage}[t]{0.45\textwidth}
    \centering
    \vspace{0cm}
    \includegraphics[width=\linewidth]{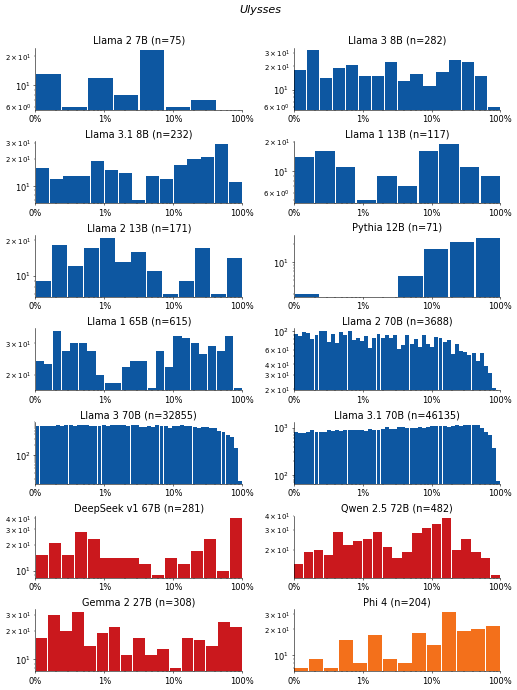}
  \end{minipage}
  \vspace{-.2cm}
  \caption{
    \textbf{\textit{Ulysses}, \citeauthor{Ulysses}.}
    For $14$ LLMs,
    (\textbf{left}) heatmaps for the sliding-window procedure and
    (\textbf{right}) corresponding distributions over suffix extraction probabilities
    ($\tau_\text{min}=0.1\%$).
  }
  \label{fig:slidingwindow:Ulysses}
\end{figure}
\FloatBarrier

\subsubsection{\textit{Sandman Slim}, \citeauthor{Sandman_Slim}}\label{app:sec:sliding:Sandman_Slim}
\vspace{-.2cm}
\begin{figure}[h]
  \centering
  \begin{minipage}[t]{0.53\textwidth}
    \centering
    \vspace{0cm}
    \includegraphics[width=\linewidth]{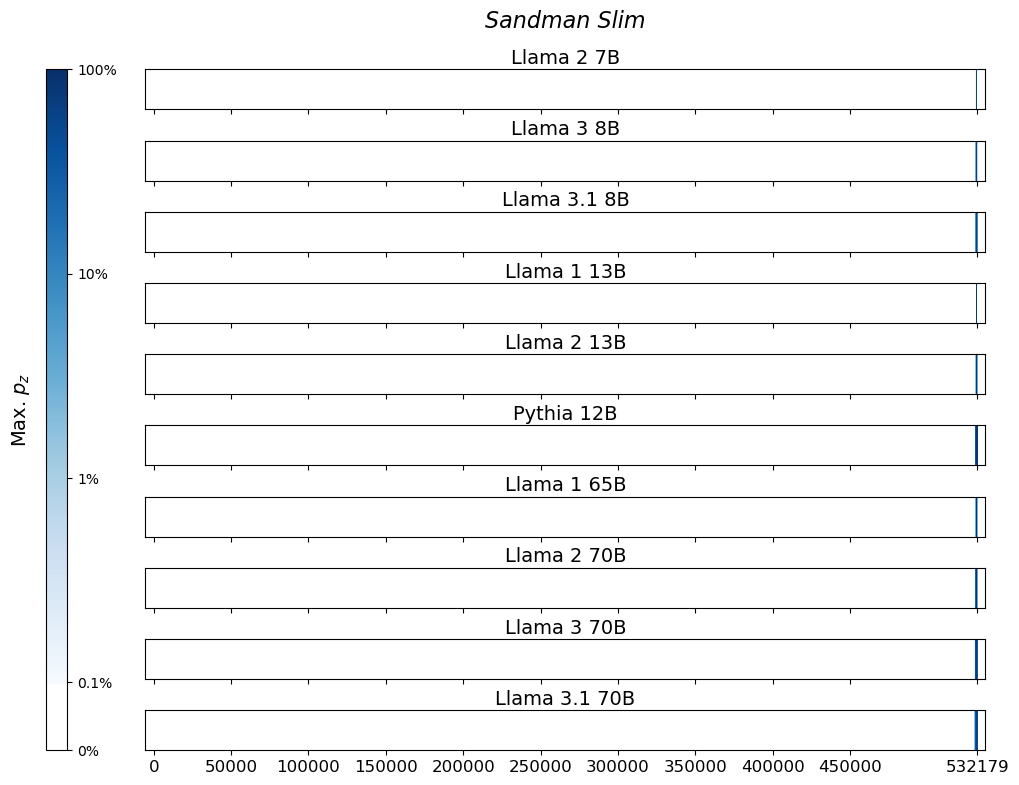}
    \includegraphics[width=\linewidth]{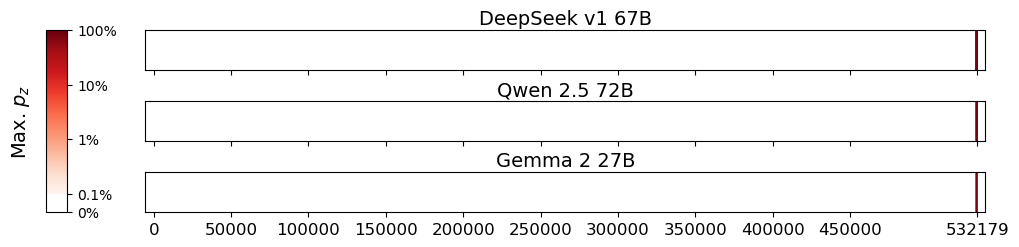}
    \includegraphics[width=\linewidth]{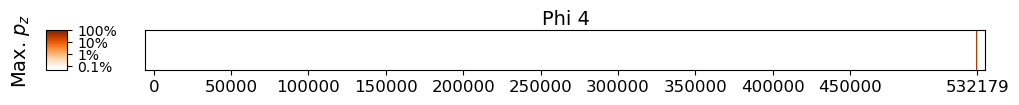}
  \end{minipage}
  \hfill
  \begin{minipage}[t]{0.45\textwidth}
    \centering
    \vspace{0cm}
    \includegraphics[width=\linewidth]{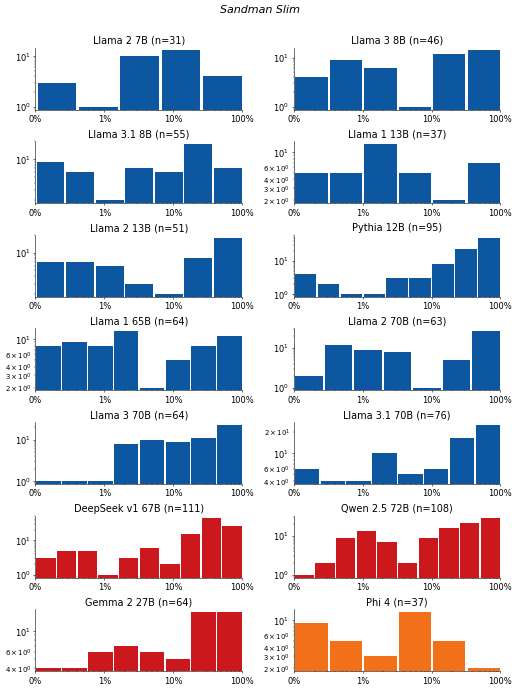}
  \end{minipage}
  \vspace{-.2cm}
  \caption{
    \textbf{\textit{Sandman Slim}, \citeauthor{Sandman_Slim}.}
    For $14$ LLMs,
    (\textbf{left}) heatmaps for the sliding-window procedure and
    (\textbf{right}) corresponding distributions over suffix extraction probabilities
    ($\tau_\text{min}=0.1\%$).
  }
  \label{fig:slidingwindow:Sandman_Slim}
\end{figure}
\FloatBarrier

\clearpage
\subsubsection{\textit{Ethnography after Antiquity}, \citeauthor{Ethnography_after_Antiquity}}\label{app:sec:sliding:Ethnography_after_Antiquity}
\vspace{-.2cm}
\begin{figure}[h]
  \centering
  \begin{minipage}[t]{0.53\textwidth}
    \centering
    \vspace{0cm}
    \includegraphics[width=\linewidth]{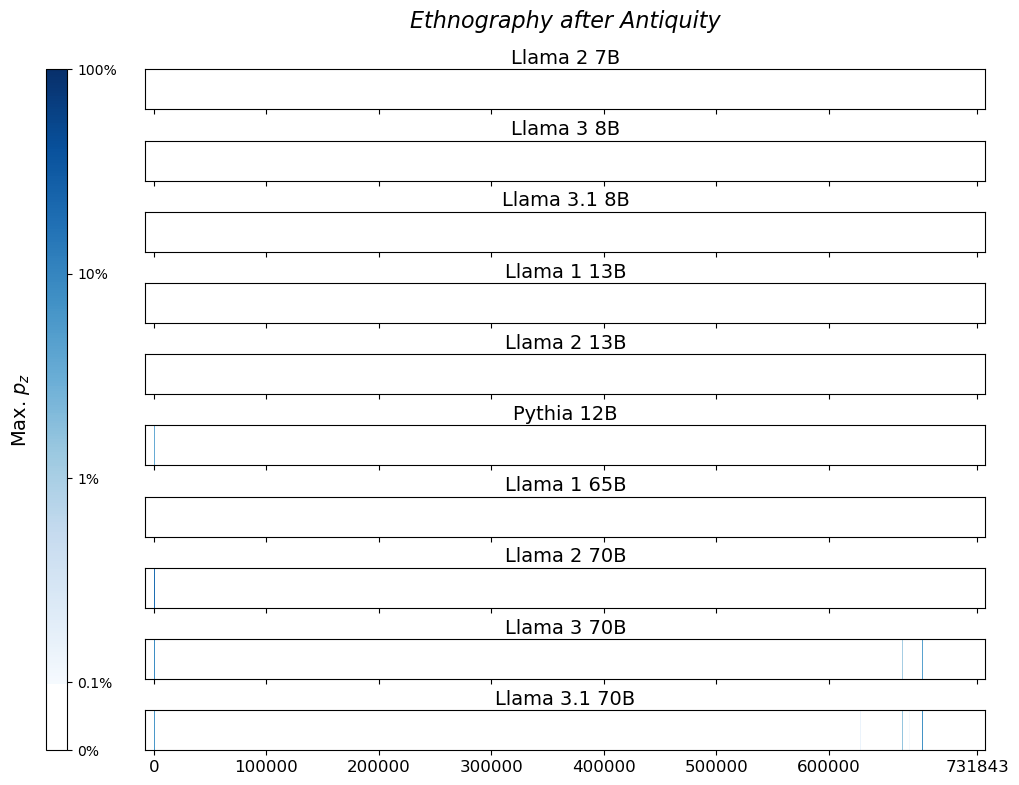}
    \includegraphics[width=\linewidth]{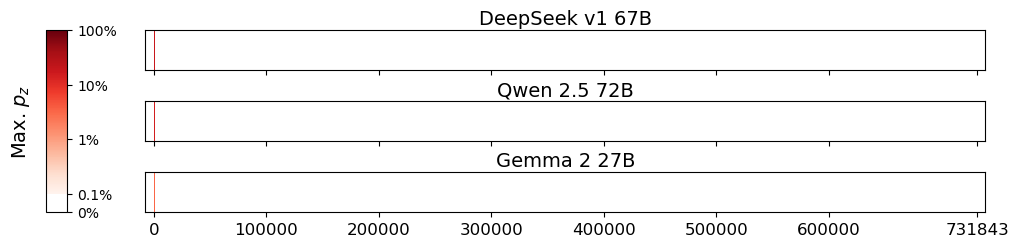}
    \includegraphics[width=\linewidth]{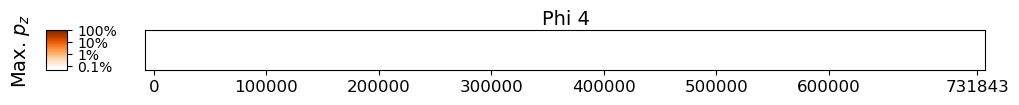}
  \end{minipage}
  \hfill
  \begin{minipage}[t]{0.45\textwidth}
    \centering
    \vspace{0cm}
    \includegraphics[width=\linewidth]{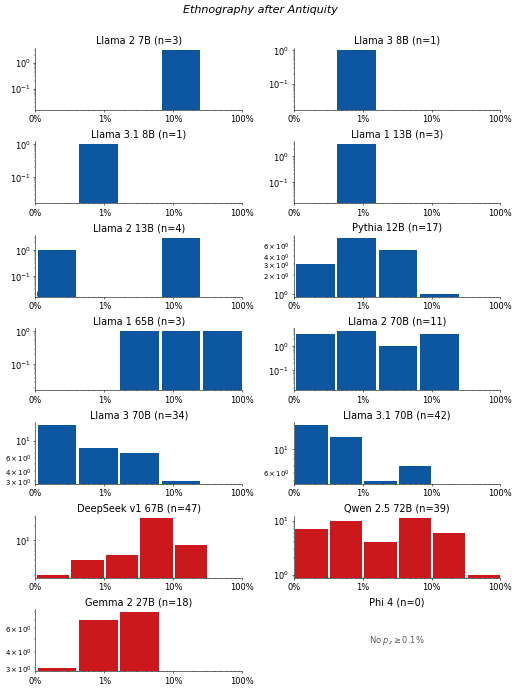}
  \end{minipage}
  \vspace{-.2cm}
  \caption{
    \textbf{\textit{Ethnography after Antiquity}, \citeauthor{Ethnography_after_Antiquity}.}
    For $14$ LLMs,
    (\textbf{left}) heatmaps for the sliding-window procedure and
    (\textbf{right}) corresponding distributions over suffix extraction probabilities
    ($\tau_\text{min}=0.1\%$).
  }
  \label{fig:slidingwindow:Ethnography_after_Antiquity}
\end{figure}
\FloatBarrier

\subsubsection{\textit{Who Is Rich?}, \citeauthor{Who_Is_Rich}}\label{app:sec:sliding:Who_Is_Rich}
\vspace{-.2cm}
\begin{figure}[h]
  \centering
  \begin{minipage}[t]{0.53\textwidth}
    \centering
    \vspace{0cm}
    \includegraphics[width=\linewidth]{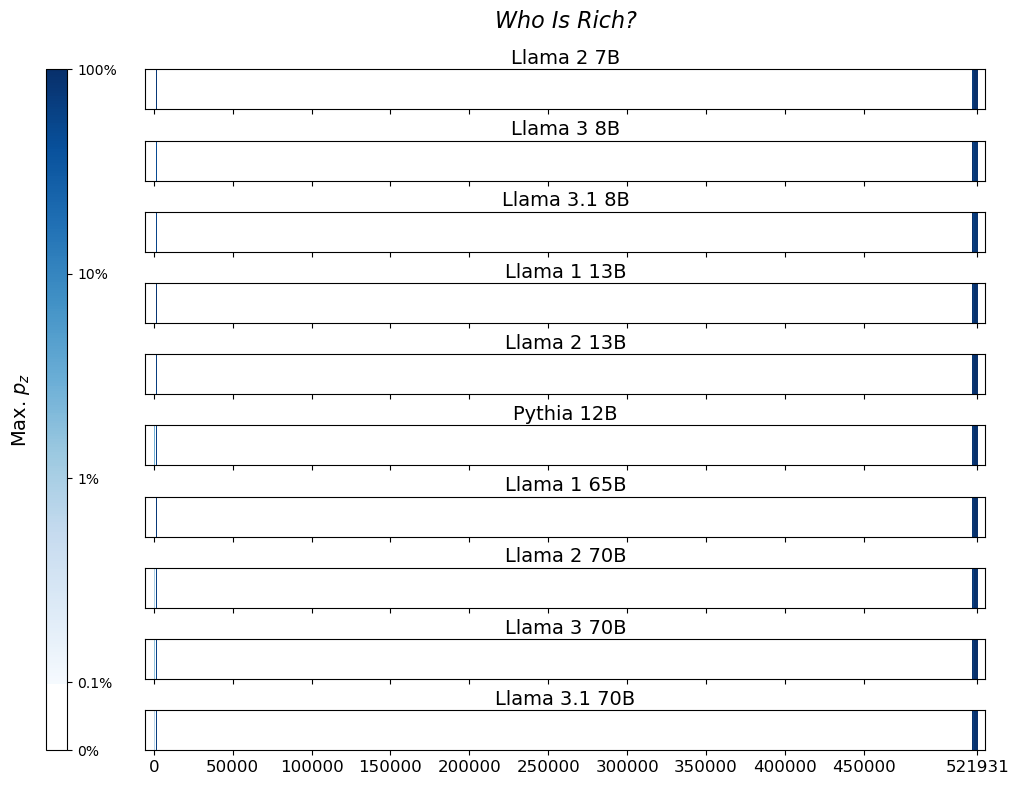}
    \includegraphics[width=\linewidth]{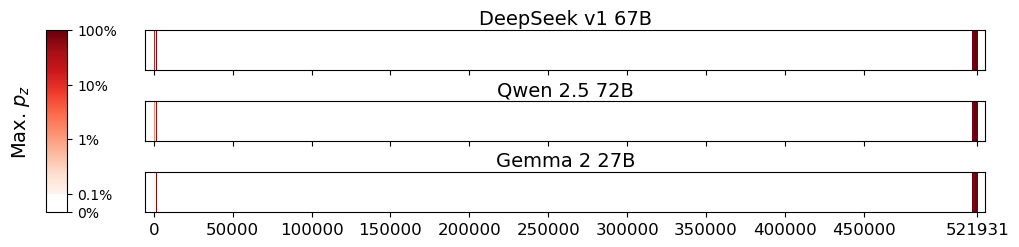}
    \includegraphics[width=\linewidth]{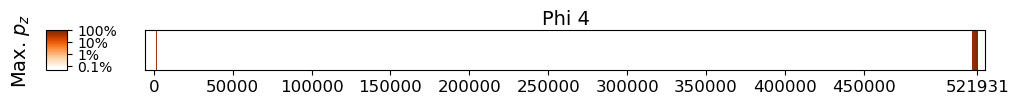}
  \end{minipage}
  \hfill
  \begin{minipage}[t]{0.45\textwidth}
    \centering
    \vspace{0cm}
    \includegraphics[width=\linewidth]{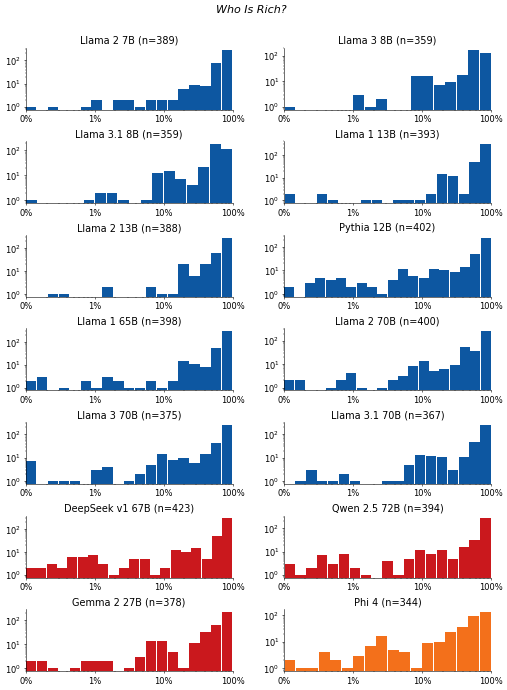}
  \end{minipage}
  \vspace{-.2cm}
  \caption{
    \textbf{\textit{Who Is Rich?}, \citeauthor{Who_Is_Rich}.}
    For $14$ LLMs,
    (\textbf{left}) heatmaps for the sliding-window procedure and
    (\textbf{right}) corresponding distributions over suffix extraction probabilities
    ($\tau_\text{min}=0.1\%$).
  }
  \label{fig:slidingwindow:Who_Is_Rich}
\end{figure}
\FloatBarrier

\clearpage
\subsubsection{\textit{The Servants of Twilight}, \citeauthor{The_Servants_of_Twilight}}\label{app:sec:sliding:The_Servants_of_Twilight}
\vspace{-.2cm}
\begin{figure}[h]
  \centering
  \begin{minipage}[t]{0.53\textwidth}
    \centering
    \vspace{0cm}
    \includegraphics[width=\linewidth]{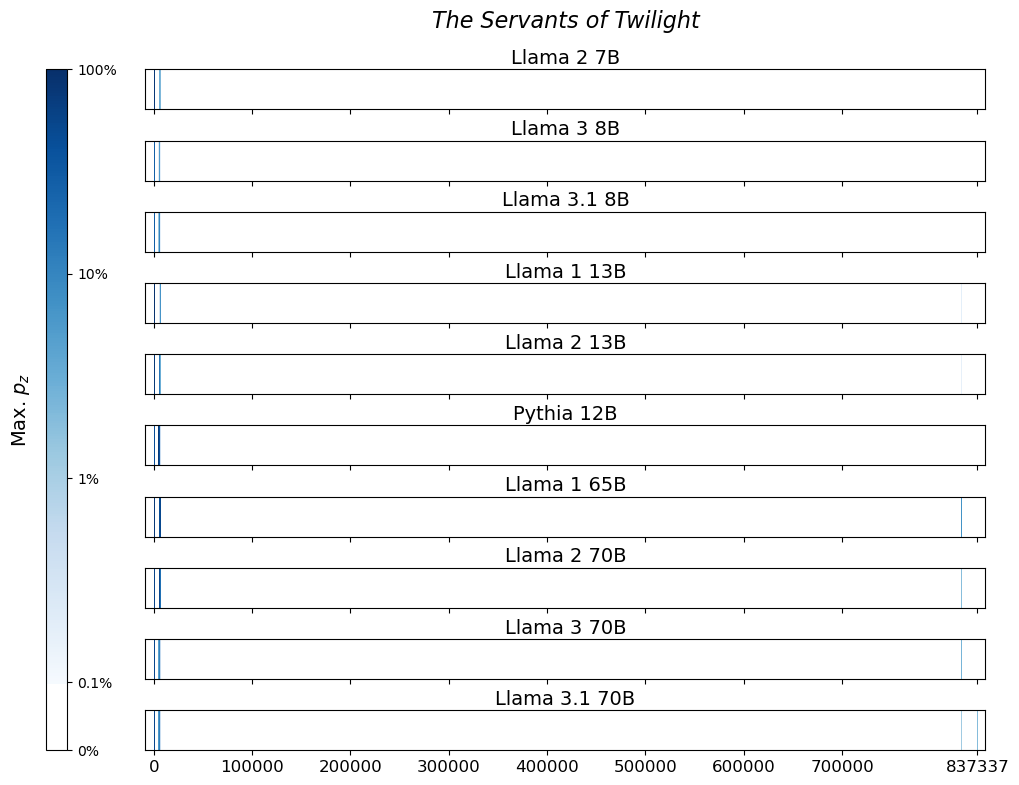}
    \includegraphics[width=\linewidth]{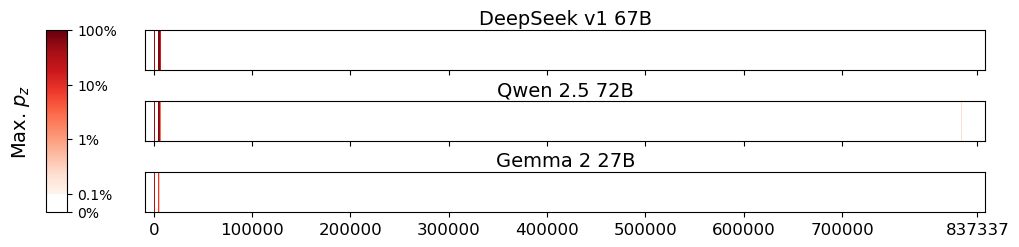}
    \includegraphics[width=\linewidth]{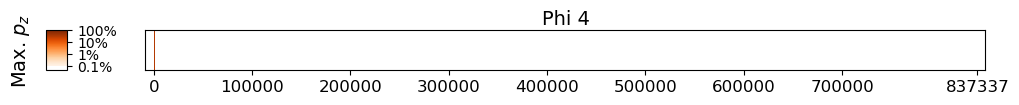}
  \end{minipage}
  \hfill
  \begin{minipage}[t]{0.45\textwidth}
    \centering
    \vspace{0cm}
    \includegraphics[width=\linewidth]{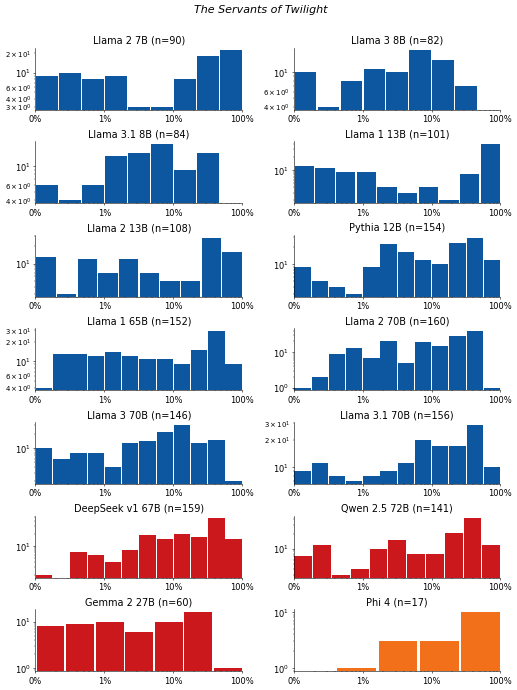}
  \end{minipage}
  \vspace{-.2cm}
  \caption{
    \textbf{\textit{The Servants of Twilight}, \citeauthor{The_Servants_of_Twilight}.}
    For $14$ LLMs,
    (\textbf{left}) heatmaps for the sliding-window procedure and
    (\textbf{right}) corresponding distributions over suffix extraction probabilities
    ($\tau_\text{min}=0.1\%$).
  }
  \label{fig:slidingwindow:The_Servants_of_Twilight}
\end{figure}
\FloatBarrier

\subsubsection{\textit{Tai Chi for Depression}, \citeauthor{Tai_Chi_for_Depression}}\label{app:sec:sliding:Tai_Chi_for_Depression}
\vspace{-.2cm}
\begin{figure}[h]
  \centering
  \begin{minipage}[t]{0.53\textwidth}
    \centering
    \vspace{0cm}
    \includegraphics[width=\linewidth]{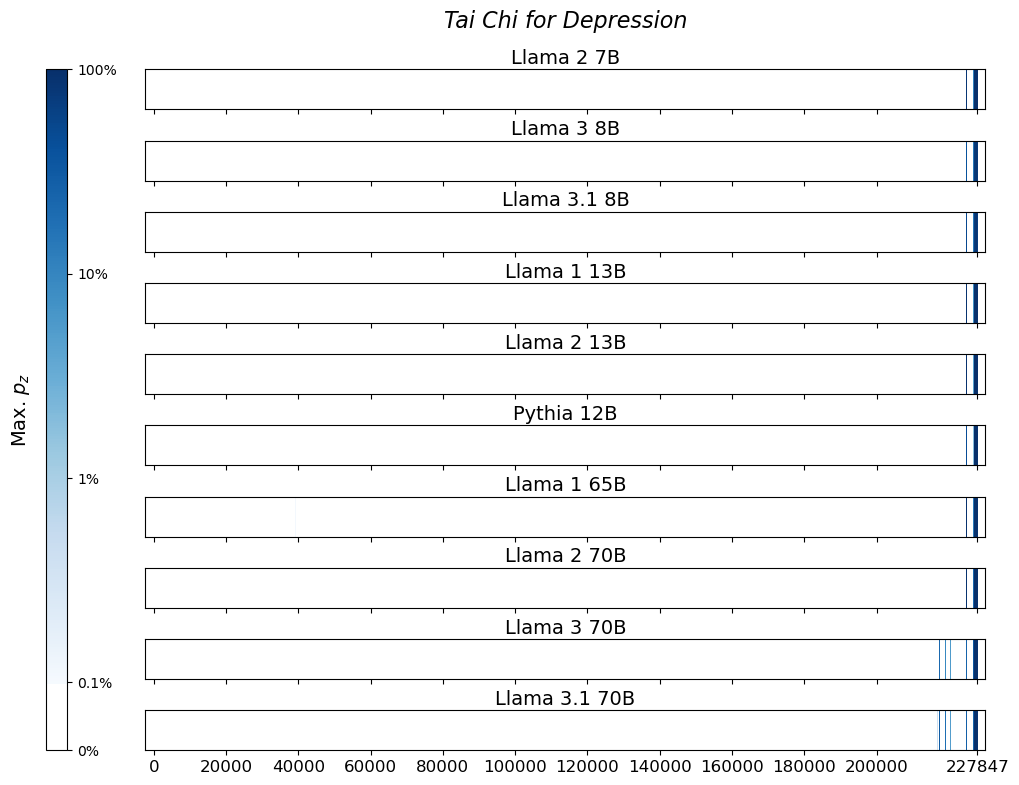}
    \includegraphics[width=\linewidth]{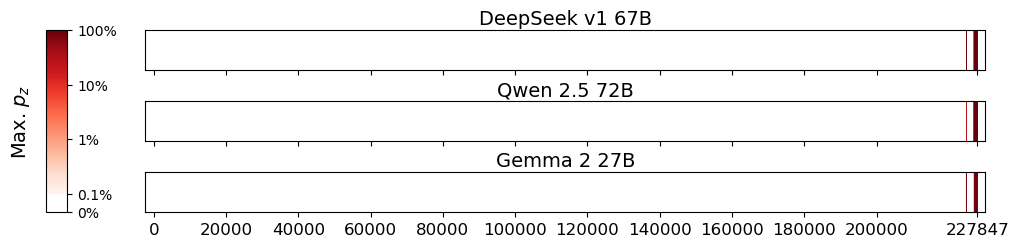}
    \includegraphics[width=\linewidth]{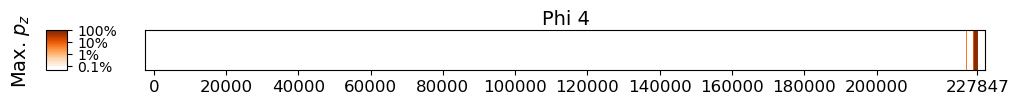}
  \end{minipage}
  \hfill
  \begin{minipage}[t]{0.45\textwidth}
    \centering
    \vspace{0cm}
    \includegraphics[width=\linewidth]{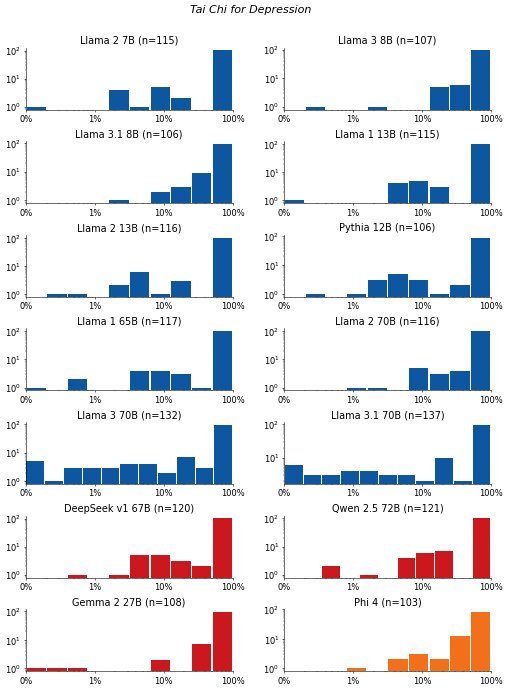}
  \end{minipage}
  \vspace{-.2cm}
  \caption{
    \textbf{\textit{Tai Chi for Depression}, \citeauthor{Tai_Chi_for_Depression}.}
    For $14$ LLMs,
    (\textbf{left}) heatmaps for the sliding-window procedure and
    (\textbf{right}) corresponding distributions over suffix extraction probabilities
    ($\tau_\text{min}=0.1\%$).
  }
  \label{fig:slidingwindow:Tai_Chi_for_Depression}
\end{figure}
\FloatBarrier

\clearpage
\subsubsection{\textit{The Tide Was Always High}, \citeauthor{The_Tide_Was_Always_High}}\label{app:sec:sliding:The_Tide_Was_Always_High}
\vspace{-.2cm}
\begin{figure}[h]
  \centering
  \begin{minipage}[t]{0.53\textwidth}
    \centering
    \vspace{0cm}
    \includegraphics[width=\linewidth]{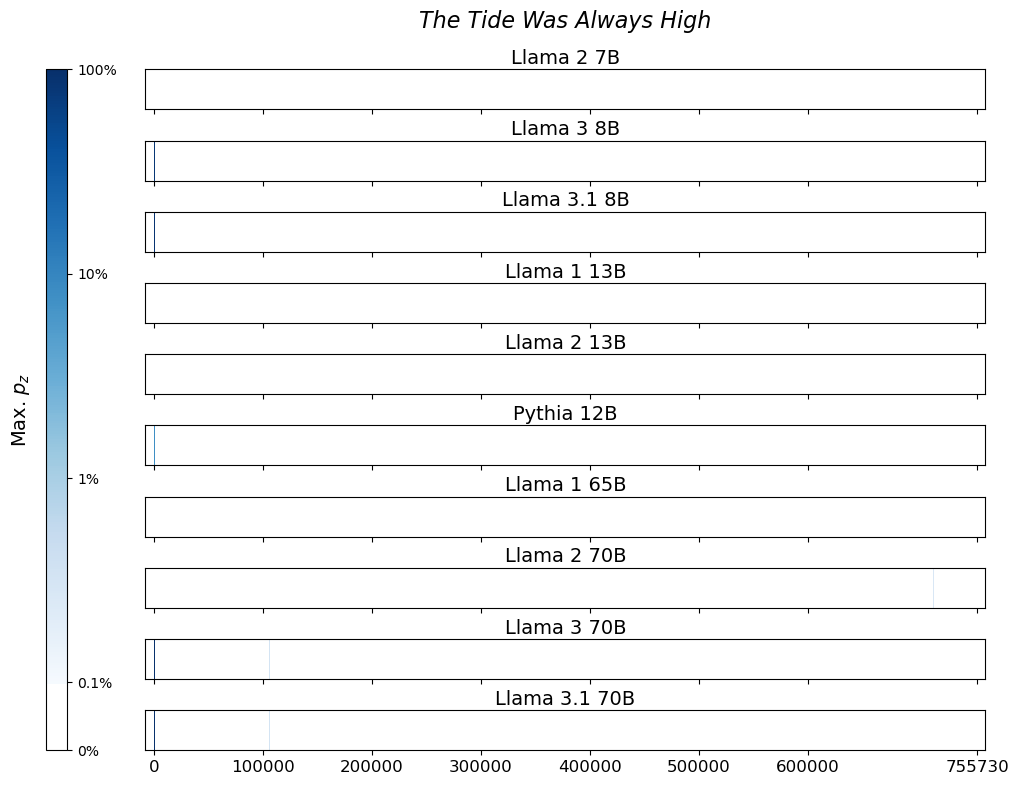}
    \includegraphics[width=\linewidth]{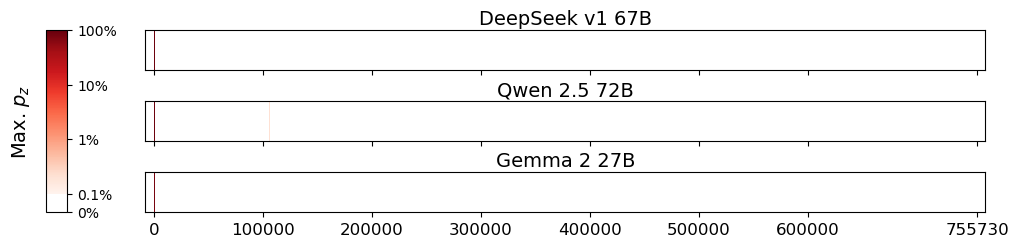}
    \includegraphics[width=\linewidth]{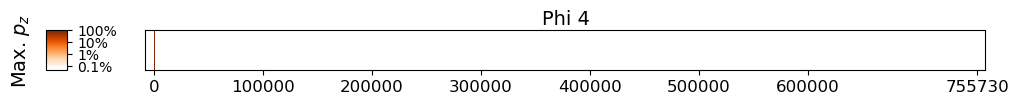}
  \end{minipage}
  \hfill
  \begin{minipage}[t]{0.45\textwidth}
    \centering
    \vspace{0cm}
    \includegraphics[width=\linewidth]{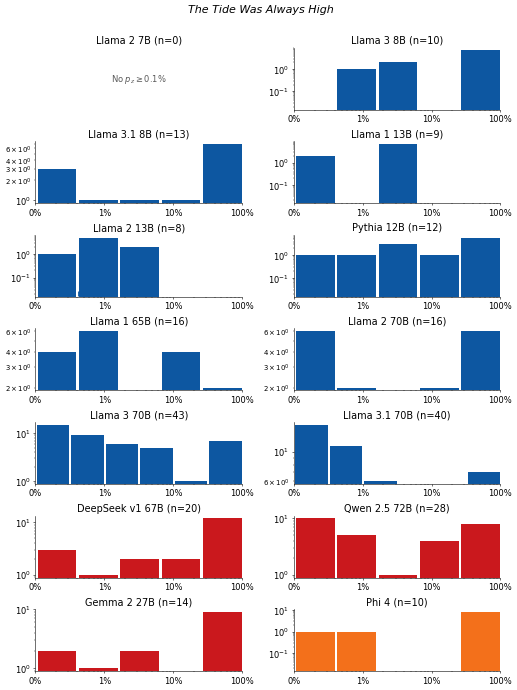}
  \end{minipage}
  \vspace{-.2cm}
  \caption{
    \textbf{\textit{The Tide Was Always High}, \citeauthor{The_Tide_Was_Always_High}.}
    For $14$ LLMs,
    (\textbf{left}) heatmaps for the sliding-window procedure and
    (\textbf{right}) corresponding distributions over suffix extraction probabilities
    ($\tau_\text{min}=0.1\%$).
  }
  \label{fig:slidingwindow:The_Tide_Was_Always_High}
\end{figure}
\FloatBarrier

\subsubsection{\textit{Girl in Translation}, \citeauthor{Girl_in_Translation}}\label{app:sec:sliding:Girl_in_Translation}
\vspace{-.2cm}
\begin{figure}[h]
  \centering
  \begin{minipage}[t]{0.53\textwidth}
    \centering
    \vspace{0cm}
    \includegraphics[width=\linewidth]{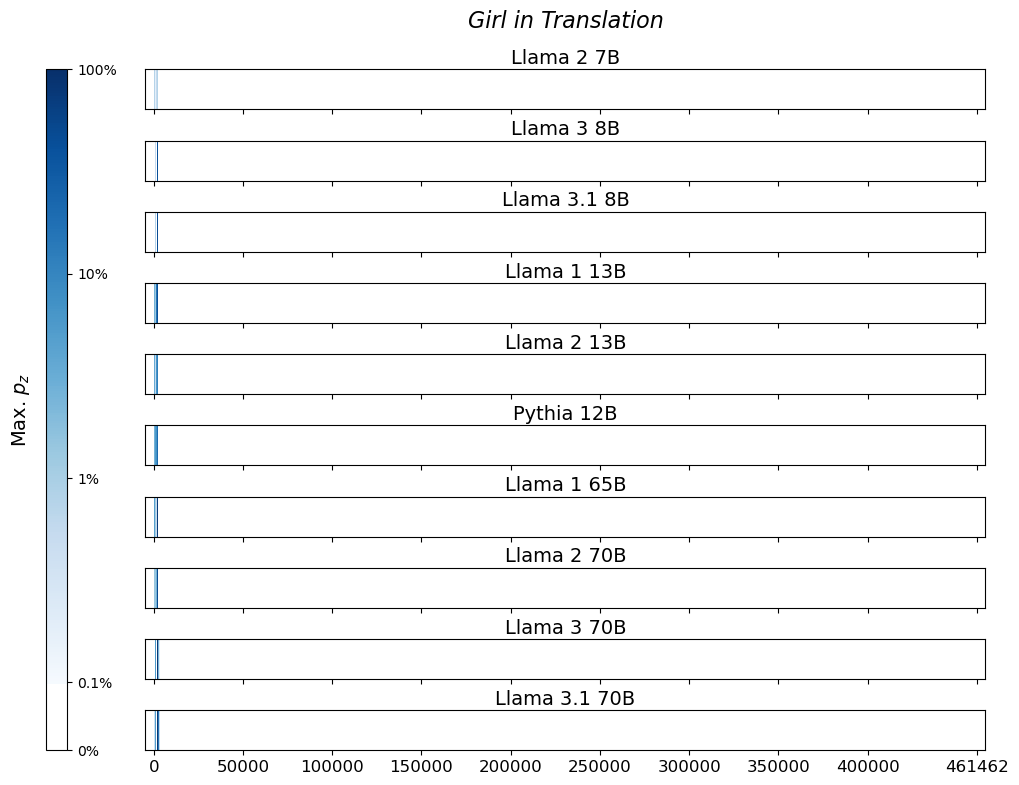}
    \includegraphics[width=\linewidth]{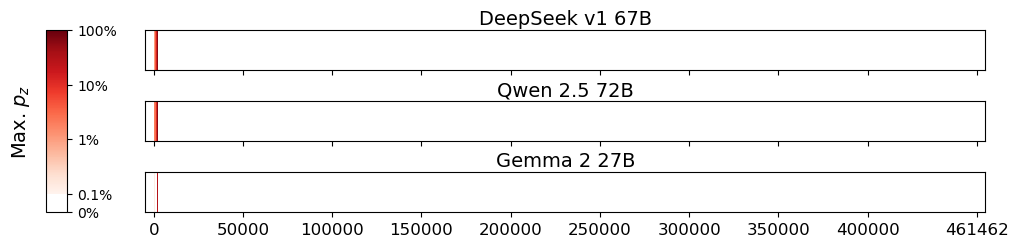}
    \includegraphics[width=\linewidth]{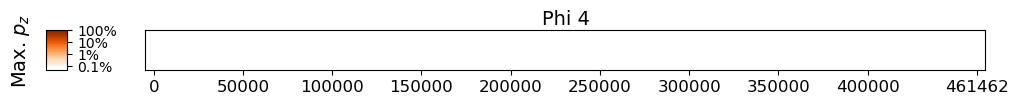}
  \end{minipage}
  \hfill
  \begin{minipage}[t]{0.45\textwidth}
    \centering
    \vspace{0cm}
    \includegraphics[width=\linewidth]{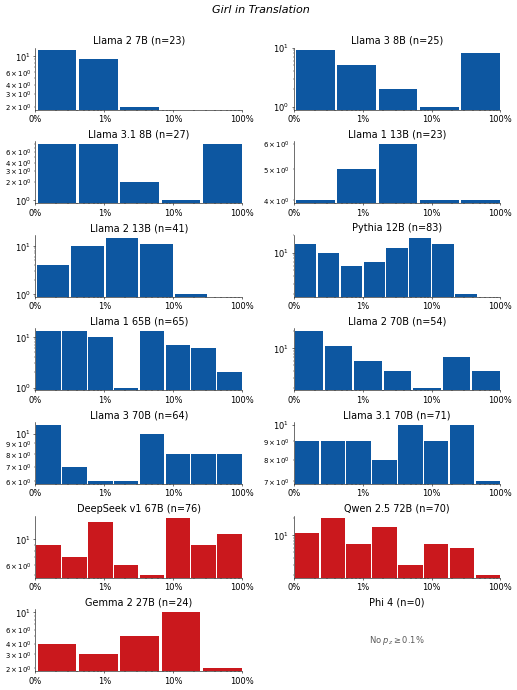}
  \end{minipage}
  \vspace{-.2cm}
  \caption{
    \textbf{\textit{Girl in Translation}, \citeauthor{Girl_in_Translation}.}
    For $14$ LLMs,
    (\textbf{left}) heatmaps for the sliding-window procedure and
    (\textbf{right}) corresponding distributions over suffix extraction probabilities
    ($\tau_\text{min}=0.1\%$).
  }
  \label{fig:slidingwindow:Girl_in_Translation}
\end{figure}
\FloatBarrier

\clearpage
\subsubsection{\textit{A Wrinkle in Time}, \citeauthor{A_Wrinkle_in_Time}}\label{app:sec:sliding:A_Wrinkle_in_Time}
\vspace{-.2cm}
\begin{figure}[h]
  \centering
  \begin{minipage}[t]{0.53\textwidth}
    \centering
    \vspace{0cm}
    \includegraphics[width=\linewidth]{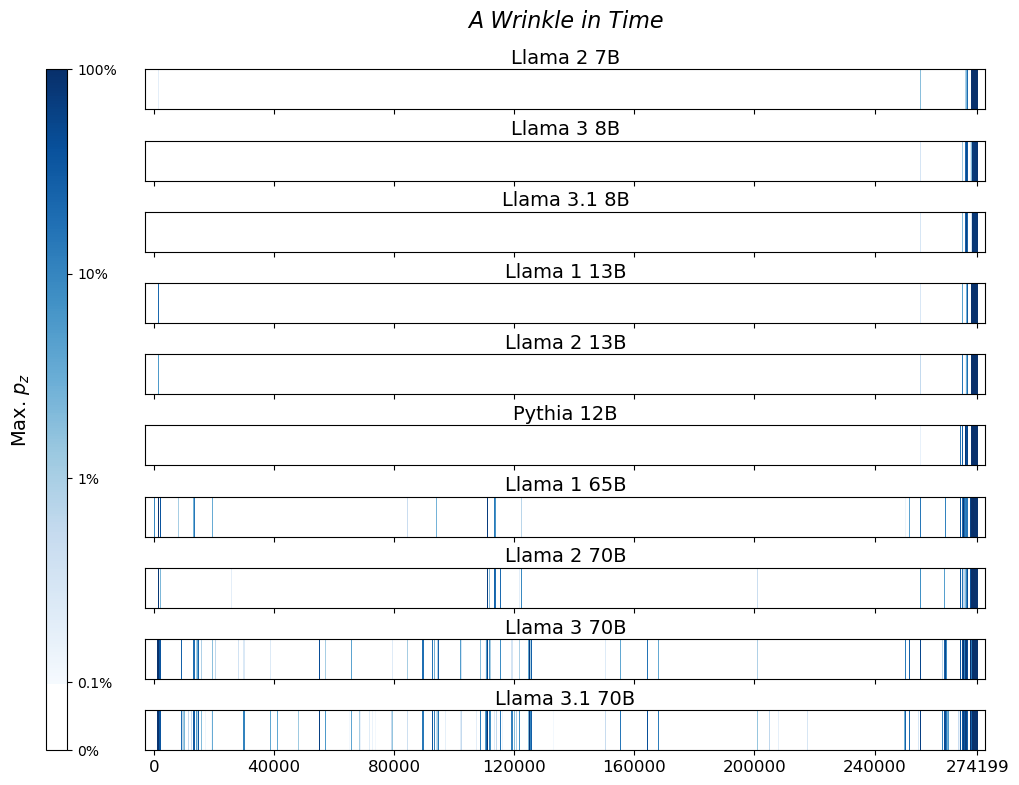}
    \includegraphics[width=\linewidth]{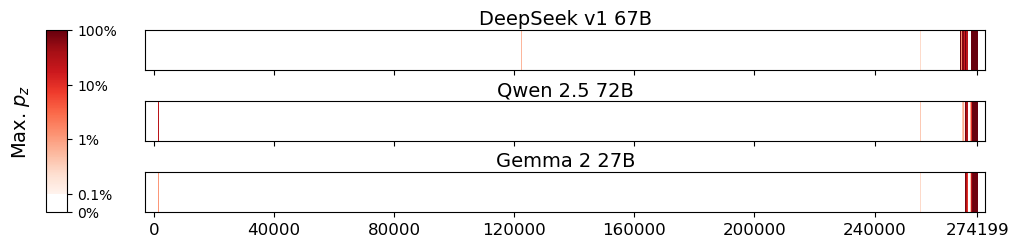}
    \includegraphics[width=\linewidth]{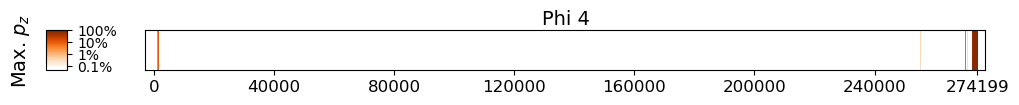}
  \end{minipage}
  \hfill
  \begin{minipage}[t]{0.45\textwidth}
    \centering
    \vspace{0cm}
    \includegraphics[width=\linewidth]{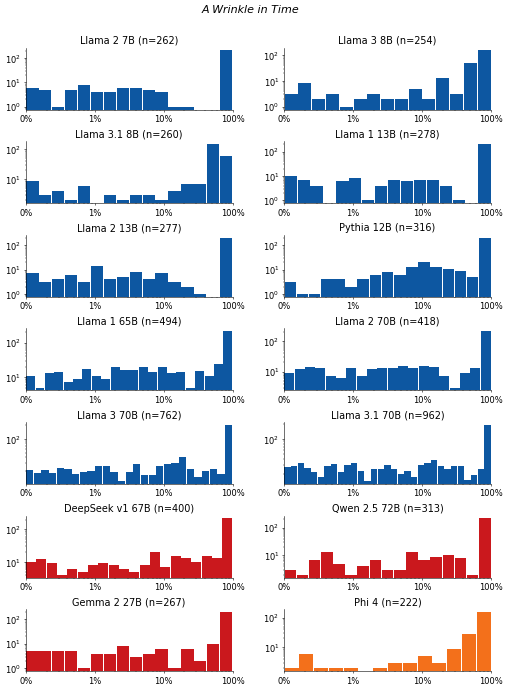}
  \end{minipage}
  \vspace{-.2cm}
  \caption{
    \textbf{\textit{A Wrinkle in Time}, \citeauthor{A_Wrinkle_in_Time}.}
    For $14$ LLMs,
    (\textbf{left}) heatmaps for the sliding-window procedure and
    (\textbf{right}) corresponding distributions over suffix extraction probabilities
    ($\tau_\text{min}=0.1\%$).
  }
  \label{fig:slidingwindow:A_Wrinkle_in_Time}
\end{figure}
\FloatBarrier

\subsubsection{\textit{Call Me Brooklyn}, \citeauthor{Call_Me_Brooklyn}}\label{app:sec:sliding:Call_Me_Brooklyn}
\vspace{-.2cm}
\begin{figure}[h]
  \centering
  \begin{minipage}[t]{0.53\textwidth}
    \centering
    \vspace{0cm}
    \includegraphics[width=\linewidth]{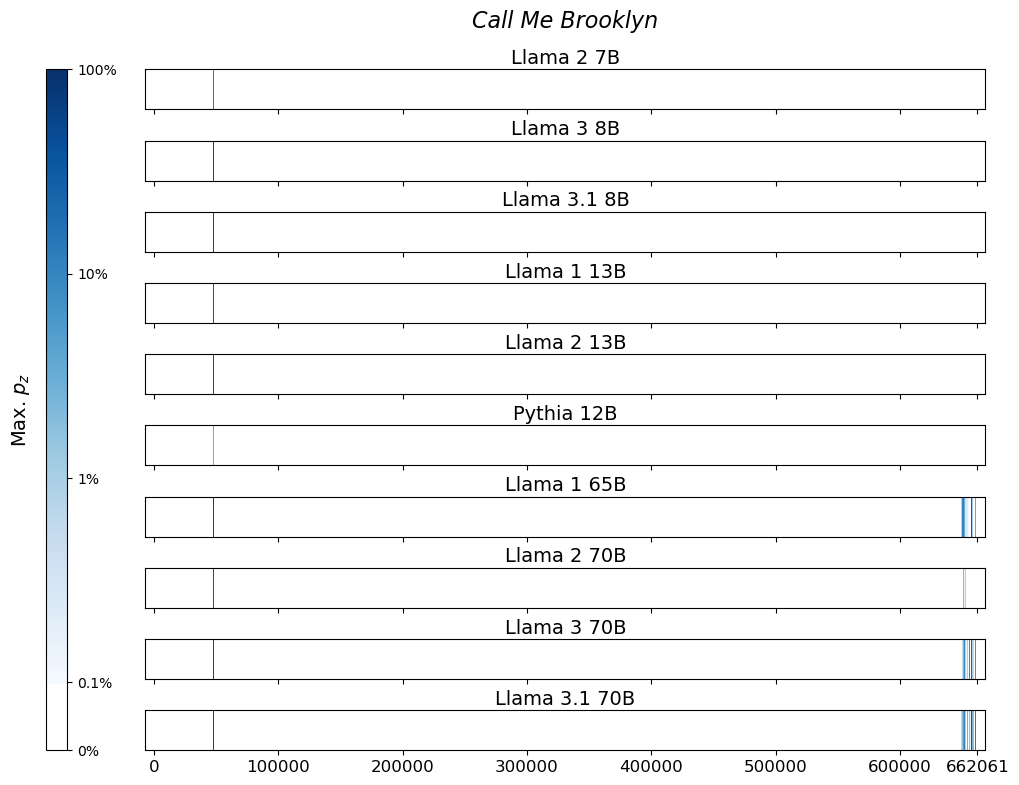}
    \includegraphics[width=\linewidth]{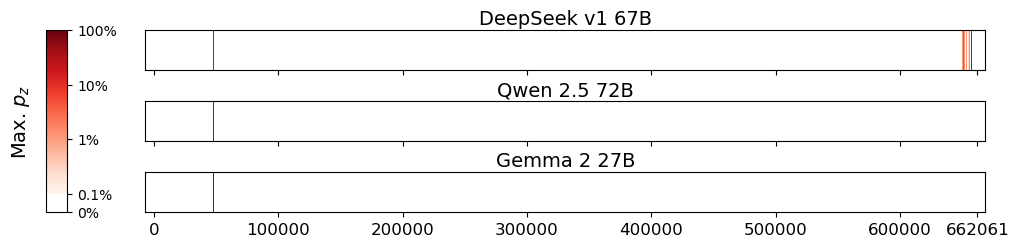}
    \includegraphics[width=\linewidth]{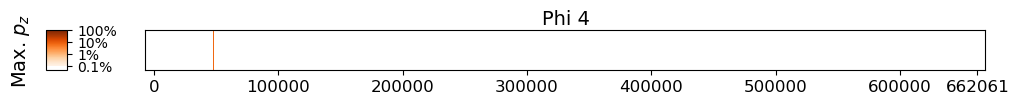}
  \end{minipage}
  \hfill
  \begin{minipage}[t]{0.45\textwidth}
    \centering
    \vspace{0cm}
    \includegraphics[width=\linewidth]{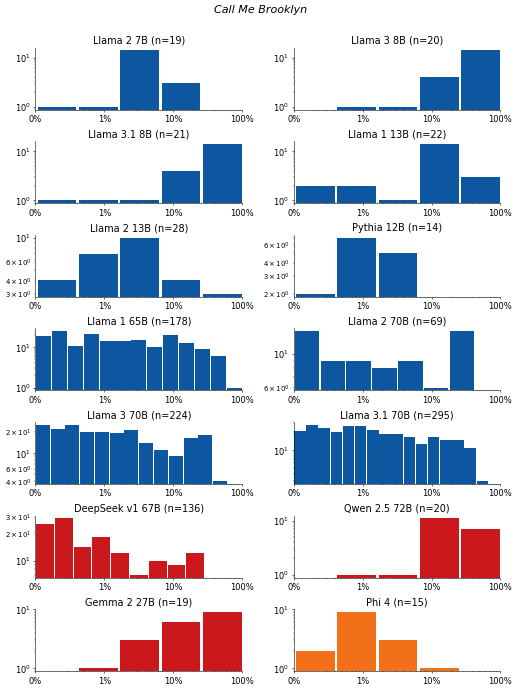}
  \end{minipage}
  \vspace{-.2cm}
  \caption{
    \textbf{\textit{Call Me Brooklyn}, \citeauthor{Call_Me_Brooklyn}.}
    For $14$ LLMs,
    (\textbf{left}) heatmaps for the sliding-window procedure and
    (\textbf{right}) corresponding distributions over suffix extraction probabilities
    ($\tau_\text{min}=0.1\%$).
  }
  \label{fig:slidingwindow:Call_Me_Brooklyn}
\end{figure}
\FloatBarrier

\clearpage
\subsubsection{\textit{Dead Wake}, \citeauthor{Dead_Wake}}\label{app:sec:sliding:Dead_Wake}
\vspace{-.2cm}
\begin{figure}[h]
  \centering
  \begin{minipage}[t]{0.53\textwidth}
    \centering
    \vspace{0cm}
    \includegraphics[width=\linewidth]{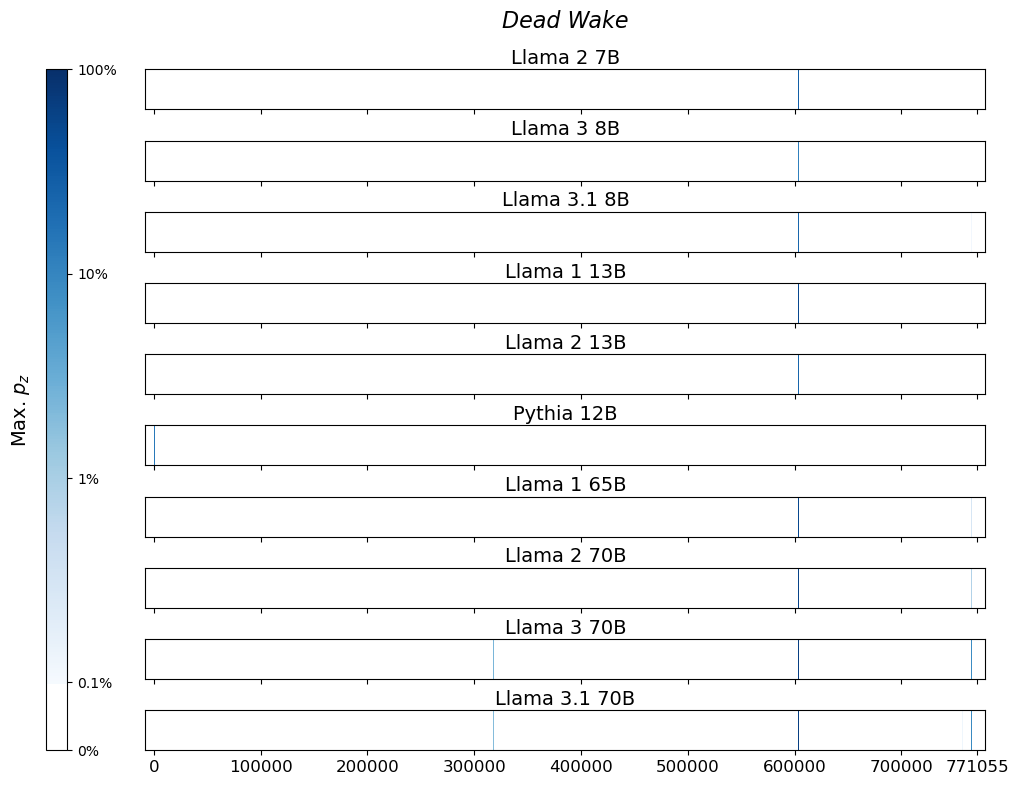}
    \includegraphics[width=\linewidth]{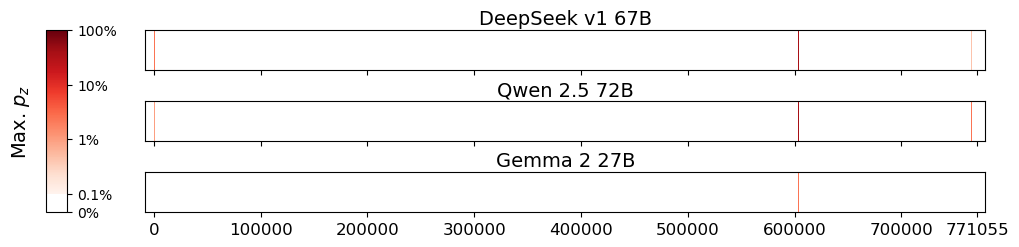}
    \includegraphics[width=\linewidth]{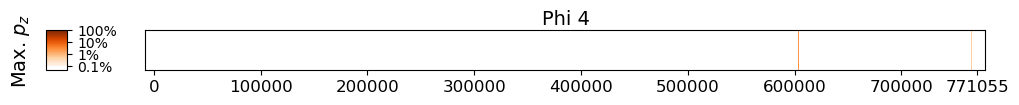}
  \end{minipage}
  \hfill
  \begin{minipage}[t]{0.45\textwidth}
    \centering
    \vspace{0cm}
    \includegraphics[width=\linewidth]{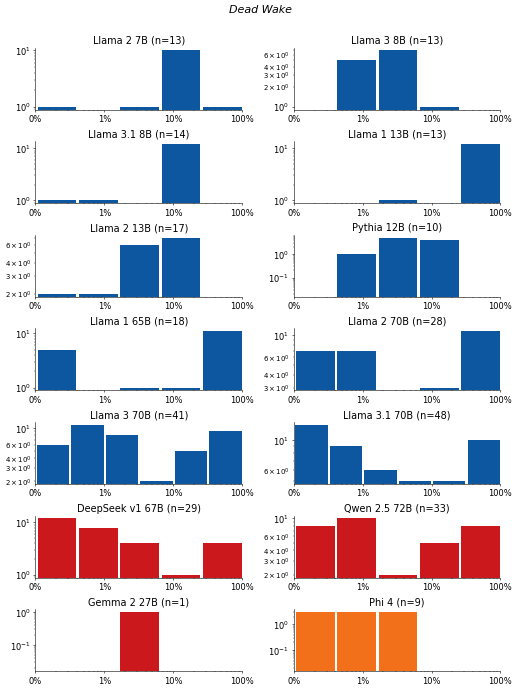}
  \end{minipage}
  \vspace{-.2cm}
  \caption{
    \textbf{\textit{Dead Wake}, \citeauthor{Dead_Wake}.}
    For $14$ LLMs,
    (\textbf{left}) heatmaps for the sliding-window procedure and
    (\textbf{right}) corresponding distributions over suffix extraction probabilities
    ($\tau_\text{min}=0.1\%$).
  }
  \label{fig:slidingwindow:Dead_Wake}
\end{figure}
\FloatBarrier

\subsubsection{\textit{The Girl with the Dragon Tattoo}, \citeauthor{The_Girl_with_the_Dragon_Tattoo}}\label{app:sec:sliding:The_Girl_with_the_Dragon_Tattoo}
\vspace{-.2cm}
\begin{figure}[h]
  \centering
  \begin{minipage}[t]{0.53\textwidth}
    \centering
    \vspace{0cm}
    \includegraphics[width=\linewidth]{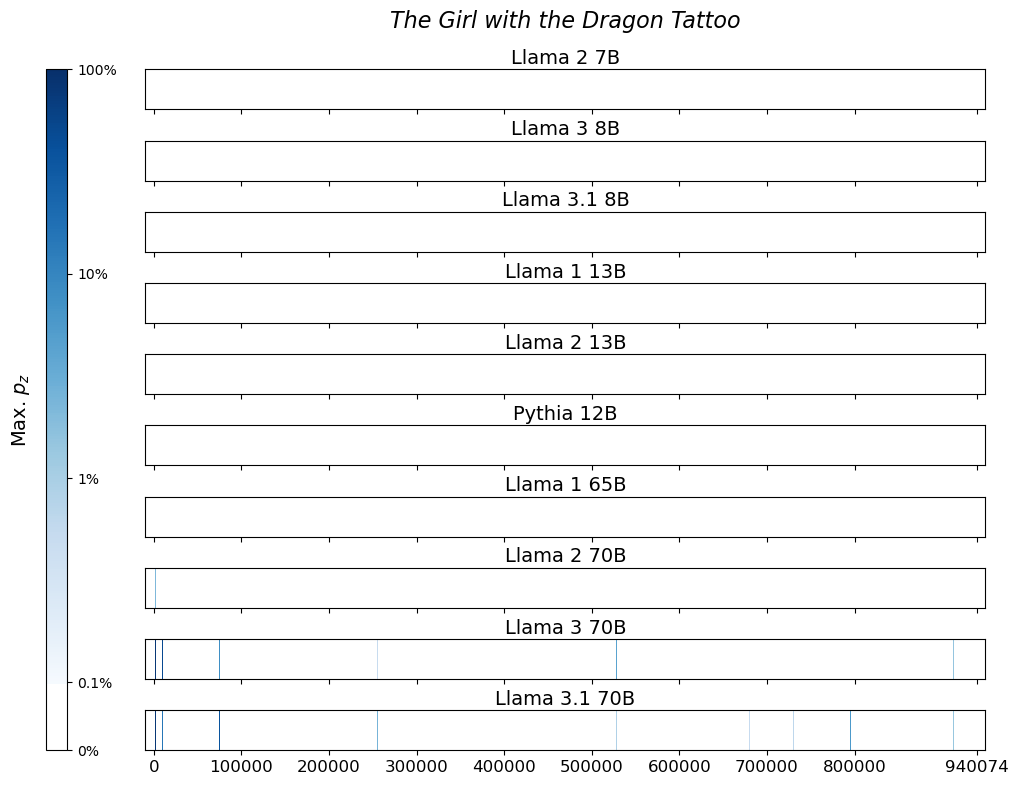}
    \includegraphics[width=\linewidth]{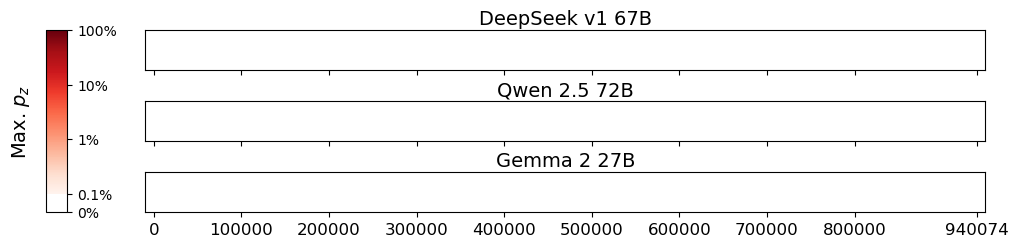}
    \includegraphics[width=\linewidth]{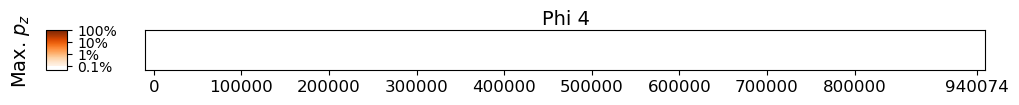}
  \end{minipage}
  \hfill
  \begin{minipage}[t]{0.45\textwidth}
    \centering
    \vspace{0cm}
    \includegraphics[width=\linewidth]{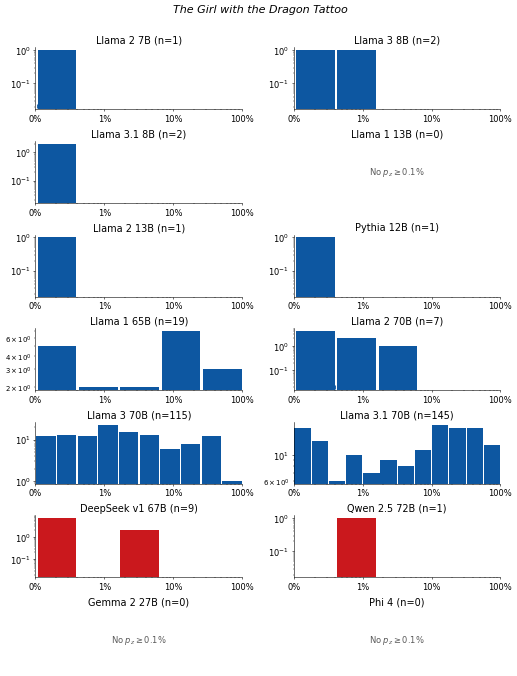}
  \end{minipage}
  \vspace{-.2cm}
  \caption{
    \textbf{\textit{The Girl with the Dragon Tattoo}, \citeauthor{The_Girl_with_the_Dragon_Tattoo}.}
    For $14$ LLMs,
    (\textbf{left}) heatmaps for the sliding-window procedure and
    (\textbf{right}) corresponding distributions over suffix extraction probabilities
    ($\tau_\text{min}=0.1\%$).
  }
  \label{fig:slidingwindow:The_Girl_with_the_Dragon_Tattoo}
\end{figure}
\FloatBarrier

\clearpage
\subsubsection{\textit{The Daughter of Odren}, \citeauthor{The_Daughter_of_Odren}}\label{app:sec:sliding:The_Daughter_of_Odren}
\vspace{-.2cm}
\begin{figure}[h]
  \centering
  \begin{minipage}[t]{0.53\textwidth}
    \centering
    \vspace{0cm}
    \includegraphics[width=\linewidth]{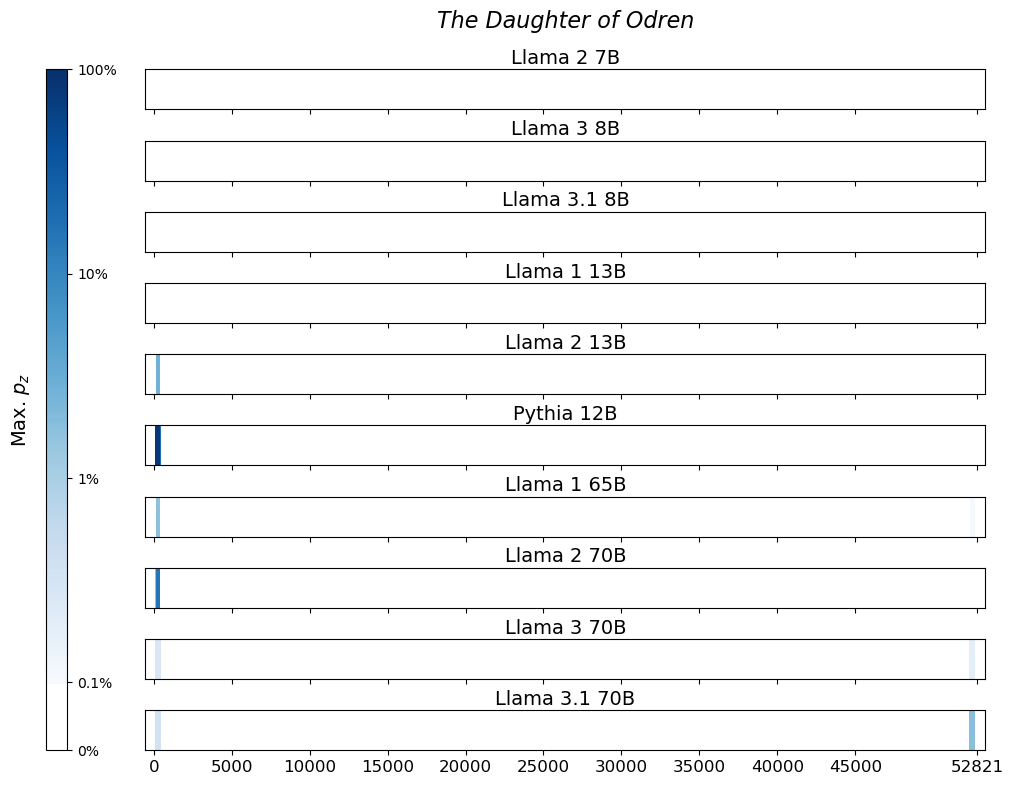}
    \includegraphics[width=\linewidth]{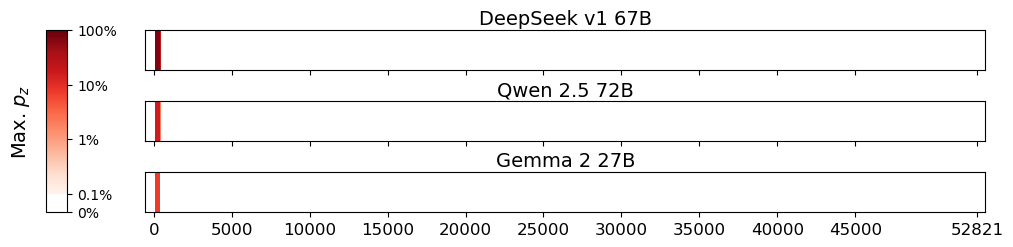}
    \includegraphics[width=\linewidth]{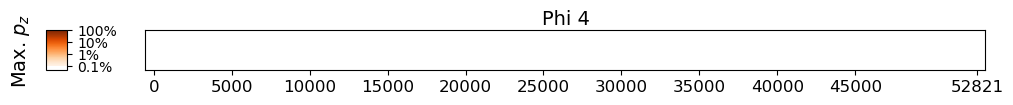}
  \end{minipage}
  \hfill
  \begin{minipage}[t]{0.45\textwidth}
    \centering
    \vspace{0cm}
    \includegraphics[width=\linewidth]{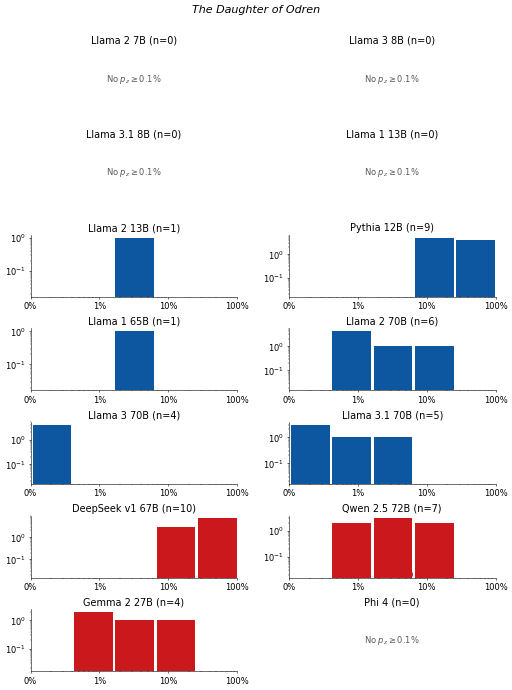}
  \end{minipage}
  \vspace{-.2cm}
  \caption{
    \textbf{\textit{The Daughter of Odren}, \citeauthor{The_Daughter_of_Odren}.}
    For $14$ LLMs,
    (\textbf{left}) heatmaps for the sliding-window procedure and
    (\textbf{right}) corresponding distributions over suffix extraction probabilities
    ($\tau_\text{min}=0.1\%$).
  }
  \label{fig:slidingwindow:The_Daughter_of_Odren}
\end{figure}
\FloatBarrier

\subsubsection{\textit{The Chronicles of Narnia}, \citeauthor{The_Chronicles_of_Narnia}}\label{app:sec:sliding:The_Chronicles_of_Narnia}
\vspace{-.2cm}
\begin{figure}[h]
  \centering
  \begin{minipage}[t]{0.53\textwidth}
    \centering
    \vspace{0cm}
    \includegraphics[width=\linewidth]{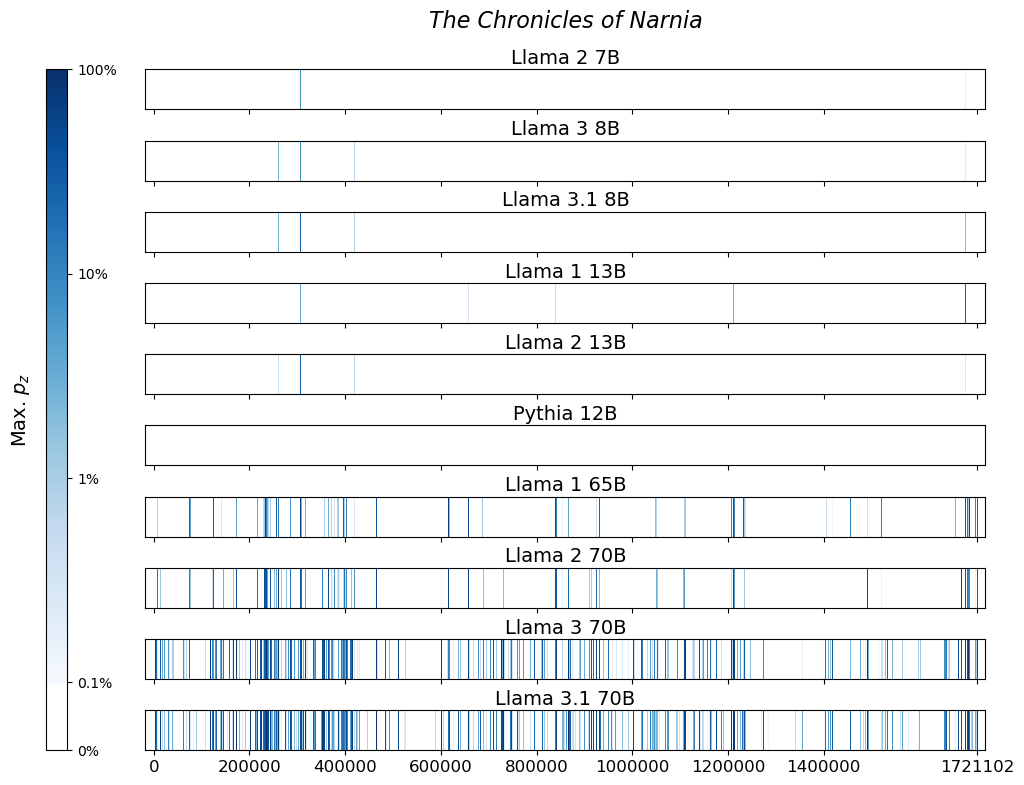}
    \includegraphics[width=\linewidth]{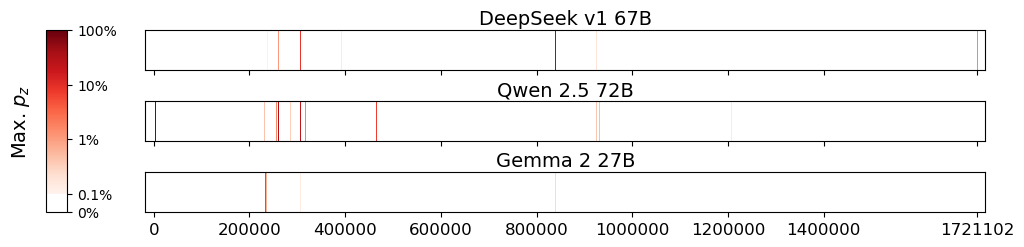}
    \includegraphics[width=\linewidth]{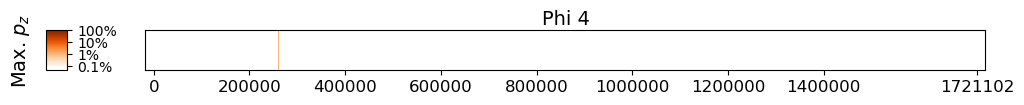}
  \end{minipage}
  \hfill
  \begin{minipage}[t]{0.45\textwidth}
    \centering
    \vspace{0cm}
    \includegraphics[width=\linewidth]{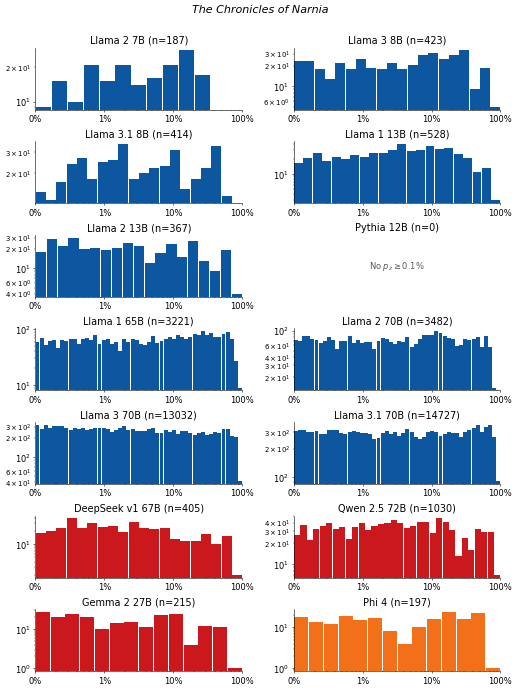}
  \end{minipage}
  \vspace{-.2cm}
  \caption{
    \textbf{\textit{The Chronicles of Narnia}, \citeauthor{The_Chronicles_of_Narnia}.}
    For $14$ LLMs,
    (\textbf{left}) heatmaps for the sliding-window procedure and
    (\textbf{right}) corresponding distributions over suffix extraction probabilities
    ($\tau_\text{min}=0.1\%$).
  }
  \label{fig:slidingwindow:The_Chronicles_of_Narnia}
\end{figure}
\FloatBarrier

\clearpage
\subsubsection{\textit{After I'm Gone}, \citeauthor{After_I_m_Gone}}\label{app:sec:sliding:After_I_m_Gone}
\vspace{-.2cm}
\begin{figure}[h]
  \centering
  \begin{minipage}[t]{0.53\textwidth}
    \centering
    \vspace{0cm}
    \includegraphics[width=\linewidth]{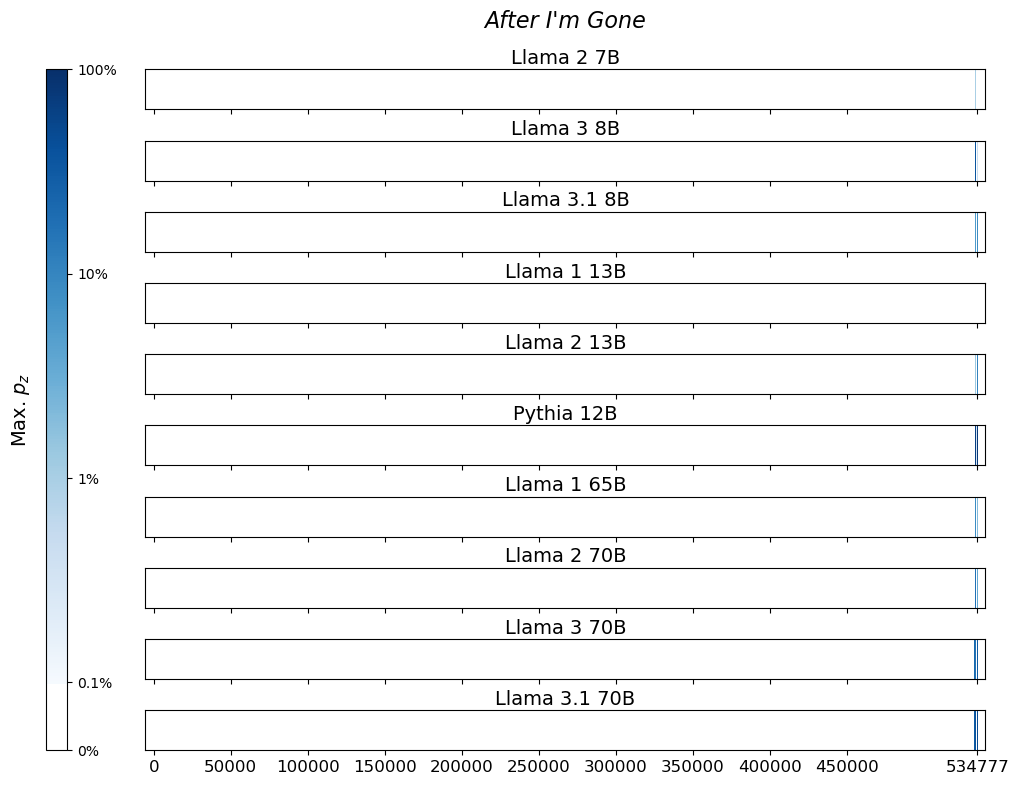}
    \includegraphics[width=\linewidth]{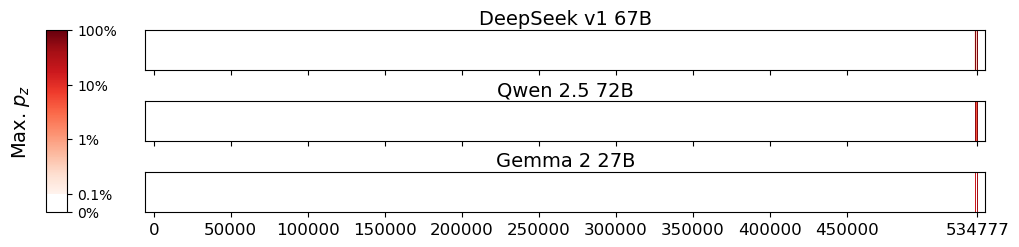}
    \includegraphics[width=\linewidth]{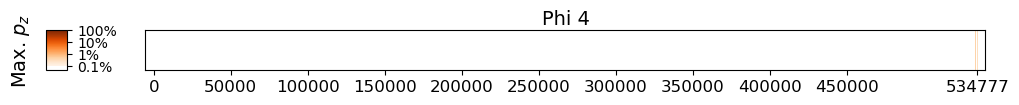}
  \end{minipage}
  \hfill
  \begin{minipage}[t]{0.45\textwidth}
    \centering
    \vspace{0cm}
    \includegraphics[width=\linewidth]{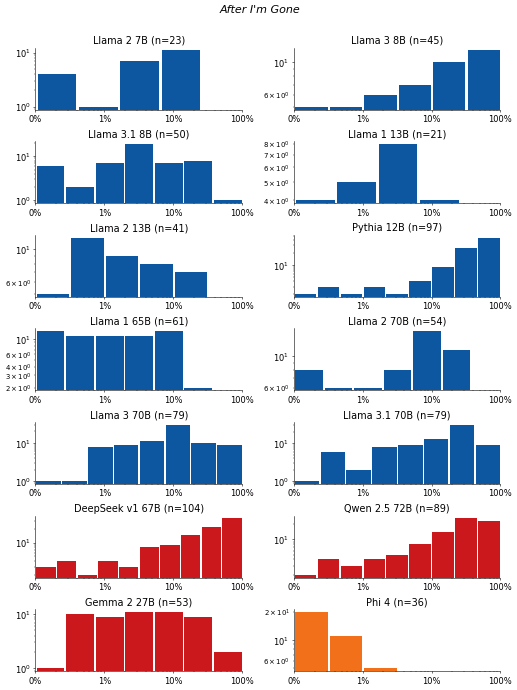}
  \end{minipage}
  \vspace{-.2cm}
  \caption{
    \textbf{\textit{After I'm Gone}, \citeauthor{After_I_m_Gone}.}
    For $14$ LLMs,
    (\textbf{left}) heatmaps for the sliding-window procedure and
    (\textbf{right}) corresponding distributions over suffix extraction probabilities
    ($\tau_\text{min}=0.1\%$).
  }
  \label{fig:slidingwindow:After_I_m_Gone}
\end{figure}
\FloatBarrier

\subsubsection{\textit{Sunburn}, \citeauthor{Sunburn}}\label{app:sec:sliding:Sunburn}
\vspace{-.2cm}
\begin{figure}[h]
  \centering
  \begin{minipage}[t]{0.53\textwidth}
    \centering
    \vspace{0cm}
    \includegraphics[width=\linewidth]{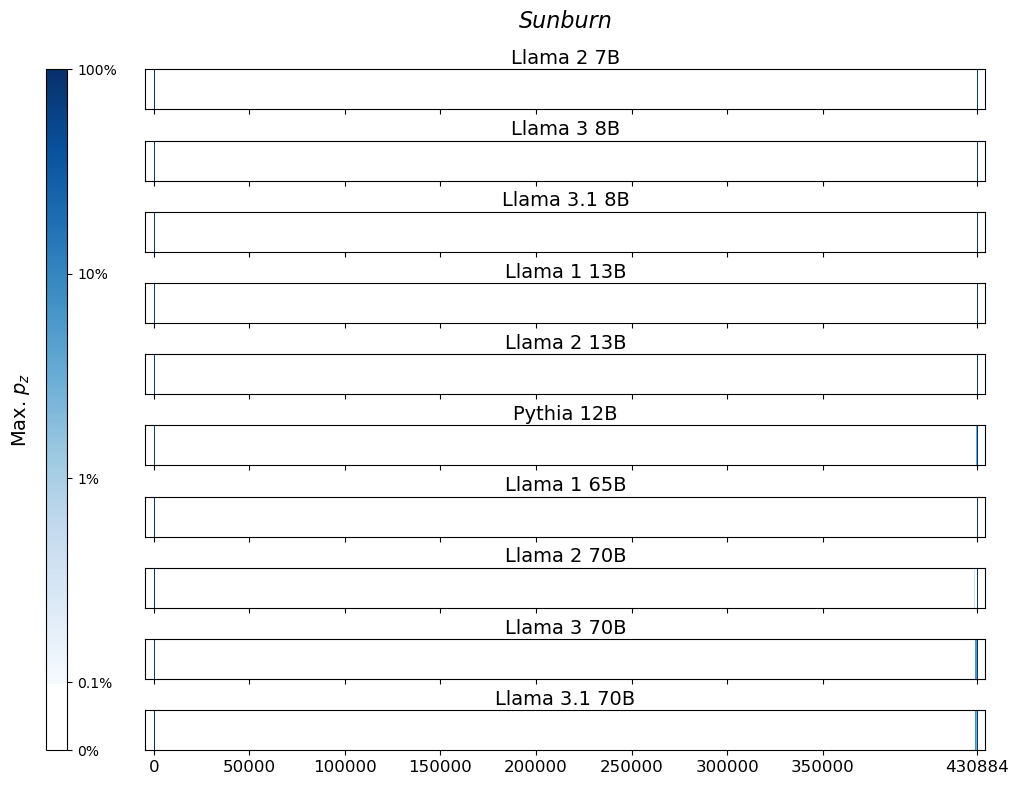}
    \includegraphics[width=\linewidth]{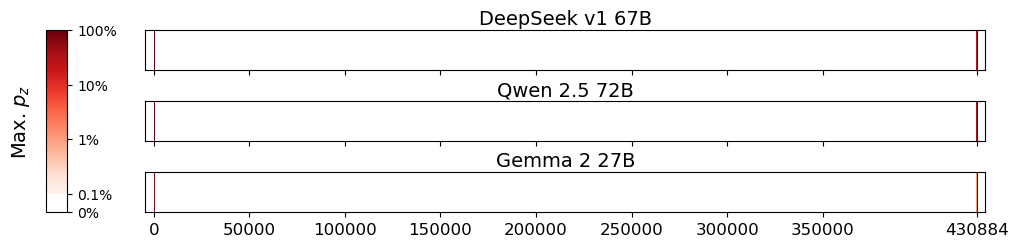}
    \includegraphics[width=\linewidth]{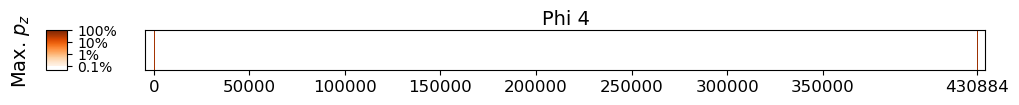}
  \end{minipage}
  \hfill
  \begin{minipage}[t]{0.45\textwidth}
    \centering
    \vspace{0cm}
    \includegraphics[width=\linewidth]{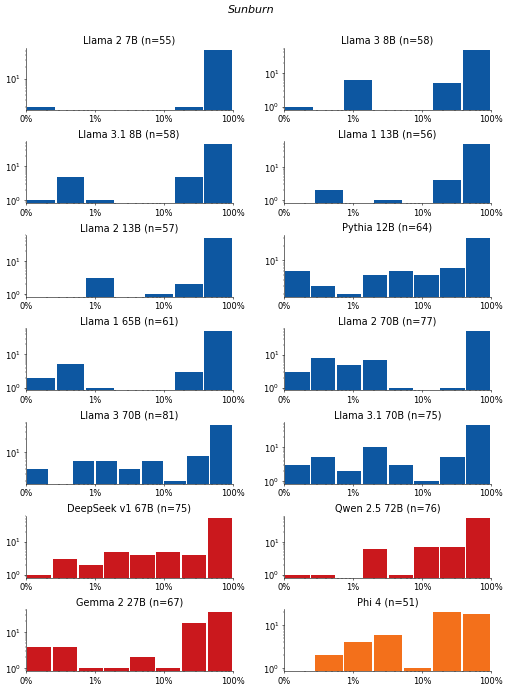}
  \end{minipage}
  \vspace{-.2cm}
  \caption{
    \textbf{\textit{Sunburn}, \citeauthor{Sunburn}.}
    For $14$ LLMs,
    (\textbf{left}) heatmaps for the sliding-window procedure and
    (\textbf{right}) corresponding distributions over suffix extraction probabilities
    ($\tau_\text{min}=0.1\%$).
  }
  \label{fig:slidingwindow:Sunburn}
\end{figure}
\FloatBarrier

\clearpage
\subsubsection{\textit{Marvel's Spider-Man: Hostile Takeover}, \citeauthor{Marvel_s_Spider-Man_Hostile_Takeover}}\label{app:sec:sliding:Marvel_s_Spider-Man_Hostile_Takeover}
\vspace{-.2cm}
\begin{figure}[h]
  \centering
  \begin{minipage}[t]{0.53\textwidth}
    \centering
    \vspace{0cm}
    \includegraphics[width=\linewidth]{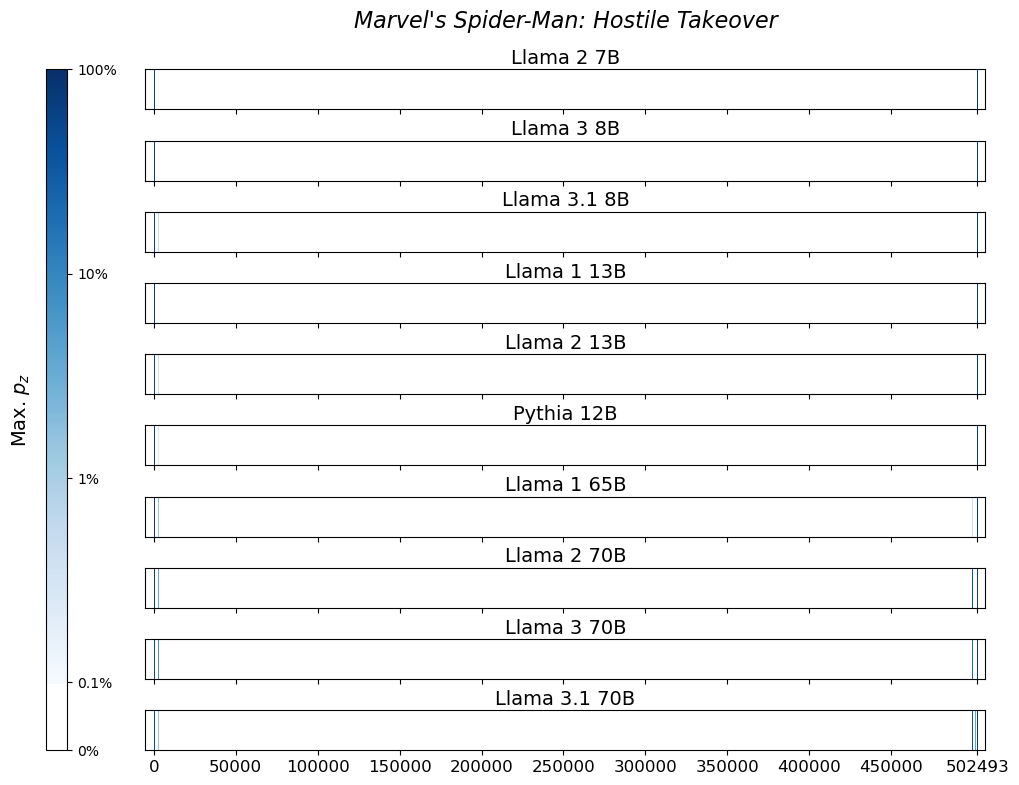}
    \includegraphics[width=\linewidth]{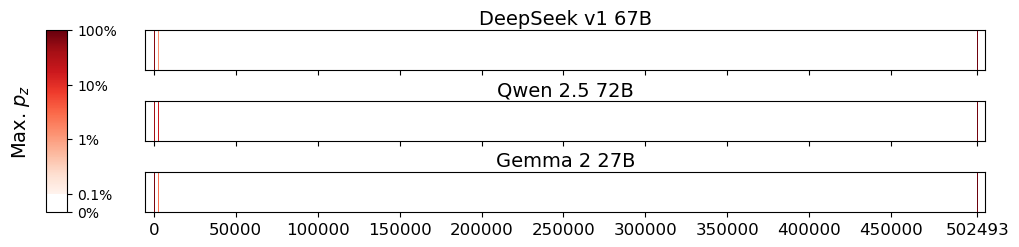}
    \includegraphics[width=\linewidth]{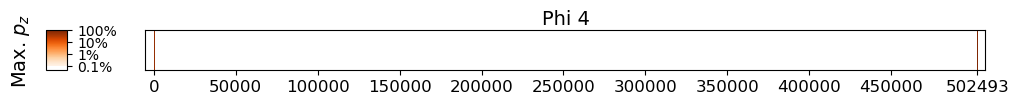}
  \end{minipage}
  \hfill
  \begin{minipage}[t]{0.45\textwidth}
    \centering
    \vspace{0cm}
    \includegraphics[width=\linewidth]{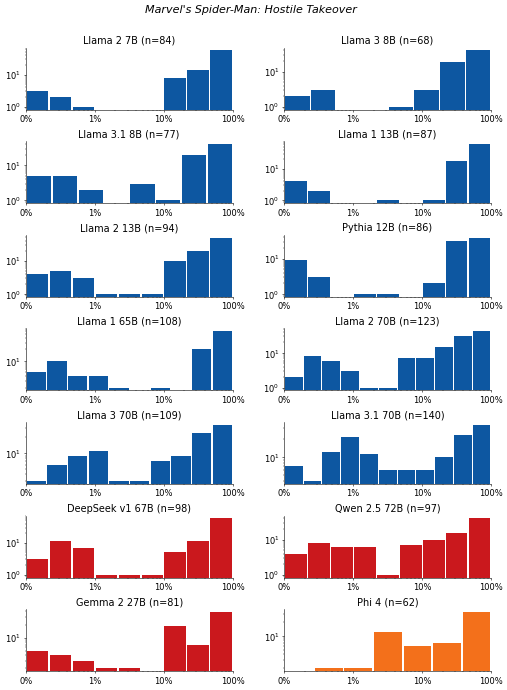}
  \end{minipage}
  \vspace{-.2cm}
  \caption{
    \textbf{\textit{Marvel's Spider-Man: Hostile Takeover}, \citeauthor{Marvel_s_Spider-Man_Hostile_Takeover}.}
    For $14$ LLMs,
    (\textbf{left}) heatmaps for the sliding-window procedure and
    (\textbf{right}) corresponding distributions over suffix extraction probabilities
    ($\tau_\text{min}=0.1\%$).
  }
  \label{fig:slidingwindow:Marvel_s_Spider-Man_Hostile_Takeover}
\end{figure}
\FloatBarrier

\subsubsection{\textit{Anastasia on Her Own}, \citeauthor{Anastasia_on_Her_Own}}\label{app:sec:sliding:Anastasia_on_Her_Own}
\vspace{-.2cm}
\begin{figure}[h]
  \centering
  \begin{minipage}[t]{0.53\textwidth}
    \centering
    \vspace{0cm}
    \includegraphics[width=\linewidth]{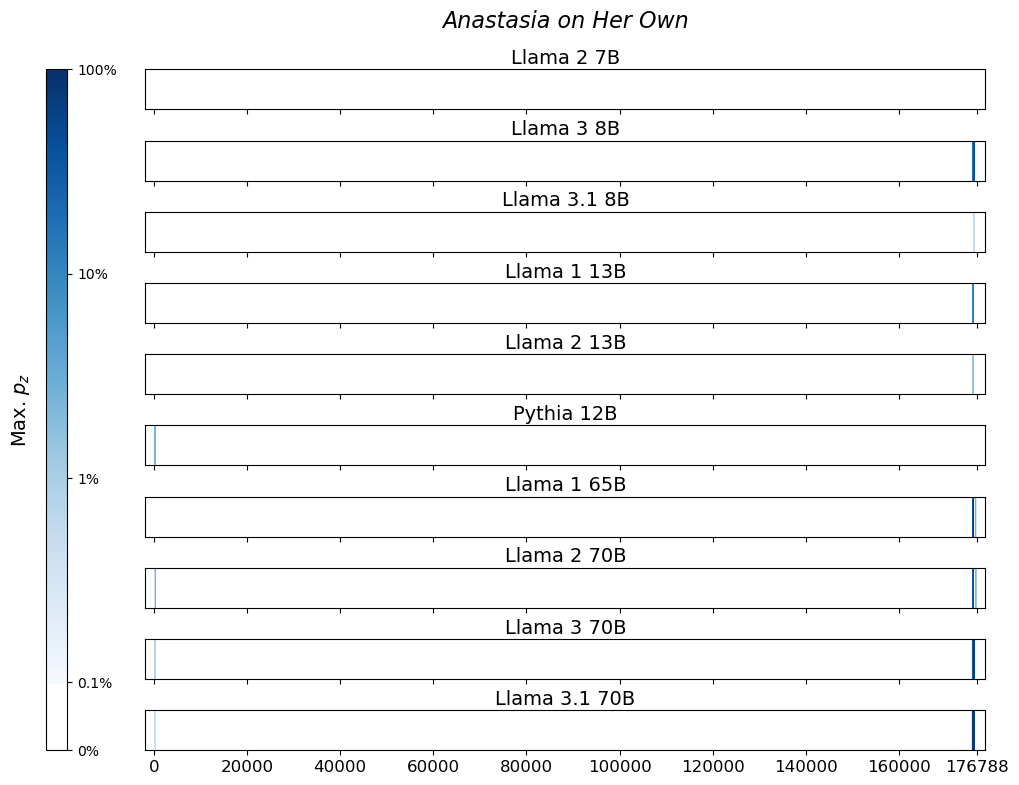}
    \includegraphics[width=\linewidth]{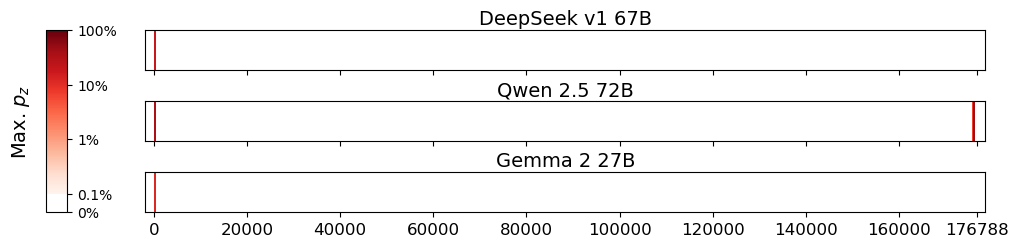}
    \includegraphics[width=\linewidth]{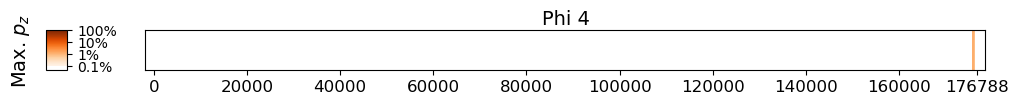}
  \end{minipage}
  \hfill
  \begin{minipage}[t]{0.45\textwidth}
    \centering
    \vspace{0cm}
    \includegraphics[width=\linewidth]{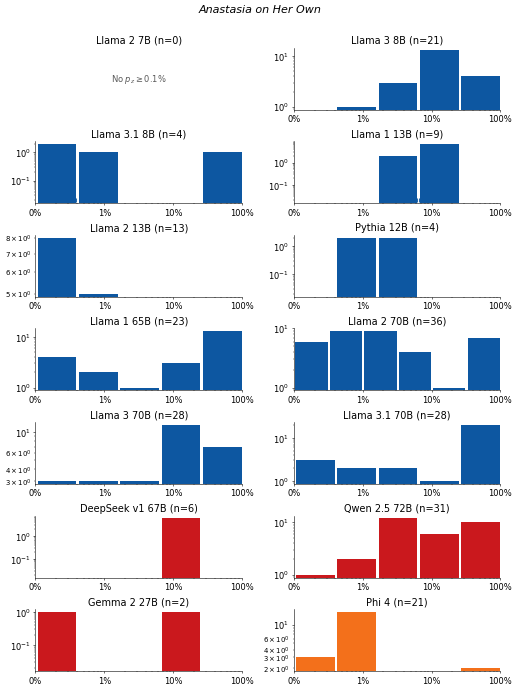}
  \end{minipage}
  \vspace{-.2cm}
  \caption{
    \textbf{\textit{Anastasia on Her Own}, \citeauthor{Anastasia_on_Her_Own}.}
    For $14$ LLMs,
    (\textbf{left}) heatmaps for the sliding-window procedure and
    (\textbf{right}) corresponding distributions over suffix extraction probabilities
    ($\tau_\text{min}=0.1\%$).
  }
  \label{fig:slidingwindow:Anastasia_on_Her_Own}
\end{figure}
\FloatBarrier

\clearpage
\subsubsection{\textit{Nixon in China}, \citeauthor{Nixon_in_China}}\label{app:sec:sliding:Nixon_in_China}
\vspace{-.2cm}
\begin{figure}[h]
  \centering
  \begin{minipage}[t]{0.53\textwidth}
    \centering
    \vspace{0cm}
    \includegraphics[width=\linewidth]{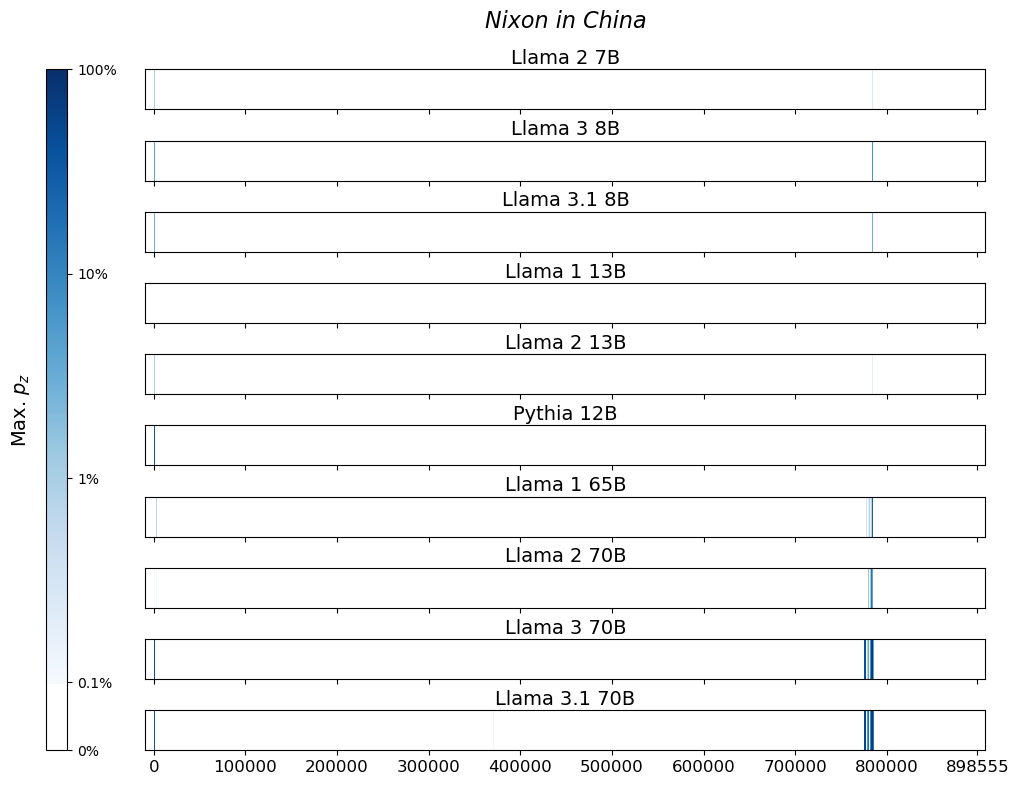}
    \includegraphics[width=\linewidth]{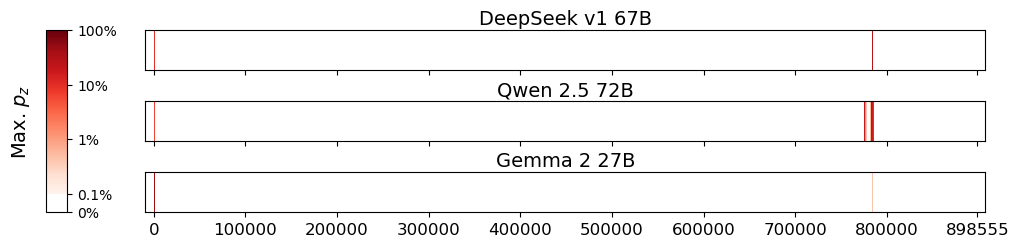}
    \includegraphics[width=\linewidth]{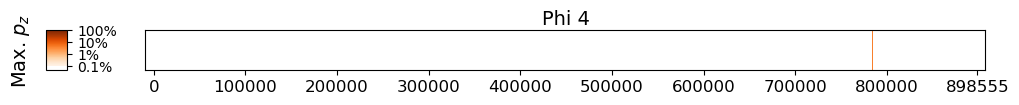}
  \end{minipage}
  \hfill
  \begin{minipage}[t]{0.45\textwidth}
    \centering
    \vspace{0cm}
    \includegraphics[width=\linewidth]{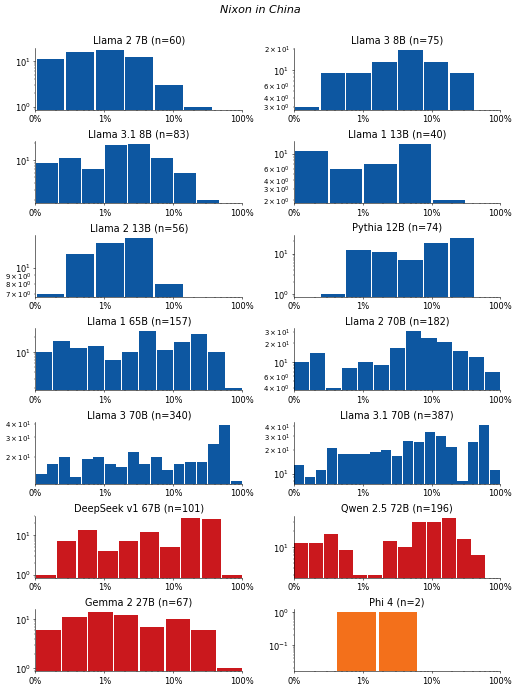}
  \end{minipage}
  \vspace{-.2cm}
  \caption{
    \textbf{\textit{Nixon in China}, \citeauthor{Nixon_in_China}.}
    For $14$ LLMs,
    (\textbf{left}) heatmaps for the sliding-window procedure and
    (\textbf{right}) corresponding distributions over suffix extraction probabilities
    ($\tau_\text{min}=0.1\%$).
  }
  \label{fig:slidingwindow:Nixon_in_China}
\end{figure}
\FloatBarrier

\subsubsection{\textit{The Doomsday Prophecy}, \citeauthor{The_Doomsday_Prophecy}}\label{app:sec:sliding:The_Doomsday_Prophecy}
\vspace{-.2cm}
\begin{figure}[h]
  \centering
  \begin{minipage}[t]{0.53\textwidth}
    \centering
    \vspace{0cm}
    \includegraphics[width=\linewidth]{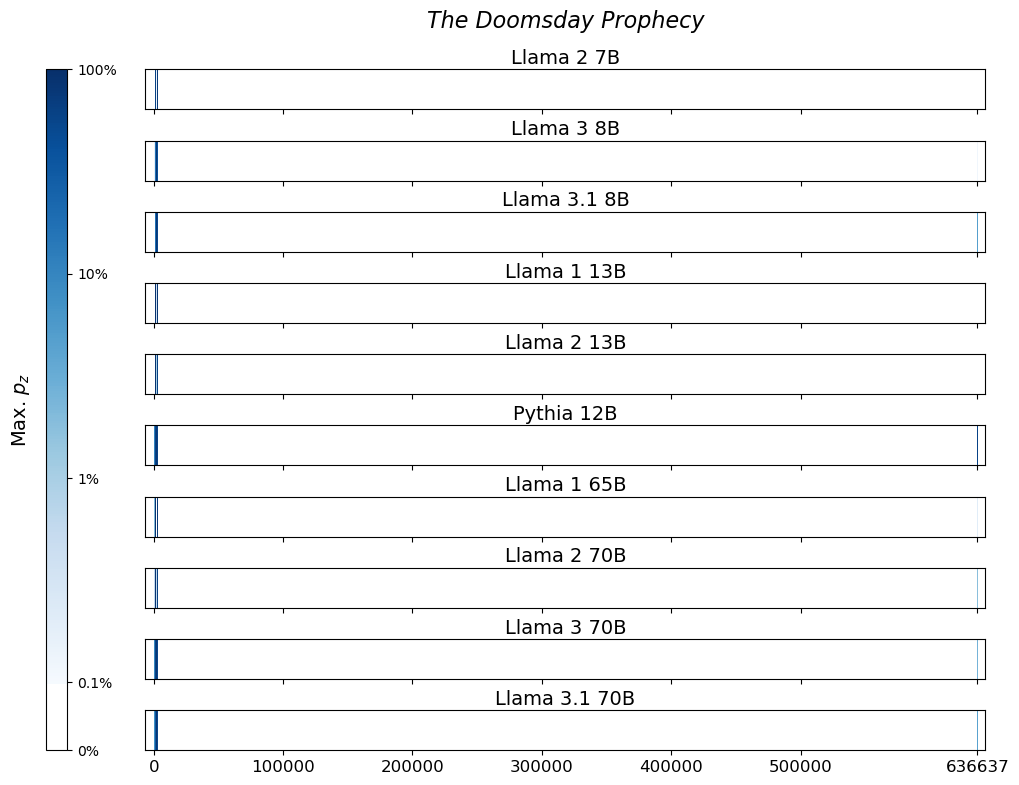}
    \includegraphics[width=\linewidth]{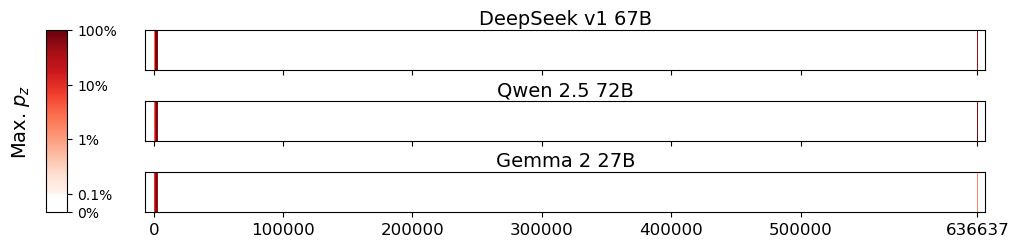}
    \includegraphics[width=\linewidth]{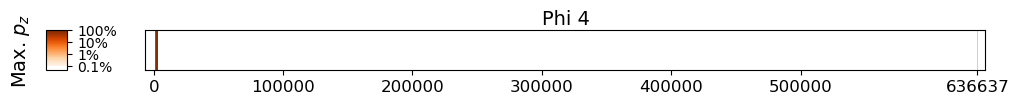}
  \end{minipage}
  \hfill
  \begin{minipage}[t]{0.45\textwidth}
    \centering
    \vspace{0cm}
    \includegraphics[width=\linewidth]{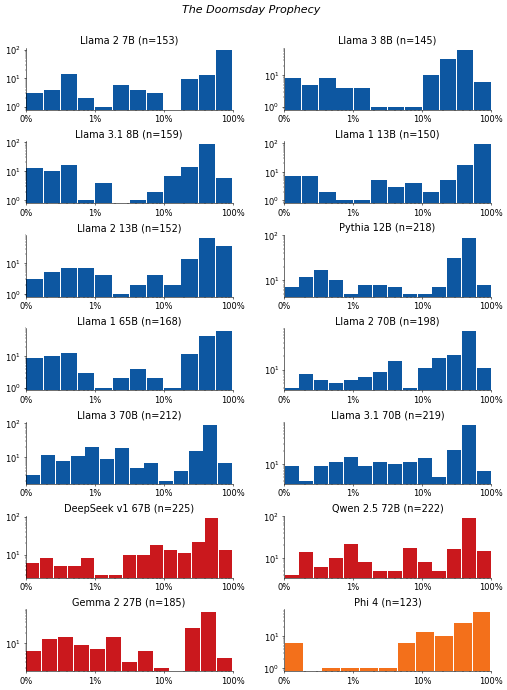}
  \end{minipage}
  \vspace{-.2cm}
  \caption{
    \textbf{\textit{The Doomsday Prophecy}, \citeauthor{The_Doomsday_Prophecy}.}
    For $14$ LLMs,
    (\textbf{left}) heatmaps for the sliding-window procedure and
    (\textbf{right}) corresponding distributions over suffix extraction probabilities
    ($\tau_\text{min}=0.1\%$).
  }
  \label{fig:slidingwindow:The_Doomsday_Prophecy}
\end{figure}
\FloatBarrier

\clearpage
\subsubsection{\textit{A Game of Thrones}, \citeauthor{A_Game_of_Thrones}}\label{app:sec:sliding:A_Game_of_Thrones}
\vspace{-.2cm}
\begin{figure}[h]
  \centering
  \begin{minipage}[t]{0.53\textwidth}
    \centering
    \vspace{0cm}
    \includegraphics[width=\linewidth]{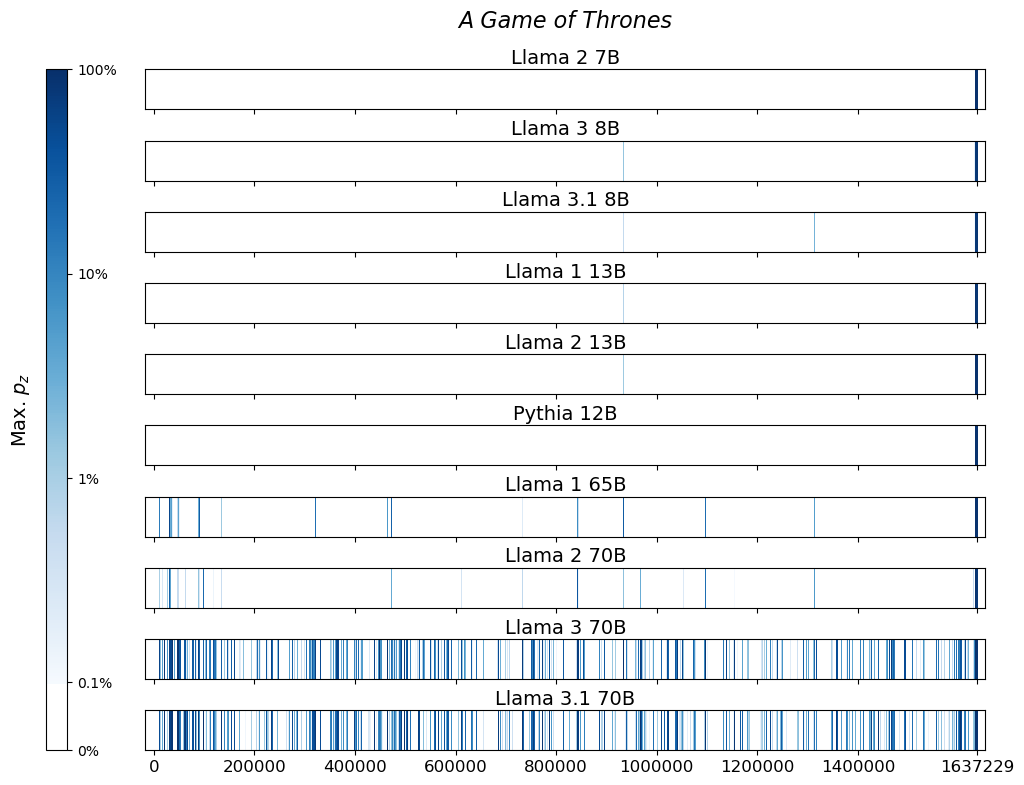}
    \includegraphics[width=\linewidth]{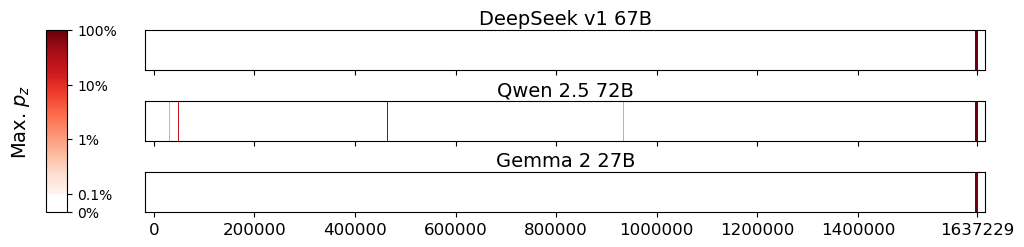}
    \includegraphics[width=\linewidth]{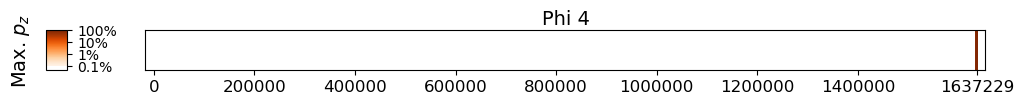}
  \end{minipage}
  \hfill
  \begin{minipage}[t]{0.45\textwidth}
    \centering
    \vspace{0cm}
    \includegraphics[width=\linewidth]{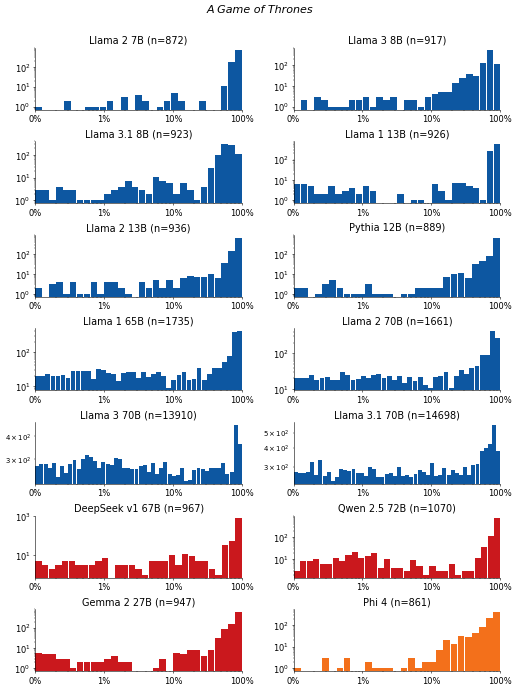}
  \end{minipage}
  \vspace{-.2cm}
  \caption{
    \textbf{\textit{A Game of Thrones}, \citeauthor{A_Game_of_Thrones}.}
    For $14$ LLMs,
    (\textbf{left}) heatmaps for the sliding-window procedure and
    (\textbf{right}) corresponding distributions over suffix extraction probabilities
    ($\tau_\text{min}=0.1\%$).
  }
  \label{fig:slidingwindow:A_Game_of_Thrones}
\end{figure}
\FloatBarrier

\subsubsection{\textit{Rough-Hewn Land}, \citeauthor{Rough-Hewn_Land}}\label{app:sec:sliding:Rough-Hewn_Land}
\vspace{-.2cm}
\begin{figure}[h]
  \centering
  \begin{minipage}[t]{0.53\textwidth}
    \centering
    \vspace{0cm}
    \includegraphics[width=\linewidth]{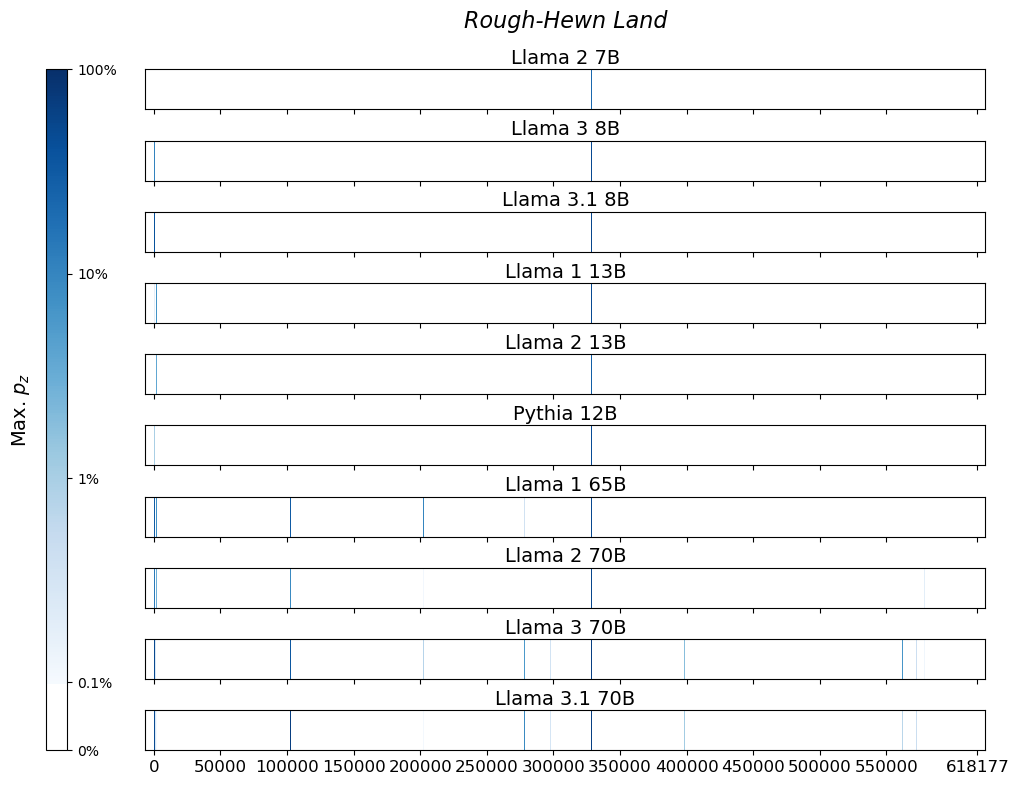}
    \includegraphics[width=\linewidth]{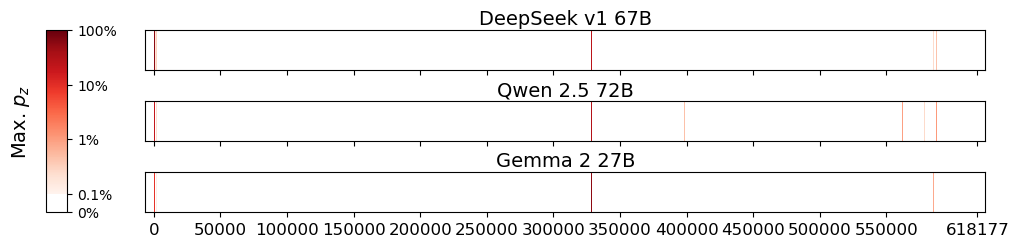}
    \includegraphics[width=\linewidth]{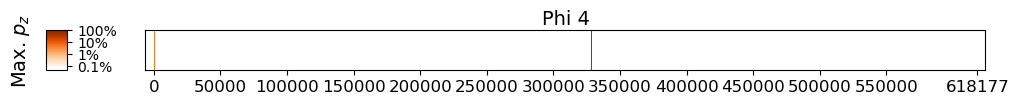}
  \end{minipage}
  \hfill
  \begin{minipage}[t]{0.45\textwidth}
    \centering
    \vspace{0cm}
    \includegraphics[width=\linewidth]{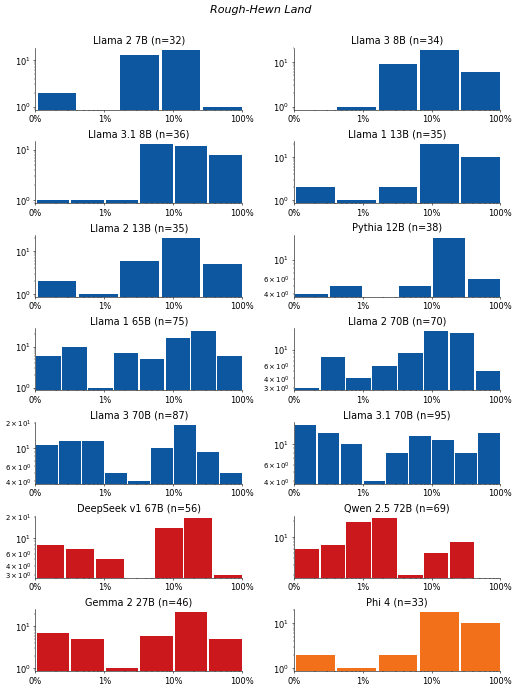}
  \end{minipage}
  \vspace{-.2cm}
  \caption{
    \textbf{\textit{Rough-Hewn Land}, \citeauthor{Rough-Hewn_Land}.}
    For $14$ LLMs,
    (\textbf{left}) heatmaps for the sliding-window procedure and
    (\textbf{right}) corresponding distributions over suffix extraction probabilities
    ($\tau_\text{min}=0.1\%$).
  }
  \label{fig:slidingwindow:Rough-Hewn_Land}
\end{figure}
\FloatBarrier

\clearpage
\subsubsection{\textit{Twilight}, \citeauthor{Twilight}}\label{app:sec:sliding:Twilight}
\vspace{-.2cm}
\begin{figure}[h]
  \centering
  \begin{minipage}[t]{0.53\textwidth}
    \centering
    \vspace{0cm}
    \includegraphics[width=\linewidth]{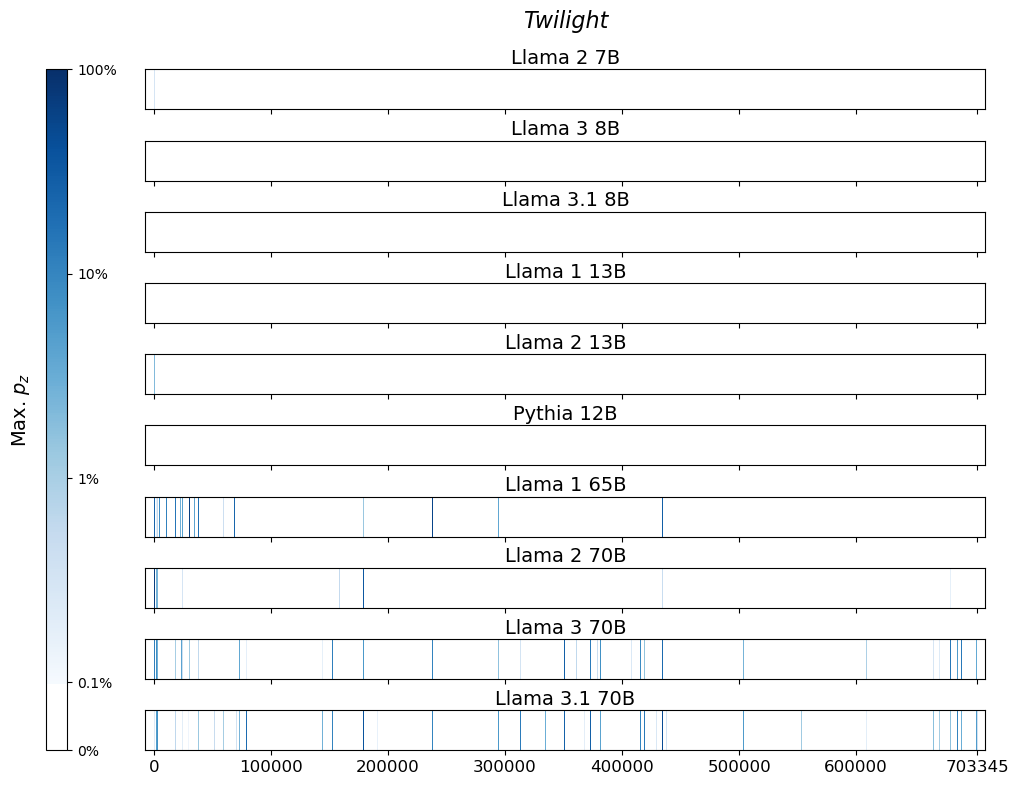}
    \includegraphics[width=\linewidth]{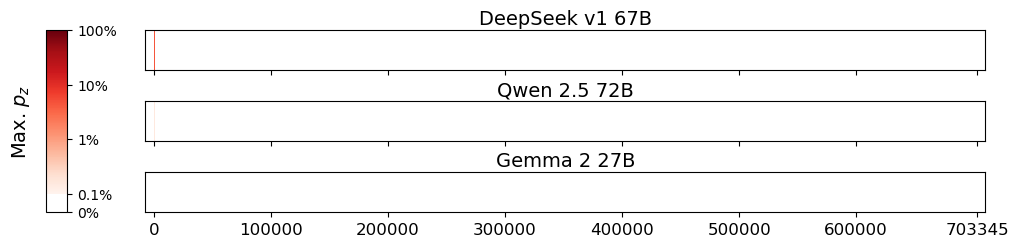}
    \includegraphics[width=\linewidth]{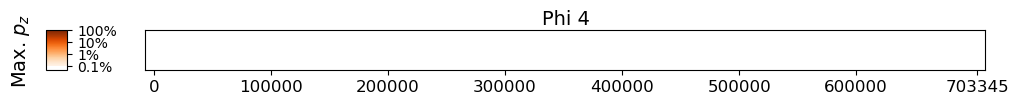}
  \end{minipage}
  \hfill
  \begin{minipage}[t]{0.45\textwidth}
    \centering
    \vspace{0cm}
    \includegraphics[width=\linewidth]{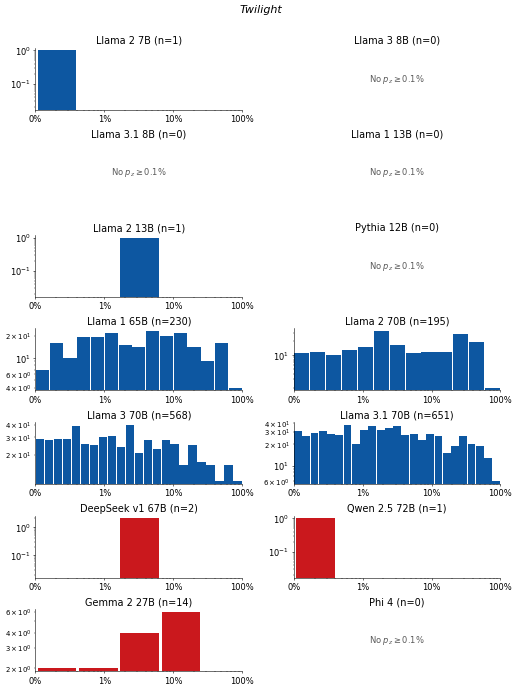}
  \end{minipage}
  \vspace{-.2cm}
  \caption{
    \textbf{\textit{Twilight}, \citeauthor{Twilight}.}
    For $14$ LLMs,
    (\textbf{left}) heatmaps for the sliding-window procedure and
    (\textbf{right}) corresponding distributions over suffix extraction probabilities
    ($\tau_\text{min}=0.1\%$).
  }
  \label{fig:slidingwindow:Twilight}
\end{figure}
\FloatBarrier

\subsubsection{\textit{The Duchess War}, \citeauthor{The_Duchess_War}}\label{app:sec:sliding:The_Duchess_War}
\vspace{-.2cm}
\begin{figure}[h]
  \centering
  \begin{minipage}[t]{0.53\textwidth}
    \centering
    \vspace{0cm}
    \includegraphics[width=\linewidth]{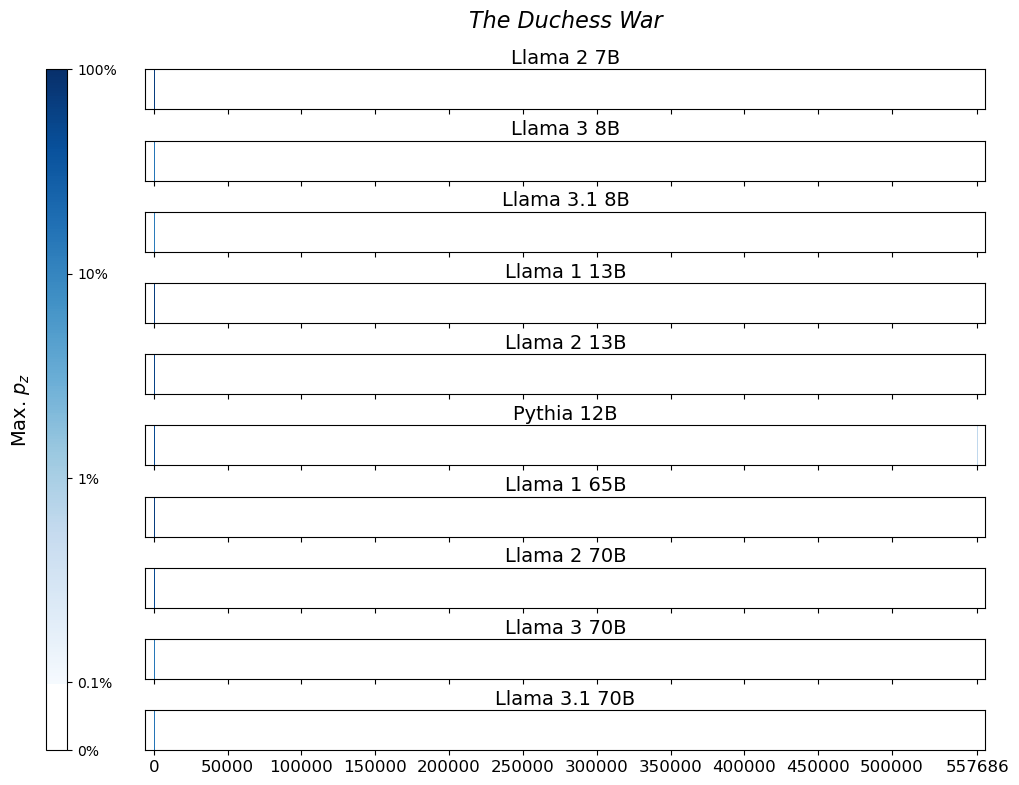}
    \includegraphics[width=\linewidth]{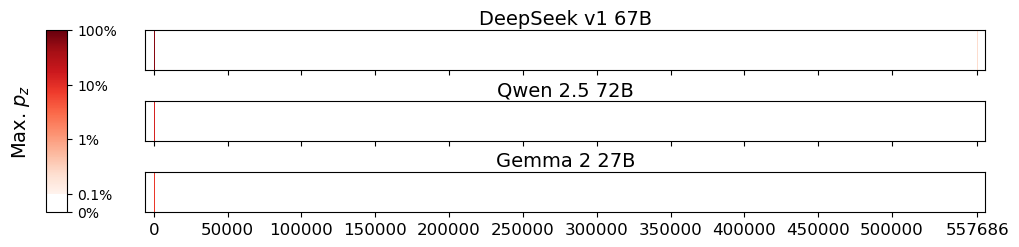}
    \includegraphics[width=\linewidth]{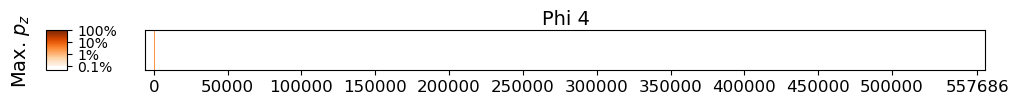}
  \end{minipage}
  \hfill
  \begin{minipage}[t]{0.45\textwidth}
    \centering
    \vspace{0cm}
    \includegraphics[width=\linewidth]{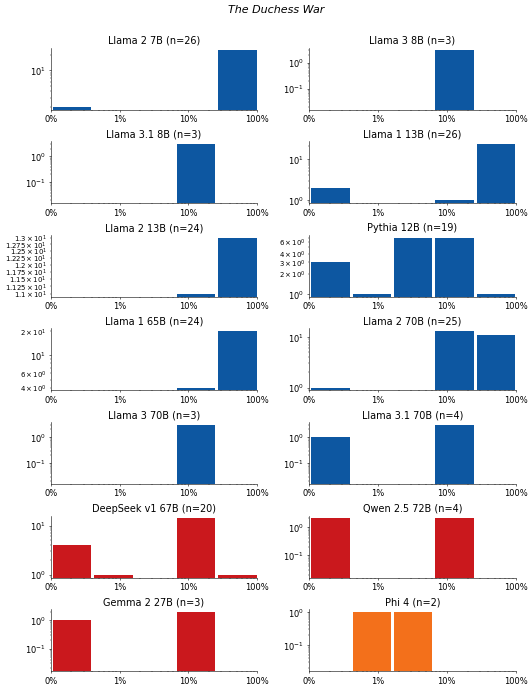}
  \end{minipage}
  \vspace{-.2cm}
  \caption{
    \textbf{\textit{The Duchess War}, \citeauthor{The_Duchess_War}.}
    For $14$ LLMs,
    (\textbf{left}) heatmaps for the sliding-window procedure and
    (\textbf{right}) corresponding distributions over suffix extraction probabilities
    ($\tau_\text{min}=0.1\%$).
  }
  \label{fig:slidingwindow:The_Duchess_War}
\end{figure}
\FloatBarrier

\clearpage
\subsubsection{\textit{On Liberty}, \citeauthor{On_Liberty}}\label{app:sec:sliding:On_Liberty}
\vspace{-.2cm}
\begin{figure}[h]
  \centering
  \begin{minipage}[t]{0.53\textwidth}
    \centering
    \vspace{0cm}
    \includegraphics[width=\linewidth]{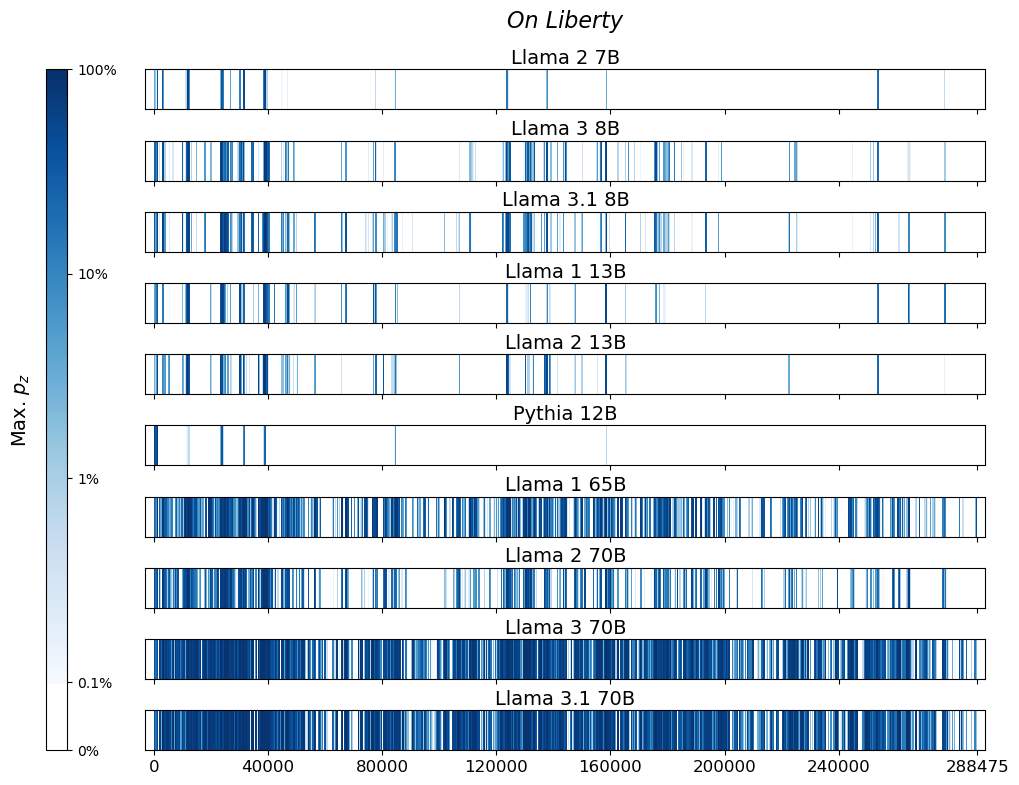}
    \includegraphics[width=\linewidth]{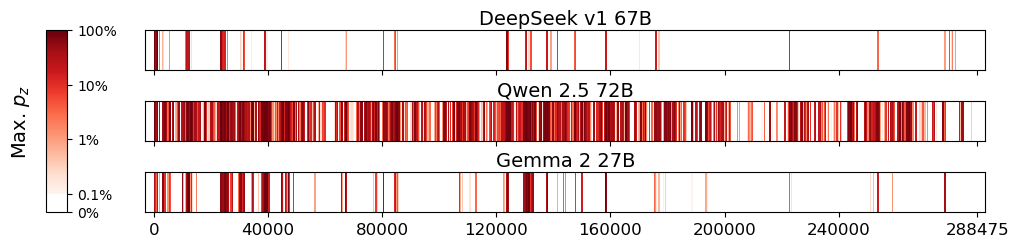}
    \includegraphics[width=\linewidth]{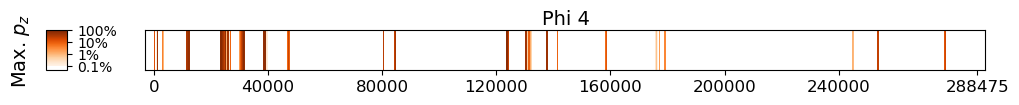}
  \end{minipage}
  \hfill
  \begin{minipage}[t]{0.45\textwidth}
    \centering
    \vspace{0cm}
    \includegraphics[width=\linewidth]{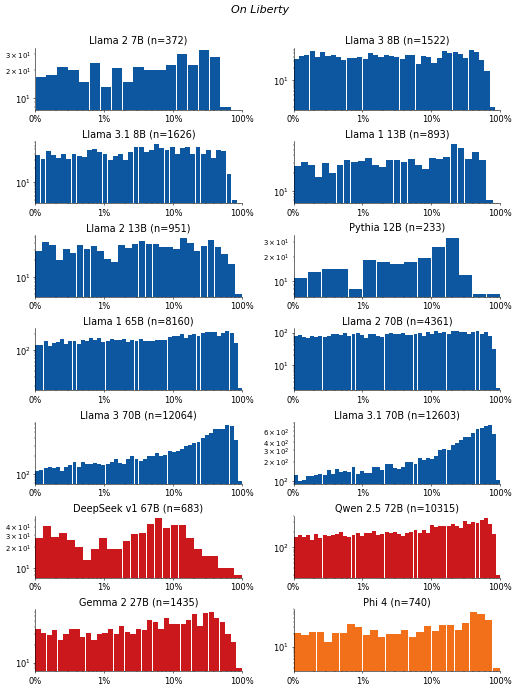}
  \end{minipage}
  \vspace{-.2cm}
  \caption{
    \textbf{\textit{On Liberty}, \citeauthor{On_Liberty}.}
    For $14$ LLMs,
    (\textbf{left}) heatmaps for the sliding-window procedure and
    (\textbf{right}) corresponding distributions over suffix extraction probabilities
    ($\tau_\text{min}=0.1\%$).
  }
  \label{fig:slidingwindow:On_Liberty}
\end{figure}
\FloatBarrier

\subsubsection{\textit{Coal Creek}, \citeauthor{Coal_Creek}}\label{app:sec:sliding:Coal_Creek}
\vspace{-.2cm}
\begin{figure}[h]
  \centering
  \begin{minipage}[t]{0.53\textwidth}
    \centering
    \vspace{0cm}
    \includegraphics[width=\linewidth]{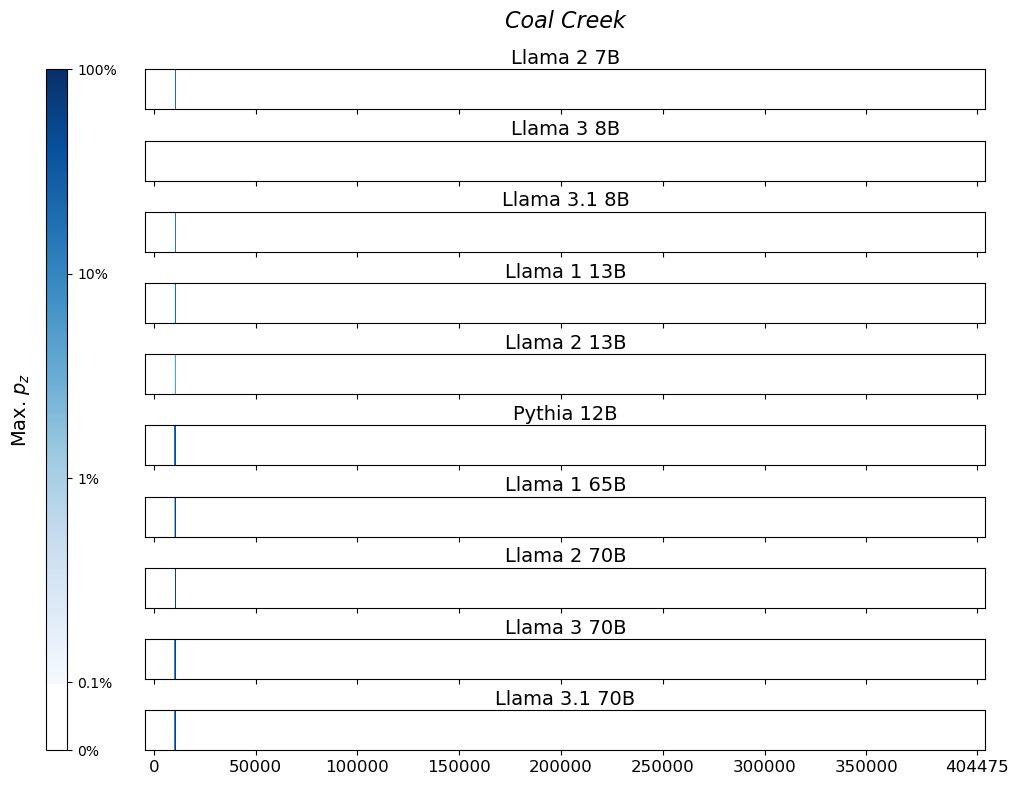}
    \includegraphics[width=\linewidth]{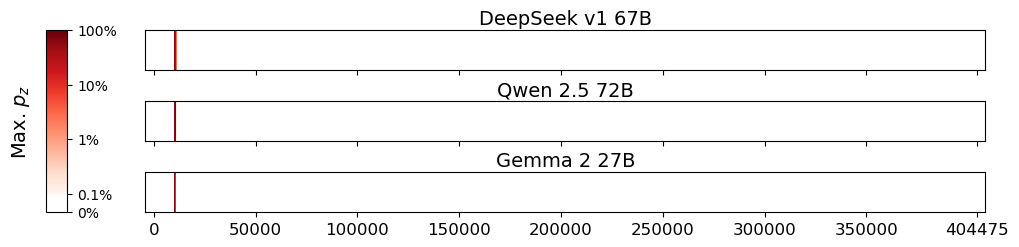}
    \includegraphics[width=\linewidth]{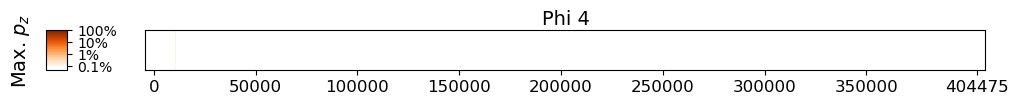}
  \end{minipage}
  \hfill
  \begin{minipage}[t]{0.45\textwidth}
    \centering
    \vspace{0cm}
    \includegraphics[width=\linewidth]{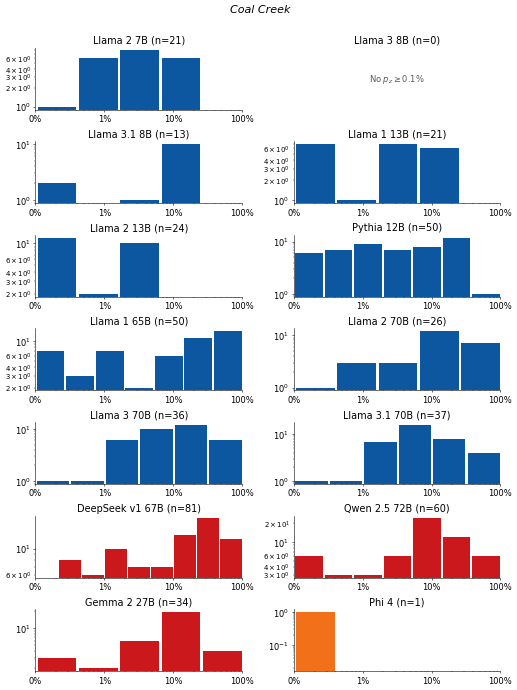}
  \end{minipage}
  \vspace{-.2cm}
  \caption{
    \textbf{\textit{Coal Creek}, \citeauthor{Coal_Creek}.}
    For $14$ LLMs,
    (\textbf{left}) heatmaps for the sliding-window procedure and
    (\textbf{right}) corresponding distributions over suffix extraction probabilities
    ($\tau_\text{min}=0.1\%$).
  }
  \label{fig:slidingwindow:Coal_Creek}
\end{figure}
\FloatBarrier

\clearpage
\subsubsection{\textit{Winnie the Pooh}, \citeauthor{Winnie_the_Pooh}}\label{app:sec:sliding:Winnie_the_Pooh}
\vspace{-.2cm}
\begin{figure}[h]
  \centering
  \begin{minipage}[t]{0.53\textwidth}
    \centering
    \vspace{0cm}
    \includegraphics[width=\linewidth]{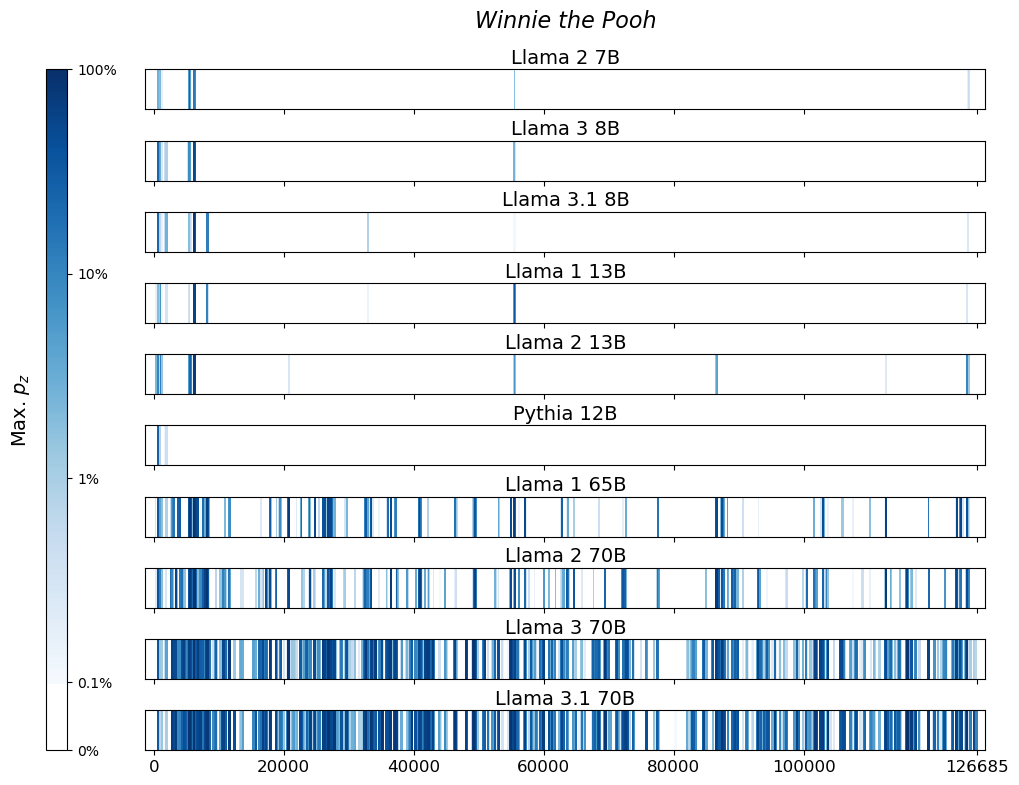}
    \includegraphics[width=\linewidth]{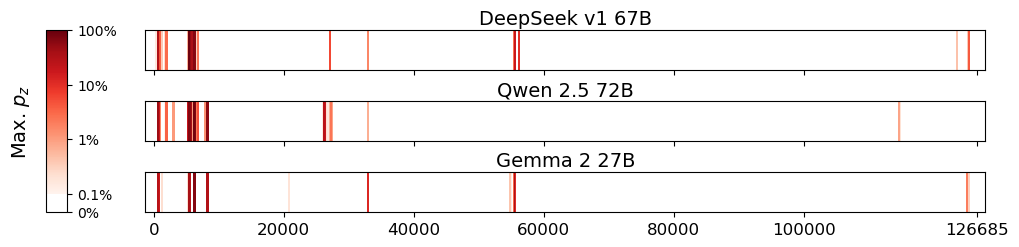}
    \includegraphics[width=\linewidth]{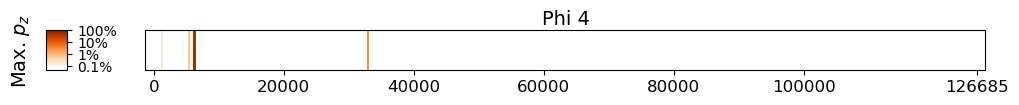}
  \end{minipage}
  \hfill
  \begin{minipage}[t]{0.45\textwidth}
    \centering
    \vspace{0cm}
    \includegraphics[width=\linewidth]{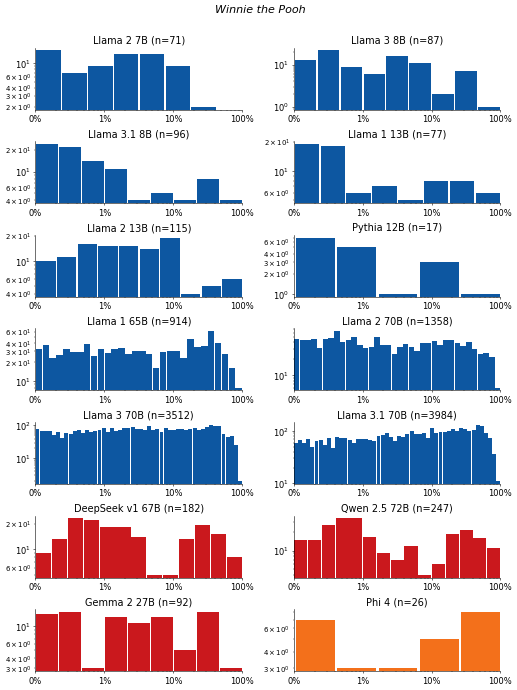}
  \end{minipage}
  \vspace{-.2cm}
  \caption{
    \textbf{\textit{Winnie the Pooh}, \citeauthor{Winnie_the_Pooh}.}
    For $14$ LLMs,
    (\textbf{left}) heatmaps for the sliding-window procedure and
    (\textbf{right}) corresponding distributions over suffix extraction probabilities
    ($\tau_\text{min}=0.1\%$).
  }
  \label{fig:slidingwindow:Winnie_the_Pooh}
\end{figure}
\FloatBarrier

\subsubsection{\textit{Catching the Sky}, \citeauthor{Catching_the_Sky}}\label{app:sec:sliding:Catching_the_Sky}
\vspace{-.2cm}
\begin{figure}[h]
  \centering
  \begin{minipage}[t]{0.53\textwidth}
    \centering
    \vspace{0cm}
    \includegraphics[width=\linewidth]{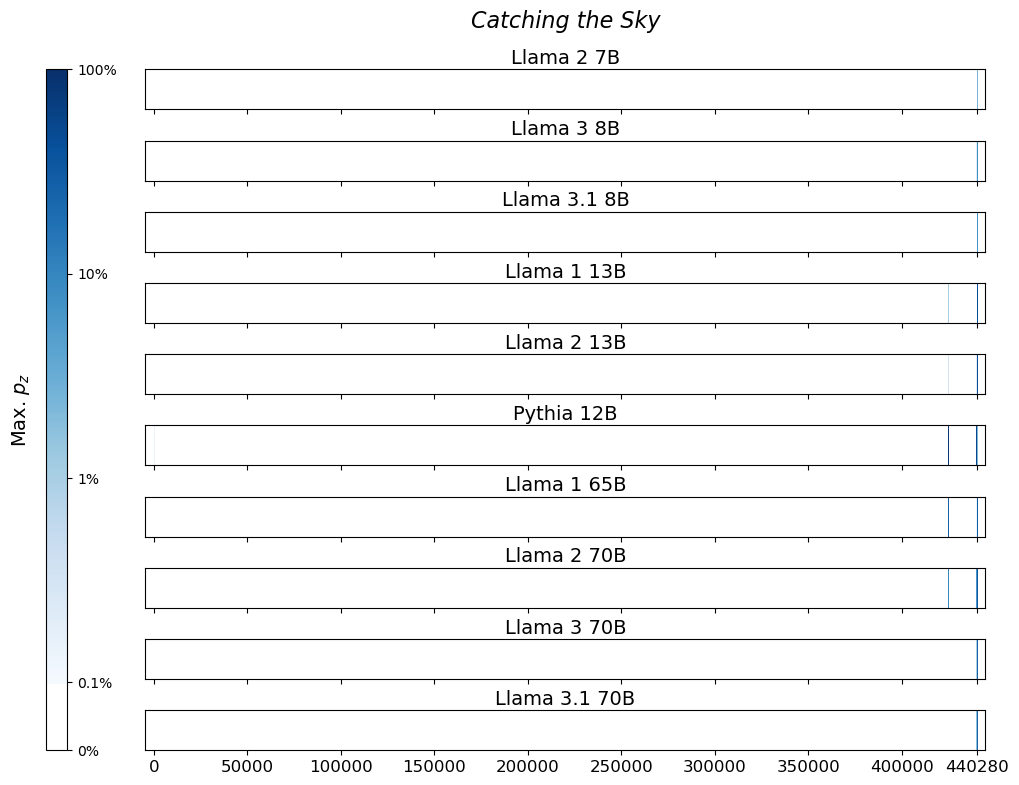}
    \includegraphics[width=\linewidth]{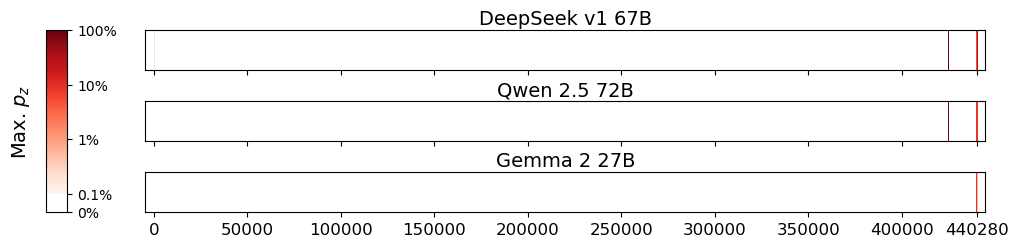}
    \includegraphics[width=\linewidth]{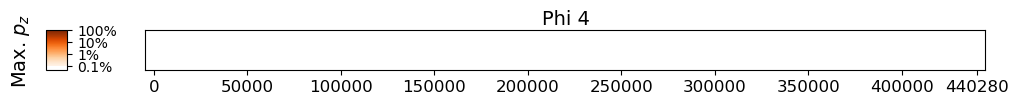}
  \end{minipage}
  \hfill
  \begin{minipage}[t]{0.45\textwidth}
    \centering
    \vspace{0cm}
    \includegraphics[width=\linewidth]{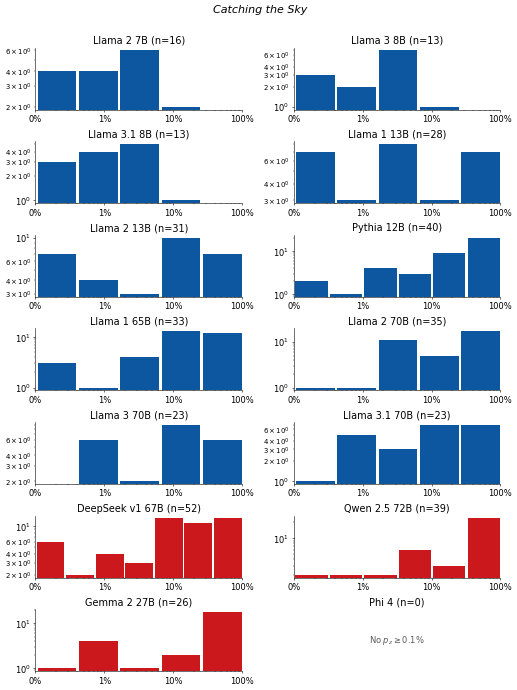}
  \end{minipage}
  \vspace{-.2cm}
  \caption{
    \textbf{\textit{Catching the Sky}, \citeauthor{Catching_the_Sky}.}
    For $14$ LLMs,
    (\textbf{left}) heatmaps for the sliding-window procedure and
    (\textbf{right}) corresponding distributions over suffix extraction probabilities
    ($\tau_\text{min}=0.1\%$).
  }
  \label{fig:slidingwindow:Catching_the_Sky}
\end{figure}
\FloatBarrier

\clearpage
\subsubsection{\textit{The Heretic Queen}, \citeauthor{The_Heretic_Queen}}\label{app:sec:sliding:The_Heretic_Queen}
\vspace{-.2cm}
\begin{figure}[h]
  \centering
  \begin{minipage}[t]{0.53\textwidth}
    \centering
    \vspace{0cm}
    \includegraphics[width=\linewidth]{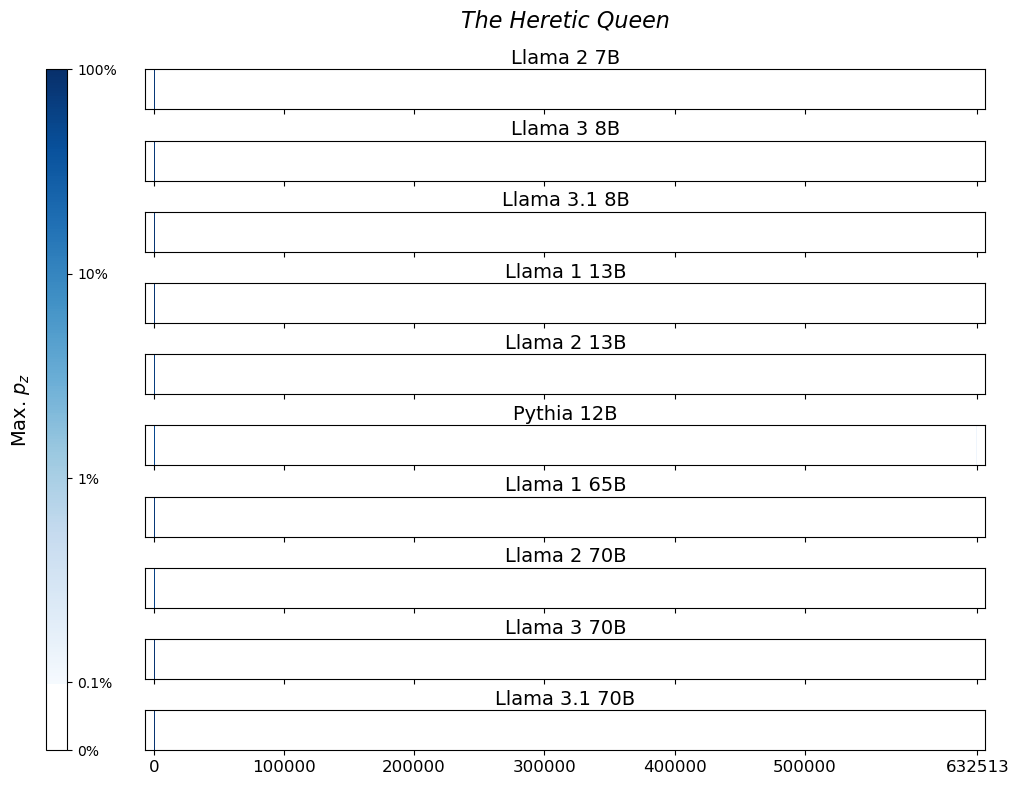}
    \includegraphics[width=\linewidth]{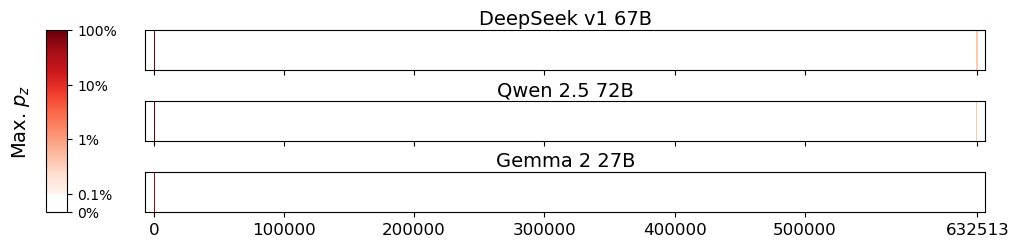}
    \includegraphics[width=\linewidth]{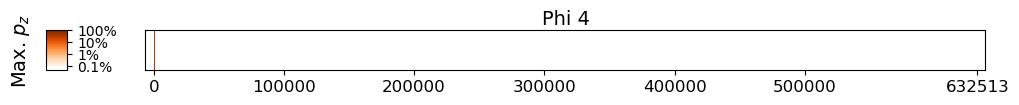}
  \end{minipage}
  \hfill
  \begin{minipage}[t]{0.45\textwidth}
    \centering
    \vspace{0cm}
    \includegraphics[width=\linewidth]{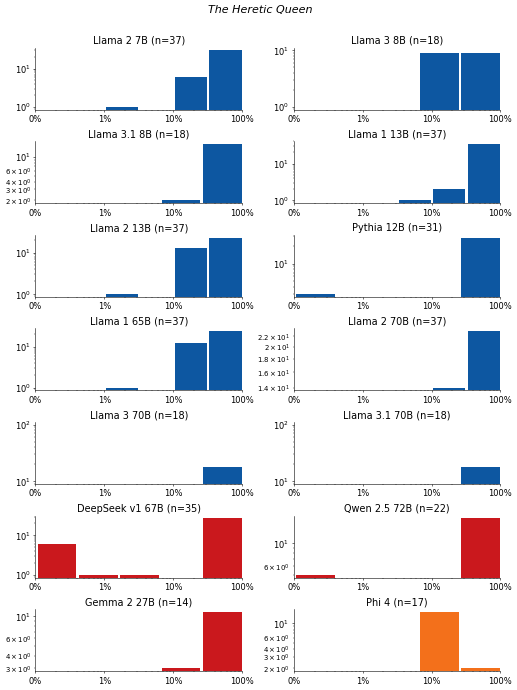}
  \end{minipage}
  \vspace{-.2cm}
  \caption{
    \textbf{\textit{The Heretic Queen}, \citeauthor{The_Heretic_Queen}.}
    For $14$ LLMs,
    (\textbf{left}) heatmaps for the sliding-window procedure and
    (\textbf{right}) corresponding distributions over suffix extraction probabilities
    ($\tau_\text{min}=0.1\%$).
  }
  \label{fig:slidingwindow:The_Heretic_Queen}
\end{figure}
\FloatBarrier

\subsubsection{\textit{The Gondola Maker}, \citeauthor{The_Gondola_Maker}}\label{app:sec:sliding:The_Gondola_Maker}
\vspace{-.2cm}
\begin{figure}[h]
  \centering
  \begin{minipage}[t]{0.53\textwidth}
    \centering
    \vspace{0cm}
    \includegraphics[width=\linewidth]{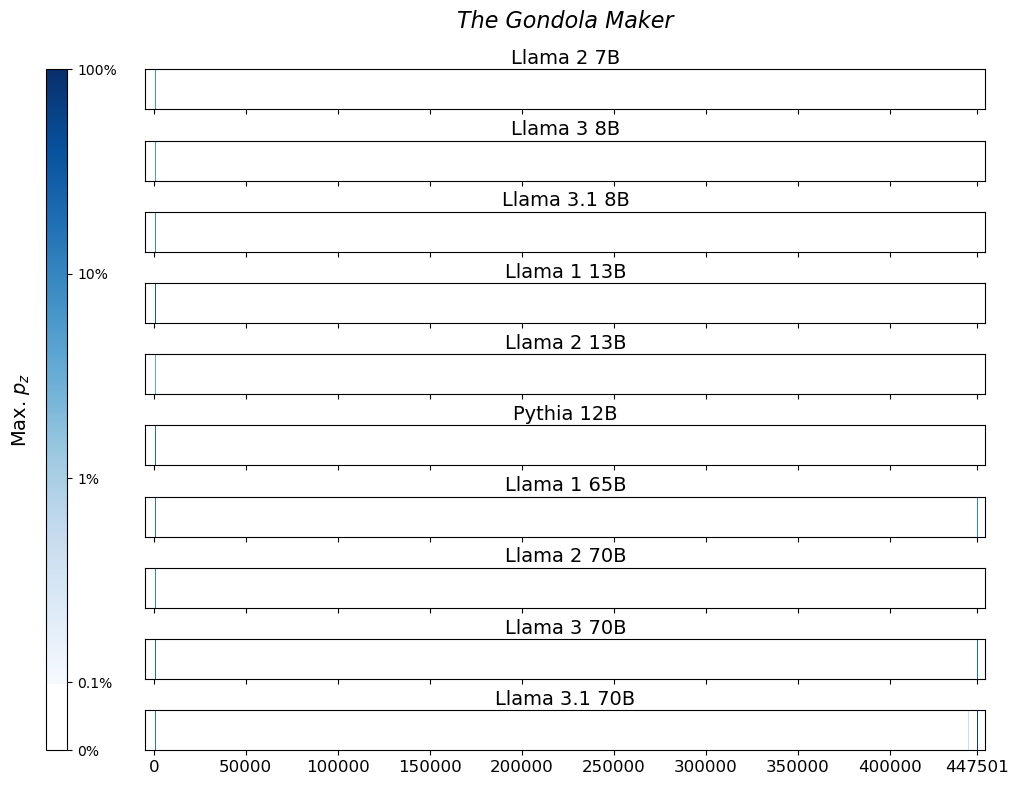}
    \includegraphics[width=\linewidth]{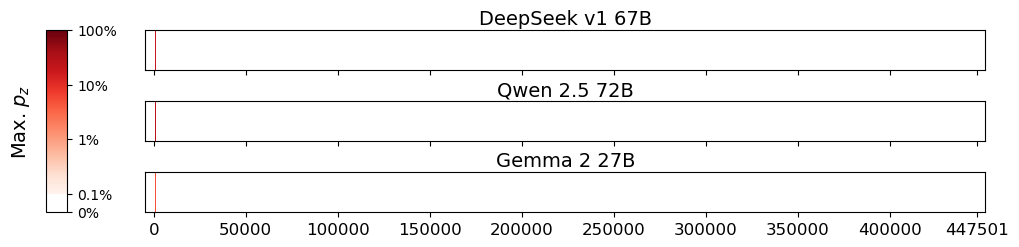}
    \includegraphics[width=\linewidth]{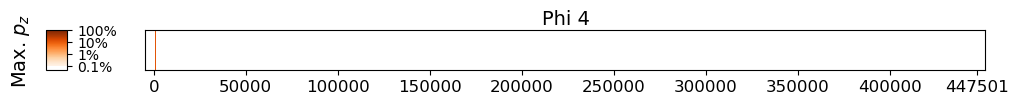}
  \end{minipage}
  \hfill
  \begin{minipage}[t]{0.45\textwidth}
    \centering
    \vspace{0cm}
    \includegraphics[width=\linewidth]{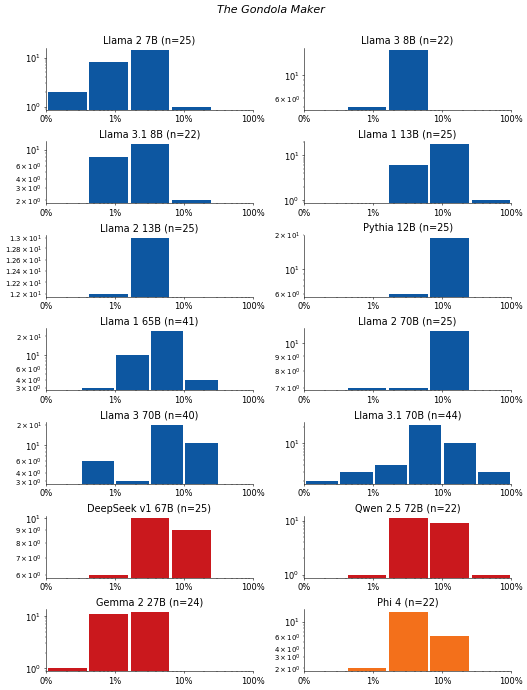}
  \end{minipage}
  \vspace{-.2cm}
  \caption{
    \textbf{\textit{The Gondola Maker}, \citeauthor{The_Gondola_Maker}.}
    For $14$ LLMs,
    (\textbf{left}) heatmaps for the sliding-window procedure and
    (\textbf{right}) corresponding distributions over suffix extraction probabilities
    ($\tau_\text{min}=0.1\%$).
  }
  \label{fig:slidingwindow:The_Gondola_Maker}
\end{figure}
\FloatBarrier

\clearpage
\subsubsection{\textit{Songs in Ordinary Time}, \citeauthor{Songs_in_Ordinary_Time}}\label{app:sec:sliding:Songs_in_Ordinary_Time}
\vspace{-.2cm}
\begin{figure}[h]
  \centering
  \begin{minipage}[t]{0.53\textwidth}
    \centering
    \vspace{0cm}
    \includegraphics[width=\linewidth]{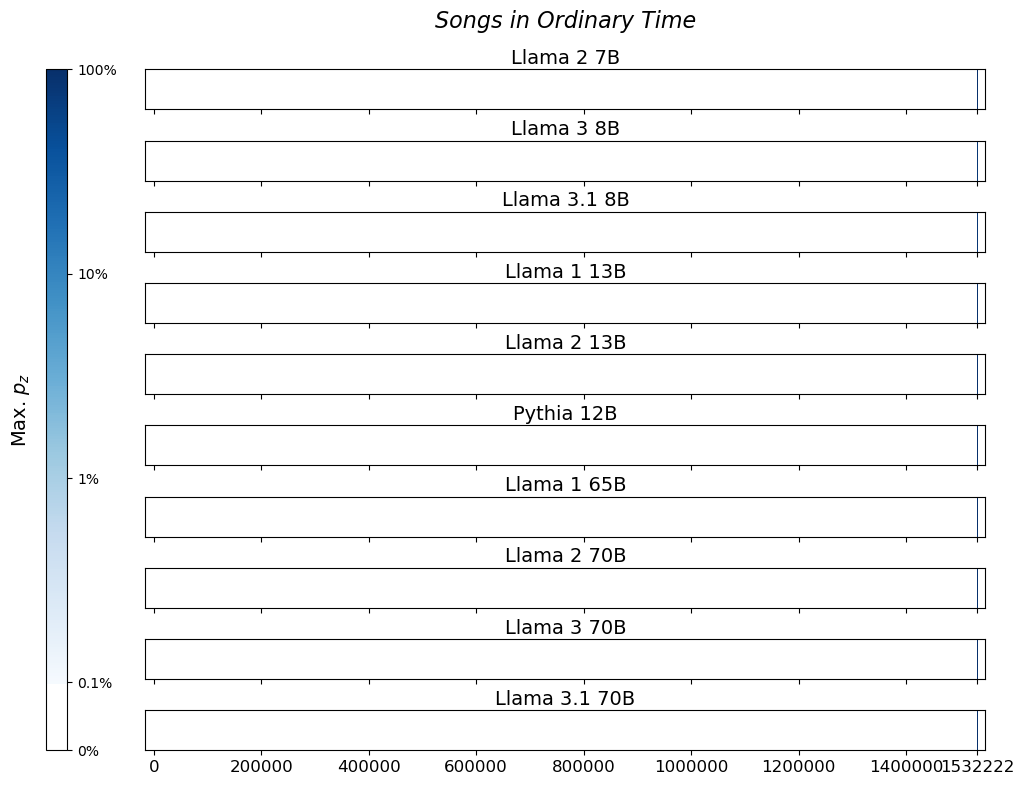}
    \includegraphics[width=\linewidth]{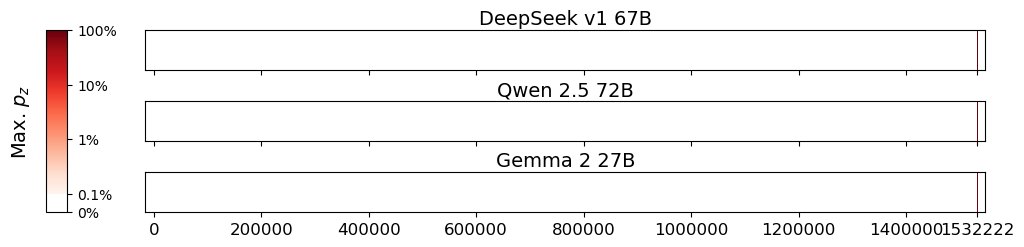}
    \includegraphics[width=\linewidth]{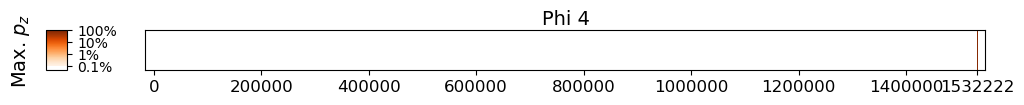}
  \end{minipage}
  \hfill
  \begin{minipage}[t]{0.45\textwidth}
    \centering
    \vspace{0cm}
    \includegraphics[width=\linewidth]{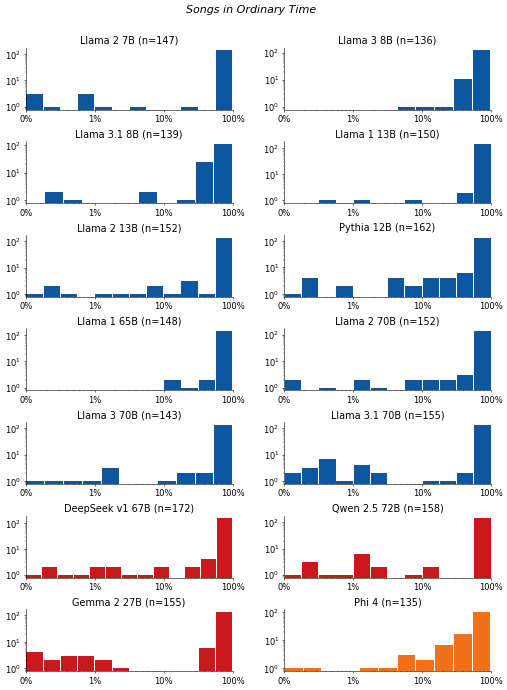}
  \end{minipage}
  \vspace{-.2cm}
  \caption{
    \textbf{\textit{Songs in Ordinary Time}, \citeauthor{Songs_in_Ordinary_Time}.}
    For $14$ LLMs,
    (\textbf{left}) heatmaps for the sliding-window procedure and
    (\textbf{right}) corresponding distributions over suffix extraction probabilities
    ($\tau_\text{min}=0.1\%$).
  }
  \label{fig:slidingwindow:Songs_in_Ordinary_Time}
\end{figure}
\FloatBarrier

\subsubsection{\textit{Beloved}, \citeauthor{Beloved}}\label{app:sec:sliding:Beloved}
\vspace{-.2cm}
\begin{figure}[h]
  \centering
  \begin{minipage}[t]{0.53\textwidth}
    \centering
    \vspace{0cm}
    \includegraphics[width=\linewidth]{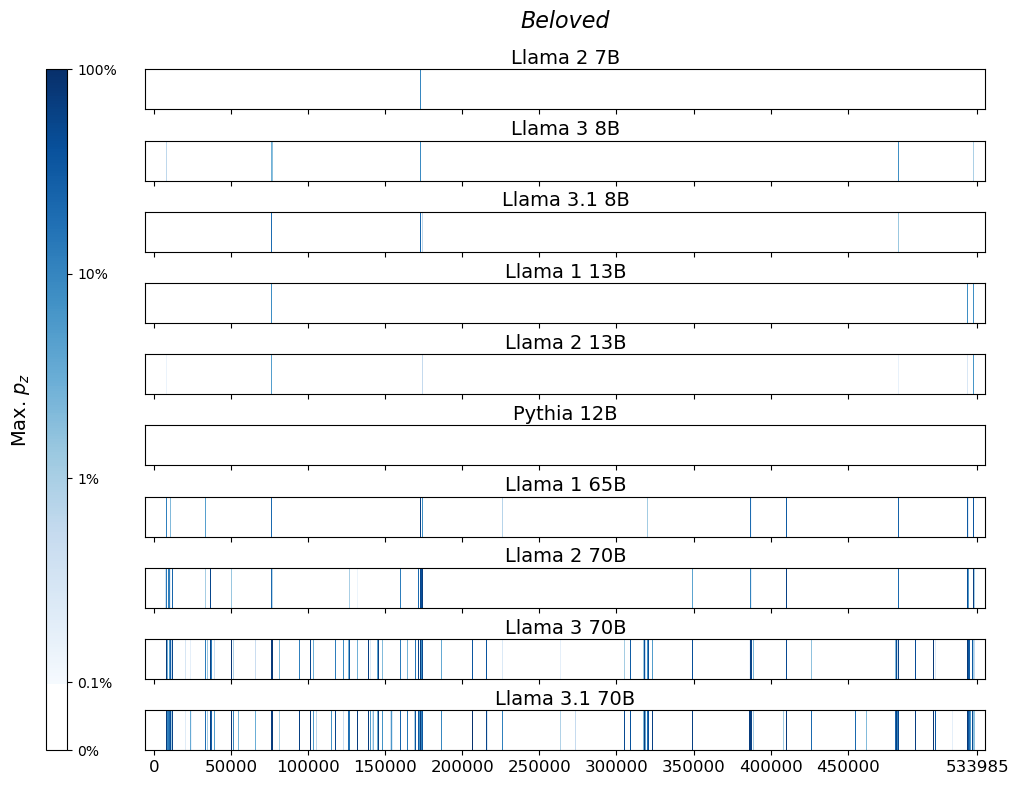}
    \includegraphics[width=\linewidth]{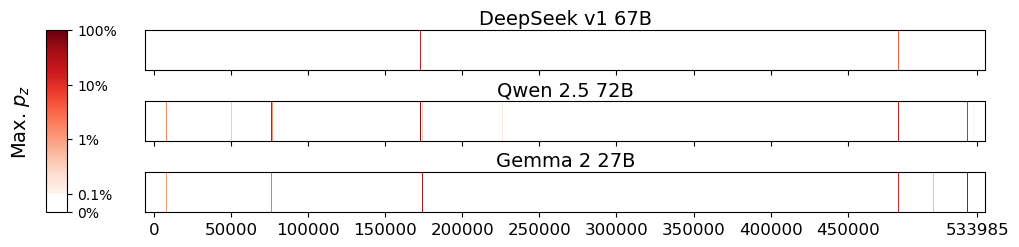}
    \includegraphics[width=\linewidth]{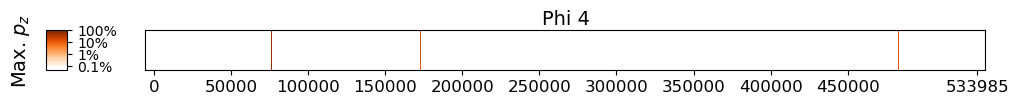}
  \end{minipage}
  \hfill
  \begin{minipage}[t]{0.45\textwidth}
    \centering
    \vspace{0cm}
    \includegraphics[width=\linewidth]{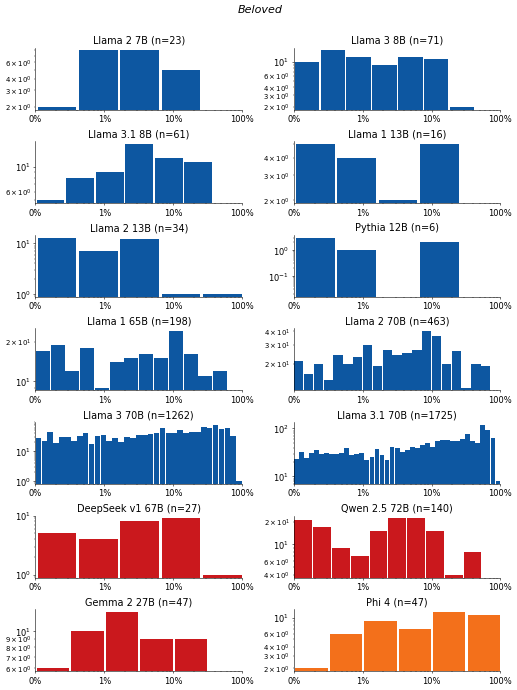}
  \end{minipage}
  \vspace{-.2cm}
  \caption{
    \textbf{\textit{Beloved}, \citeauthor{Beloved}.}
    For $14$ LLMs,
    (\textbf{left}) heatmaps for the sliding-window procedure and
    (\textbf{right}) corresponding distributions over suffix extraction probabilities
    ($\tau_\text{min}=0.1\%$).
  }
  \label{fig:slidingwindow:Beloved}
\end{figure}
\FloatBarrier

\clearpage
\subsubsection{\textit{Norwegian Wood}, \citeauthor{Norwegian_Wood}}\label{app:sec:sliding:Norwegian_Wood}
\vspace{-.2cm}
\begin{figure}[h]
  \centering
  \begin{minipage}[t]{0.53\textwidth}
    \centering
    \vspace{0cm}
    \includegraphics[width=\linewidth]{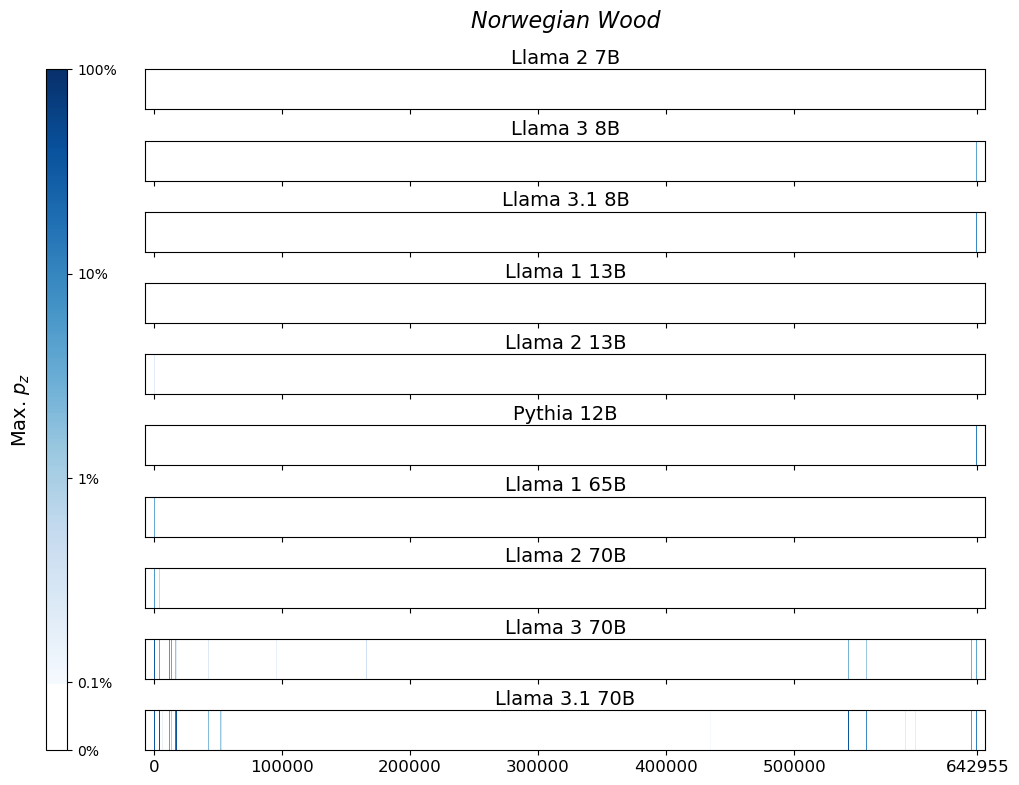}
    \includegraphics[width=\linewidth]{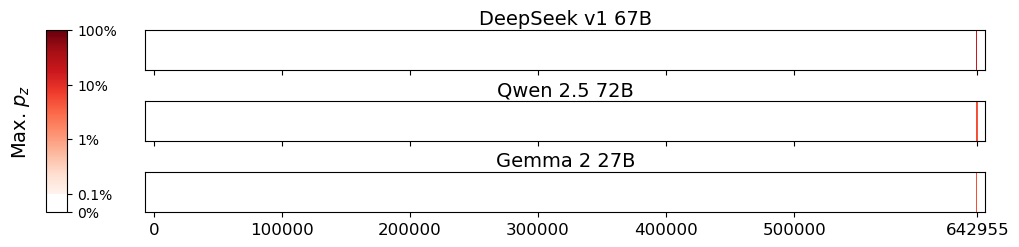}
    \includegraphics[width=\linewidth]{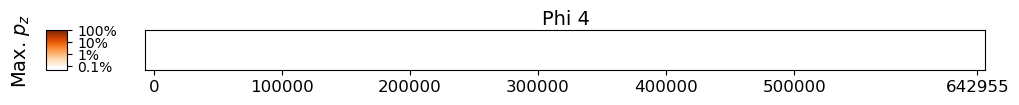}
  \end{minipage}
  \hfill
  \begin{minipage}[t]{0.45\textwidth}
    \centering
    \vspace{0cm}
    \includegraphics[width=\linewidth]{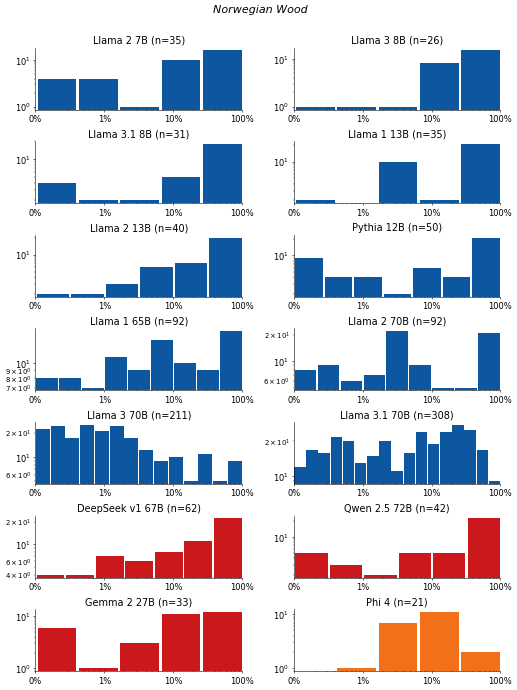}
  \end{minipage}
  \vspace{-.2cm}
  \caption{
    \textbf{\textit{Norwegian Wood}, \citeauthor{Norwegian_Wood}.}
    For $14$ LLMs,
    (\textbf{left}) heatmaps for the sliding-window procedure and
    (\textbf{right}) corresponding distributions over suffix extraction probabilities
    ($\tau_\text{min}=0.1\%$).
  }
  \label{fig:slidingwindow:Norwegian_Wood}
\end{figure}
\FloatBarrier

\subsubsection{\textit{Eat More Plants}, \citeauthor{Eat_More_Plants}}\label{app:sec:sliding:Eat_More_Plants}
\vspace{-.2cm}
\begin{figure}[h]
  \centering
  \begin{minipage}[t]{0.53\textwidth}
    \centering
    \vspace{0cm}
    \includegraphics[width=\linewidth]{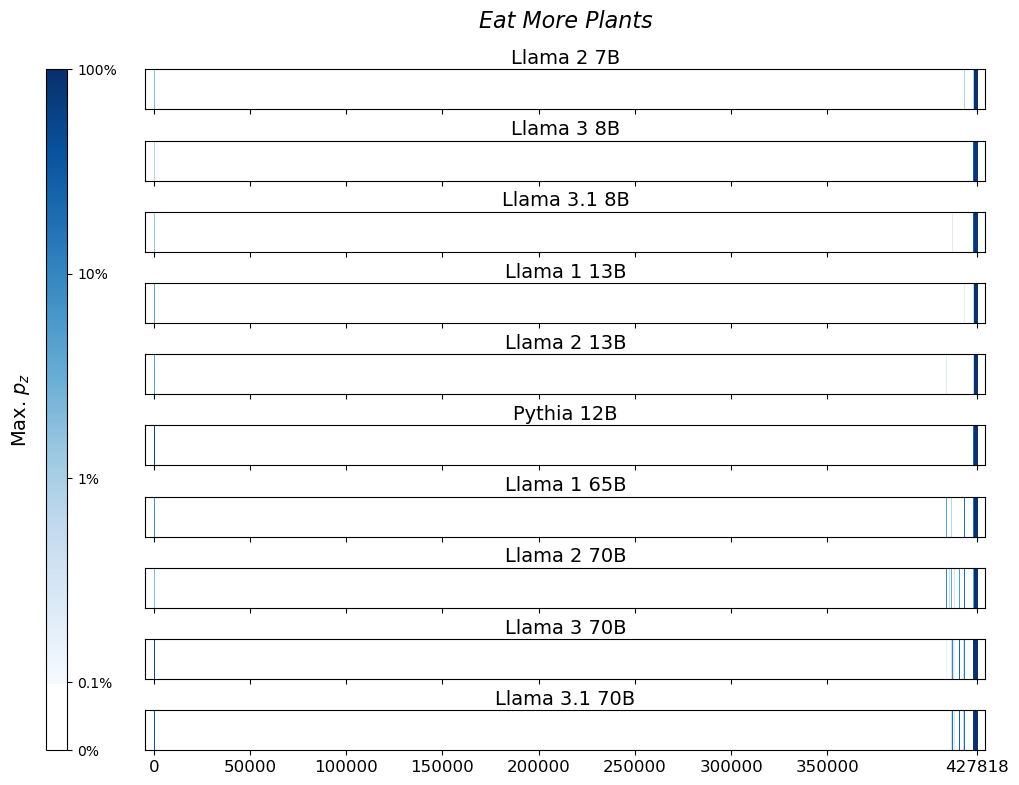}
    \includegraphics[width=\linewidth]{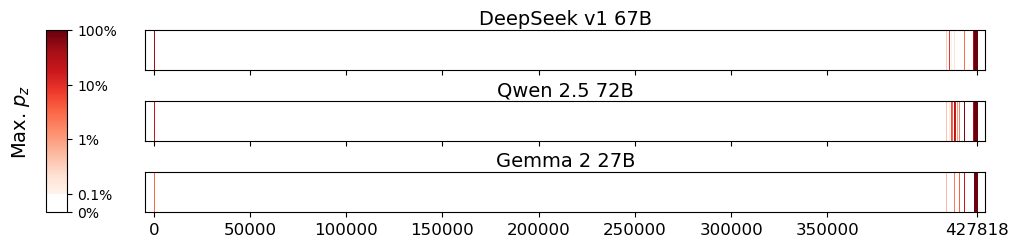}
    \includegraphics[width=\linewidth]{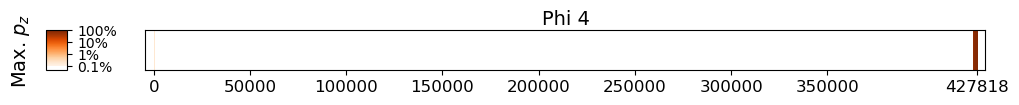}
  \end{minipage}
  \hfill
  \begin{minipage}[t]{0.45\textwidth}
    \centering
    \vspace{0cm}
    \includegraphics[width=\linewidth]{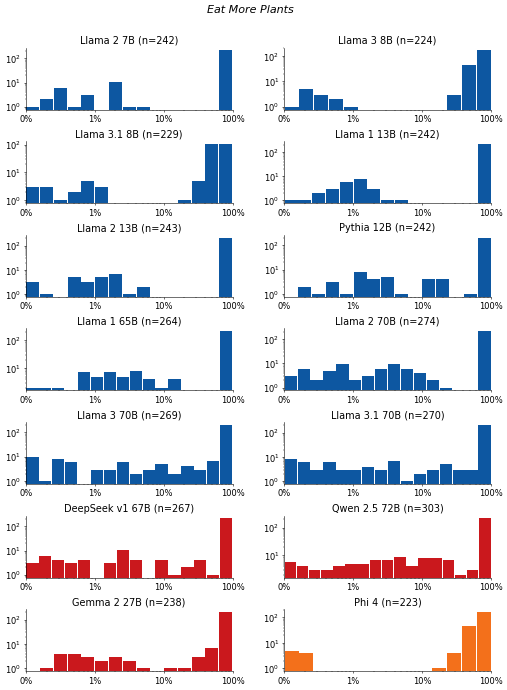}
  \end{minipage}
  \vspace{-.2cm}
  \caption{
    \textbf{\textit{Eat More Plants}, \citeauthor{Eat_More_Plants}.}
    For $14$ LLMs,
    (\textbf{left}) heatmaps for the sliding-window procedure and
    (\textbf{right}) corresponding distributions over suffix extraction probabilities
    ($\tau_\text{min}=0.1\%$).
  }
  \label{fig:slidingwindow:Eat_More_Plants}
\end{figure}
\FloatBarrier

\clearpage
\subsubsection{\textit{Polaris}, \citeauthor{Polaris}}\label{app:sec:sliding:Polaris}
\vspace{-.2cm}
\begin{figure}[h]
  \centering
  \begin{minipage}[t]{0.53\textwidth}
    \centering
    \vspace{0cm}
    \includegraphics[width=\linewidth]{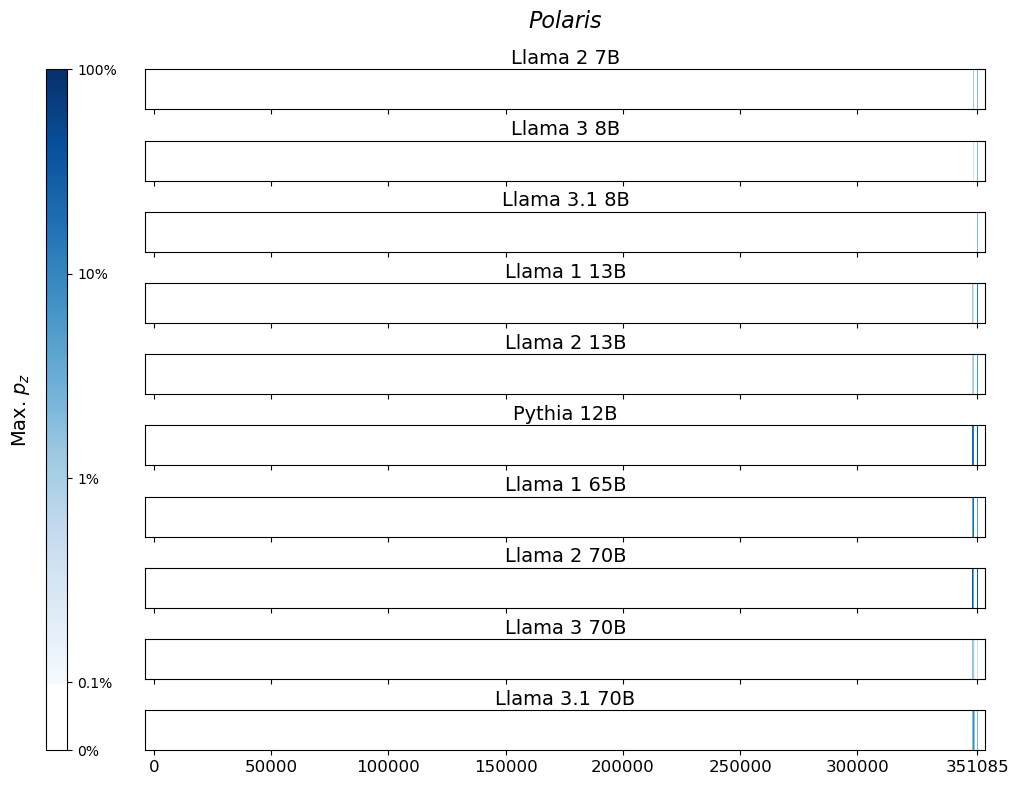}
    \includegraphics[width=\linewidth]{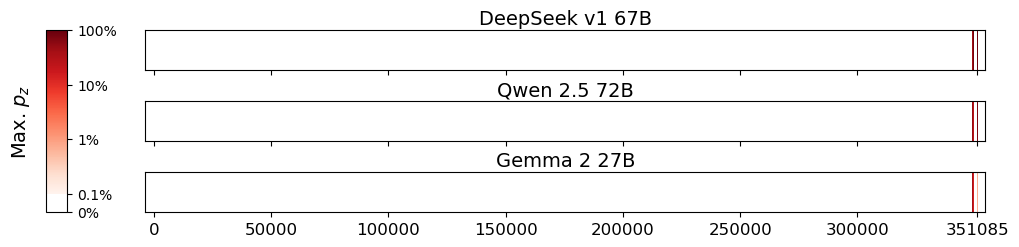}
    \includegraphics[width=\linewidth]{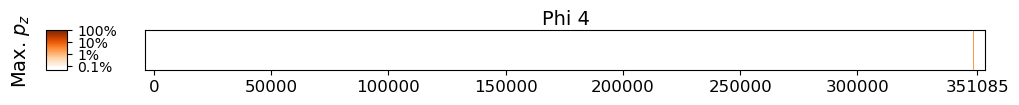}
  \end{minipage}
  \hfill
  \begin{minipage}[t]{0.45\textwidth}
    \centering
    \vspace{0cm}
    \includegraphics[width=\linewidth]{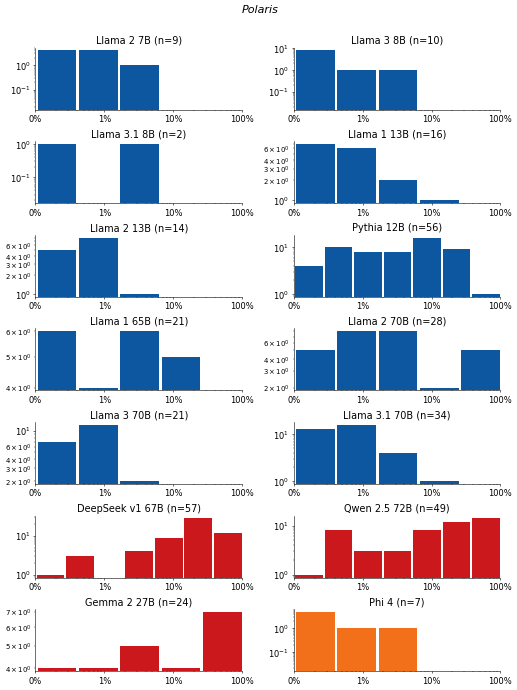}
  \end{minipage}
  \vspace{-.2cm}
  \caption{
    \textbf{\textit{Polaris}, \citeauthor{Polaris}.}
    For $14$ LLMs,
    (\textbf{left}) heatmaps for the sliding-window procedure and
    (\textbf{right}) corresponding distributions over suffix extraction probabilities
    ($\tau_\text{min}=0.1\%$).
  }
  \label{fig:slidingwindow:Polaris}
\end{figure}
\FloatBarrier

\subsubsection{\textit{Pagans}, \citeauthor{Pagans}}\label{app:sec:sliding:Pagans}
\vspace{-.2cm}
\begin{figure}[h]
  \centering
  \begin{minipage}[t]{0.53\textwidth}
    \centering
    \vspace{0cm}
    \includegraphics[width=\linewidth]{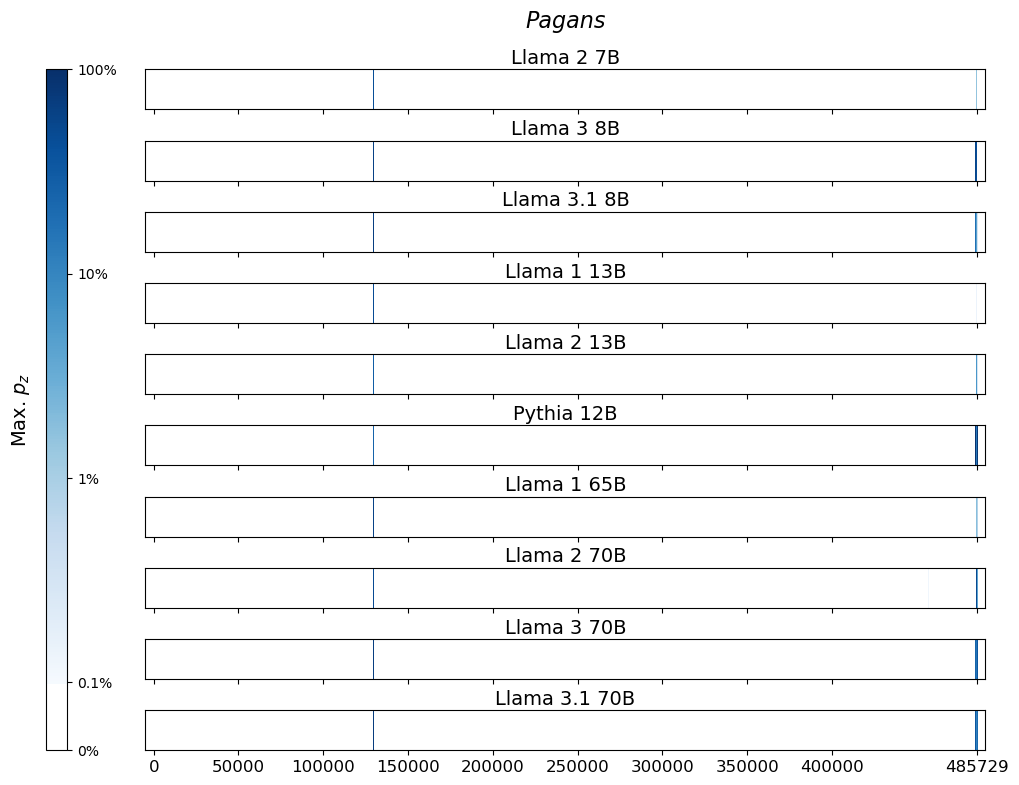}
    \includegraphics[width=\linewidth]{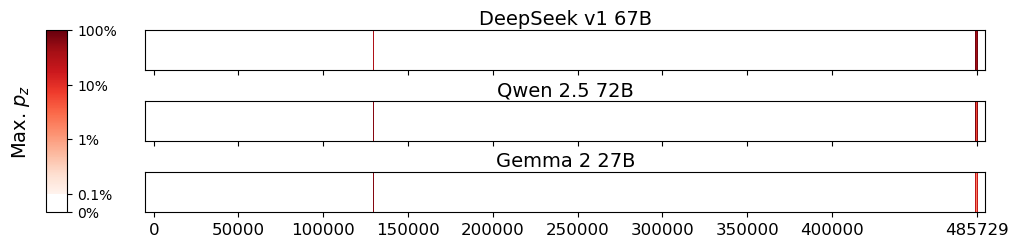}
    \includegraphics[width=\linewidth]{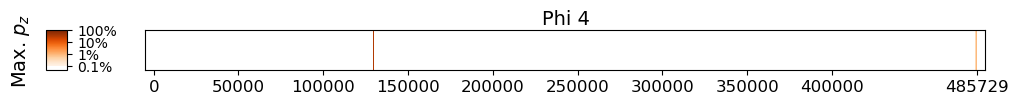}
  \end{minipage}
  \hfill
  \begin{minipage}[t]{0.45\textwidth}
    \centering
    \vspace{0cm}
    \includegraphics[width=\linewidth]{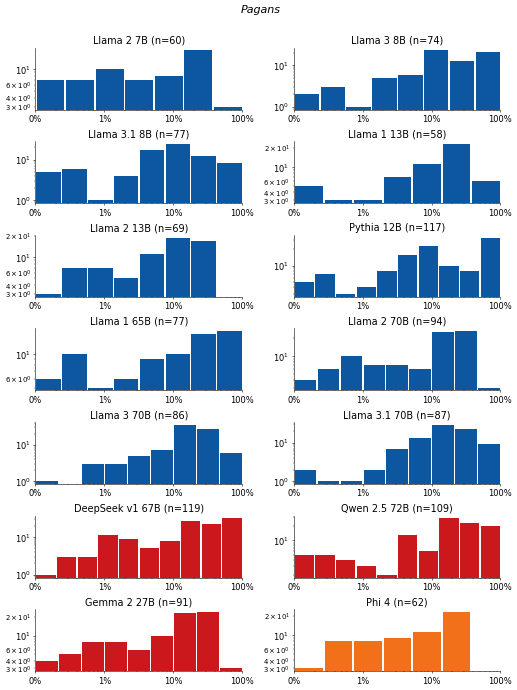}
  \end{minipage}
  \vspace{-.2cm}
  \caption{
    \textbf{\textit{Pagans}, \citeauthor{Pagans}.}
    For $14$ LLMs,
    (\textbf{left}) heatmaps for the sliding-window procedure and
    (\textbf{right}) corresponding distributions over suffix extraction probabilities
    ($\tau_\text{min}=0.1\%$).
  }
  \label{fig:slidingwindow:Pagans}
\end{figure}
\FloatBarrier

\clearpage
\subsubsection{\textit{Windfall}, \citeauthor{Windfall}}\label{app:sec:sliding:Windfall}
\vspace{-.2cm}
\begin{figure}[h]
  \centering
  \begin{minipage}[t]{0.53\textwidth}
    \centering
    \vspace{0cm}
    \includegraphics[width=\linewidth]{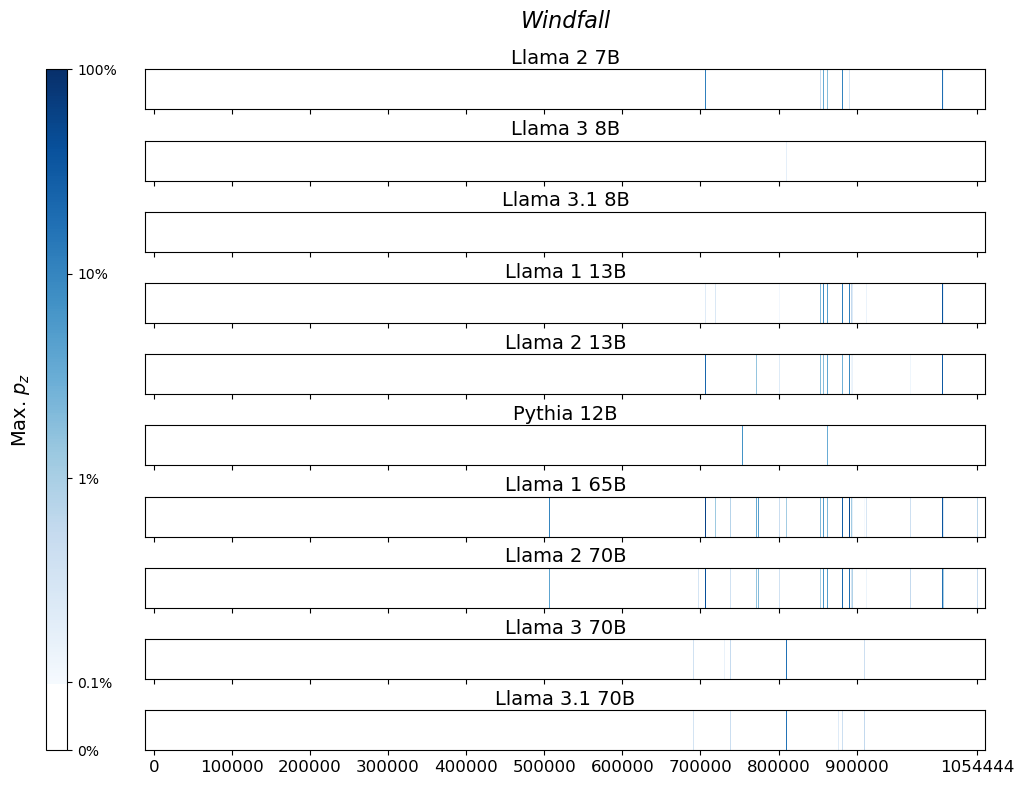}
    \includegraphics[width=\linewidth]{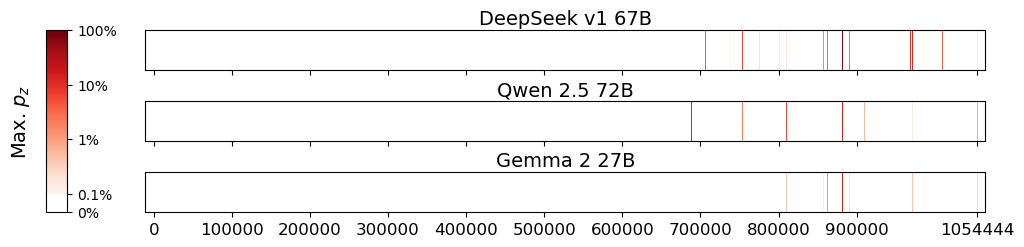}
    \includegraphics[width=\linewidth]{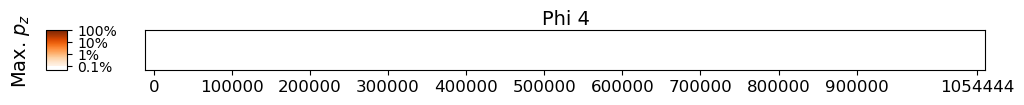}
  \end{minipage}
  \hfill
  \begin{minipage}[t]{0.45\textwidth}
    \centering
    \vspace{0cm}
    \includegraphics[width=\linewidth]{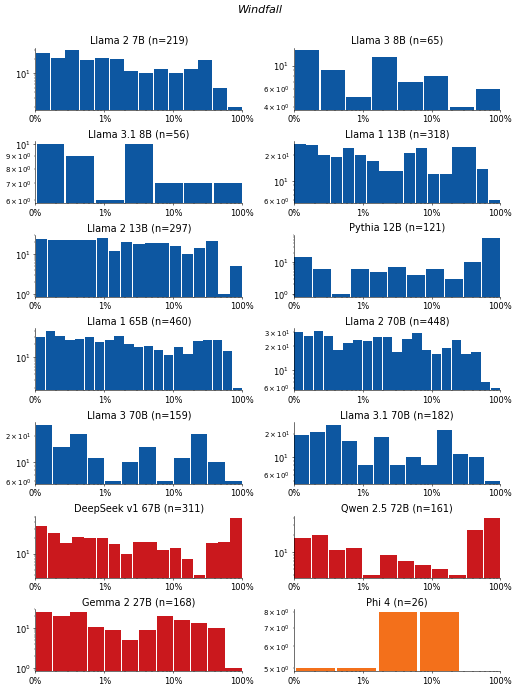}
  \end{minipage}
  \vspace{-.2cm}
  \caption{
    \textbf{\textit{Windfall}, \citeauthor{Windfall}.}
    For $14$ LLMs,
    (\textbf{left}) heatmaps for the sliding-window procedure and
    (\textbf{right}) corresponding distributions over suffix extraction probabilities
    ($\tau_\text{min}=0.1\%$).
  }
  \label{fig:slidingwindow:Windfall}
\end{figure}
\FloatBarrier

\subsubsection{\textit{The Memory Police}, \citeauthor{The_Memory_Police}}\label{app:sec:sliding:The_Memory_Police}
\vspace{-.2cm}
\begin{figure}[h]
  \centering
  \begin{minipage}[t]{0.53\textwidth}
    \centering
    \vspace{0cm}
    \includegraphics[width=\linewidth]{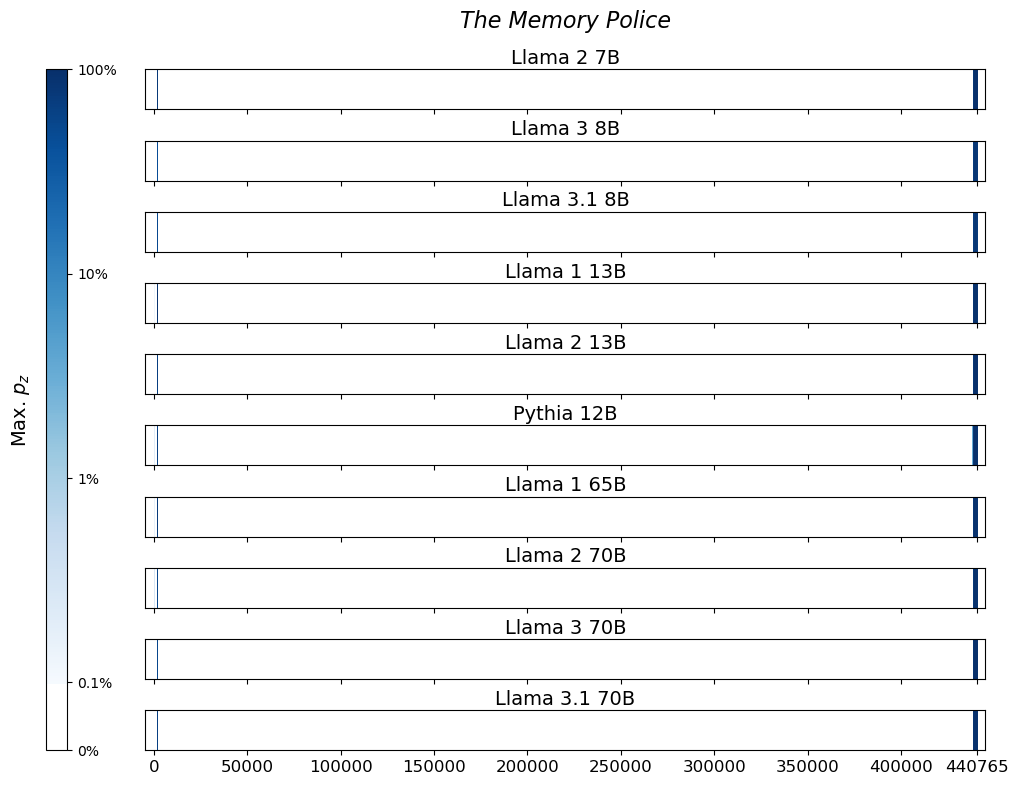}
    \includegraphics[width=\linewidth]{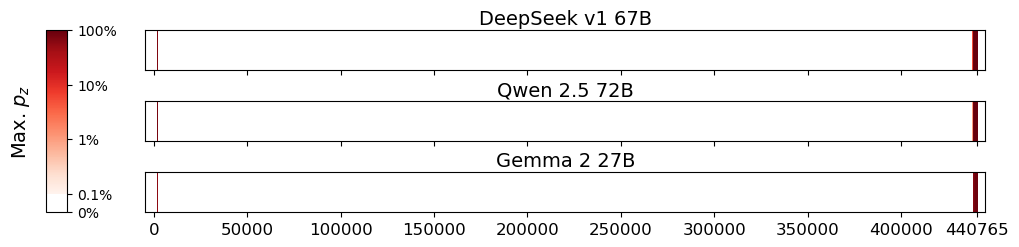}
    \includegraphics[width=\linewidth]{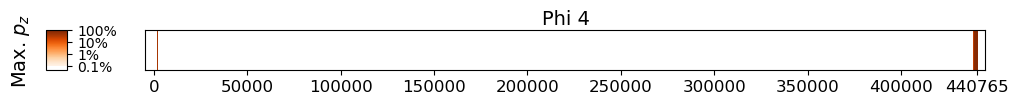}
  \end{minipage}
  \hfill
  \begin{minipage}[t]{0.45\textwidth}
    \centering
    \vspace{0cm}
    \includegraphics[width=\linewidth]{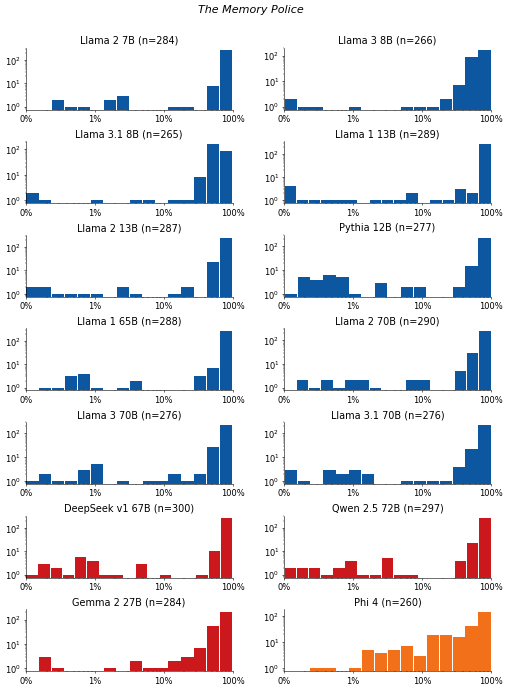}
  \end{minipage}
  \vspace{-.2cm}
  \caption{
    \textbf{\textit{The Memory Police}, \citeauthor{The_Memory_Police}.}
    For $14$ LLMs,
    (\textbf{left}) heatmaps for the sliding-window procedure and
    (\textbf{right}) corresponding distributions over suffix extraction probabilities
    ($\tau_\text{min}=0.1\%$).
  }
  \label{fig:slidingwindow:The_Memory_Police}
\end{figure}
\FloatBarrier

\clearpage
\subsubsection{\textit{Winter Sisters}, \citeauthor{Winter_Sisters}}\label{app:sec:sliding:Winter_Sisters}
\vspace{-.2cm}
\begin{figure}[h]
  \centering
  \begin{minipage}[t]{0.53\textwidth}
    \centering
    \vspace{0cm}
    \includegraphics[width=\linewidth]{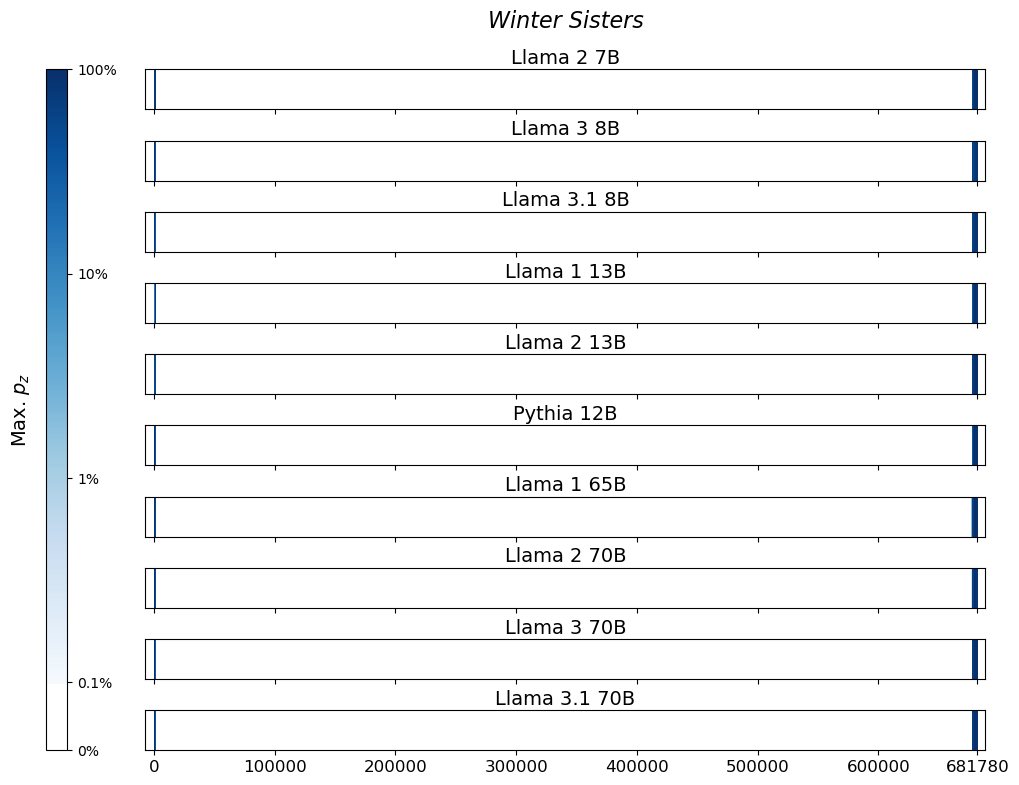}
    \includegraphics[width=\linewidth]{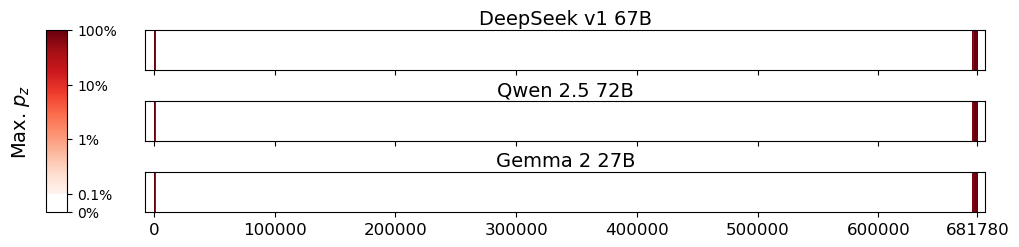}
    \includegraphics[width=\linewidth]{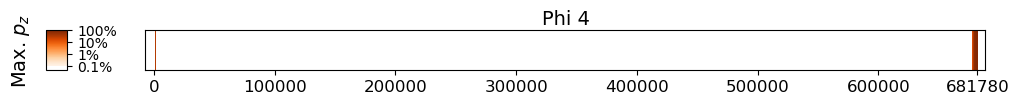}
  \end{minipage}
  \hfill
  \begin{minipage}[t]{0.45\textwidth}
    \centering
    \vspace{0cm}
    \includegraphics[width=\linewidth]{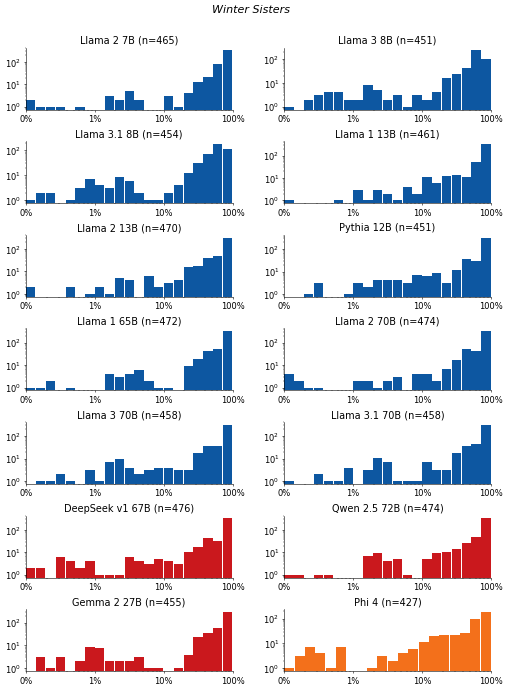}
  \end{minipage}
  \vspace{-.2cm}
  \caption{
    \textbf{\textit{Winter Sisters}, \citeauthor{Winter_Sisters}.}
    For $14$ LLMs,
    (\textbf{left}) heatmaps for the sliding-window procedure and
    (\textbf{right}) corresponding distributions over suffix extraction probabilities
    ($\tau_\text{min}=0.1\%$).
  }
  \label{fig:slidingwindow:Winter_Sisters}
\end{figure}
\FloatBarrier

\subsubsection{\textit{Nineteen Eighty-Four}, \citeauthor{Nineteen_Eighty-Four}}\label{app:sec:sliding:Nineteen_Eighty-Four}
\vspace{-.2cm}
\begin{figure}[h]
  \centering
  \begin{minipage}[t]{0.53\textwidth}
    \centering
    \vspace{0cm}
    \includegraphics[width=\linewidth]{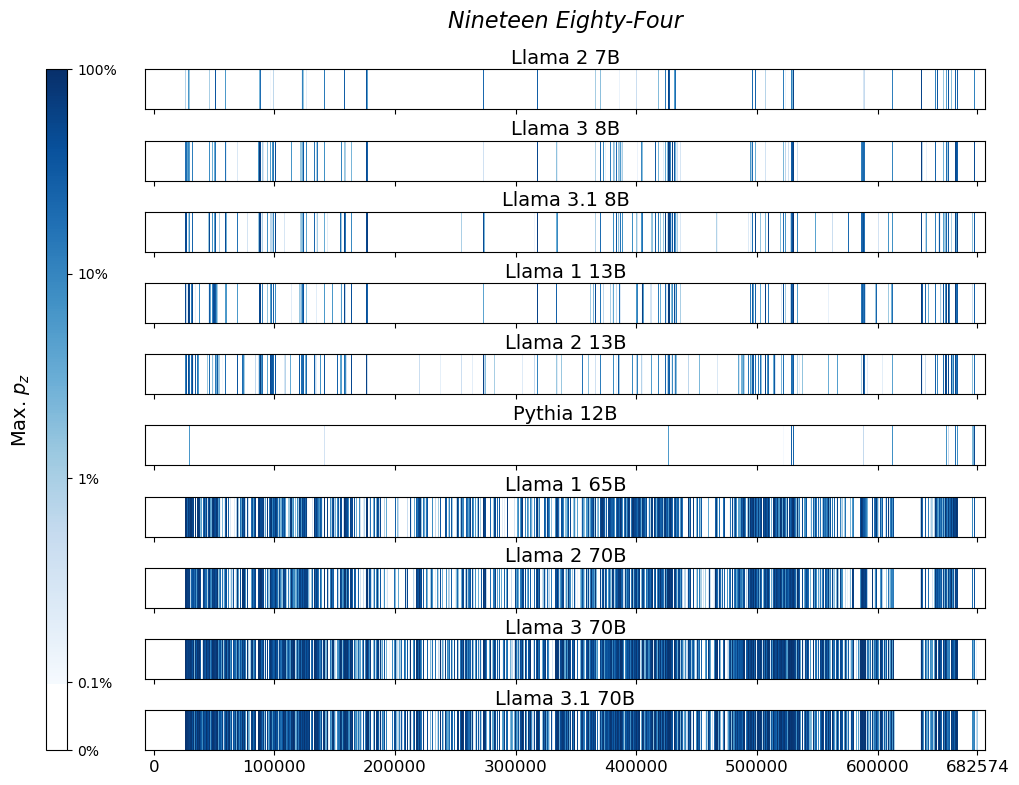}
    \includegraphics[width=\linewidth]{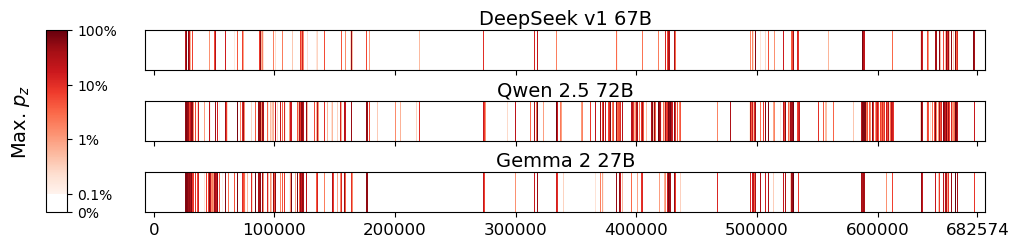}
    \includegraphics[width=\linewidth]{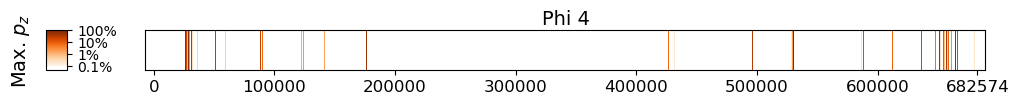}
  \end{minipage}
  \hfill
  \begin{minipage}[t]{0.45\textwidth}
    \centering
    \vspace{0cm}
    \includegraphics[width=\linewidth]{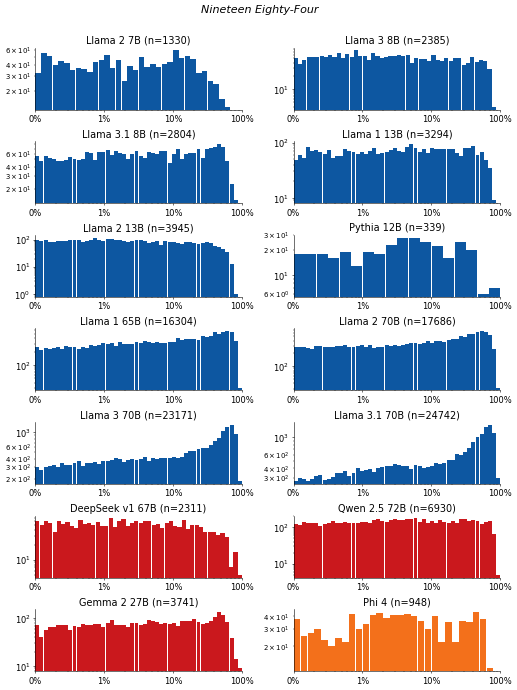}
  \end{minipage}
  \vspace{-.2cm}
  \caption{
    \textbf{\textit{Nineteen Eighty-Four}, \citeauthor{Nineteen_Eighty-Four}.}
    For $14$ LLMs,
    (\textbf{left}) heatmaps for the sliding-window procedure and
    (\textbf{right}) corresponding distributions over suffix extraction probabilities
    ($\tau_\text{min}=0.1\%$).
  }
  \label{fig:slidingwindow:Nineteen_Eighty-Four}
\end{figure}
\FloatBarrier

\clearpage
\subsubsection{\textit{Fight Club}, \citeauthor{Fight_Club}}\label{app:sec:sliding:Fight_Club}
\vspace{-.2cm}
\begin{figure}[h]
  \centering
  \begin{minipage}[t]{0.53\textwidth}
    \centering
    \vspace{0cm}
    \includegraphics[width=\linewidth]{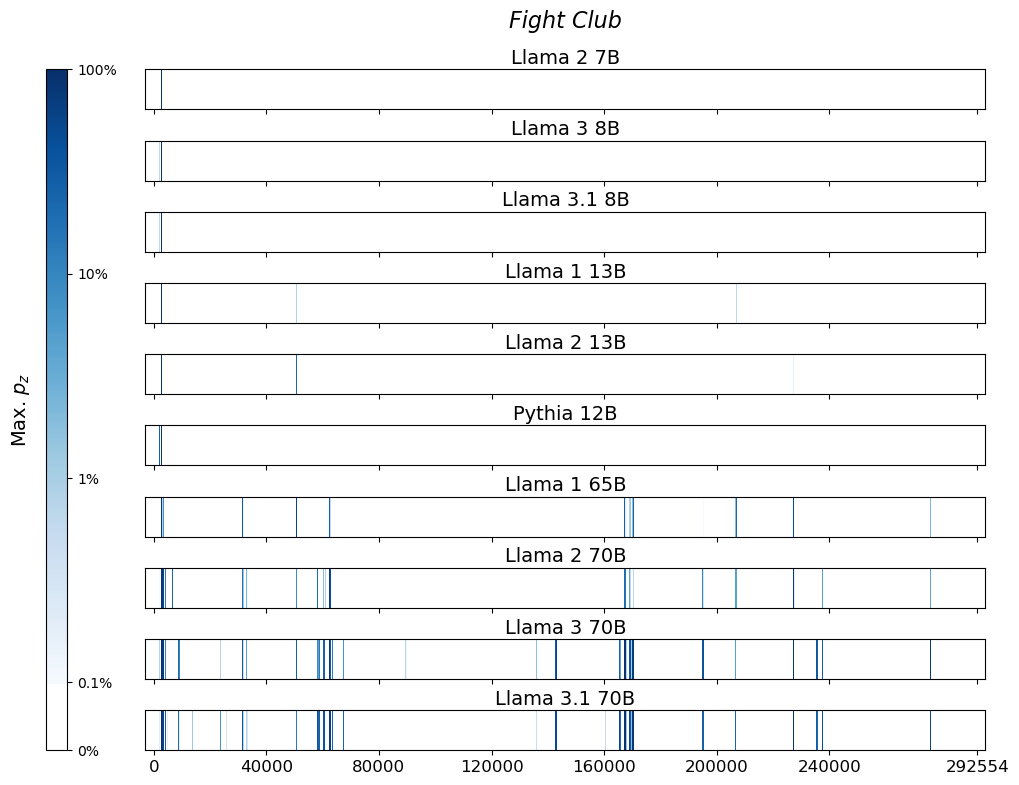}
    \includegraphics[width=\linewidth]{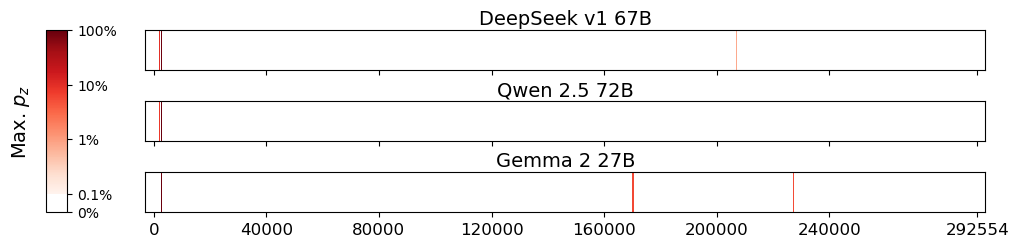}
    \includegraphics[width=\linewidth]{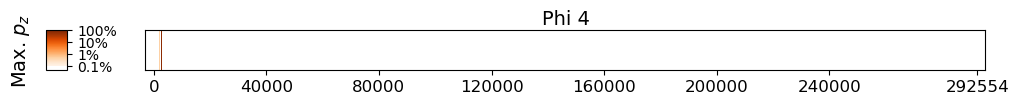}
  \end{minipage}
  \hfill
  \begin{minipage}[t]{0.45\textwidth}
    \centering
    \vspace{0cm}
    \includegraphics[width=\linewidth]{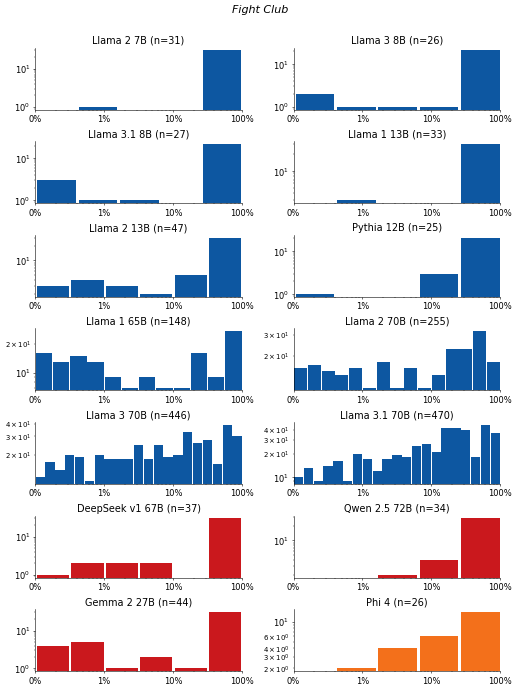}
  \end{minipage}
  \vspace{-.2cm}
  \caption{
    \textbf{\textit{Fight Club}, \citeauthor{Fight_Club}.}
    For $14$ LLMs,
    (\textbf{left}) heatmaps for the sliding-window procedure and
    (\textbf{right}) corresponding distributions over suffix extraction probabilities
    ($\tau_\text{min}=0.1\%$).
  }
  \label{fig:slidingwindow:Fight_Club}
\end{figure}
\FloatBarrier

\subsubsection{\textit{The Complete Joy of Homebrewing}, \citeauthor{The_Complete_Joy_of_Homebrewing}}\label{app:sec:sliding:The_Complete_Joy_of_Homebrewing}
\vspace{-.2cm}
\begin{figure}[h]
  \centering
  \begin{minipage}[t]{0.53\textwidth}
    \centering
    \vspace{0cm}
    \includegraphics[width=\linewidth]{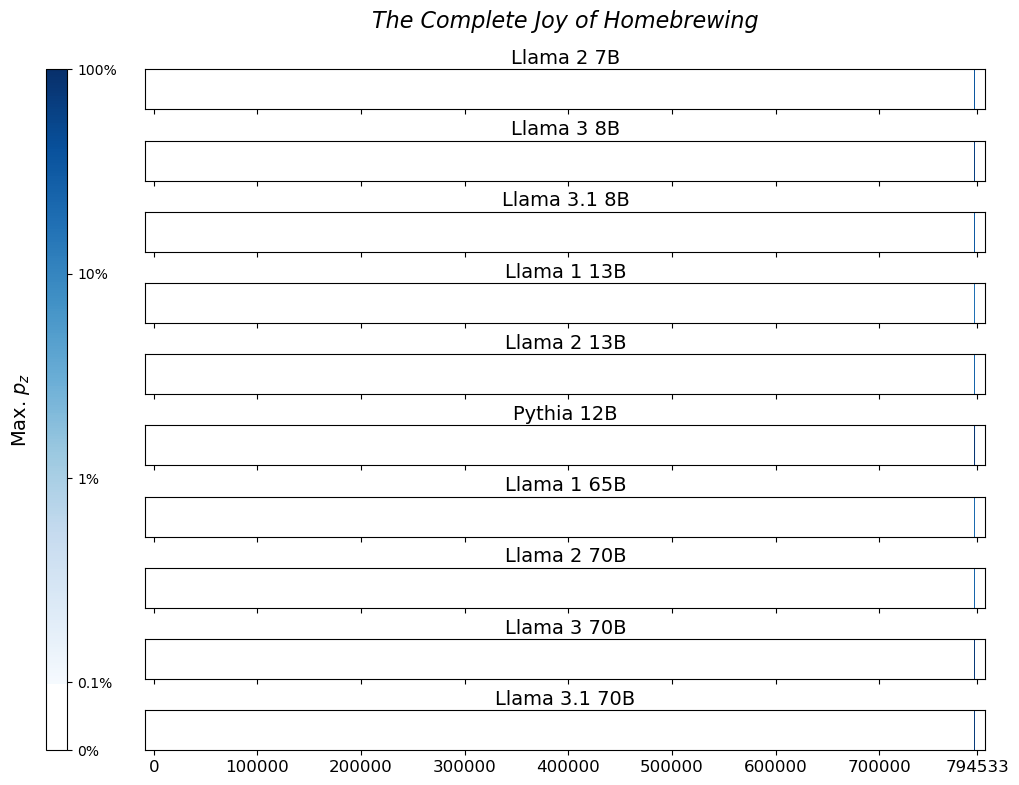}
    \includegraphics[width=\linewidth]{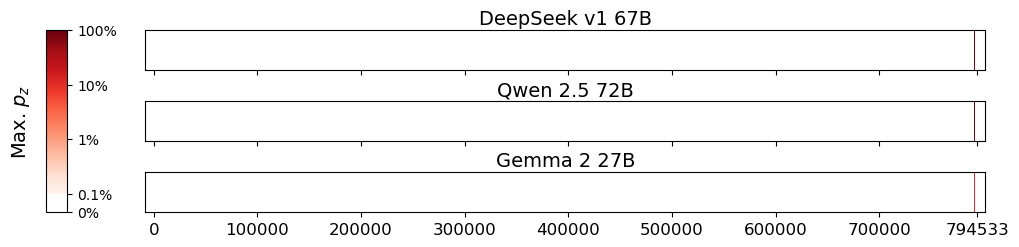}
    \includegraphics[width=\linewidth]{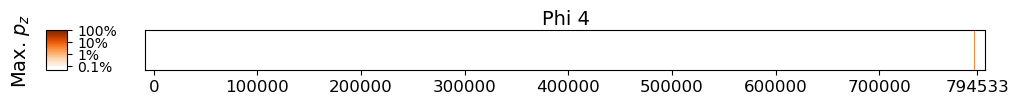}
  \end{minipage}
  \hfill
  \begin{minipage}[t]{0.45\textwidth}
    \centering
    \vspace{0cm}
    \includegraphics[width=\linewidth]{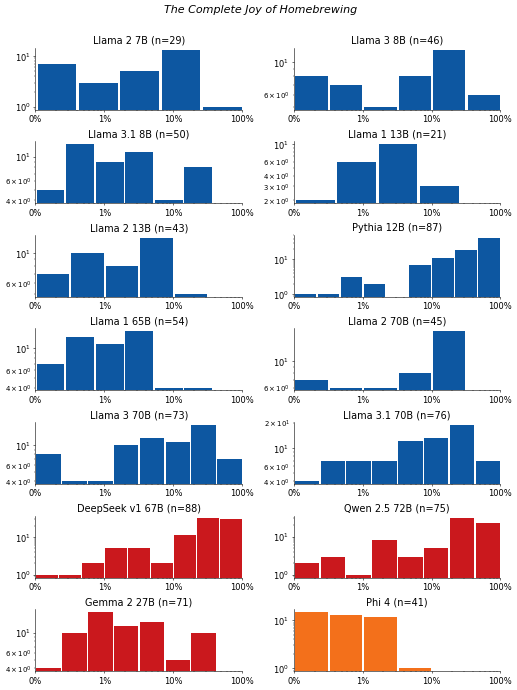}
  \end{minipage}
  \vspace{-.2cm}
  \caption{
    \textbf{\textit{The Complete Joy of Homebrewing}, \citeauthor{The_Complete_Joy_of_Homebrewing}.}
    For $14$ LLMs,
    (\textbf{left}) heatmaps for the sliding-window procedure and
    (\textbf{right}) corresponding distributions over suffix extraction probabilities
    ($\tau_\text{min}=0.1\%$).
  }
  \label{fig:slidingwindow:The_Complete_Joy_of_Homebrewing}
\end{figure}
\FloatBarrier

\clearpage
\subsubsection{\textit{The Cult of Loving Kindness}, \citeauthor{The_Cult_of_Loving_Kindness}}\label{app:sec:sliding:The_Cult_of_Loving_Kindness}
\vspace{-.2cm}
\begin{figure}[h]
  \centering
  \begin{minipage}[t]{0.53\textwidth}
    \centering
    \vspace{0cm}
    \includegraphics[width=\linewidth]{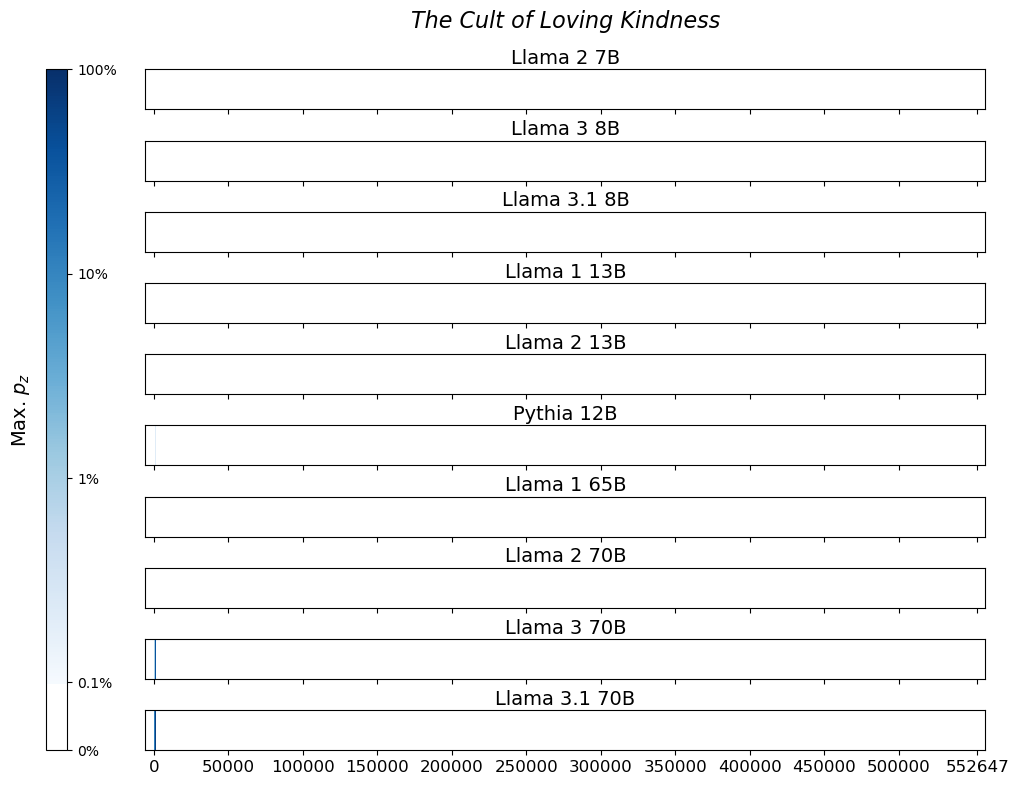}
    \includegraphics[width=\linewidth]{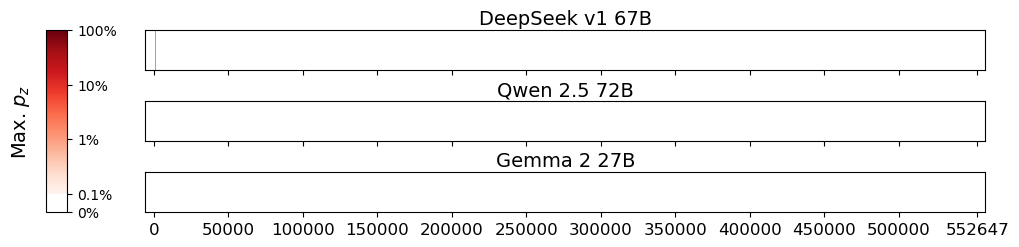}
    \includegraphics[width=\linewidth]{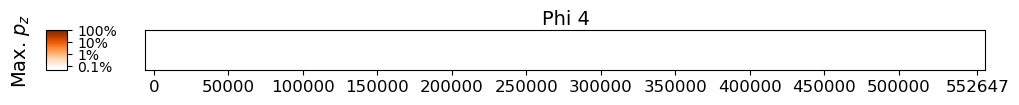}
  \end{minipage}
  \hfill
  \begin{minipage}[t]{0.45\textwidth}
    \centering
    \vspace{0cm}
    \includegraphics[width=\linewidth]{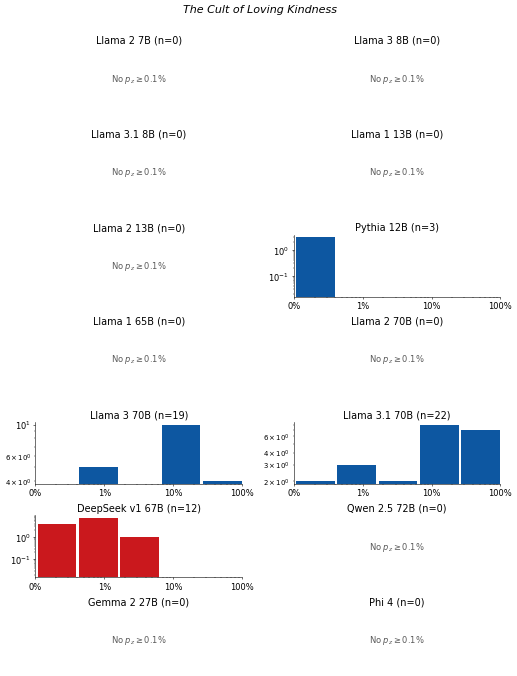}
  \end{minipage}
  \vspace{-.2cm}
  \caption{
    \textbf{\textit{The Cult of Loving Kindness}, \citeauthor{The_Cult_of_Loving_Kindness}.}
    For $14$ LLMs,
    (\textbf{left}) heatmaps for the sliding-window procedure and
    (\textbf{right}) corresponding distributions over suffix extraction probabilities
    ($\tau_\text{min}=0.1\%$).
  }
  \label{fig:slidingwindow:The_Cult_of_Loving_Kindness}
\end{figure}
\FloatBarrier

\subsubsection{\textit{Along Came a Spider}, \citeauthor{Along_Came_a_Spider}}\label{app:sec:sliding:Along_Came_a_Spider}
\vspace{-.2cm}
\begin{figure}[h]
  \centering
  \begin{minipage}[t]{0.53\textwidth}
    \centering
    \vspace{0cm}
    \includegraphics[width=\linewidth]{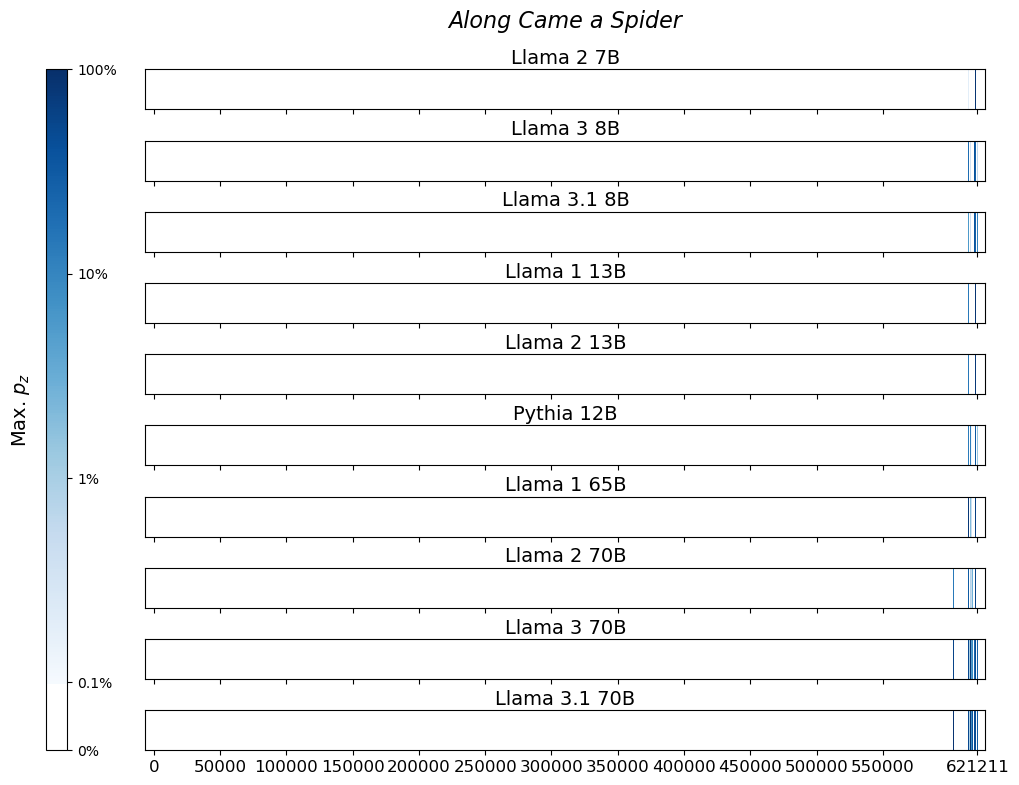}
    \includegraphics[width=\linewidth]{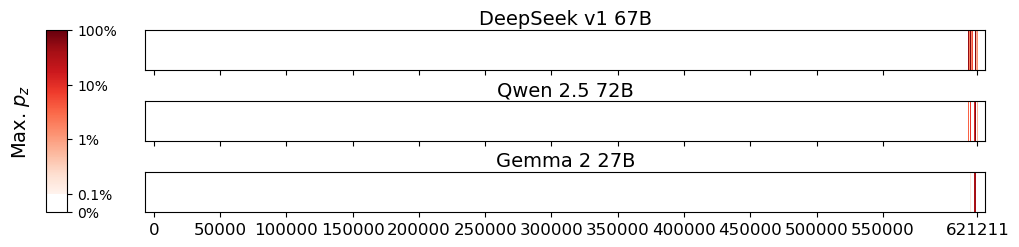}
    \includegraphics[width=\linewidth]{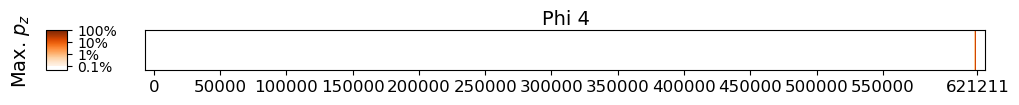}
  \end{minipage}
  \hfill
  \begin{minipage}[t]{0.45\textwidth}
    \centering
    \vspace{0cm}
    \includegraphics[width=\linewidth]{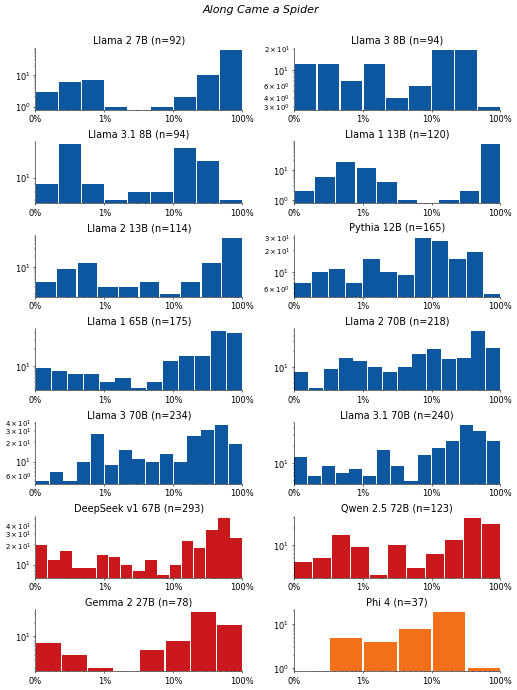}
  \end{minipage}
  \vspace{-.2cm}
  \caption{
    \textbf{\textit{Along Came a Spider}, \citeauthor{Along_Came_a_Spider}.}
    For $14$ LLMs,
    (\textbf{left}) heatmaps for the sliding-window procedure and
    (\textbf{right}) corresponding distributions over suffix extraction probabilities
    ($\tau_\text{min}=0.1\%$).
  }
  \label{fig:slidingwindow:Along_Came_a_Spider}
\end{figure}
\FloatBarrier

\clearpage
\subsubsection{\textit{Payard Cookies}, \citeauthor{Payard_Cookies}}\label{app:sec:sliding:Payard_Cookies}
\vspace{-.2cm}
\begin{figure}[h]
  \centering
  \begin{minipage}[t]{0.53\textwidth}
    \centering
    \vspace{0cm}
    \includegraphics[width=\linewidth]{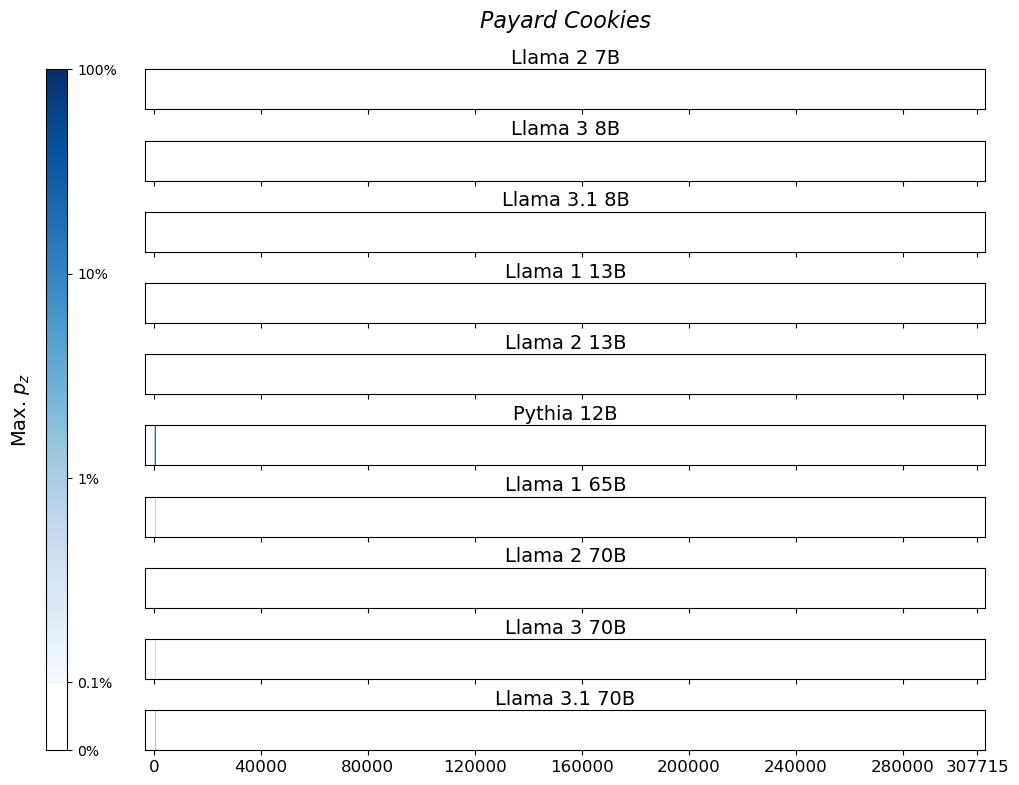}
    \includegraphics[width=\linewidth]{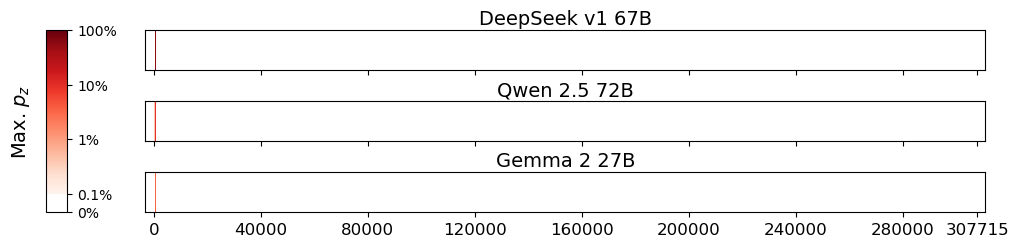}
    \includegraphics[width=\linewidth]{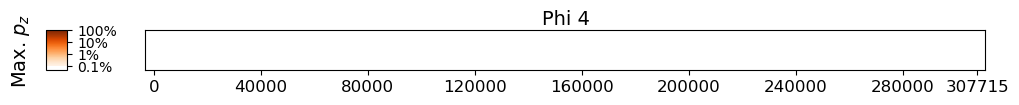}
  \end{minipage}
  \hfill
  \begin{minipage}[t]{0.45\textwidth}
    \centering
    \vspace{0cm}
    \includegraphics[width=\linewidth]{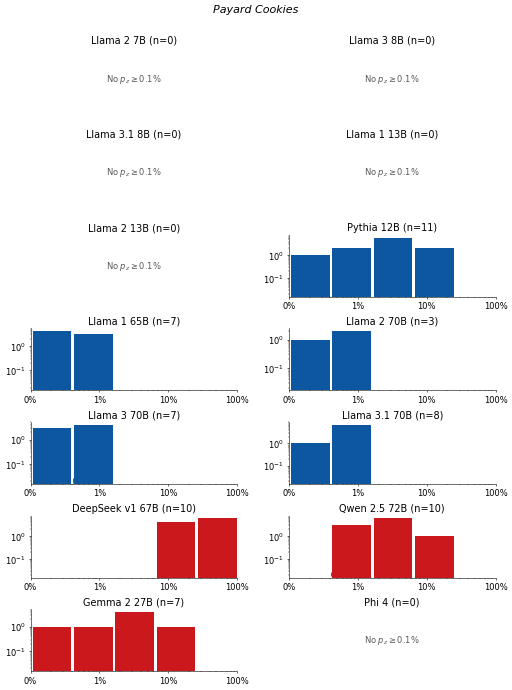}
  \end{minipage}
  \vspace{-.2cm}
  \caption{
    \textbf{\textit{Payard Cookies}, \citeauthor{Payard_Cookies}.}
    For $14$ LLMs,
    (\textbf{left}) heatmaps for the sliding-window procedure and
    (\textbf{right}) corresponding distributions over suffix extraction probabilities
    ($\tau_\text{min}=0.1\%$).
  }
  \label{fig:slidingwindow:Payard_Cookies}
\end{figure}
\FloatBarrier

\subsubsection{\textit{Essential Pepin Desserts}, \citeauthor{Essential_Pepin_Desserts}}\label{app:sec:sliding:Essential_Pepin_Desserts}
\vspace{-.2cm}
\begin{figure}[h]
  \centering
  \begin{minipage}[t]{0.53\textwidth}
    \centering
    \vspace{0cm}
    \includegraphics[width=\linewidth]{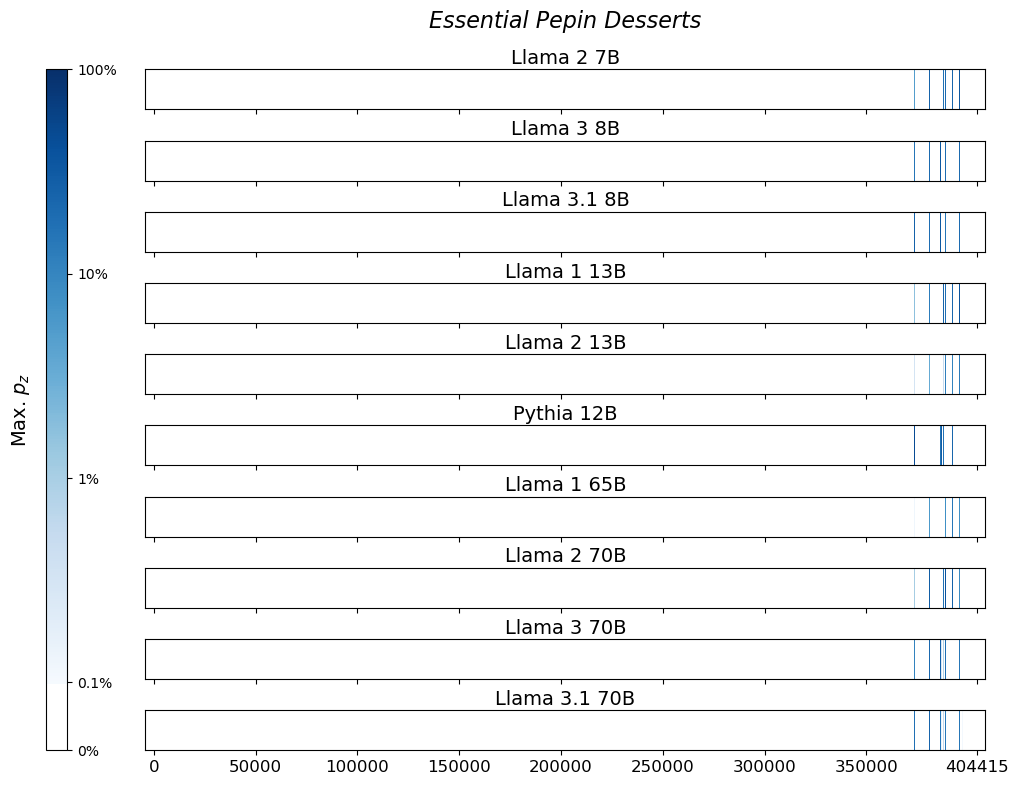}
    \includegraphics[width=\linewidth]{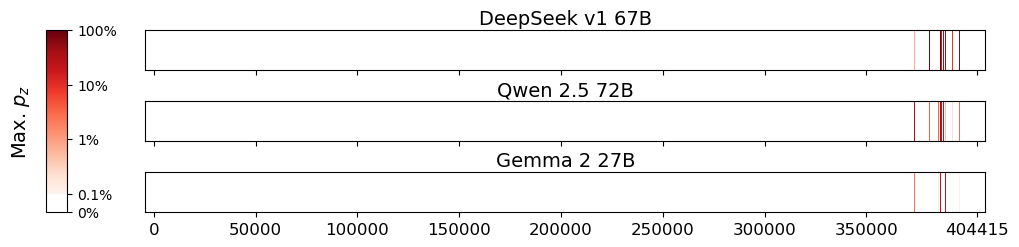}
    \includegraphics[width=\linewidth]{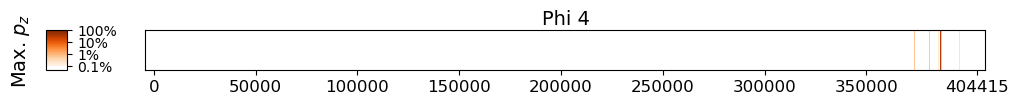}
  \end{minipage}
  \hfill
  \begin{minipage}[t]{0.45\textwidth}
    \centering
    \vspace{0cm}
    \includegraphics[width=\linewidth]{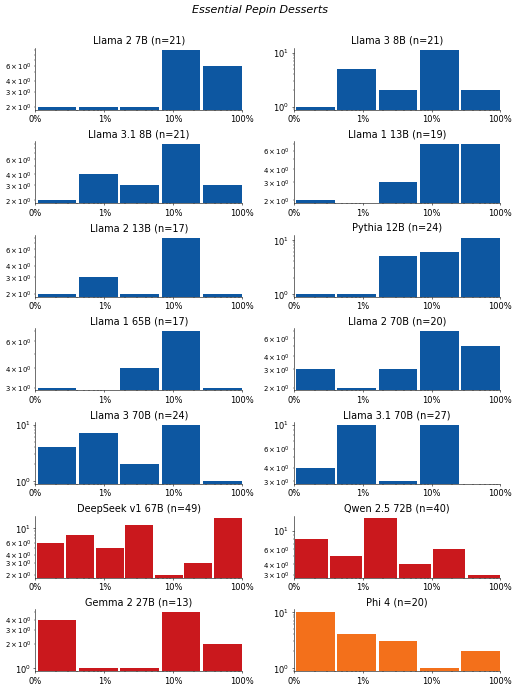}
  \end{minipage}
  \vspace{-.2cm}
  \caption{
    \textbf{\textit{Essential Pepin Desserts}, \citeauthor{Essential_Pepin_Desserts}.}
    For $14$ LLMs,
    (\textbf{left}) heatmaps for the sliding-window procedure and
    (\textbf{right}) corresponding distributions over suffix extraction probabilities
    ($\tau_\text{min}=0.1\%$).
  }
  \label{fig:slidingwindow:Essential_Pepin_Desserts}
\end{figure}
\FloatBarrier

\clearpage
\subsubsection{\textit{Why New Orleans Matters}, \citeauthor{Why_New_Orleans_Matters}}\label{app:sec:sliding:Why_New_Orleans_Matters}
\vspace{-.2cm}
\begin{figure}[h]
  \centering
  \begin{minipage}[t]{0.53\textwidth}
    \centering
    \vspace{0cm}
    \includegraphics[width=\linewidth]{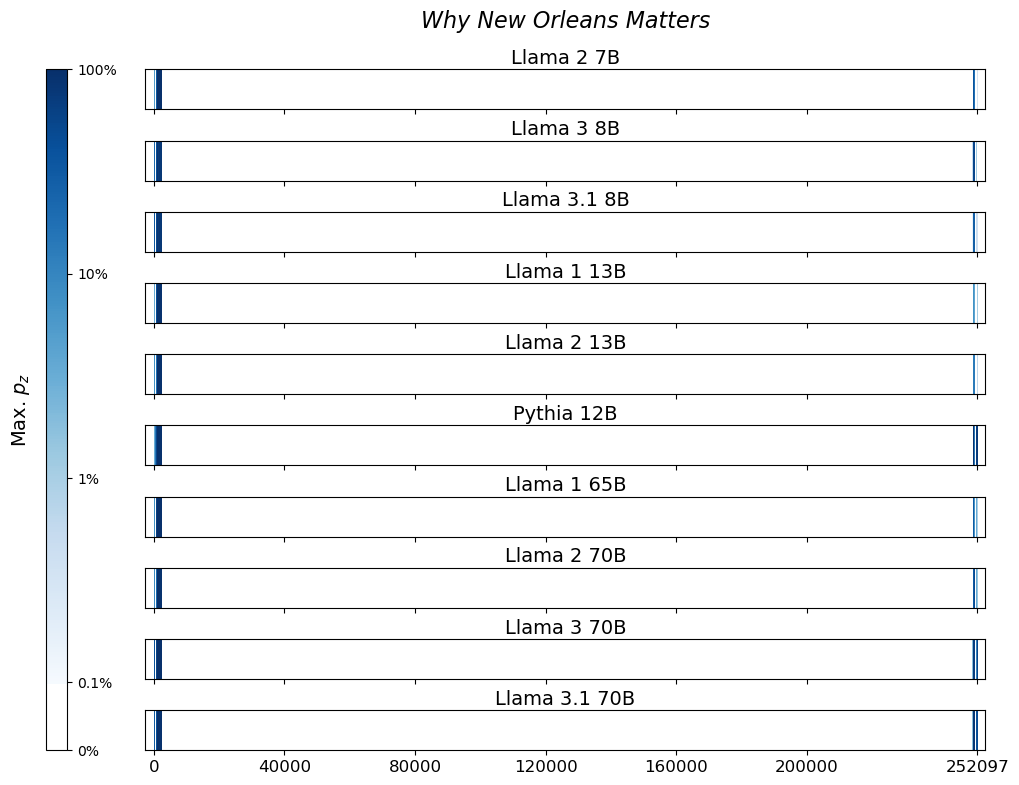}
    \includegraphics[width=\linewidth]{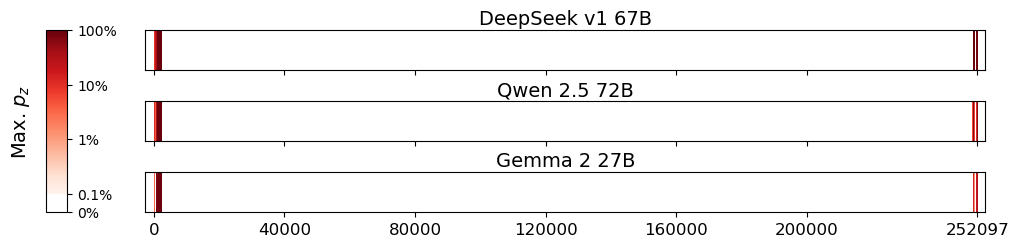}
    \includegraphics[width=\linewidth]{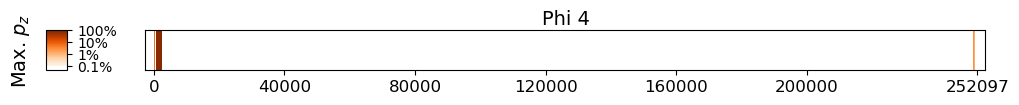}
  \end{minipage}
  \hfill
  \begin{minipage}[t]{0.45\textwidth}
    \centering
    \vspace{0cm}
    \includegraphics[width=\linewidth]{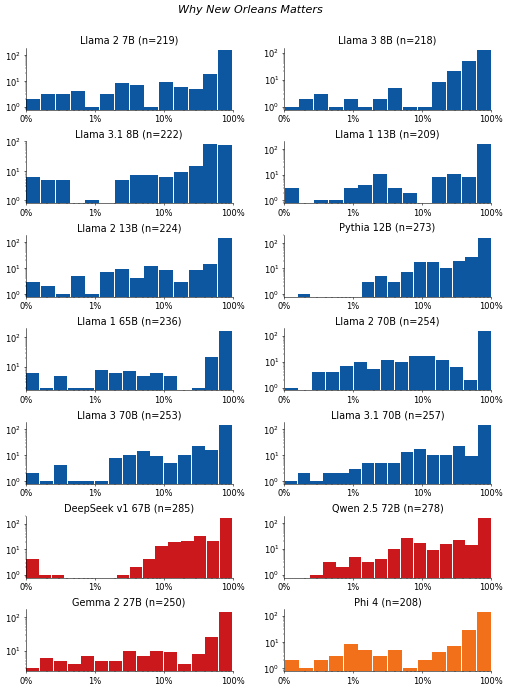}
  \end{minipage}
  \vspace{-.2cm}
  \caption{
    \textbf{\textit{Why New Orleans Matters}, \citeauthor{Why_New_Orleans_Matters}.}
    For $14$ LLMs,
    (\textbf{left}) heatmaps for the sliding-window procedure and
    (\textbf{right}) corresponding distributions over suffix extraction probabilities
    ($\tau_\text{min}=0.1\%$).
  }
  \label{fig:slidingwindow:Why_New_Orleans_Matters}
\end{figure}
\FloatBarrier

\subsubsection{\textit{Enlightenment Now}, \citeauthor{Enlightenment_Now}}\label{app:sec:sliding:Enlightenment_Now}
\vspace{-.2cm}
\begin{figure}[h]
  \centering
  \begin{minipage}[t]{0.53\textwidth}
    \centering
    \vspace{0cm}
    \includegraphics[width=\linewidth]{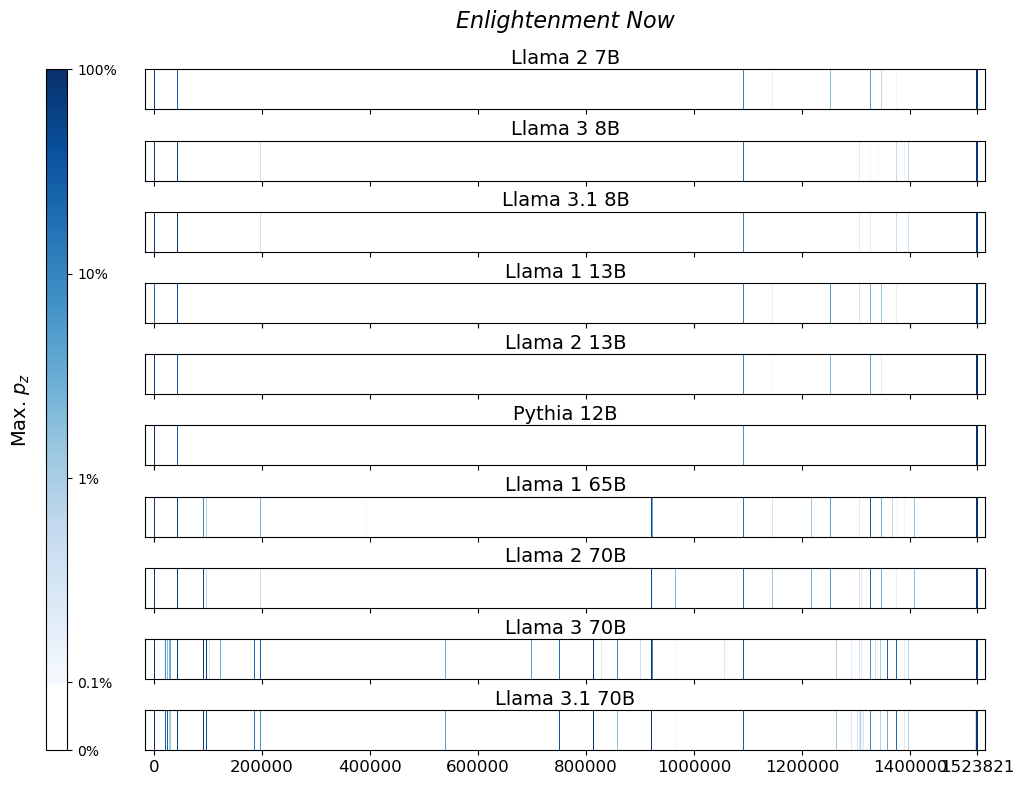}
    \includegraphics[width=\linewidth]{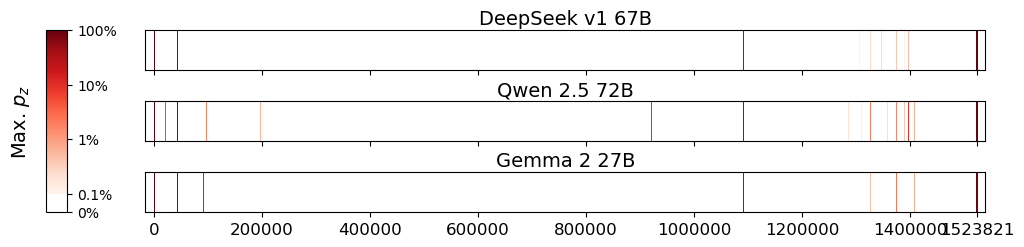}
    \includegraphics[width=\linewidth]{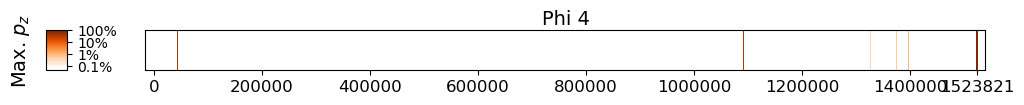}
  \end{minipage}
  \hfill
  \begin{minipage}[t]{0.45\textwidth}
    \centering
    \vspace{0cm}
    \includegraphics[width=\linewidth]{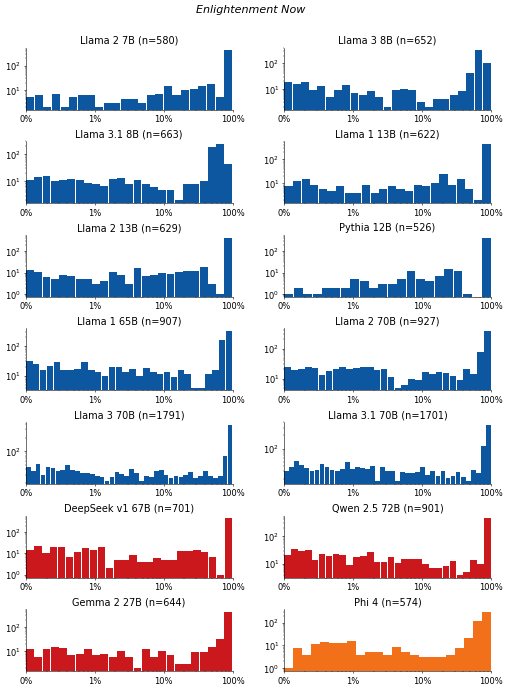}
  \end{minipage}
  \vspace{-.2cm}
  \caption{
    \textbf{\textit{Enlightenment Now}, \citeauthor{Enlightenment_Now}.}
    For $14$ LLMs,
    (\textbf{left}) heatmaps for the sliding-window procedure and
    (\textbf{right}) corresponding distributions over suffix extraction probabilities
    ($\tau_\text{min}=0.1\%$).
  }
  \label{fig:slidingwindow:Enlightenment_Now}
\end{figure}
\FloatBarrier

\clearpage
\subsubsection{\textit{Competitive Strategy}, \citeauthor{Competitive_Strategy}}\label{app:sec:sliding:Competitive_Strategy}
\vspace{-.2cm}
\begin{figure}[h]
  \centering
  \begin{minipage}[t]{0.53\textwidth}
    \centering
    \vspace{0cm}
    \includegraphics[width=\linewidth]{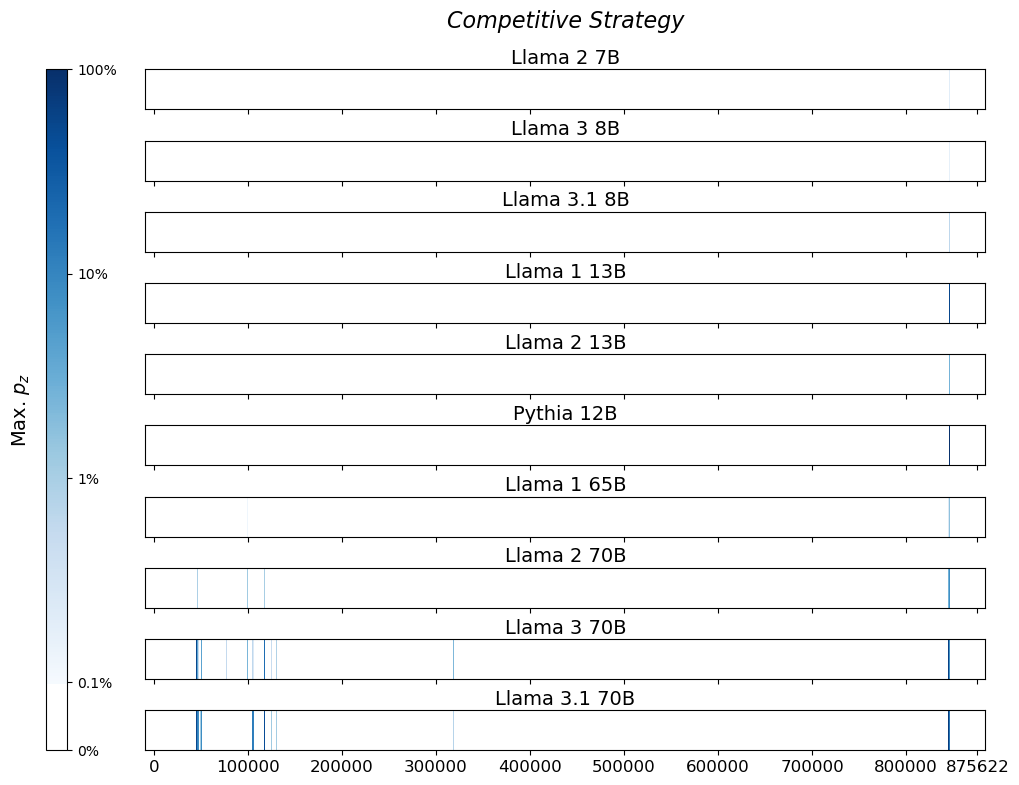}
    \includegraphics[width=\linewidth]{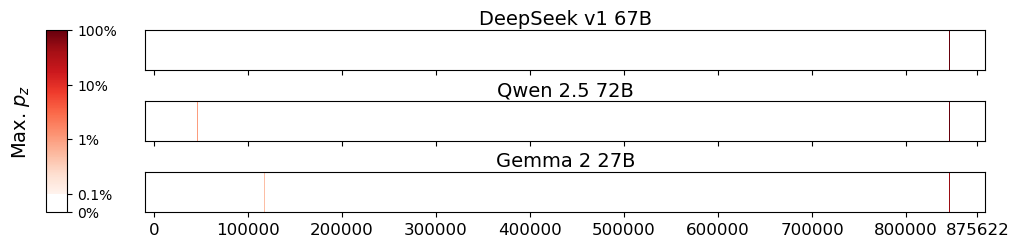}
    \includegraphics[width=\linewidth]{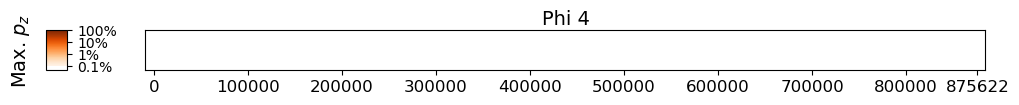}
  \end{minipage}
  \hfill
  \begin{minipage}[t]{0.45\textwidth}
    \centering
    \vspace{0cm}
    \includegraphics[width=\linewidth]{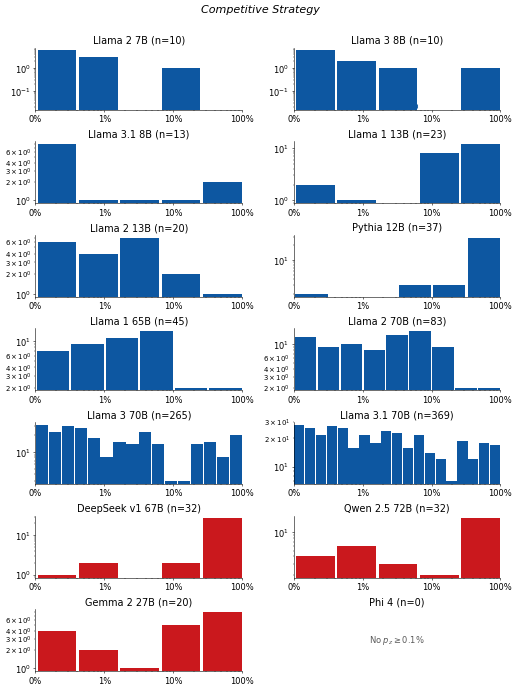}
  \end{minipage}
  \vspace{-.2cm}
  \caption{
    \textbf{\textit{Competitive Strategy}, \citeauthor{Competitive_Strategy}.}
    For $14$ LLMs,
    (\textbf{left}) heatmaps for the sliding-window procedure and
    (\textbf{right}) corresponding distributions over suffix extraction probabilities
    ($\tau_\text{min}=0.1\%$).
  }
  \label{fig:slidingwindow:Competitive_Strategy}
\end{figure}
\FloatBarrier

\subsubsection{\textit{Night Watch}, \citeauthor{Night_Watch}}\label{app:sec:sliding:Night_Watch}
\vspace{-.2cm}
\begin{figure}[h]
  \centering
  \begin{minipage}[t]{0.53\textwidth}
    \centering
    \vspace{0cm}
    \includegraphics[width=\linewidth]{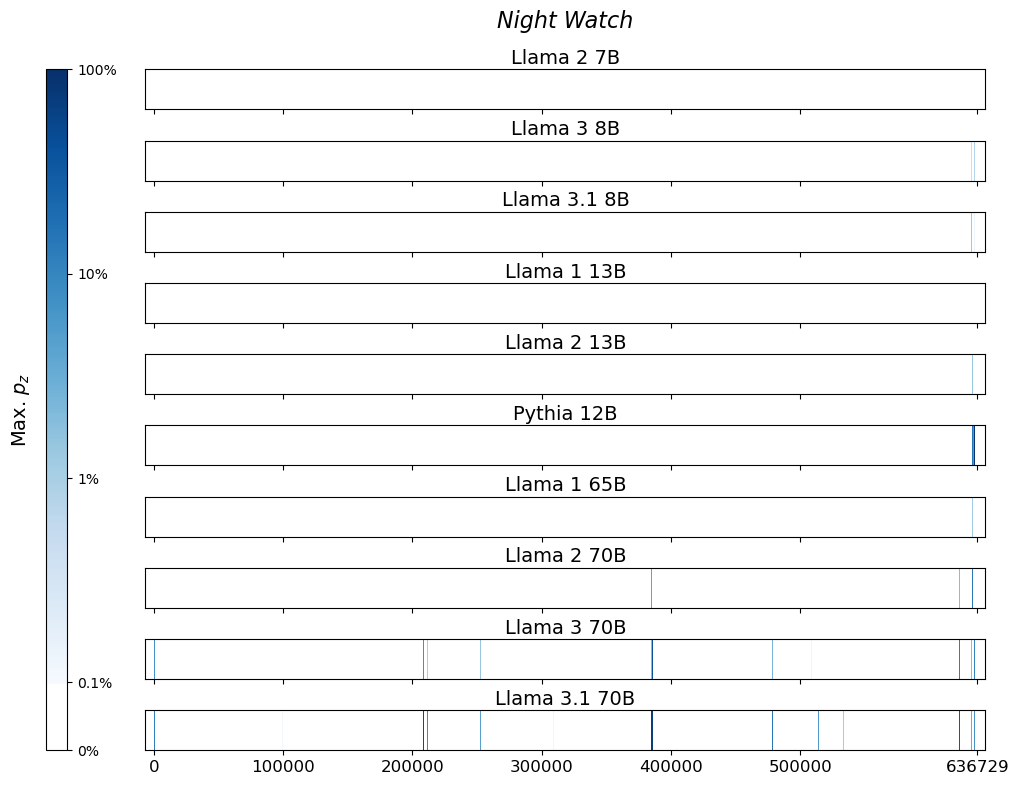}
    \includegraphics[width=\linewidth]{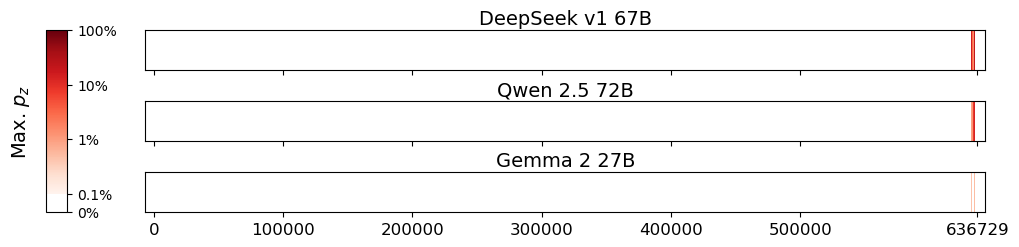}
    \includegraphics[width=\linewidth]{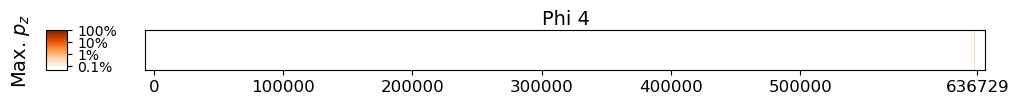}
  \end{minipage}
  \hfill
  \begin{minipage}[t]{0.45\textwidth}
    \centering
    \vspace{0cm}
    \includegraphics[width=\linewidth]{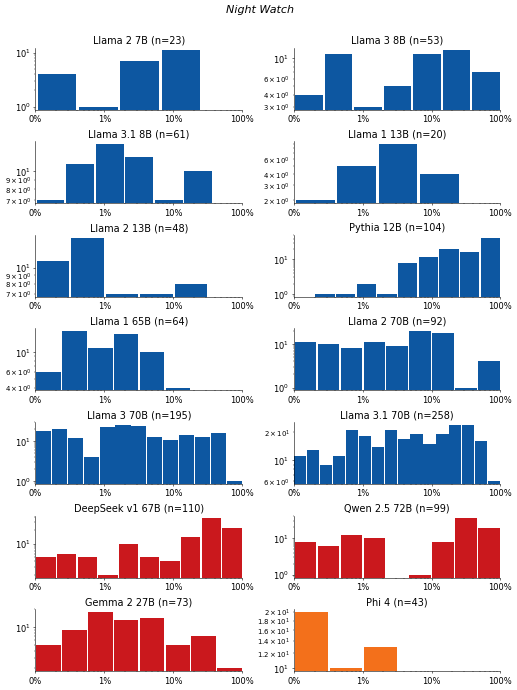}
  \end{minipage}
  \vspace{-.2cm}
  \caption{
    \textbf{\textit{Night Watch}, \citeauthor{Night_Watch}.}
    For $14$ LLMs,
    (\textbf{left}) heatmaps for the sliding-window procedure and
    (\textbf{right}) corresponding distributions over suffix extraction probabilities
    ($\tau_\text{min}=0.1\%$).
  }
  \label{fig:slidingwindow:Night_Watch}
\end{figure}
\FloatBarrier

\clearpage
\subsubsection{\textit{The Subtle Knife}, \citeauthor{The_Subtle_Knife}}\label{app:sec:sliding:The_Subtle_Knife}
\vspace{-.2cm}
\begin{figure}[h]
  \centering
  \begin{minipage}[t]{0.53\textwidth}
    \centering
    \vspace{0cm}
    \includegraphics[width=\linewidth]{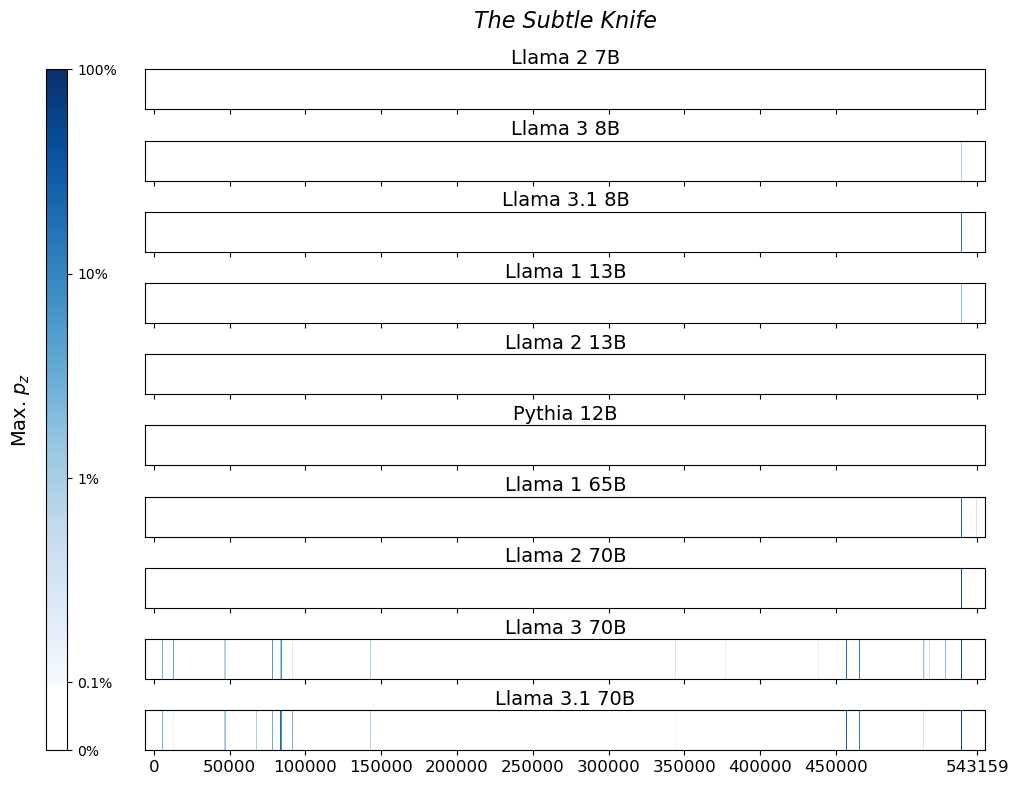}
    \includegraphics[width=\linewidth]{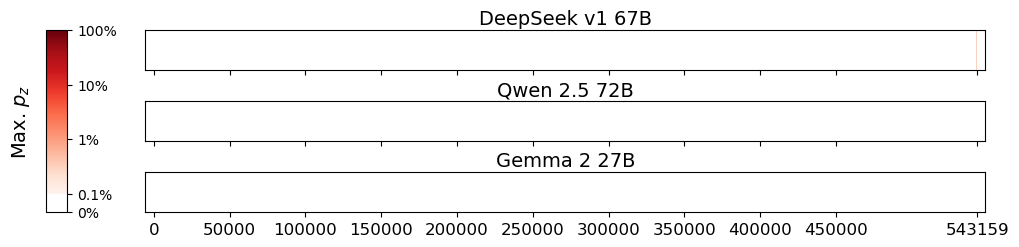}
    \includegraphics[width=\linewidth]{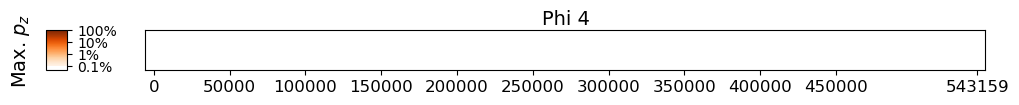}
  \end{minipage}
  \hfill
  \begin{minipage}[t]{0.45\textwidth}
    \centering
    \vspace{0cm}
    \includegraphics[width=\linewidth]{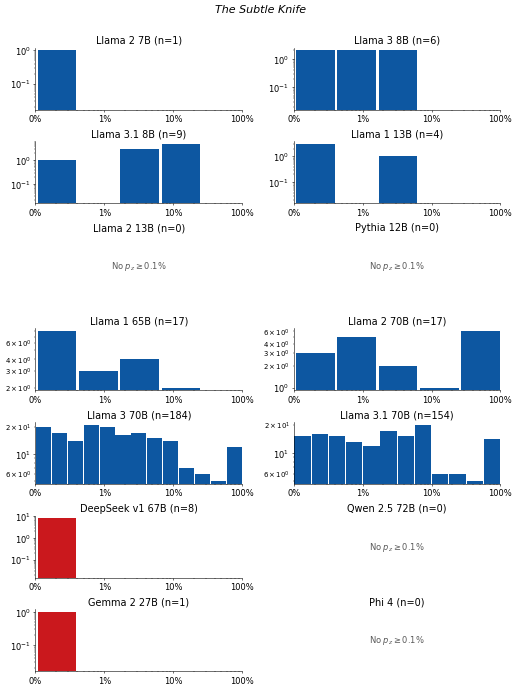}
  \end{minipage}
  \vspace{-.2cm}
  \caption{
    \textbf{\textit{The Subtle Knife}, \citeauthor{The_Subtle_Knife}.}
    For $14$ LLMs,
    (\textbf{left}) heatmaps for the sliding-window procedure and
    (\textbf{right}) corresponding distributions over suffix extraction probabilities
    ($\tau_\text{min}=0.1\%$).
  }
  \label{fig:slidingwindow:The_Subtle_Knife}
\end{figure}
\FloatBarrier

\subsubsection{\textit{The Seductions of Darwin}, \citeauthor{The_Seductions_of_Darwin}}\label{app:sec:sliding:The_Seductions_of_Darwin}
\vspace{-.2cm}
\begin{figure}[h]
  \centering
  \begin{minipage}[t]{0.53\textwidth}
    \centering
    \vspace{0cm}
    \includegraphics[width=\linewidth]{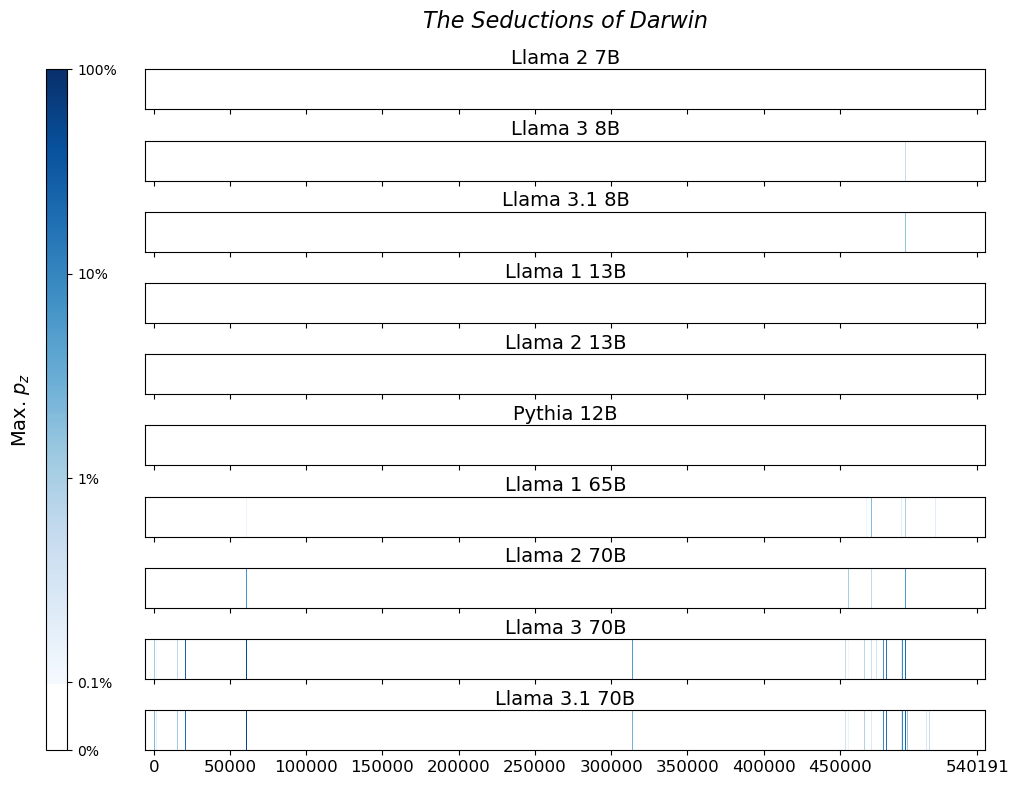}
    \includegraphics[width=\linewidth]{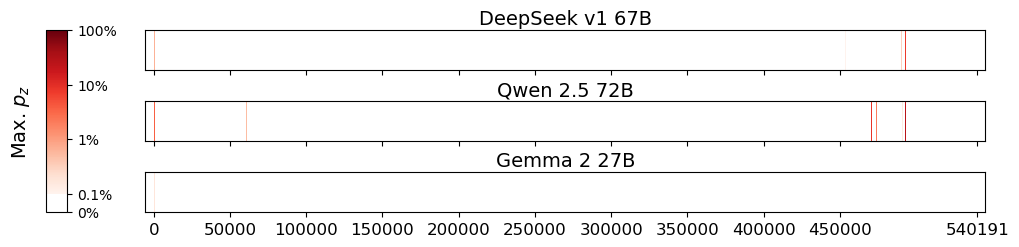}
    \includegraphics[width=\linewidth]{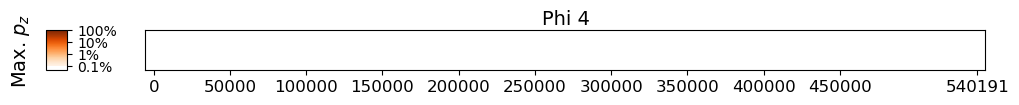}
  \end{minipage}
  \hfill
  \begin{minipage}[t]{0.45\textwidth}
    \centering
    \vspace{0cm}
    \includegraphics[width=\linewidth]{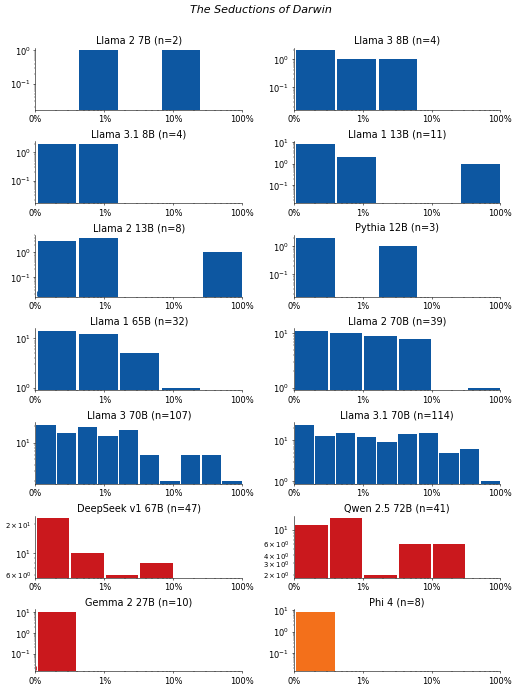}
  \end{minipage}
  \vspace{-.2cm}
  \caption{
    \textbf{\textit{The Seductions of Darwin}, \citeauthor{The_Seductions_of_Darwin}.}
    For $14$ LLMs,
    (\textbf{left}) heatmaps for the sliding-window procedure and
    (\textbf{right}) corresponding distributions over suffix extraction probabilities
    ($\tau_\text{min}=0.1\%$).
  }
  \label{fig:slidingwindow:The_Seductions_of_Darwin}
\end{figure}
\FloatBarrier

\clearpage
\subsubsection{\textit{Kitchen Table Wisdom}, \citeauthor{Kitchen_Table_Wisdom}}\label{app:sec:sliding:Kitchen_Table_Wisdom}
\vspace{-.2cm}
\begin{figure}[h]
  \centering
  \begin{minipage}[t]{0.53\textwidth}
    \centering
    \vspace{0cm}
    \includegraphics[width=\linewidth]{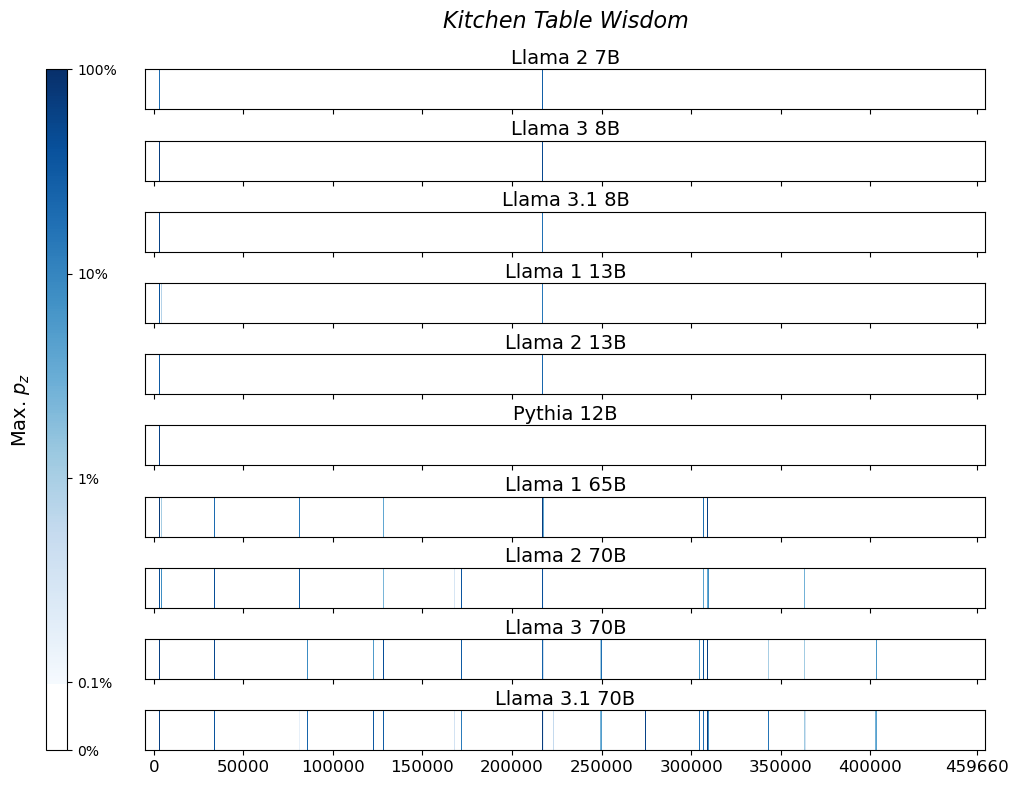}
    \includegraphics[width=\linewidth]{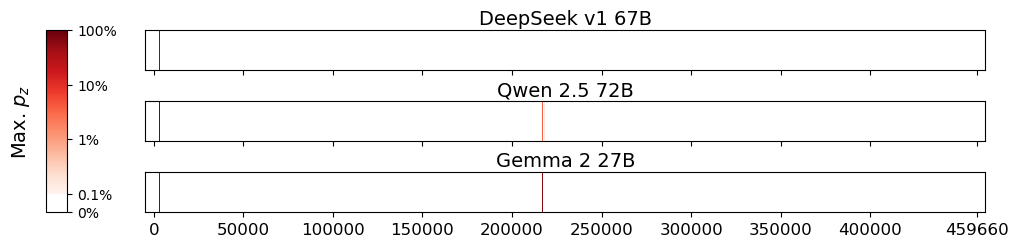}
    \includegraphics[width=\linewidth]{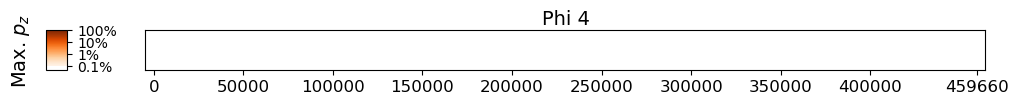}
  \end{minipage}
  \hfill
  \begin{minipage}[t]{0.45\textwidth}
    \centering
    \vspace{0cm}
    \includegraphics[width=\linewidth]{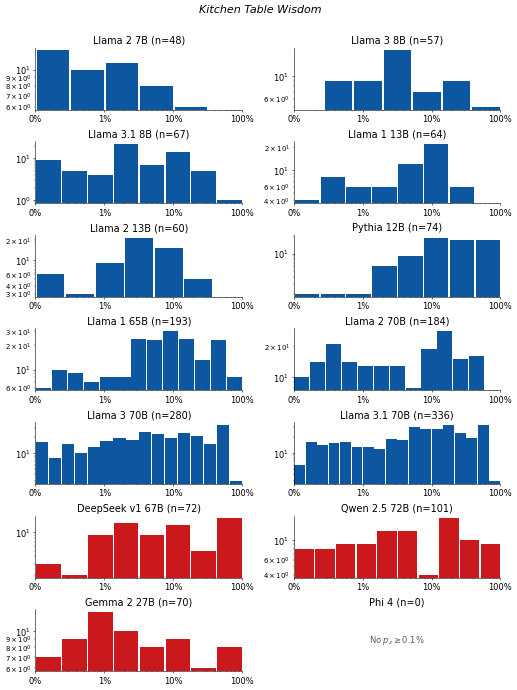}
  \end{minipage}
  \vspace{-.2cm}
  \caption{
    \textbf{\textit{Kitchen Table Wisdom}, \citeauthor{Kitchen_Table_Wisdom}.}
    For $14$ LLMs,
    (\textbf{left}) heatmaps for the sliding-window procedure and
    (\textbf{right}) corresponding distributions over suffix extraction probabilities
    ($\tau_\text{min}=0.1\%$).
  }
  \label{fig:slidingwindow:Kitchen_Table_Wisdom}
\end{figure}
\FloatBarrier

\subsubsection{\textit{Backroads Boss Lady}, \citeauthor{Backroads_Boss_Lady}}\label{app:sec:sliding:Backroads_Boss_Lady}
\vspace{-.2cm}
\begin{figure}[h]
  \centering
  \begin{minipage}[t]{0.53\textwidth}
    \centering
    \vspace{0cm}
    \includegraphics[width=\linewidth]{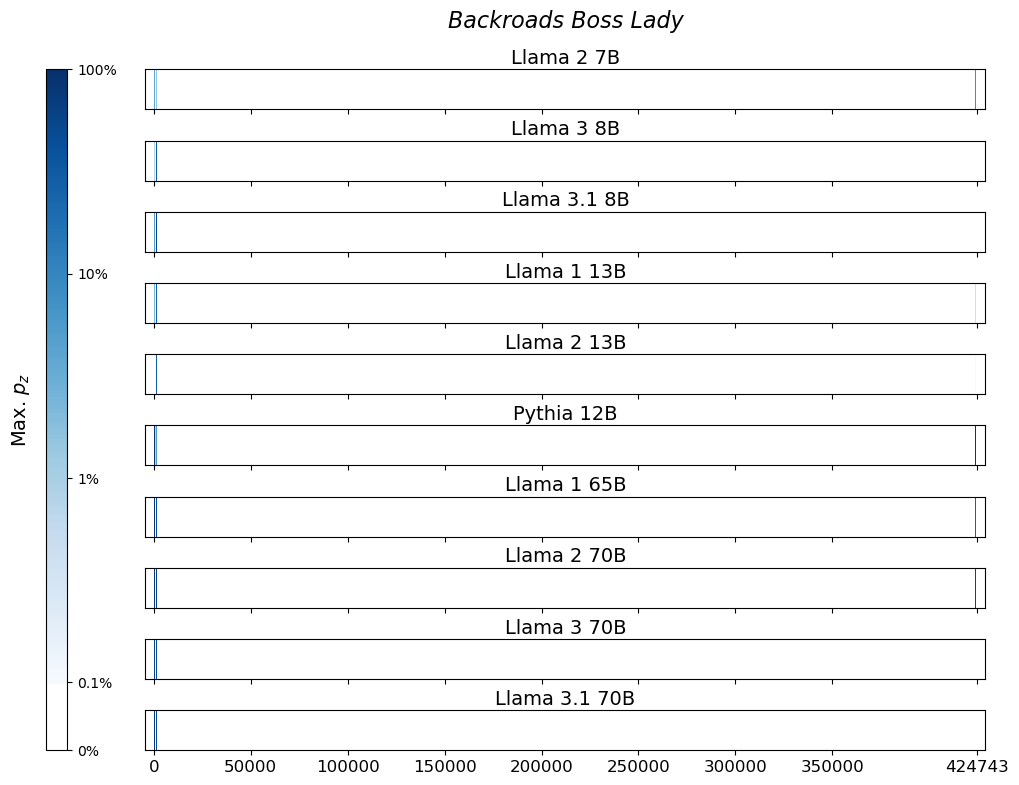}
    \includegraphics[width=\linewidth]{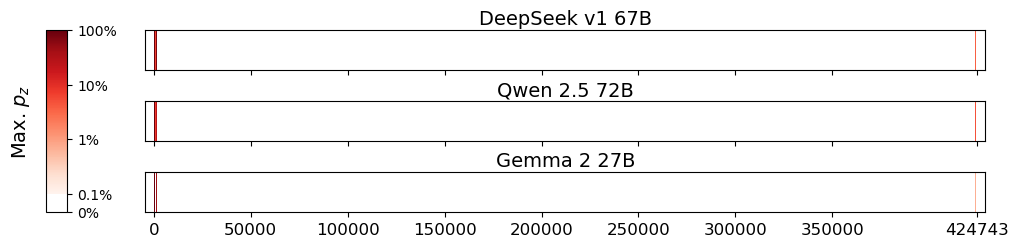}
    \includegraphics[width=\linewidth]{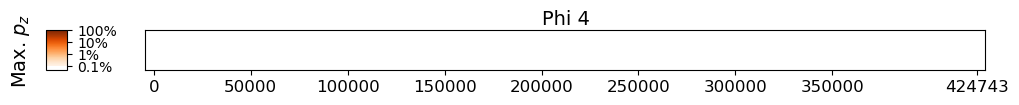}
  \end{minipage}
  \hfill
  \begin{minipage}[t]{0.45\textwidth}
    \centering
    \vspace{0cm}
    \includegraphics[width=\linewidth]{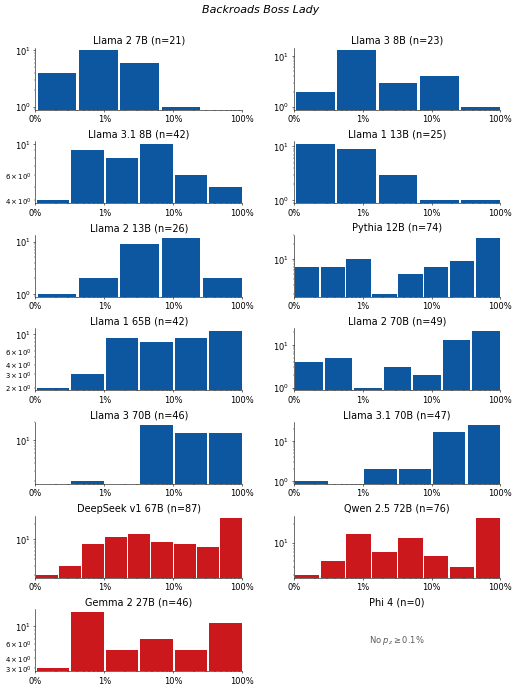}
  \end{minipage}
  \vspace{-.2cm}
  \caption{
    \textbf{\textit{Backroads Boss Lady}, \citeauthor{Backroads_Boss_Lady}.}
    For $14$ LLMs,
    (\textbf{left}) heatmaps for the sliding-window procedure and
    (\textbf{right}) corresponding distributions over suffix extraction probabilities
    ($\tau_\text{min}=0.1\%$).
  }
  \label{fig:slidingwindow:Backroads_Boss_Lady}
\end{figure}
\FloatBarrier

\clearpage
\subsubsection{\textit{Soft in the Head}, \citeauthor{Soft_in_the_Head}}\label{app:sec:sliding:Soft_in_the_Head}
\vspace{-.2cm}
\begin{figure}[h]
  \centering
  \begin{minipage}[t]{0.53\textwidth}
    \centering
    \vspace{0cm}
    \includegraphics[width=\linewidth]{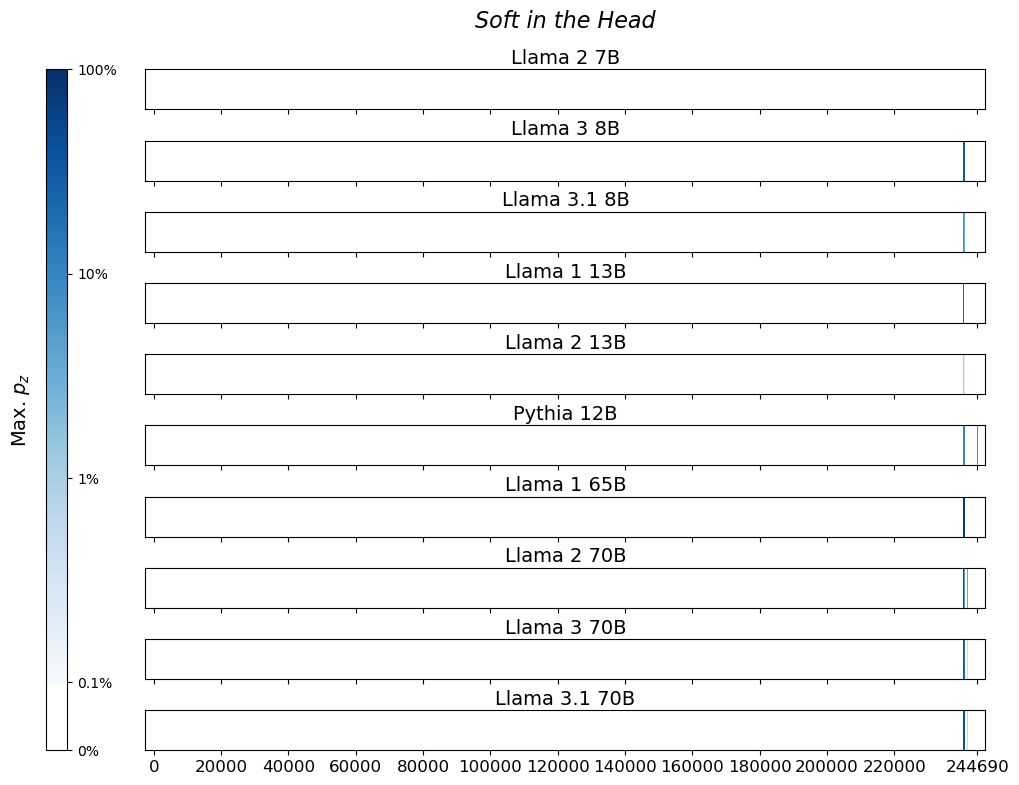}
    \includegraphics[width=\linewidth]{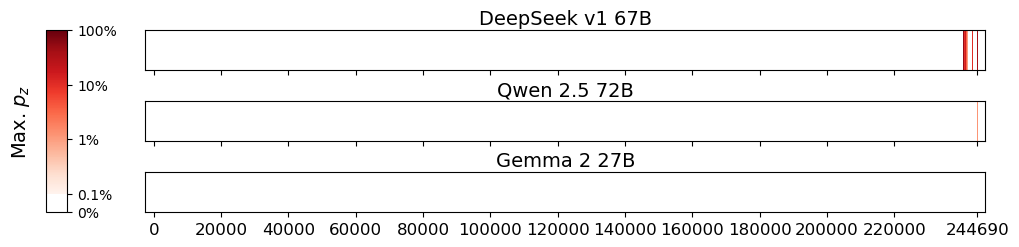}
    \includegraphics[width=\linewidth]{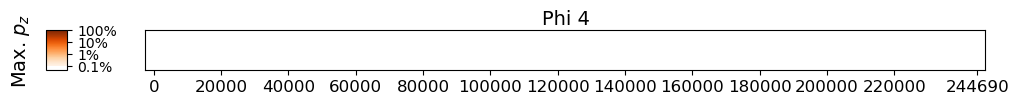}
  \end{minipage}
  \hfill
  \begin{minipage}[t]{0.45\textwidth}
    \centering
    \vspace{0cm}
    \includegraphics[width=\linewidth]{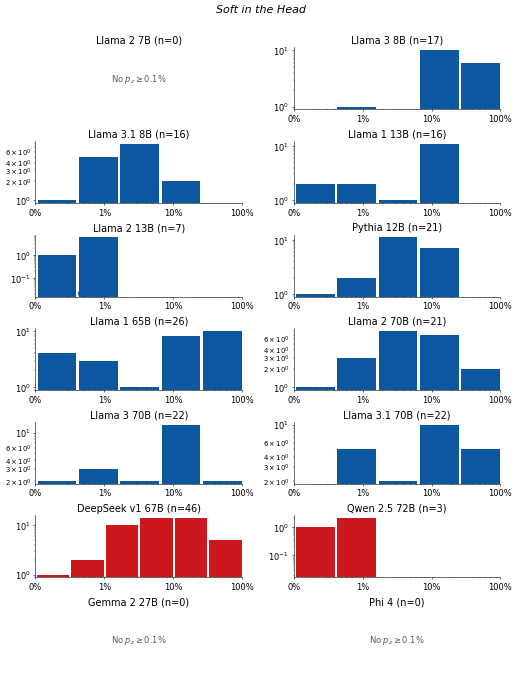}
  \end{minipage}
  \vspace{-.2cm}
  \caption{
    \textbf{\textit{Soft in the Head}, \citeauthor{Soft_in_the_Head}.}
    For $14$ LLMs,
    (\textbf{left}) heatmaps for the sliding-window procedure and
    (\textbf{right}) corresponding distributions over suffix extraction probabilities
    ($\tau_\text{min}=0.1\%$).
  }
  \label{fig:slidingwindow:Soft_in_the_Head}
\end{figure}
\FloatBarrier

\subsubsection{\textit{The Making of a Mediterranean Emirate}, \citeauthor{The_Making_of_a_Mediterranean_Emirate}}\label{app:sec:sliding:The_Making_of_a_Mediterranean_Emirate}
\vspace{-.2cm}
\begin{figure}[h]
  \centering
  \begin{minipage}[t]{0.53\textwidth}
    \centering
    \vspace{0cm}
    \includegraphics[width=\linewidth]{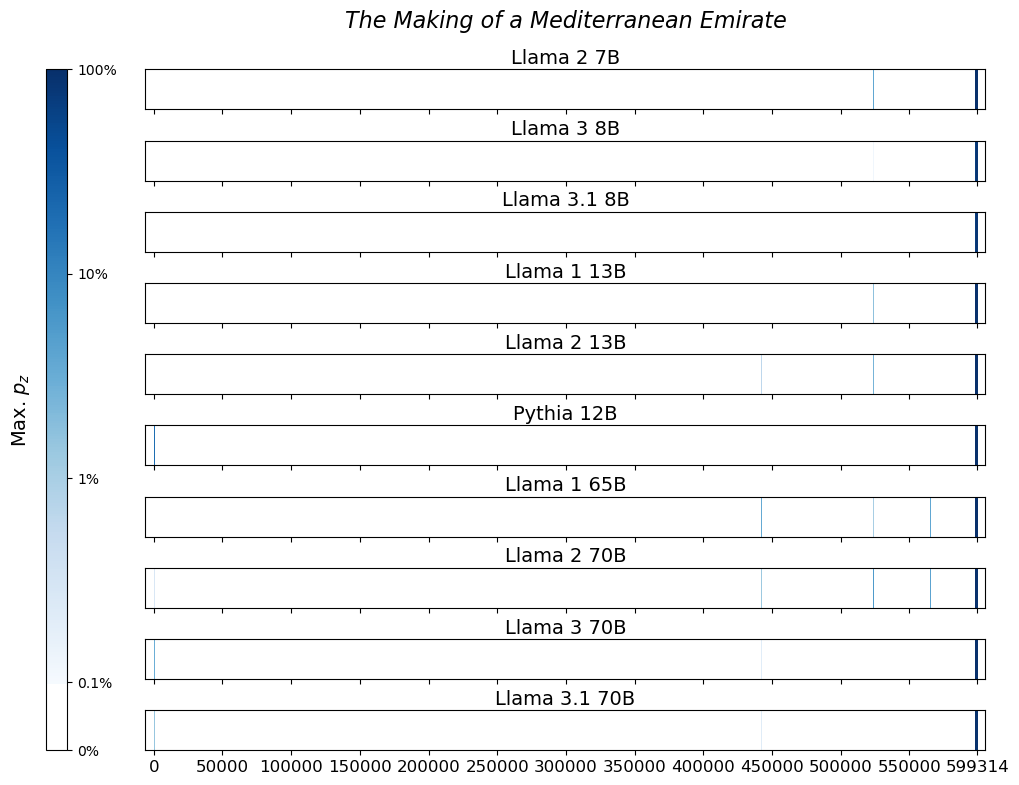}
    \includegraphics[width=\linewidth]{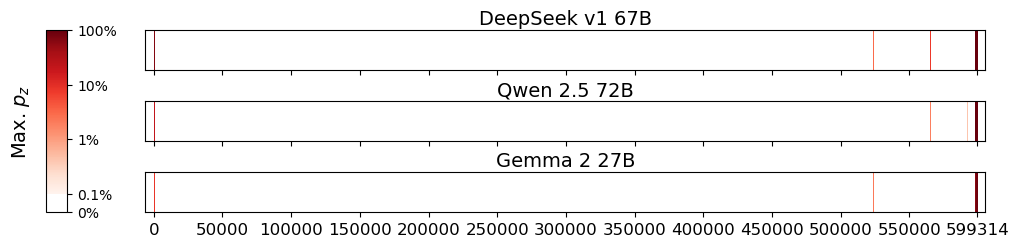}
    \includegraphics[width=\linewidth]{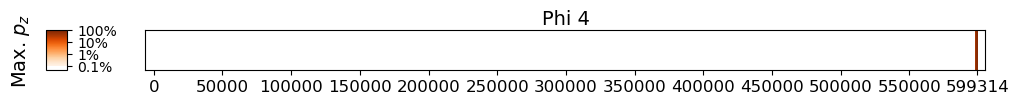}
  \end{minipage}
  \hfill
  \begin{minipage}[t]{0.45\textwidth}
    \centering
    \vspace{0cm}
    \includegraphics[width=\linewidth]{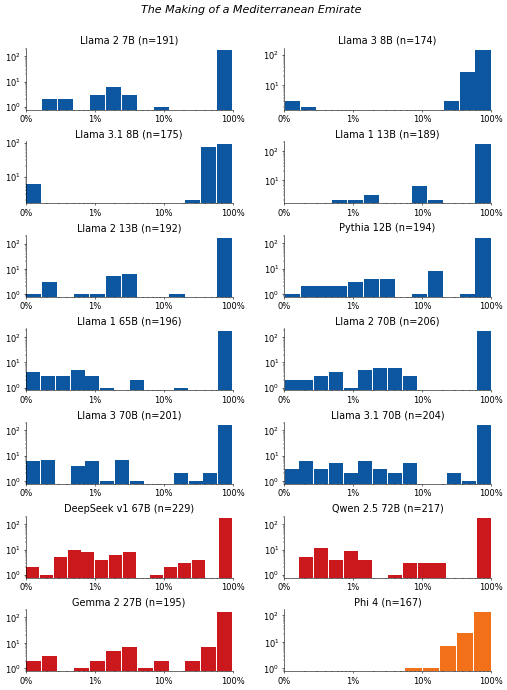}
  \end{minipage}
  \vspace{-.2cm}
  \caption{
    \textbf{\textit{The Making of a Mediterranean Emirate}, \citeauthor{The_Making_of_a_Mediterranean_Emirate}.}
    For $14$ LLMs,
    (\textbf{left}) heatmaps for the sliding-window procedure and
    (\textbf{right}) corresponding distributions over suffix extraction probabilities
    ($\tau_\text{min}=0.1\%$).
  }
  \label{fig:slidingwindow:The_Making_of_a_Mediterranean_Emirate}
\end{figure}
\FloatBarrier

\clearpage
\subsubsection{\textit{Harry Potter and the Sorcerer's Stone}, \citeauthor{Harry_Potter_and_the_Sorcerer_s_Stone}}\label{app:sec:sliding:Harry_Potter_and_the_Sorcerer_s_Stone}
\vspace{-.2cm}
\begin{figure}[h]
  \centering
  \begin{minipage}[t]{0.53\textwidth}
    \centering
    \vspace{0cm}
    \includegraphics[width=\linewidth]{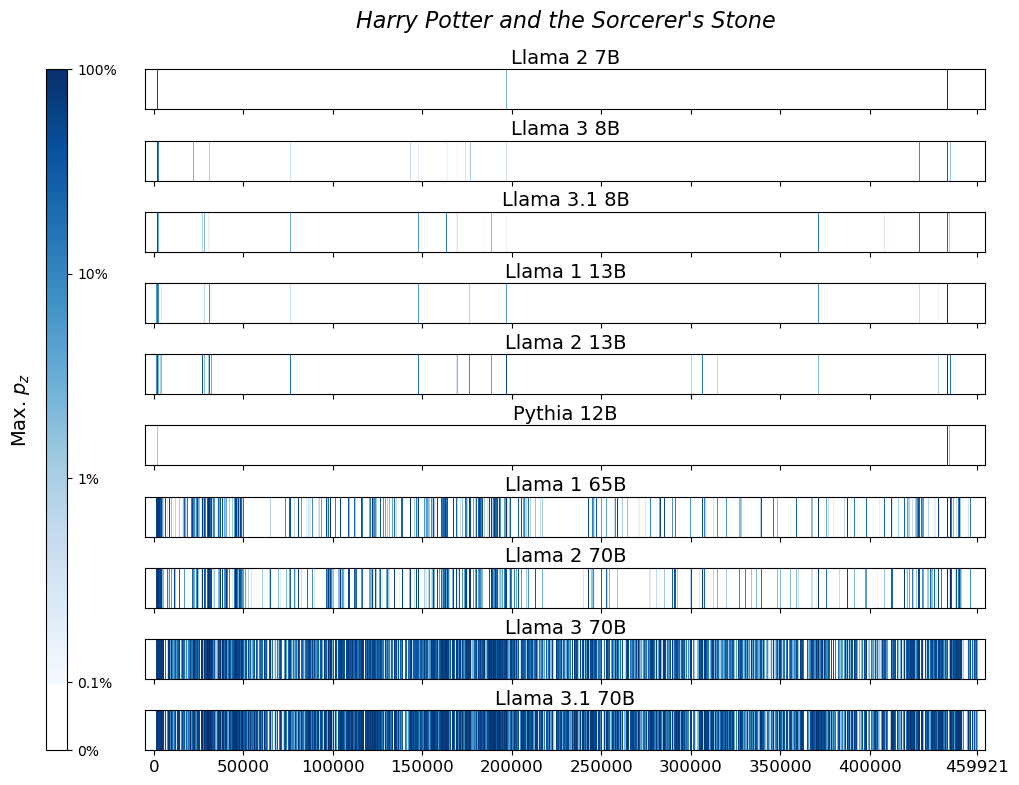}
    \includegraphics[width=\linewidth]{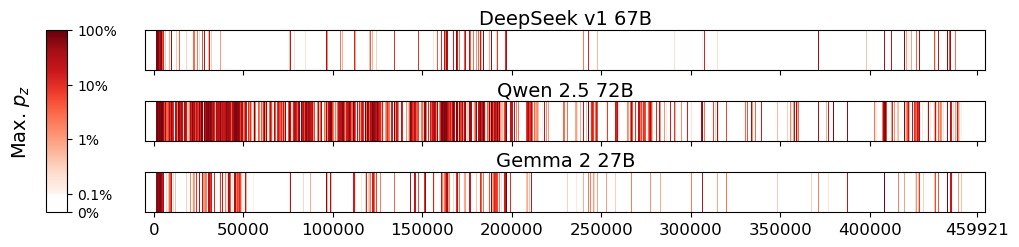}
    \includegraphics[width=\linewidth]{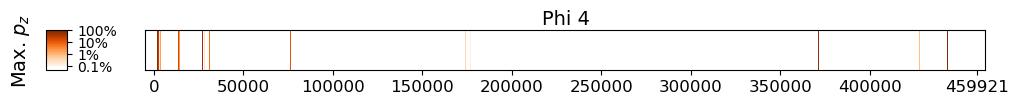}
  \end{minipage}
  \hfill
  \begin{minipage}[t]{0.45\textwidth}
    \centering
    \vspace{0cm}
    \includegraphics[width=\linewidth]{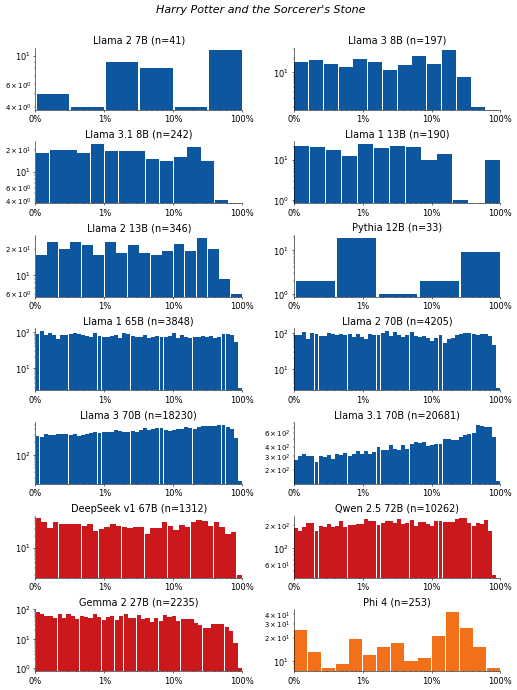}
  \end{minipage}
  \vspace{-.2cm}
  \caption{
    \textbf{\textit{Harry Potter and the Sorcerer's Stone}, \citeauthor{Harry_Potter_and_the_Sorcerer_s_Stone}.}
    For $14$ LLMs,
    (\textbf{left}) heatmaps for the sliding-window procedure and
    (\textbf{right}) corresponding distributions over suffix extraction probabilities
    ($\tau_\text{min}=0.1\%$).
  }
  \label{fig:slidingwindow:Harry_Potter_and_the_Sorcerer_s_Stone}
\end{figure}
\FloatBarrier

\subsubsection{\textit{Harry Potter and the Chamber of Secrets}, \citeauthor{Harry_Potter_and_the_Chamber_of_Secrets}}\label{app:sec:sliding:Harry_Potter_and_the_Chamber_of_Secrets}
\vspace{-.2cm}
\begin{figure}[h]
  \centering
  \begin{minipage}[t]{0.53\textwidth}
    \centering
    \vspace{0cm}
    \includegraphics[width=\linewidth]{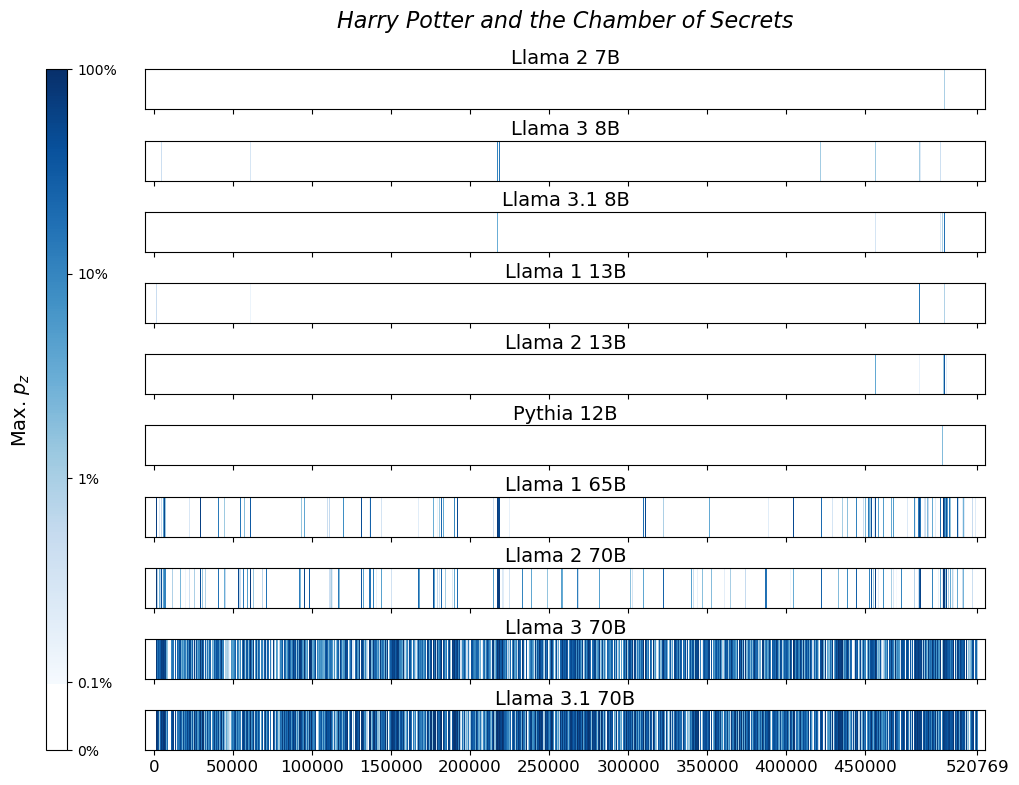}
    \includegraphics[width=\linewidth]{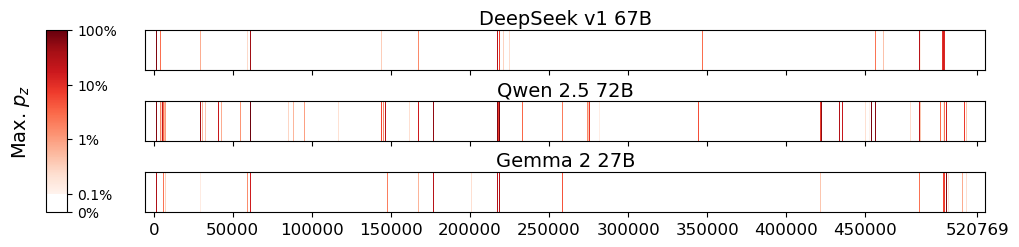}
    \includegraphics[width=\linewidth]{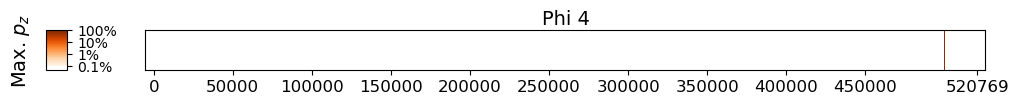}
  \end{minipage}
  \hfill
  \begin{minipage}[t]{0.45\textwidth}
    \centering
    \vspace{0cm}
    \includegraphics[width=\linewidth]{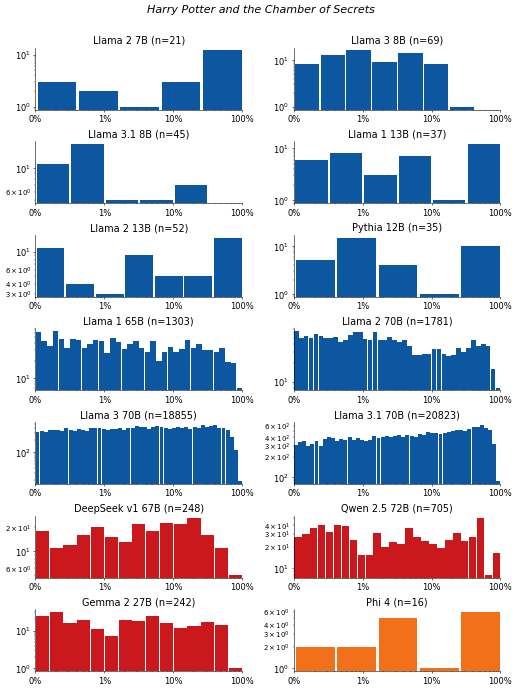}
  \end{minipage}
  \vspace{-.2cm}
  \caption{
    \textbf{\textit{Harry Potter and the Chamber of Secrets}, \citeauthor{Harry_Potter_and_the_Chamber_of_Secrets}.}
    For $14$ LLMs,
    (\textbf{left}) heatmaps for the sliding-window procedure and
    (\textbf{right}) corresponding distributions over suffix extraction probabilities
    ($\tau_\text{min}=0.1\%$).
  }
  \label{fig:slidingwindow:Harry_Potter_and_the_Chamber_of_Secrets}
\end{figure}
\FloatBarrier

\clearpage
\subsubsection{\textit{Harry Potter and the Goblet of Fire}, \citeauthor{Harry_Potter_and_the_Goblet_of_Fire}}\label{app:sec:sliding:Harry_Potter_and_the_Goblet_of_Fire}
\vspace{-.2cm}
\begin{figure}[h]
  \centering
  \begin{minipage}[t]{0.53\textwidth}
    \centering
    \vspace{0cm}
    \includegraphics[width=\linewidth]{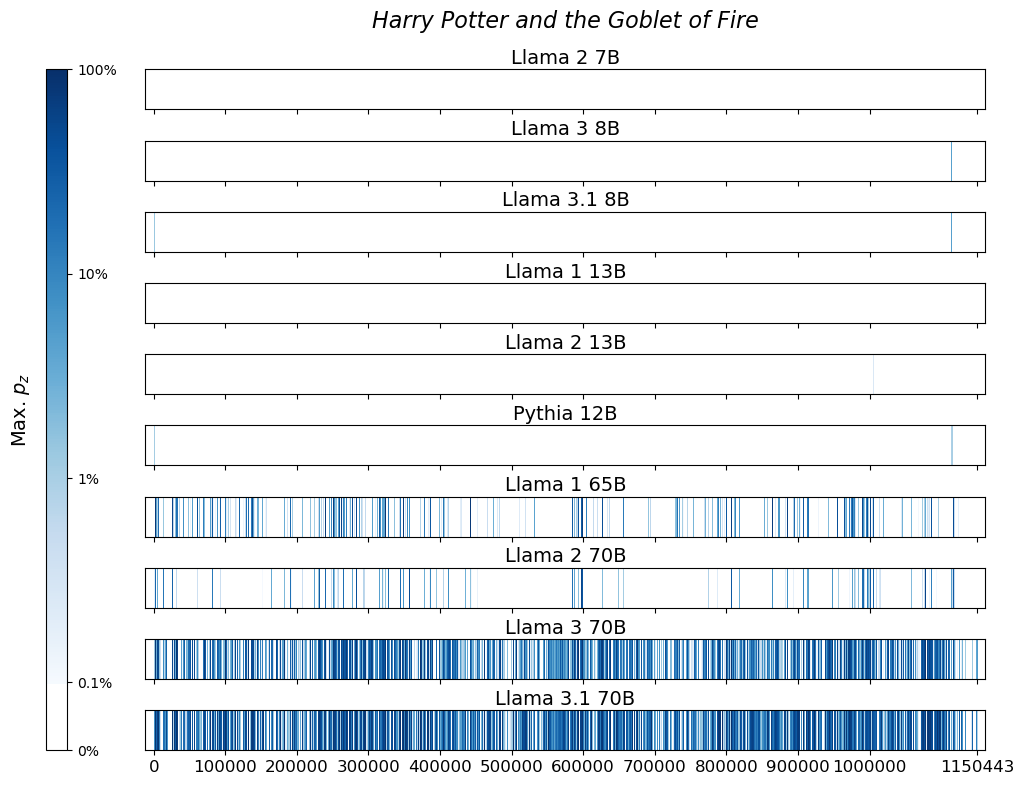}
    \includegraphics[width=\linewidth]{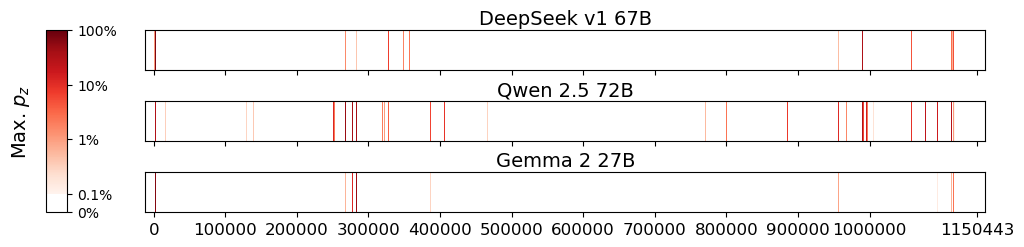}
    \includegraphics[width=\linewidth]{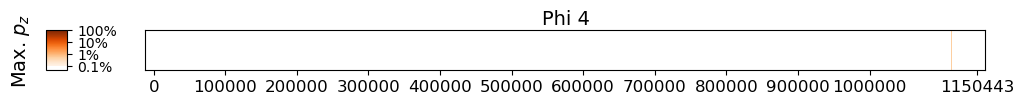}
  \end{minipage}
  \hfill
  \begin{minipage}[t]{0.45\textwidth}
    \centering
    \vspace{0cm}
    \includegraphics[width=\linewidth]{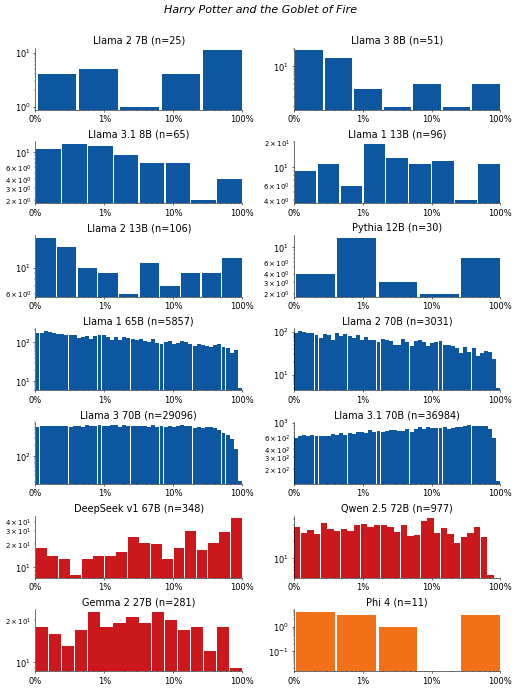}
  \end{minipage}
  \vspace{-.2cm}
  \caption{
    \textbf{\textit{Harry Potter and the Goblet of Fire}, \citeauthor{Harry_Potter_and_the_Goblet_of_Fire}.}
    For $14$ LLMs,
    (\textbf{left}) heatmaps for the sliding-window procedure and
    (\textbf{right}) corresponding distributions over suffix extraction probabilities
    ($\tau_\text{min}=0.1\%$).
  }
  \label{fig:slidingwindow:Harry_Potter_and_the_Goblet_of_Fire}
\end{figure}
\FloatBarrier

\subsubsection{\textit{Harry Potter and the Deathly Hallows}, \citeauthor{Harry_Potter_and_the_Deathly_Hallows}}\label{app:sec:sliding:Harry_Potter_and_the_Deathly_Hallows}
\vspace{-.2cm}
\begin{figure}[h]
  \centering
  \begin{minipage}[t]{0.53\textwidth}
    \centering
    \vspace{0cm}
    \includegraphics[width=\linewidth]{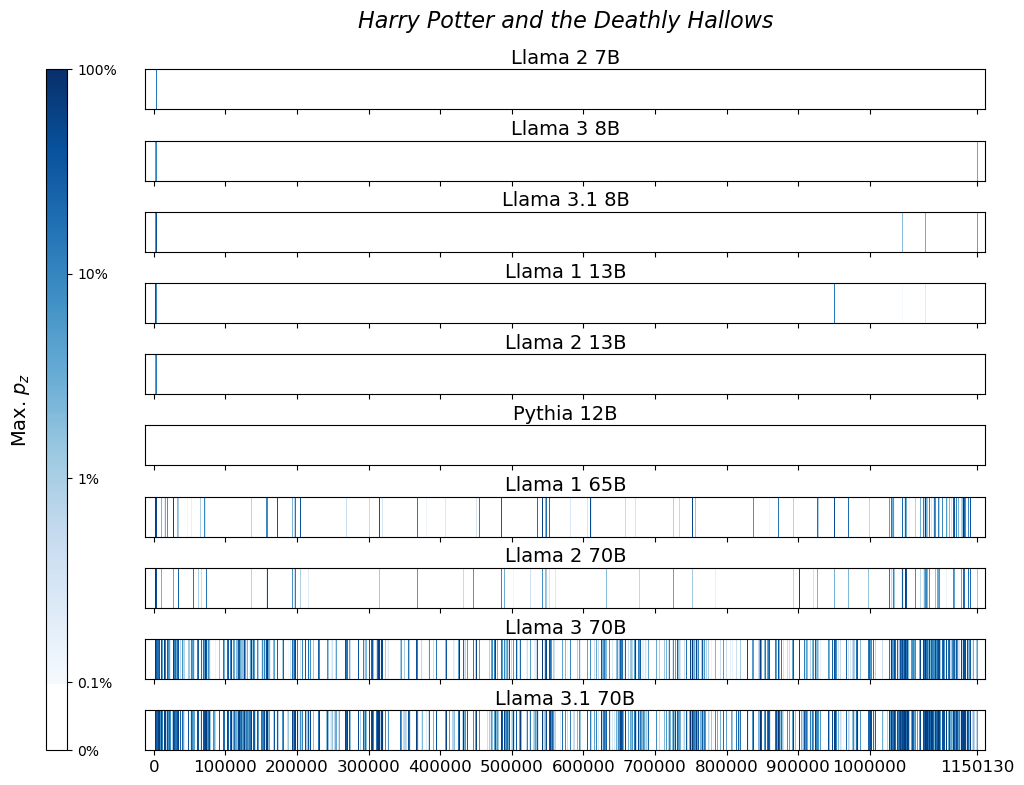}
    \includegraphics[width=\linewidth]{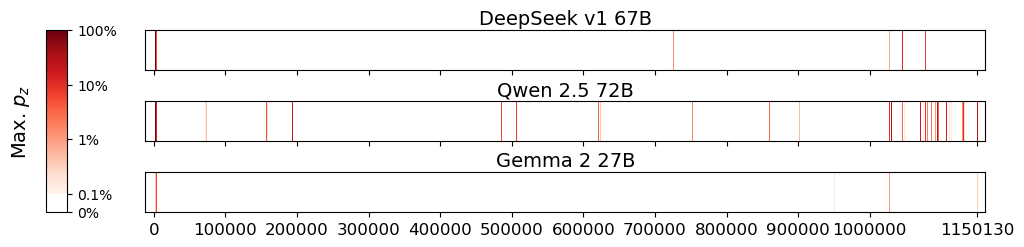}
    \includegraphics[width=\linewidth]{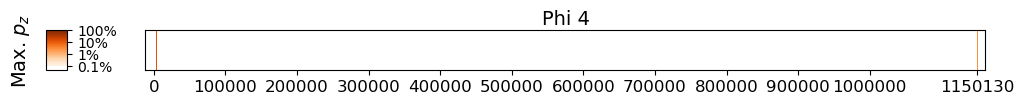}
  \end{minipage}
  \hfill
  \begin{minipage}[t]{0.45\textwidth}
    \centering
    \vspace{0cm}
    \includegraphics[width=\linewidth]{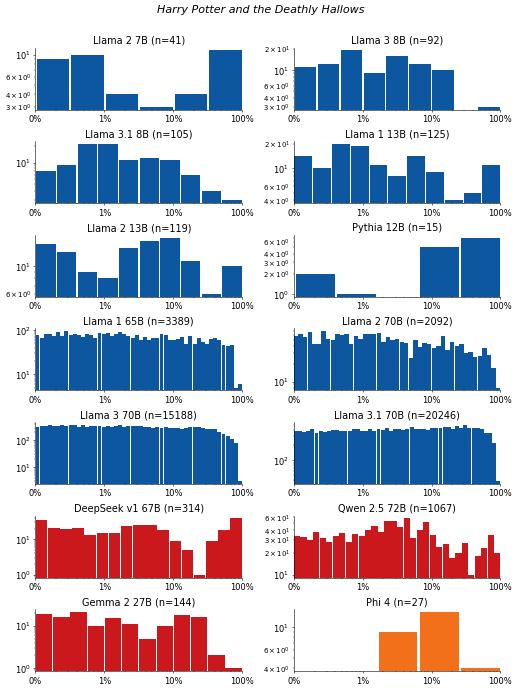}
  \end{minipage}
  \vspace{-.2cm}
  \caption{
    \textbf{\textit{Harry Potter and the Deathly Hallows}, \citeauthor{Harry_Potter_and_the_Deathly_Hallows}.}
    For $14$ LLMs,
    (\textbf{left}) heatmaps for the sliding-window procedure and
    (\textbf{right}) corresponding distributions over suffix extraction probabilities
    ($\tau_\text{min}=0.1\%$).
  }
  \label{fig:slidingwindow:Harry_Potter_and_the_Deathly_Hallows}
\end{figure}
\FloatBarrier

\clearpage
\subsubsection{\textit{Born to Walk}, \citeauthor{Born_to_Walk}}\label{app:sec:sliding:Born_to_Walk}
\vspace{-.2cm}
\begin{figure}[h]
  \centering
  \begin{minipage}[t]{0.53\textwidth}
    \centering
    \vspace{0cm}
    \includegraphics[width=\linewidth]{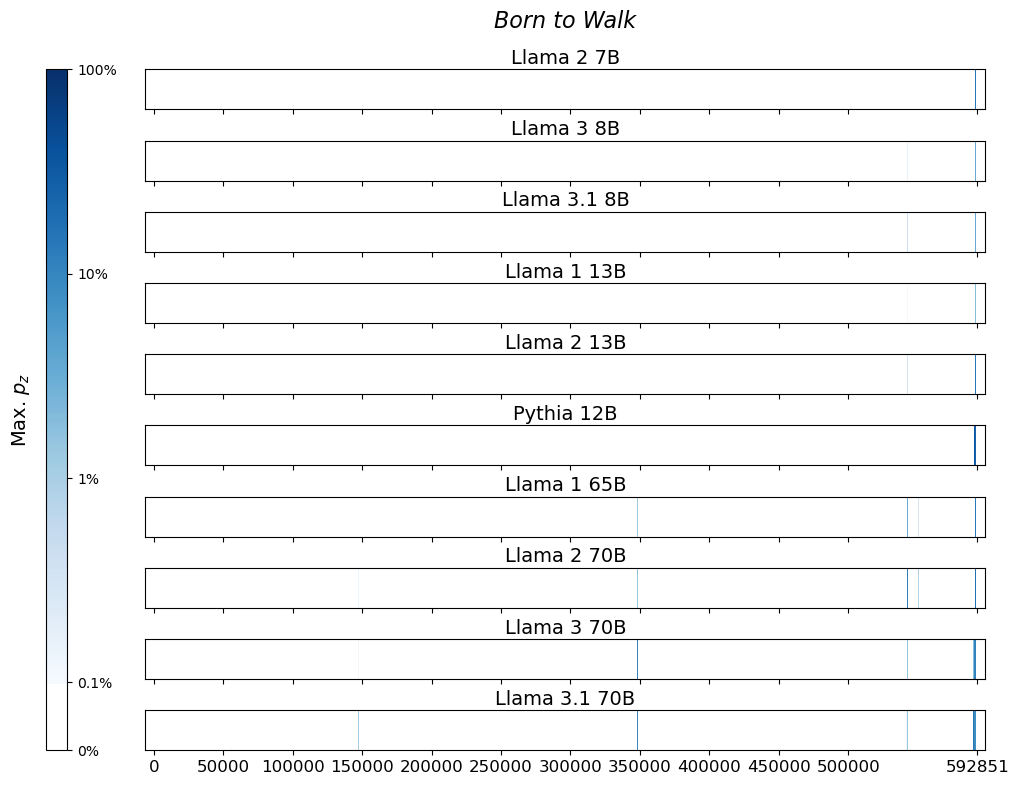}
    \includegraphics[width=\linewidth]{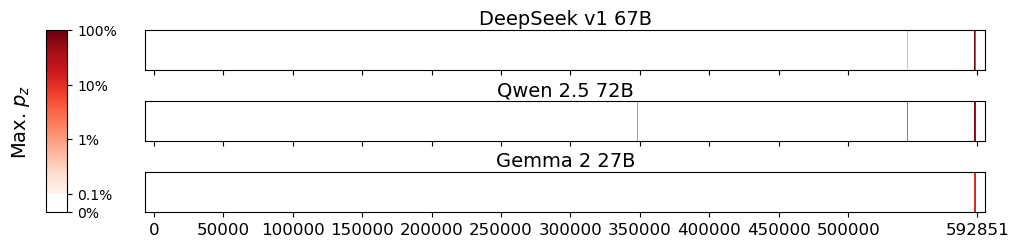}
    \includegraphics[width=\linewidth]{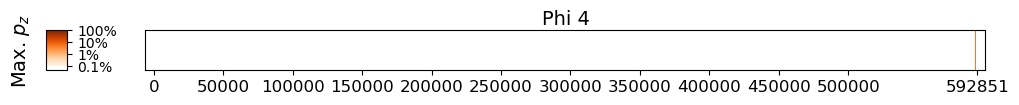}
  \end{minipage}
  \hfill
  \begin{minipage}[t]{0.45\textwidth}
    \centering
    \vspace{0cm}
    \includegraphics[width=\linewidth]{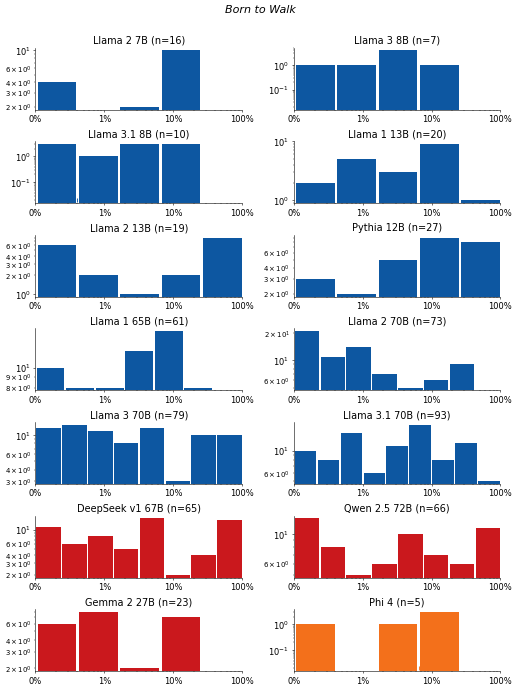}
  \end{minipage}
  \vspace{-.2cm}
  \caption{
    \textbf{\textit{Born to Walk}, \citeauthor{Born_to_Walk}.}
    For $14$ LLMs,
    (\textbf{left}) heatmaps for the sliding-window procedure and
    (\textbf{right}) corresponding distributions over suffix extraction probabilities
    ($\tau_\text{min}=0.1\%$).
  }
  \label{fig:slidingwindow:Born_to_Walk}
\end{figure}
\FloatBarrier

\subsubsection{\textit{The Pretender}, \citeauthor{The_Pretender}}\label{app:sec:sliding:The_Pretender}
\vspace{-.2cm}
\begin{figure}[h]
  \centering
  \begin{minipage}[t]{0.53\textwidth}
    \centering
    \vspace{0cm}
    \includegraphics[width=\linewidth]{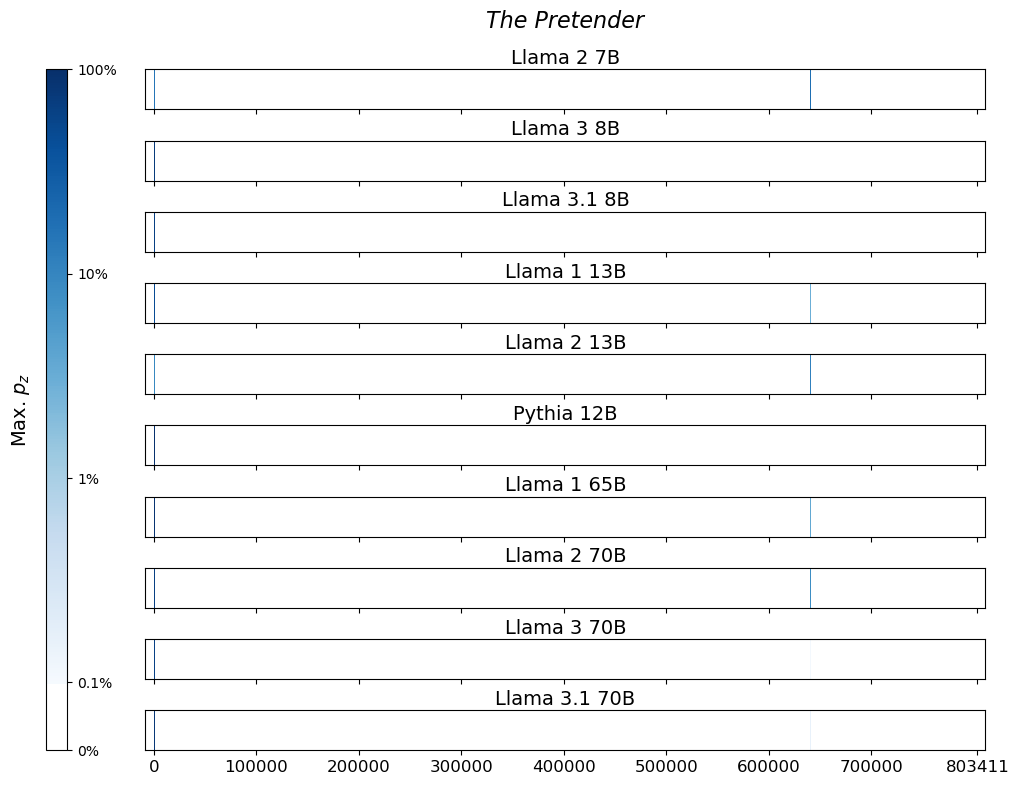}
    \includegraphics[width=\linewidth]{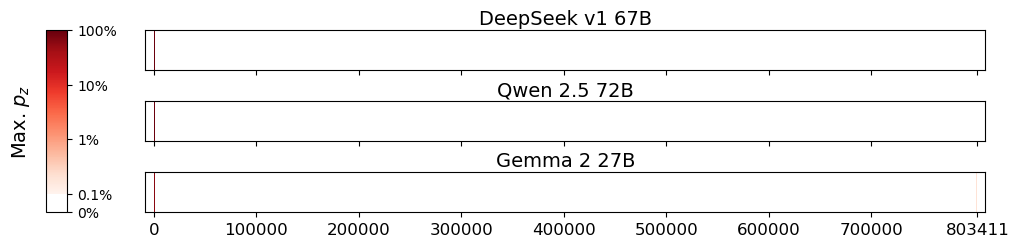}
    \includegraphics[width=\linewidth]{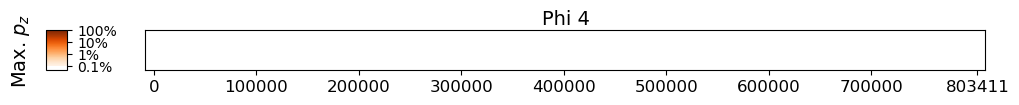}
  \end{minipage}
  \hfill
  \begin{minipage}[t]{0.45\textwidth}
    \centering
    \vspace{0cm}
    \includegraphics[width=\linewidth]{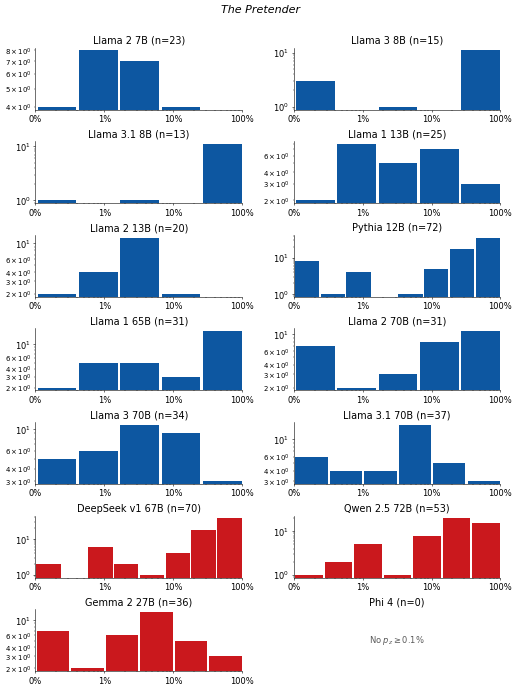}
  \end{minipage}
  \vspace{-.2cm}
  \caption{
    \textbf{\textit{The Pretender}, \citeauthor{The_Pretender}.}
    For $14$ LLMs,
    (\textbf{left}) heatmaps for the sliding-window procedure and
    (\textbf{right}) corresponding distributions over suffix extraction probabilities
    ($\tau_\text{min}=0.1\%$).
  }
  \label{fig:slidingwindow:The_Pretender}
\end{figure}
\FloatBarrier

\clearpage
\subsubsection{\textit{Toscanini}, \citeauthor{Toscanini}}\label{app:sec:sliding:Toscanini}
\vspace{-.2cm}
\begin{figure}[h]
  \centering
  \begin{minipage}[t]{0.53\textwidth}
    \centering
    \vspace{0cm}
    \includegraphics[width=\linewidth]{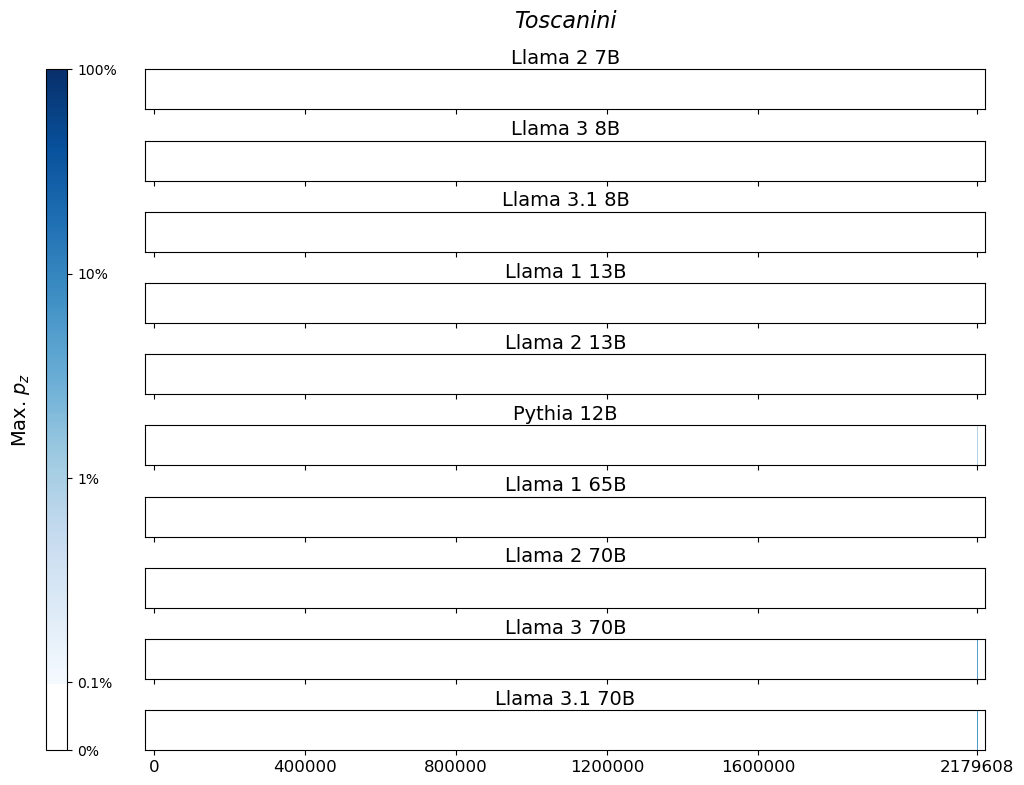}
    \includegraphics[width=\linewidth]{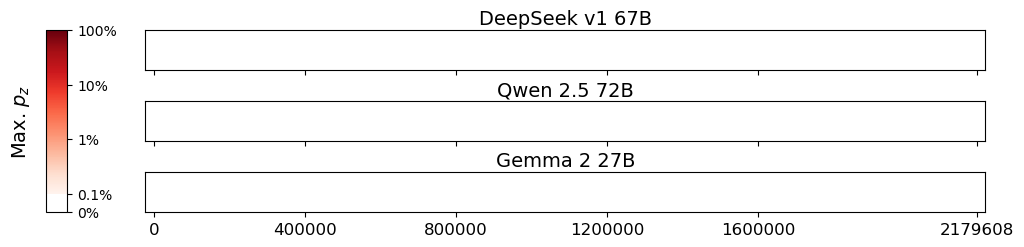}
    \includegraphics[width=\linewidth]{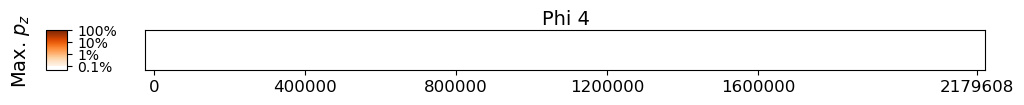}
  \end{minipage}
  \hfill
  \begin{minipage}[t]{0.45\textwidth}
    \centering
    \vspace{0cm}
    \includegraphics[width=\linewidth]{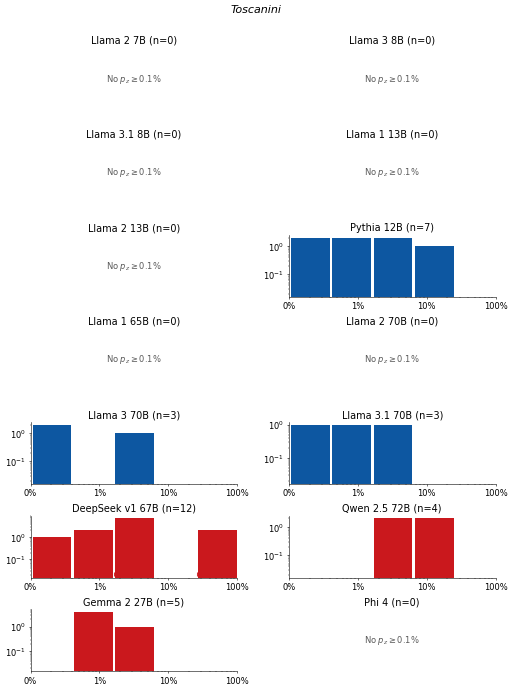}
  \end{minipage}
  \vspace{-.2cm}
  \caption{
    \textbf{\textit{Toscanini}, \citeauthor{Toscanini}.}
    For $14$ LLMs,
    (\textbf{left}) heatmaps for the sliding-window procedure and
    (\textbf{right}) corresponding distributions over suffix extraction probabilities
    ($\tau_\text{min}=0.1\%$).
  }
  \label{fig:slidingwindow:Toscanini}
\end{figure}
\FloatBarrier

\subsubsection{\textit{Cosmos}, \citeauthor{Cosmos}}\label{app:sec:sliding:Cosmos}
\vspace{-.2cm}
\begin{figure}[h]
  \centering
  \begin{minipage}[t]{0.53\textwidth}
    \centering
    \vspace{0cm}
    \includegraphics[width=\linewidth]{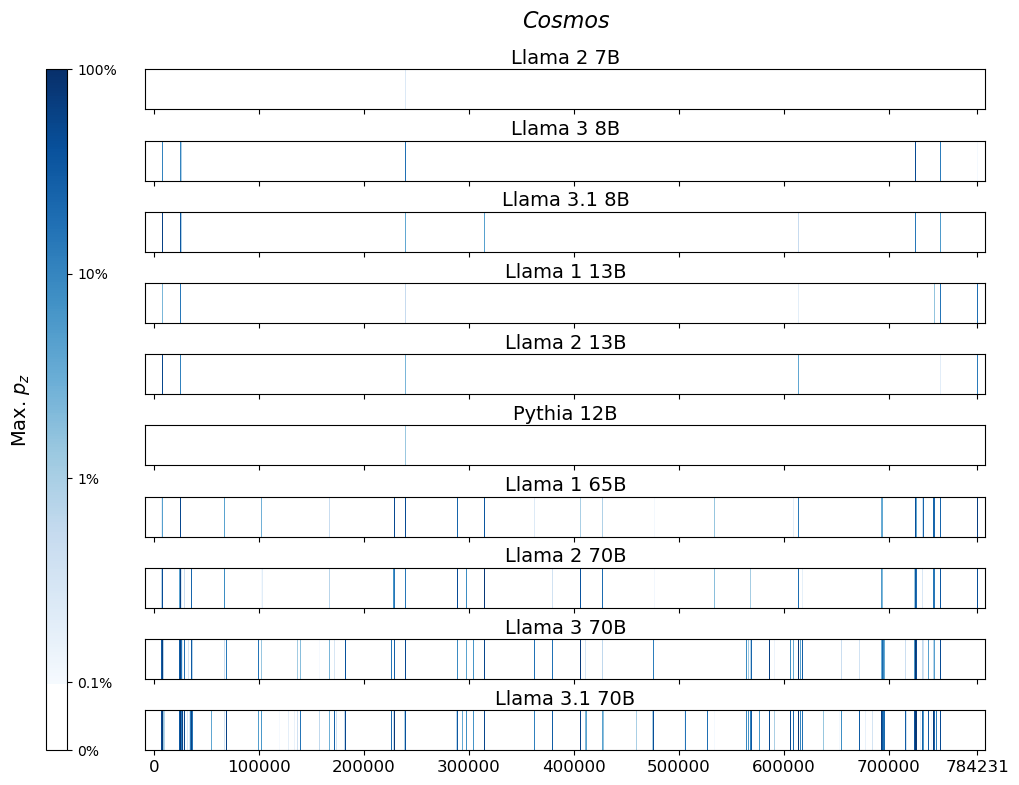}
    \includegraphics[width=\linewidth]{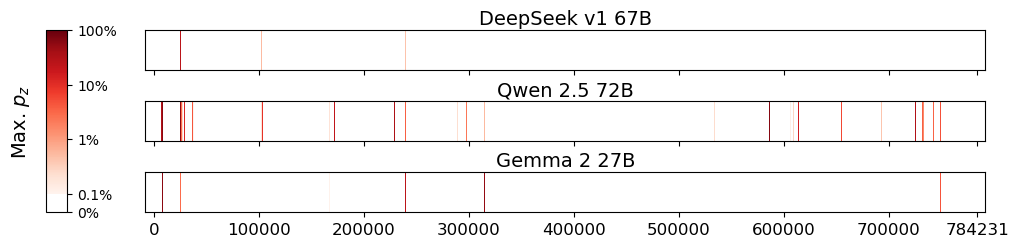}
    \includegraphics[width=\linewidth]{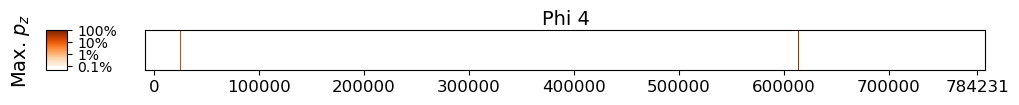}
  \end{minipage}
  \hfill
  \begin{minipage}[t]{0.45\textwidth}
    \centering
    \vspace{0cm}
    \includegraphics[width=\linewidth]{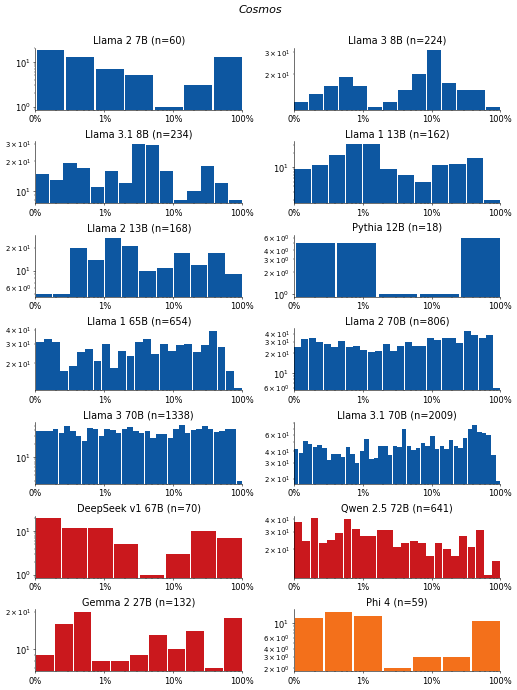}
  \end{minipage}
  \vspace{-.2cm}
  \caption{
    \textbf{\textit{Cosmos}, \citeauthor{Cosmos}.}
    For $14$ LLMs,
    (\textbf{left}) heatmaps for the sliding-window procedure and
    (\textbf{right}) corresponding distributions over suffix extraction probabilities
    ($\tau_\text{min}=0.1\%$).
  }
  \label{fig:slidingwindow:Cosmos}
\end{figure}
\FloatBarrier

\clearpage
\subsubsection{\textit{Middle India}, \citeauthor{Middle_India}}\label{app:sec:sliding:Middle_India}
\vspace{-.2cm}
\begin{figure}[h]
  \centering
  \begin{minipage}[t]{0.53\textwidth}
    \centering
    \vspace{0cm}
    \includegraphics[width=\linewidth]{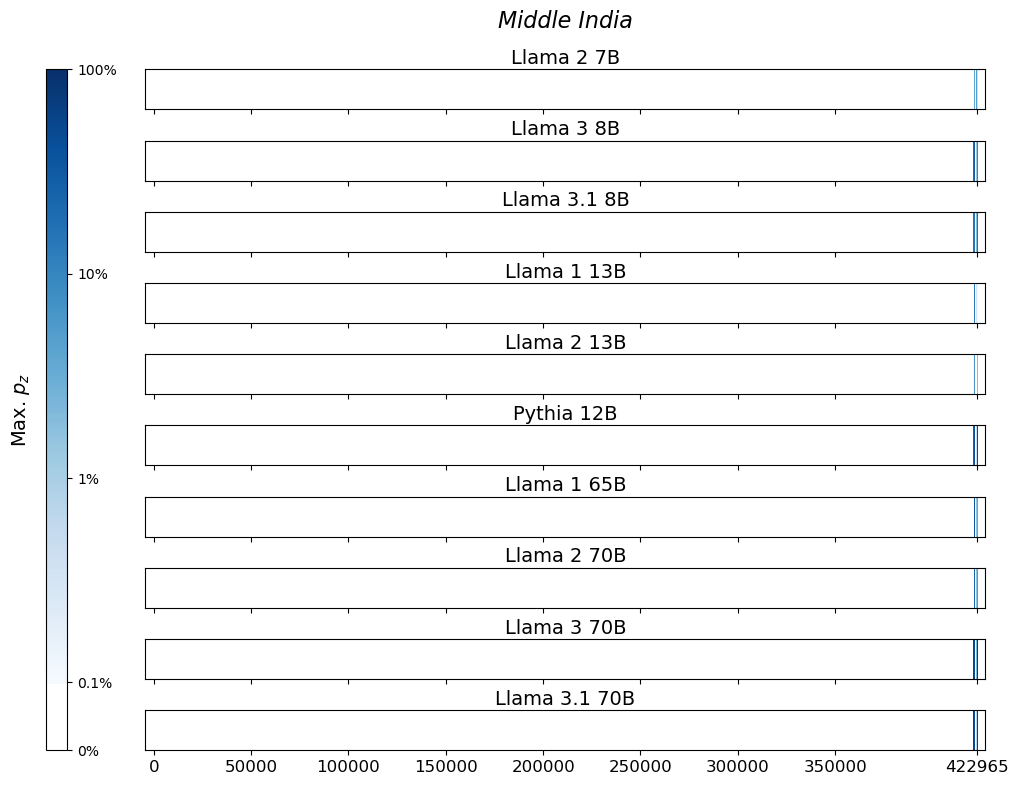}
    \includegraphics[width=\linewidth]{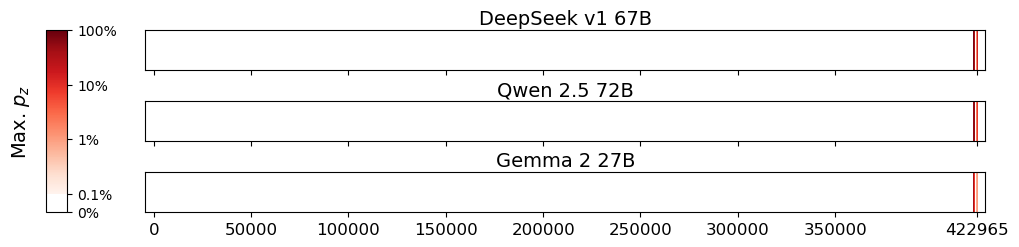}
    \includegraphics[width=\linewidth]{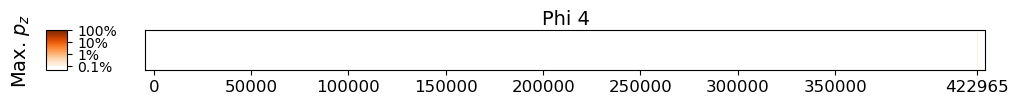}
  \end{minipage}
  \hfill
  \begin{minipage}[t]{0.45\textwidth}
    \centering
    \vspace{0cm}
    \includegraphics[width=\linewidth]{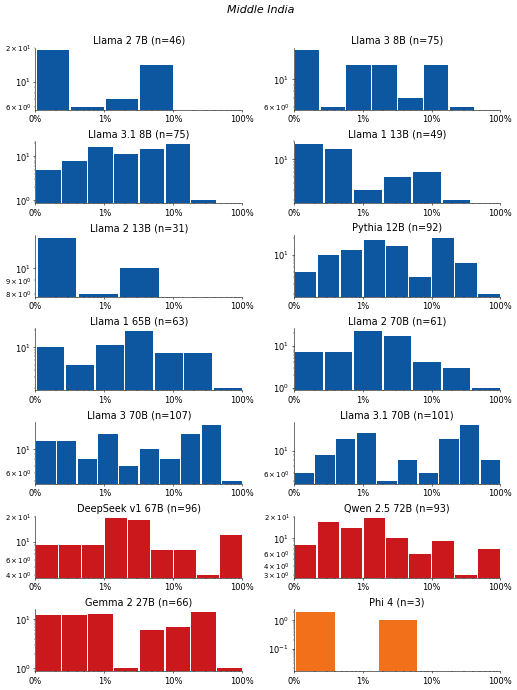}
  \end{minipage}
  \vspace{-.2cm}
  \caption{
    \textbf{\textit{Middle India}, \citeauthor{Middle_India}.}
    For $14$ LLMs,
    (\textbf{left}) heatmaps for the sliding-window procedure and
    (\textbf{right}) corresponding distributions over suffix extraction probabilities
    ($\tau_\text{min}=0.1\%$).
  }
  \label{fig:slidingwindow:Middle_India}
\end{figure}
\FloatBarrier

\subsubsection{\textit{The Catcher in the Rye}, \citeauthor{The_Catcher_in_the_Rye}}\label{app:sec:sliding:The_Catcher_in_the_Rye}
\vspace{-.2cm}
\begin{figure}[h]
  \centering
  \begin{minipage}[t]{0.53\textwidth}
    \centering
    \vspace{0cm}
    \includegraphics[width=\linewidth]{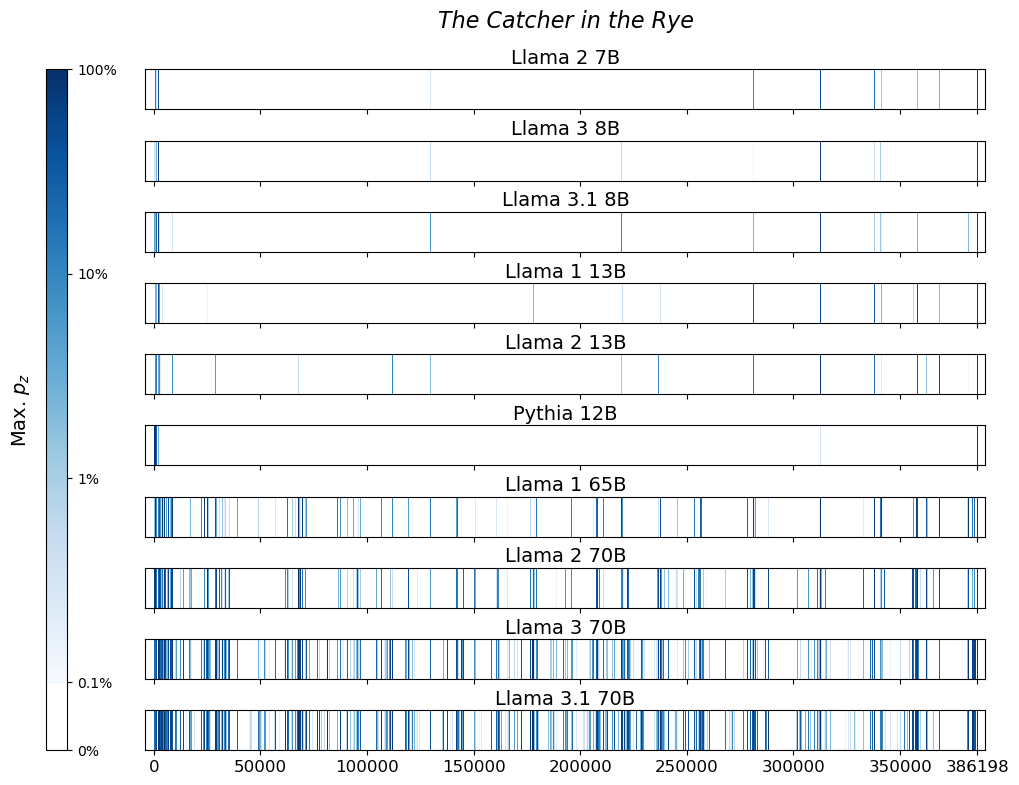}
    \includegraphics[width=\linewidth]{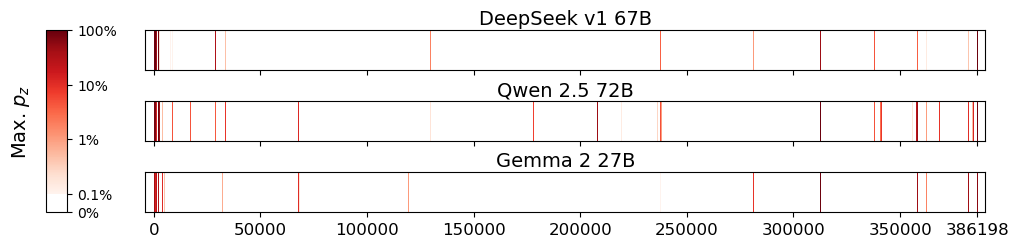}
    \includegraphics[width=\linewidth]{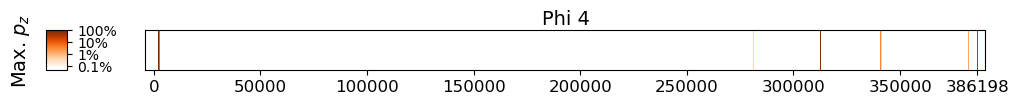}
  \end{minipage}
  \hfill
  \begin{minipage}[t]{0.45\textwidth}
    \centering
    \vspace{0cm}
    \includegraphics[width=\linewidth]{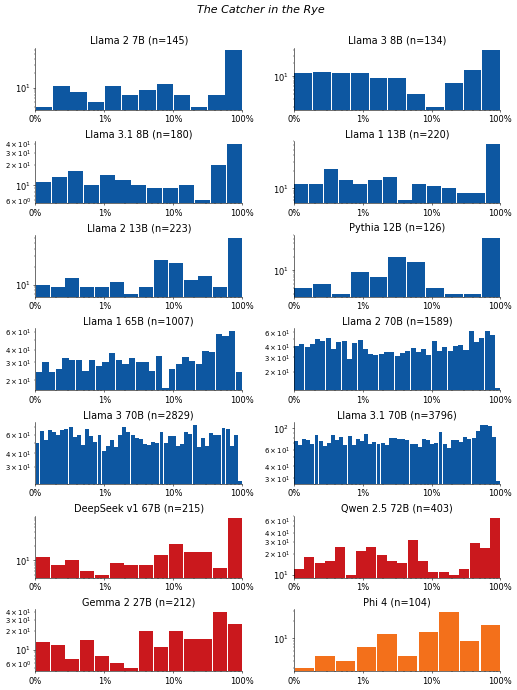}
  \end{minipage}
  \vspace{-.2cm}
  \caption{
    \textbf{\textit{The Catcher in the Rye}, \citeauthor{The_Catcher_in_the_Rye}.}
    For $14$ LLMs,
    (\textbf{left}) heatmaps for the sliding-window procedure and
    (\textbf{right}) corresponding distributions over suffix extraction probabilities
    ($\tau_\text{min}=0.1\%$).
  }
  \label{fig:slidingwindow:The_Catcher_in_the_Rye}
\end{figure}
\FloatBarrier

\clearpage
\subsubsection{\textit{Lean In}, \citeauthor{Lean_In}}\label{app:sec:sliding:Lean_In}
\vspace{-.2cm}
\begin{figure}[h]
  \centering
  \begin{minipage}[t]{0.53\textwidth}
    \centering
    \vspace{0cm}
    \includegraphics[width=\linewidth]{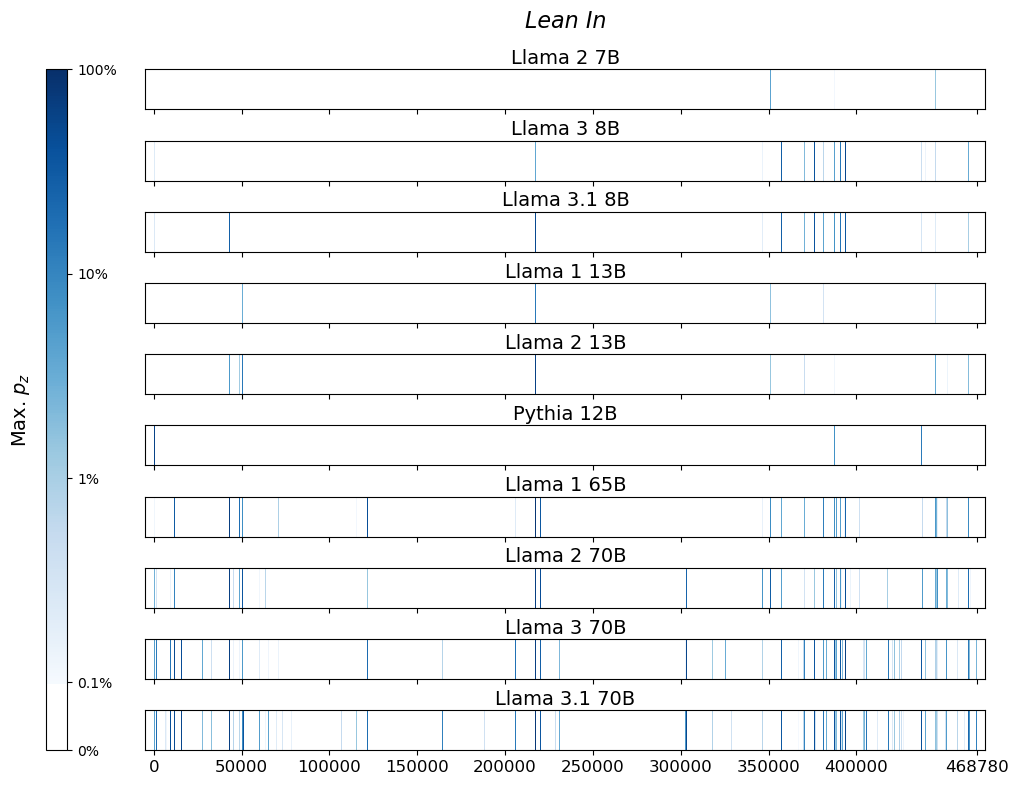}
    \includegraphics[width=\linewidth]{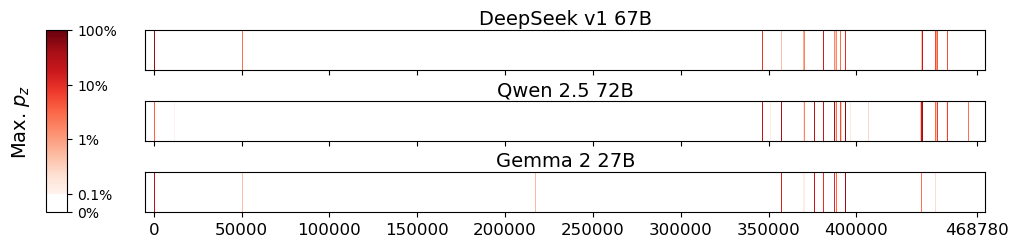}
    \includegraphics[width=\linewidth]{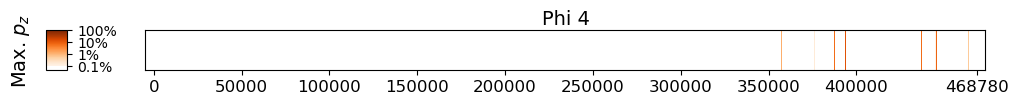}
  \end{minipage}
  \hfill
  \begin{minipage}[t]{0.45\textwidth}
    \centering
    \vspace{0cm}
    \includegraphics[width=\linewidth]{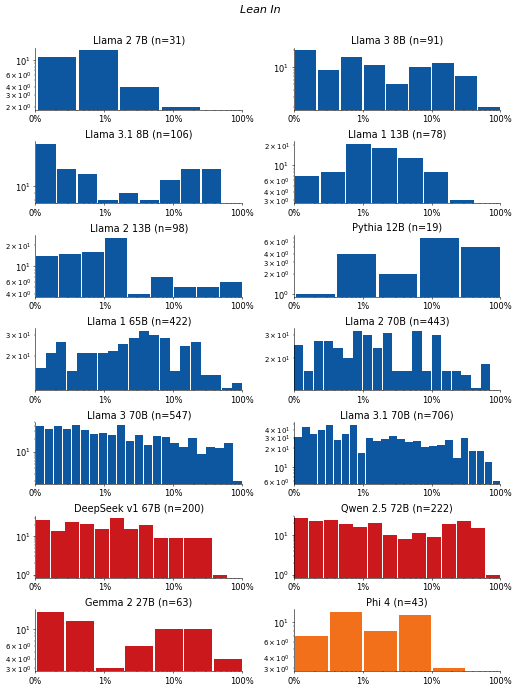}
  \end{minipage}
  \vspace{-.2cm}
  \caption{
    \textbf{\textit{Lean In}, \citeauthor{Lean_In}.}
    For $14$ LLMs,
    (\textbf{left}) heatmaps for the sliding-window procedure and
    (\textbf{right}) corresponding distributions over suffix extraction probabilities
    ($\tau_\text{min}=0.1\%$).
  }
  \label{fig:slidingwindow:Lean_In}
\end{figure}
\FloatBarrier

\subsubsection{\textit{The DevOps Adoption Playbook}, \citeauthor{The_DevOps_Adoption_Playbook}}\label{app:sec:sliding:The_DevOps_Adoption_Playbook}
\vspace{-.2cm}
\begin{figure}[h]
  \centering
  \begin{minipage}[t]{0.53\textwidth}
    \centering
    \vspace{0cm}
    \includegraphics[width=\linewidth]{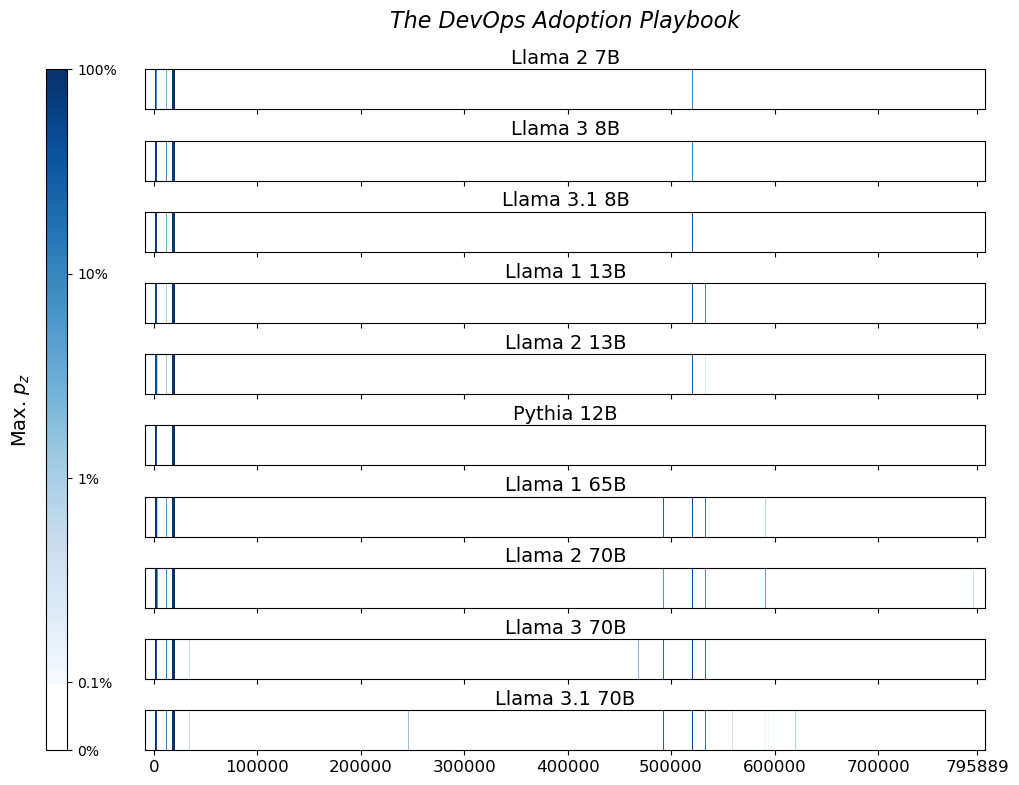}
    \includegraphics[width=\linewidth]{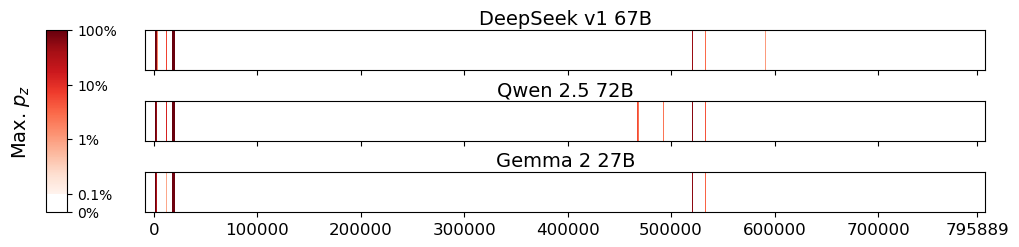}
    \includegraphics[width=\linewidth]{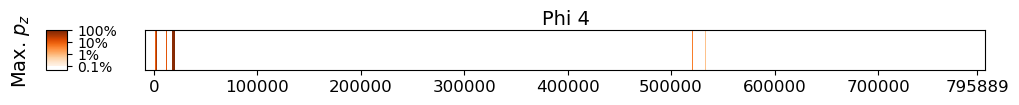}
  \end{minipage}
  \hfill
  \begin{minipage}[t]{0.45\textwidth}
    \centering
    \vspace{0cm}
    \includegraphics[width=\linewidth]{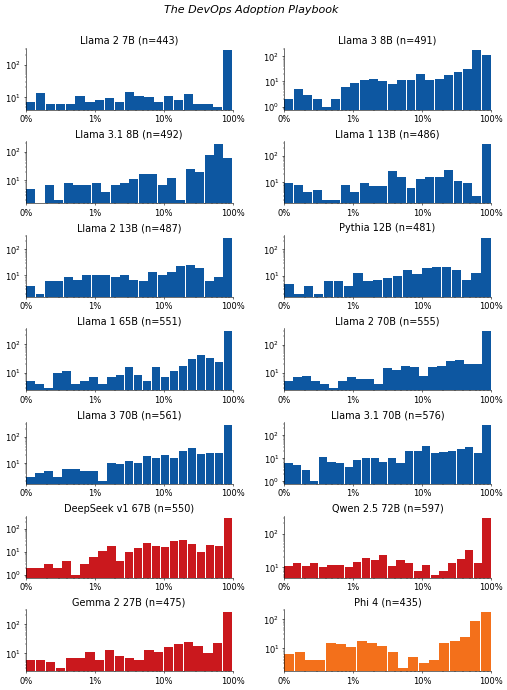}
  \end{minipage}
  \vspace{-.2cm}
  \caption{
    \textbf{\textit{The DevOps Adoption Playbook}, \citeauthor{The_DevOps_Adoption_Playbook}.}
    For $14$ LLMs,
    (\textbf{left}) heatmaps for the sliding-window procedure and
    (\textbf{right}) corresponding distributions over suffix extraction probabilities
    ($\tau_\text{min}=0.1\%$).
  }
  \label{fig:slidingwindow:The_DevOps_Adoption_Playbook}
\end{figure}
\FloatBarrier

\clearpage
\subsubsection{\textit{Frankenstein}, \citeauthor{Frankenstein}}\label{app:sec:sliding:Frankenstein}
\vspace{-.2cm}
\begin{figure}[h]
  \centering
  \begin{minipage}[t]{0.53\textwidth}
    \centering
    \vspace{0cm}
    \includegraphics[width=\linewidth]{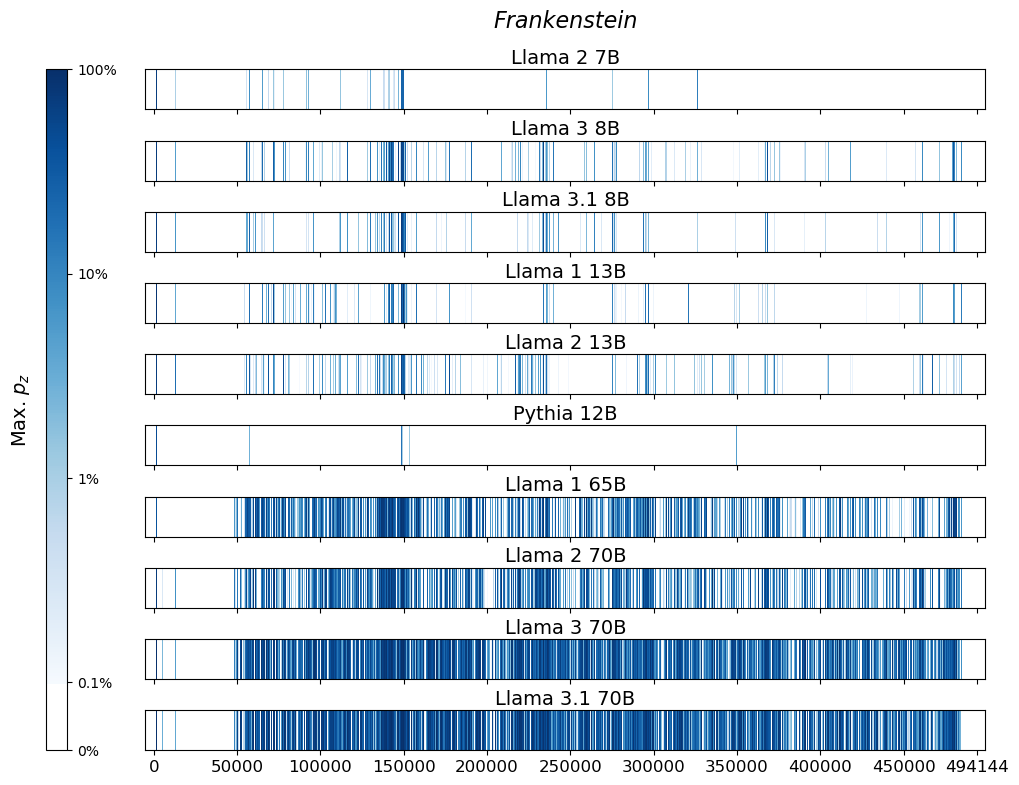}
    \includegraphics[width=\linewidth]{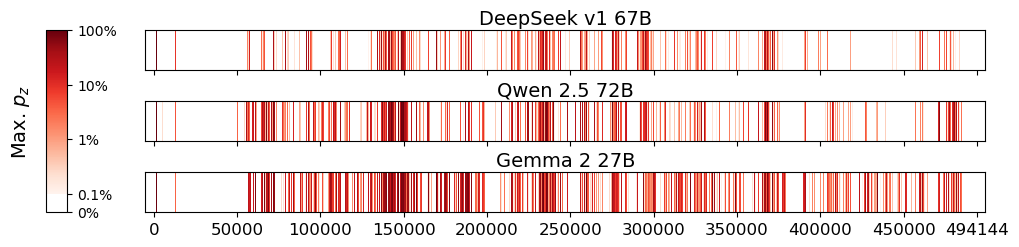}
    \includegraphics[width=\linewidth]{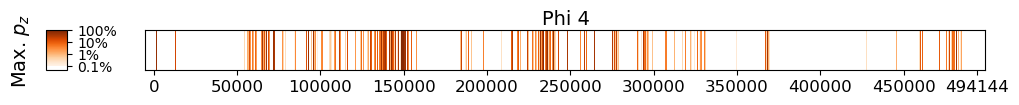}
  \end{minipage}
  \hfill
  \begin{minipage}[t]{0.45\textwidth}
    \centering
    \vspace{0cm}
    \includegraphics[width=\linewidth]{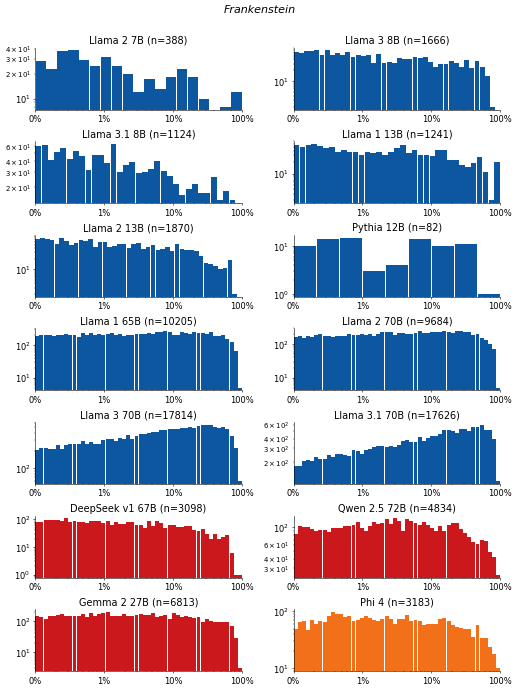}
  \end{minipage}
  \vspace{-.2cm}
  \caption{
    \textbf{\textit{Frankenstein}, \citeauthor{Frankenstein}.}
    For $14$ LLMs,
    (\textbf{left}) heatmaps for the sliding-window procedure and
    (\textbf{right}) corresponding distributions over suffix extraction probabilities
    ($\tau_\text{min}=0.1\%$).
  }
  \label{fig:slidingwindow:Frankenstein}
\end{figure}
\FloatBarrier

\subsubsection{\textit{Sally Ride: America's First Woman in Space}, \citeauthor{Sally_Ride_America_s_First_Woman_in_Space}}\label{app:sec:sliding:Sally_Ride_America_s_First_Woman_in_Space}
\vspace{-.2cm}
\begin{figure}[h]
  \centering
  \begin{minipage}[t]{0.53\textwidth}
    \centering
    \vspace{0cm}
    \includegraphics[width=\linewidth]{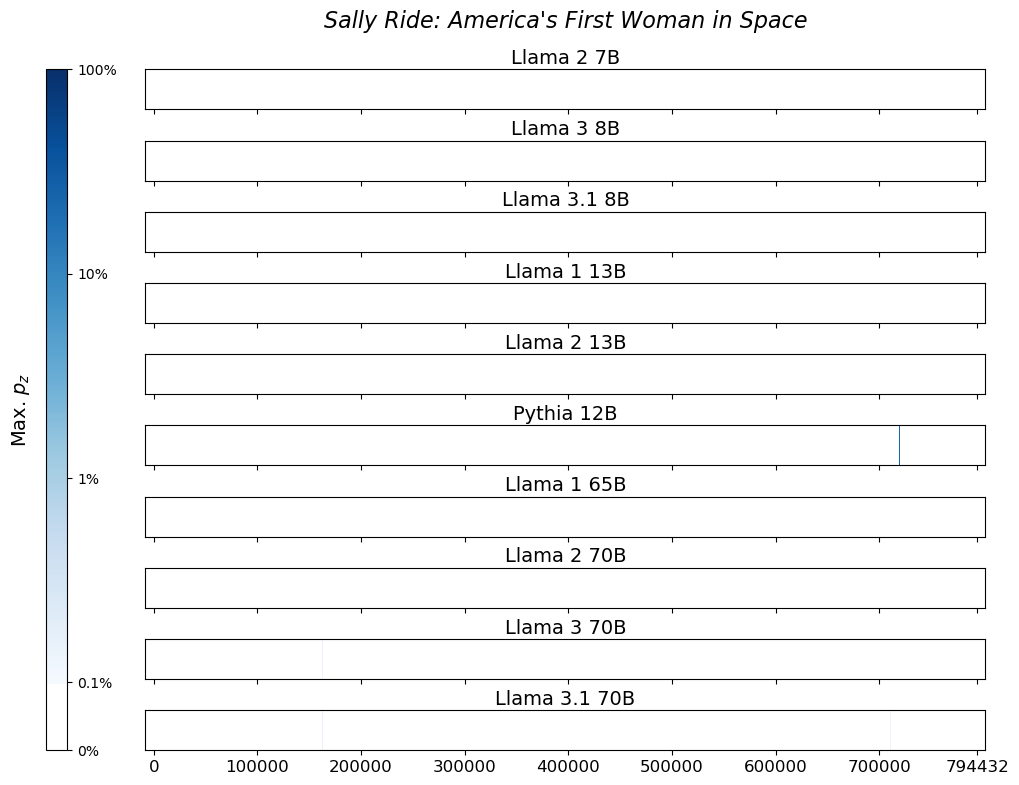}
    \includegraphics[width=\linewidth]{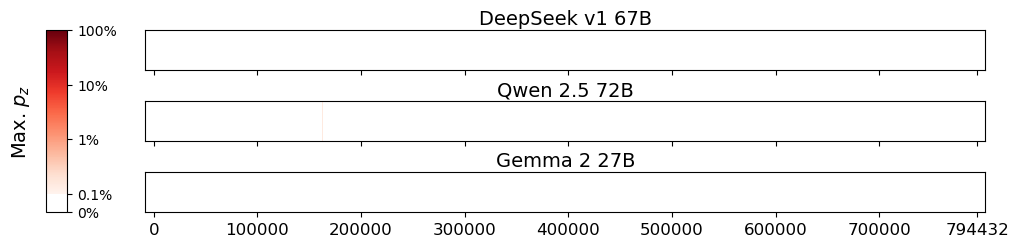}
    \includegraphics[width=\linewidth]{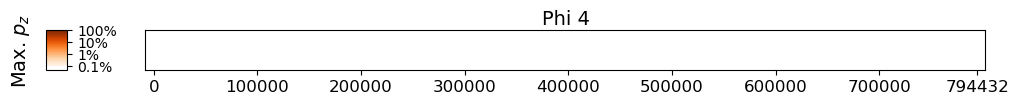}
  \end{minipage}
  \hfill
  \begin{minipage}[t]{0.45\textwidth}
    \centering
    \vspace{0cm}
    \includegraphics[width=\linewidth]{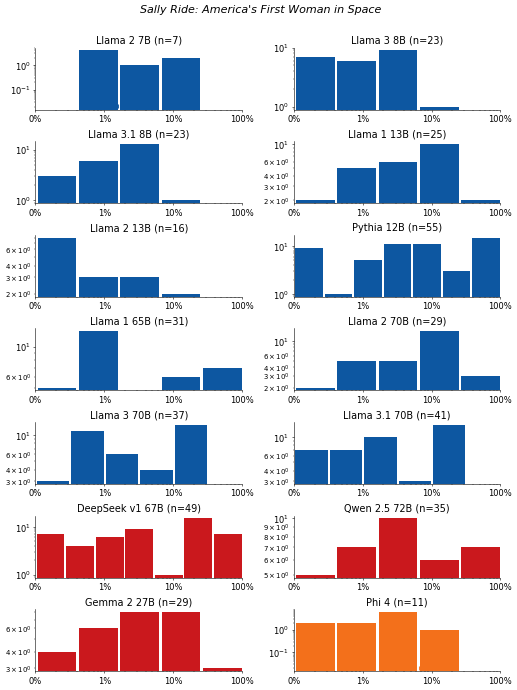}
  \end{minipage}
  \vspace{-.2cm}
  \caption{
    \textbf{\textit{Sally Ride: America's First Woman in Space}, \citeauthor{Sally_Ride_America_s_First_Woman_in_Space}.}
    For $14$ LLMs,
    (\textbf{left}) heatmaps for the sliding-window procedure and
    (\textbf{right}) corresponding distributions over suffix extraction probabilities
    ($\tau_\text{min}=0.1\%$).
  }
  \label{fig:slidingwindow:Sally_Ride_America_s_First_Woman_in_Space}
\end{figure}
\FloatBarrier

\clearpage
\subsubsection{\textit{A Perfectly Good Family}, \citeauthor{A_Perfectly_Good_Family}}\label{app:sec:sliding:A_Perfectly_Good_Family}
\vspace{-.2cm}
\begin{figure}[h]
  \centering
  \begin{minipage}[t]{0.53\textwidth}
    \centering
    \vspace{0cm}
    \includegraphics[width=\linewidth]{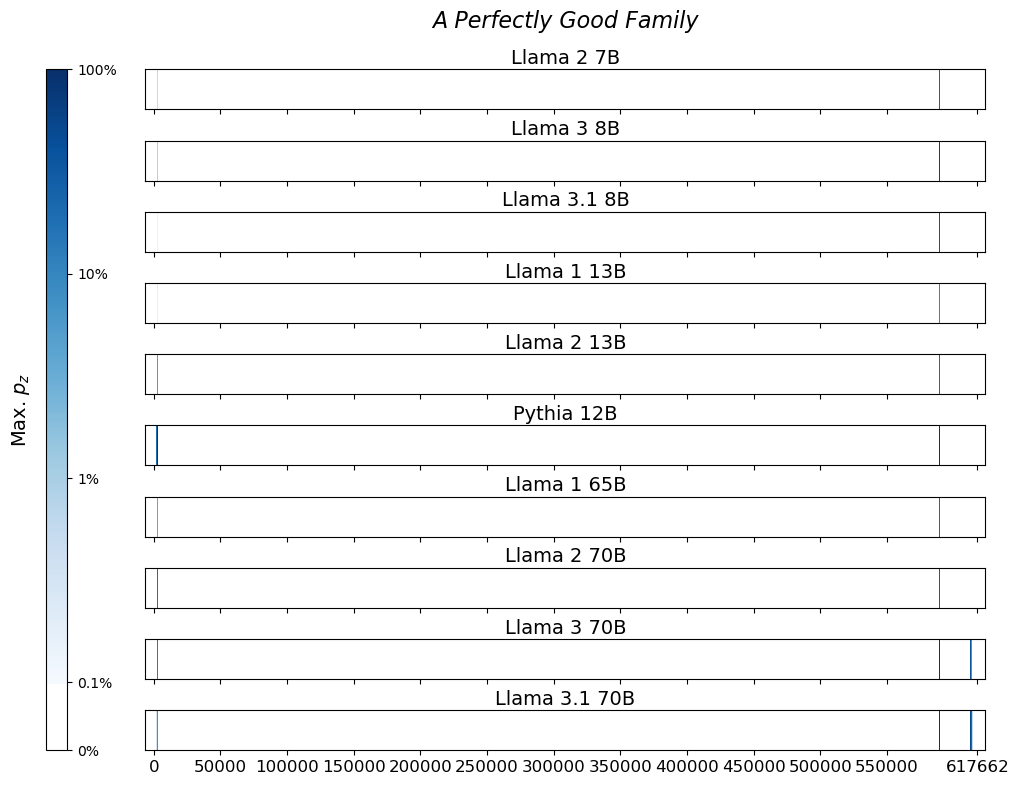}
    \includegraphics[width=\linewidth]{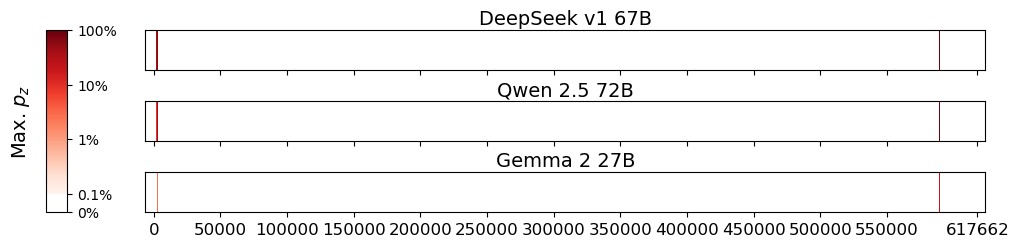}
    \includegraphics[width=\linewidth]{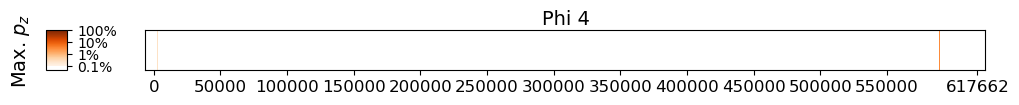}
  \end{minipage}
  \hfill
  \begin{minipage}[t]{0.45\textwidth}
    \centering
    \vspace{0cm}
    \includegraphics[width=\linewidth]{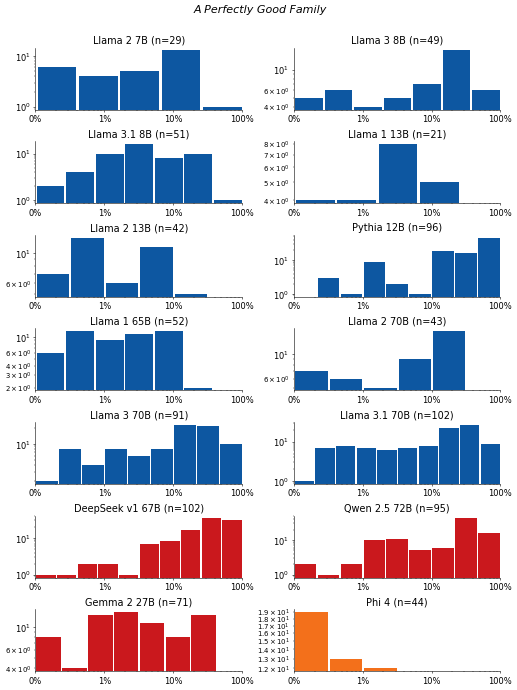}
  \end{minipage}
  \vspace{-.2cm}
  \caption{
    \textbf{\textit{A Perfectly Good Family}, \citeauthor{A_Perfectly_Good_Family}.}
    For $14$ LLMs,
    (\textbf{left}) heatmaps for the sliding-window procedure and
    (\textbf{right}) corresponding distributions over suffix extraction probabilities
    ($\tau_\text{min}=0.1\%$).
  }
  \label{fig:slidingwindow:A_Perfectly_Good_Family}
\end{figure}
\FloatBarrier

\subsubsection{\textit{The Bedwetter}, \citeauthor{The_Bedwetter}}\label{app:sec:sliding:The_Bedwetter}
\vspace{-.2cm}
\begin{figure}[h]
  \centering
  \begin{minipage}[t]{0.53\textwidth}
    \centering
    \vspace{0cm}
    \includegraphics[width=\linewidth]{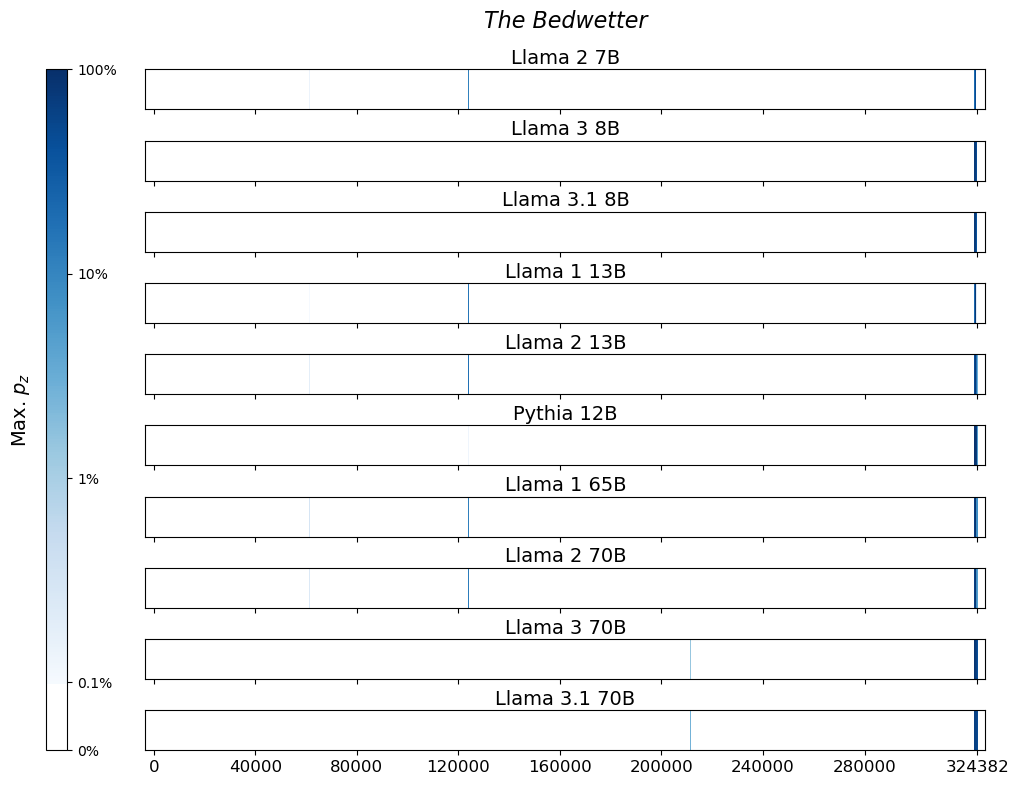}
    \includegraphics[width=\linewidth]{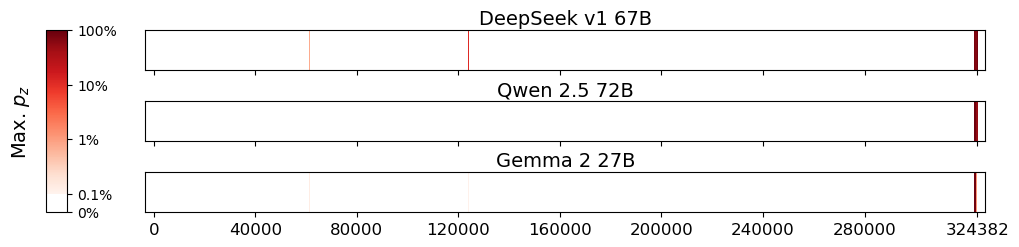}
    \includegraphics[width=\linewidth]{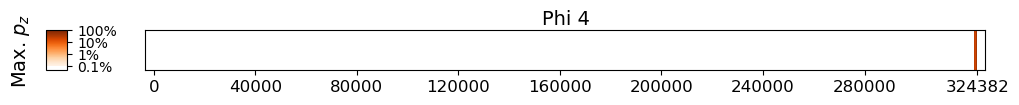}
  \end{minipage}
  \hfill
  \begin{minipage}[t]{0.45\textwidth}
    \centering
    \vspace{0cm}
    \includegraphics[width=\linewidth]{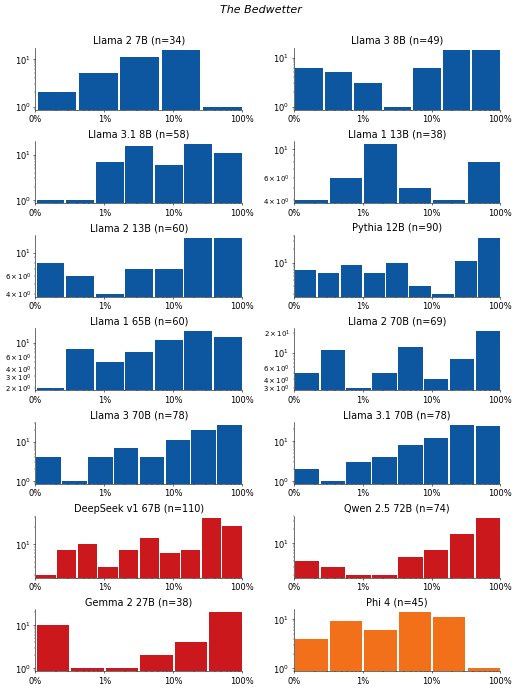}
  \end{minipage}
  \vspace{-.2cm}
  \caption{
    \textbf{\textit{The Bedwetter}, \citeauthor{The_Bedwetter}.}
    For $14$ LLMs,
    (\textbf{left}) heatmaps for the sliding-window procedure and
    (\textbf{right}) corresponding distributions over suffix extraction probabilities
    ($\tau_\text{min}=0.1\%$).
  }
  \label{fig:slidingwindow:The_Bedwetter}
\end{figure}
\FloatBarrier

\clearpage
\subsubsection{\textit{On the Road with Bob Dylan}, \citeauthor{On_the_Road_with_Bob_Dylan}}\label{app:sec:sliding:On_the_Road_with_Bob_Dylan}
\vspace{-.2cm}
\begin{figure}[h]
  \centering
  \begin{minipage}[t]{0.53\textwidth}
    \centering
    \vspace{0cm}
    \includegraphics[width=\linewidth]{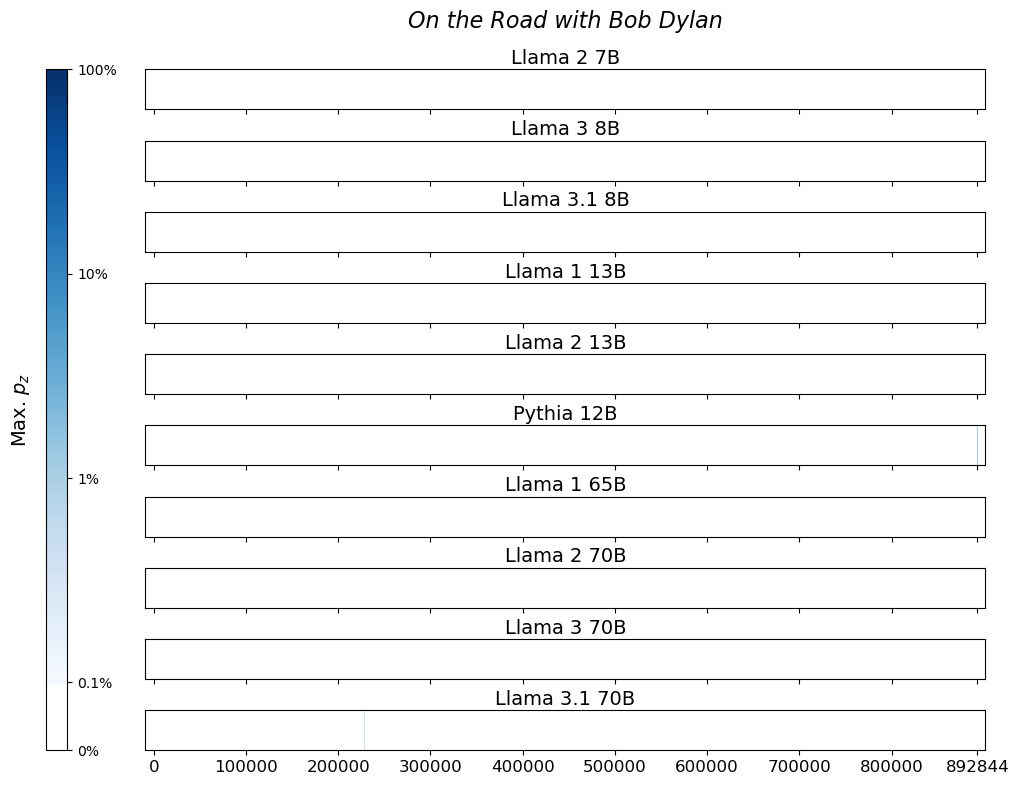}
    \includegraphics[width=\linewidth]{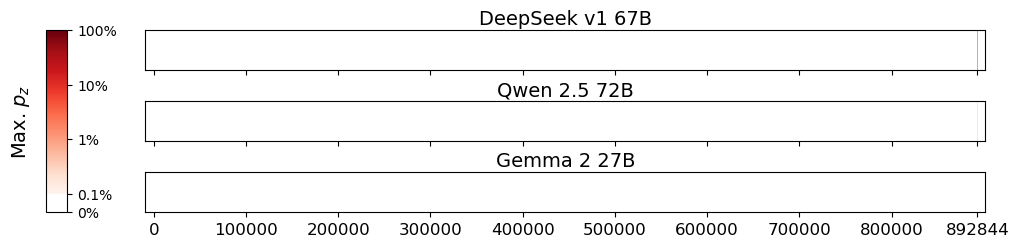}
    \includegraphics[width=\linewidth]{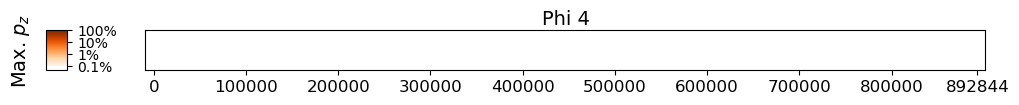}
  \end{minipage}
  \hfill
  \begin{minipage}[t]{0.45\textwidth}
    \centering
    \vspace{0cm}
    \includegraphics[width=\linewidth]{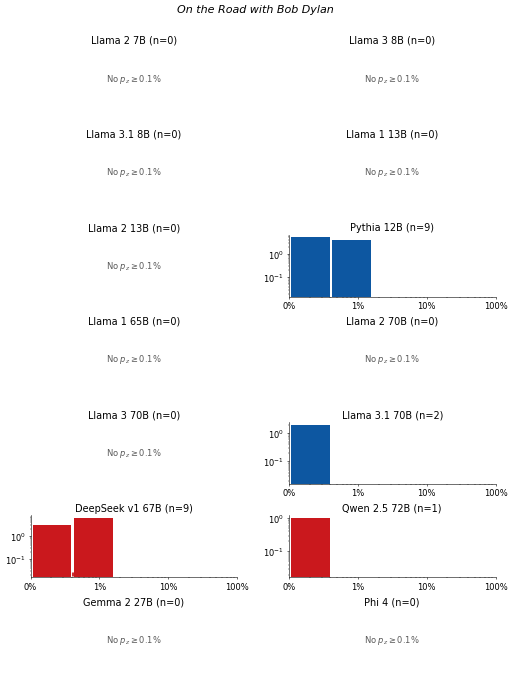}
  \end{minipage}
  \vspace{-.2cm}
  \caption{
    \textbf{\textit{On the Road with Bob Dylan}, \citeauthor{On_the_Road_with_Bob_Dylan}.}
    For $14$ LLMs,
    (\textbf{left}) heatmaps for the sliding-window procedure and
    (\textbf{right}) corresponding distributions over suffix extraction probabilities
    ($\tau_\text{min}=0.1\%$).
  }
  \label{fig:slidingwindow:On_the_Road_with_Bob_Dylan}
\end{figure}
\FloatBarrier

\subsubsection{\textit{The Night Children}, \citeauthor{The_Night_Children}}\label{app:sec:sliding:The_Night_Children}
\vspace{-.2cm}
\begin{figure}[h]
  \centering
  \begin{minipage}[t]{0.53\textwidth}
    \centering
    \vspace{0cm}
    \includegraphics[width=\linewidth]{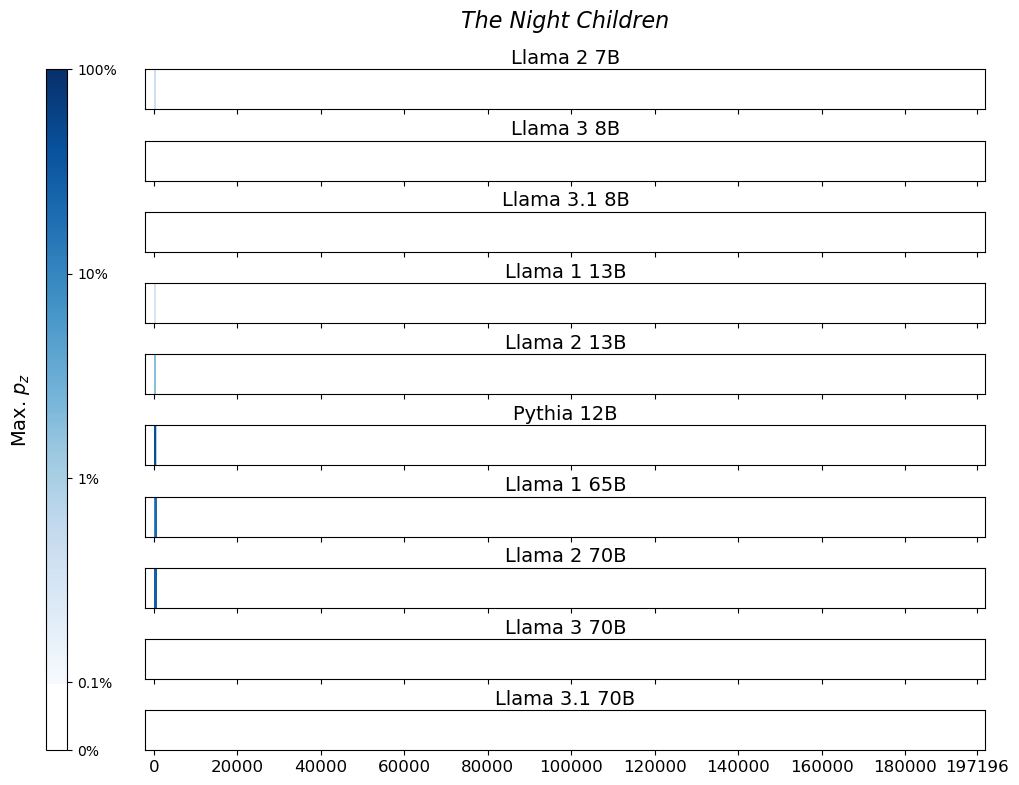}
    \includegraphics[width=\linewidth]{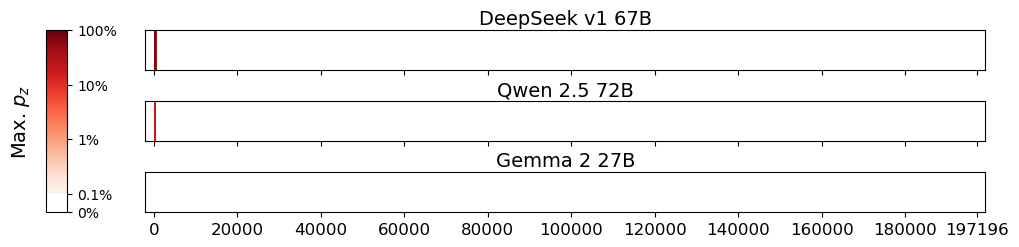}
    \includegraphics[width=\linewidth]{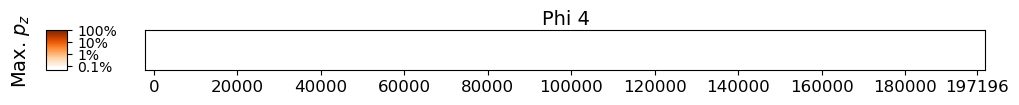}
  \end{minipage}
  \hfill
  \begin{minipage}[t]{0.45\textwidth}
    \centering
    \vspace{0cm}
    \includegraphics[width=\linewidth]{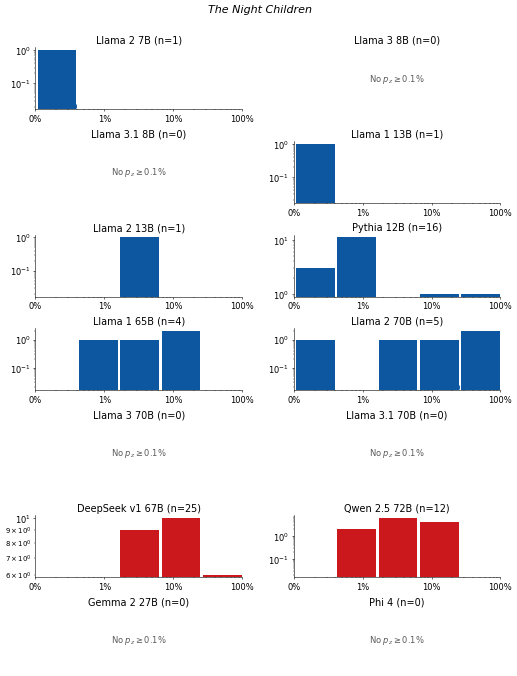}
  \end{minipage}
  \vspace{-.2cm}
  \caption{
    \textbf{\textit{The Night Children}, \citeauthor{The_Night_Children}.}
    For $14$ LLMs,
    (\textbf{left}) heatmaps for the sliding-window procedure and
    (\textbf{right}) corresponding distributions over suffix extraction probabilities
    ($\tau_\text{min}=0.1\%$).
  }
  \label{fig:slidingwindow:The_Night_Children}
\end{figure}
\FloatBarrier

\clearpage
\subsubsection{\textit{White Teeth}, \citeauthor{White_Teeth}}\label{app:sec:sliding:White_Teeth}
\vspace{-.2cm}
\begin{figure}[h]
  \centering
  \begin{minipage}[t]{0.53\textwidth}
    \centering
    \vspace{0cm}
    \includegraphics[width=\linewidth]{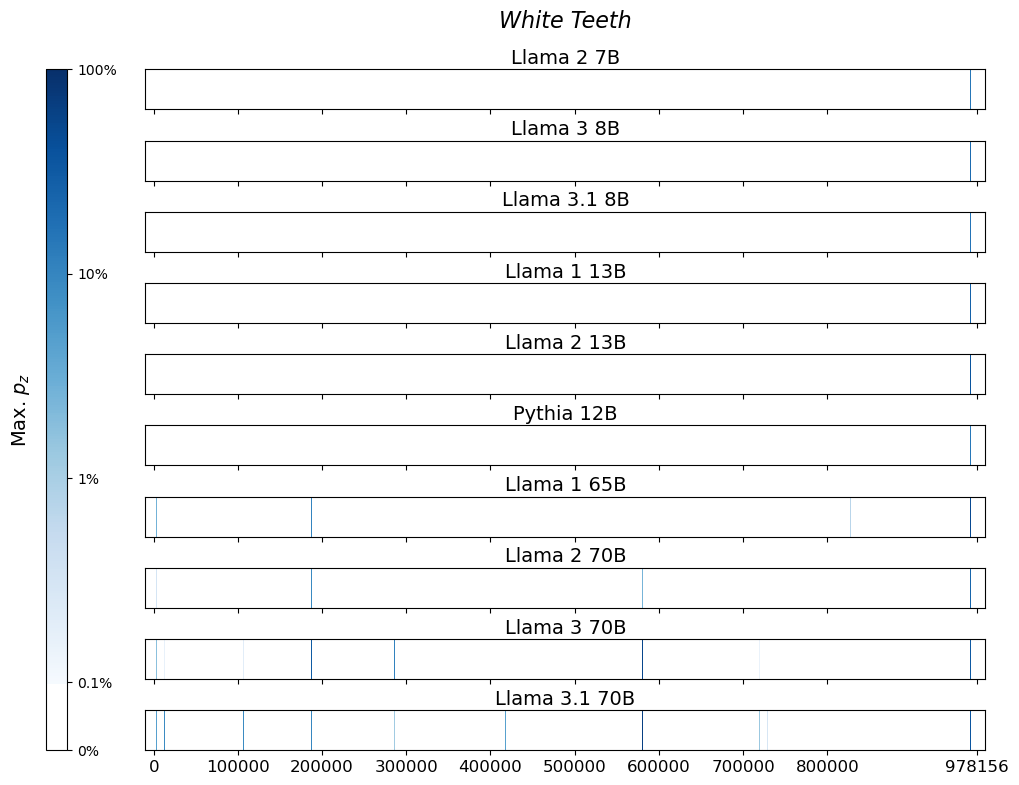}
    \includegraphics[width=\linewidth]{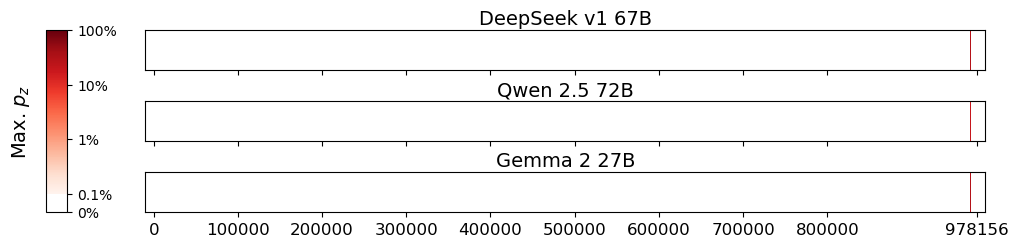}
    \includegraphics[width=\linewidth]{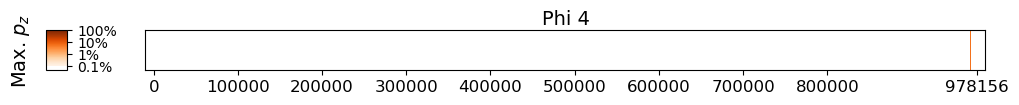}
  \end{minipage}
  \hfill
  \begin{minipage}[t]{0.45\textwidth}
    \centering
    \vspace{0cm}
    \includegraphics[width=\linewidth]{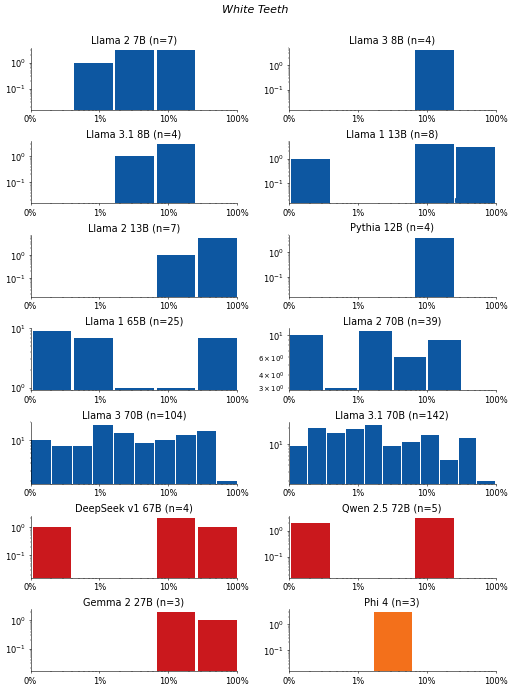}
  \end{minipage}
  \vspace{-.2cm}
  \caption{
    \textbf{\textit{White Teeth}, \citeauthor{White_Teeth}.}
    For $14$ LLMs,
    (\textbf{left}) heatmaps for the sliding-window procedure and
    (\textbf{right}) corresponding distributions over suffix extraction probabilities
    ($\tau_\text{min}=0.1\%$).
  }
  \label{fig:slidingwindow:White_Teeth}
\end{figure}
\FloatBarrier

\subsubsection{\textit{No Visible Bruises}, \citeauthor{No_Visible_Bruises}}\label{app:sec:sliding:No_Visible_Bruises}
\vspace{-.2cm}
\begin{figure}[h]
  \centering
  \begin{minipage}[t]{0.53\textwidth}
    \centering
    \vspace{0cm}
    \includegraphics[width=\linewidth]{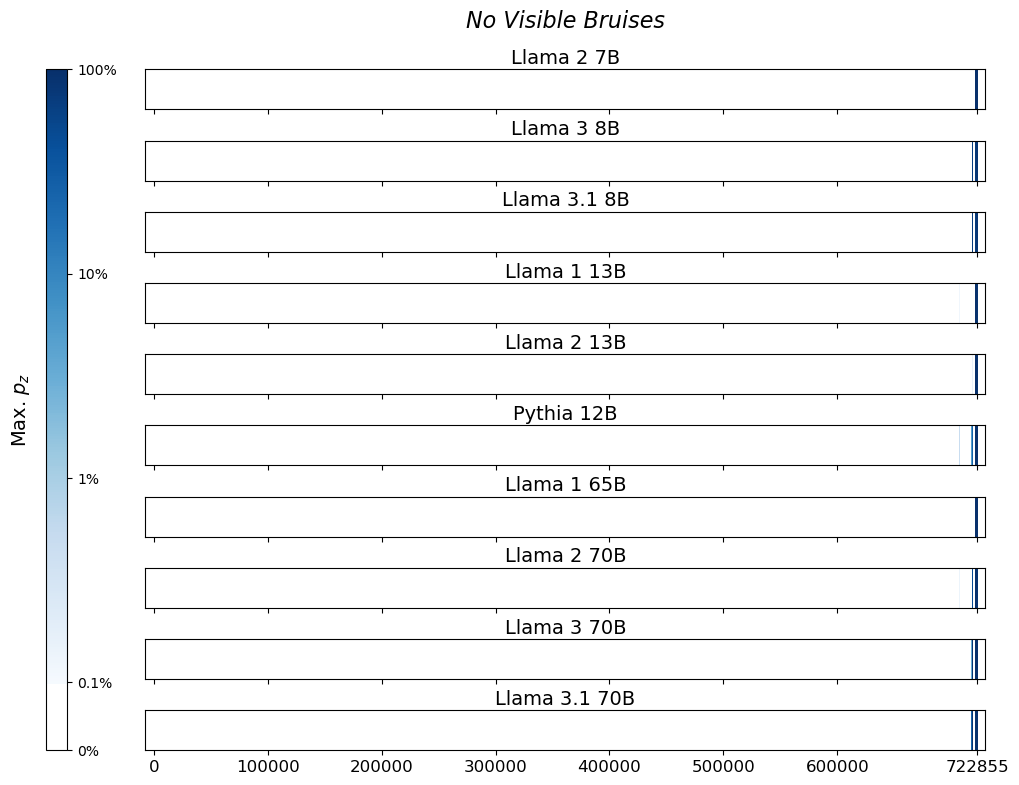}
    \includegraphics[width=\linewidth]{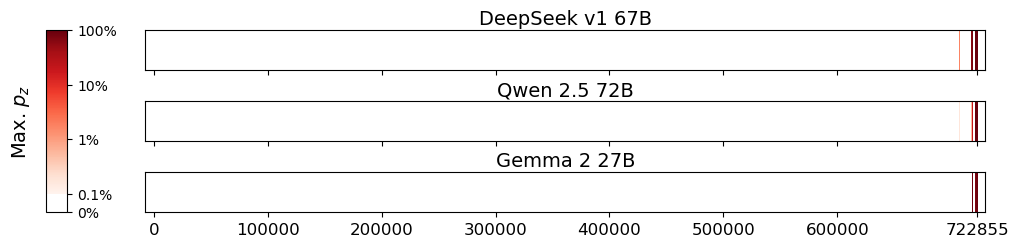}
    \includegraphics[width=\linewidth]{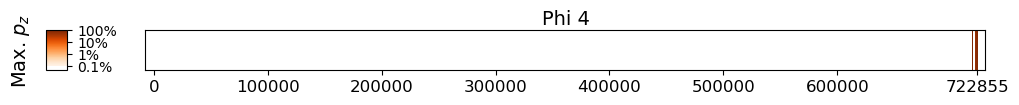}
  \end{minipage}
  \hfill
  \begin{minipage}[t]{0.45\textwidth}
    \centering
    \vspace{0cm}
    \includegraphics[width=\linewidth]{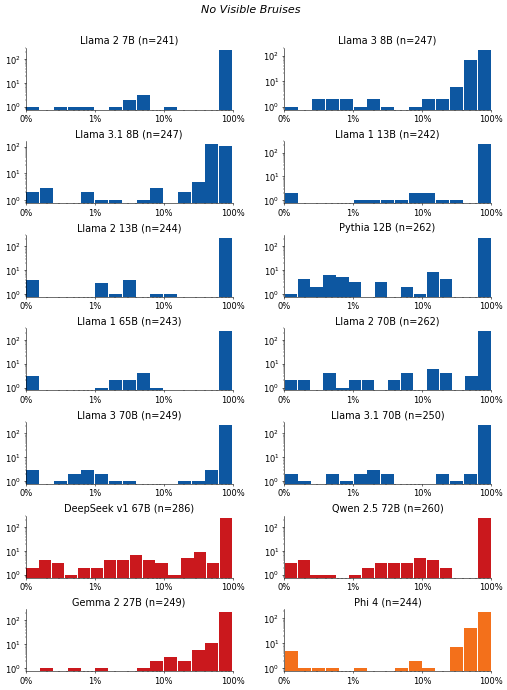}
  \end{minipage}
  \vspace{-.2cm}
  \caption{
    \textbf{\textit{No Visible Bruises}, \citeauthor{No_Visible_Bruises}.}
    For $14$ LLMs,
    (\textbf{left}) heatmaps for the sliding-window procedure and
    (\textbf{right}) corresponding distributions over suffix extraction probabilities
    ($\tau_\text{min}=0.1\%$).
  }
  \label{fig:slidingwindow:No_Visible_Bruises}
\end{figure}
\FloatBarrier

\clearpage
\subsubsection{\textit{The Grapes of Wrath}, \citeauthor{The_Grapes_of_Wrath}}\label{app:sec:sliding:The_Grapes_of_Wrath}
\vspace{-.2cm}
\begin{figure}[h]
  \centering
  \begin{minipage}[t]{0.53\textwidth}
    \centering
    \vspace{0cm}
    \includegraphics[width=\linewidth]{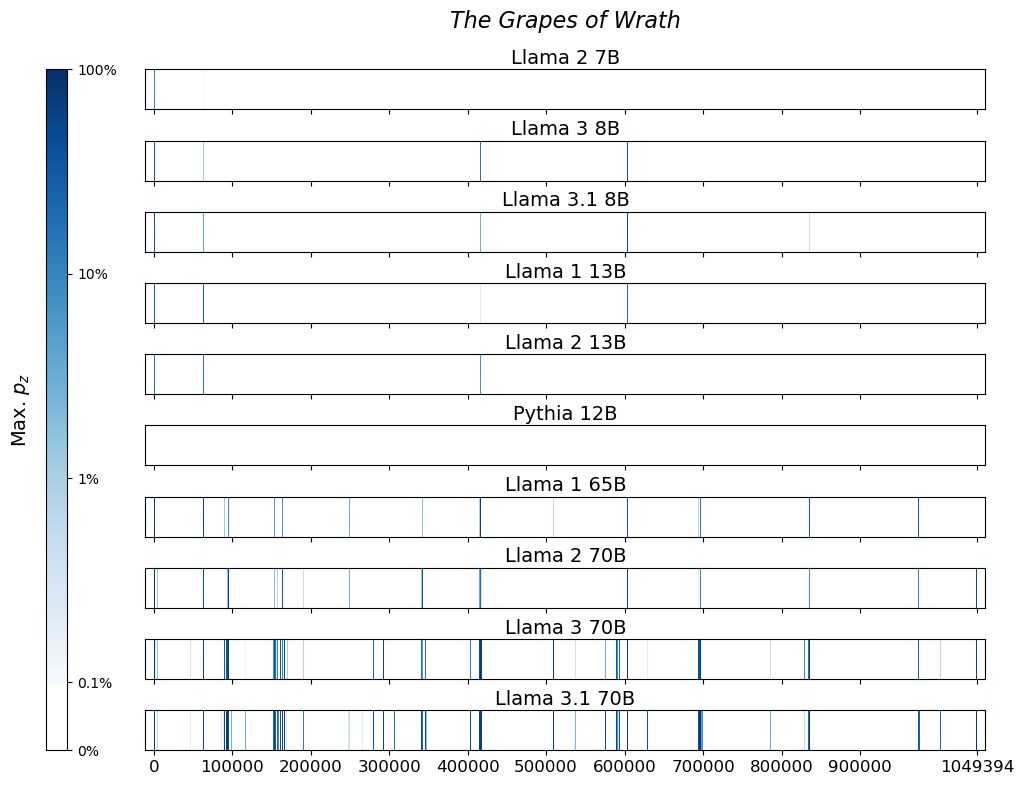}
    \includegraphics[width=\linewidth]{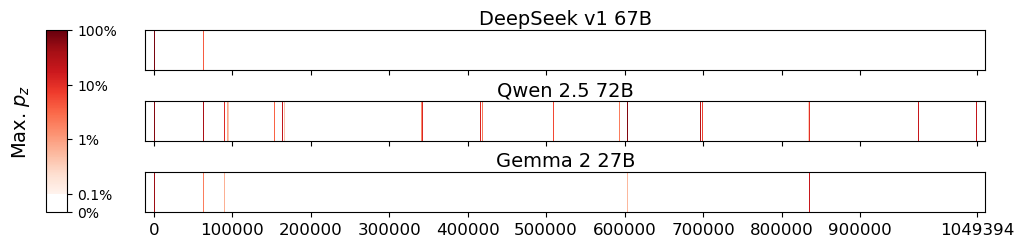}
    \includegraphics[width=\linewidth]{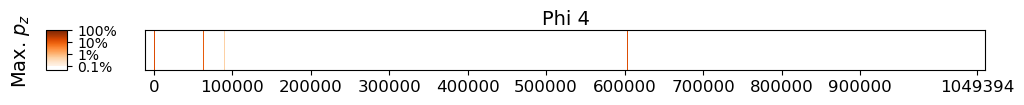}
  \end{minipage}
  \hfill
  \begin{minipage}[t]{0.45\textwidth}
    \centering
    \vspace{0cm}
    \includegraphics[width=\linewidth]{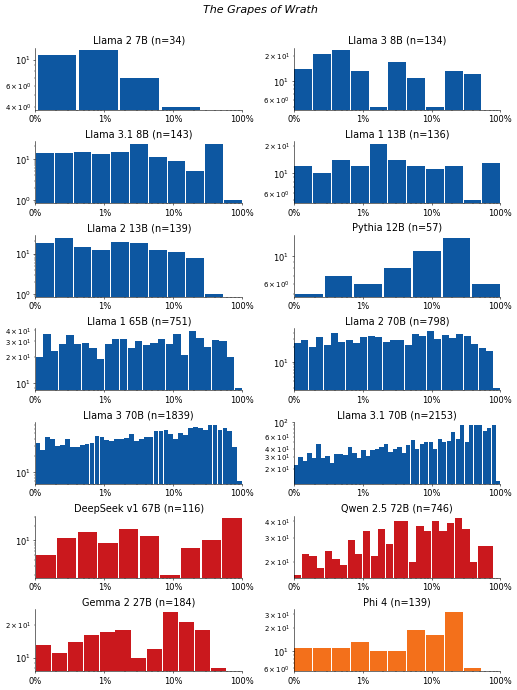}
  \end{minipage}
  \vspace{-.2cm}
  \caption{
    \textbf{\textit{The Grapes of Wrath}, \citeauthor{The_Grapes_of_Wrath}.}
    For $14$ LLMs,
    (\textbf{left}) heatmaps for the sliding-window procedure and
    (\textbf{right}) corresponding distributions over suffix extraction probabilities
    ($\tau_\text{min}=0.1\%$).
  }
  \label{fig:slidingwindow:The_Grapes_of_Wrath}
\end{figure}
\FloatBarrier

\subsubsection{\textit{Jesse James}, \citeauthor{Jesse_James}}\label{app:sec:sliding:Jesse_James}
\vspace{-.2cm}
\begin{figure}[h]
  \centering
  \begin{minipage}[t]{0.53\textwidth}
    \centering
    \vspace{0cm}
    \includegraphics[width=\linewidth]{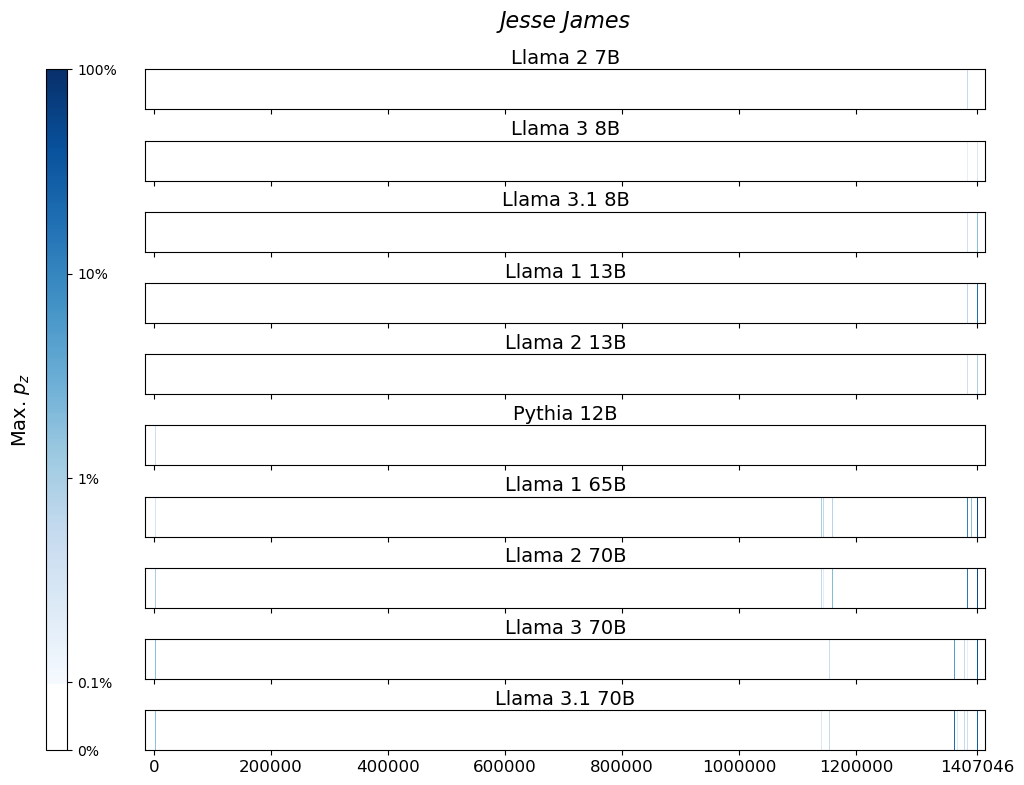}
    \includegraphics[width=\linewidth]{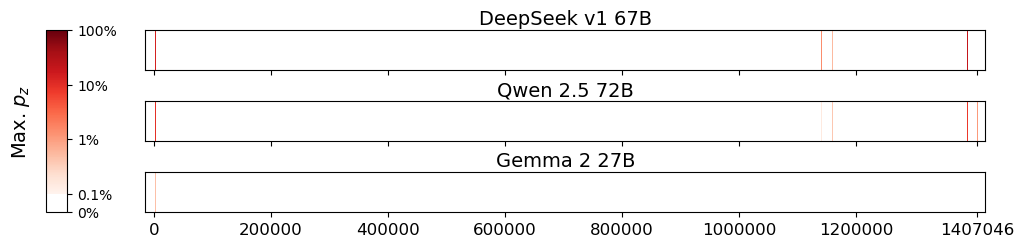}
    \includegraphics[width=\linewidth]{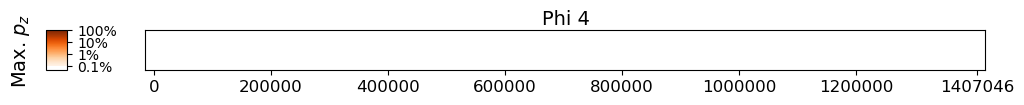}
  \end{minipage}
  \hfill
  \begin{minipage}[t]{0.45\textwidth}
    \centering
    \vspace{0cm}
    \includegraphics[width=\linewidth]{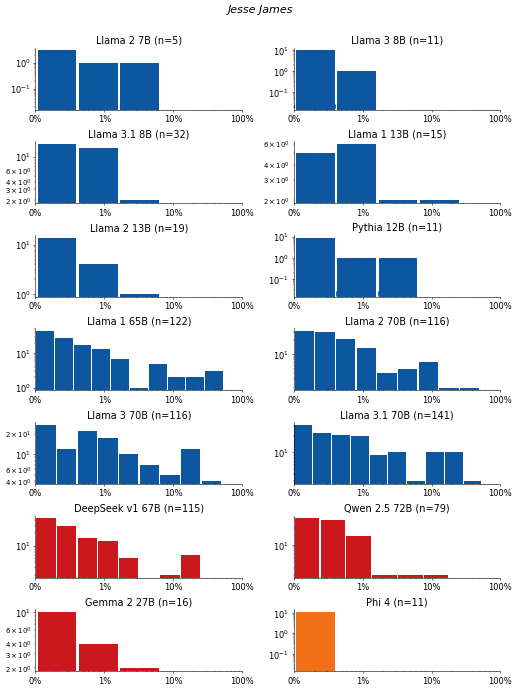}
  \end{minipage}
  \vspace{-.2cm}
  \caption{
    \textbf{\textit{Jesse James}, \citeauthor{Jesse_James}.}
    For $14$ LLMs,
    (\textbf{left}) heatmaps for the sliding-window procedure and
    (\textbf{right}) corresponding distributions over suffix extraction probabilities
    ($\tau_\text{min}=0.1\%$).
  }
  \label{fig:slidingwindow:Jesse_James}
\end{figure}
\FloatBarrier

\clearpage
\subsubsection{\textit{Zombie Halloween}, \citeauthor{Zombie_Halloween}}\label{app:sec:sliding:Zombie_Halloween}
\vspace{-.2cm}
\begin{figure}[h]
  \centering
  \begin{minipage}[t]{0.53\textwidth}
    \centering
    \vspace{0cm}
    \includegraphics[width=\linewidth]{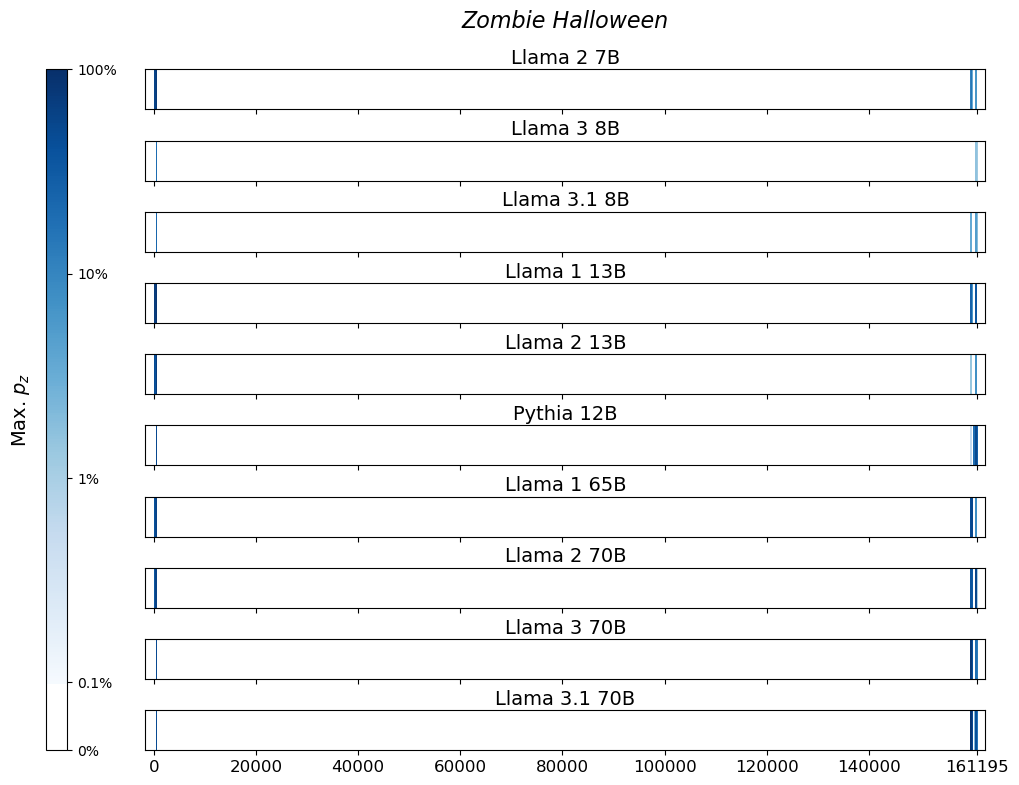}
    \includegraphics[width=\linewidth]{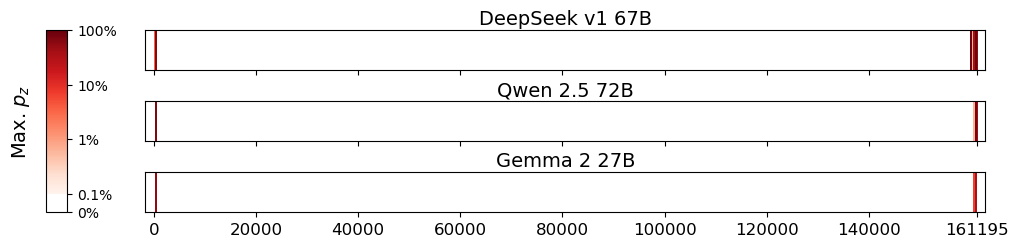}
    \includegraphics[width=\linewidth]{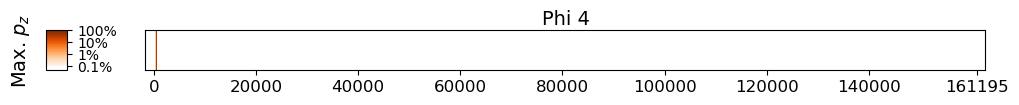}
  \end{minipage}
  \hfill
  \begin{minipage}[t]{0.45\textwidth}
    \centering
    \vspace{0cm}
    \includegraphics[width=\linewidth]{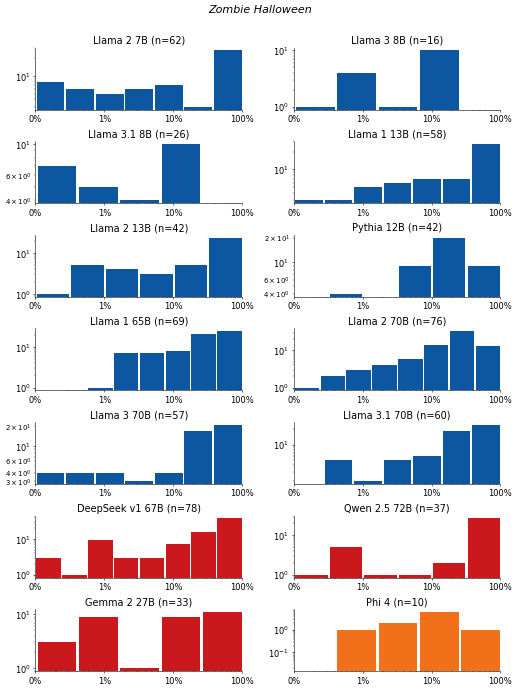}
  \end{minipage}
  \vspace{-.2cm}
  \caption{
    \textbf{\textit{Zombie Halloween}, \citeauthor{Zombie_Halloween}.}
    For $14$ LLMs,
    (\textbf{left}) heatmaps for the sliding-window procedure and
    (\textbf{right}) corresponding distributions over suffix extraction probabilities
    ($\tau_\text{min}=0.1\%$).
  }
  \label{fig:slidingwindow:Zombie_Halloween}
\end{figure}
\FloatBarrier

\subsubsection{\textit{Fear of Music}, \citeauthor{Fear_of_Music}}\label{app:sec:sliding:Fear_of_Music}
\vspace{-.2cm}
\begin{figure}[h]
  \centering
  \begin{minipage}[t]{0.53\textwidth}
    \centering
    \vspace{0cm}
    \includegraphics[width=\linewidth]{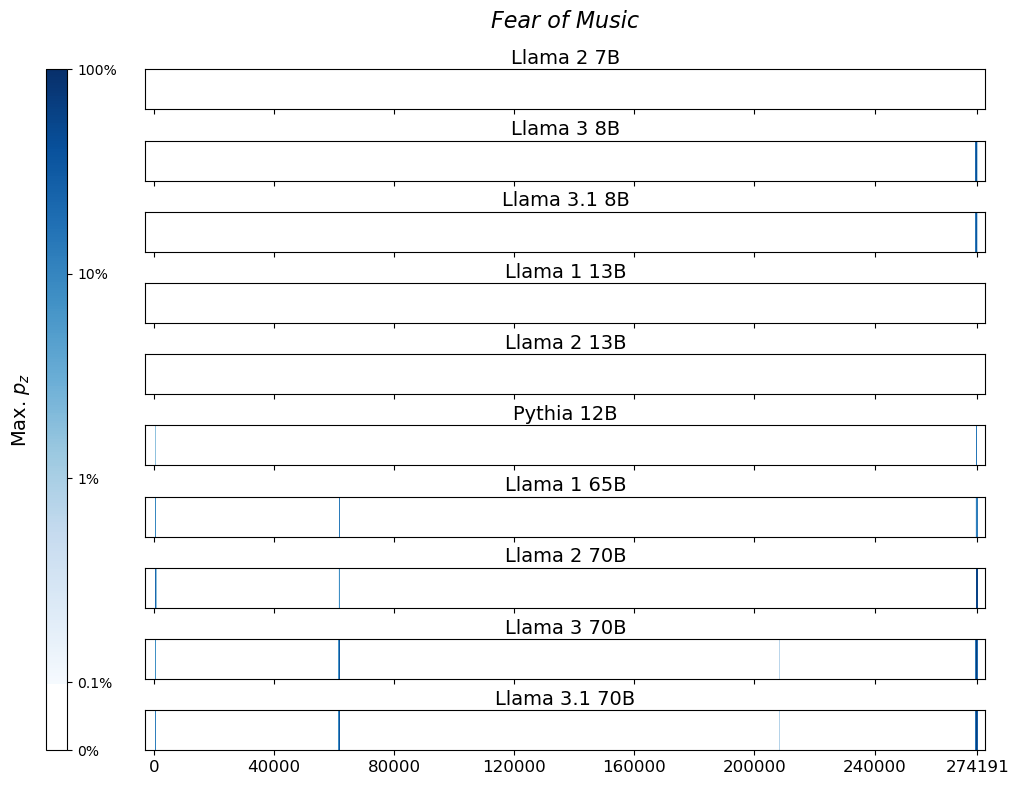}
    \includegraphics[width=\linewidth]{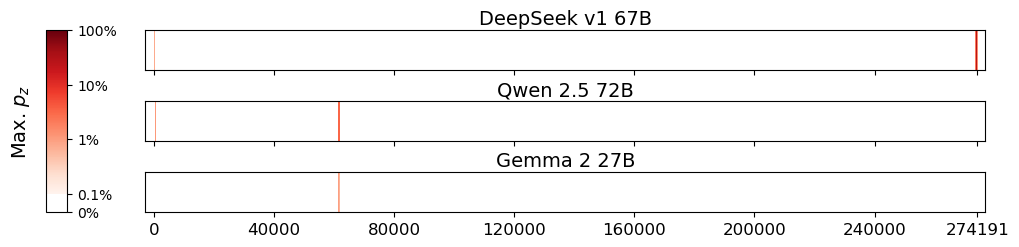}
    \includegraphics[width=\linewidth]{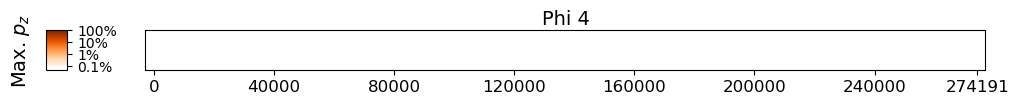}
  \end{minipage}
  \hfill
  \begin{minipage}[t]{0.45\textwidth}
    \centering
    \vspace{0cm}
    \includegraphics[width=\linewidth]{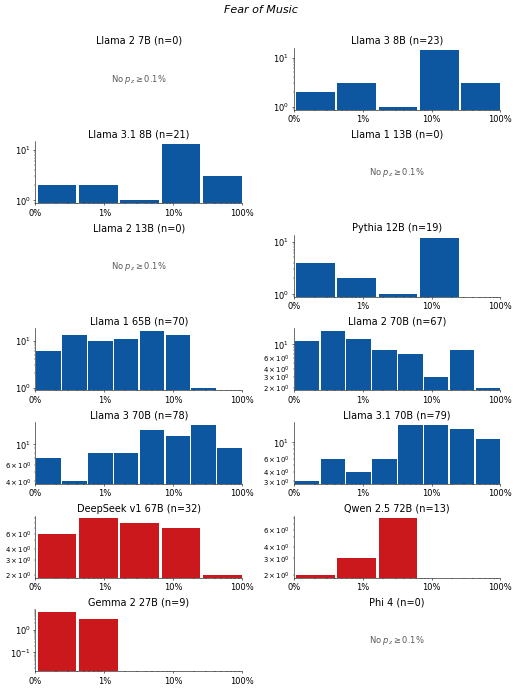}
  \end{minipage}
  \vspace{-.2cm}
  \caption{
    \textbf{\textit{Fear of Music}, \citeauthor{Fear_of_Music}.}
    For $14$ LLMs,
    (\textbf{left}) heatmaps for the sliding-window procedure and
    (\textbf{right}) corresponding distributions over suffix extraction probabilities
    ($\tau_\text{min}=0.1\%$).
  }
  \label{fig:slidingwindow:Fear_of_Music}
\end{figure}
\FloatBarrier

\clearpage
\subsubsection{\textit{Pearl Harbor}, \citeauthor{Pearl_Harbor}}\label{app:sec:sliding:Pearl_Harbor}
\vspace{-.2cm}
\begin{figure}[h]
  \centering
  \begin{minipage}[t]{0.53\textwidth}
    \centering
    \vspace{0cm}
    \includegraphics[width=\linewidth]{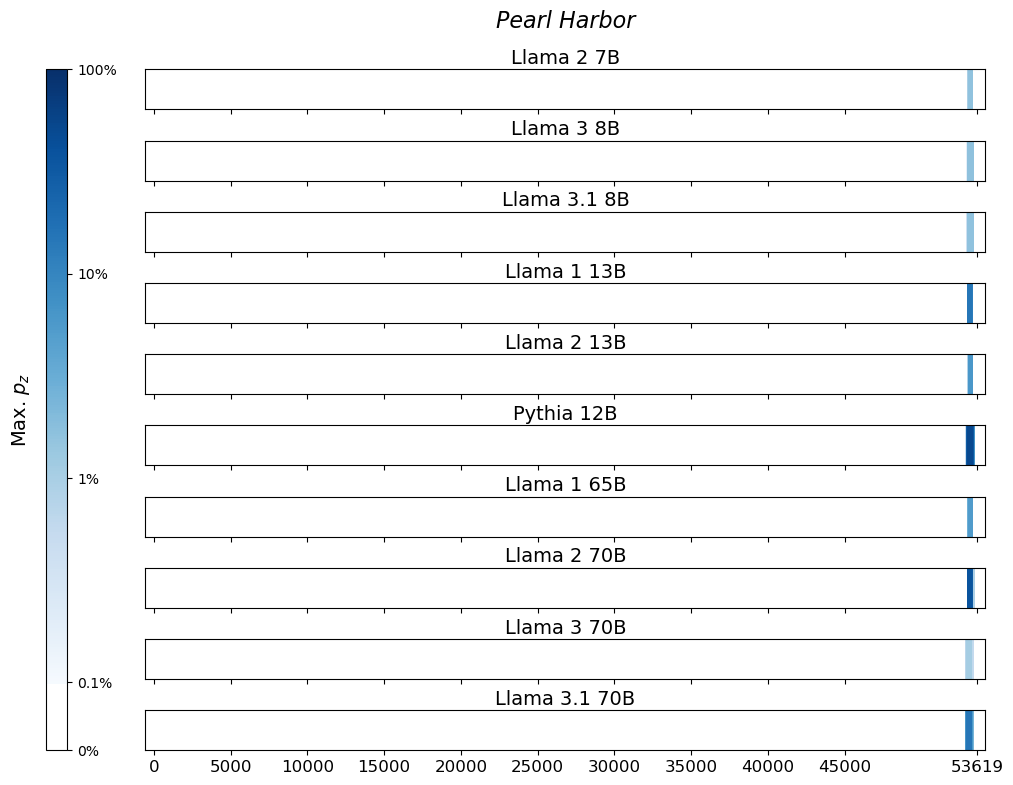}
    \includegraphics[width=\linewidth]{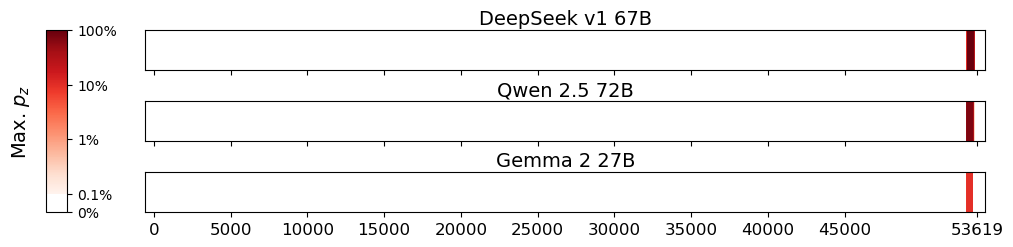}
    \includegraphics[width=\linewidth]{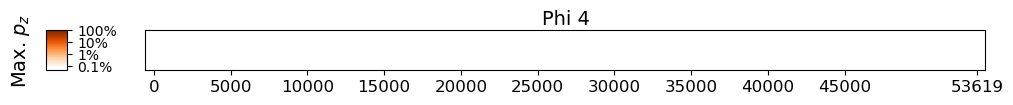}
  \end{minipage}
  \hfill
  \begin{minipage}[t]{0.45\textwidth}
    \centering
    \vspace{0cm}
    \includegraphics[width=\linewidth]{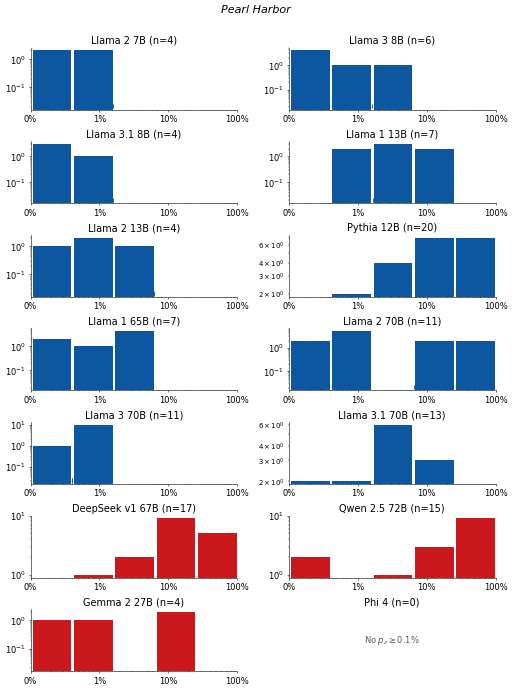}
  \end{minipage}
  \vspace{-.2cm}
  \caption{
    \textbf{\textit{Pearl Harbor}, \citeauthor{Pearl_Harbor}.}
    For $14$ LLMs,
    (\textbf{left}) heatmaps for the sliding-window procedure and
    (\textbf{right}) corresponding distributions over suffix extraction probabilities
    ($\tau_\text{min}=0.1\%$).
  }
  \label{fig:slidingwindow:Pearl_Harbor}
\end{figure}
\FloatBarrier

\subsubsection{\textit{Chesapeake Requiem}, \citeauthor{Chesapeake_Requiem}}\label{app:sec:sliding:Chesapeake_Requiem}
\vspace{-.2cm}
\begin{figure}[h]
  \centering
  \begin{minipage}[t]{0.53\textwidth}
    \centering
    \vspace{0cm}
    \includegraphics[width=\linewidth]{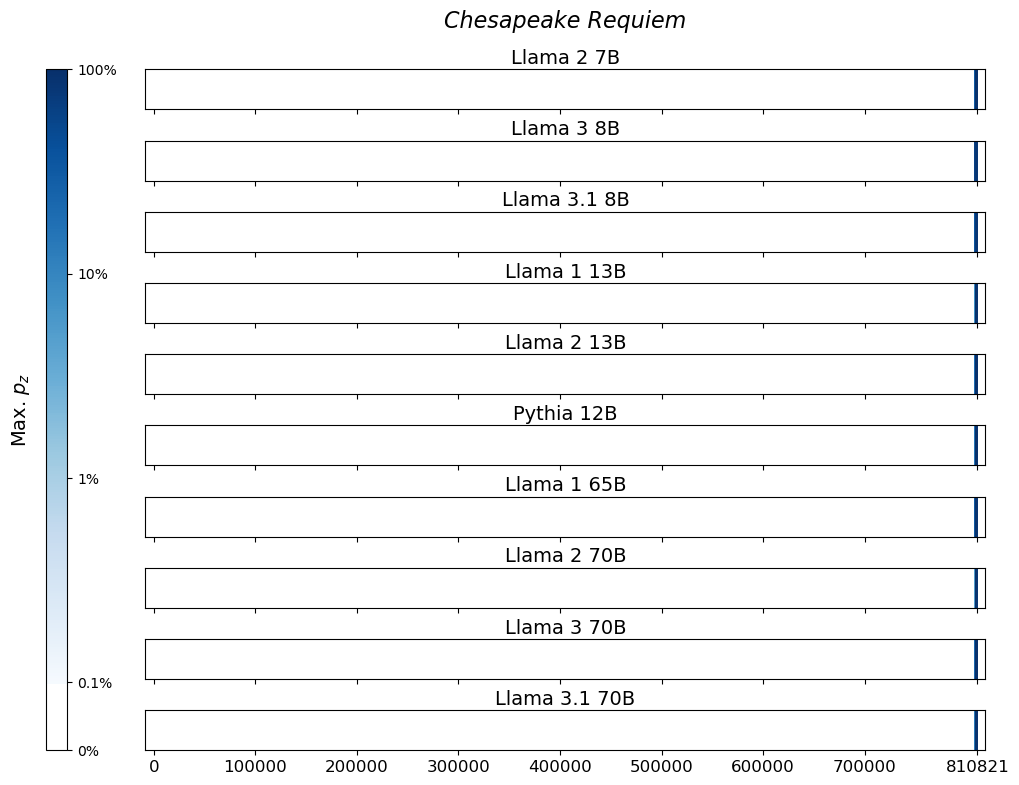}
    \includegraphics[width=\linewidth]{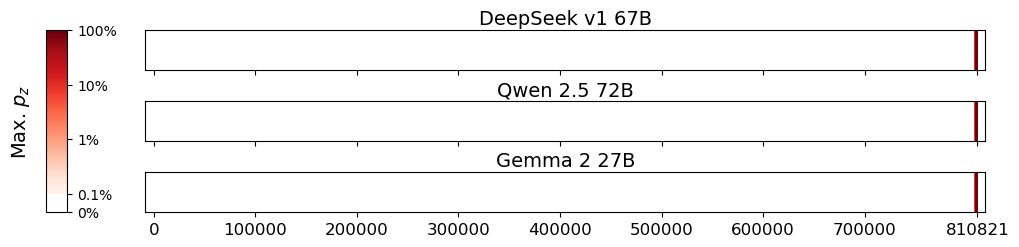}
    \includegraphics[width=\linewidth]{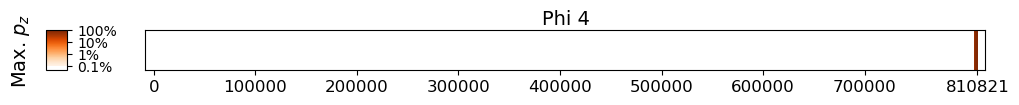}
  \end{minipage}
  \hfill
  \begin{minipage}[t]{0.45\textwidth}
    \centering
    \vspace{0cm}
    \includegraphics[width=\linewidth]{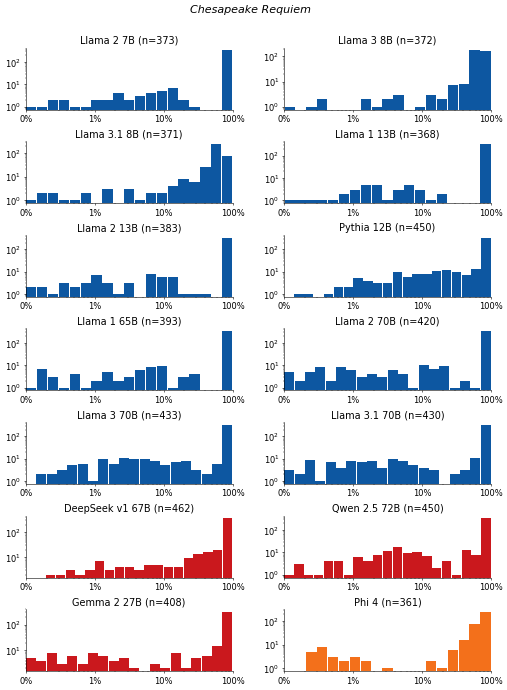}
  \end{minipage}
  \vspace{-.2cm}
  \caption{
    \textbf{\textit{Chesapeake Requiem}, \citeauthor{Chesapeake_Requiem}.}
    For $14$ LLMs,
    (\textbf{left}) heatmaps for the sliding-window procedure and
    (\textbf{right}) corresponding distributions over suffix extraction probabilities
    ($\tau_\text{min}=0.1\%$).
  }
  \label{fig:slidingwindow:Chesapeake_Requiem}
\end{figure}
\FloatBarrier

\clearpage
\subsubsection{\textit{The Goldfinch}, \citeauthor{The_Goldfinch}}\label{app:sec:sliding:The_Goldfinch}
\vspace{-.2cm}
\begin{figure}[h]
  \centering
  \begin{minipage}[t]{0.53\textwidth}
    \centering
    \vspace{0cm}
    \includegraphics[width=\linewidth]{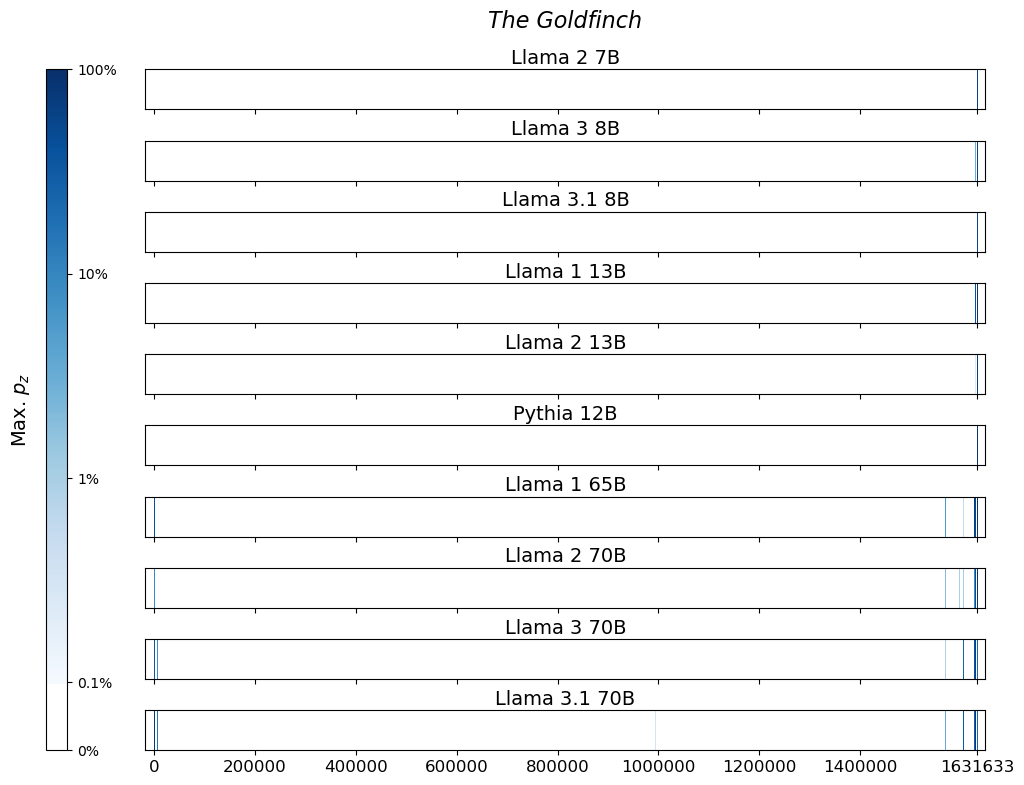}
    \includegraphics[width=\linewidth]{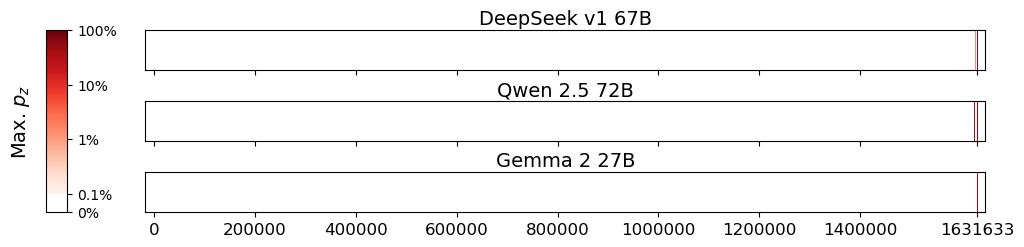}
    \includegraphics[width=\linewidth]{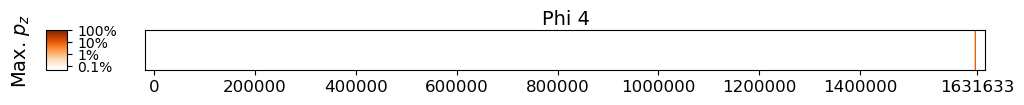}
  \end{minipage}
  \hfill
  \begin{minipage}[t]{0.45\textwidth}
    \centering
    \vspace{0cm}
    \includegraphics[width=\linewidth]{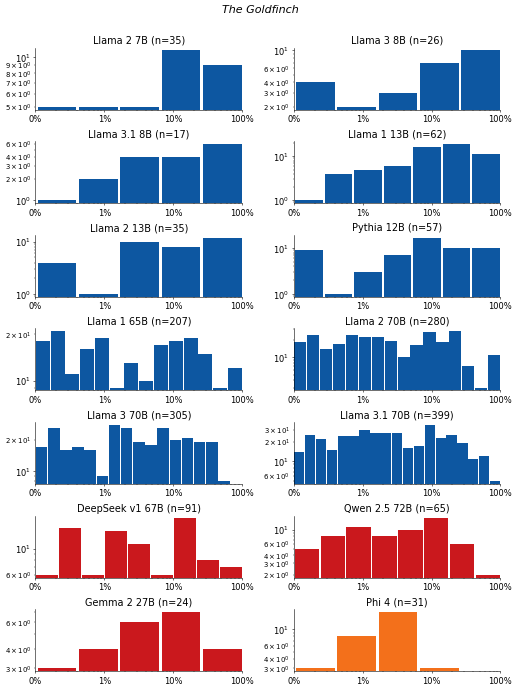}
  \end{minipage}
  \vspace{-.2cm}
  \caption{
    \textbf{\textit{The Goldfinch}, \citeauthor{The_Goldfinch}.}
    For $14$ LLMs,
    (\textbf{left}) heatmaps for the sliding-window procedure and
    (\textbf{right}) corresponding distributions over suffix extraction probabilities
    ($\tau_\text{min}=0.1\%$).
  }
  \label{fig:slidingwindow:The_Goldfinch}
\end{figure}
\FloatBarrier

\subsubsection{\textit{Unglued}, \citeauthor{Unglued}}\label{app:sec:sliding:Unglued}
\vspace{-.2cm}
\begin{figure}[h]
  \centering
  \begin{minipage}[t]{0.53\textwidth}
    \centering
    \vspace{0cm}
    \includegraphics[width=\linewidth]{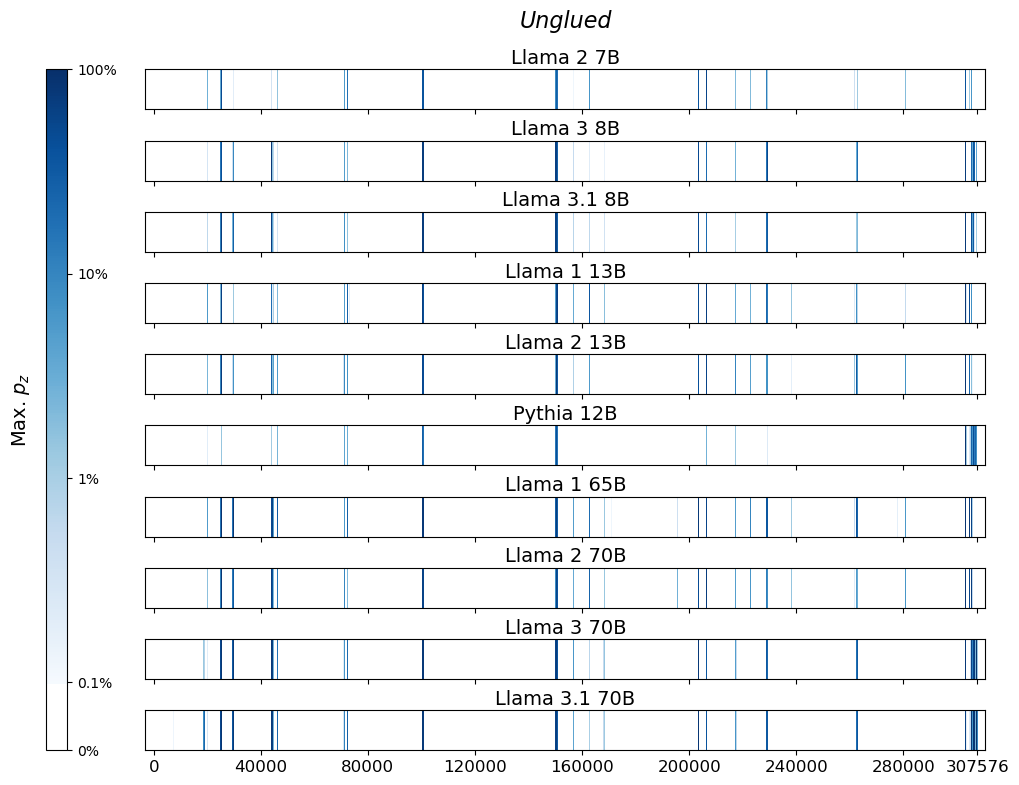}
    \includegraphics[width=\linewidth]{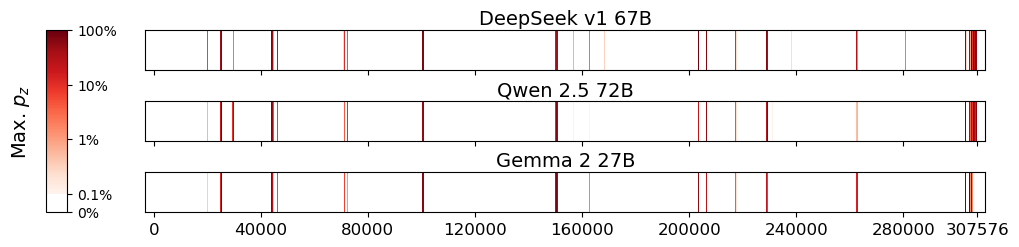}
    \includegraphics[width=\linewidth]{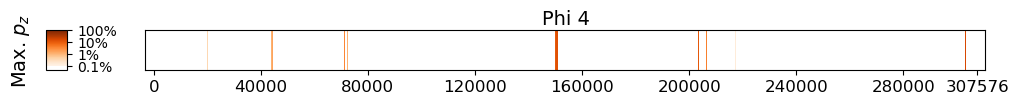}
  \end{minipage}
  \hfill
  \begin{minipage}[t]{0.45\textwidth}
    \centering
    \vspace{0cm}
    \includegraphics[width=\linewidth]{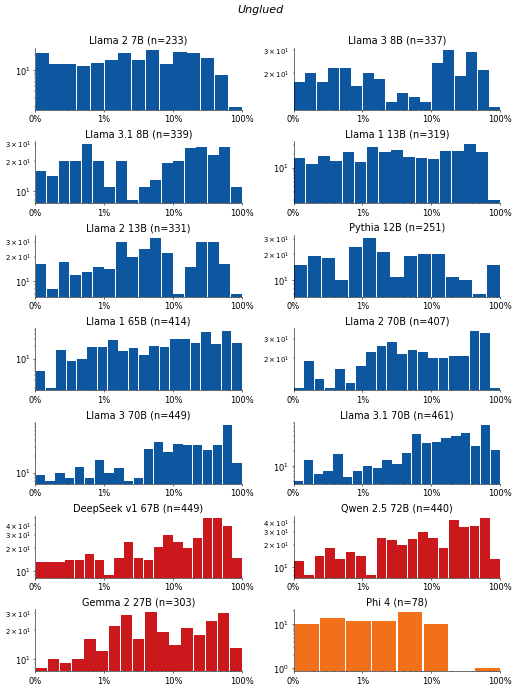}
  \end{minipage}
  \vspace{-.2cm}
  \caption{
    \textbf{\textit{Unglued}, \citeauthor{Unglued}.}
    For $14$ LLMs,
    (\textbf{left}) heatmaps for the sliding-window procedure and
    (\textbf{right}) corresponding distributions over suffix extraction probabilities
    ($\tau_\text{min}=0.1\%$).
  }
  \label{fig:slidingwindow:Unglued}
\end{figure}
\FloatBarrier

\clearpage
\subsubsection{\textit{Embraced}, \citeauthor{Embraced}}\label{app:sec:sliding:Embraced}
\vspace{-.2cm}
\begin{figure}[h]
  \centering
  \begin{minipage}[t]{0.53\textwidth}
    \centering
    \vspace{0cm}
    \includegraphics[width=\linewidth]{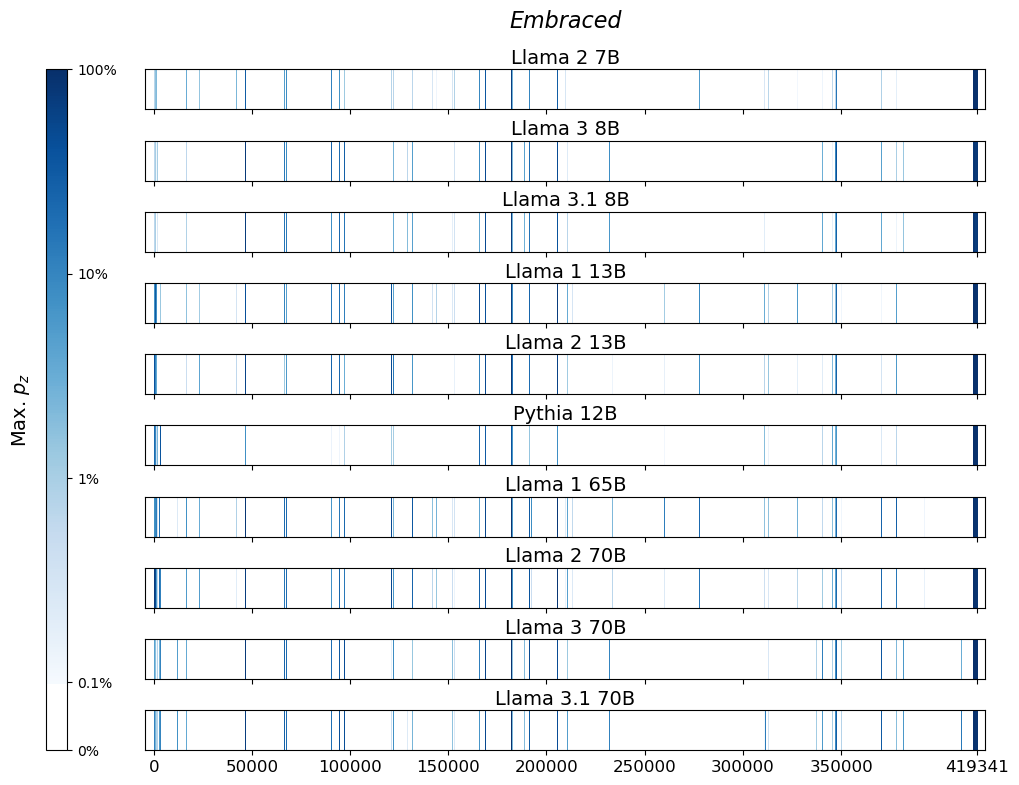}
    \includegraphics[width=\linewidth]{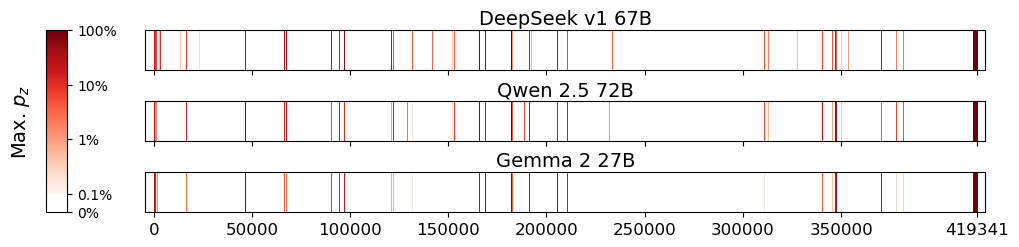}
    \includegraphics[width=\linewidth]{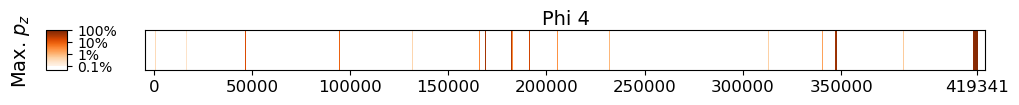}
  \end{minipage}
  \hfill
  \begin{minipage}[t]{0.45\textwidth}
    \centering
    \vspace{0cm}
    \includegraphics[width=\linewidth]{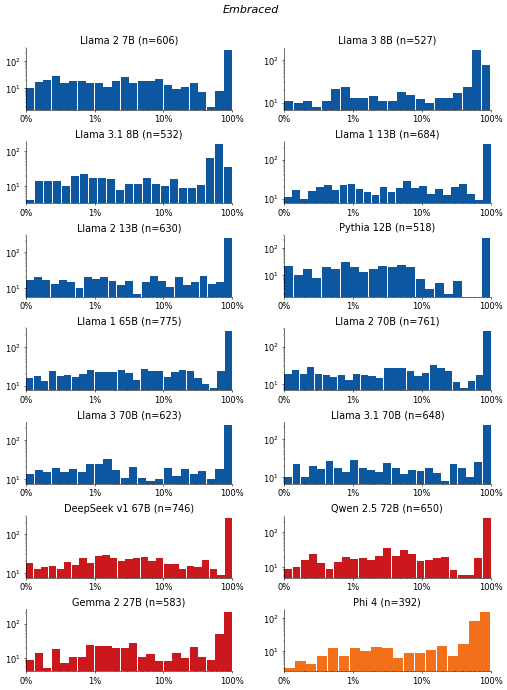}
  \end{minipage}
  \vspace{-.2cm}
  \caption{
    \textbf{\textit{Embraced}, \citeauthor{Embraced}.}
    For $14$ LLMs,
    (\textbf{left}) heatmaps for the sliding-window procedure and
    (\textbf{right}) corresponding distributions over suffix extraction probabilities
    ($\tau_\text{min}=0.1\%$).
  }
  \label{fig:slidingwindow:Embraced}
\end{figure}
\FloatBarrier

\subsubsection{\textit{Birding with Yeats}, \citeauthor{Birding_with_Yeats}}\label{app:sec:sliding:Birding_with_Yeats}
\vspace{-.2cm}
\begin{figure}[h]
  \centering
  \begin{minipage}[t]{0.53\textwidth}
    \centering
    \vspace{0cm}
    \includegraphics[width=\linewidth]{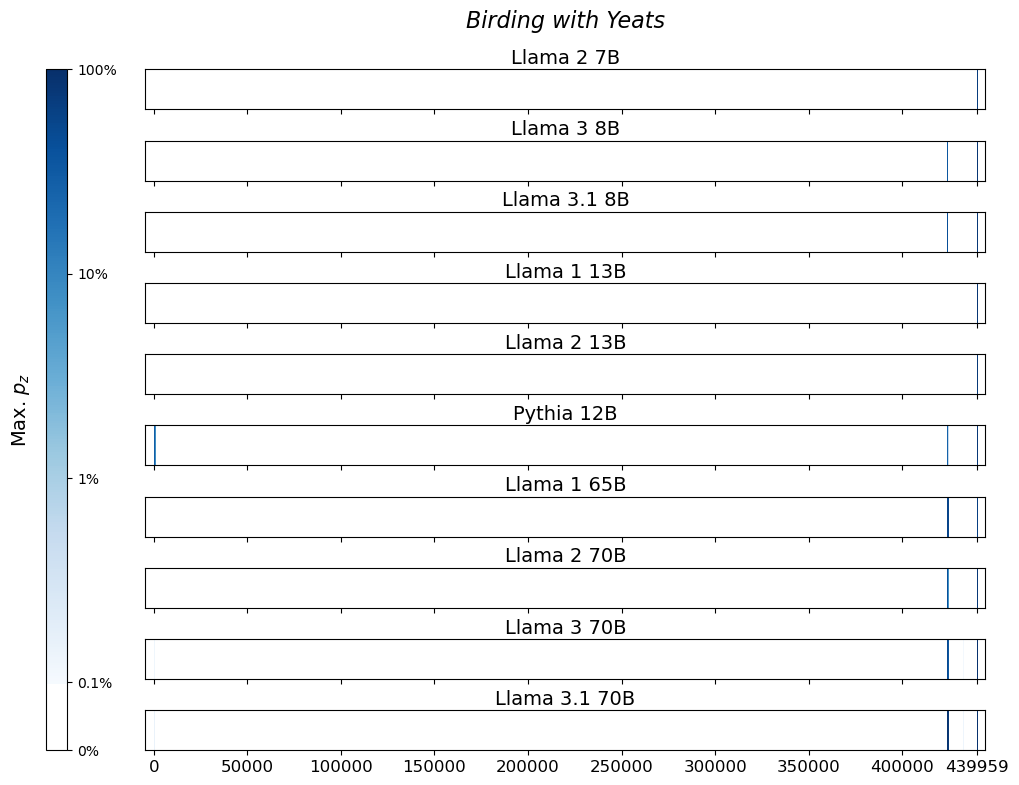}
    \includegraphics[width=\linewidth]{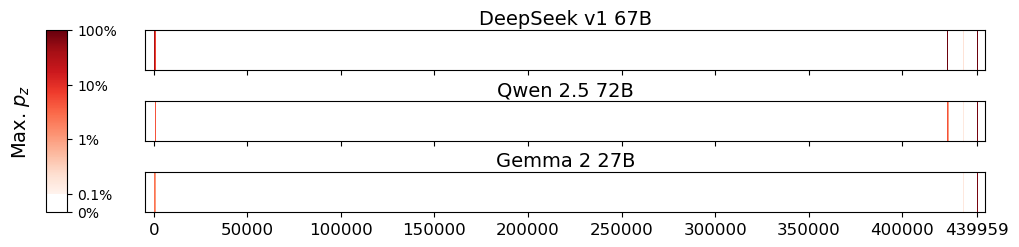}
    \includegraphics[width=\linewidth]{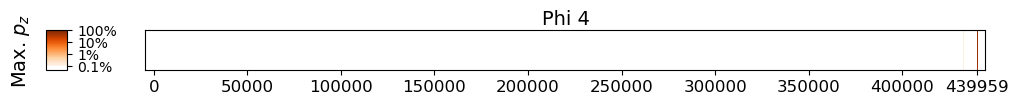}
  \end{minipage}
  \hfill
  \begin{minipage}[t]{0.45\textwidth}
    \centering
    \vspace{0cm}
    \includegraphics[width=\linewidth]{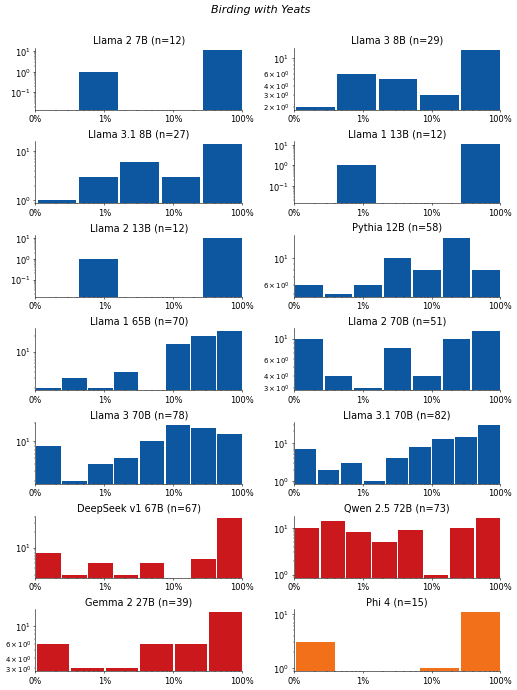}
  \end{minipage}
  \vspace{-.2cm}
  \caption{
    \textbf{\textit{Birding with Yeats}, \citeauthor{Birding_with_Yeats}.}
    For $14$ LLMs,
    (\textbf{left}) heatmaps for the sliding-window procedure and
    (\textbf{right}) corresponding distributions over suffix extraction probabilities
    ($\tau_\text{min}=0.1\%$).
  }
  \label{fig:slidingwindow:Birding_with_Yeats}
\end{figure}
\FloatBarrier

\clearpage
\subsubsection{\textit{The Hobbit}, \citeauthor{The_Hobbit}}\label{app:sec:sliding:The_Hobbit}
\vspace{-.2cm}
\begin{figure}[h]
  \centering
  \begin{minipage}[t]{0.53\textwidth}
    \centering
    \vspace{0cm}
    \includegraphics[width=\linewidth]{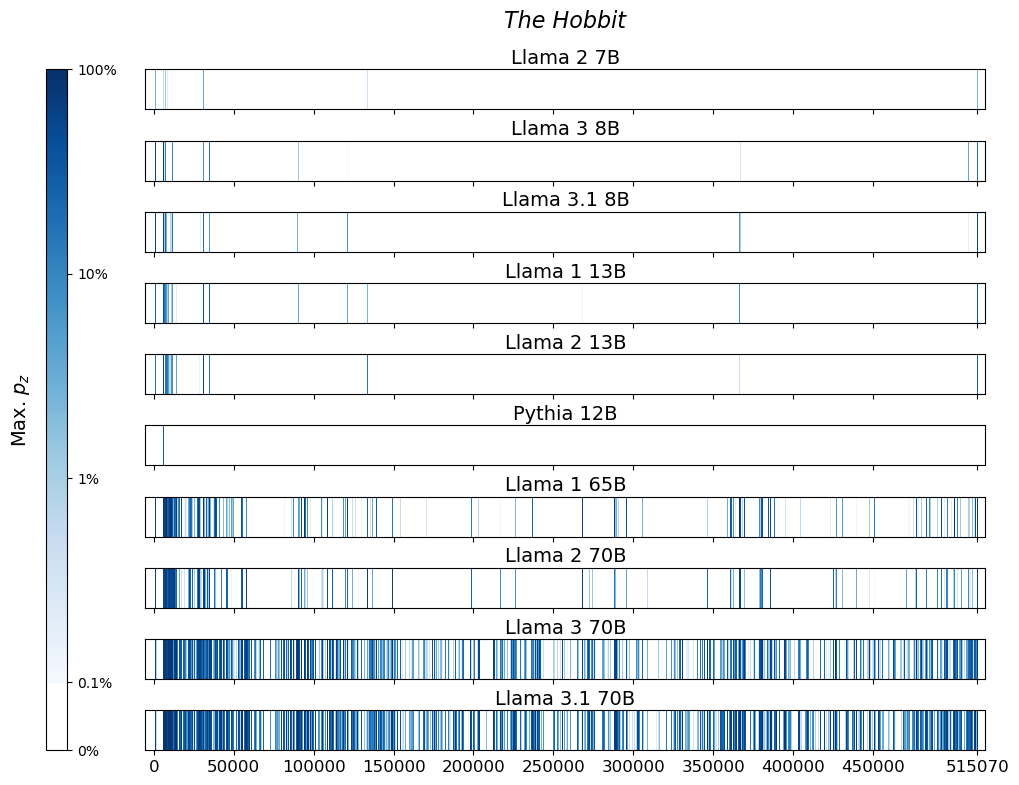}
    \includegraphics[width=\linewidth]{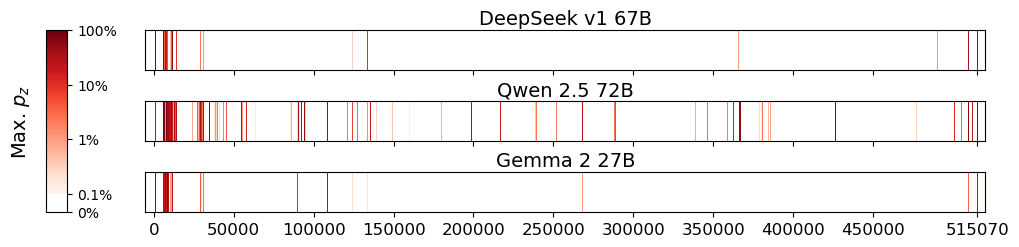}
    \includegraphics[width=\linewidth]{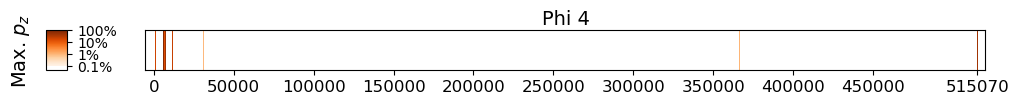}
  \end{minipage}
  \hfill
  \begin{minipage}[t]{0.45\textwidth}
    \centering
    \vspace{0cm}
    \includegraphics[width=\linewidth]{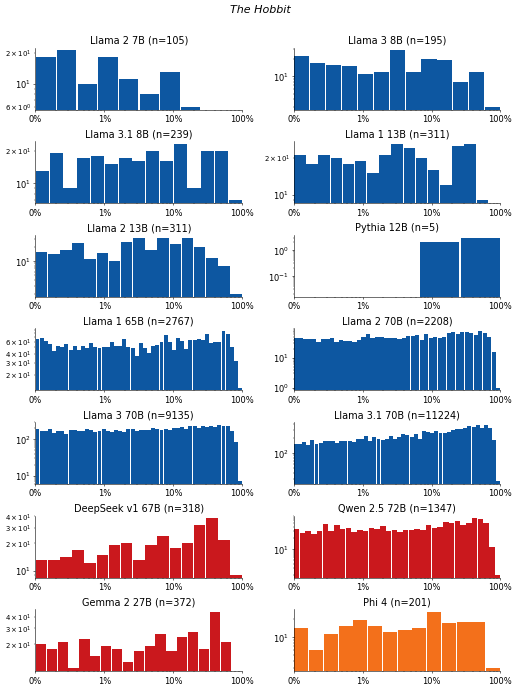}
  \end{minipage}
  \vspace{-.2cm}
  \caption{
    \textbf{\textit{The Hobbit}, \citeauthor{The_Hobbit}.}
    For $14$ LLMs,
    (\textbf{left}) heatmaps for the sliding-window procedure and
    (\textbf{right}) corresponding distributions over suffix extraction probabilities
    ($\tau_\text{min}=0.1\%$).
  }
  \label{fig:slidingwindow:The_Hobbit}
\end{figure}
\FloatBarrier

\subsubsection{\textit{The Fellowship of the Ring}, \citeauthor{The_Fellowship_of_the_Ring}}\label{app:sec:sliding:The_Fellowship_of_the_Ring}
\vspace{-.2cm}
\begin{figure}[h]
  \centering
  \begin{minipage}[t]{0.53\textwidth}
    \centering
    \vspace{0cm}
    \includegraphics[width=\linewidth]{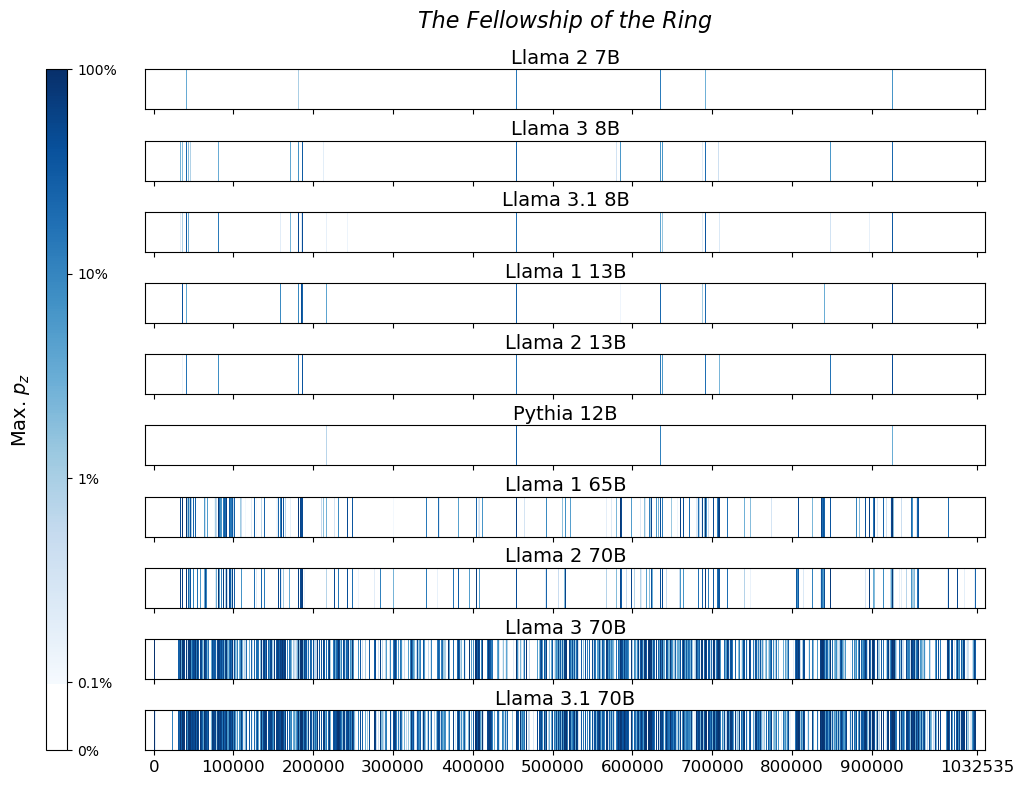}
    \includegraphics[width=\linewidth]{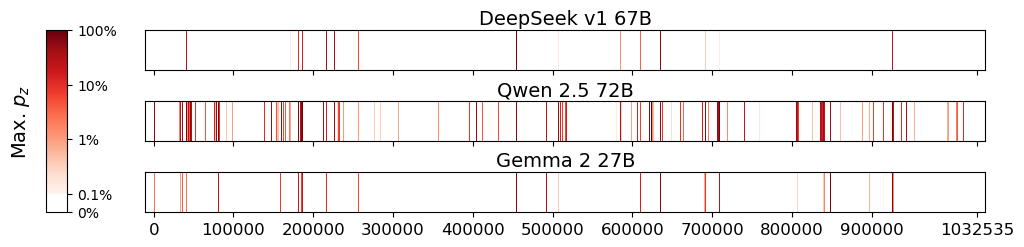}
    \includegraphics[width=\linewidth]{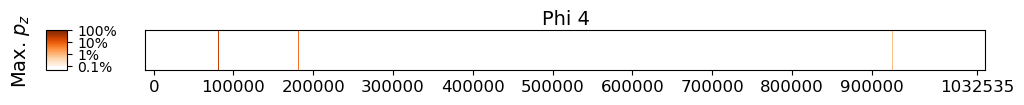}
  \end{minipage}
  \hfill
  \begin{minipage}[t]{0.45\textwidth}
    \centering
    \vspace{0cm}
    \includegraphics[width=\linewidth]{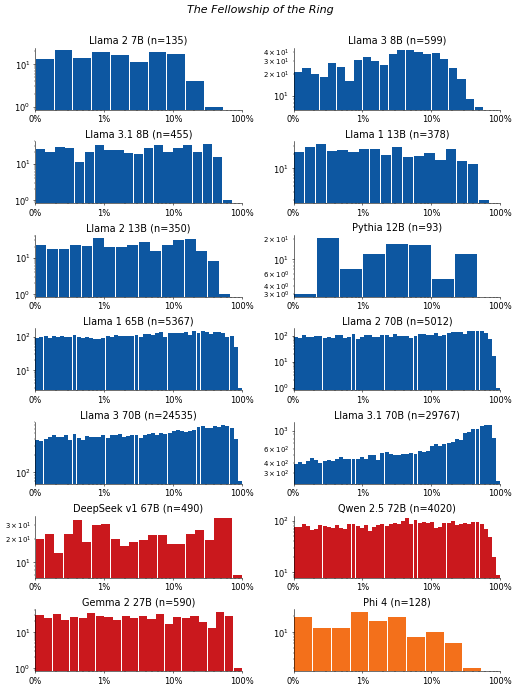}
  \end{minipage}
  \vspace{-.2cm}
  \caption{
    \textbf{\textit{The Fellowship of the Ring}, \citeauthor{The_Fellowship_of_the_Ring}.}
    For $14$ LLMs,
    (\textbf{left}) heatmaps for the sliding-window procedure and
    (\textbf{right}) corresponding distributions over suffix extraction probabilities
    ($\tau_\text{min}=0.1\%$).
  }
  \label{fig:slidingwindow:The_Fellowship_of_the_Ring}
\end{figure}
\FloatBarrier

\clearpage
\subsubsection{\textit{Tree and Leaf}, \citeauthor{Tree_and_Leaf}}\label{app:sec:sliding:Tree_and_Leaf}
\vspace{-.2cm}
\begin{figure}[h]
  \centering
  \begin{minipage}[t]{0.53\textwidth}
    \centering
    \vspace{0cm}
    \includegraphics[width=\linewidth]{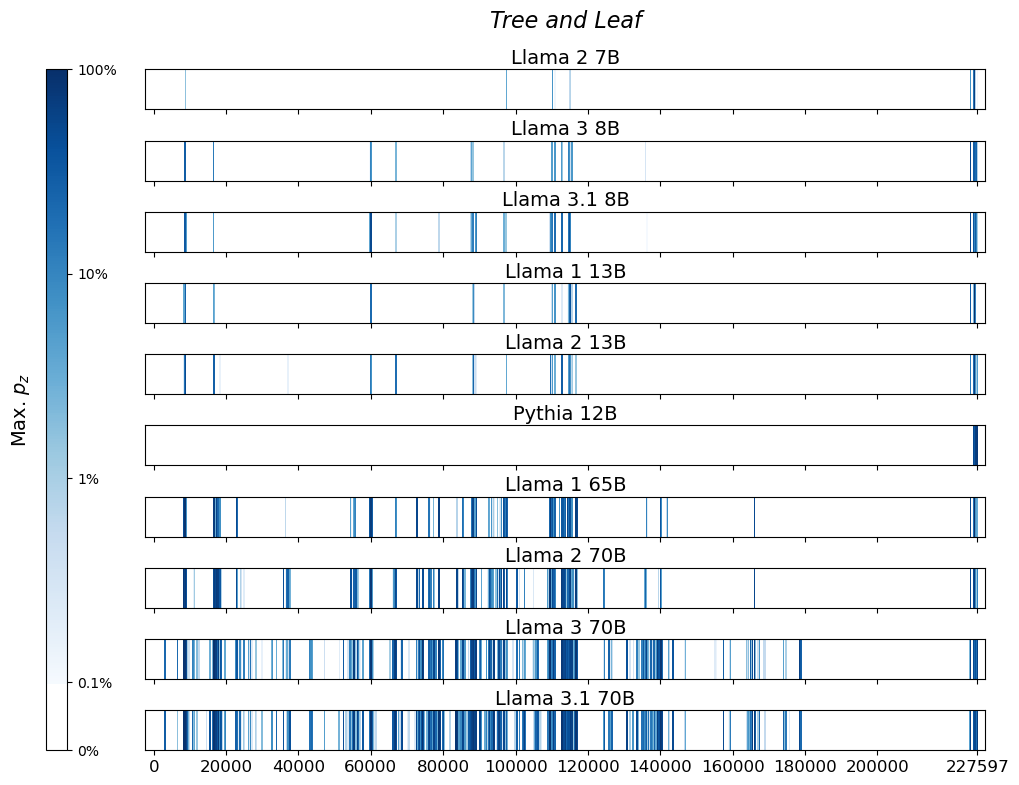}
    \includegraphics[width=\linewidth]{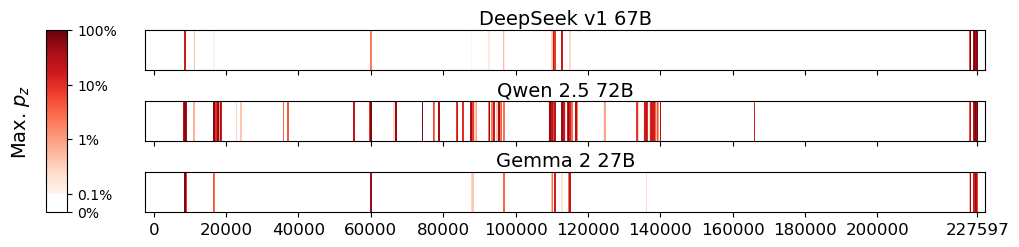}
    \includegraphics[width=\linewidth]{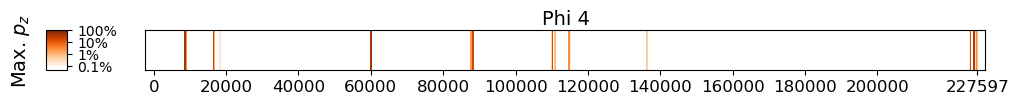}
  \end{minipage}
  \hfill
  \begin{minipage}[t]{0.45\textwidth}
    \centering
    \vspace{0cm}
    \includegraphics[width=\linewidth]{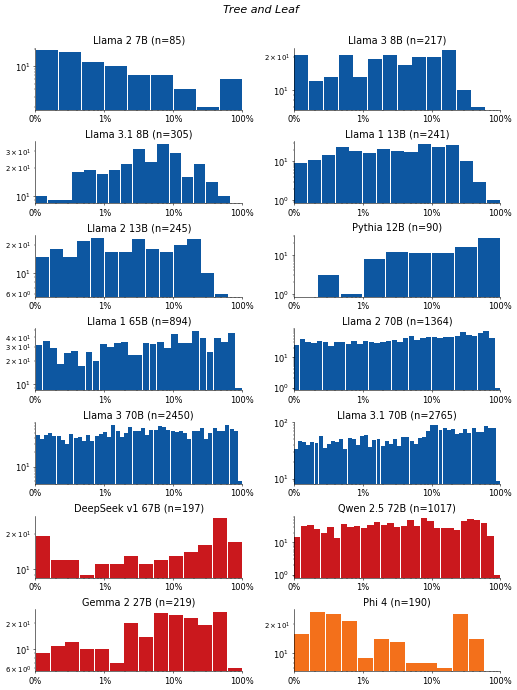}
  \end{minipage}
  \vspace{-.2cm}
  \caption{
    \textbf{\textit{Tree and Leaf}, \citeauthor{Tree_and_Leaf}.}
    For $14$ LLMs,
    (\textbf{left}) heatmaps for the sliding-window procedure and
    (\textbf{right}) corresponding distributions over suffix extraction probabilities
    ($\tau_\text{min}=0.1\%$).
  }
  \label{fig:slidingwindow:Tree_and_Leaf}
\end{figure}
\FloatBarrier

\subsubsection{\textit{Noodles Every Day}, \citeauthor{Noodles_Every_Day}}\label{app:sec:sliding:Noodles_Every_Day}
\vspace{-.2cm}
\begin{figure}[h]
  \centering
  \begin{minipage}[t]{0.53\textwidth}
    \centering
    \vspace{0cm}
    \includegraphics[width=\linewidth]{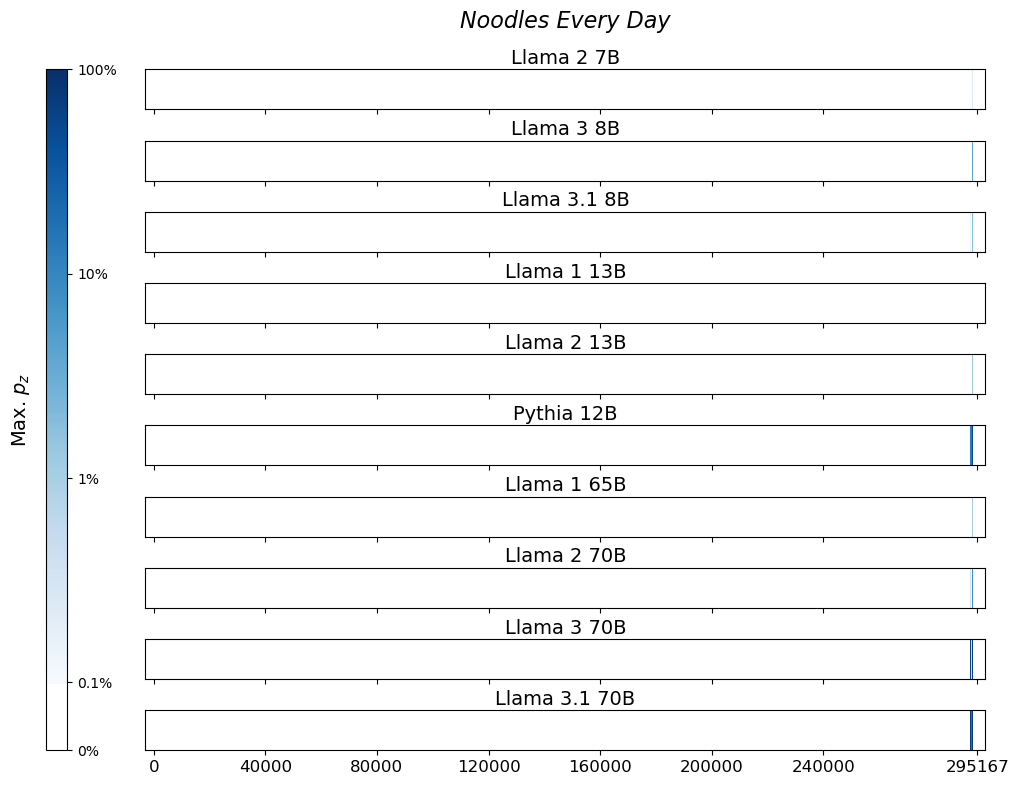}
    \includegraphics[width=\linewidth]{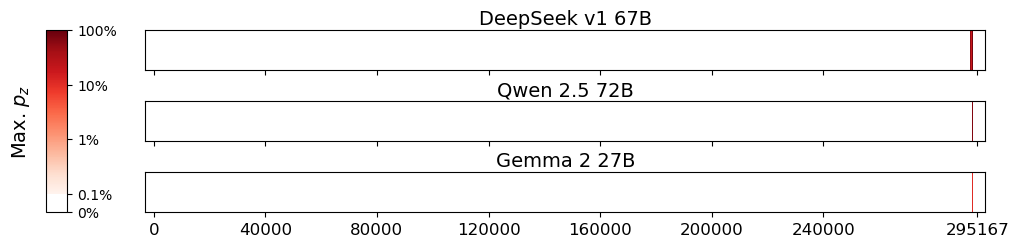}
    \includegraphics[width=\linewidth]{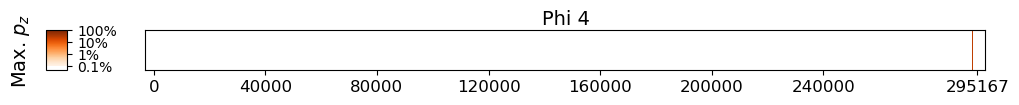}
  \end{minipage}
  \hfill
  \begin{minipage}[t]{0.45\textwidth}
    \centering
    \vspace{0cm}
    \includegraphics[width=\linewidth]{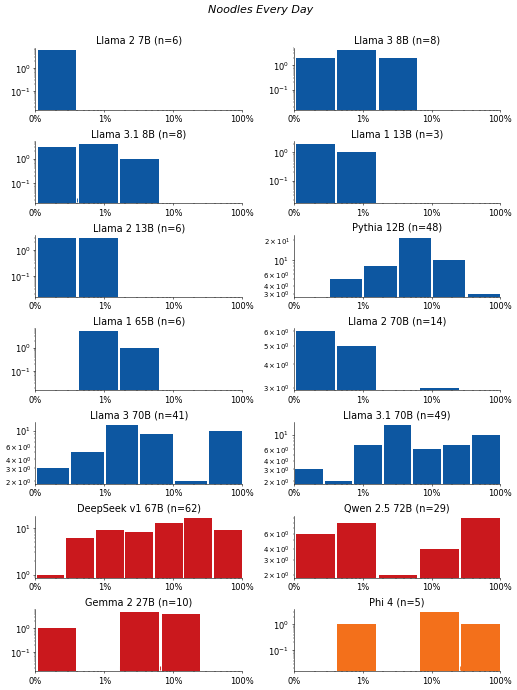}
  \end{minipage}
  \vspace{-.2cm}
  \caption{
    \textbf{\textit{Noodles Every Day}, \citeauthor{Noodles_Every_Day}.}
    For $14$ LLMs,
    (\textbf{left}) heatmaps for the sliding-window procedure and
    (\textbf{right}) corresponding distributions over suffix extraction probabilities
    ($\tau_\text{min}=0.1\%$).
  }
  \label{fig:slidingwindow:Noodles_Every_Day}
\end{figure}
\FloatBarrier

\clearpage
\subsubsection{\textit{Billionaire Democracy}, \citeauthor{Billionaire_Democracy}}\label{app:sec:sliding:Billionaire_Democracy}
\vspace{-.2cm}
\begin{figure}[h]
  \centering
  \begin{minipage}[t]{0.53\textwidth}
    \centering
    \vspace{0cm}
    \includegraphics[width=\linewidth]{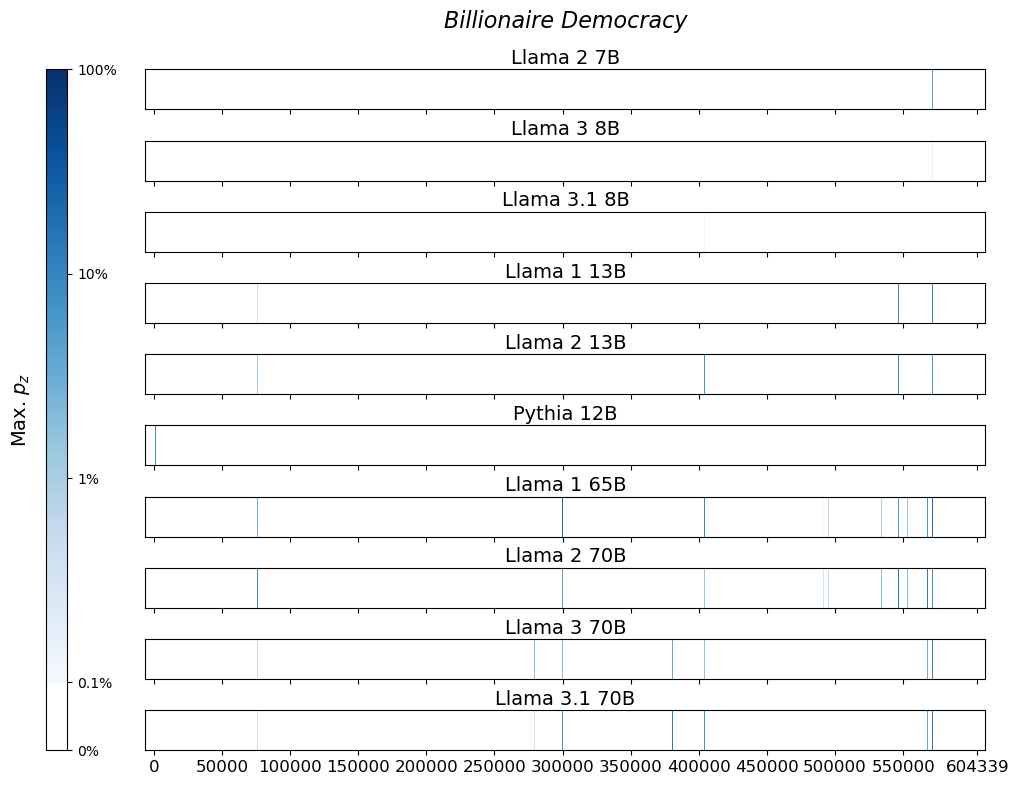}
    \includegraphics[width=\linewidth]{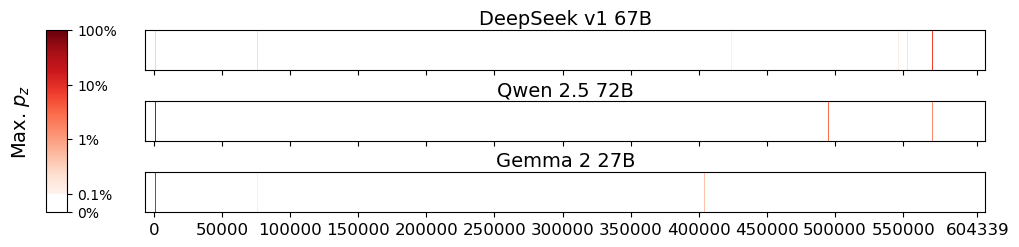}
    \includegraphics[width=\linewidth]{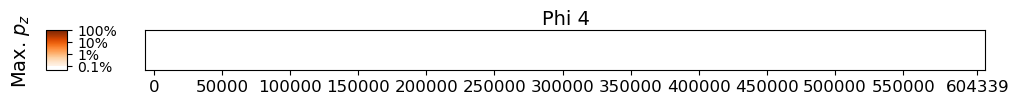}
  \end{minipage}
  \hfill
  \begin{minipage}[t]{0.45\textwidth}
    \centering
    \vspace{0cm}
    \includegraphics[width=\linewidth]{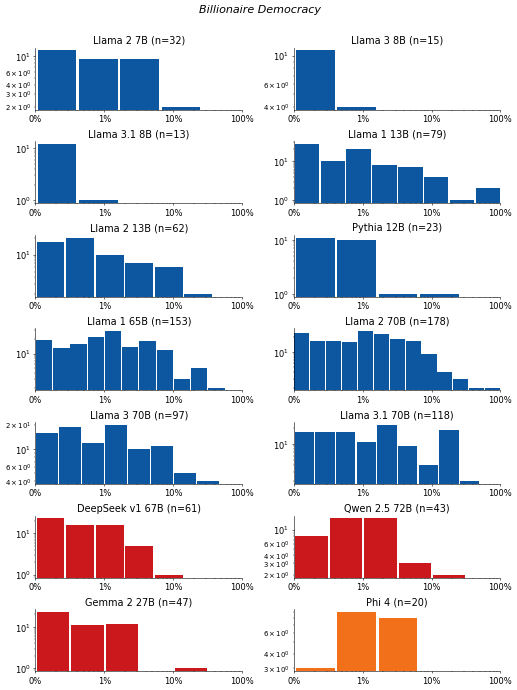}
  \end{minipage}
  \vspace{-.2cm}
  \caption{
    \textbf{\textit{Billionaire Democracy}, \citeauthor{Billionaire_Democracy}.}
    For $14$ LLMs,
    (\textbf{left}) heatmaps for the sliding-window procedure and
    (\textbf{right}) corresponding distributions over suffix extraction probabilities
    ($\tau_\text{min}=0.1\%$).
  }
  \label{fig:slidingwindow:Billionaire_Democracy}
\end{figure}
\FloatBarrier

\subsubsection{\textit{Portugal's Guerrilla Wars in Africa}, \citeauthor{Portugal_s_Guerrilla_Wars_in_Africa}}\label{app:sec:sliding:Portugal_s_Guerrilla_Wars_in_Africa}
\vspace{-.2cm}
\begin{figure}[h]
  \centering
  \begin{minipage}[t]{0.53\textwidth}
    \centering
    \vspace{0cm}
    \includegraphics[width=\linewidth]{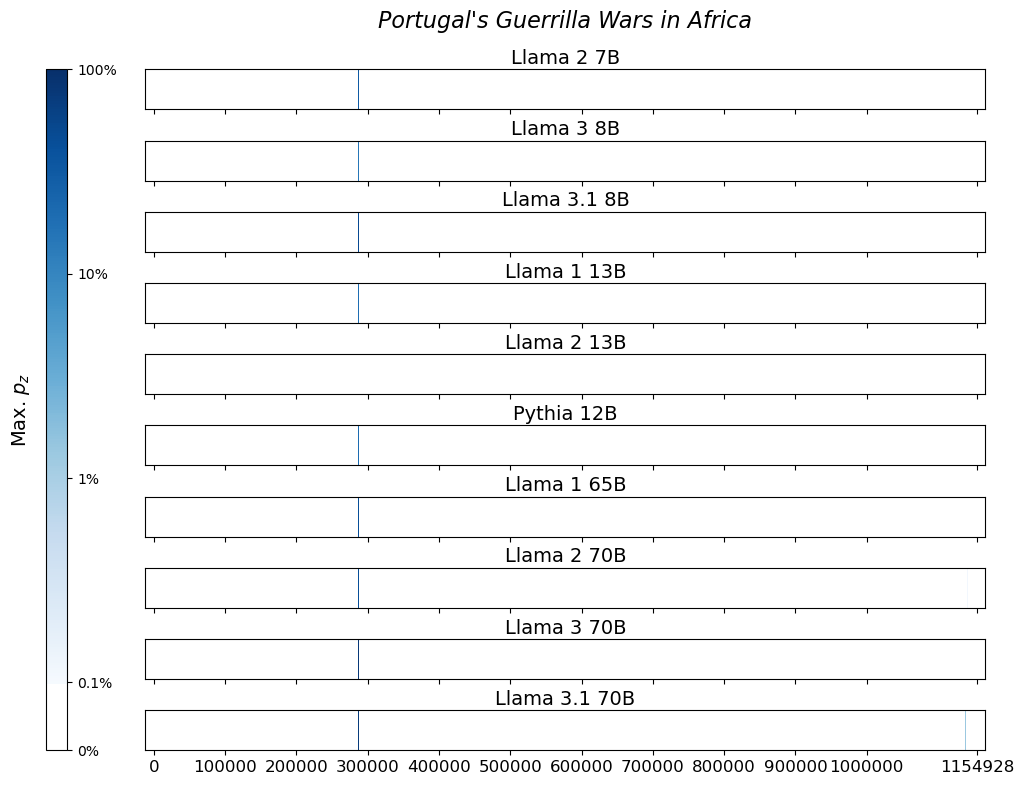}
    \includegraphics[width=\linewidth]{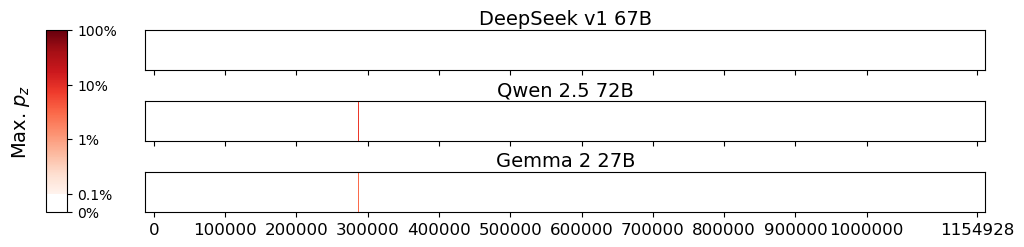}
    \includegraphics[width=\linewidth]{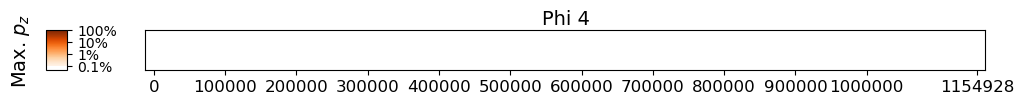}
  \end{minipage}
  \hfill
  \begin{minipage}[t]{0.45\textwidth}
    \centering
    \vspace{0cm}
    \includegraphics[width=\linewidth]{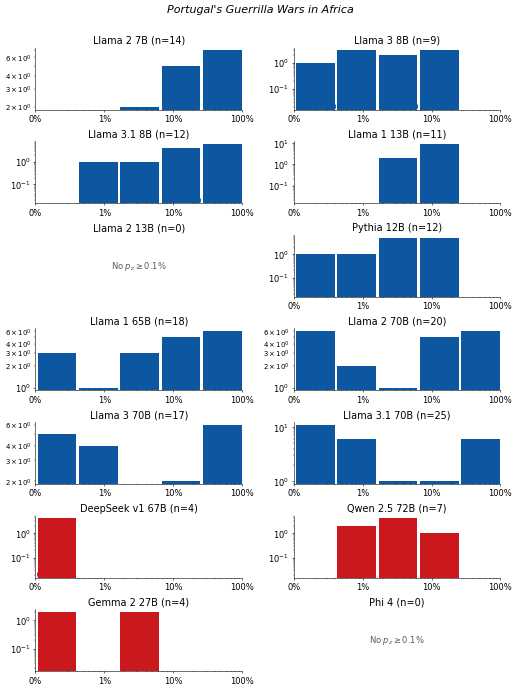}
  \end{minipage}
  \vspace{-.2cm}
  \caption{
    \textbf{\textit{Portugal's Guerrilla Wars in Africa}, \citeauthor{Portugal_s_Guerrilla_Wars_in_Africa}.}
    For $14$ LLMs,
    (\textbf{left}) heatmaps for the sliding-window procedure and
    (\textbf{right}) corresponding distributions over suffix extraction probabilities
    ($\tau_\text{min}=0.1\%$).
  }
  \label{fig:slidingwindow:Portugal_s_Guerrilla_Wars_in_Africa}
\end{figure}
\FloatBarrier

\clearpage
\subsubsection{\textit{Slaughterhouse-Five}, \citeauthor{Slaughterhouse-Five}}\label{app:sec:sliding:Slaughterhouse-Five}
\vspace{-.2cm}
\begin{figure}[h]
  \centering
  \begin{minipage}[t]{0.53\textwidth}
    \centering
    \vspace{0cm}
    \includegraphics[width=\linewidth]{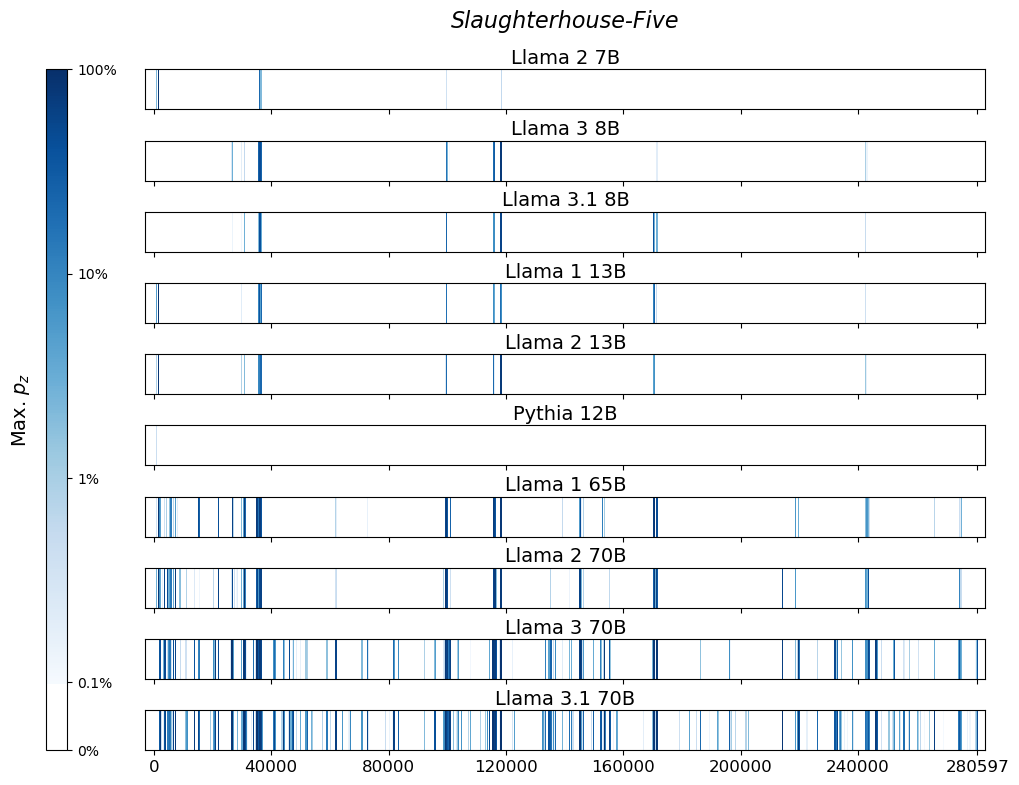}
    \includegraphics[width=\linewidth]{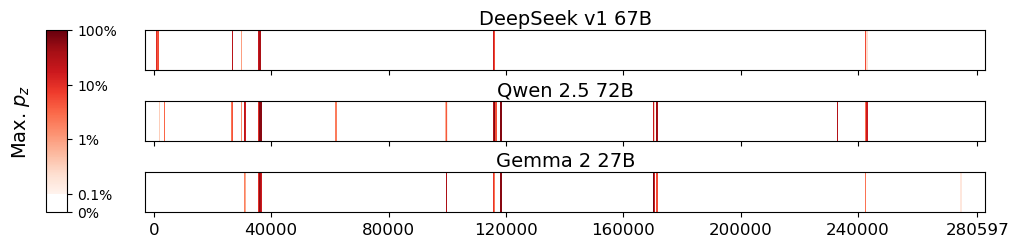}
    \includegraphics[width=\linewidth]{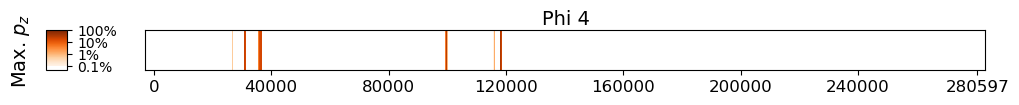}
  \end{minipage}
  \hfill
  \begin{minipage}[t]{0.45\textwidth}
    \centering
    \vspace{0cm}
    \includegraphics[width=\linewidth]{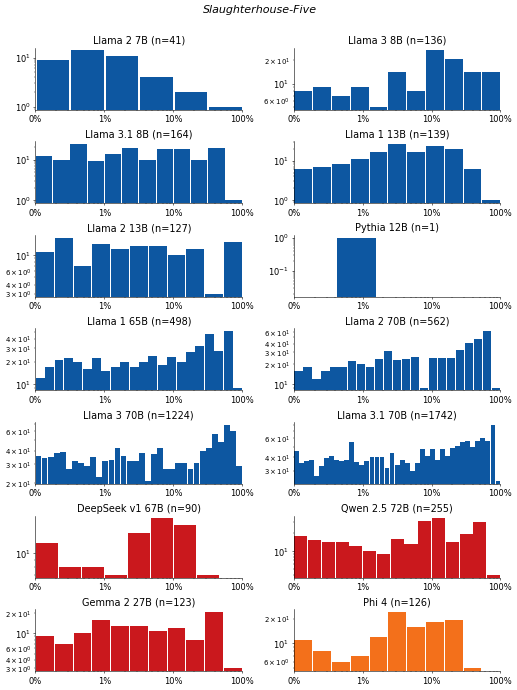}
  \end{minipage}
  \vspace{-.2cm}
  \caption{
    \textbf{\textit{Slaughterhouse-Five}, \citeauthor{Slaughterhouse-Five}.}
    For $14$ LLMs,
    (\textbf{left}) heatmaps for the sliding-window procedure and
    (\textbf{right}) corresponding distributions over suffix extraction probabilities
    ($\tau_\text{min}=0.1\%$).
  }
  \label{fig:slidingwindow:Slaughterhouse-Five}
\end{figure}
\FloatBarrier

\subsubsection{\textit{Animal Rights}, \citeauthor{Animal_Rights}}\label{app:sec:sliding:Animal_Rights}
\vspace{-.2cm}
\begin{figure}[h]
  \centering
  \begin{minipage}[t]{0.53\textwidth}
    \centering
    \vspace{0cm}
    \includegraphics[width=\linewidth]{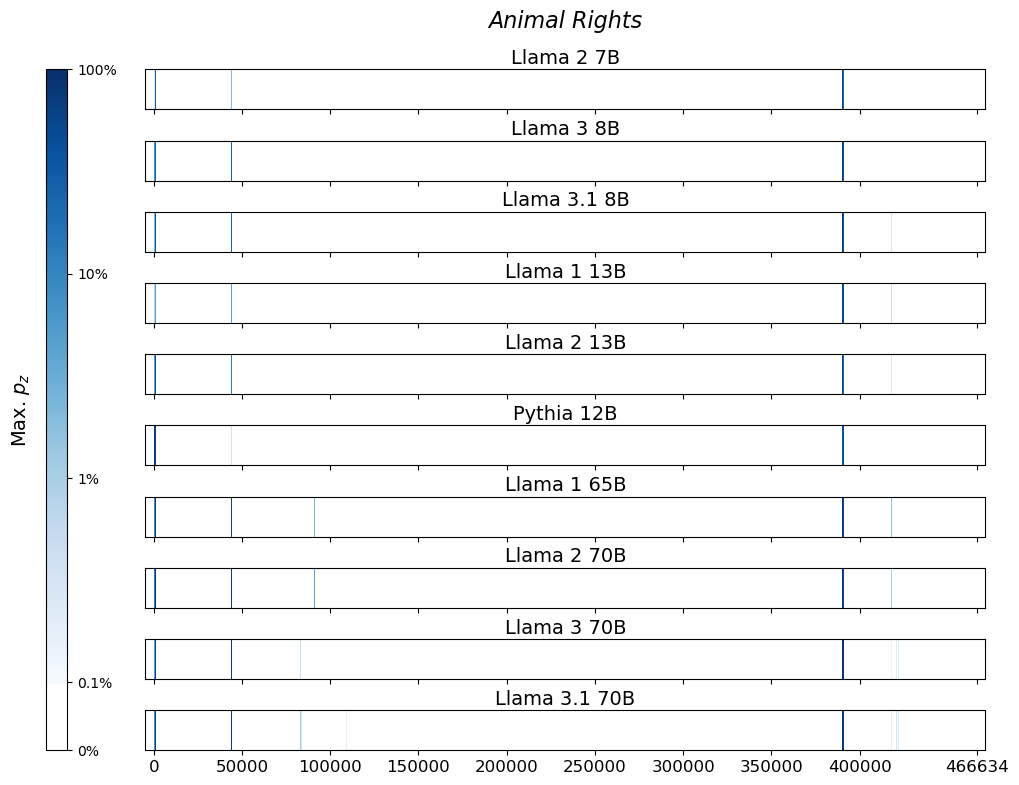}
    \includegraphics[width=\linewidth]{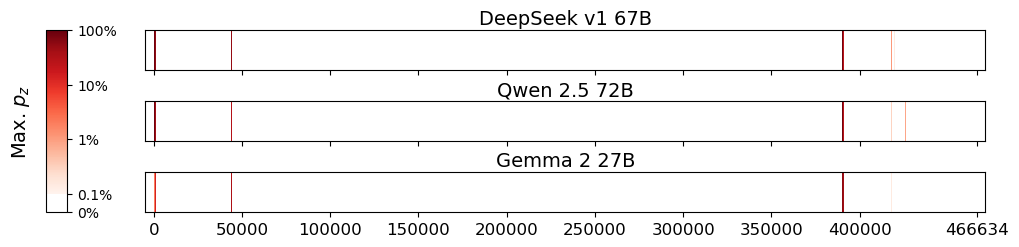}
    \includegraphics[width=\linewidth]{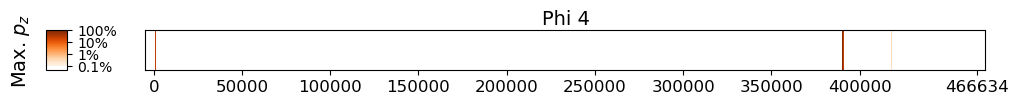}
  \end{minipage}
  \hfill
  \begin{minipage}[t]{0.45\textwidth}
    \centering
    \vspace{0cm}
    \includegraphics[width=\linewidth]{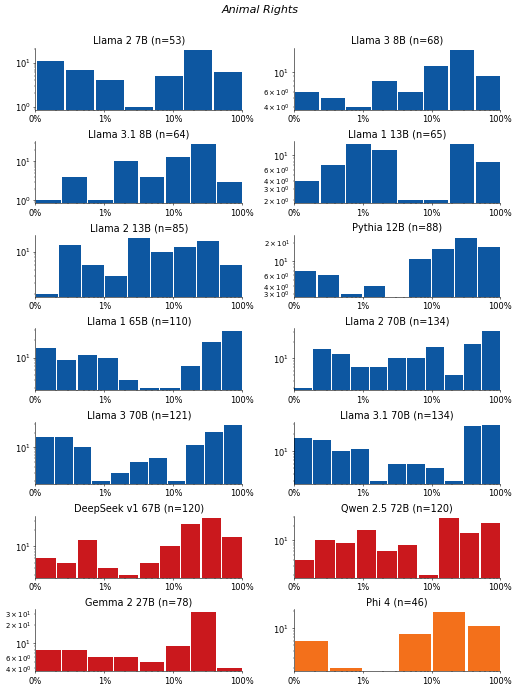}
  \end{minipage}
  \vspace{-.2cm}
  \caption{
    \textbf{\textit{Animal Rights}, \citeauthor{Animal_Rights}.}
    For $14$ LLMs,
    (\textbf{left}) heatmaps for the sliding-window procedure and
    (\textbf{right}) corresponding distributions over suffix extraction probabilities
    ($\tau_\text{min}=0.1\%$).
  }
  \label{fig:slidingwindow:Animal_Rights}
\end{figure}
\FloatBarrier

\clearpage
\subsubsection{\textit{Men We Reaped}, \citeauthor{Men_We_Reaped}}\label{app:sec:sliding:Men_We_Reaped}
\vspace{-.2cm}
\begin{figure}[h]
  \centering
  \begin{minipage}[t]{0.53\textwidth}
    \centering
    \vspace{0cm}
    \includegraphics[width=\linewidth]{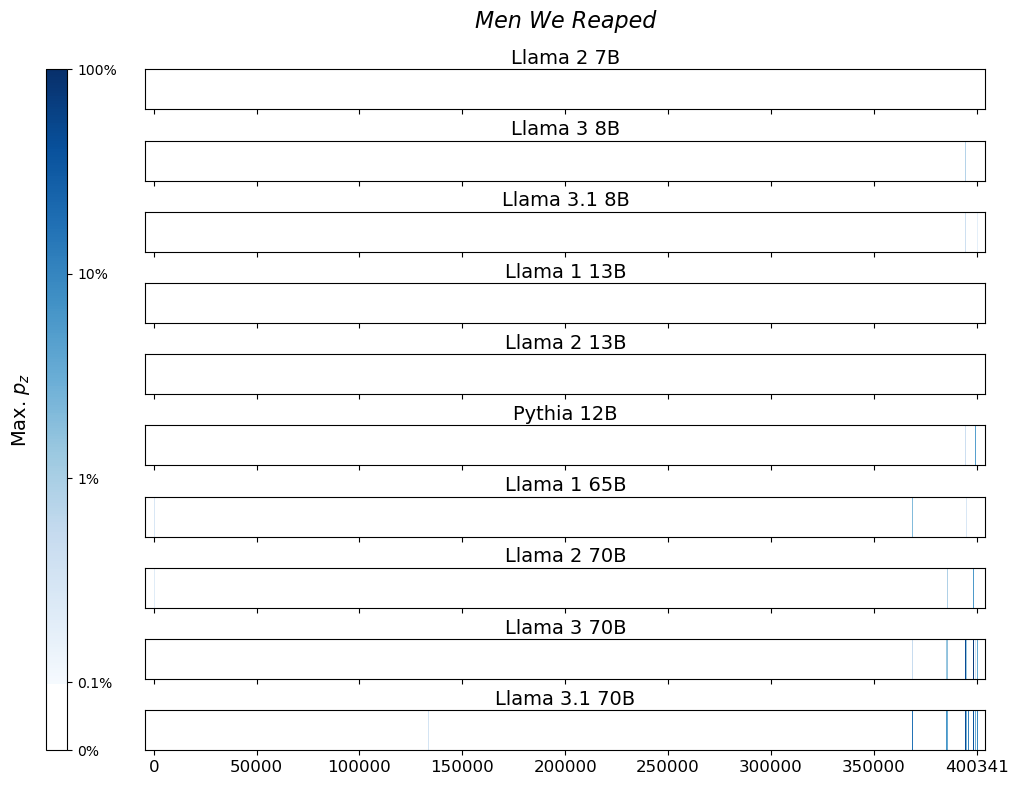}
    \includegraphics[width=\linewidth]{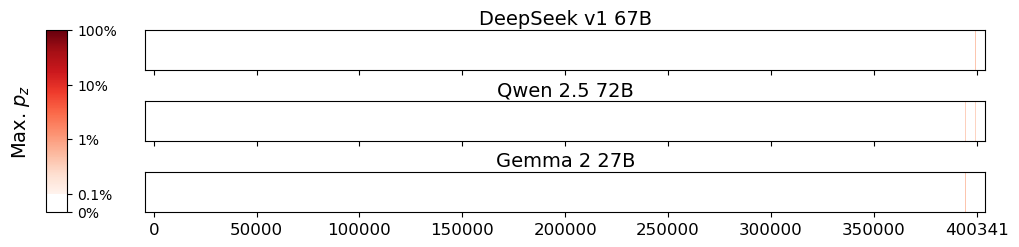}
    \includegraphics[width=\linewidth]{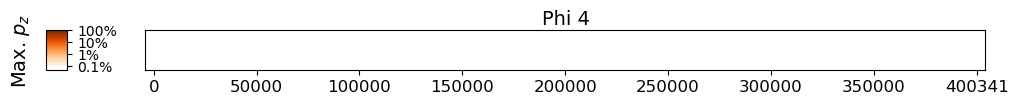}
  \end{minipage}
  \hfill
  \begin{minipage}[t]{0.45\textwidth}
    \centering
    \vspace{0cm}
    \includegraphics[width=\linewidth]{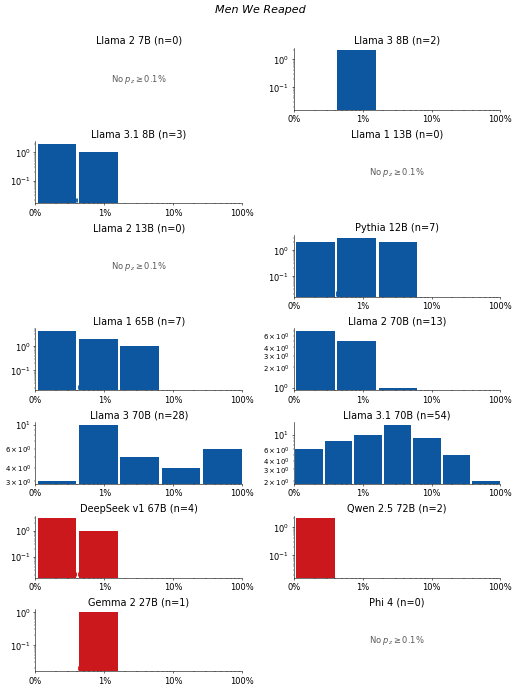}
  \end{minipage}
  \vspace{-.2cm}
  \caption{
    \textbf{\textit{Men We Reaped}, \citeauthor{Men_We_Reaped}.}
    For $14$ LLMs,
    (\textbf{left}) heatmaps for the sliding-window procedure and
    (\textbf{right}) corresponding distributions over suffix extraction probabilities
    ($\tau_\text{min}=0.1\%$).
  }
  \label{fig:slidingwindow:Men_We_Reaped}
\end{figure}
\FloatBarrier

\subsubsection{\textit{Charlotte's Web}, \citeauthor{Charlotte_s_Web}}\label{app:sec:sliding:Charlotte_s_Web}
\vspace{-.2cm}
\begin{figure}[h]
  \centering
  \begin{minipage}[t]{0.53\textwidth}
    \centering
    \vspace{0cm}
    \includegraphics[width=\linewidth]{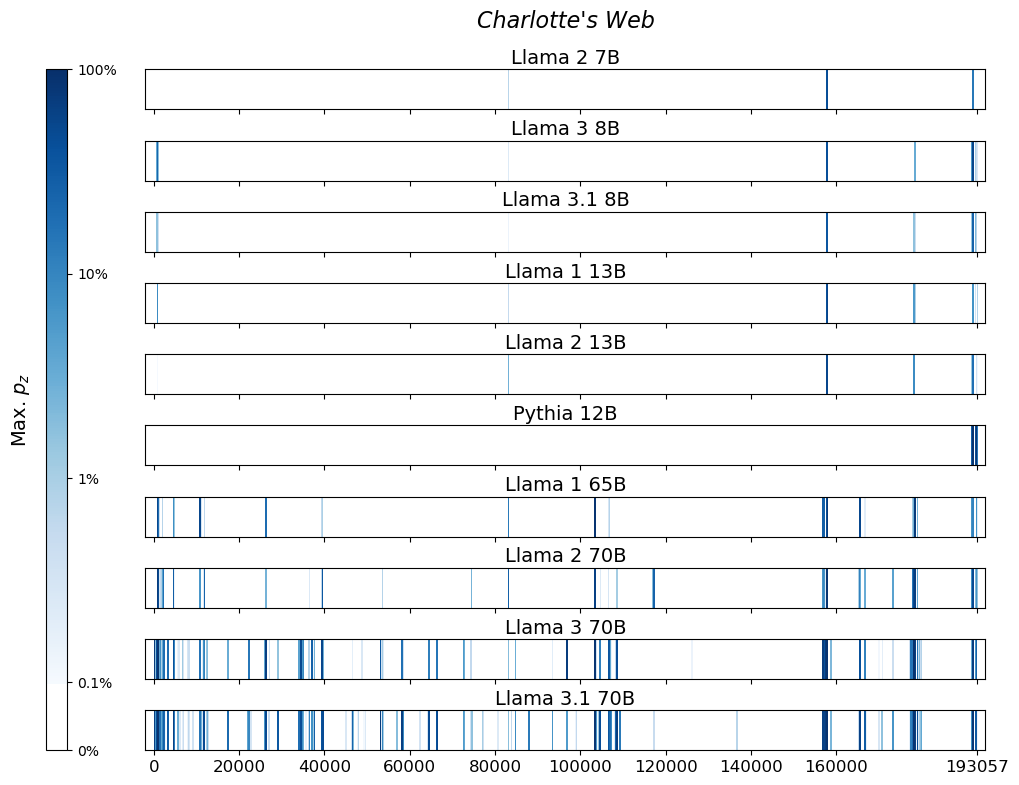}
    \includegraphics[width=\linewidth]{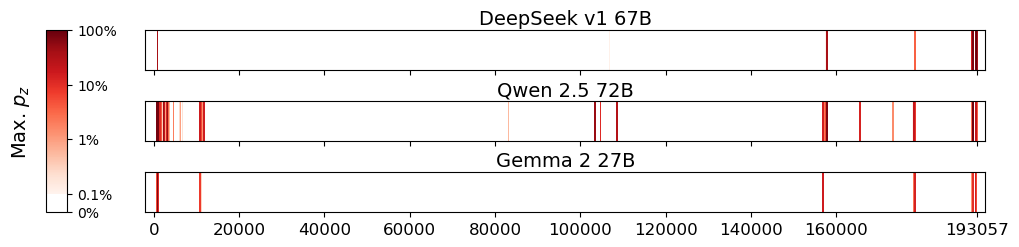}
    \includegraphics[width=\linewidth]{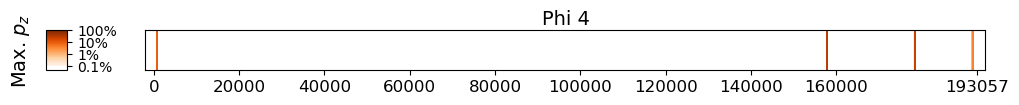}
  \end{minipage}
  \hfill
  \begin{minipage}[t]{0.45\textwidth}
    \centering
    \vspace{0cm}
    \includegraphics[width=\linewidth]{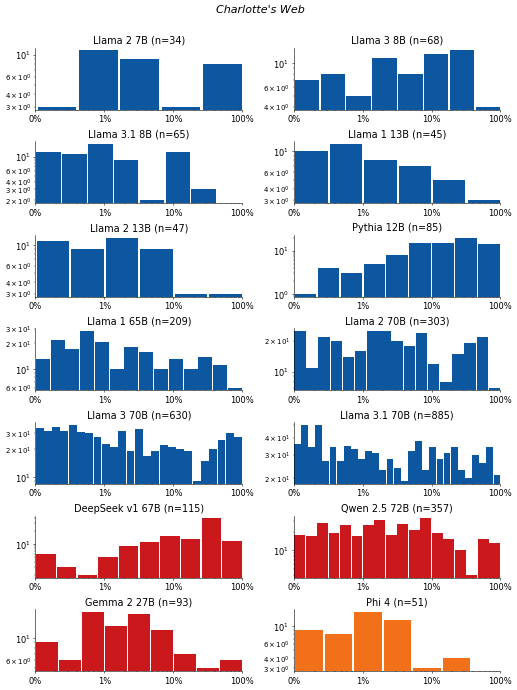}
  \end{minipage}
  \vspace{-.2cm}
  \caption{
    \textbf{\textit{Charlotte's Web}, \citeauthor{Charlotte_s_Web}.}
    For $14$ LLMs,
    (\textbf{left}) heatmaps for the sliding-window procedure and
    (\textbf{right}) corresponding distributions over suffix extraction probabilities
    ($\tau_\text{min}=0.1\%$).
  }
  \label{fig:slidingwindow:Charlotte_s_Web}
\end{figure}
\FloatBarrier

\clearpage
\subsubsection{\textit{A Return to Love}, \citeauthor{A_Return_to_Love}}\label{app:sec:sliding:A_Return_to_Love}
\vspace{-.2cm}
\begin{figure}[h]
  \centering
  \begin{minipage}[t]{0.53\textwidth}
    \centering
    \vspace{0cm}
    \includegraphics[width=\linewidth]{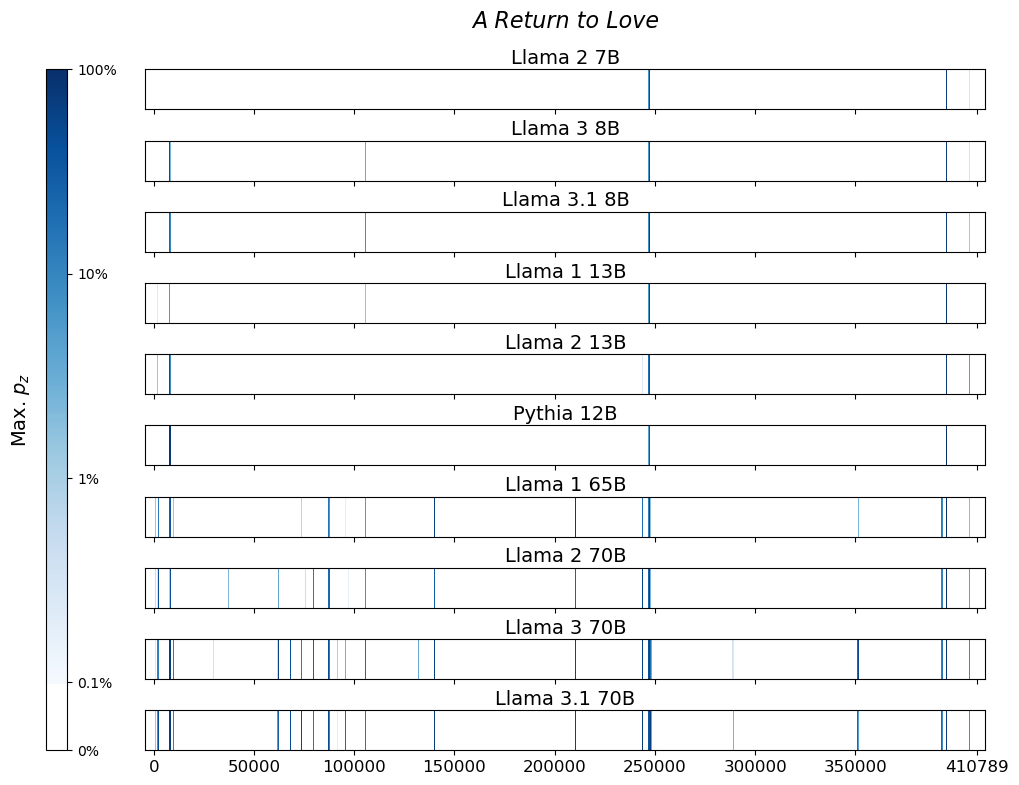}
    \includegraphics[width=\linewidth]{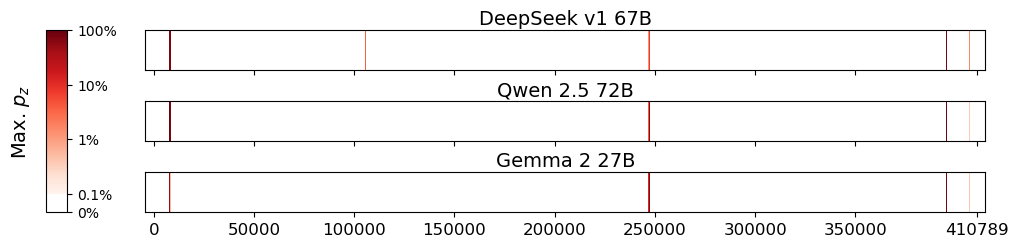}
    \includegraphics[width=\linewidth]{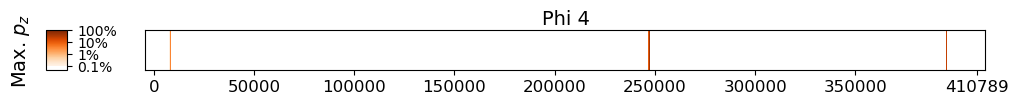}
  \end{minipage}
  \hfill
  \begin{minipage}[t]{0.45\textwidth}
    \centering
    \vspace{0cm}
    \includegraphics[width=\linewidth]{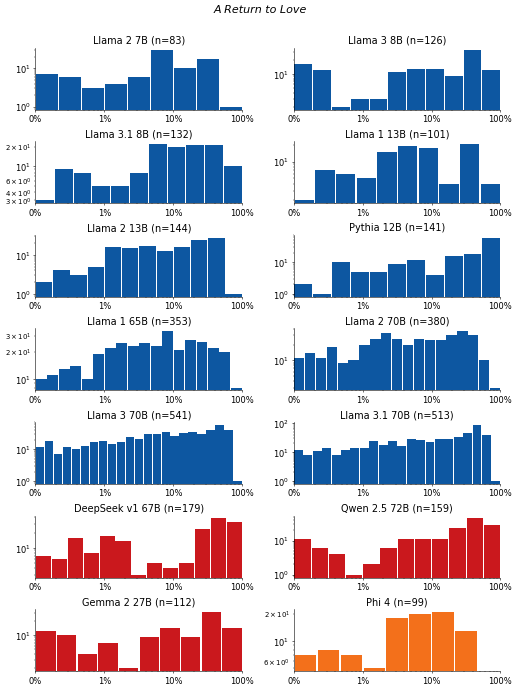}
  \end{minipage}
  \vspace{-.2cm}
  \caption{
    \textbf{\textit{A Return to Love}, \citeauthor{A_Return_to_Love}.}
    For $14$ LLMs,
    (\textbf{left}) heatmaps for the sliding-window procedure and
    (\textbf{right}) corresponding distributions over suffix extraction probabilities
    ($\tau_\text{min}=0.1\%$).
  }
  \label{fig:slidingwindow:A_Return_to_Love}
\end{figure}
\FloatBarrier

\subsubsection{\textit{Another Brooklyn}, \citeauthor{Another_Brooklyn}}\label{app:sec:sliding:Another_Brooklyn}
\vspace{-.2cm}
\begin{figure}[h]
  \centering
  \begin{minipage}[t]{0.53\textwidth}
    \centering
    \vspace{0cm}
    \includegraphics[width=\linewidth]{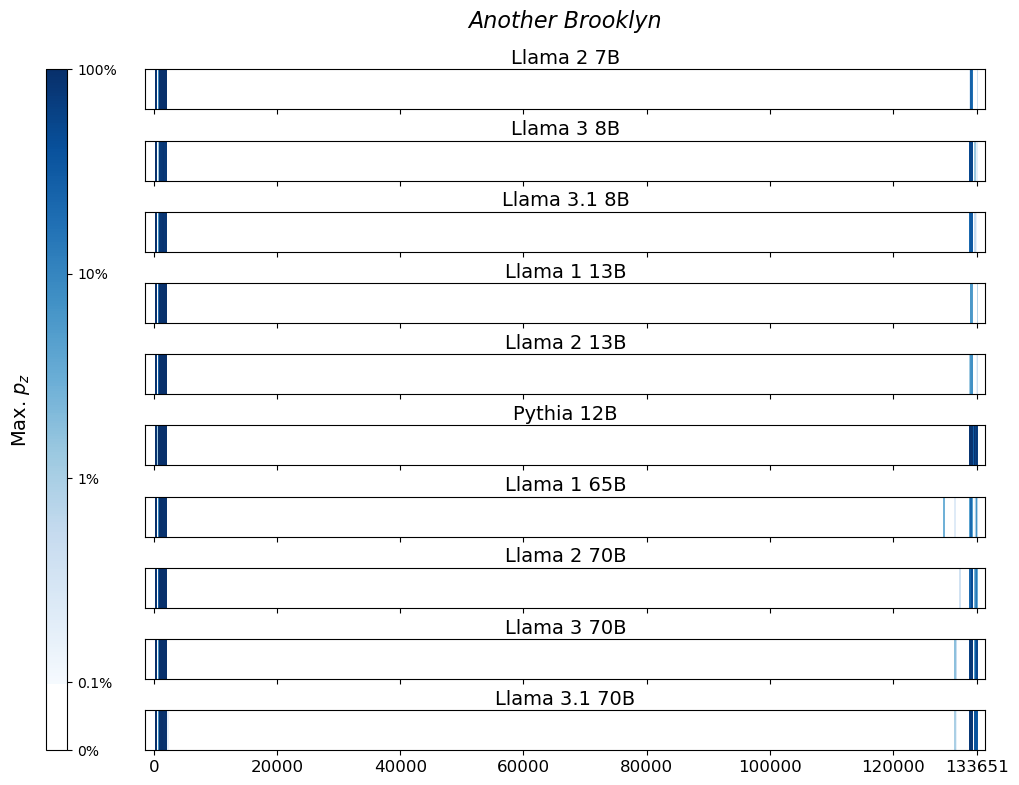}
    \includegraphics[width=\linewidth]{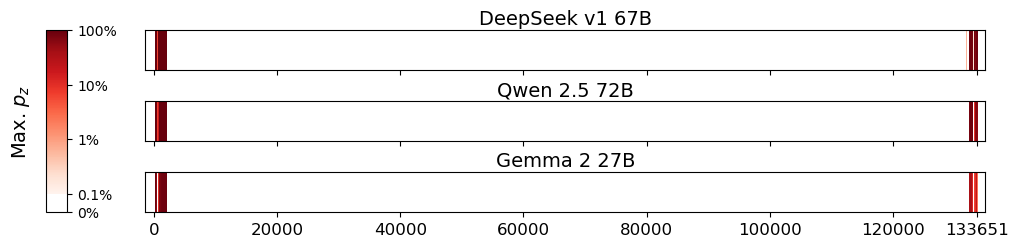}
    \includegraphics[width=\linewidth]{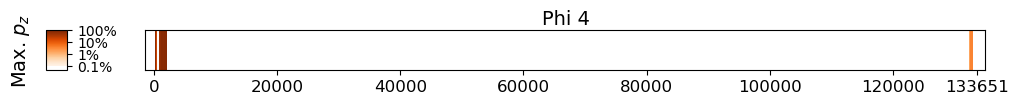}
  \end{minipage}
  \hfill
  \begin{minipage}[t]{0.45\textwidth}
    \centering
    \vspace{0cm}
    \includegraphics[width=\linewidth]{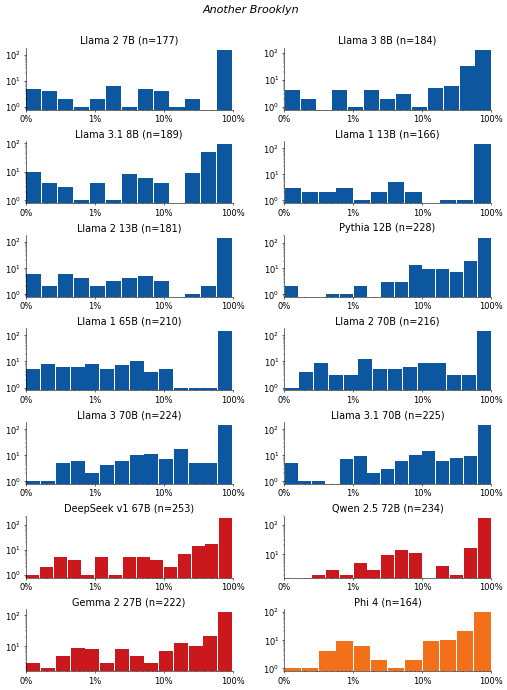}
  \end{minipage}
  \vspace{-.2cm}
  \caption{
    \textbf{\textit{Another Brooklyn}, \citeauthor{Another_Brooklyn}.}
    For $14$ LLMs,
    (\textbf{left}) heatmaps for the sliding-window procedure and
    (\textbf{right}) corresponding distributions over suffix extraction probabilities
    ($\tau_\text{min}=0.1\%$).
  }
  \label{fig:slidingwindow:Another_Brooklyn}
\end{figure}
\FloatBarrier

\clearpage
\subsubsection{\textit{Brown Girl Dreaming}, \citeauthor{Brown_Girl_Dreaming}}\label{app:sec:sliding:Brown_Girl_Dreaming}
\vspace{-.2cm}
\begin{figure}[h]
  \centering
  \begin{minipage}[t]{0.53\textwidth}
    \centering
    \vspace{0cm}
    \includegraphics[width=\linewidth]{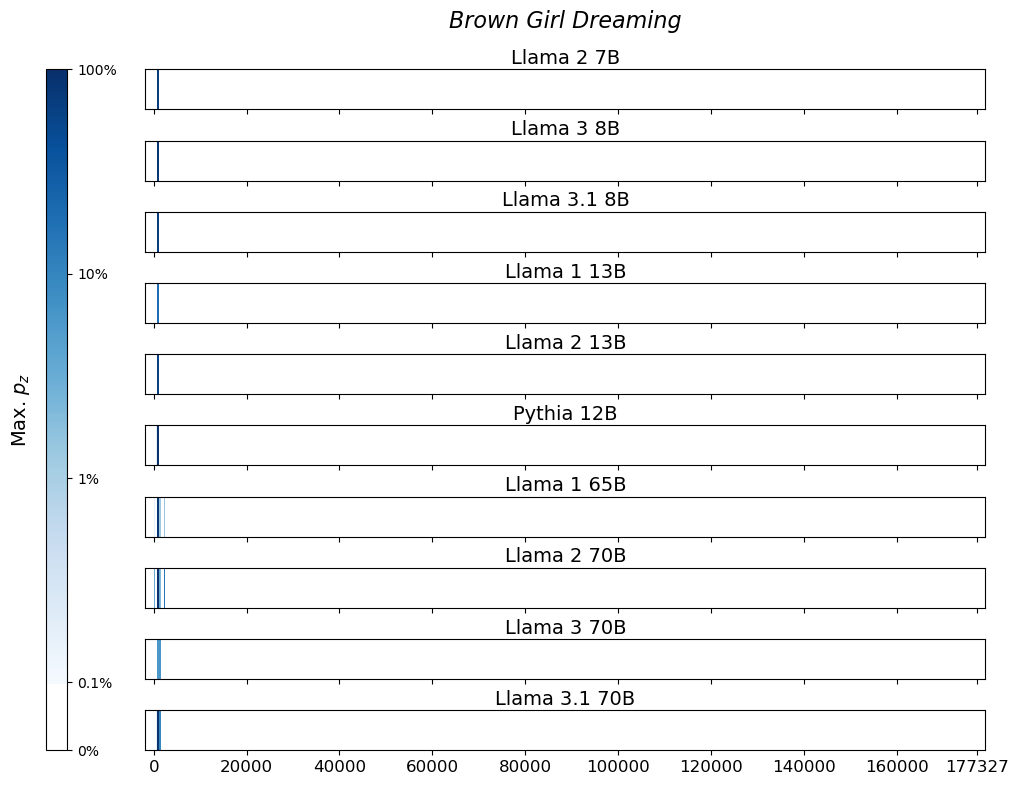}
    \includegraphics[width=\linewidth]{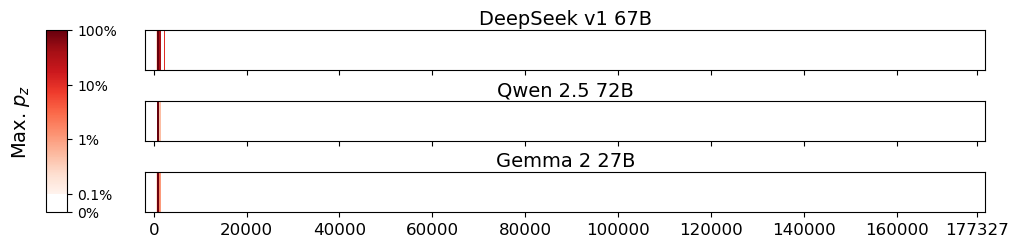}
    \includegraphics[width=\linewidth]{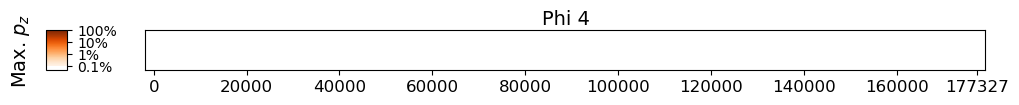}
  \end{minipage}
  \hfill
  \begin{minipage}[t]{0.45\textwidth}
    \centering
    \vspace{0cm}
    \includegraphics[width=\linewidth]{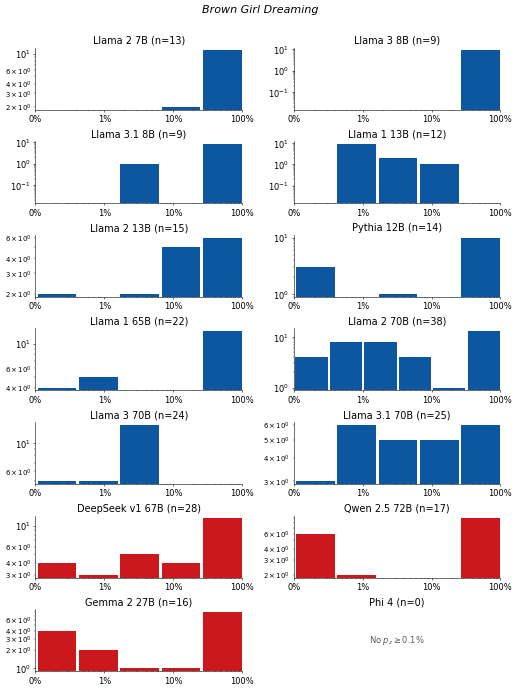}
  \end{minipage}
  \vspace{-.2cm}
  \caption{
    \textbf{\textit{Brown Girl Dreaming}, \citeauthor{Brown_Girl_Dreaming}.}
    For $14$ LLMs,
    (\textbf{left}) heatmaps for the sliding-window procedure and
    (\textbf{right}) corresponding distributions over suffix extraction probabilities
    ($\tau_\text{min}=0.1\%$).
  }
  \label{fig:slidingwindow:Brown_Girl_Dreaming}
\end{figure}
\FloatBarrier

\subsubsection{\textit{A Little Life}, \citeauthor{A_Little_Life}}\label{app:sec:sliding:A_Little_Life}
\vspace{-.2cm}
\begin{figure}[h]
  \centering
  \begin{minipage}[t]{0.53\textwidth}
    \centering
    \vspace{0cm}
    \includegraphics[width=\linewidth]{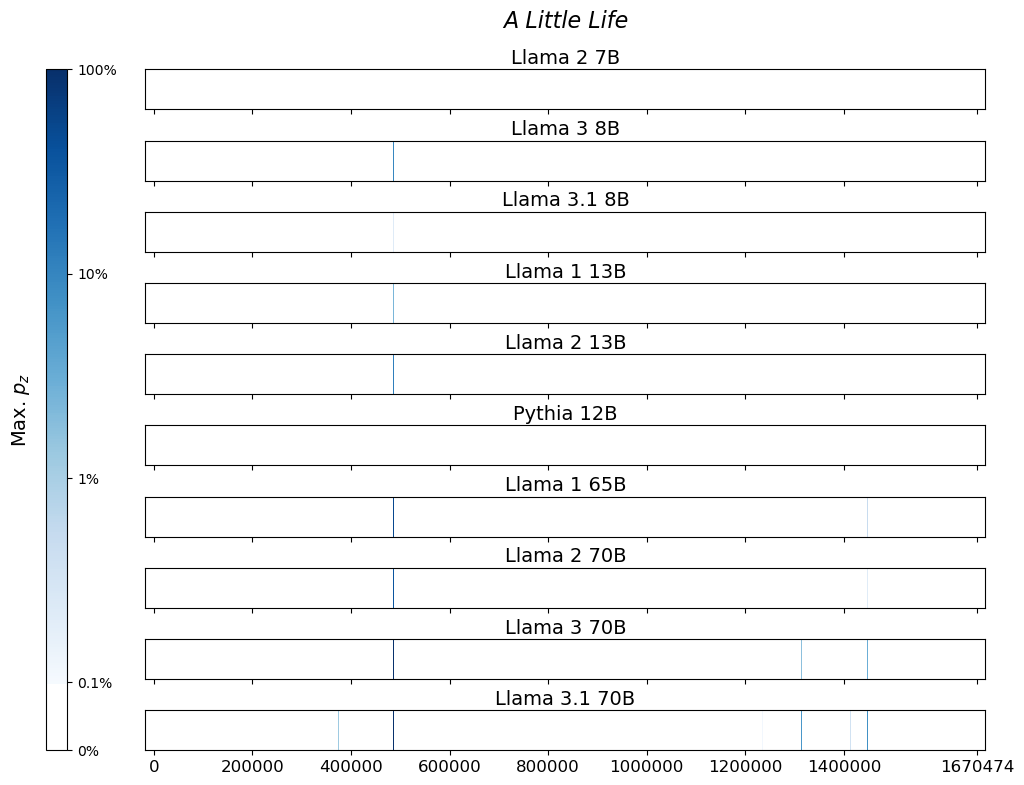}
    \includegraphics[width=\linewidth]{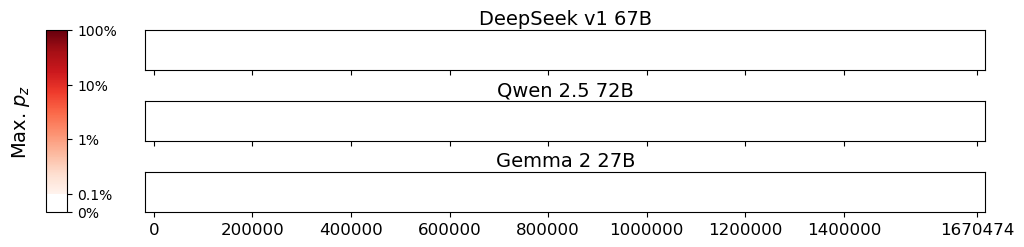}
    \includegraphics[width=\linewidth]{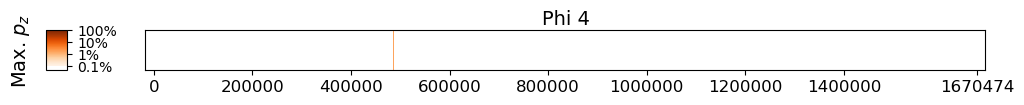}
  \end{minipage}
  \hfill
  \begin{minipage}[t]{0.45\textwidth}
    \centering
    \vspace{0cm}
    \includegraphics[width=\linewidth]{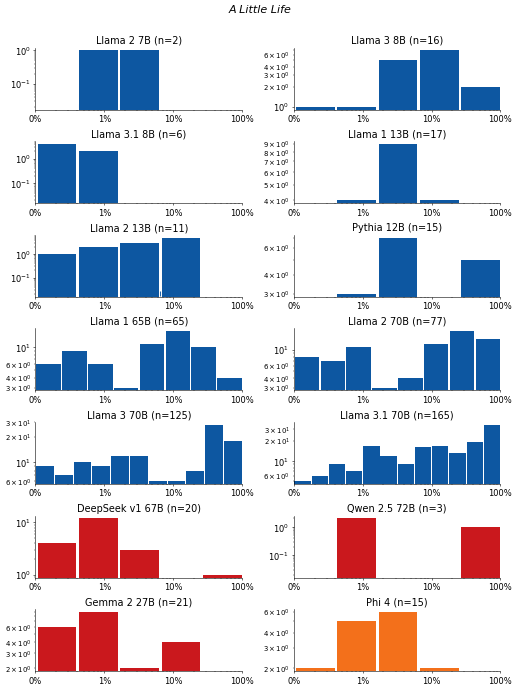}
  \end{minipage}
  \vspace{-.2cm}
  \caption{
    \textbf{\textit{A Little Life}, \citeauthor{A_Little_Life}.}
    For $14$ LLMs,
    (\textbf{left}) heatmaps for the sliding-window procedure and
    (\textbf{right}) corresponding distributions over suffix extraction probabilities
    ($\tau_\text{min}=0.1\%$).
  }
  \label{fig:slidingwindow:A_Little_Life}
\end{figure}
\FloatBarrier

\clearpage
\subsubsection{\textit{The Art of Bonsai}, \citeauthor{The_Art_of_Bonsai}}\label{app:sec:sliding:The_Art_of_Bonsai}
\begin{figure}[h]
  \vspace{-.2cm}
  \centering
  \begin{minipage}[t]{0.53\textwidth}
    \centering
    \vspace{0cm}
    \includegraphics[width=\linewidth]{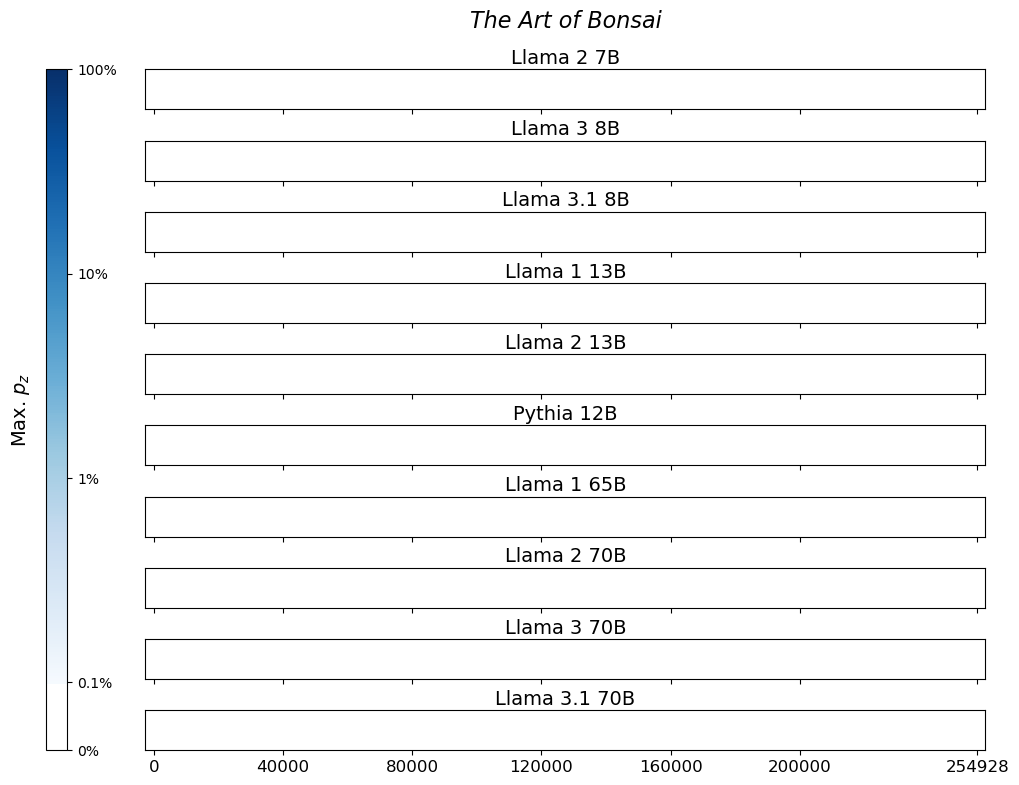}
    \includegraphics[width=\linewidth]{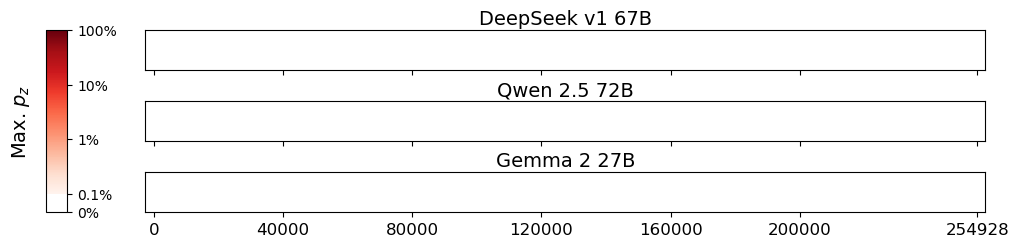}
    \includegraphics[width=\linewidth]{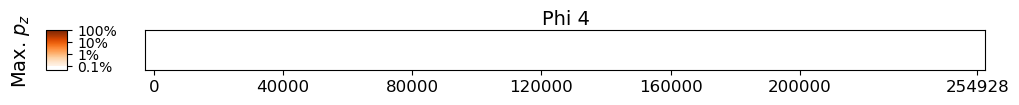}
  \end{minipage}
  \hfill
  \begin{minipage}[t]{0.45\textwidth}
    \centering
    \vspace{0cm}
    \includegraphics[width=\linewidth]{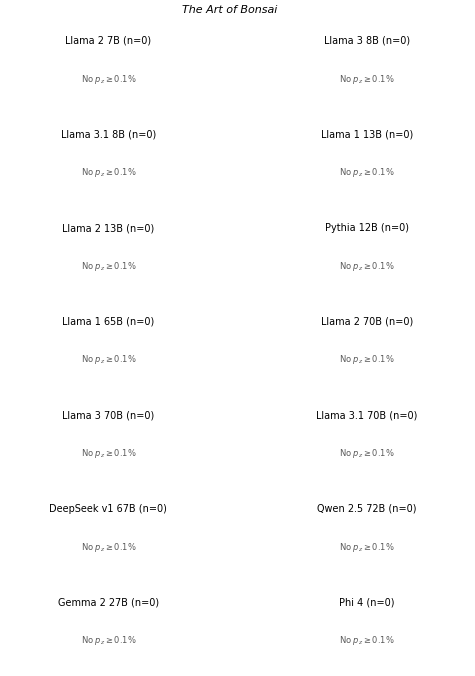}
  \end{minipage}
  \vspace{-.4cm}
  \caption{
    \textbf{\textit{The Art of Bonsai}, \citeauthor{The_Art_of_Bonsai}.}
    For $14$ LLMs,
    (\textbf{left}) heatmaps for the sliding-window procedure and
    (\textbf{right}) corresponding distributions over suffix extraction probabilities
    ($\tau_\text{min}=0.1\%$).
  }
  \label{fig:slidingwindow:The_Art_of_Bonsai}
\end{figure}
\FloatBarrier

\subsubsection{\textit{A People's History of the United States}, \citeauthor{A_People_s_History_of_the_United_States}}\label{app:sec:sliding:A_People_s_History_of_the_United_States}
\begin{figure}[h]
  \vspace{-.2cm}
  \centering
  \begin{minipage}[t]{0.53\textwidth}
    \centering
    \vspace{0cm}
    \includegraphics[width=\linewidth]{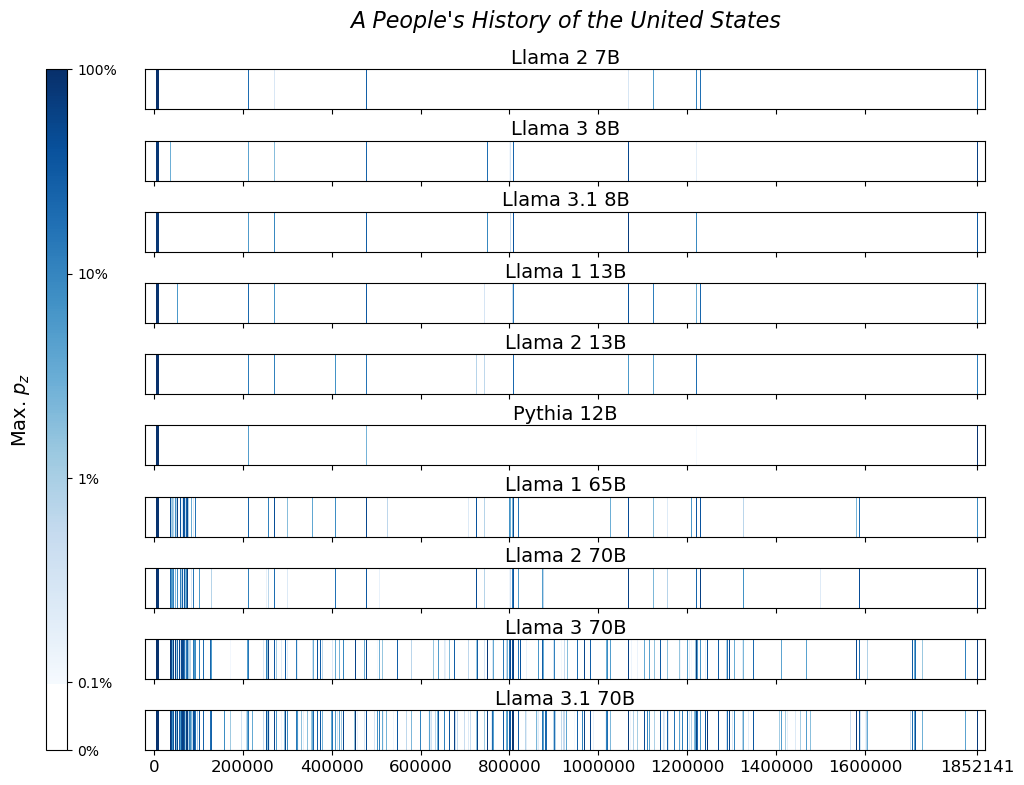}
    \includegraphics[width=\linewidth]{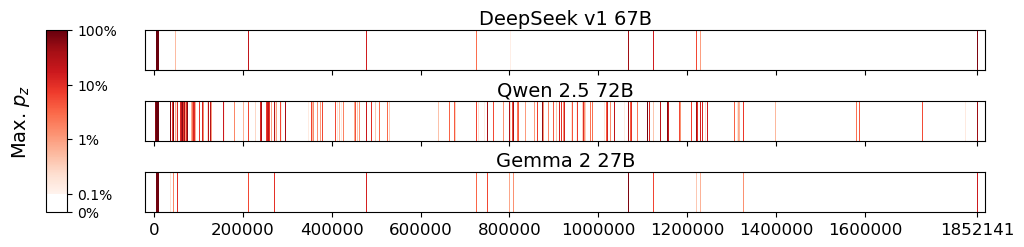}
    \includegraphics[width=\linewidth]{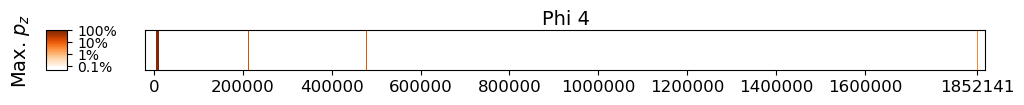}
  \end{minipage}
  \hfill
  \begin{minipage}[t]{0.45\textwidth}
    \centering
    \vspace{0cm}
    \includegraphics[width=\linewidth]{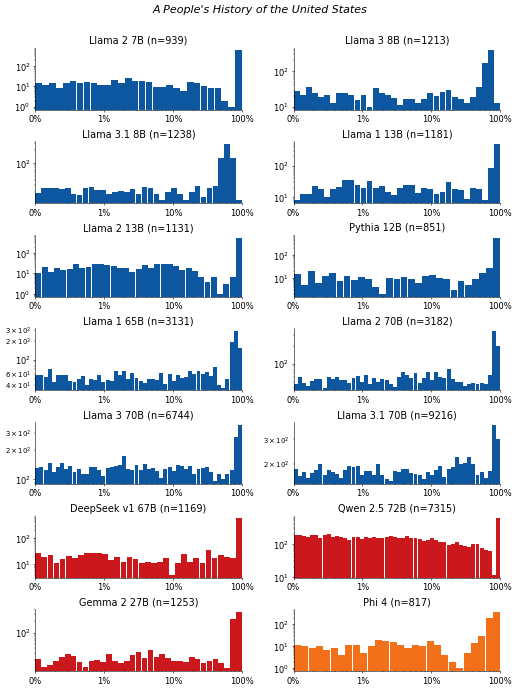}
  \end{minipage}
  \vspace{-.2cm}
  \caption{
    \textbf{\textit{A People's History of the United States}, \citeauthor{A_People_s_History_of_the_United_States}.}
    For $14$ LLMs,
    (\textbf{left}) heatmaps for the sliding-window procedure and
    (\textbf{right}) corresponding distributions over suffix extraction probabilities
    ($\tau_\text{min}=0.1\%$).
  }
  \label{fig:slidingwindow:A_People_s_History_of_the_United_States}
\end{figure}
\FloatBarrier

\clearpage
\subsubsection{\textit{The Future of the Internet and How to Stop It}, \citeauthor{The_Future_of_the_Internet_and_How_to_Stop_It}}\label{app:sec:sliding:The_Future_of_the_Internet_and_How_to_Stop_It}
\vspace{-.2cm}
\begin{figure}[h]
  \centering
  \begin{minipage}[t]{0.53\textwidth}
    \centering
    \vspace{0cm}
    \includegraphics[width=\linewidth]{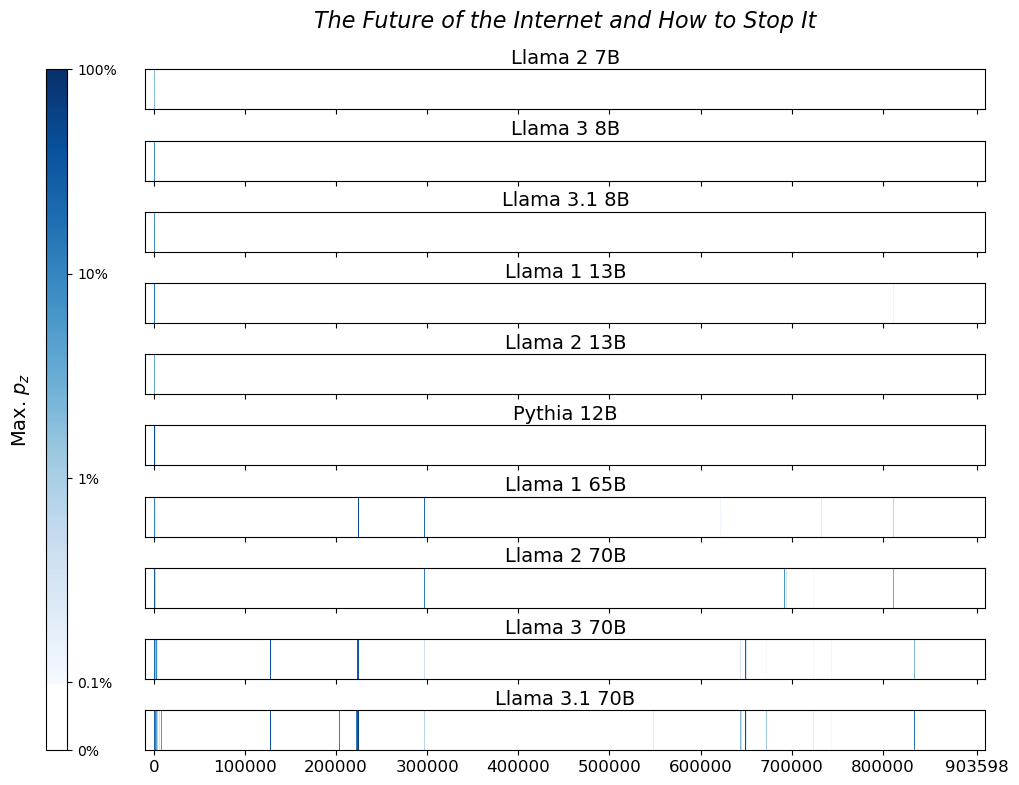}
    \includegraphics[width=\linewidth]{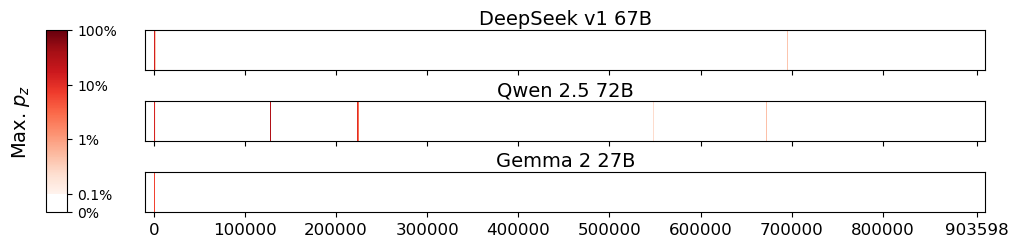}
    \includegraphics[width=\linewidth]{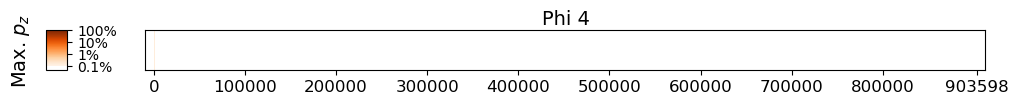}
  \end{minipage}
  \hfill
  \begin{minipage}[t]{0.45\textwidth}
    \centering
    \vspace{0cm}
    \includegraphics[width=\linewidth]{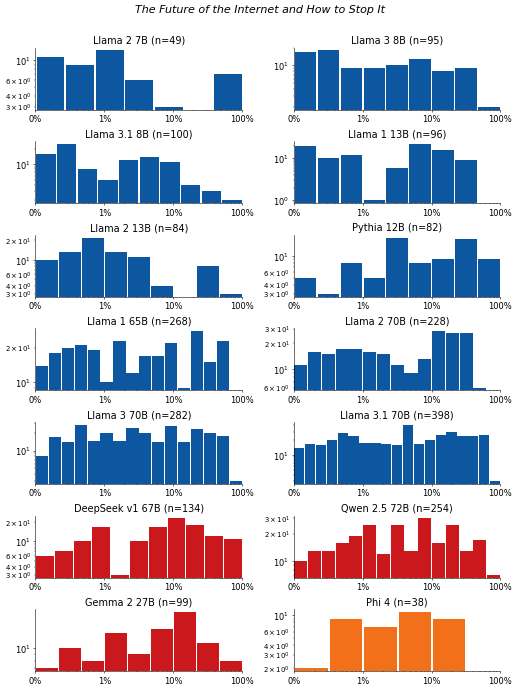}
  \end{minipage}
  \vspace{-.2cm}
  \caption{
    \textbf{\textit{The Future of the Internet and How to Stop It}, \citeauthor{The_Future_of_the_Internet_and_How_to_Stop_It}.}
    For $14$ LLMs,
    (\textbf{left}) heatmaps for the sliding-window procedure and
    (\textbf{right}) corresponding distributions over suffix extraction probabilities
    ($\tau_\text{min}=0.1\%$).
  }
  \label{fig:slidingwindow:The_Future_of_the_Internet_and_How_to_Stop_It}
\end{figure}
\FloatBarrier

\subsubsection{\textit{The Book Thief}, \citeauthor{The_Book_Thief}}\label{app:sec:sliding:The_Book_Thief}
\vspace{-.2cm}
\begin{figure}[h]
  \centering
  \begin{minipage}[t]{0.53\textwidth}
    \centering
    \vspace{0cm}
    \includegraphics[width=\linewidth]{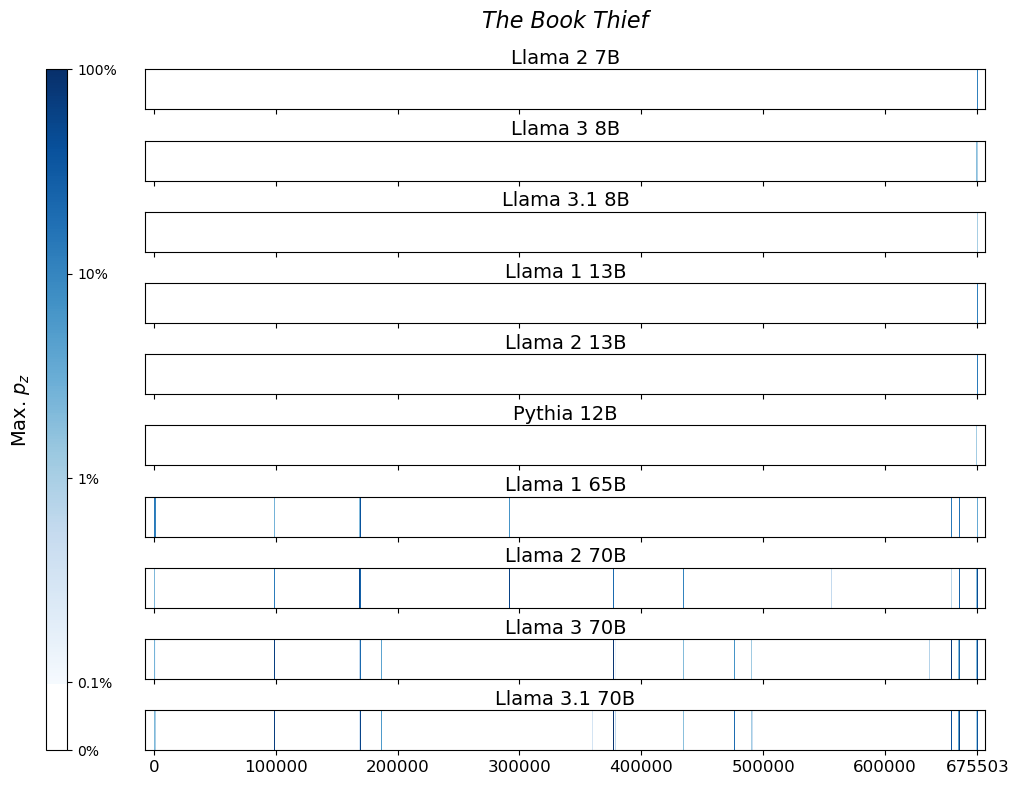}
    \includegraphics[width=\linewidth]{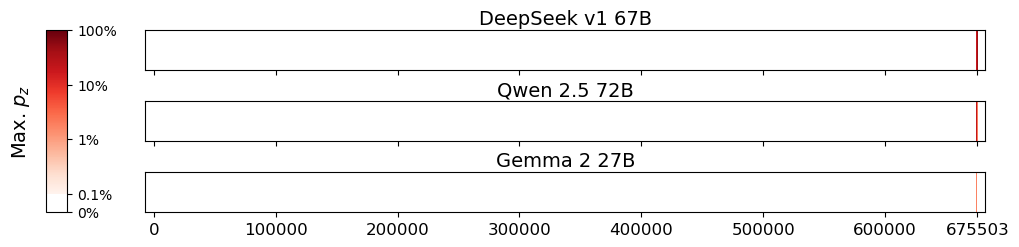}
    \includegraphics[width=\linewidth]{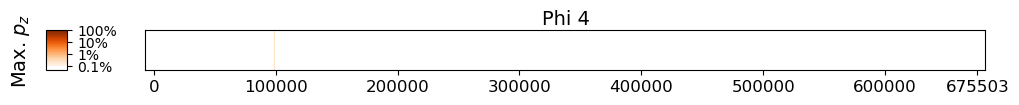}
  \end{minipage}
  \hfill
  \begin{minipage}[t]{0.45\textwidth}
    \centering
    \vspace{0cm}
    \includegraphics[width=\linewidth]{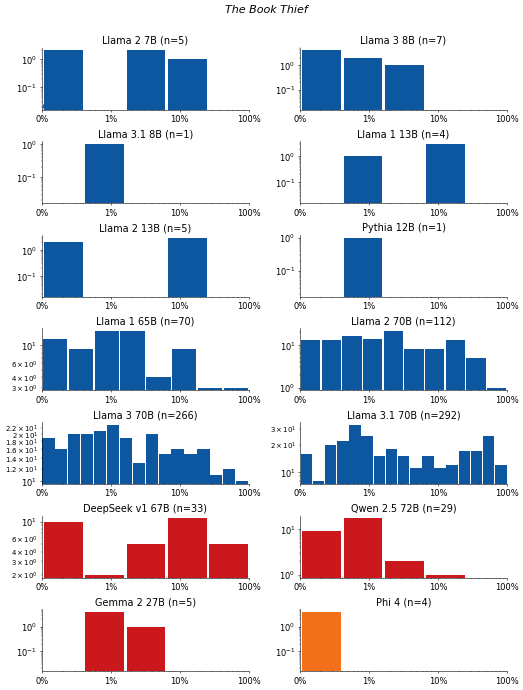}
  \end{minipage}
  \vspace{-.2cm}
  \caption{
    \textbf{\textit{The Book Thief}, \citeauthor{The_Book_Thief}.}
    For $14$ LLMs,
    (\textbf{left}) heatmaps for the sliding-window procedure and
    (\textbf{right}) corresponding distributions over suffix extraction probabilities
    ($\tau_\text{min}=0.1\%$).
  }
  \label{fig:slidingwindow:The_Book_Thief}
\end{figure}
\FloatBarrier

\clearpage
\subsection{Estimating how much of a book is memorized by a model}\label{app:sec:sliding-window:percentage}

\FloatBarrier

We estimate book-specific memorization for a given model $\theta$ and decoding scheme $\phi$ with a metric for \newterm{extraction coverage}.
We define this metric and provide measurements for the different models and books we evaluate.

\subsubsection{Quantifying extraction coverage}\label{app:sec:coverage}

We compute extraction coverage as the fraction of characters in a book that lie within at least one suffix whose extraction probability $p_\vz$ exceeds a threshold $\tau$.

\begin{definition}[\textbf{Extraction coverage}]
\label{app:def:cov}
Let $\mathsf{span}(\vz)$ denote the set of character indices spanned by a $50$-token suffix of a given sequence $\vz \in \sZ(B)$, where $B$ is a book. 
Also let $\tau_\text{min}$ denote the minimum probability we consider to reflect extraction success. 
For book $B$, LLM $\theta$, decoding scheme $\phi$, and probability threshold $\tau \in [\tau_\text{min}, 1]$,\looseness=-1
\begin{align}
\label{app:eq:cov}
\mathsf{coverage}_{B,\theta,\phi}(\tau) 
= \frac{1}{L_B}\,\Bigl|\bigcup_{\vz \in \sZ(B):\, p_{\vz,\theta,\phi} \ge \tau}\, \mathsf{span}(\vz)\Bigr|,
\end{align}

where $L_B$ is the total number of characters in the book.\looseness=-1
\end{definition}

For details on $\tau_\text{min}$, see Equation~\ref{eq:probsuccess} (Sections~\ref{sec:background:pz} \&~\ref{sec:validity}; Appendices~\ref{app:sec:rates} \&~\ref{app:sec:validity}). 
In the main text, we omit $\theta$ and $\phi$ for brevity, since we always specify the exact model and decoding scheme being considered. 
 
The results of our sliding-window procedure can be used to compute Equation~\ref{app:eq:cov}.
This procedure returns a list of $p_\vz$ sorted by start position in the book (i.e., $\sZ(B)$). 
We pick a threshold $\tau$ and filter the list of $p_\vz$ to only contain sequences whose extraction probability is $\geq\!\tau$.
Our sliding-window procedure keeps track of character-level metadata so that we can visualize heatmaps; 
we have information about the start/end character indexes for the prefixes and suffixes. 
Based on these intervals, we can compute which characters in the whole book are covered with $p_\vz\!\geq\!\tau$, and then divide that by the total number of characters in the book.\looseness=-1 

High extraction coverage does not necessarily mean we can extract the whole book in one go, or that we have identified the minimal set of sequences to try to extract to get maximal coverage of the book. 
This is simply a procedure for estimating the proportion of total memorized text for a book, in order to do cross-book and cross-model comparisons, as in Section~\ref{sec:book-procedure:averages}.\looseness=-1

\subsubsection{Extended extraction coverage results}\label{app:sec:coverage-results}

We display extraction coverage in two ways, which support different types of comparisons.\looseness=-1 
\begin{enumerate}[leftmargin=0.65cm]
    \item \textbf{Model-level histograms.} 
    For each model, we plot the distribution of extraction coverage across the $200$ evaluated books. 
    This provides a high-level sense of how coverage varies across books, and enables comparisons of this distribution across models. 
    For example, in the main paper we show two histograms (for \textsc{Llama 1 65B} and \textsc{Llama 3.1 70B}) at $\tau\!=\!0.1\%$, the minimum extraction probability that we validate in this work.  
    These results indicate that most books have little extraction coverage. 
    But there are clear exceptions. 
    They also reveal that some models exhibit substantially higher coverage than others---in this case, \textsc{Llama 3.1 70B} shows higher maximum coverage and more books with significant coverage, i.e., it has memorized much more book text than \textsc{Llama 1 65B}.
    
    \item \textbf{Book-level percentages.} 
    For each book, we compare extraction coverage across different models. 
    This not only indicates how memorized a given book is by a particular model, but also highlights variation in memorization of a given book across models. 
    We provide examples of this comparison in the table in Figure~\ref{fig:rates} (right), for \emph{Harry Potter and the Sorcerer's Stone} and \emph{Sandman Slim}, evaluated with \textsc{Llama 3.1 70B}, \textsc{Llama 1 65B}, and \textsc{Pythia 12B}.
\end{enumerate}

\begin{figure*}[t]
\centering
\begin{minipage}[t]{0.4\textwidth}\centering
  \includegraphics[width=\linewidth]{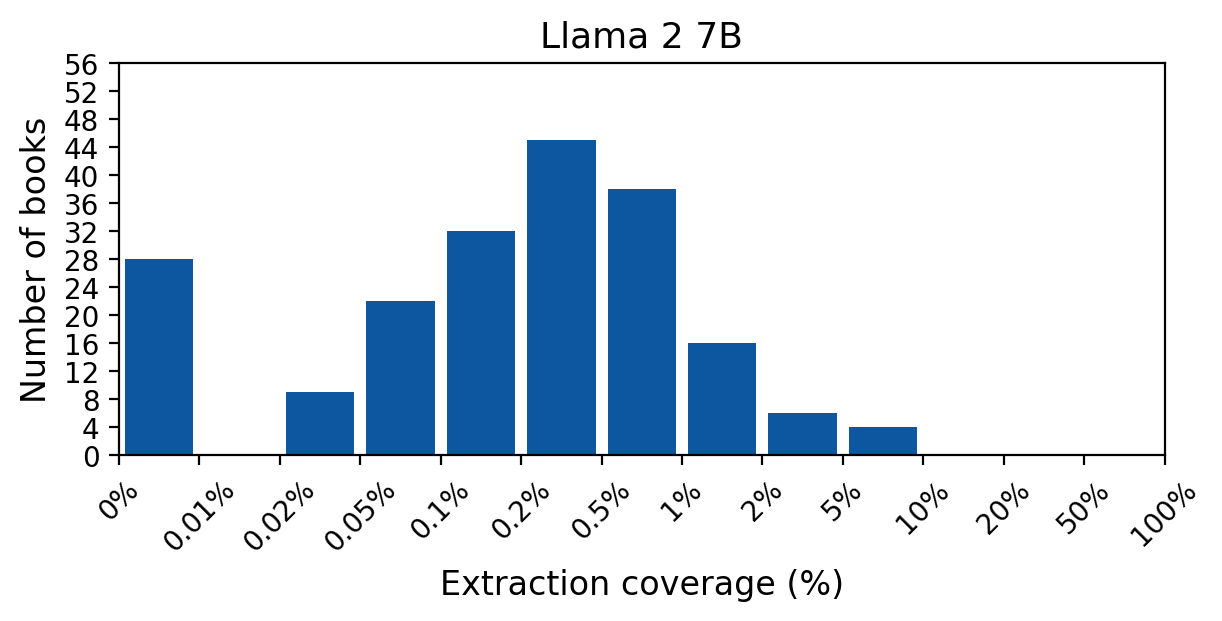}
\end{minipage}\hfill
\begin{minipage}[t]{0.4\textwidth}\centering
  \includegraphics[width=\linewidth]{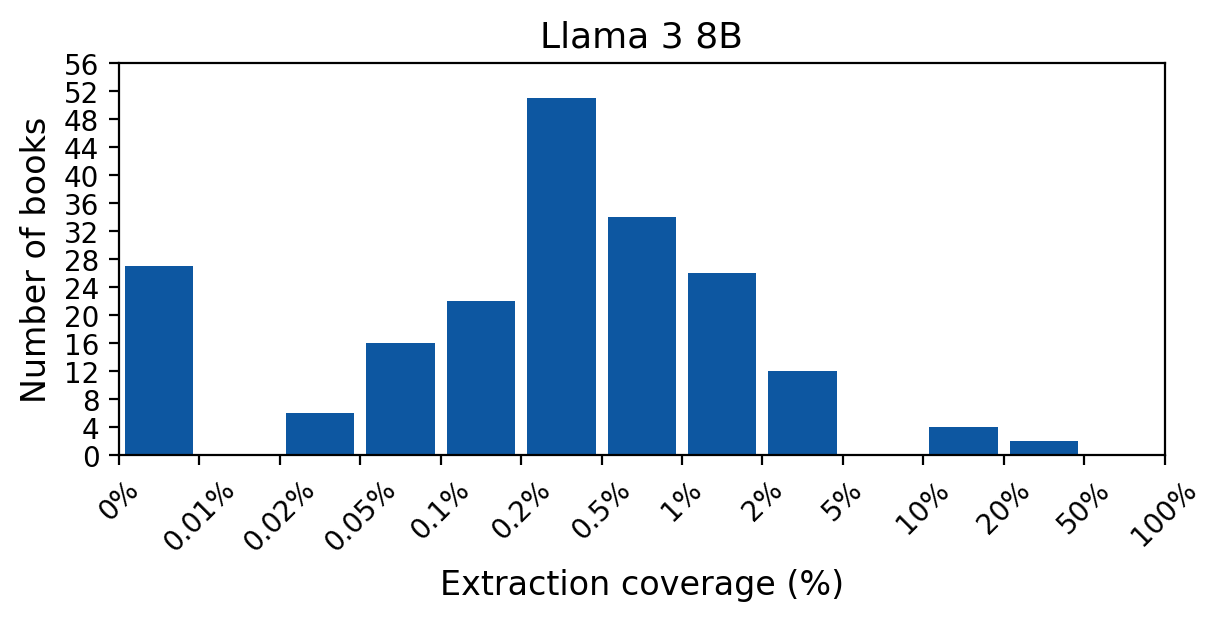}
\end{minipage}
\vspace{0.1cm}
\begin{minipage}[t]{0.4\textwidth}\centering
  \includegraphics[width=\linewidth]{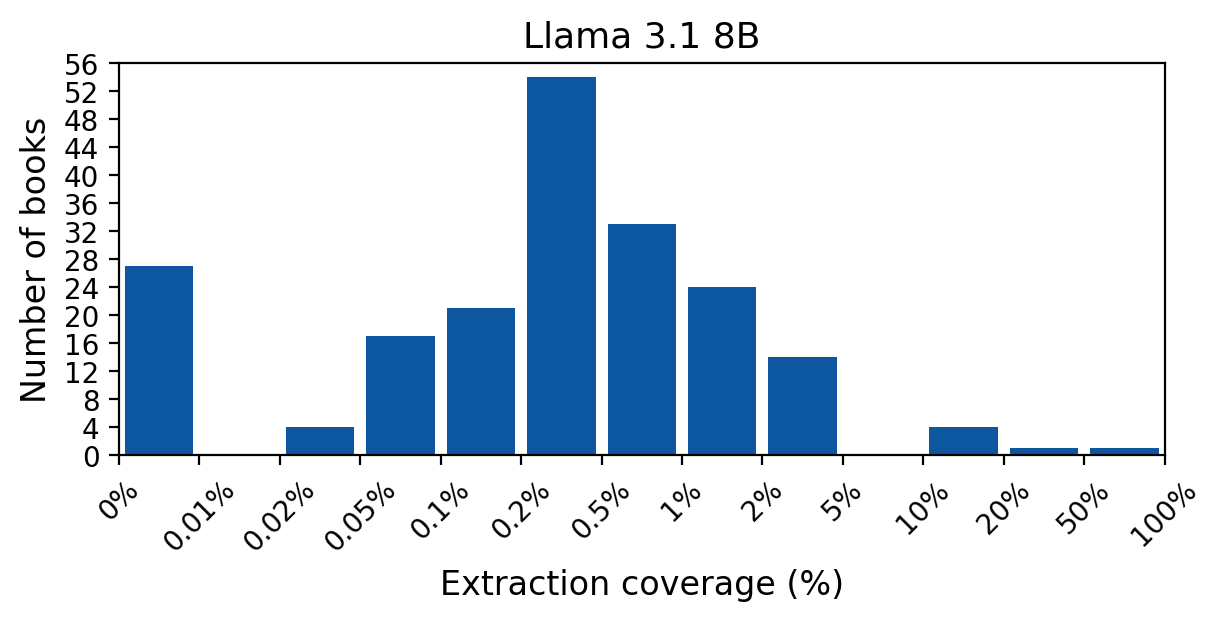}
\end{minipage}\hfill
\begin{minipage}[t]{0.4\textwidth}\centering
  \includegraphics[width=\linewidth]{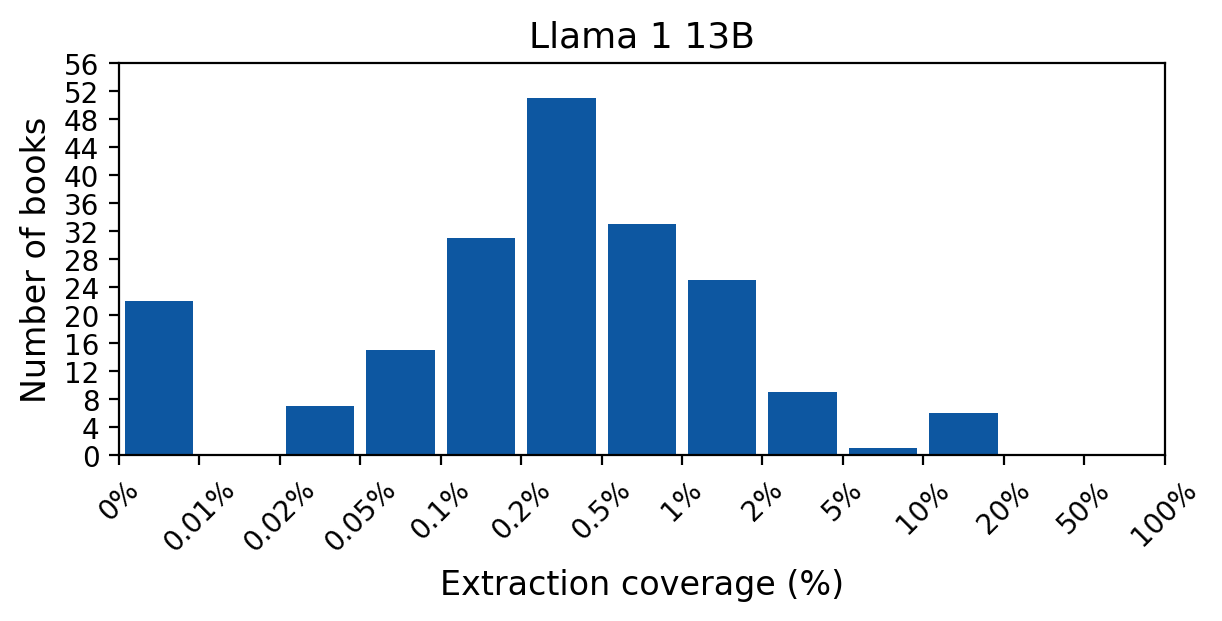}
\end{minipage}
\vspace{0.1cm}
\begin{minipage}[t]{0.4\textwidth}\centering
  \includegraphics[width=\linewidth]{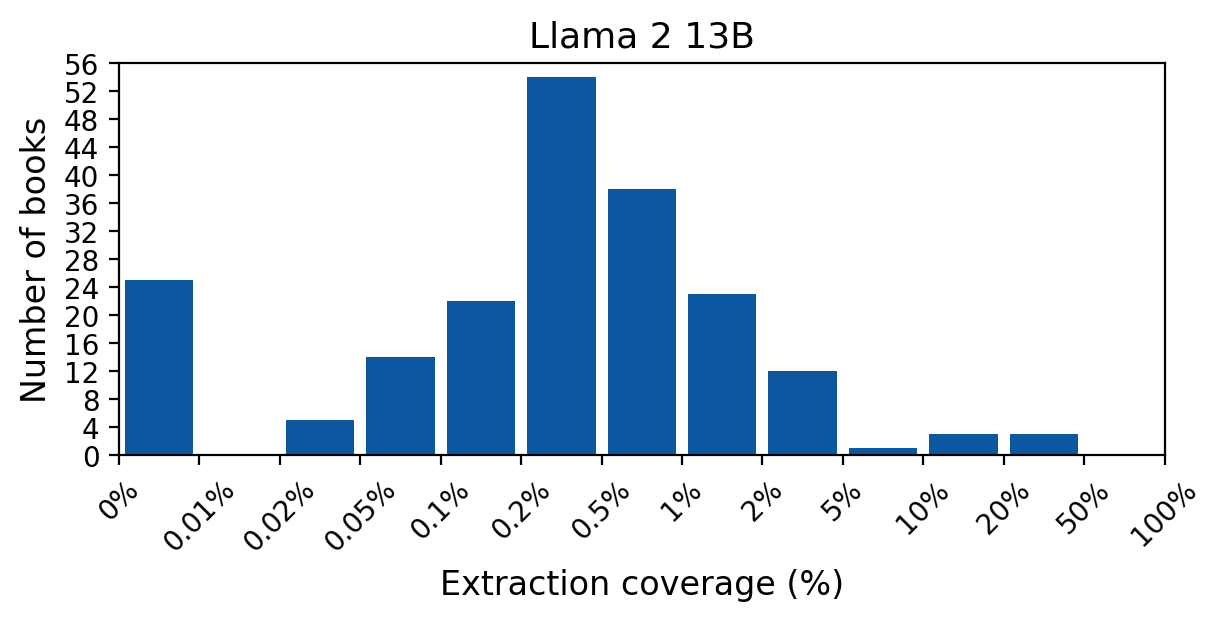}
\end{minipage}\hfill
\begin{minipage}[t]{0.4\textwidth}\centering
  \includegraphics[width=\linewidth]{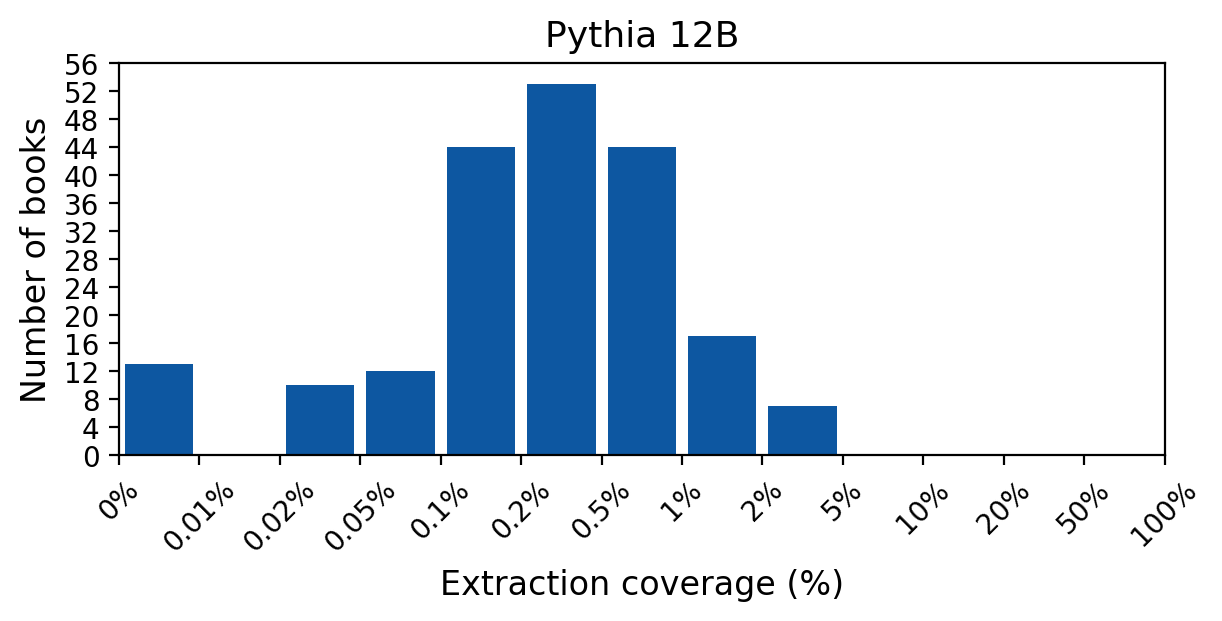}
\end{minipage}
\vspace{0.1cm}
\begin{minipage}[t]{0.4\textwidth}\centering
  \includegraphics[width=\linewidth]{figure/coverage_histograms/coverage-hist-huggyllama-65b_manualbins.png}
\end{minipage}\hfill
\begin{minipage}[t]{0.4\textwidth}\centering
  \includegraphics[width=\linewidth]{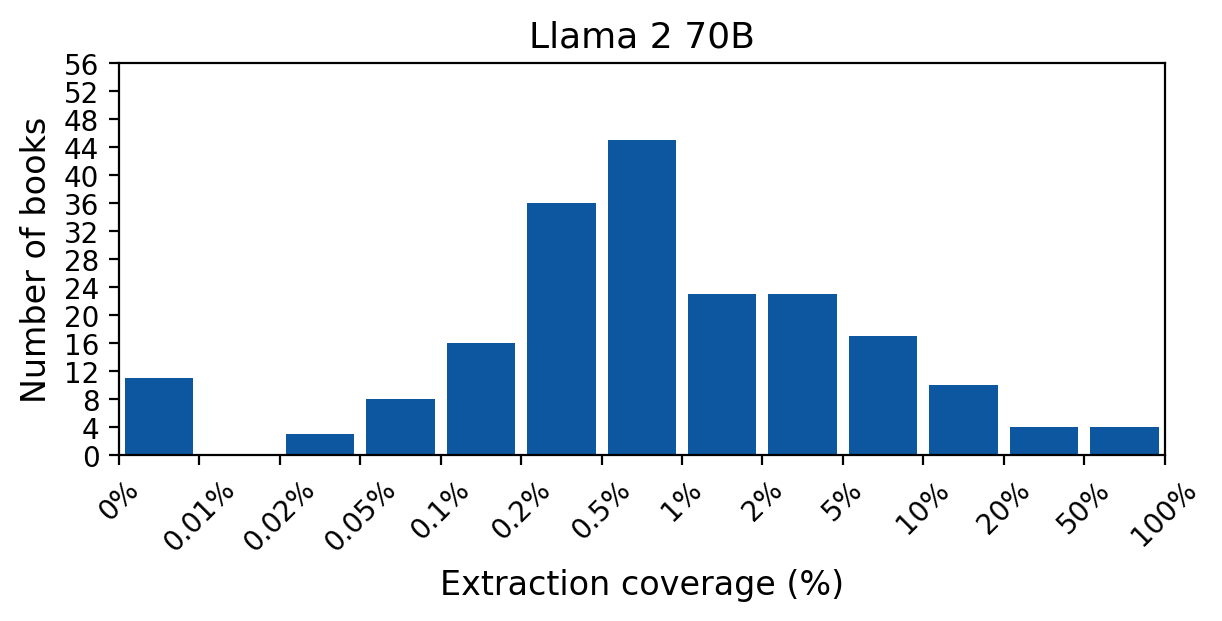}
\end{minipage}
\vspace{0.1cm}
\begin{minipage}[t]{0.4\textwidth}\centering
  \includegraphics[width=\linewidth]{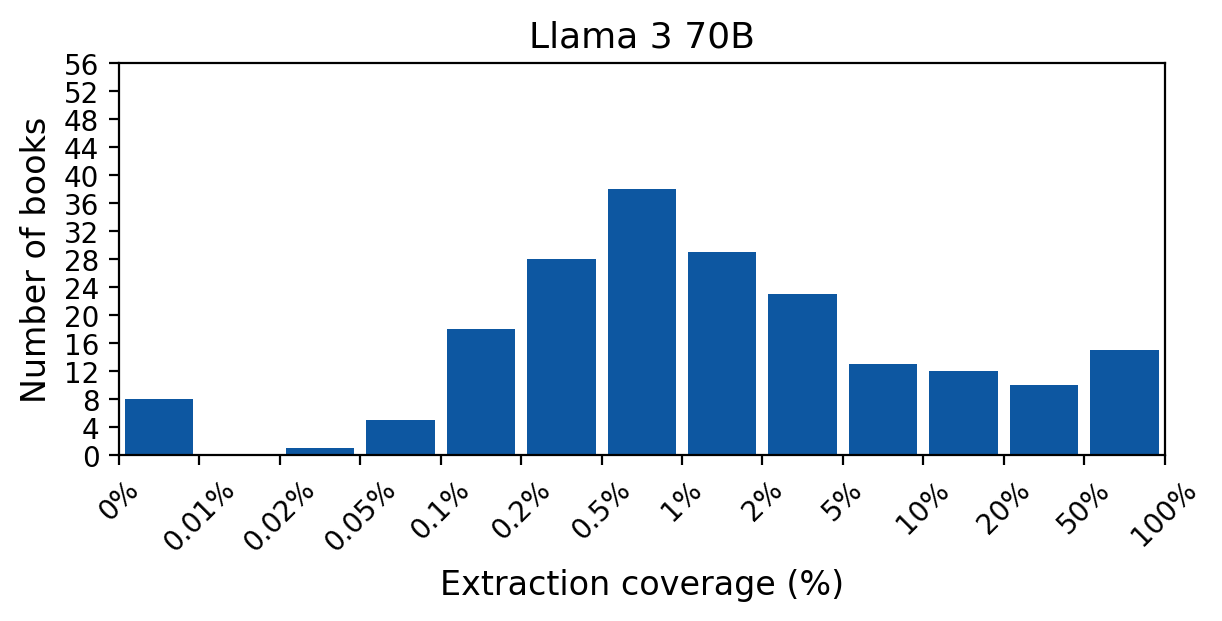}
\end{minipage}\hfill
\begin{minipage}[t]{0.4\textwidth}\centering
  \includegraphics[width=\linewidth]{figure/coverage_histograms/coverage-hist-Llama-3.1-70b_manualbins.png}
\end{minipage}
\vspace{0.1cm}
\begin{minipage}[t]{0.4\textwidth}\centering
  \includegraphics[width=\linewidth]{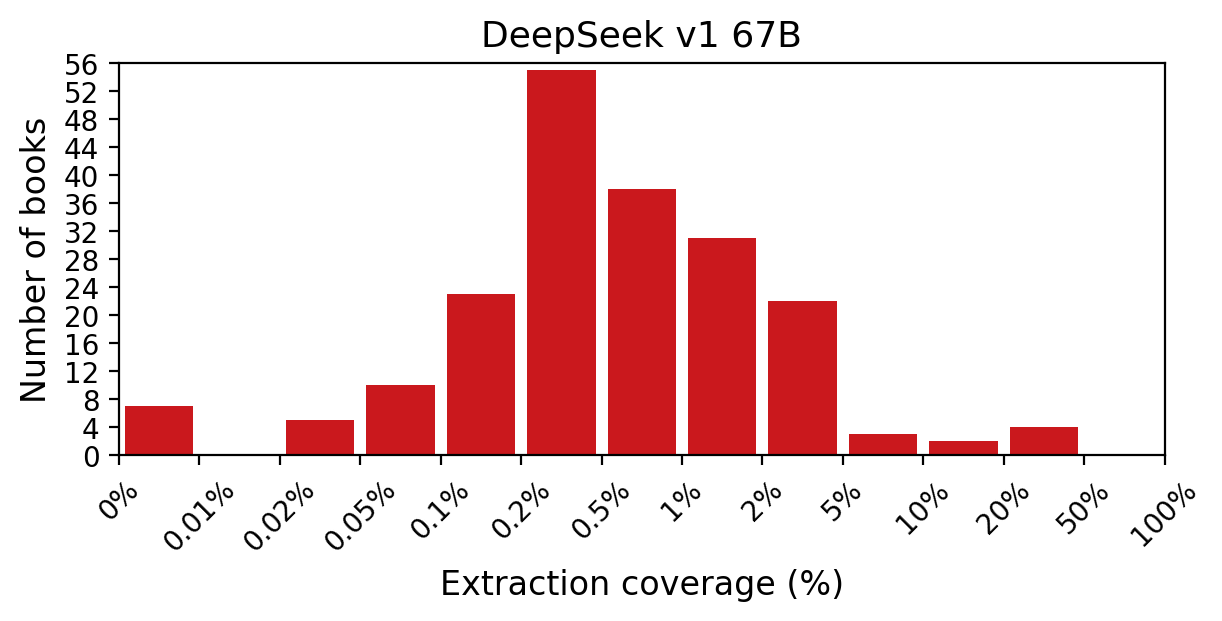}
\end{minipage}\hfill
\begin{minipage}[t]{0.4\textwidth}\centering
  \includegraphics[width=\linewidth]{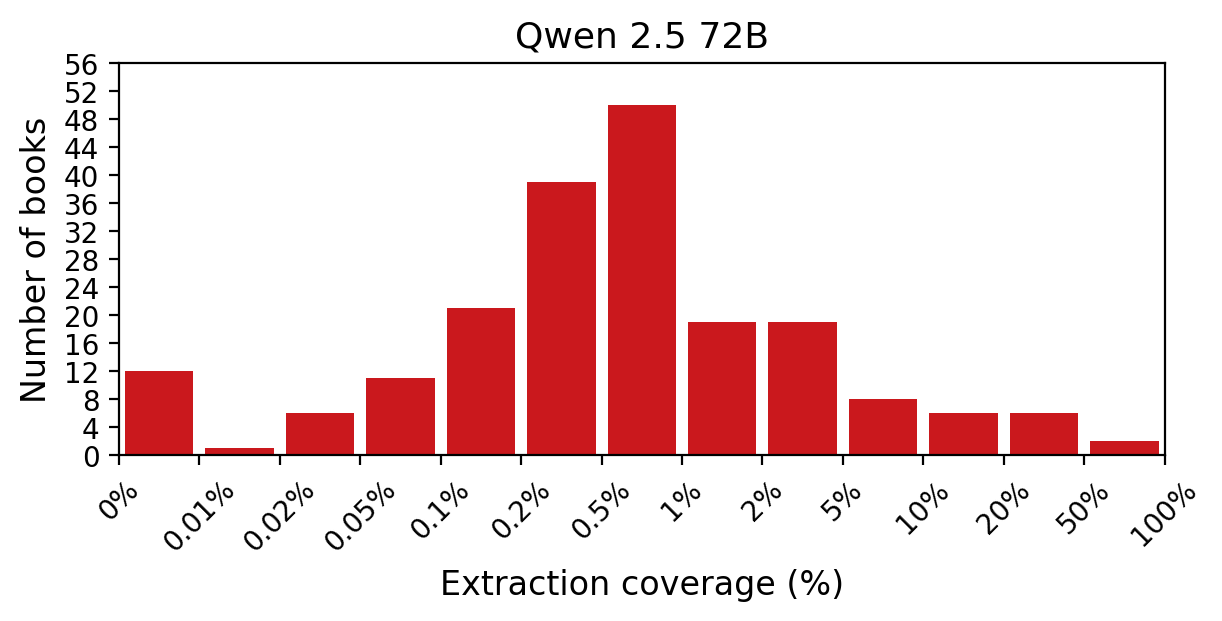}
\end{minipage}
\vspace{0.1cm}
\begin{minipage}[t]{0.4\textwidth}\centering
  \includegraphics[width=\linewidth]{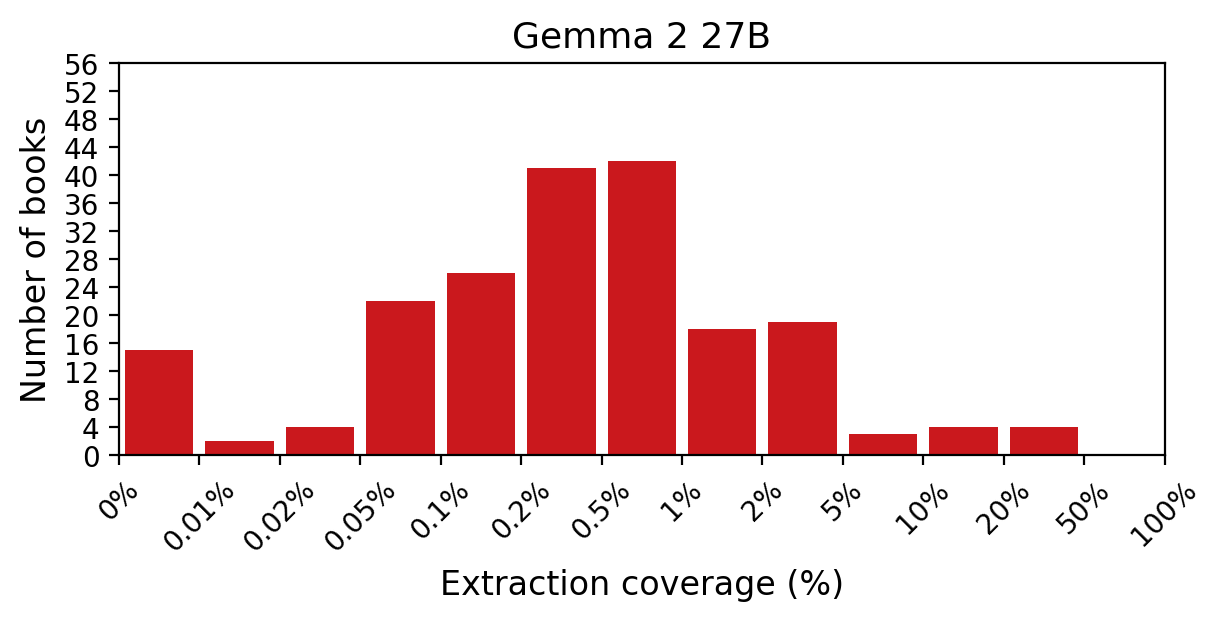}
\end{minipage}\hfill
\begin{minipage}[t]{0.4\textwidth}\centering
  \includegraphics[width=\linewidth]{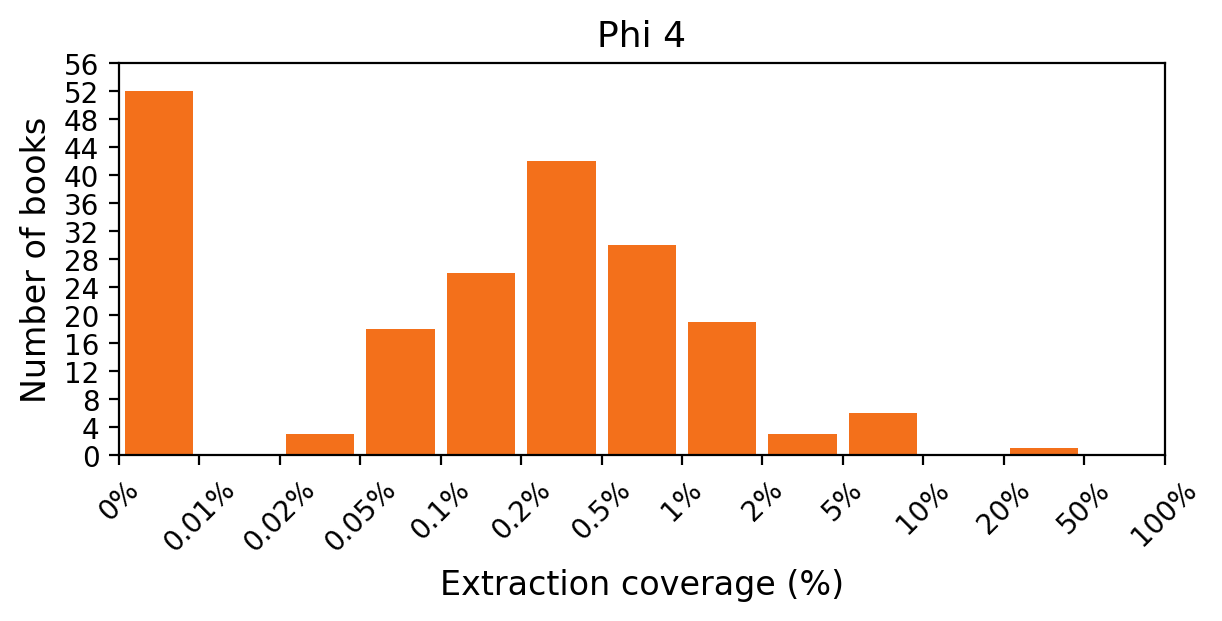}
\end{minipage}
\caption{\textbf{Extraction coverage for $200$ books and $14$ LLMs.} 
Computing Definition~\ref{app:def:cov} for the results in Appendix~\ref{app:sec:sliding-window:results} ($50$-token prefix $+$ $50$-token suffix; top-$k$ decoding with $T\!=\!1$, $k\!=\!40$; $\tau=\tau_\text{min}=0.1\%$).  
As elsewhere, we divide results into our three categories by color: LLMs that we know with certainty were trained on \texttt{Books3} (\textcolor{seabornbluemid}{blue}), LLMs that we can confidently conclude were trained on (at least parts of) some books that were also contained in \texttt{Books3} (\textcolor{seabornredmid}{red}), and an LLM whose developers claim was not trained on whole copyrighted books (\textcolor{seabornorangemid}{orange}).\looseness=-1
}
\label{fig:coverage:histogram-grid}
\end{figure*}

Here, we provide more extensive extraction coverage results. 
We include both types of comparisons for the $14$ models for which detailed sliding-window results are given in Appendix~\ref{app:sec:sliding-window:results}.
All results are shown for $\tau=\tau_\text{min}=0.1\%$, which corresponds to the total (largest) amount of extraction coverage revealed by our experiments.
Figure~\ref{fig:coverage:histogram-grid} plots histograms of extraction coverage across all $200$ books for each model, 
and Table~\ref{app:tab:coverage} reports per-book extraction coverage results for each model.

\newcommand{\pct}[1]{%
  \begingroup
  \dimen0=#1pt\relax
  \ifdim\dimen0>10pt \tiny{\textcolor{red}{#1\%}}%
  \else\ifdim\dimen0>5pt \textcolor{Purple}{#1\%}%
  \else\ifdim\dimen0>2.5pt \textcolor{blue}{#1\%}%
  \else #1\%%
  \fi\fi\fi
  \endgroup
}

\newcommand{\bookcolwidth}{0.135\textwidth}
\newcolumntype{L}[1]{>{\raggedright\arraybackslash\tiny}p{#1}}
\newcolumntype{C}[1]{>{\centering\arraybackslash\tiny}p{#1}}
\newlength\ModelColW
\setlength{\ModelColW}{.8cm}
\setlength{\tabcolsep}{1pt}
\rowcolors{2}{gray!15}{white}

\begin{center}
\begin{longtable}{L{\bookcolwidth} *{14}{C{\ModelColW}}}
\caption{\textbf{Extraction coverage for $200$ books and $14$ LLMs.}
We use Definition~\ref{app:def:cov} to compute extraction coverage for the results in Appendix~\ref{app:sec:sliding-window:results}, which use $100$-token sequences ($50$-token prefix $+$ $50$-token suffix) and top-$k$ decoding ($T\!=\!1$, $k\!=\!40$).
We set $\tau=\tau_\text{min}=0.1\%$ to compute the total amount of extraction coverage that we observe.
\looseness=-1}
\footnotesize
\label{app:tab:coverage}\\
\toprule
\multicolumn{15}{c}{\textbf{Proportion of each book that is extractable with $p_\vz\ge 0.1\%$} \textbf{for different models}}\\
\midrule
\textbf{Book} &
\makecell{\textbf{\textsc{Llama}}\\\textbf{\textsc{2 7B}}} &
\makecell{\textbf{\textsc{Llama}}\\\textbf{\textsc{3 8B}}} &
\makecell{\textbf{\textsc{Llama}}\\\textbf{\textsc{3.1 8B}}} &
\makecell{\textbf{\textsc{Llama}}\\\textbf{\textsc{1 13B}}} &
\makecell{\textbf{\textsc{Llama}}\\\textbf{\textsc{2 13B}}} &
\makecell{\textbf{\textsc{Pythia}}\\\textbf{\textsc{12B}}} &
\makecell{\textbf{\textsc{Llama}}\\\textbf{\textsc{1 65B}}} &
\makecell{\textbf{\textsc{Llama}}\\\textbf{\textsc{2 70B}}} &
\makecell{\textbf{\textsc{Llama}}\\\textbf{\textsc{3 70B}}} &
\makecell{\textbf{\textsc{Llama}}\\\textbf{\textsc{3.1 70B}}} &
\makecell{\textbf{\textsc{DS}}\\\textbf{\textsc{v1 67B}}} &
\makecell{\textbf{\textsc{Qwen}}\\\textbf{\textsc{2.5 72B}}} &
\makecell{\textbf{\textsc{Gemma}}\\\textbf{\textsc{2 27B}}} &
\makecell{\textbf{\textsc{Phi}}\\\textbf{\textsc{4}}} \\
\midrule
\endfirsthead

\multicolumn{15}{l}{\footnotesize\emph{Table \thetable\ (continued). Proportion of each book extractable with $p_\vz\ge 0.1\%$ for different models.}}\\
\toprule
\textbf{Book} &
\makecell{\textbf{\textsc{Llama}}\\\textbf{\textsc{2 7B}}} &
\makecell{\textbf{\textsc{Llama}}\\\textbf{\textsc{3 8B}}} &
\makecell{\textbf{\textsc{Llama}}\\\textbf{\textsc{3.1 8B}}} &
\makecell{\textbf{\textsc{Llama}}\\\textbf{\textsc{1 13B}}} &
\makecell{\textbf{\textsc{Llama}}\\\textbf{\textsc{2 13B}}} &
\makecell{\textbf{\textsc{Pythia}}\\\textbf{\textsc{12B}}} &
\makecell{\textbf{\textsc{Llama}}\\\textbf{\textsc{1 65B}}} &
\makecell{\textbf{\textsc{Llama}}\\\textbf{\textsc{2 70B}}} &
\makecell{\textbf{\textsc{Llama}}\\\textbf{\textsc{3 70B}}} &
\makecell{\textbf{\textsc{Llama}}\\\textbf{\textsc{3.1 70B}}} &
\makecell{\textbf{\textsc{DS}}\\\textbf{\textsc{v1 67B}}} &
\makecell{\textbf{\textsc{Qwen}}\\\textbf{\textsc{2.5 72B}}} &
\makecell{\textbf{\textsc{Gemma}}\\\textbf{\textsc{2 27B}}} &
\makecell{\textbf{\textsc{Phi}}\\\textbf{\textsc{4}}} \\
\midrule
\endhead

\midrule
\multicolumn{15}{r}{\footnotesize\emph{Continued on next page}}\\
\endfoot

\bottomrule
\endlastfoot
\textit{A Game of Thrones} & \pct{0.65} & \pct{0.75} & \pct{0.78} & \pct{0.76} & \pct{0.78} & \pct{0.67} & \pct{3.34} & \pct{3.07} & \pct{37.39} & \pct{39.39} & \pct{0.81} & \pct{1.25} & \pct{0.87} & \pct{0.59} \\
\textit{A Little Life} & \pct{0.02} & \pct{0.03} & \pct{0.03} & \pct{0.05} & \pct{0.05} & \pct{0.04} & \pct{0.16} & \pct{0.16} & \pct{0.35} & \pct{0.46} & \pct{0.04} & \pct{0.03} & \pct{0.07} & \pct{0.04} \\
\textit{A People's History of the United States} & \pct{1.30} & \pct{2.01} & \pct{2.05} & \pct{1.91} & \pct{1.66} & \pct{0.92} & \pct{6.86} & \pct{6.77} & \pct{17.91} & \pct{26.27} & \pct{1.71} & \pct{18.82} & \pct{2.05} & \pct{1.01} \\
\textit{A Perfectly Good Family} & \pct{0.17} & \pct{0.24} & \pct{0.24} & \pct{0.14} & \pct{0.18} & \pct{0.27} & \pct{0.21} & \pct{0.19} & \pct{0.42} & \pct{0.58} & \pct{0.28} & \pct{0.28} & \pct{0.24} & \pct{0.22} \\
\textit{A Return to Love} & \pct{0.64} & \pct{0.85} & \pct{0.85} & \pct{0.85} & \pct{0.94} & \pct{0.62} & \pct{2.76} & \pct{3.04} & \pct{4.35} & \pct{4.02} & \pct{1.08} & \pct{0.80} & \pct{0.73} & \pct{0.58} \\
\textit{A Visit from the Goon Squad} & \pct{0.07} & \pct{0.00} & \pct{0.10} & \pct{0.07} & \pct{0.07} & \pct{0.20} & \pct{0.29} & \pct{0.36} & \pct{0.74} & \pct{1.06} & \pct{0.19} & \pct{0.22} & \pct{0.17} & \pct{0.11} \\
\textit{A Wrinkle in Time} & \pct{1.44} & \pct{1.62} & \pct{1.57} & \pct{1.61} & \pct{1.69} & \pct{1.75} & \pct{4.54} & \pct{3.93} & \pct{10.00} & \pct{14.59} & \pct{2.32} & \pct{2.09} & \pct{1.57} & \pct{1.44} \\
\textit{After I'm Gone} & \pct{0.19} & \pct{0.28} & \pct{0.28} & \pct{0.16} & \pct{0.27} & \pct{0.36} & \pct{0.39} & \pct{0.31} & \pct{0.40} & \pct{0.46} & \pct{0.41} & \pct{0.36} & \pct{0.28} & \pct{0.24} \\
\end{longtable}
\end{center}
\rowcolors{2}{white}{white}
\normalsize

\FloatBarrier

\section{Measurement validity}\label{app:sec:validity}

In the follow-on work of~\citet{cooper2026firstprinciples}, we take up this topic in significantly more detail.
Here, we provide more detail on the validity experiments introduced in Section~\ref{sec:validity}. 
These experiments support that our measurements capture true instances of extraction. 
On this basis, we expand our analysis to LLMs with undisclosed information about training on books---\textsc{DeepSeek v1 67B}, \textsc{Qwen 2.5 72B}, and \textsc{Gemma 2 27B}. 
Because our controls show that our method does not mistake chance generation for extraction, we can be confident that when these LLMs register extraction success with our measurements, it reflects memorization.\looseness=-1 

Given the length of this appendix, we provide a summary: 
\begin{itemize}[leftmargin=0.65cm]
    \item \textbf{Extraction of known training data is evidence of memorization.} 
    For models where the training data are known, high-probability generation of a verbatim suffix given its prefix is widely accepted as evidence of memorization~\citep{carlini2021extracting, carlini2023quantifying, hayes2025measuringmemorizationlanguagemodels}. 
    Probabilistic extraction provides a more fine-grained version of the same test: rather than returning a single greedy continuation, it assigns a probability to the target suffix and reveals whether it surpasses a minimum acceptable threshold.\looseness=-1

    \item \textbf{We adopt a deliberately conservative minimum threshold.} 
    Probabilistic extraction produces a continuous measure of extraction, but requires a cutoff. 
    We set $\tau_\text{min}$ very high to ensure we do not over-count memorization. 
    This makes our results lower bounds: we under-count memorization (Section~\ref{sec:background}).
    We capture more instances of memorization than greedy decoding, but we still do not over-count. 
    We defer more fine-grained settings of $\tau_\text{min}$ to future work. 
    
    \item \textbf{Validating against non-training data}
    The majority of our experiments---those on \textsc{Llama} and \textsc{Pythia} models---fit the case of extracting known training data from models. 
    It is \emph{known} that \texttt{Books3} was included in the training data, and we are running experiments to extract \texttt{Books3} sequences. 
    But we can also do more to ensure validity: 
    we can test our procedure on known \emph{non}-training data, and show that suffix probabilities fall below our chosen $\tau_\text{min}$. 
    We (and \citet{hayes2025measuringmemorizationlanguagemodels}) do this for an additional sanity check. 
    In particular, we perform over $300$ negative control experiments (Section~\ref{sec:validity:controls} \& Appendix~\ref{app:sec:validity:controls}). 
    Work on greedy-decoded discoverable extraction often does not do this; it assumes validity or uses shorter prefix lengths to induce less conditioning on the output~\citep{nasr2023scalable} (Appendix~\ref{app:sec:validity:membership}).  
    Overall, our results confirm that our choice of $\tau_\text{min}$ is very conservative; 
    we could set it lower without risking false positives.
    However, we are deliberately cautious, and defer this to future work  (Section~\ref{sec:validity:implications}). 

    \item \textbf{Stress-testing extraction with longer prefixes.}
    When validating $\tau_\text{min}$ on known training data, we can also ask whether longer prefixes make memorization more discoverable. 
    Because suffix probabilities are conditional on the prefix, extending the prompt can further condition the model's output toward a certain part of its learned distribution (which can also further be affected by the decoding algorithm). 
    As a result (as others have shown~\citep{carlini2023quantifying}), for experiments on \emph{known} training data, when a piece of training data is memorized, a longer prefix prompt can raise the probability of the target suffix and expose it as memorized (via greedy decoding). 
    This memorization may not be \emph{discoverable} with only a $50$-token prompt, but may become discoverable with a longer prompt. 
    Our own experiments confirm these patterns, but also show that, for a specific sequence,  that longer prompts do not always lead to successful extraction: 
    if a sequence is not memorized, we are unable to find a long enough prefix to make it it extractable (Appendix~\ref{app:sec:validity:baseline}). 
    This supports the conclusion that probabilistic extraction captures true memorization, and clarifies the limits of analogies that treat LLMs as ``monkeys at the typewriter'' (Appendix~\ref{app:sec:validity:monkey}).

    \item \textbf{Bringing the pieces together.}
    Our validity experiments combine baselines, negative controls, and varied-length prefix tests. 
    For models known to be trained on \texttt{Books3}, we run baselines with different prefix lengths as points of comparison (Appendix~\ref{app:sec:validity:baseline}). 
    We then add two sets of negative controls: (i) \textsc{Phi 4}, a model not trained on whole copyrighted books (Appendices~\ref{app:sec:sliding-window} \& \ref{app:sec:validity}), and (ii) books published after the training cutoff for LLMs trained on \texttt{Books3} (Appendix~\ref{app:sec:validity:cutoff}). 
    Across more than $300$ such experiments, suffix probabilities for non-training data are essentially zero, and nonzero values are orders of magnitude below $\tau_\text{min}$. 
    As a result, our procedure does not mistakenly register non-training data as extractable, reinforcing that what we count as extraction reflects true instances of memorization. 

    \item \textbf{Applying the procedure to models with unknown training data.}
    Based on these results, we can also apply the procedure to models whose training data are undisclosed.  
    For sequences that register as extractable in these models, we can confidently interpret them as evidence of memorization, regardless of whether the training-data source was \texttt{Books3} directly or overlapping material from elsewhere (Appendix~\ref{app:sec:validity:membership}). 
    Importantly, as we state throughout, for our experiments on these models, our claims about membership in the training data \emph{are only about specific extracted sequences}, not whole books.
\end{itemize}

\paragraph{Appendix outline.}
\begin{itemize}[leftmargin=0.65cm]
    \item \textbf{Appendix~\ref{app:sec:validity:motivations}.} Motivation for these experiments.
    \item \textbf{Appendix~\ref{app:sec:validity:setup}.} Setup.
    \item \textbf{Appendix~\ref{app:sec:validity:baseline}.} Additional baseline experiments for LLMs known to be trained on \texttt{Books3}.
    \item \textbf{Appendix~\ref{app:sec:validity:controls}.} Additional negative control experiments on non-training data.
    \item \textbf{Appendix~\ref{app:sec:validity:monkey}.} Additional discussion on ``the monkey at the typewriter.''
    \item \textbf{Appendix~\ref{app:sec:validity:membership}.} Discussion of relationship to prior literature on membership inference.
\end{itemize}

\subsection{Motivating our validity experiments}\label{app:sec:validity:motivations}

The main point of these experiments is to validate that our measurements capture true instances of extraction, rather than chance generation of verbatim suffixes. 
They help us make confident claims about both our results for models trained on \texttt{Books3} and our application of the procedure to models with undisclosed training data. 
Here, we expand on these motivations (Appendix~\ref{app:sec:validity:motivations:extend}), connect our work to the literature on greedy-decoded discoverable extraction (Appendix~\ref{app:sec:validity:motivations:lit}), and clarify the role of varying prefix length in our baselines and negative controls (Appendix~\ref{app:sec:validity:motivations:vary}). \looseness=-1 

\subsubsection{Setting $\tau_\text{min}$ and extending to models with unknown training data}\label{app:sec:validity:motivations:extend}

We can be fairly confident that \textsc{DeepSeek v1 67B},  \textsc{Qwen 2.5 72B}, and \textsc{Gemma 2 27B} were trained on at least some books:
the Pile~\citep{gao2020pile} and several other common LLM pre-training datasets either directly include \texttt{Books3} or other book corpora~\citep{lee2023explainers, lee2023talkin}. 
Other work on extraction with uncertain training data treats the Pile as part of a proxy for the true training set~\citep{nasr2023scalable, nasr2025scalable}.
However, without disclosure, we cannot be absolutely certain. 
And without knowing whether \texttt{Books3} was in fact included, we cannot---without additional validity experiments---make claims about memorization that are as strong as those we make for models known to be trained on \texttt{Books3}. 
This is because membership in the training data is a prerequisite for memorization; 
by definition, a model cannot memorize sequences it has never seen.\looseness=-1

This is also why observing $p_\vz$ on non-training data is important for setting $\tau_\text{min}$. 
By comparing suffix probabilities from negative controls on non-training data (Section~\ref{sec:validity:controls}, Appendix~\ref{app:sec:validity:controls}) with those from baselines on known training data (Section~\ref{sec:validity:baseline}, Appendix~\ref{app:sec:validity:baseline}), we can draw a conservative line for what $p_\vz$ is high enough to be indicative of memorization.\looseness=-1 

That is, our negative controls clarify how to set the threshold that we use to count extraction success, $\tau_\text{min}$---the minimum $p_\vz$ for which we interpret our measurements as evidence of memorization.
Running our procedure on \emph{non}-training data, we confirm that the values of $p_\vz$ we observe are far below our chosen cutoff. 
That way, when we count extraction success for a given sequence with $p_\vz\geq\tau_\text{min}$, we can be confident that it does not reflect chance generation. 
Our results in Section~\ref{sec:validity} and the rest of this Appendix confirm this:
$\tau_\text{min}=0.1\%$ is conservative---well above any $p_\vz$ observed in negative control experiments.\looseness=-1

We therefore use these results to justify our choice of $\tau_\text{min}$ in this paper. 
More extensive experiments would be needed to lower $\tau_\text{min}$ or calibrate per LLM. 
Here, we prefer to be conservative: 
we would rather under-count memorization than risk over-counting~\citep{carlini2022membershipinferenceattacksprinciples, hayes2025strongmia}, and defer fine-grained threshold settings to future work.\looseness=-1

\subsubsection{Relating greedy-decoded and probabilistic discoverable extraction}\label{app:sec:validity:motivations:lit}

To our knowledge, we run the largest set of validity experiments in the ML literature on memorization and extraction. 
Such experiments are often not performed in work on extraction~\citep{nasr2023scalable, nasr2025scalable, carlini2021extracting, karamolegkou2023copyrightviolationslargelanguage}, with important exceptions~\citep{carlini2023quantifying}. 
\citet{hayes2025measuringmemorizationlanguagemodels} conducted one small-scale test on non-training data for a small \textsc{Pythia} model, and used this to support validity claims for probabilistic extraction from much larger and higher quality \textsc{Llama 1} models.

Given these norms, we initially omitted our negative controls from the main paper, deferring discussion to the Appendix. 
However, after receiving some feedback about earlier drafts of this work that indicated serious misunderstandings about the nature of memorization and extraction, we realized that including the results of these experiments in the main text could help some readers better understand our results and claims. 
We now clearly delineate two sets of negative controls, both in this appendix and in the main paper text (Section~\ref{sec:validity:controls}):
the experiments on \textsc{Phi 4} (Appendix~\ref{app:sec:validity:phi4}) and the experiments on books published after training-date cutoffs (Appendix~\ref{app:sec:validity:cutoff}). 

Here, we clarify the relationship between prior definitions and our approach, and explain how this motivates our validity experiments.

\paragraph{Greedy-decoded discoverable extraction.}
It is widely accepted that successful greedy-decoded discoverable extraction on known training data is evidence of memorization;
memorization remains the most parsimonious explanation~\citep{hayes2025measuringmemorizationlanguagemodels, carlini2021extracting, nasr2023scalable, carlini2025blog, cooper2024files}. 
In this procedure, greedy decoding always selects the top-ranked token at each step, making it a deterministic approximation of the most likely sequence. 
\citet{carlini2021extracting} describe this choice as a tractable proxy for the $\argmax$ in their definition of model knowledge extraction:

\begin{definition}[\textbf{Model Knowledge Extraction}, from \citet{carlini2021extracting}]
A string $s$ is extractable from an LM $f_\theta$ if there exists a prefix $c$ such that:
\[
s \leftarrow \argmax_{s': |s'|=N} f_\theta(s' | c)
\]
We abuse notation slightly here to denote by $f_\theta(s' | c)$ the likelihood of an entire sequence $s'$. 
\emph{Since computing the most likely sequence $s$ is intractable for large $N$}, the $\argmax$ in [the definition]  can be replaced by an appropriate sampling strategy \emph{(e.g., greedy sampling)} that reflects the way in which the model $f_\theta$ generates text in practical applications.
\end{definition}

We add emphasis to two areas in this definition, which we discuss below.
First, we note that this definition does not mention training data; 
it is just a definition about a model having knowledge of string $s$.
\citet{carlini2021extracting} connect extraction \emph{of training data} to memorization through a second definition:\looseness=-1

\begin{definition}[\textbf{$k$-Eidetic Memorization}, from \citet{carlini2021extracting}]
A string $s$ is $k$-eidetic memorized (for $k\geq1$) by an LM $f_\theta$ if $s$ is extractable from $f_\theta$ and $s$ appears in at most $k$ examples in the training data $X: |\{ x \in X : s \subseteq x \} | \leq k$.
\end{definition}

Successful extraction, under these definitions, is evidence for memorization. 
Ideally, $s$ should be computed as the \emph{most likely sequence} under $f_\theta$.
However, since this is not tractable to compute in practice, \emph{greedy decoding} is suggested as a reasonable proxy.
It forces top-1 choices regardless of their actual probabilities under the model.

\paragraph{The relationship to probabilistic extraction.}
Greedy decoding is equivalent to top-$k$ decoding with $k\!=\!1$, which  only considers the token with top-$1$-ranked logit.
(Temperature has no effect in this setting, since both temperature and top-$k$ are rank-preserving.)
Probabilistic extraction relaxes this by considering a wider set of high-probability continuations (e.g., top-$40$). 
This better reflects the model's high-probability distribution and avoids the problem where greedy decoding yields globally low-probability sequences. 
\citet{hayes2025measuringmemorizationlanguagemodels} motivate their metric precisely from this point: 
the greedy-decoded suffix is \emph{locally} optimal, but it may lead to a \emph{globally} relatively low probability sequence under the model.\looseness=-1

A hypothetical example illustrates this gap: a greedy-decoded suffix can have vanishing total probability, while a slightly different path within the top-$k$ yields a much higher-probability suffix.  
Consider a greedy-decoded suffix where the first token has $99\%$  probability and is selected, and then every top-$1$ token after it for $49$ tokens has probability $1\%$; 
the total probability for the $50$-token suffix would be $0.99 \times (0.01)^{49} = 9.9 \times 10^{-99}$. 
Now imagine we instead were \emph{not} restricted to always picking the top-$1$ probability token; 
we relax this requirement to, for example, allow to select from the top-$k$ tokens.
We consider probabilities computed with respect to the base $T\!=\!1$ distribution.  
For the first token position, the top-$2$ token might have probability $1\%$, and then the remaining top-$1$ $49$ tokens could each have probability $99\%$; 
the total probability for this suffix would be $0.01 \times (0.99)^{49}\!\approx\!0.006$ (or about $0.6\%$).\looseness=-1  

The point is, by allowing for a bit more flexibility in the decoding strategy, the model can potentially produce much higher probability sequences;
when we observe that such ($\geq\tau_\text{min}$) sequences match the verbatim suffix from the training data, then we count them as extracted.
As long as we choose $\tau_\text{min}$ appropriately---i.e., such that it does not capture low-probability sequences that are perhaps reflective of generalization---we should obtain higher fidelity measurements of memorization.
We should have fewer false negatives than greedy-decoded discoverable extraction.

\paragraph{Greedy-decoding is a degenerate case of probabilistic extraction.}
Greedy decoding can be seen as a degenerate case of probabilistic extraction, corresponding to $\tau_\text{min}=100\%$. 
Each step assigns all probability mass to a single token under a normalized top-$1$ distribution, even if that token's true model probability is very low. 
Probabilistic extraction surfaces this mismatch more clearly and is a better proxy for identifying high-probability memorized sequences. 
Because it includes the greedy suffix when it matches the training suffix, it never misses those cases, while also capturing others that greedy decoding fails to expose. 

\subsubsection{The role of prefix length in our experiments}\label{app:sec:validity:motivations:vary}

We run our validity experiments with various prefix lengths, not just for $100$-token sequences with $50$-token prefixes and $50$-token suffixes. 
This is a deliberate choice because additional context conditions the model's output distribution toward a particular output space. 
For memorized sequences, this is known to surface memorization that may not appear for shorter prefixes. 
\citet{carlini2023quantifying} call this the \newterm{discoverability phenomenon}: ``some memorization only becomes apparent \ldots when the model is prompted with a sufficiently long context.'' 
Through the lens of probabilistic extraction, a longer training-data prefix can raise the probability of the target suffix in Equation~\ref{eq:pz}, such that it surpasses $\tau_\text{min}$. 
(It is worth noting that any conditional extraction procedure---i.e., one that uses prompts, rather than unconditionally generating from the LLM---statistically biases the output.
This is one of the reasons that extraction under-counts memorization; 
it only searches a part of the learned distribution.)\looseness=-1

By testing various prefix lengths on known training data, we can estimate an upper bound on memorization (with respect to our extraction procedure) for a given work. 
We  expect longer prefixes to expose more memorization, but that increasing prefix length will not typically induce all suffix probabilities to $100\%$;
that is, we will observe more extraction only if there is more memorization to uncover. 
It is not the case that one can simply make the prefix arbitrarily long and then ``extract'' any target suffix. 
We discuss this further in Appendices~\ref{app:sec:validity:baseline} and~\ref{app:sec:validity:monkey}. 
This logic also strengthens the interpretation of our negative controls.
In this case, our measurements should not capture extraction for unique sequences of non-training data---even when we intentionally condition with longer prompts. 
Again, this shows that arbitrarily long suffixes do not all of a sudden mean we will register extraction with our measurements.
For these experiments, we should not be able to do so because the sequences were not included in the training data.\looseness=-1 

\subsection{Setup}\label{app:sec:validity:setup}

\paragraph{\texttt{Books3} books.} 
Running experiments with longer sequences is significantly more expensive. 
We therefore limit our results on varied prefixes to a small selection of books from \texttt{Books3}:
\emph{Beloved}~\citep{Beloved}, \emph{Harry Potter and the Sorcerer's Stone}~\citep{Harry_Potter_and_the_Sorcerer_s_Stone}, \emph{This Is How You Lose Her}~\citep{This_Is_How_You_Lose_Her}, and \emph{We Were Eight Years in Power}~\citep{We_Were_Eight_Years_in_Power}. 
We picked these $4$ books because each represents (roughly speaking) a different pattern for memorization in our sliding-window results on $50$-token prefixes (Appendix~\ref{app:sec:sliding-window}) for the model that exhibits the most memorization: \textsc{Llama 3.1 70B}.
\emph{This Is How You Lose Her} is barely memorized at all (Appendix~\ref{app:sec:sliding:This_Is_How_You_Lose_Her}). 
\emph{Beloved} has fragmented memorization ``hot-spots;'' there are many throughout the book, but they are rather disconnected, separated by lots of regions that do not register any memorization (Appendix~\ref{app:sec:sliding:Beloved}).
\emph{We Were Eight Years in Power} has several ``hot-spots;'' some are fragmented, but some indicate long, continuous regions of memorized text (Appendix~\ref{app:sec:sliding:We_Were_Eight_Years_in_Power}.
\emph{Harry Potter and the Sorcerer's Stone} is effectively memorized entirely, though not all suffixes have equally high extraction probability (Appendix~\ref{app:sec:sliding:Harry_Potter_and_the_Sorcerer_s_Stone}). 

\paragraph{Non-training data.} 
The large majority of sequences in the in-copyright \texttt{Books3} books above should not be training data for \textsc{Phi 4}; 
exceptions are highly duplicated text, like boilerplate copyright notices and occasional popular quotes that are included from other sources (e.g., the Bible, newspaper articles, earlier books). 
We also purchased a set of $4$ books published in 2025---after the release date of the newest model we test, \textsc{Qwen 2.5 72B} (September 19, 2024). 
(See Table~\ref{app:tab:cutoff-books}.)
We run these books through models known to have been trained on \texttt{Books3} as another point of comparison.
Similar to the above note about \textsc{Phi 4}, none of the unique text in these books should ping as extractable.
This may not be true for non-unique text: 
boilerplate text and popular or otherwise duplicated quotes from earlier sources.\looseness=-1  

\begin{table}[t!]
\caption{\textbf{Books that are non-training data.} 
The $4$ books that post-date training cutoff for all LLMs we test. 
We use these books for one set of our negative control experiments. 
See Section~\ref{sec:validity} and Appendix~\ref{app:sec:validity:cutoff}.}
\label{app:tab:cutoff-books} 
\centering
\rowcolors{2}{gray!15}{white}
\scriptsize
\begin{tabular}{cp{3cm}p{5.5cm}p{2cm}p{1cm}}  
\toprule
 & \textbf{Author} & \textbf{Title} & \textbf{Publication Date} & \textbf{Status} \\
\midrule
1 & Gareth Brown & \emph{The Society of Unknowable Objects}~\citep{unknowableobjects} & July 31, 2025 & \textcopyright\\
2 & Emily Henry & \emph{Great Big Beautiful Life}~\citep{greatbig} & April 22, 2025 & \textcopyright \\
3 & Ocean Vuong & \emph{The Emperor of Gladness}~\citep{emperorgladness} & May 13, 2025 & \textcopyright\\
4 & Sarah Wynn-Williams & \emph{Careless People}~\citep{careless} & March 11, 2025 & \textcopyright\\
\bottomrule
\end{tabular}
\vspace{-.5cm}
\end{table}

Overall, we want to see how changing prefix length impacts how much more memorization we can discover. 
As we increase prefix length, we expect to discover more memorization of \texttt{Books3} text for models trained on \texttt{Books3}, and to effectively observe no major changes for \textsc{Phi 4} for unique text from these books nor for  unique text from the 2025 books. 

\paragraph{Models.} 
Following from above, we run these experiments on $3$ models known to have been trained on \texttt{Books3}. 
We include \textsc{Llama 3.1 70B}, because this is the model that has memorized the most of all \textsc{Llama} models we test. 
We also include \textsc{Llama 3.1 8B}, as it is a much smaller model from the same family. 
It memorizes less than \textsc{Llama 3.1 70B}. 
(See Appendix~\ref{app:sec:sliding-window}.)
While it may be possible that increasing prefix length for \textsc{Llama 3.1 70B} surfaces many more books that are effectively entirely memorized (like \emph{Harry Potter and the Sorcerer's Stone}), we do not expect this to be the case for \textsc{Llama 3.1 8B}, since it is a much smaller capacity model.
Together, this will clarify how it is not the case that just making prefixes longer means that one could generate any target suffix they so desire---regardless of whether it was memorized. 
We also include \textsc{Llama 2 13B}. 
This model is a similar size as \textsc{Phi 4}, which we use in our negative controls. 
They have similar capacity, but were clearly trained very differently.
While we do not know all of the differences, we do know that \textsc{Llama 2 13B} was trained on \texttt{Books3}, while \textsc{Phi 4} was not. 
For one experiment on non-training data, we also include \textsc{Qwen 2.5 72B} as a point of comparison. 

\paragraph{Procedure.} 
We use the same sliding-window procedure as in Section~\ref{sec:book-procedure:method}, documented in detail in Appendix~\ref{app:sec:sliding-window:procedure}. 
We inherit all of those setup details; 
the only difference is that we run for sequences of different lengths (and associated different prefix lengths) to extract $50$-token suffixes: 
$75$ tokens ($25$-token prefixes $+$ $50$-token suffixes); 
$100$ tokens ($50$-token prefixes $+$ $50$-token suffixes, i.e., the typical setting); 
$150$ tokens ($100$-token prefixes $+$ $50$-token suffixes); 
$250$ tokens ($200$-token prefixes $+$ $50$-token suffixes);
$450$ tokens ($400$-token prefixes $+$ $50$-token suffixes); and,
$850$ tokens ($800$-token prefixes $+$ $50$-token suffixes).
%
On our hardware, it takes approximately $13$ hours to run \emph{We Were Eight Years in Power} through \textsc{Llama 3.1 70B} with $850$-token sequences (compared to $\sim\!1.5$ hours with our typical setup with $100$ tokens, see Appendix~\ref{app:sec:sliding:We_Were_Eight_Years_in_Power}).
This cost is the main reason we do not expand these experiments to a larger set of books and models.
\vspace{-.1cm}
\subsection{Results of varying prefix length on known training data}\label{app:sec:validity:baseline}
\vspace{-.1cm}

We next provide extended results for baseline experiments that vary prefix lengths on known training data. 
Our results yield the outcomes we expect:
we are able to surface more memorization with increased prefix length---to a point.
Except in rare cases like \emph{Harry Potter and the Sorcerer's Stone} and \textsc{Llama 3.1 70B}, in which the entire book is memorized, arbitrarily increasing prefix length does not make every training-data suffix extractable. 
This makes sense, as not all training data are memorized. 
These results also make clear that absence of extraction is not itself evidence that a sequence is \emph{not} a member of the training data.
Most sequences from \texttt{Books3} are not extractable even from models like \textsc{Llama 3.1 70B}, which memorize a lot.
So the claim we make with respect to absence of extraction signal is in this one direction.
When we make claims about training data membership, these are either based on ground-truth knowledge (e.g., knowing that  \texttt{Books3} was in the training data and that 2025 books were not), or are based on comparisons to our negative controls (Appendix~\ref{app:sec:validity:membership}).\looseness=-1

\begin{figure*}[t]
\centering
\begin{subfigure}{0.48\linewidth}
\includegraphics[width=\linewidth]{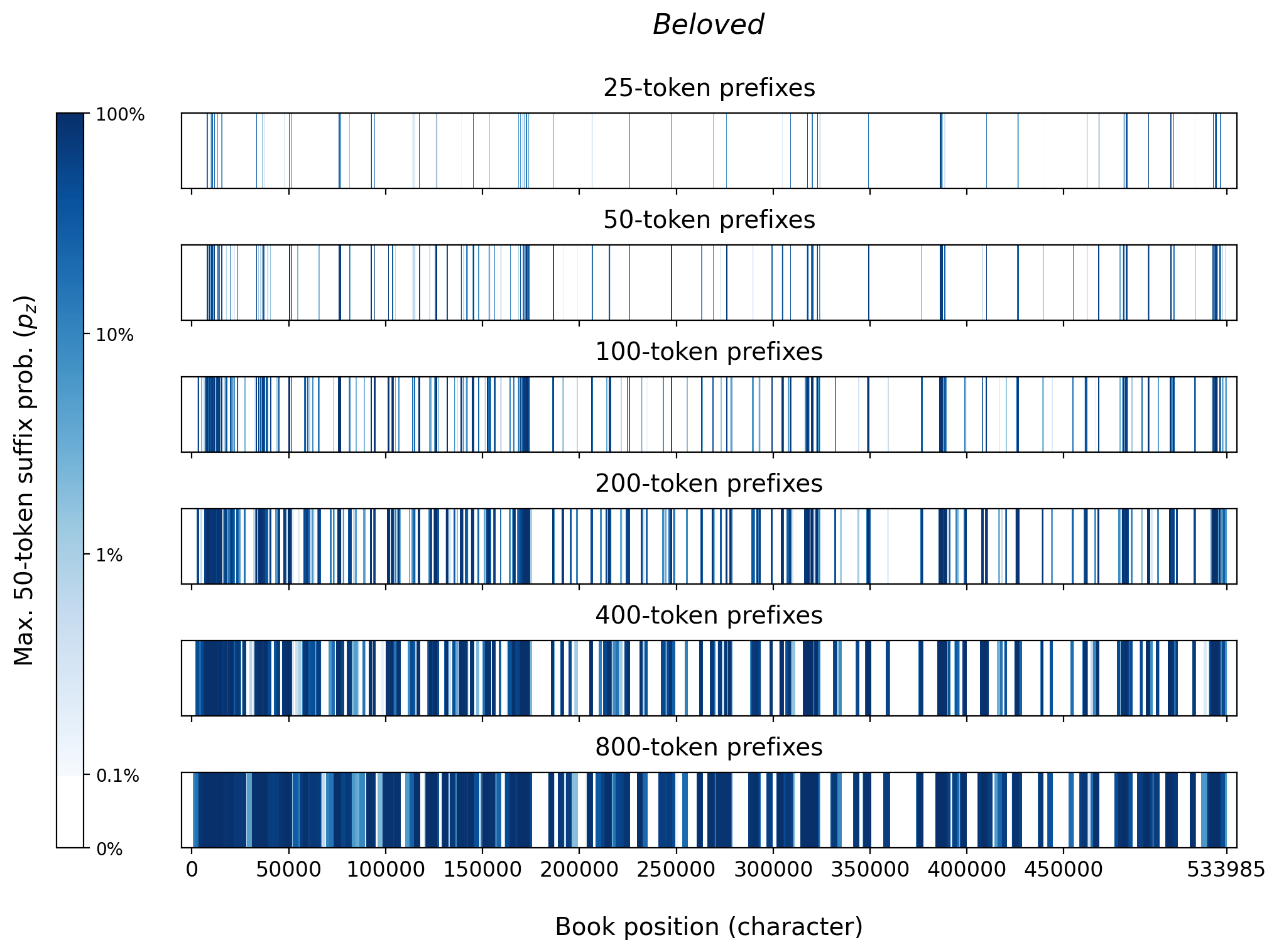}    
\end{subfigure}
\hspace{.25cm}
\begin{subfigure}{0.48\linewidth}
\includegraphics[width=\linewidth]{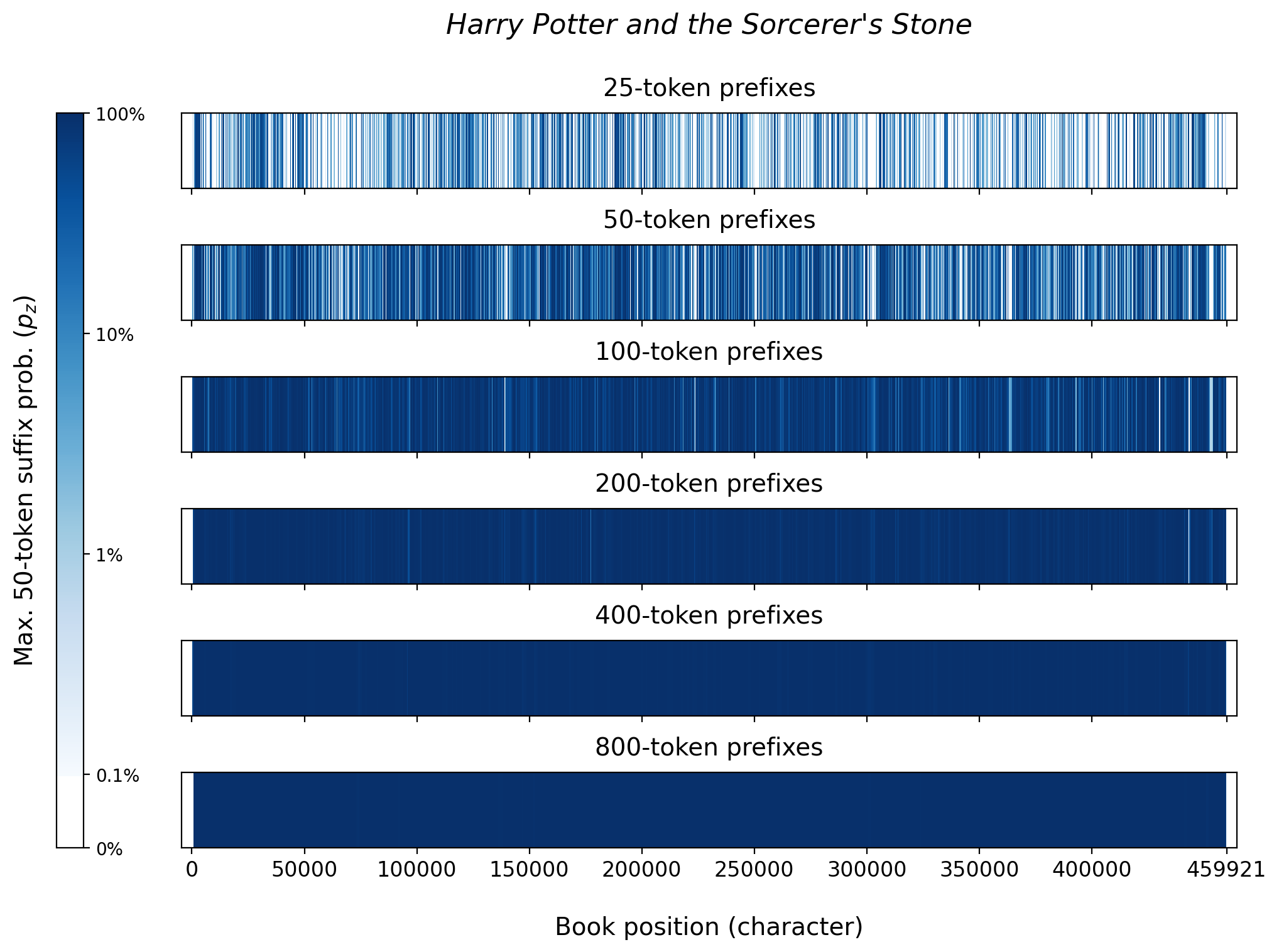}
\end{subfigure}
\begin{subfigure}{0.48\linewidth}
\includegraphics[width=\linewidth]{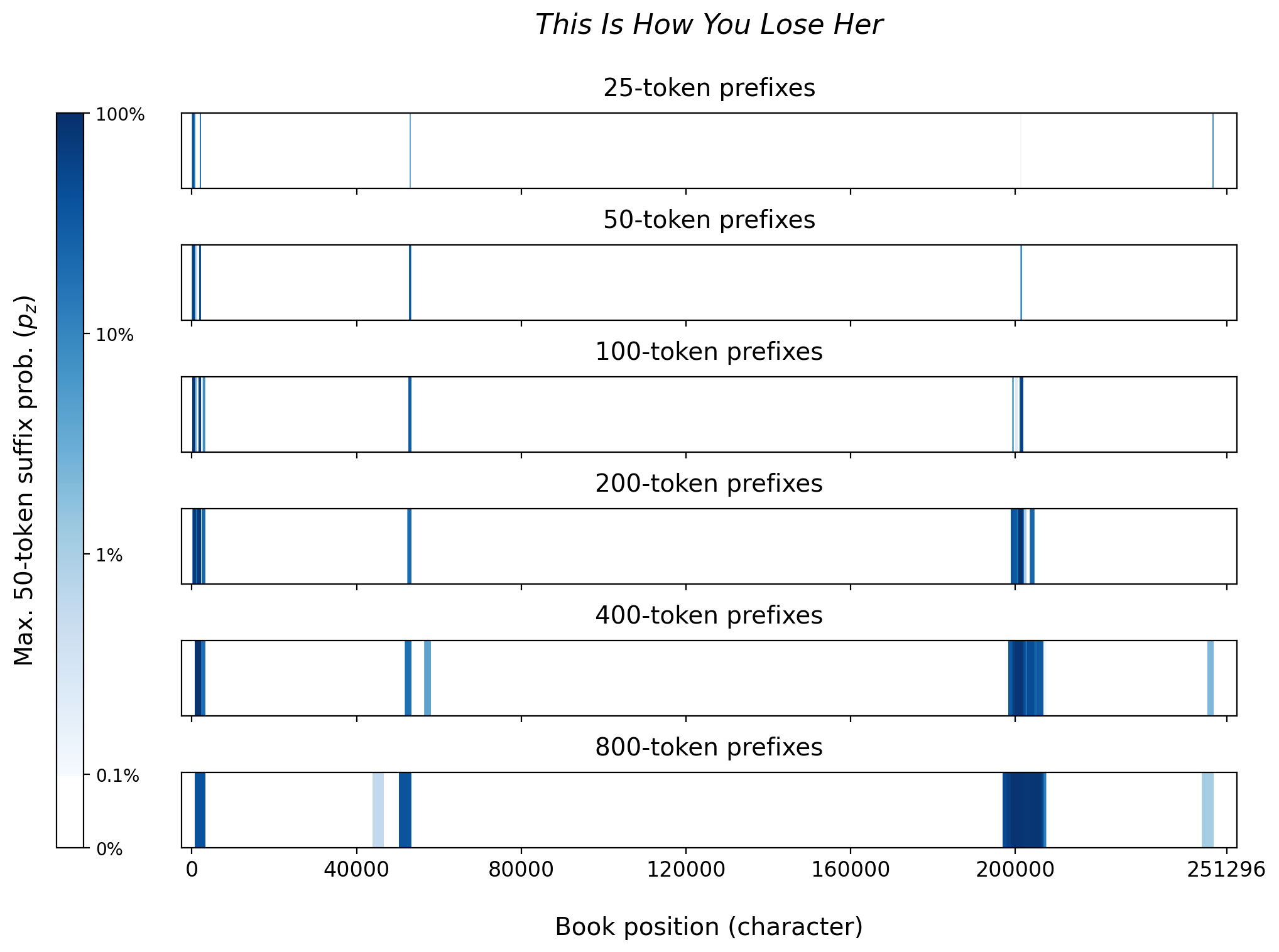}    
\end{subfigure}
\hspace{.25cm}
\begin{subfigure}{0.48\linewidth}
\includegraphics[width=\linewidth]{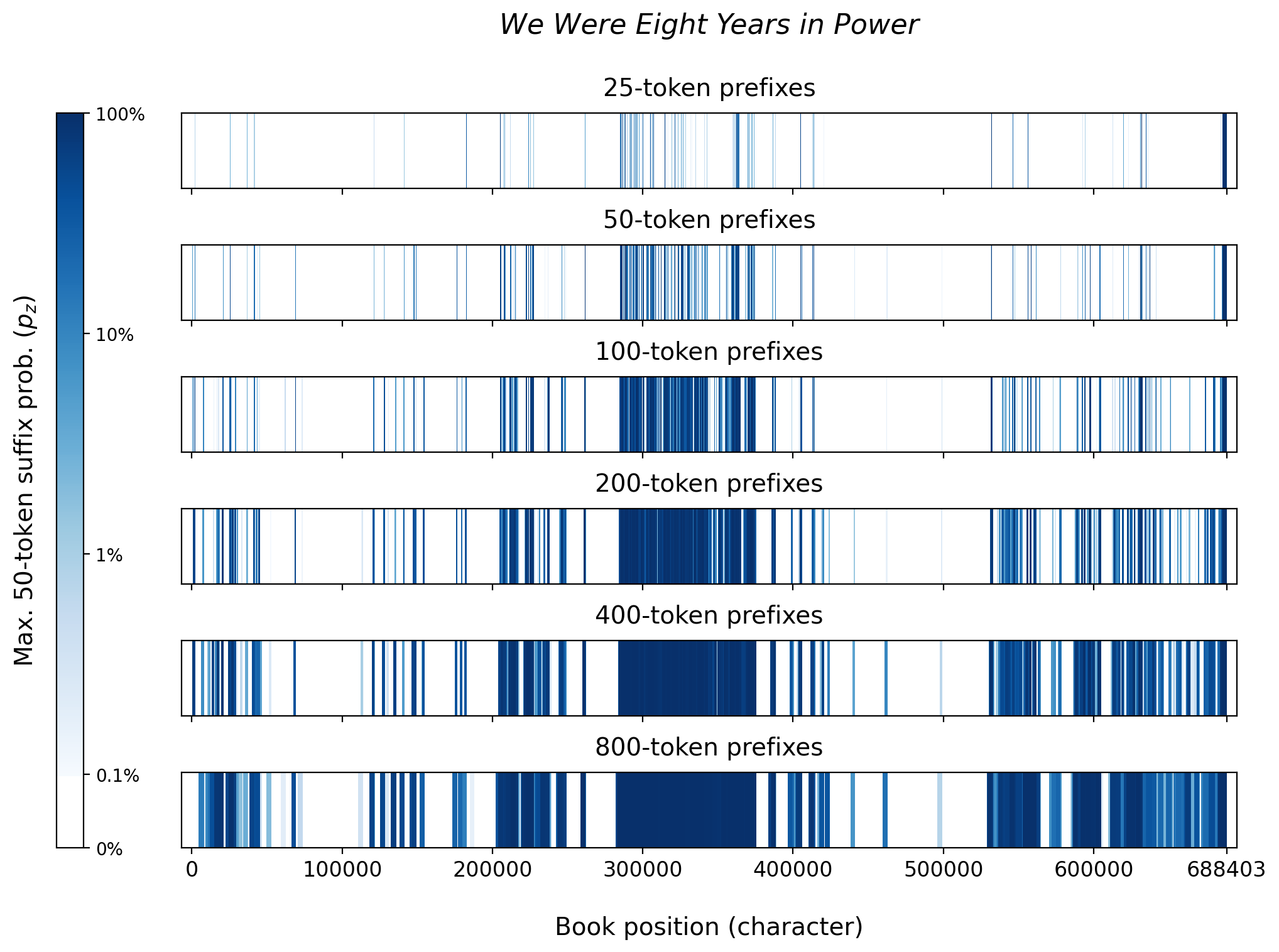}
\end{subfigure}
\caption{\textbf{Varying prefix for the \textsc{Llama 3.1 70B} baseline.} 
We run the sliding-window procedure for various prefix lengths for  $4$ books from \texttt{Books3}: \emph{Beloved}~\citep{Beloved}, \emph{Harry Potter and the Sorcerer's Stone}~\citep{Harry_Potter_and_the_Sorcerer_s_Stone}, \emph{This Is How You Lose Her}~\citep{This_Is_How_You_Lose_Her}, and \emph{We Were Eight Years in Power}~\citep{We_Were_Eight_Years_in_Power}.
\textsc{Llama} models are known to have been trained on \texttt{Books3}. 
We run probabilistic extraction on \textsc{Llama 3.1 70B} with top-$k$ ($T\!=\!1$, $k\!=\!40$) decoding for $50$-token suffixes and prefix lengths in $\{25, 50, 100, 200, 400, 800\}$. 
Extraction signal generally increases as prefix length increases.
However, this only appears to be the case if there is indeed more memorization to surface; 
increasing prefix length does not always make previously unextractable sequences extractable. 
There are also diminishing increases as the prefix lengthens.\looseness=-1
}
\label{fig:validity:baseline:llama-3.1-70b}
\vspace{-.25cm}
\end{figure*}

\begin{figure*}[t]
\centering
\begin{subfigure}{0.48\linewidth}
\includegraphics[width=\linewidth]{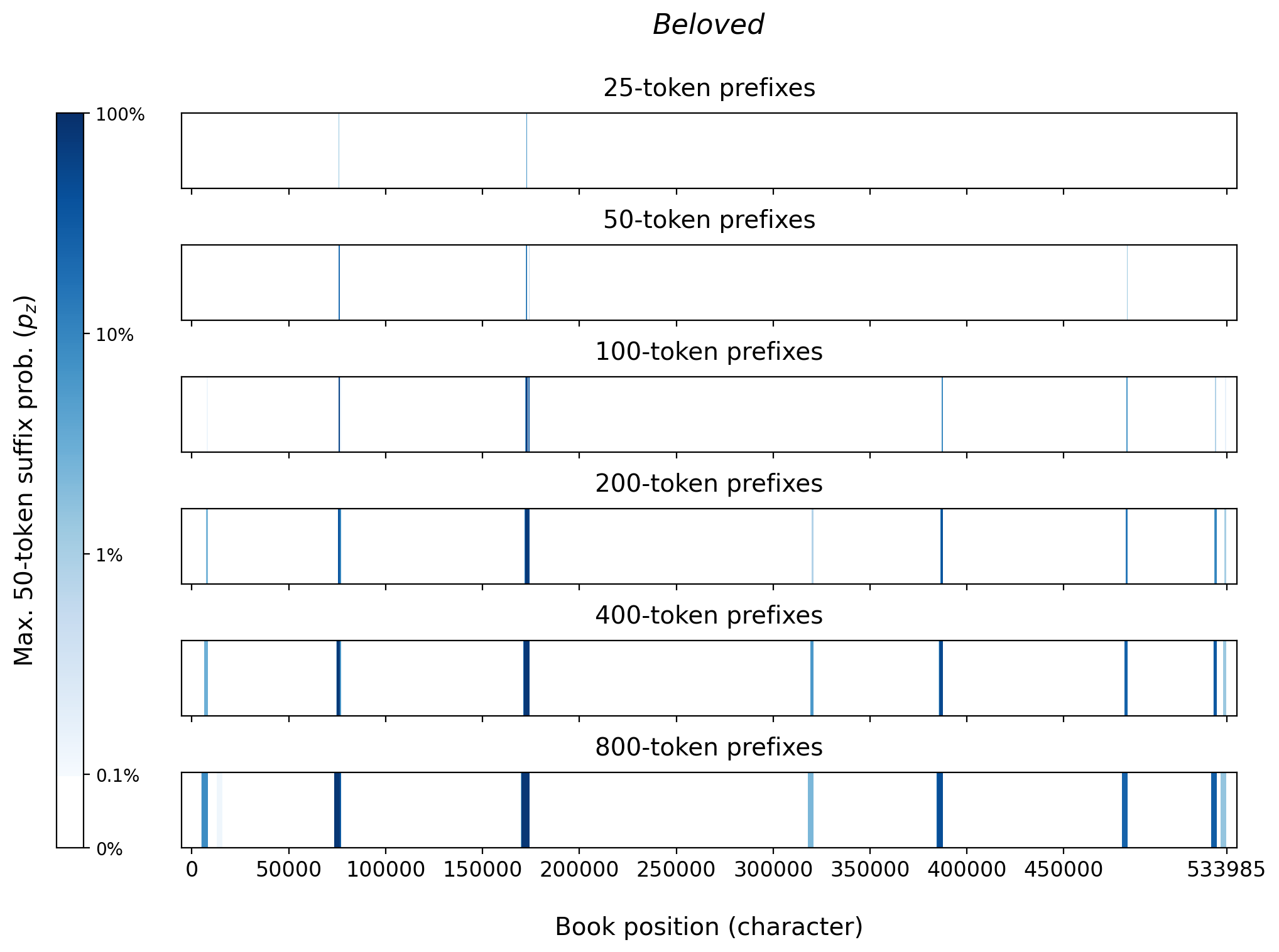}    
\end{subfigure}
\hspace{.25cm}
\begin{subfigure}{0.48\linewidth}
\includegraphics[width=\linewidth]{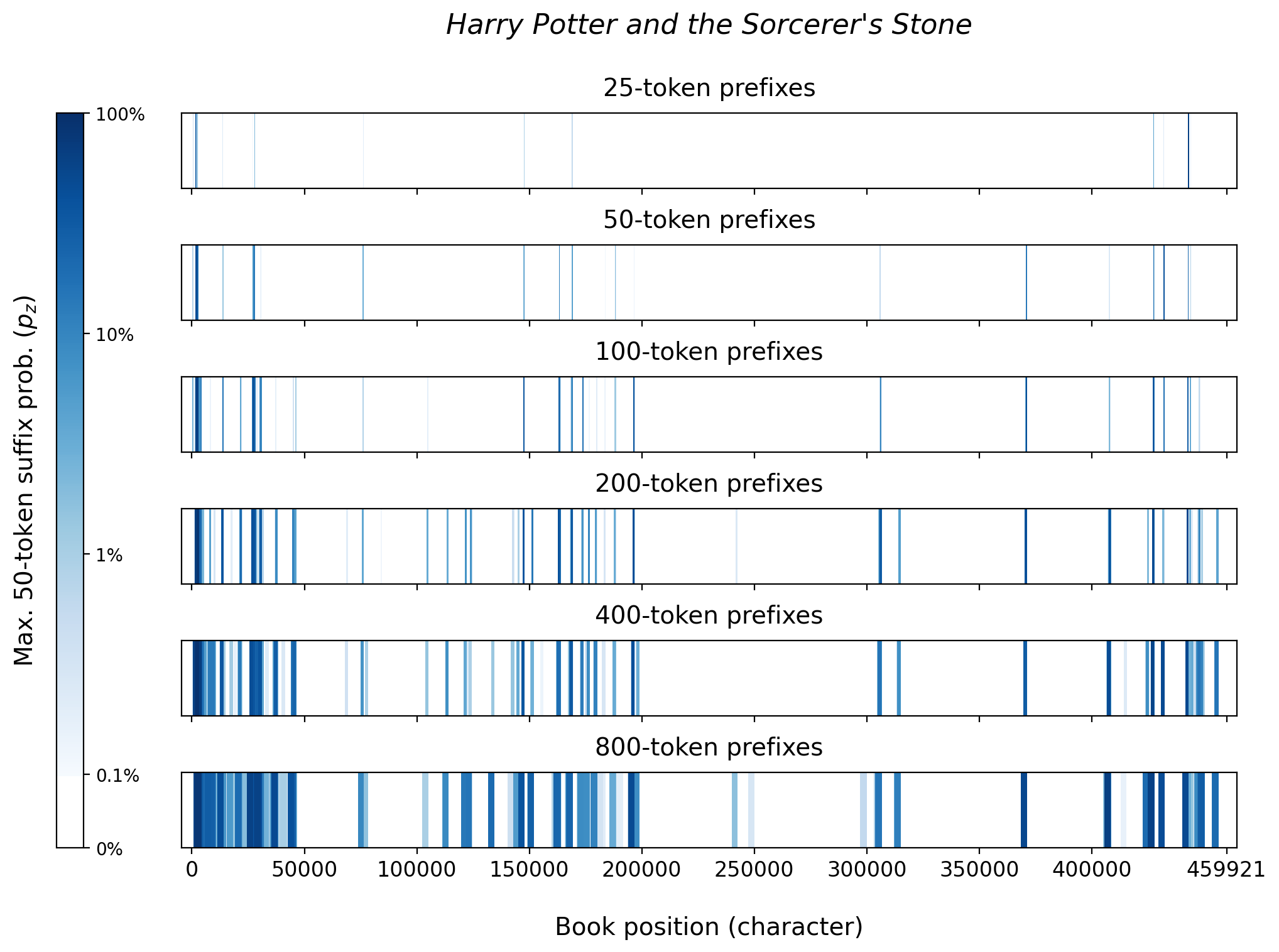}
\end{subfigure}
\begin{subfigure}{0.48\linewidth}
\includegraphics[width=\linewidth]{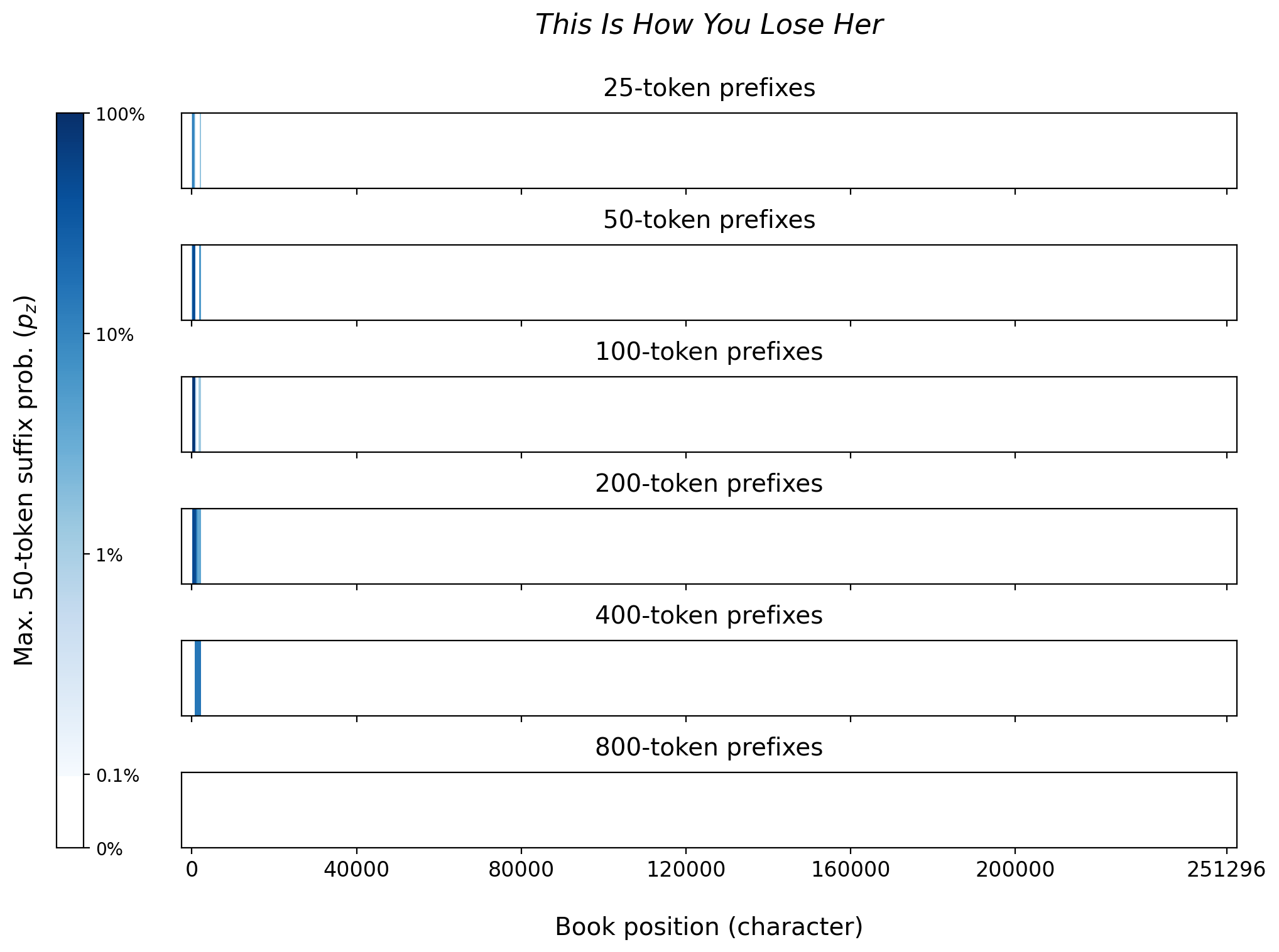}    
\end{subfigure}
\hspace{.25cm}
\begin{subfigure}{0.48\linewidth}
\includegraphics[width=\linewidth]{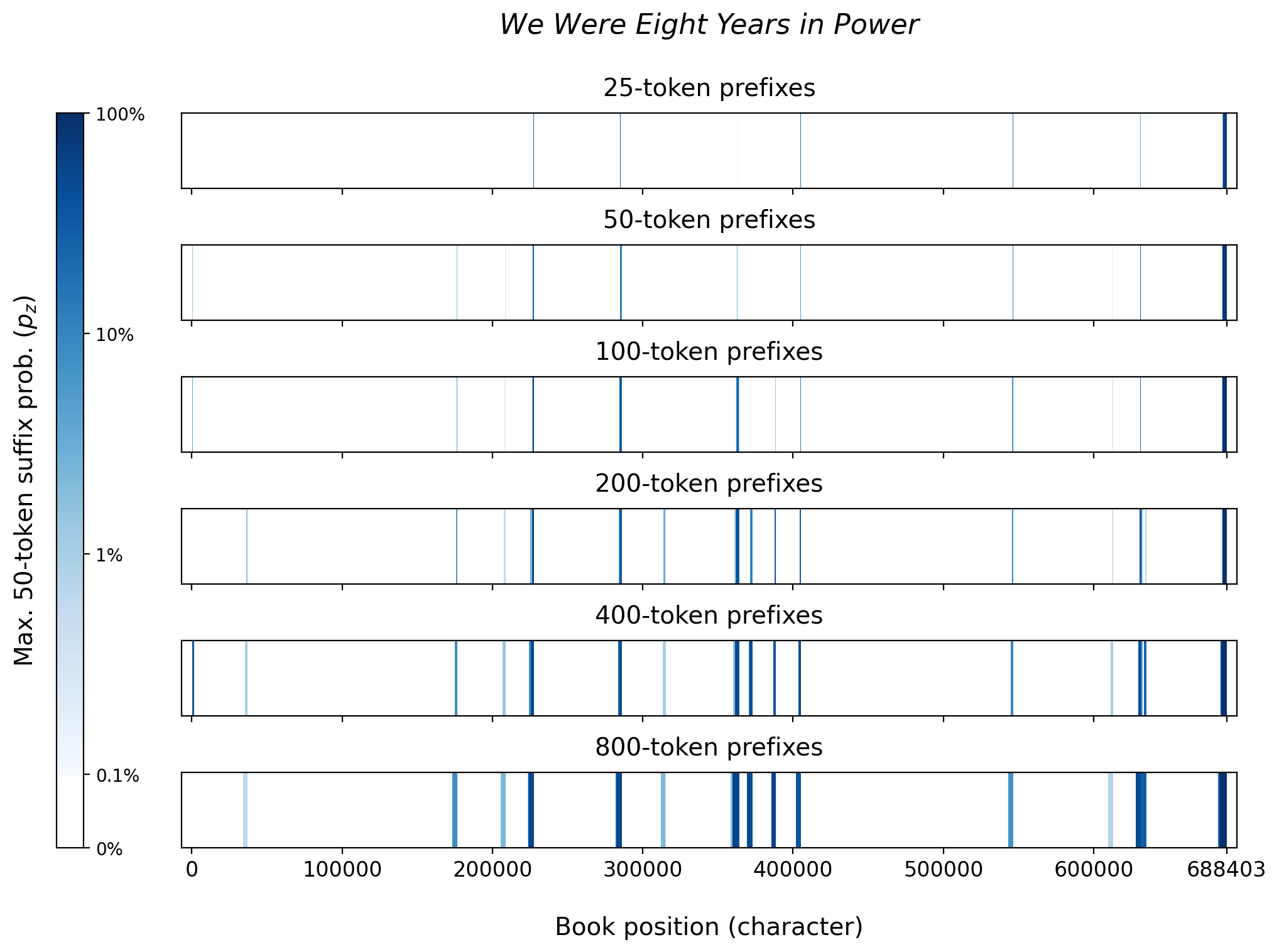}
\end{subfigure}
\caption{\textbf{Varying prefix for the \textsc{Llama 3.1 8B} baseline.} 
We run the sliding-window procedure for various prefix lengths for  $4$ books from \texttt{Books3}: \emph{Beloved}~\citep{Beloved}, \emph{Harry Potter and the Sorcerer's Stone}~\citep{Harry_Potter_and_the_Sorcerer_s_Stone}, \emph{This Is How You Lose Her}~\citep{This_Is_How_You_Lose_Her}, and \emph{We Were Eight Years in Power}~\citep{We_Were_Eight_Years_in_Power}.
\textsc{Llama} models are known to have been trained on \texttt{Books3}. 
We run probabilistic extraction on \textsc{Llama 3.1 70B} with top-$k$ ($T\!=\!1$, $k\!=\!40$) decoding for $50$-token suffixes and prefix lengths in $\{25, 50, 100, 200, 400, 800\}$. 
Extraction signal generally increases as prefix length increases.
However, this only appears to be the case if there is indeed more memorization to surface; 
increasing prefix length does not always make previously unextractable sequences extractable. 
There are also diminishing increases as the prefix lengthens.\looseness=-1
}
\label{fig:validity:baseline:llama-3.1-8b}
\vspace{-.25cm}
\end{figure*}

\begin{figure*}[t!]
\centering
\begin{subfigure}{0.48\linewidth}
\includegraphics[width=\linewidth]{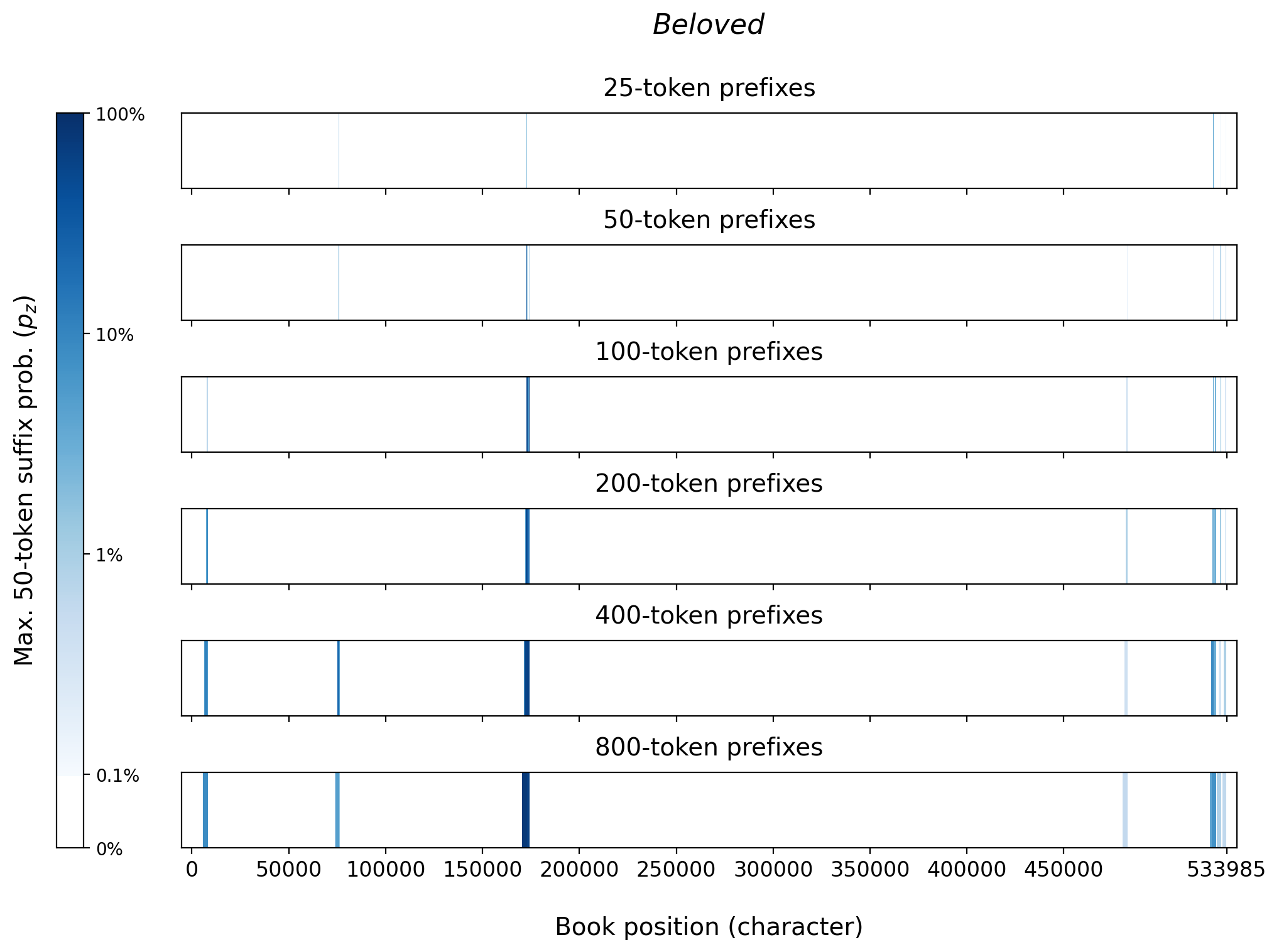} 
\end{subfigure}
\hspace{.25cm}
\begin{subfigure}{0.48\linewidth}
\includegraphics[width=\linewidth]{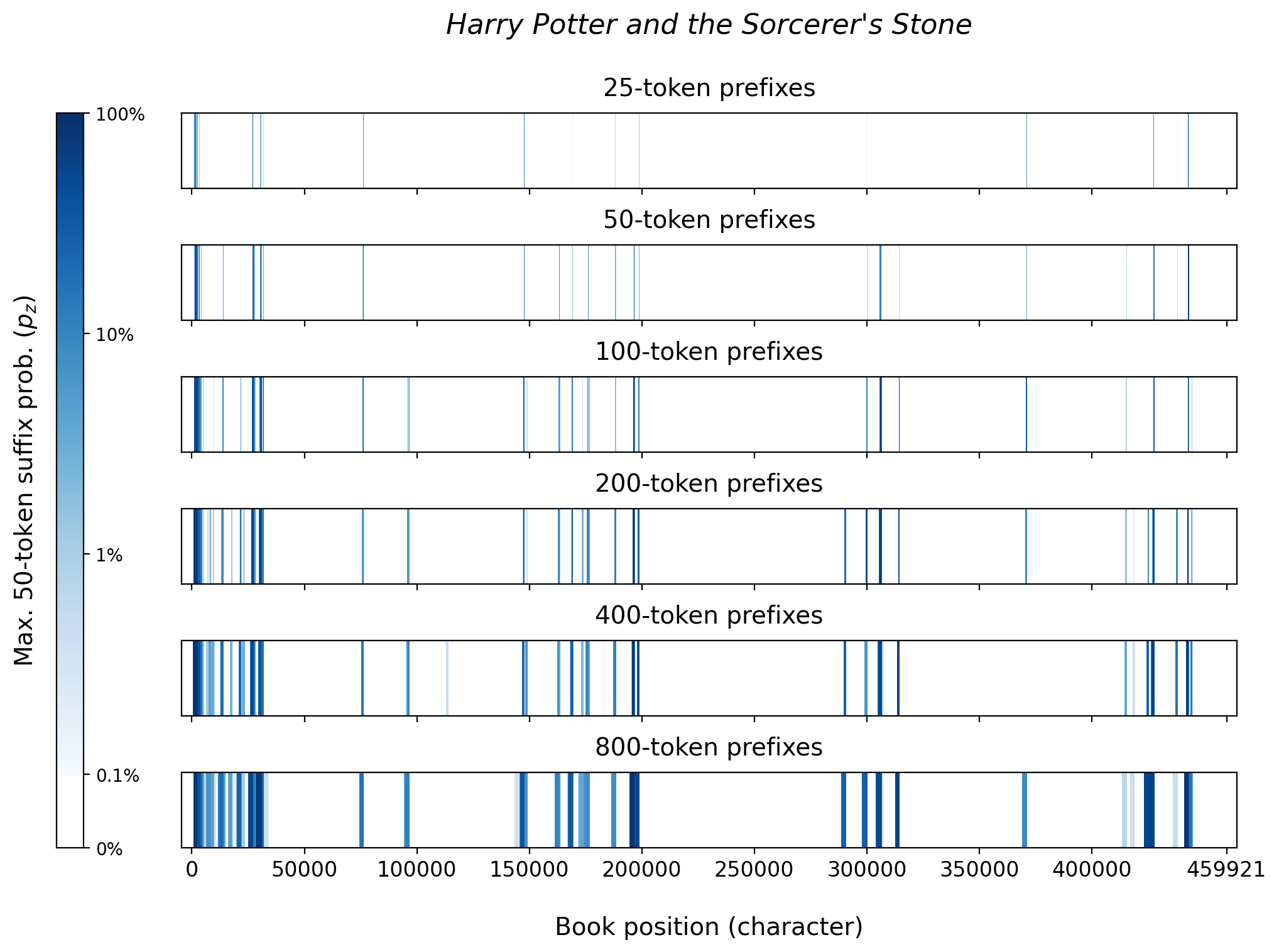}
\end{subfigure}
\begin{subfigure}{0.48\linewidth}
\includegraphics[width=\linewidth]{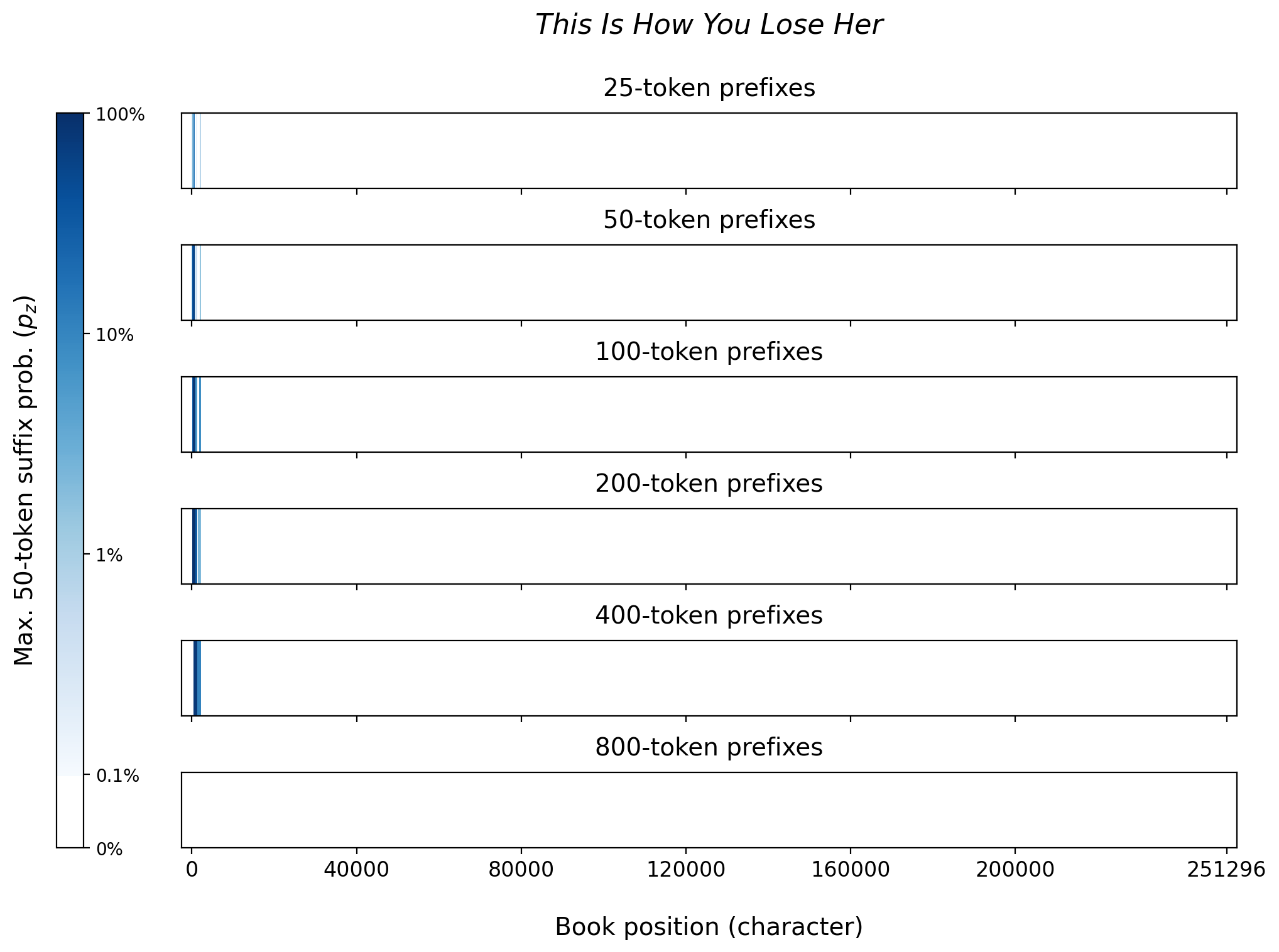} 
\end{subfigure}
\hspace{.25cm}
\begin{subfigure}{0.48\linewidth}
\includegraphics[width=\linewidth]{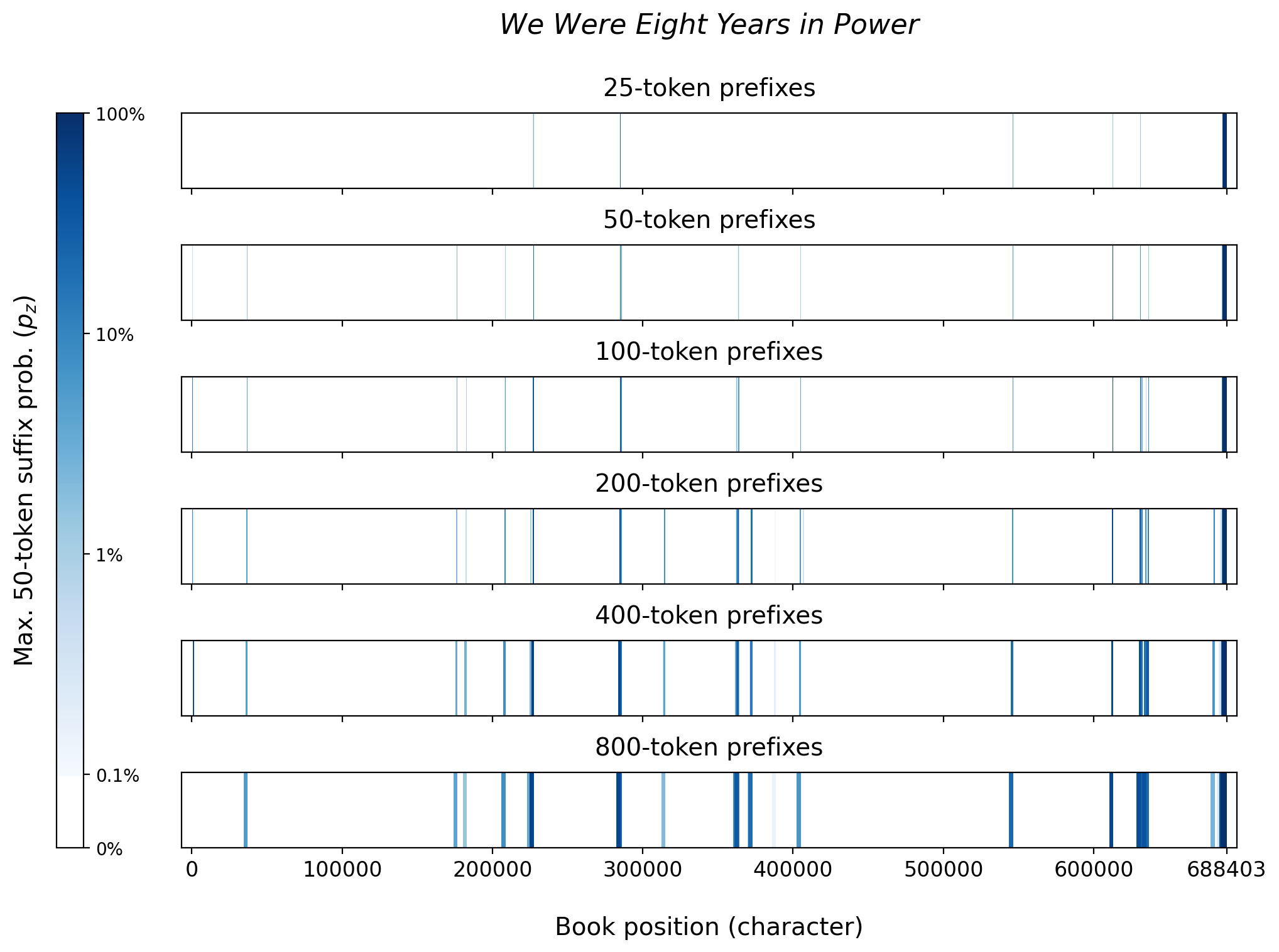}
\end{subfigure}
\caption{\textbf{Varying prefix for the \textsc{Llama 2 13B} baseline.} 
We run the sliding-window procedure for various prefix lengths for  $4$ books from \texttt{Books3}: \emph{Beloved}~\citep{Beloved}, \emph{Harry Potter and the Sorcerer's Stone}~\citep{Harry_Potter_and_the_Sorcerer_s_Stone}, \emph{This Is How You Lose Her}~\citep{This_Is_How_You_Lose_Her}, and \emph{We Were Eight Years in Power}~\citep{We_Were_Eight_Years_in_Power}.
\textsc{Llama} models are known to have been trained on \texttt{Books3}. 
We run probabilistic extraction on \textsc{Llama 3.1 70B} with top-$k$ ($T\!=\!1$, $k\!=\!40$) decoding for $50$-token suffixes and prefix lengths in $\{25, 50, 100, 200, 400, 800\}$. 
Extraction signal generally increases as prefix length increases.
However, this only appears to be the case if there is indeed more memorization to surface; 
increasing prefix length does not always make previously unextractable sequences extractable. 
There are also diminishing increases as the prefix lengthens.\looseness=-1
}
\label{fig:validity:baseline:llama-2-13b}
\vspace{-0cm}
\end{figure*}

We include results for $3$ different models on $4$ books from \texttt{Books3}, as documented in Appendix~\ref{app:sec:validity:setup}.
The largest model we test is \textsc{Llama 3.1 70B} (Figure~\ref{fig:validity:baseline:llama-3.1-70b}). 
For a point of comparison with a much smaller capacity model, we also include the smaller model from the same family---\textsc{Llama 3.1 8B} (Figure~\ref{fig:validity:baseline:llama-3.1-8b}) and 
\textsc{Llama 2 13B} (Figure~\ref{fig:validity:baseline:llama-2-13b}), since this model is of a similar size as \textsc{Phi 4}  (Section~\ref{sec:validity:controls}, Appendix~\ref{app:sec:validity:phi4}).\looseness=-1

\clearpage
\begin{figure}[t]
    \centering
    \includegraphics[width=0.8\linewidth]{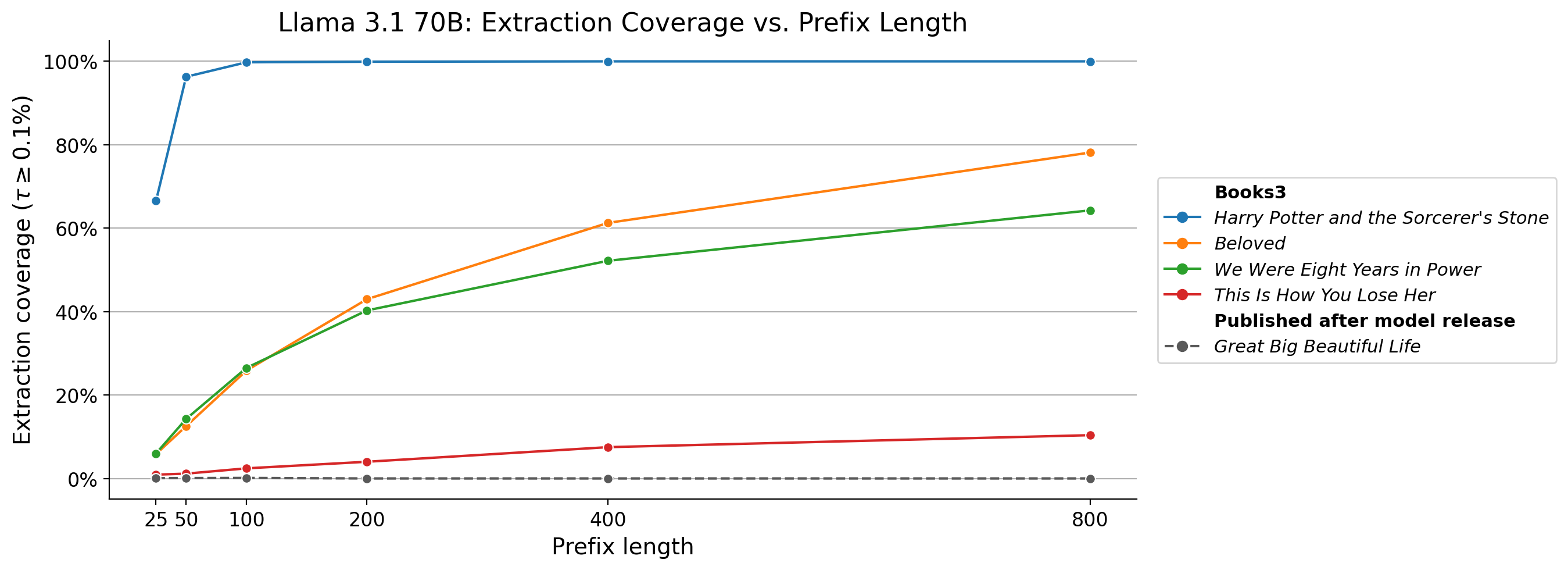}
    \caption{\textbf{Extraction coverage by prefix length.}
    Plotting extraction coverage for $\tau\!=\!\tau_\text{min}\!=\!0.1\%$ (Equation~\ref{app:eq:cov}) for the $5$ books we test with varying prefix lengths and \textsc{Llama 3.1 70B}. 
    Four of these books are in \texttt{Books3} and thus in \textsc{Llama}'s training data.
    The remaining book, \emph{Great Big Beautiful Life}, is used in our negative controls on non-training data; 
    extraction coverage is $0\%$ regardless of prefix length. 
    As we increase prefix length, extraction coverage for each book in the training data begins to level off.
    While \emph{Harry Potter and the Sorcerer's Stone} is entirely memorized by the model, it does not appear to be the case that \emph{Beloved}, \emph{We Were Eight Years in Power}, or \emph{This Is How You Lose Her} would be entirely extractable with respect to $50$-token suffixes and even longer prefixes.}
    \label{app:fig:prefix-coverage}
\end{figure}

\FloatBarrier

These results illustrate the high-level points described above. 
While for \textsc{Llama 3.1 70B} we are able to extract a lot more memorized sequences, we are able to extract relatively less from \textsc{Llama 3.1 8B}. 
That model has memorized less overall, which makes sense because it is a significantly smaller model. 
For \textsc{Llama 3.1 70B} and for some books, there is very little overall memorization that we can expose, regardless of prefix length.
We can see this with the results in Figure~\ref{fig:validity:baseline:llama-3.1-70b} for \emph{This Is How You Lose Her}. 
We also observe diminishing returns for increasing prompt length.
This suggests that there is a maximum number of sequences we can extract from each work, even in the limit of setting the prefix length especially long. 
We can show this with plots of extraction coverage (Equation~\ref{app:eq:cov}) as a function of prefix length, as in Figure~\ref{app:fig:prefix-coverage} (duplicated from Figure~\ref{fig:prefix-coverage} in the main paper).

\FloatBarrier

\subsection{Results of negative control experiments}\label{app:sec:validity:controls}

We provide additional results for our varied-prefix-length negative control experiments (Section~\ref{sec:validity:controls}) on both \textsc{Phi 4} (Appendix~\ref{app:sec:validity:phi4}) and non-training data that post-dates training cutoffs (Appendix~\ref{app:sec:validity:cutoff}).
While these experiments provide strong evidence for the validity of our procedure, we would want to run much more extensive experiments to responsibly lower $\tau_\text{min}$ or to make it LLM-specific. 
We defer this to future work, and here opt to pick a conservative global $\tau_\text{min}\!=\!0.1\%$, which we discuss also in Section~\ref{sec:validity}.
The experiments that we run here are of a more substantial scale than the original probabilistic discoverable extraction paper by \citet{hayes2025measuringmemorizationlanguagemodels}. 
That work performed one experiment on non-training data on a small \textsc{Pythia} model.
The paper uses the results of this experiment as a basis to both validate the experiments on known training data in \textsc{Pythia} models, and to apply the methodology to much larger (and higher quality) \textsc{Llama 1} and \textsc{OPT} models---i.e., to make claims about extraction of not-known-with-certainty training data, just as we do here with \textsc{DeepSeek v1 67B}, \textsc{Qwen 2.5 72B}, and \textsc{Gemma 2 27B}. 

\citet{hayes2025measuringmemorizationlanguagemodels} finds that ``\ldots the number of queries $n$ needed to generate unseen test data is orders of magnitude larger than for generating training data. [They] find that it is generally challenging to generate test data \ldots This supports that, in [their] measurements of [probabilistic] discoverable extraction, matches between training-example targets and generated suffixes are almost surely due to memorization.''
We observe the same pattern, though instead of discussing it with respect to the number of queries $n$, as elsewhere, we frame our discussion in terms of $p_\vz$.
In our results, the $p_\vz$ that we observe on non-training data are many orders of magnitude lower than what we deem extraction success in our experiments on known training data. 
That is, our measurements for extraction are not erroneously capturing chance generation of non-training data. 
As in \citet{hayes2025measuringmemorizationlanguagemodels}, this gives us assurance that we can use our approach on models for which we do not know exactly what was included in the training data.
We discuss this further in Appendices~\ref{app:sec:validity:monkey} and~\ref{app:sec:validity:membership}.\looseness=-1 

\subsubsection{Results on \textsc{Phi 4}}\label{app:sec:validity:phi4}

We ran a large number of negative control experiments using \textsc{Phi 4}~\citep{phi4}.
As noted in Appendix~\ref{app:sec:sliding-window:setup}, this model was \emph{not} trained on whole copyrighted books.
It was trained on a mix of synthetic data, licensed websites, and public domain books. 
When running our sliding-window extraction procedure on this model for copyrighted books, we can expect (generally speaking, with important exceptions) that unique text from copyrighted books will not register as extractable with our measurements.

\begin{figure*}[t]
\centering
\begin{subfigure}{0.48\linewidth}
\includegraphics[width=\linewidth]{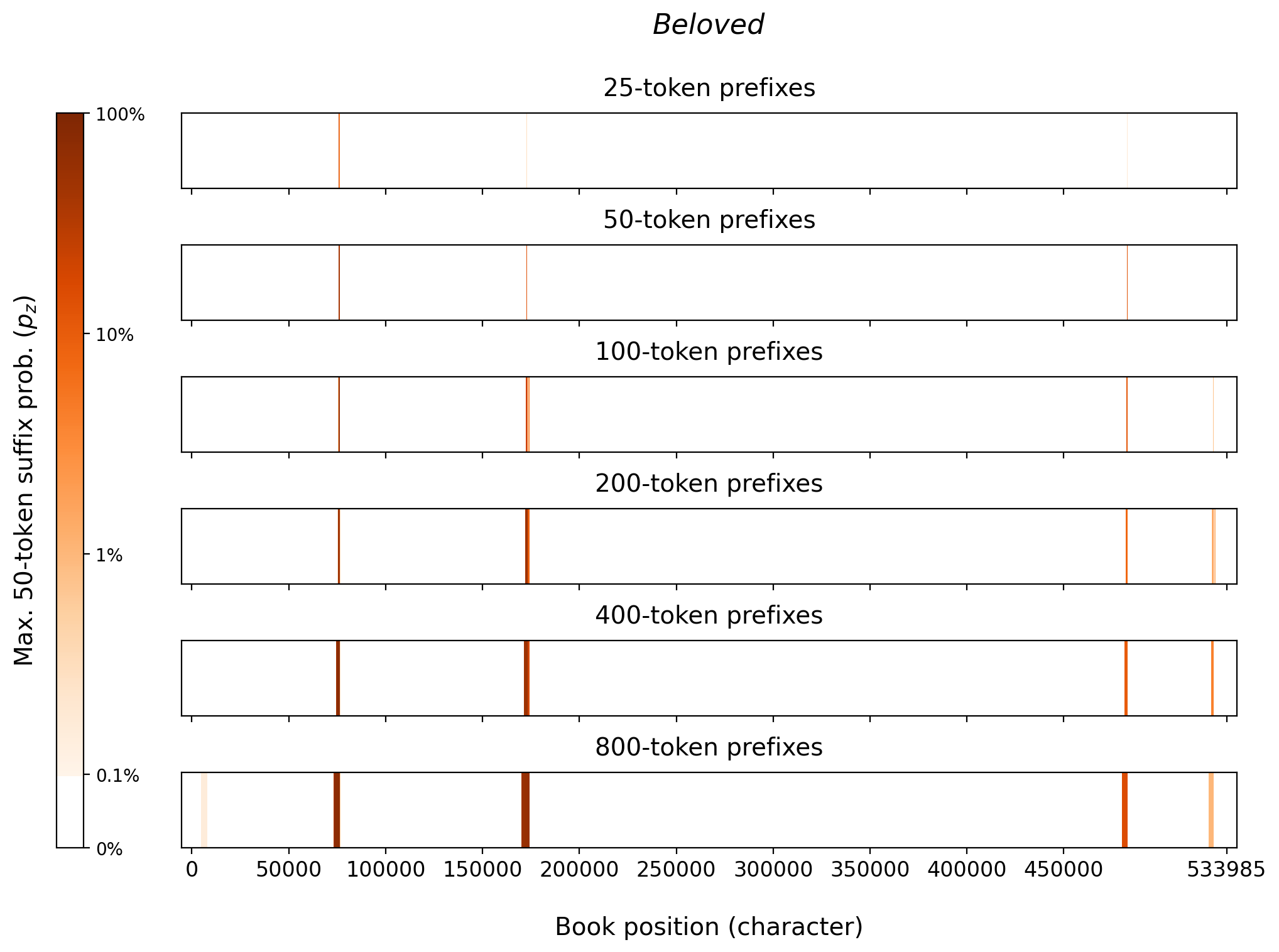}    
\end{subfigure}
\hspace{.25cm}
\begin{subfigure}{0.48\linewidth}
\includegraphics[width=\linewidth]{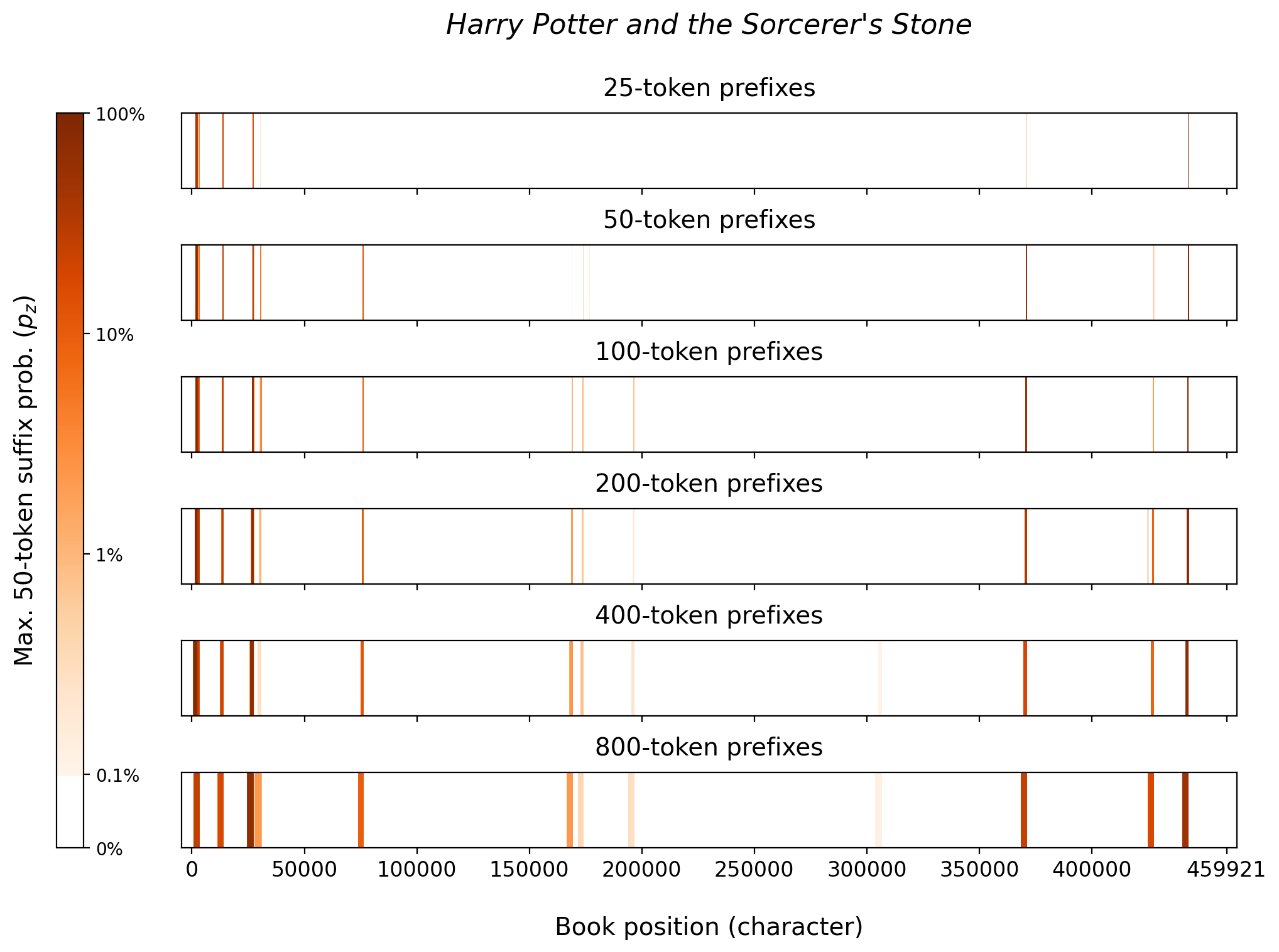}
\end{subfigure}
\begin{subfigure}{0.48\linewidth}
\includegraphics[width=\linewidth]{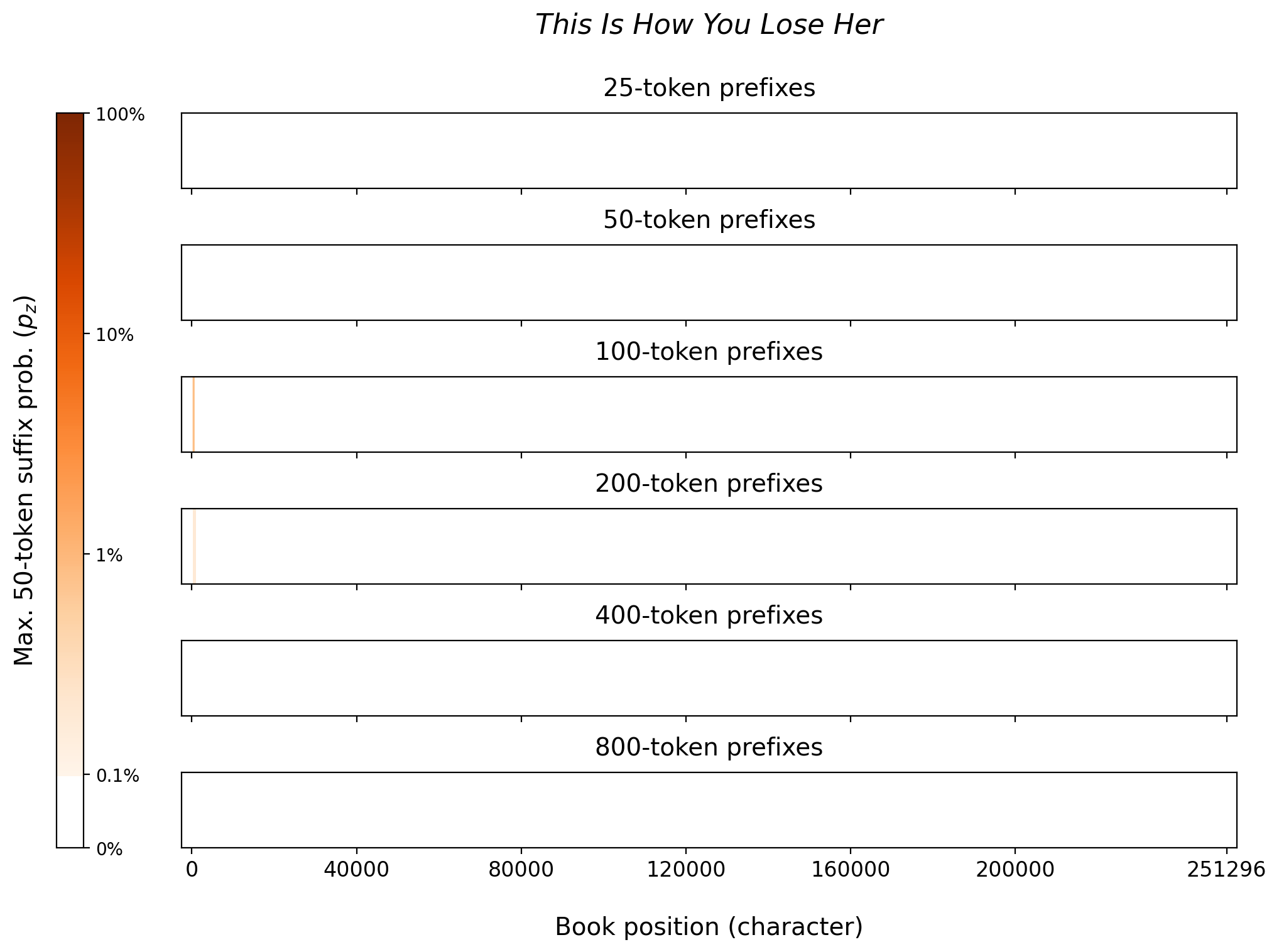}    
\end{subfigure}
\hspace{.25cm}
\begin{subfigure}{0.48\linewidth}
\includegraphics[width=\linewidth]{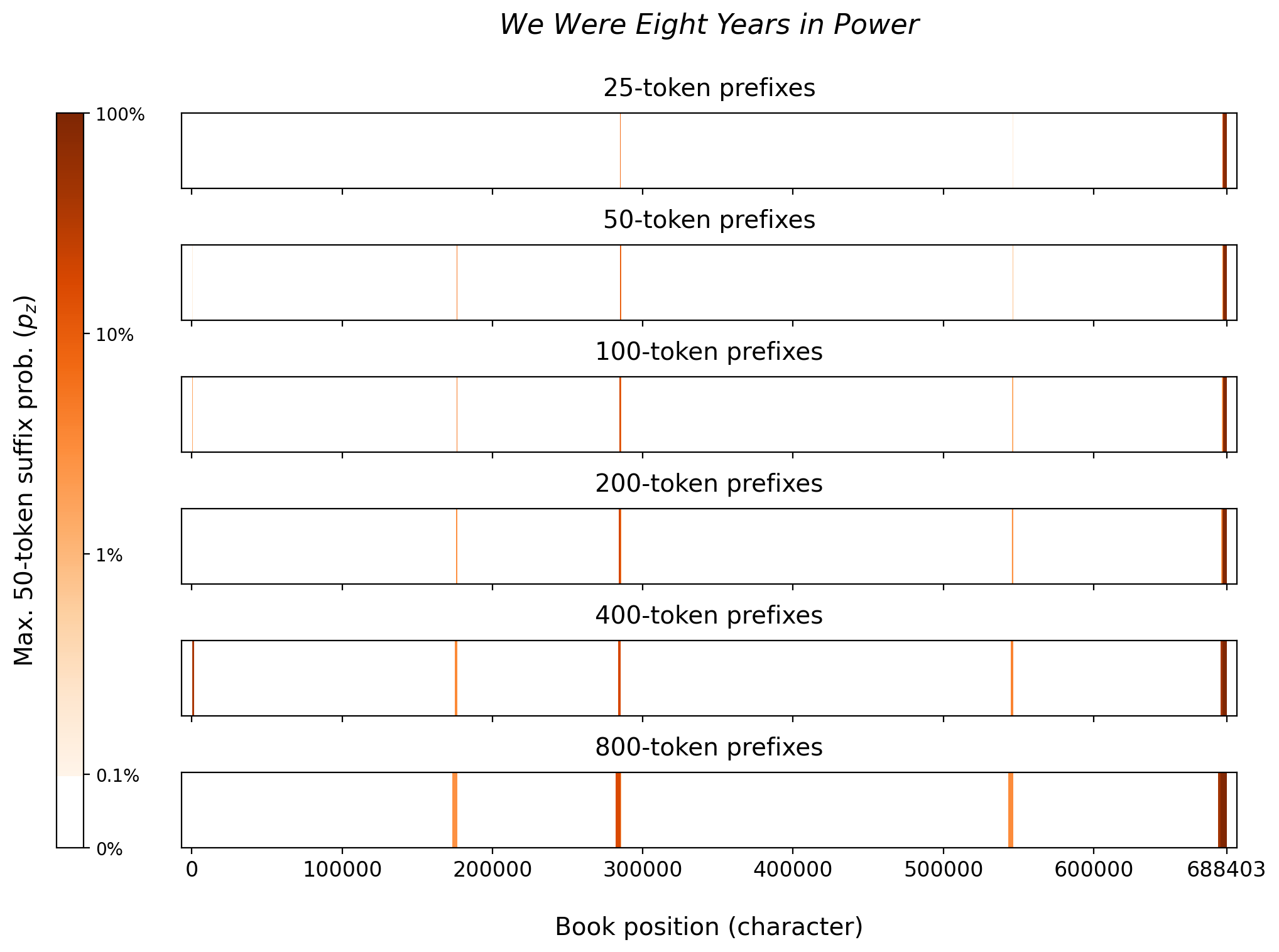}
\end{subfigure}
\caption{\textbf{\textsc{Phi 4} negative controls.} 
Sliding-window procedure for various prefix lengths. 
We show results for $4$ books from \texttt{Books3}: \emph{Beloved}~\citep{Beloved}, \emph{Harry Potter and the Sorcerer's Stone}~\citep{Harry_Potter_and_the_Sorcerer_s_Stone}, \emph{This Is How You Lose Her}~\citep{This_Is_How_You_Lose_Her}, and \emph{We Were Eight Years in Power}~\citep{We_Were_Eight_Years_in_Power}.
\textsc{Phi 4} was deliberately not trained on whole copyrighted books.
We run probabilistic extraction on \textsc{Llama 3.1 70B} with top-$k$ ($T\!=\!1$, $k\!=\!40$) decoding for $50$-token suffixes and prefix lengths in $\{25, 50, 100, 200, 400, 800\}$. 
Extraction signal generally increases as prefix length increases.
However, this only appears to be the case if there is indeed more memorization to surface; 
increasing prefix length does not always make previously unextractable sequences extractable. 
There are also diminishing increases as the prefix lengthens.
We manually verified that every extractable sequence comes from highly duplicated text that can easily be found on the Internet;
such sources reasonably were a part of \textsc{Phi 4}'s training data.
}
\label{fig:validity:control:phi-4}
\vspace{-.25cm}
\end{figure*}

For our main results using the sliding-window procedure with $100$-token sequences (Appendix~\ref{app:sec:sliding-window:results}), we include \textsc{Phi 4} as one of the $14$ models that we evaluate.
Here, in Figure~\ref{fig:validity:control:phi-4}, we show additional results for experiments on varied prefixes and the $4$ in-copyright books from \texttt{Books3}. 
We are able to extract non-trivial amounts of public domain books from this model (e.g., Appendix~\ref{app:sec:sliding:The_Great_Gatsby}).
In contrast, we are not able to extract much unique text from in-copyright books.
As elsewhere, we extract boilerplate text like copyright notices and publishers' addresses. 
And we are also able to extract famous or otherwise popular quotes---either quoted in in-copyright books, or from the books themselves.\looseness=-1

We (painstakingly) verify that every sequence that we extract from \textsc{Phi 4} can be easily found in multiple sources on the Internet. 
That is, the quotes from books that we extract from \textsc{Phi 4} are either in the public domain, quoted widely on blogs or platforms like Goodreads, or both. 
This is indicative (though not definitive) that these quotes could have found their way into \textsc{Phi 4}'s training data from these online sources, rather from the books themselves. 

For further support, we can also compare the text at the book locations for which we extract these sequences to heatmaps other models. 
We note that, almost always, these sequences are extractable (sometimes with enormous probability) from \emph{all} models we test. 
Again, while this is not definitive, it is suggestive that this text is highly duplicated in sources that are a part of common pre-training corpora---that they are not necessarily memorized \emph{because} they are in \texttt{Books3}, but because of their duplicated inclusion in the training data from other sources.\looseness=-1 

We also observe similar challenges---to a somewhat lesser extent---for our negative control experiments on books published after LLM training-date cutoffs.
These books sometimes cite earlier sources---other books, news articles, social media posts, etc.---that pre-date the training-date cutoffs and are widely duplicated online. 
So, even this seemingly ``clean'' case also requires significant manual verification to confirm validity, which we discuss in Appendix~\ref{app:sec:validity:cutoff}.\looseness=-1

Altogether, it is reasonable to conclude that these are true instances of extraction. These results reinforce that our procedure detects true memorization when it exists and does not spuriously flag text where memorization is impossible. 
Even so, these results show just how challenging it is to run completely clean negative controls for models trained on web-scraped data~\citep{lee2023explainers}, for which duplicates of the same piece of text across various sources complicate what training-data membership means.
A given book may not have been a member of the training data, but this does not mean that all text from that book was not.
In these cases, it seems extraordinarilyy likely that these sequences are members of the training data (and memorized). 
We discuss this further in Appendix~\ref{app:sec:validity:membership}.

With the exception of the copyright notice, we are unable to extract any sequences from \emph{This Is How You Lose Her}~\citep{This_Is_How_You_Lose_Her}.
The sequences that we extract from \emph{Beloved} are all posted (multiple times) on platforms like Goodreads.
We discuss some of these quotes in Section~\ref{sec:book:compare}~\citep{belovedquote1, belovedquote2, belovedquote3}.
\emph{Harry Potter and the Sorcerer's Stone}~\citep{Harry_Potter_and_the_Sorcerer_s_Stone} is excerpted and quoted in an extraordinary number of places on the Internet. 
We were able to find every extracted sequence on $3$ different web pages (after which we stopped searching/counting).\looseness=-1

\FloatBarrier

We discuss \emph{When We Were Eight Years in Power}~\citep{We_Were_Eight_Years_in_Power} in Section~\ref{sec:validity:controls}. 
As noted there, regardless of prefix length, there are only $4$ areas---other than the copyright notice at the beginning of the book---that show extractable sequences. 
The sequences at the end of the book contain a list of ordered (page) numbers. 
This kind of list is a very common artifact from converting an \textsc{epub} file to \textsc{txt}---as was done for the collection of \texttt{Books3}~\citep{reisnerbooks3}. 
Like the copyright notice, such ordered lists of numbers are text that, while high probability, we can discount as being particularly interesting with respect to our negative controls~\citep{carlini2021extracting}.
The other $3$ are all text that is \emph{not} unique to book, but rather highly duplicated text from the Internet (some of which is in the public domain):\looseness=-1 
\begin{itemize}[leftmargin=0.65cm]
    \item $\sim\!176$K: An exact quote from Malcom X's speech, ``Who Taught You To Hate Yourself?'' (May 5, 1962)~\citep{malcolmx}
    \item $\sim\!280$K: An exact quote of Deuteronomy 15:125 from the Bible -- King James Version (KJV)~\citep{deuteronomy}.
    \item $\sim\!546$K: A exact quote from Barack Obama's 2004 DNC Keynote: ``hope of a young naval lieutenant bravely patrolling the Mekong Delta; the hope of a mill worker's son who dares to defy the odds; the hope of a skinny kid with a funny name who believes that America has a place for''~\citep{obama2004}
    \vspace{-.2cm}
\end{itemize}
All of these sequences are also highly memorized for other models we test. 

\begin{figure*}[t]
\centering
    \includegraphics[width=\linewidth]{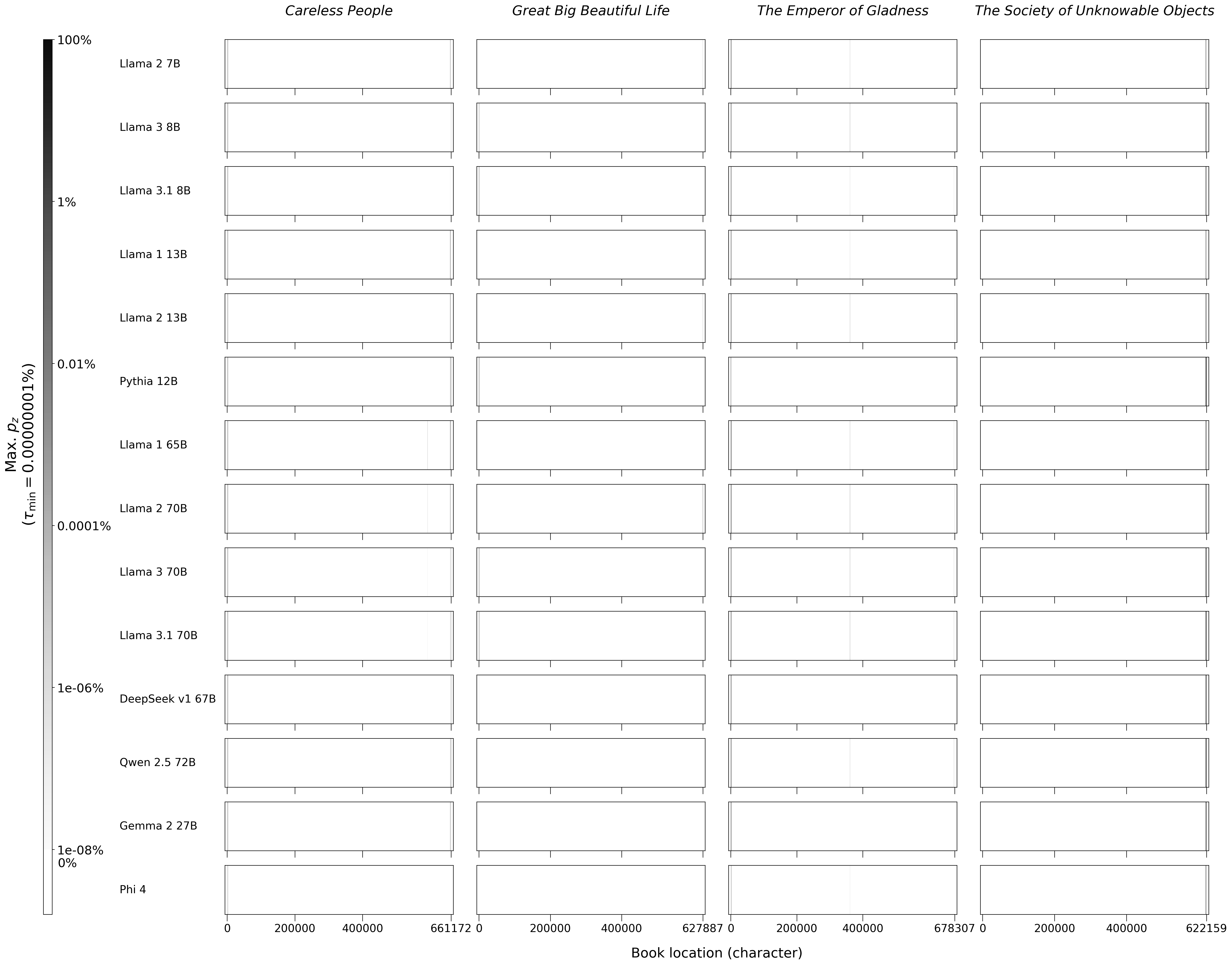}
    \caption{\textbf{Post-training cutoff negative controls.}
    We measure the probability of generating target $50$-token suffixes for $50$-token prefixes (using top-$k$, as elsewhere) for  $4$ books published in 2025: 
    \emph{Careless People}~\citep{careless}, \emph{Great Big Beautiful Life}~\citep{greatbig}, \emph{The Emperor of Gladness}~\citep{emperorgladness}, and \emph{The Society of Unknowable Objects}~\citep{unknowableobjects}. 
    All of these books were published after the release dates for all models we evaluate. 
    These experiments serve as a validity check that our measurements for probabilistic extraction are capturing true instances of extraction. 
    By definition, no unique sequences in these books can be memorized because they are not members of the training data; 
    we should not be able to (and do not) obtain high, non-zero probabilities---probabilities that we would consider to be successful instances of extraction for known or suspected training data. 
    Note that, in contrast to our other heatmaps, we set a more permissive (i.e., lower) minimum for $p_\vz$ ($p_\vz\geq0.00000001\%$, or $1$ in $10$ billion generations), in order to see if there are sequences that can be generated even with fairly low probability that we would \emph{not} count as successful extraction.\looseness=-1} 
    \label{fig:2025-books}
    \vspace{-.5cm}
\end{figure*}
\FloatBarrier

\subsubsection{Results on books from after training cutoffs}\label{app:sec:validity:cutoff}

We run additional negative control experiments on $4$ books that post-date the training of all models.
As a result, these books could not have been included in the training data;
we similarly should not expect to extract unique text from them for the different models we evaluate. 
However, as with our experiments on \textsc{Phi 4}, there are again cases where boilerplate text and popular quotes from other sources come up in our measurements.

Nevertheless, we do not detect extraction for \emph{any} text unique to any of these $4$ books for any model we test. 
In Figure~\ref{fig:2025-books}, we show results for all $4$ books using $100$-token sequences ($50$-token prefixes $+$ $50$-token suffixes) for the $14$ models that we include for our sliding-window experiments on \texttt{Books3} books (Appendix~\ref{app:sec:sliding-window}). 
Note that in Figure~\ref{fig:2025-books}, we expand the range of $p_\vz$ that we plot to orders of magnitude below $\tau_\text{min}{=}0.1\%$.
Even so, there are only a select few regions that register with non-zero probabilities that surpass $\tau_\text{min}$, and they are duplicated generic text---in this case, the copyright notice---and some references drawn from pre-2025 sources. 
For example, for \emph{The Emperor of Gladness}~\citep{emperorgladness}, we extract a verbatim quote from the Bible (Mark 5:13). 
For \emph{Great Big Beautiful Life}~\citep{greatbig} and \emph{The Society of Unknowable Objects}~\citep{unknowableobjects}, we only extract boilerplate text. 

For \emph{Careless People}~\citep{careless}, we extract the ``careless people'' quote from \emph{The Great Gatsby}~\citep{The_Great_Gatsby} (Figure~\ref{fig:header}). 
It is quoted at the beginning of the book (e.g., $p_\vz{=}2.8\%$ for \textsc{Llama 3.1 70B}):
\begin{quote}
\small
    \textbf{Prefix}:  to all this.\verb|\n&\n|For my grandmother Eileen\verb|\n|Who regularly reminds us to \verb|“|live an ordinary life\verb|”| and \verb|“|enjoy the good times.\verb|”\nEpigraph\n|\textcolor{blue}{They were careless people, Tom and Daisy—they smashed up things and creatures and then}

    \textbf{Suffix}: \textcolor{blue}{retreated back into their money or their vast carelessness, or whatever it was that kept them together, and let other people clean up the mess they had made.}\verb|\n—|F. SCOTT FITZGERALD, THE GREAT GATSBY\verb|\n|
\end{quote}
This is what we register as extracted close to the $0$-character position in the heatmaps. 
(We highlight the portion from \emph{The Great Gatsby} in \textcolor{blue}{blue}.) 
The book also quotes Mark Zuckerberg in several locations---quotes that we confirm are often drawn from Facebook posts that  are widely duplicated on social media platforms and in online newspaper articles. 
These make up the remaining extracted sequences. 
For example, one quote is from a 2016 post about a SpaceX rocket explosion~\citep{spacex} (e.g., $p_\vz{=}13.5\%$ for \textsc{Llama 1 65B}):
\begin{quote}
\small
    \textbf{Prefix}: s SpaceX rocket explodes on the launchpad, completely destroying the Internet.org satellite it\verb|’|s supposed to be putting in orbit. Mark posted,\verb|\n|\textcolor{blue}{As I’m here in Africa, I’m deeply disappointed to}

    \textbf{Suffix}: \textcolor{blue}{hear that SpaceX’s launch failure destroyed our satellite that would have provided connectivity to so many entrepreneurs and everyone else across the continent. Fortunately, we have developed other technologies like Aquila that will connect people as well.}
\end{quote}
(We highlight the quoted portion from post in \textcolor{blue}{blue};
even just a few words from the quote in the prefix are sufficient to generate the suffix with high probability.) 
Another is from a widely quoted (in news articles) Facebook press release~\citep{release}:
\begin{quote}
\small
    \textbf{Suffix}: people based on their emotional state.\verb|\n|The analysis done by an Australian researcher was intended to help marketers understand how people express themselves on Facebook. It was never used to target ads and was based on data that was anonymous and aggregated
\end{quote}
An example of a sequence below $\tau_\text{min}$---i.e., that we would not count as extractable, but that we include with our expanded range in Figure~\ref{fig:2025-books}--- for \textsc{Llama 1 65} exhibits $p_\vz{=}0.02\%$ for a Mark Zuckerberg post about the Charlie Hebdo headquarters attack:
\begin{quote}
\small
    need to reject—a group of extremists trying to silence the voices and opinions of everyone else around the world. I won’t let that happen on Facebook. I’m committed to building a service where you can speak freely without fear of violence~\citep{hebdo}
\end{quote}
Mark Zuckerberg's manifesto is also quoted~\citep{manifesto}.

We know the \emph{The Great Gatsby} quote is in \texttt{Books3}, and widely quoted elsewhere.  
Similar to our observations for \textsc{Phi 4}, it is reasonable to assume that these other highly duplicated quotes are similarly in the training data, and that these are true instances of extraction. 
Overall, consistent with \citet{hayes2025measuringmemorizationlanguagemodels}, we do not observe chance generation of non-training data until we significantly lower the probability threshold. 
In Figure~\ref{fig:2025-books}, we observe no such instances for $p_\vz\geq10^{-8}\%$, which is $7$ orders of magnitude smaller than our chosen $\tau_\text{min}$.\looseness=-1 

\begin{figure*}[t!]
\centering
\includegraphics[width=.75\linewidth]{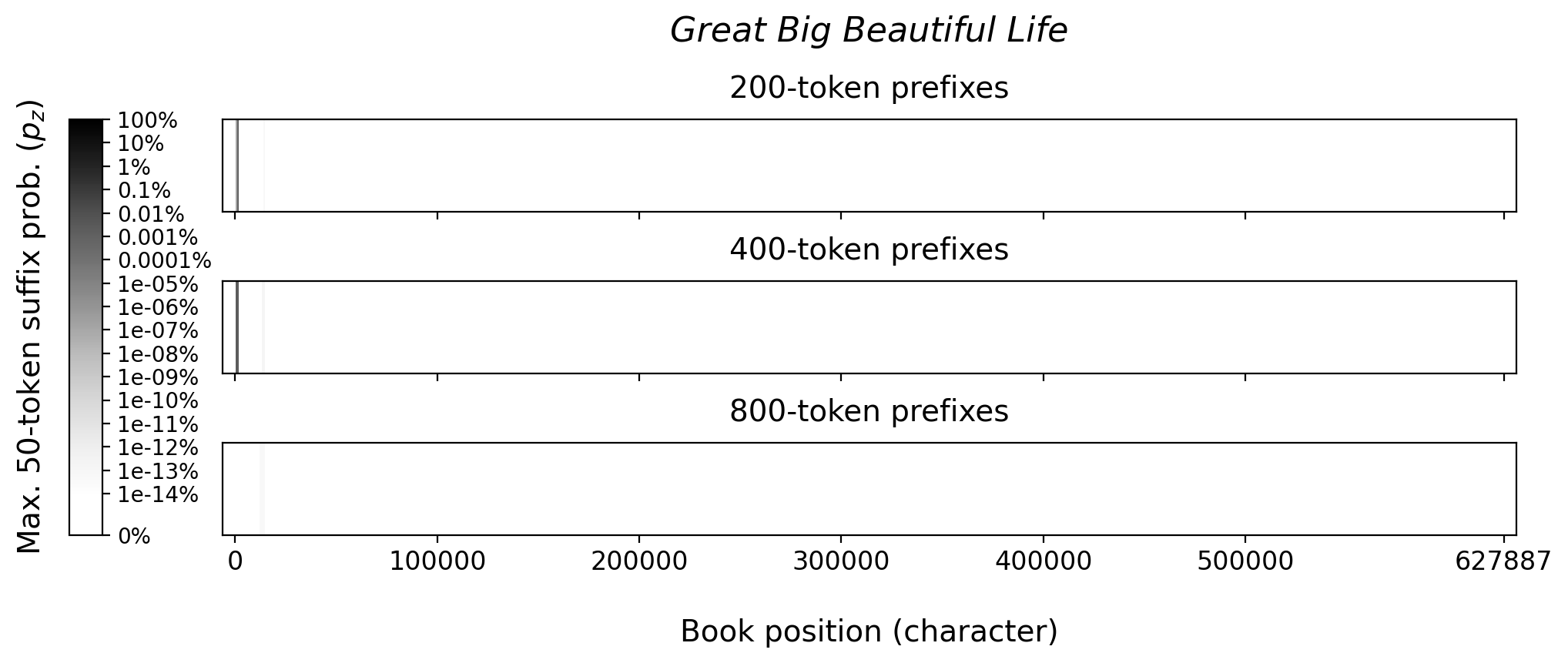}
\vspace{-.2cm}
\caption{\textbf{Chance generation of non-training data.}
We show \textsc{Llama 3.1 70B} and top-$k$ ($T\!=\!1$, $k\!=\!40$) decoding on \emph{Great Big Beautiful Life}~\cite{greatbig} (published spring 2025), with respect to $50$-token suffixes and prefix lengths in $\{200, 400, 800\}$ tokens. 
Except for the copyright notice at the beginning, we do not register any $p_\vz\!\geq\!\tau_\text{min}\!=\!0.1\%$.
We register no other suffix probabilities $p_\vz\!\geq\!10^{-13}\%$, and only $1$ (out of ${\sim}62{,}000$ sequences) for which $p_\vz{\geq}10^{-14}\%$. (See the faint gray band near $20$K characters.)}
\label{app:fig:greatbig:main}
\end{figure*}

\begin{figure*}[t]
\centering
\begin{subfigure}{0.48\linewidth}
\includegraphics[width=\linewidth]{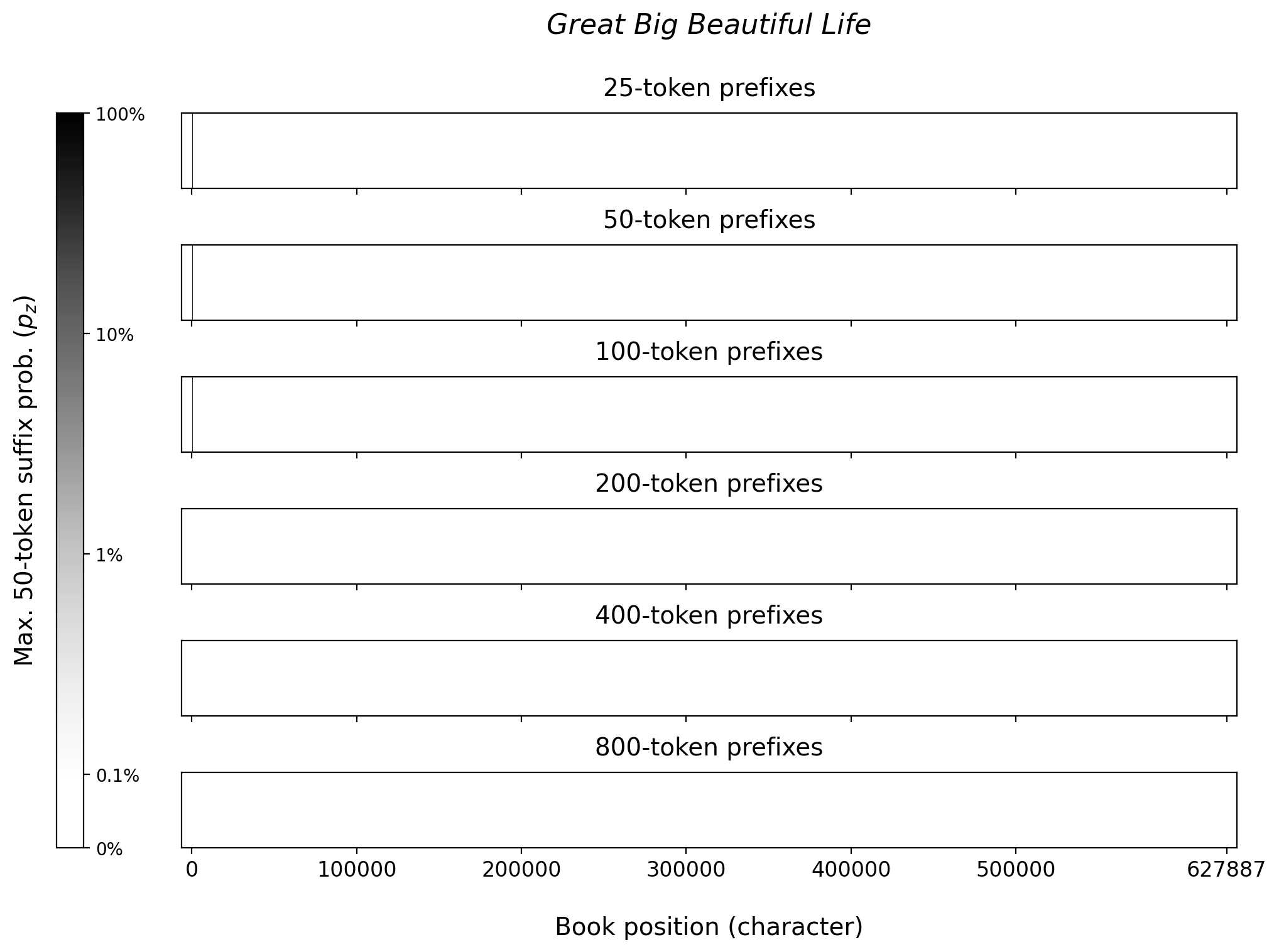}  
\caption{\textsc{Llama 3.1 70B}}
\end{subfigure}
\hspace{.25cm}
\begin{subfigure}{0.48\linewidth}
\includegraphics[width=\linewidth]{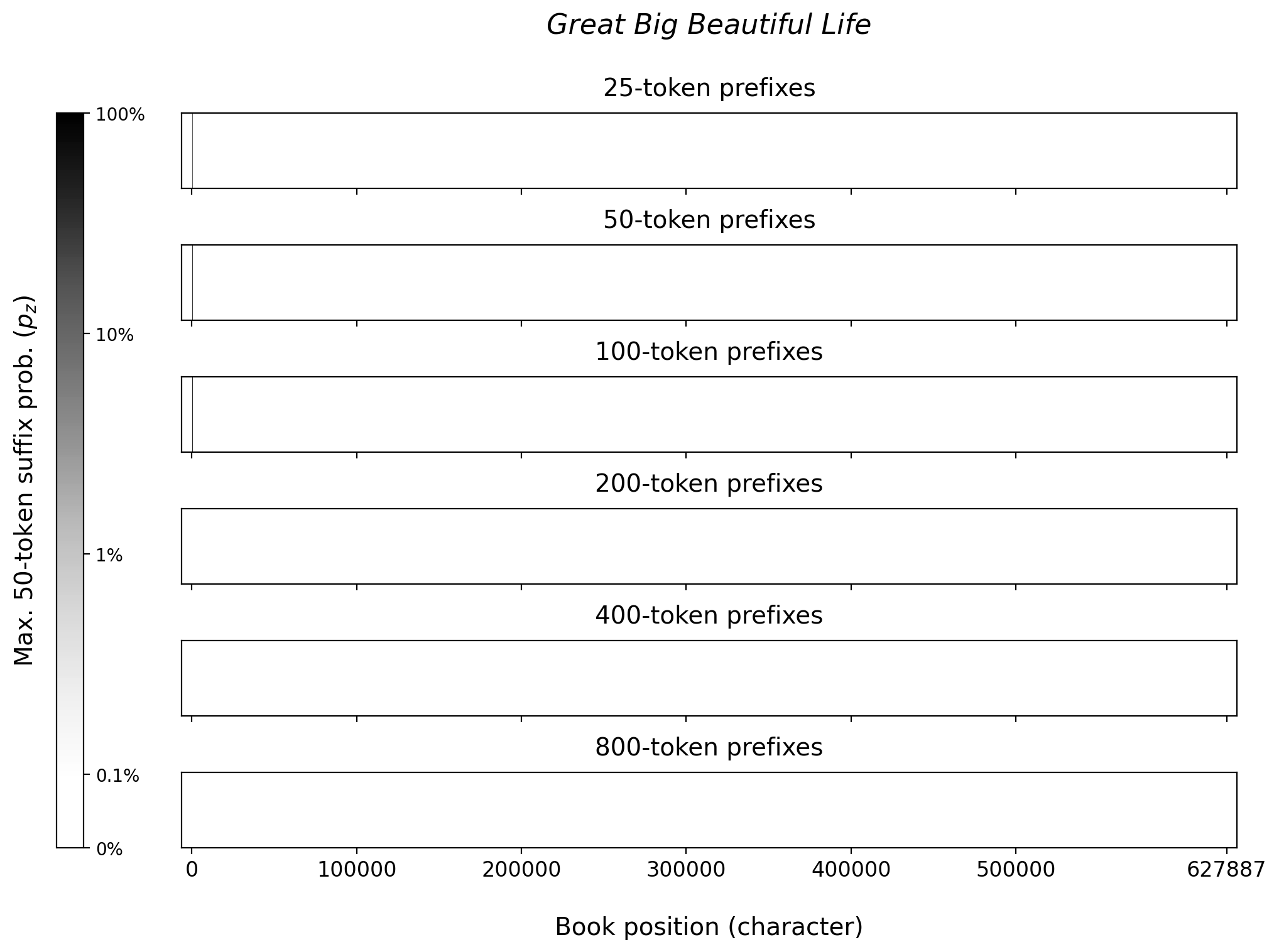}
\caption{\textsc{Llama 3.1 70B}}
\end{subfigure}
\begin{subfigure}{0.48\linewidth}
\includegraphics[width=\linewidth]{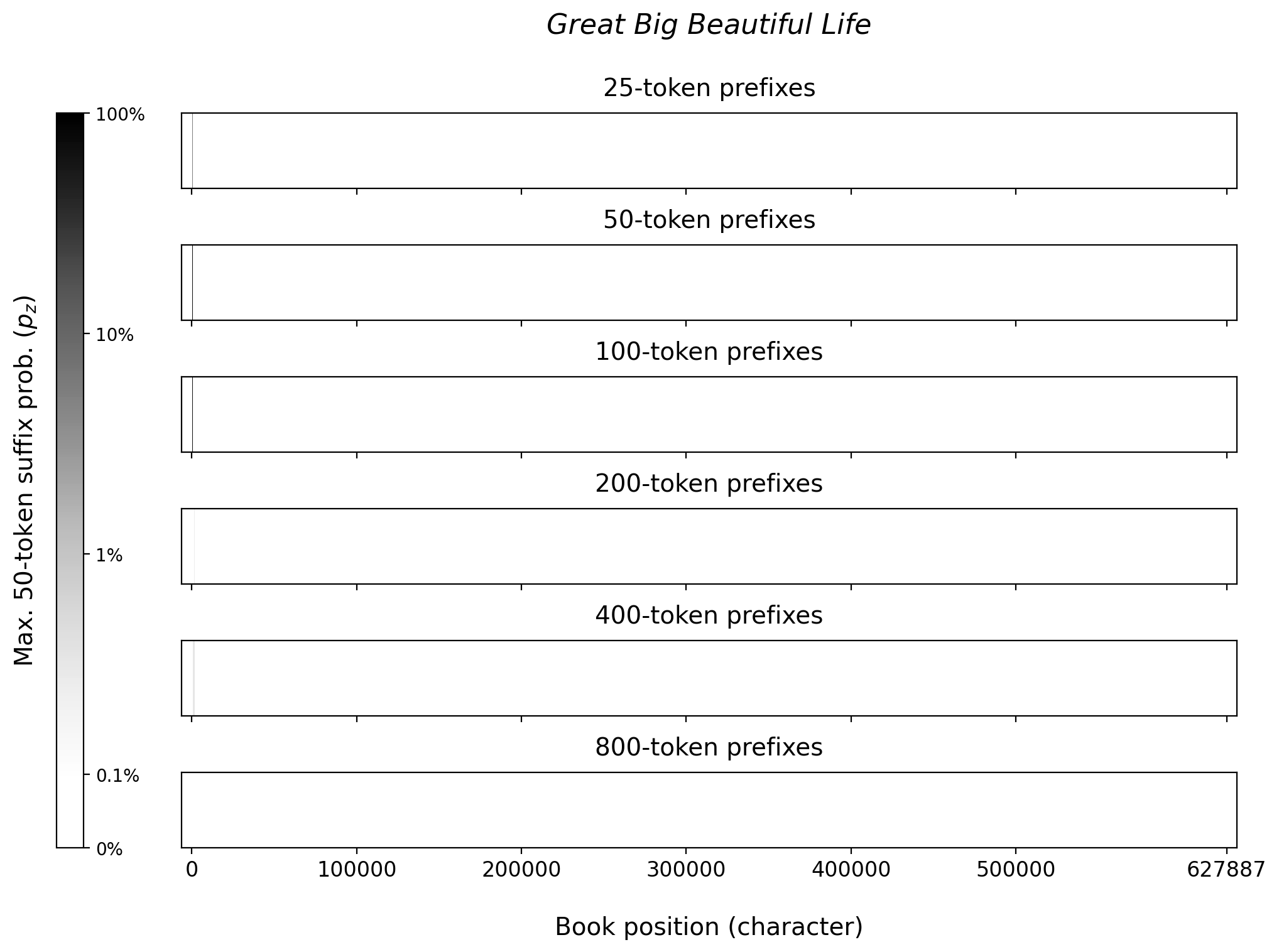}
\caption{\textsc{Llama 2 13B}}
\end{subfigure}
\hspace{.25cm}
\begin{subfigure}{0.48\linewidth}
\includegraphics[width=\linewidth]{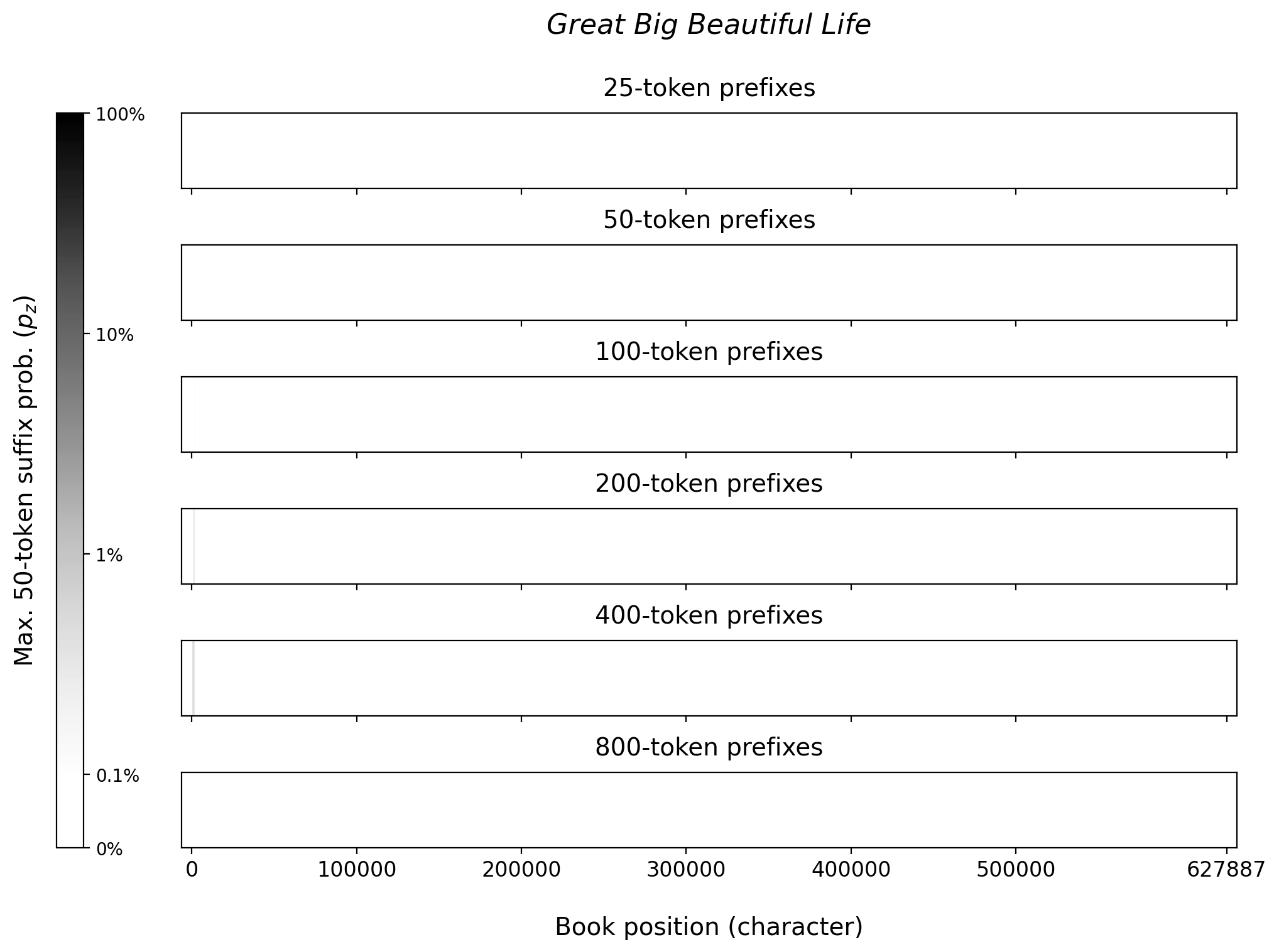}
\caption{\textsc{Phi 4}}
\end{subfigure}
\begin{subfigure}{0.48\linewidth}
\includegraphics[width=\linewidth]{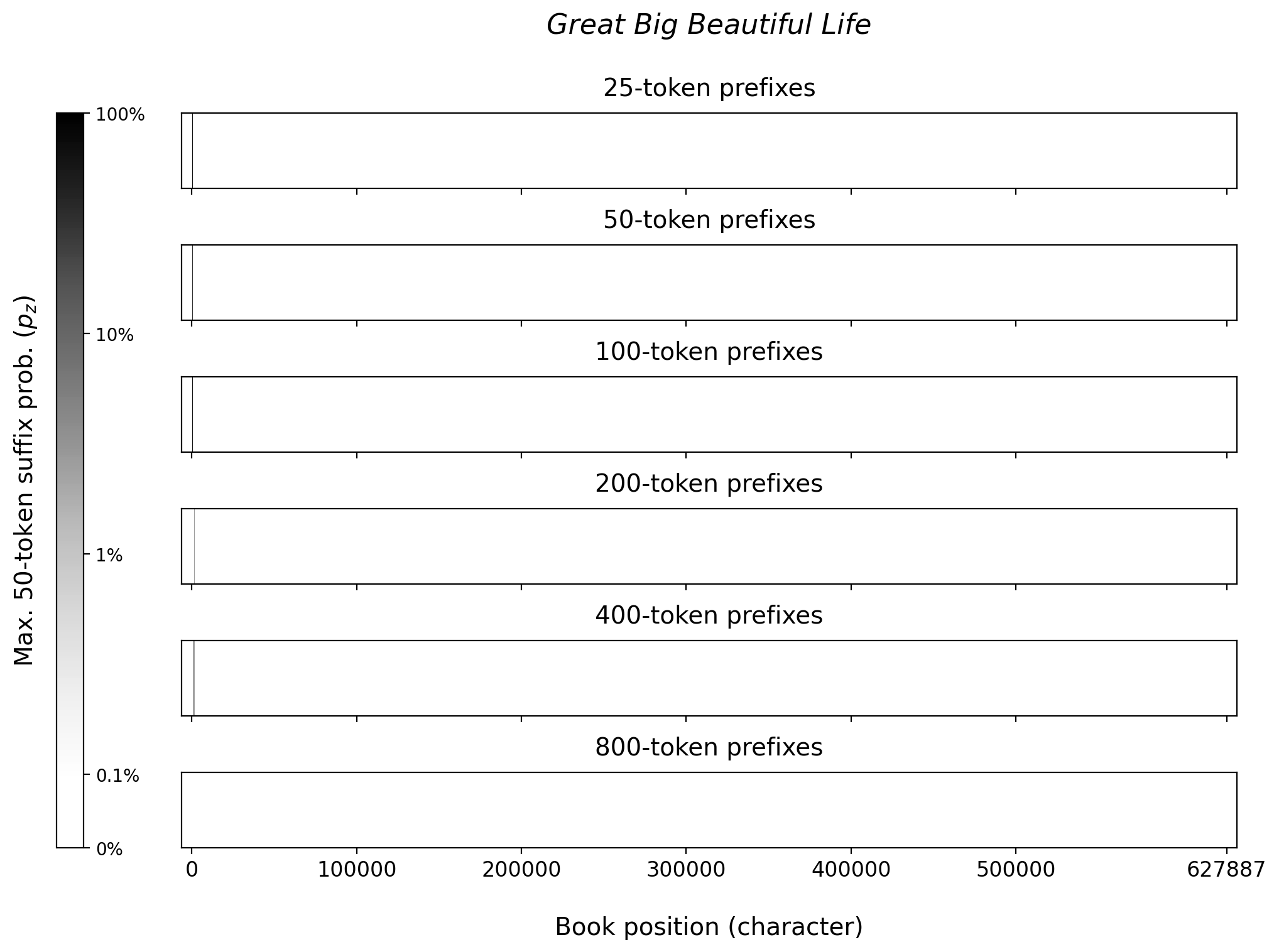}
\caption{\textsc{Qwen 2.5 72B}}
\end{subfigure}
\caption{\textbf{Varied-prefix, post-training-cutoff negative controls.} 
We show results for \emph{Great Big Beautiful Life}~\citep{greatbig}, which was published in spring 2025---after the release of all of the models we test. 
(The newest model is from the \textsc{Qwen 2.5} series, which was released in late September 2024.)
This book is not in any of the models' training data.
We run probabilistic extraction on (\textbf{a}) \textsc{Llama 3.1 70B}, (\textbf{b}) \textsc{Llama 3.1 8B}, (\textbf{c}) \textsc{Llama 2 13B}, (\textbf{d}) \textsc{Phi 4}, and  (\textbf{e}) \textsc{Qwen 2.5 72B} with top-$k$ ($T\!=\!1$, $k\!=\!40$) decoding for $50$-token suffixes and prefix lengths in $\{25, 50, 100, 200, 400, 800\}$. 
We are unable to extract any unique (e.g., non-copyright notice) sequences using any model or prefix length.\looseness=-1 
}
\label{fig:validity:control:greatbig}
\vspace{-.25cm}
\end{figure*}

We also include the last book, \emph{Great Big Beautiful Life}~\citep{greatbig}, in our varied prefix experiments as another point of comparison. 
We discuss these results for \textsc{Llama 3.1 70B} in Section~\ref{sec:validity:controls} and in Figure~\ref{app:fig:prefix-coverage}, and provide the full set of results for $5$ models in Figure~\ref{fig:validity:control:greatbig}. 
We extract no sequences (beyond boilerplate text) from any of the $5$ models we test.
In Figure~\ref{app:fig:greatbig:main}, we show an extended $p_\vz$ range to highlight how low we need to go beneath $\tau_\text{min}$ to register chance generation of non-training data. 
We extend to $10^{-14}\%$---$13$ orders of magnitude below $\tau_\text{min}{=}0.1\%$---and find exactly $1$ sequence that surpasses this threshold for $200$--$800$-token prefixes.\looseness=-1

For example, for $800$-token prefixes, we register $1$ sequence with $p_\vz{=}6.06 \times 10^{-14}\%$. 
We report this as on the order of $10^{-14}\%$ in the main text because that is all we can reliably do for a number this small.
The mantissa ($6.06$) is unreliable at this magnitude. 
This is because, even though we perform all $\log$ probability computations on CPU in full \texttt{float64} precision, producing the logits during inference on GPU (that we post-process on CPU) involves mixed-precision computations with the model weights/activations in \texttt{bfloat16}~\citep{gdm_bf16_range} and accumulation operations (matrix multiplications, normalization) in \texttt{float32}~\citep{hf_bf16_accum}.
Each computation introduces small rounding errors that accumulate and propagate to the final logits.
When we exponentiate to get probabilities, these errors result in only the exponent (not the mantissa) being reliable. 

\FloatBarrier

\subsection{Limits on ``the monkey at the typewriter'' metaphor}\label{app:sec:validity:monkey}

A common trope is to compare an LLM to a ``monkey at the typewriter''~\citep{borelmonkey, simpsons}.
The basic idea is that, given infinite attempts, an LLM could presumably output \emph{any} continuation of a piece of text.
Like the monkey, it could by happenstance output a pristine copy of a Shakespeare play. 
This metaphor seems founded in the idea that LLMs are probabilistic. 
The monkey is randomly typing letters, and so, too, is an LLM, since there is an element of randomness in the generation process. 
For a given piece of text (the context), the LLM samples the next token to output based on the next-token conditional probabilities.
That sampling is random, with respect to the conditional probability distribution of the possible next tokens.\looseness=-1 

Already here, this description of what an LLM is doing to generate text should start to sound very different from a ``monkey at the typewriter.'' 
In that thought experiment, what the monkey is doing is random in the sense that \emph{every} next character is equally likely to be produced at \emph{every} step.
But clearly, that is \emph{not} what is happening with an LLM. 
The LLM has learned probabilistic relationships from the natural-language and other text (e.g., code) it was trained on.
These relationships condition outputs given inputs.
So while there is sampling involved---and thus probabilistic outcomes---those probabilities are not ``random'' in the same sense as the monkey typing. 
If an LLM could be used to produce a copy of a Shakespeare play (or \emph{Harry Potter and the Sorcerer's Stone}, Section~\ref{sec:book:seed} \& Appendix~\ref{app:sec:reconstruct}), it would not be because it behaves randomly like the monkey. 
(One could set the temperature to an enormous number, effectively inducing ``monkey'' behavior via the decoding scheme, but this is not how LLMs are used in practice.) 

\paragraph{A ``monkey'' or a ``toothpick''?}
These probabilistic relationships can also be easily misunderstood in the opposite direction. 
One could imagine that a sufficient amount of context used as input would serve to bias the LLM's output to a particular, deterministic outcome:  
with a long enough prompt of the user's choosing, they could get an LLM to produce \emph{any} particular, desired target output.
This is the basic idea in Tyler Cowen's ``toothpick argument''~\citep{toothpick}: 

\begin{quote}
\small
If you stare at just the exact right part of the toothpick, and measure the
length from the tip, expressed in terms of the appropriate unit and
converted into binary, and then translated into English, you can find any
message you want. 
You just have to pinpoint your gaze very very exactly
(I call this ``a prompt'').

In fact, on your toothpick you can find the lead article from today's \emph{New
York Times}. 
With enough squinting, measuring, and translating.

By producing the toothpick, they put the message there and thus they gave
you NYT access, even though you are not a paid subscriber. 
You simply
need to know how to stare (and translate), or in other words how to prompt.

So let's sue the toothpick company!
\end{quote}

The ``toothpick'' misunderstanding of an LLM is mutually exclusive with the ``monkey'' misunderstanding. 
For the toothpick, prior context has a determinative impact on the output; for the monkey, there is no context long enough that could ever have a determinative impact on the output.  
Note that in the toothpick metaphor, it should also be possible to get an LLM to generate outputs that are exact copies of \emph{non}-training data---in this case, ``the lead article from \emph{today's} [paper]'' (emphasis added). 
(One could, for an open-weight model, manually bias logits.
This could force an LLM to produce certain outputs, effectively inducing ``toothpick'' behavior, but this is also not how LLMs are used in practice.)\looseness=-1 

\paragraph{Where our results come in.}
There is plenty to say about misplaced metaphors about generative AI.
We defer to others who have written extensively about this~\citep{lee2023talkin, cooper2023report, cooper2024files}.
Here, we highlight how our specific results make specific points clear about why these two particular metaphors are both misguided. 
The key point is that our experiments show that (a) there will be at least some target suffixes that could \emph{never} be generated by an LLM, even with infinite prompts to the model with a given prefix, and (b) a specific desired target suffix cannot \emph{always} be produced by an LLM, given a sufficiently long prompt.  
The evidence for both comes from the same results:
for a given LLM $\theta$ and decoding scheme $\phi$, some suffixes are \emph{impossible} to generate, regardless of (a) how many times one independently prompts the model with a given prefix or (b) how long of a prefix one uses to condition the output.\looseness=-1 

Our results make this very clear in practice, but we can also make this point conceptually.
Intuitively, for a sequence $\vz$, if the LLM $\theta$ with decoding scheme $\phi$ assigns 
the verbatim suffix (given the prefix) zero probability (i.e., $p_{\vz}{=}0$), then no amount of independent
sampling trials will ever produce that suffix. 
If instead $p_{\vz}>0$, then repeated independent 
sampling ensures that the probability of seeing $\vz$ at least once tends to $1$ as the number 
of trials grows. 
Therefore, for any finite set of sequences $\sZ$, the maximum number of 
verbatim suffixes that can ever be generated is exactly those with nonzero probability:\looseness=-1
\[
\lim_{n \to \infty} \mathbb{E}\!\left[\#\text{ unique verbatim suffixes produced}\right]
\;=\; \big|\{\vz \in \sZ : p_{\vz} > 0\}\big|.
\]

\begin{proof}
For a single sequence $\vz$ with per-trial, verbatim-suffix generation probability $p_{\vz}$, 
the probability of generating the suffix at least once after $n$ independent trials is
\[
\Pr(\text{verbatim suffix produced in }n \text{ independent trials}) \;=\; 1 - (1-p_{\vz})^n.
\]
If $p_{\vz}=0$, this expression is $0$ for all $n$. 
If $p_{\vz}>0$, then $(1-p_{\vz})^n \to 0$ as $n \to \infty$, so $\Pr(\text{verbatim suffix produced in }n \text{ independent trials})\to 1$.
Now let $\sZ$ be a finite set of sequences. 
By linearity of expectation,
\[
\mathbb{E}[\# \text{verbatim suffixes produced in }n \text{ independent trials}] 
\;=\; \sum_{\vz\in\sZ} \big[1 - (1-p_{\vz})^n\big].
\]
Taking $n \to \infty$,
\[
\lim_{n\to\infty} \mathbb{E}[\# \text{verbatim suffix produced in }n \text{ independent trials}] 
\;=\; \sum_{\vz\in\sZ} \mathbf{1}[p_{\vz}>0].
\]
\end{proof}

Therefore, for this fixed finite set \(\sZ\), there is a maximum attainable suffix (given prefix) generation rate: 
the fraction of \(\sZ\) whose suffixes can ever be generated.  
This limit is independent of the target success level \(p\) in \((n,p)\)-discoverable extraction (Definition~\ref{app:def:np_discoverable_extraction}).  
As \(n\) grows large, the rate reflects not only extractable (memorized) sequences---i.e., those with sufficiently high probability such that we consider them to be instances of extraction (in this work, $\tau_\text{min}\!\geq\!0.1\%$)---but also any $\vz$ with $0 < p_{\vz} < \tau_{\text{min}}$ that we would not count as extraction.\looseness=-1

\paragraph{Sequences that cannot be generated in practice.}
There are various ways that a given suffix could have $p_\vz\!=\!0$.
It might be the case that, for a particular decoding scheme, it is impossible to sample certain tokens, making some suffixes unreachable.
This is the case for top-$k$ decoding, which only considers the top-$k$-highest probability tokens at each sampling iteration.
Any token not included in the top-$k$ tokens forecloses possible output generation paths
(Appendix~\ref{app:sec:background:compute}). 
For $k\!=\!40$ and for \textsc{Llama 1} models, only $40$ out of $32{,}000$ tokens could be generated in a given sampling iteration. 

For another, numbers are represented in computers with a finite number of bits. 
This means that there is a minimum number that can be represented ($>0$) before it gets rounded down to $0$. 
For instance, imagine a token vocabulary of $32{,}000$ tokens (as with \textsc{Llama 1} models).
If the model just generated tokens completing randomly (by setting temperature very high, like a ``monkey at the typewriter''), then at each iteration every token would have probability $1/32{,}000$.
When generating $71$ such tokens in a row (i.e., a $71$-token suffix), the probability of any such suffix is ${(1/32{,}000)}^{71}\approx1.3626\times 10^{-320}$. 
Once we try to generate a $72$nd token, the probability for the whole sequence becomes ${(1/32{,}000)}^{72}=0$:
in terms of how the probability gets represented in the computer, there is underflow (i.e., rounding down to $0$). 
As a result, even with completely-random temperature sampling, not every sequence could possibly be generated.\looseness=-1

We often observe $0$-probability suffixes regardless of prefix length.
This is true for non-training data (Section~\ref{sec:validity:controls}, Appendix~\ref{app:sec:validity:cutoff}) and for training data that are not memorized (Section~\ref{sec:validity:baseline}, Appendix~\ref{app:sec:validity:baseline}).
Even if we add more context to try to condition it in a particular direction---there are many cases where we simply are unable to raise $p_\vz$ above $0$ (let alone above $\tau_\text{min}$, the minimum probability we consider evidence for extraction).\looseness=-1 
\subsection{Additional notes on extraction and membership inference}\label{app:sec:validity:membership}

There are several points that are worth addressing with respect to the broader literature on \newterm{membership inference} and \newterm{extraction}.
Here, we briefly situate our work---particularly our work with respect to validity---in relation to prior work. 
This is important to do for clarity reasons, because our work and the types of claims we make are very different from prior work.
Prior work tends to focus on \emph{average} extraction rates over \emph{random} samples of known or proxy training data.
Since we are making claims about \emph{specific} works from a \emph{specific} population of text, the ways that we go about evaluating and validating our work are necessarily different.
This is not just a detail or aside;
it has to do with the entire purpose of this project, and significantly influences our measurement approach in ways that contrast with other work on membership inference.

\subsubsection{The relationship between membership inference and extraction}\label{app:sec:validity:membership:inference}

As noted in \citet{carlini2021extracting}, ``Given a set of samples from the model, the problem of training data extraction reduces to one of membership inference: predict whether each sample was present in the training data.''
This makes sense intuitively. 
Extraction is defined with respect to memorization: 
it is only possible to extract memorized training data, and (tautologically) training data can only be memorized if they are included---i.e., are \newterm{members}---of the training dataset. 
To demonstrate extraction is therefore to demonstrate memorization, and memorization implies membership. 

Not only does extraction imply membership of a piece of data, it actually reproduces that data in model outputs.
In this respect, it is a stronger evidentiary claim (and, at least in principle, a harder one to make): 
membership inference asks a one‐bit question---was this sequence a member of the training data?---while extraction produces the sequence itself. 
(At the same time, reliable membership inference with even the strongest \newterm{membership inference attacks} is extraordinarily challenging for LLMs~\citep{hayes2025strongmia}, while extraction is relatively straightforward.) 
Put differently, one can make membership inference claims \emph{without} extraction (e.g., with classifiers, see below); 
these claims can involve evidence where extraction might fail.
So, even though successful extraction always implies membership, this does not go in both directions:
membership does not imply successful extraction (Appendix~\ref{app:sec:validity:membership:nonmember}).\looseness=-1 

\citet{carlini2021extracting} goes on to note that ``In their most basic form, past membership inference attacks rely on the observation that models tend to assign higher confidence to examples that are present in the training data.
Therefore, a potentially high-precision membership inference classifier is to simply choose examples that are assigned the
highest likelihood by the model.'' 
At a high level, this is clearly directly related to extraction. 
Greedy‐decoded discoverable extraction approximates the highest‐likelihood sequence (albeit pretty poorly) with greedy decoding~\citep{carlini2023quantifying}. 
Probabilistic extraction does the same but allows more flexibility than greedy decoding, and can identify sequences with higher likelihood than greedy‐decoded ones~\citep{hayes2025measuringmemorizationlanguagemodels}.

\subsubsection{We infer  membership for sequences, not whole books}\label{app:sec:validity:membership:known}

In much work on extraction, researchers do not demonstrate extraction to \emph{infer} membership. 
Instead, they run extraction experiments on \emph{known} training data~\citep{hayes2025measuringmemorizationlanguagemodels, carlini2023quantifying, lee2022dedup, hayes2025strongmia}, i.e., where ground-truth membership is not in doubt. 
This makes sense as a setup for drawing valid conclusions about extraction and memorization: 
membership is a prerequisite for memorization, and memorization is a prerequisite for extraction. 
Testing extraction on known member sequences therefore enables straightforward conclusions about memorization.
Here, memorization is the most plausible explanation for extraction. 
As Carlini explains, when a sufficiently large and unique piece of training data is extracted, ``the only possible explanation is that the model has somewhere internally stored [that piece of training data]''~\citep{carlini2025blog}. 
As Cooper and Grimmelmann similarly note, ``in order to be able to extract memorized content from a model at generation time, that memorized content must be encoded in the model's parameters. There is nowhere else it could be. A model is not a magical portal that pulls fresh information from some parallel universe into our own''~\citep[p. 25]{cooper2024files}. 
(See Section~\ref{sec:background}.)\looseness=-1

Most of our experiments fall into this category, as the large majority of our results are for \textsc{Llama} and \textsc{Pythia} models, where it is known that \texttt{Books3} was included in the training data (Sections~\ref{sec:book-procedure}--\ref{sec:book:compare}; Appendices~\ref{app:sec:sliding-window:setup} \&~\ref{app:sec:validity:setup}). 
This is precisely the motivation behind our baseline experiments (Section~\ref{sec:validity}, Appendix~\ref{app:sec:validity:baseline}), which we use as a basis of comparison for our negative controls (Section~\ref{sec:validity}, Appendix~\ref{app:sec:validity:controls}). 
Not only do we know that these models were trained on the books in \texttt{Books3}, we know that they were trained precisely on the \texttt{Books3} corpus.\looseness=-1 

Nevertheless, while most work on greedy-decoded discoverable extraction does not do this, we perform validity experiments anyway.
We need to set a minimum valid extraction probability $\tau_\text{min}$ (Appendix~\ref{app:sec:validity:motivations}). 
In this paper, we use a very high global $\tau_\text{min}$ to avoid false positives, and we validate this choice with our negative controls. 
\citet{hayes2025measuringmemorizationlanguagemodels} do not set an explicit $\tau_\text{min}$, and do not make strong claims about when probabilistic extraction is valid;
however, they do run a negative control experiment to show more generally that probabilistic extraction is a reasonable metric.\looseness=-1

In the original version of this paper, we included the negative controls on \textsc{Phi 4} (Appendix~\ref{app:sec:validity:phi4}) to support our claims. 
Combined with the extensive analysis we perform on known training data for \textsc{Llama}, this is how we chose a global $\tau_\text{min}$: 
a conservative threshold that would not over-count extraction (and thus memorization). 
Even before including our experiments on post-cutoff books, our results provided ample evidence that non-training data exhibit $0$ or orders of magnitude smaller probabilities $p_\vz$ than training data. 
The post-cutoff negative controls provide further confirmation.

This was true both for smaller, lower quality models, and higher quality models that exhibit lower loss (on average). 
While larger models like \textsc{Llama 3.1 70B} fit their training data better than smaller ones, and exhibit lower loss on average,  they do not (and cannot) exhibit low loss everywhere, such that our measurements of $\log$ probabilities would be confounded. 
This is also clear from our experiments on known training data---\texttt{Books3} books.
For many books and large models, we are unable to extract any unique sequences;
there are effectively no nonzero sequence probabilities.  
It is plausible that future work will find lower, model-specific $\tau_\text{min}$ thresholds, which may perhaps need to be set higher for larger models than for smaller ones. 
For now, however, the threshold we set is so conservative that these nuances do not affect our conclusions.\looseness=-1

\subsubsection{Inferring membership for LLMs with undisclosed training data}\label{app:sec:validity:membership:unknown}

Given our extensive results on known training data---including experiments to vet their validity---we can apply our methodology with confidence to models whose training data are not known with certainty. 
We do this for three model families: \textsc{DeepSeek v1}, \textsc{Qwen 2.5}, and \textsc{Gemma 2}. 
For these models, validated extraction serves as evidence of membership. 
These experiments differ from those on models known to be trained on \texttt{Books3}, but we ensure their validity through extensive negative controls and baselines. 

This approach is consistent with prior work. 
\citet{carlini2021extracting} performed experiments on \textsc{GPT-2}. 
\citet{nasr2023scalable, nasr2025scalable} extracted training data from open-weight models with undisclosed training data and from ChatGPT, a closed system. 
In both cases, the authors relied on proxy datasets and strategies to reduce bias in their results, such as shorter prompts. 
More generally, this is an important and reasonable thing to try to achieve.
As researchers, we should be developing valid techniques for extraction that are useful in cases for which we do not know ground-truth membership.
If membership inference required knowing membership in advance, there would be nothing to infer; 
at best---when it works---it would confirm what we already know. 

We do not need to use shorter prompts in our work because we validate our work with extensive negative controls. 
Further, we manually verified that every sequence extracted from \textsc{Phi 4} and from post-cutoff books was widely duplicated online. 
Our validity experiments go beyond most prior work, which often does not include negative controls. 
This is intentional; 
our work has the potential to impact copyright litigation, not just the scientific literature.
We therefore believe it is important to hold our work to a higher bar. 
Importantly, since training data for these models are undisclosed, we make claims only at the sequence level, not at the level of whole books (Appendix~\ref{app:sec:validity:membership:sequence}). 

Finally, the population of data we study differs from most prior work, which evaluates memorization broadly across diverse training data (websites, code, spam, etc.)~\citep{carlini2021extracting, carlini2023quantifying, nasr2023scalable, nasr2025scalable, hayes2025measuringmemorizationlanguagemodels}. 
Such work must account for false positives like ``trivial memorization'' (e.g., number lists) or ``repeated substrings'' (e.g., ``I love you'' loops). 
LLMs assign high probability to both of these types of sequences, even if they are not memorized.
We note these phenomena explicitly where we identify them, but they are generally not relevant our domain of long-form narrative text. 

The rare cases we identify of trivial memorization---lists of page numbers (an artifact of converting \textsc{epub} to \textsc{txt}), chapter indices---are excluded from our claims about extraction of unique creative expression. 
We identify one case of repeated substrings.
\citet{The_Bedwetter} exhibits elevated probability in a part of the book that repeats ``I will not wet the bed.'' four times in a row. 
But this sequence does not surpass $\tau_\text{min}$, anyway, so we do not count it as extracted.
(We also exclude boilerplate instances of true memorization from our claims, like copyright notices and author biographies.) 
Across millions of sequences, these account for only a few hundred cases. 
This highlights why the population under study matters for validity: 
our focus on books yields different considerations than prior work, and we address these systematically. 
Unlike the work done in \citet{carlini2021extracting}, these types of edge cases are extremely rare in our population of interest. 

\paragraph{Prior work on copyrighted text and LLMs.}
In the technical literature, there are a few earlier papers that address the generation of copyrighted text by LLMs.
While this work (e.g., \citet{karamolegkou2023copyrightviolationslargelanguage}) is important initial work, for the reasons discussed in this appendix, much of it arguably does not directly study memorization, even if it is framed in those terms.
In particular, without ground-truth information about training data membership and without negative controls to calibrate false positives, it is difficult to interpret reported overlap-based results as valid evidence of extraction (and thus memorization).
(Further, in the case of \citet{karamolegkou2023copyrightviolationslargelanguage}, results for very short-form generations---which are insufficient on their own to support extraction claims---are presented alongside longer recovered text within the same plots, making it difficult to distinguish stronger from weaker forms of evidence.)

\subsubsection{Our inferences about membership are about sequences, not whole books}\label{app:sec:validity:membership:sequence}

We emphasize that our claims about membership are made at the level of suffixes, not entire books. 
For models such as \textsc{DeepSeek v1}, \textsc{Qwen 2.5}, and \textsc{Gemma 2}, where training data are undisclosed, we cannot establish whole-book membership. 
Instead, our extraction results are about specific sequences drawn from \texttt{Books3}, a known pretraining dataset, and it is at this level that we make membership claims. 
In some cases, our results are suggestive that entire books were included in training, given that we can extract high-probability suffixes spanning a full work (e.g., Appendix~\ref{app:sec:sliding:Harry_Potter_and_the_Sorcerer_s_Stone}), but we do not generalize from suffixes to whole-book membership. 

We also do not claim that we know the provenance of those extracted sequences.
For \textsc{Llama} models we know that the training data included \texttt{Books3}, and thus at least one copy of each sequence from the books we test was included in the training data via \texttt{Books3}.
But this is about all we do know with certainty.
We do not know very much about these models' training data details more generally; 
knowledge of the inclusion of \texttt{Books3} is a rare exception, not the norm. 
Books sequences may have been included in the training data via other unknown sources: 
other book corpora, widely quoted excerpts in web scrapes, etc. 
The same could be said for these other models, where we do not know if \texttt{Books3} was included in the training data. 

Altogether, these observations highlight that membership questions for LLMs are rarely straightforward. 
This is already widely known in the literature, with respect to the presence of duplicates in training datasets~\citep{lee2022dedup, carlini2023quantifying, hayes2025measuringmemorizationlanguagemodels}. 
This discussion illustrates how these challenges manifest specifically for books, and why membership claims should be scoped carefully. 
Our measurements are at the level of individual suffixes, but that is not necessarily the only level at which membership is of interest. 
For copyright, membership may be meaningful in different ways at different scopes---an entire book, a chapter, a single sentence, etc.~\citep{lee2023talkin}. 

\subsubsection{Absence of extraction does not imply anything about membership}\label{app:sec:validity:membership:nonmember}

We state this with respect to describing the problem of membership inference in Appendix~\ref{app:sec:validity:membership:inference}, but it bears repeating: 
extraction implies membership, but membership does not imply extraction.
This is clear in our results; 
many books in \texttt{Books3} have no extractable unique text for \textsc{Llama} models, even though those models were trained on \texttt{Books3}.
So when we see $p_\vz\!=\!0\%$, we should not conclude that a given sequence (or the source that contains it) was not included in the training data.
This is trivially true for \textsc{Llama} models, but more generally true for models with undisclosed training data.
For example, \textsc{Qwen 2.5} may have been trained on \texttt{Books3};
just because it exhibits low extraction probabilities on \texttt{Books3} books is not evidence that it was not trained on this corpus.
Further, other extraction techniques than the ones we use here could possibly surface additional memorization;
so absence of extraction \emph{with our measurement procedure} is not definitive that a particular sequence was not memorized.\looseness=-1 

Further still, $p_\vz\!=\!0\%$ does not necessarily imply anything about how much a particular sequence may have influenced the model or its outputs.
Attribution of generations to training data is a different problem, both with respect to technical work in ML and implications for copyright.
We do not study these topics in this work.
At a very high level, one can see this distinction through our brief results in Appendix~\ref{app:sec:reconstruct:fail}. 
We run our reconstruction procedure on \emph{Sandman Slim}~\citep{Sandman_Slim} and \textsc{Llama 3.1 70B}.
This model does not memorize this book at all, with respect to our extraction procedure. 
We run this experiment to show that it fails---that reconstruction is not something that trivially works for any book. 
Nevertheless, even though we fail to reconstruct any text from the book, it is clear that the generation we produce was influenced by the text (in an informal sense). 
This type of influence is not something that we capture or attempt to capture---either formally or informally---in our measurements in this paper.\looseness=-1

\subsubsection{Extractability and copyright}\label{app:sec:validity:membership:copyright}

Separate from the above technical considerations about membership and extraction, there are other considerations for copyright.
A sequence can count as memorized in a technical sense---in this paper, when $p_\vz\!\geq\!\tau_\text{min}\!=\!0.1\%$---but this does not necessarily mean that all types of memorization are relevant to copyright. 
As we discuss in Sections~\ref{sec:copyright} and~\ref{sec:takeaways}, sequences that have very low extraction probabilities may matter less for copyright than those with high ones, since the latter are far easier to extract reliably in model outputs.

\section{Recovering books with one seed prompt}\label{app:sec:reconstruct}

Given the extent of memorization that we observe for \emph{Harry Potter and the Sorcerer's Stone}~\citep{Harry_Potter_and_the_Sorcerer_s_Stone} (as well as other books) and \textsc{Llama 3.1 70B}, it seemed possible that we could recover the entire book near-verbatim using only a single seed prompt of ground-truth text drawn from the book.

This result was a natural follow-on to our memorization experiments, but it is not the main focus of our paper.
Therefore, this part of our paper reflects only a preliminary investigation into recovering large book segments from an LLM.
It is an existence proof that this type of result is possible, not a comprehensive investigation into how to make this type of method work for different texts. 

\subsection{Recovery method}\label{app:sec:reconstruct:method}

We successfully recover the book near-verbatim with a variety of seed prompts and variations on the same strategy.
Overall, we observe that this result is stable---across different hardware, and for different settings.
However, the procedure is relatively sensitive to how it is configured;
for example, if too long of a stride is used, then it we observe divergence from the true text.
We have preliminary results that significantly improve the stability of this approach.
We defer investigating this to future work.\looseness=-1

\paragraph{Seed prompt.} 
We obtained high-fidelity recoveries of the book with several seed prompts, including the first $390$ tokens of the the first chapter (the first few paragraphs) and the first line of the first chapter ($60$ tokens): 

\begin{quote}
\small
Mr. and Mrs. Dursley, of number four, Privet Drive, were proud to say that they were perfectly normal, thank you very much. They were the last people you'd expect to be involved in anything strange or mysterious, because they just didn't hold with such nonsense.
\end{quote}

With some additional exploration of how to configure generation, we also were successful with prompting with just ``Mr. and Mrs. D'' and ``Chapter ONE:\verb|\n|THE BOY'' (both $7$ tokens).
The former is sufficient for the model to output ``ursley'' and then the rest of the first line;
the same is true for the latter, which outputs the rest of the first chapter's title (``THE BOY WHO LIVED''), and then proceeds to output the first substantive line of the book.
(These token counts include the first token, which is the beginning-of-sequence---\texttt{BOS}---token.)

\paragraph{Overall approach.}
We take a sliding-context approach that we discuss in Section~\ref{sec:book:seed}.
We take an initial seed $s$ and top it up to a minimum of $50$ tokens (if it is less than that). 
On some hardware, for stability, we normalize the format of that longer seed (e.g., making sure that em dashes are consistently represented) before proceeding to long-form generation.
In all cases, we do not guide the procedure beyond providing the initial context. 
We use beam search to generate $50$ tokens at a time, and use a maximum context window of $3000$ tokens for each generation.
This means that we append the $50$ tokens we generate to the existing context, and clip the first $50$ tokens for iterations that exceed the maximum context. 
(This is most of the iterations; only at the start do we have a context that can be less than $3000$ tokens).\looseness=-1

We succeeded with  $9$ tries in total.
(Our first major success occurred in the third run, when we recovered ten chapters---chapters 1-9, and 12---near-verbatim.) 
The first $3$ runs were debugging and setup (e.g., testing the key idea with greedy decoding and getting a sense of how short a reasonable seed prompt could be).
The last $6$ involved changing the size of the number of maximum context tokens (we settled on $3000$) and the number of beam candidates ($8$), as well as minor tweaks to handling \texttt{EOS} tokens.\looseness=-1

Handling \texttt{EOS} tokens was the only (minor) complication, and it was  simple to resolve. 
\texttt{EOS} tokens tend to be predicted at the ends of chapters;
we therefore remove them and manually replace them with \texttt{"CHAPTER \{n+1\}"}, with \texttt{\{n+1\}} spelled out (e.g., \texttt{"TWO"}, \texttt{"THREE"}). 
Other attempts involved \texttt{"Chapter"} (instead of all caps) and using the number \texttt{\{n+1\}}  (e.g., \texttt{"2"}) instead of spelling out the chapter number (e.g., \texttt{"TWO"}).
Other solutions could just down-weight the logits for \texttt{EOS}.

Occasionally, the model does not predict an \texttt{EOS} at the end of a chapter, which leads to some misalignment later on (e.g., inserting \texttt{"CHAPTER FOUR"} for the fifth chapter).
This can cause the model to repeat segments of chapters that have already been produced.
The fix here was also simple. 
We keep track of how many tokens have been generated since the last \texttt{EOS} token, and if that number surpasses $10{,}000$, we assume that we have missed the end of a chapter and account for this (i.e., move the chapter counter ahead even though we did not see an \texttt{EOS}).
In future runs, we were able to remove this logic entirely. 
We defer discussion to future work. 

The exact code that we ran using the first $390$ tokens as the seed prompt is in Listing~\ref{app:code:hp}.
Since this code involves beam search, our results should be (in theory) deterministic.
Because in practice hardware non-determinism can complicate this, we ran a (non-exhaustive) test by executing the code on two different sets of $4$ GPUs.
Both yielded the same exact output.
(Though we also spot-checked some of the underlying logits, and there were some slight differences.)\looseness=-1

\paragraph{Comparing to the ground-truth text.}
The substantive diff between the version of \emph{Harry Potter and the Sorcerer's Stone}~\citep{Harry_Potter_and_the_Sorcerer_s_Stone} that we have from \texttt{Books3} and our generated output is minimal.
We do some very minor text normalization of the ground-truth book.
That is, we trim the beginning and end of the ground-truth book to not contain front matter or anything after the end of the last chapter. 
We remove \texttt{\_} from both documents (used for italics in the \texttt{Books3} version) and align ellipses to \texttt{"..."} (which appear as \texttt{" . . ."} in the \texttt{Books3} version.) 
We then compute various similarity metrics betwen the two documents. 
The cosine similarity of \textsf{TF-IDF} vectors of each document is an astounding $0.9999$.
\textsf{TF-IDF} is a limited metric, as it treats documents as bags of words and thus fails to capture word order.
So, as two additional points of comparison, we compute similarity using greedy longest common substring (LCS) matching at both the word and sentence levels (via \texttt{difflib}'s \textsf{SequenceMatcher}). 
A score of $1$ indicates a perfect match and $0$ indicates no match. 
We obtain a word-level similarity of $0.992$ and a sentence-level similarity (which is more sensitive to formatting differences) of $0.934$.

The qualitative differences involve minor formatting changes (that we do not normalize) and localization differences. 
For example, the \texttt{Books3} version has British spelling (e.g., \texttt{"Mummy"} instead of \texttt{"Mommy"}). 
These formatting and localization differences are the large majority of the diff between the two texts. 
Occasionally, the model also misses single lines, e.g., ``There was a loud 'Oooooh!''' (Figure~\ref{fig:diff}).\looseness=-1

One of the most interesting outcomes (that we have not investigated in detail) is very occasional apparent hallucinations.
For example, in the first chapter, we observe the following diff:

\begin{figure}[h]
    \centering
    \includegraphics[width=0.9\linewidth]{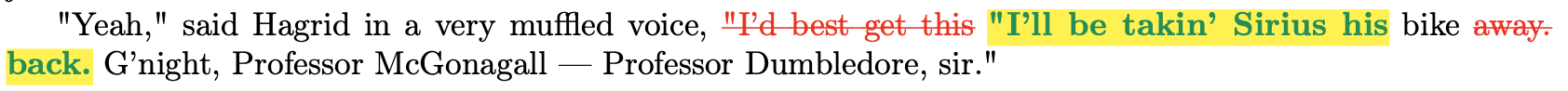}
    \caption{\textbf{Portion of the diff from the first chapter of \emph{Harry Potter and the Sorcerer's Stone}.} 
    Crossed-out red text is in the ground-truth book in \texttt{Books3}, but was not generated by our recovery procedure; 
    text highlighted in yellow/green is in the recovered text but not the ground-truth text.\looseness=-1}
    \label{app:fig:sirius}
\end{figure}

Selecting the top sequence candidate with beam search always led to this particular outcome in Figure~\ref{app:fig:sirius}, regardless of how we configured our method.
Given how specific this particular change was, we investigated it further.
It turns out that the generated text reflect the ground-truth from editions of the book printed before 2004.
In later editions, this line was changed to the one that we do not generate---the one that is in the ground-truth version we have access to in \texttt{Books3}.
This change was done for plot continuity reasons, based on an inconsistency that emerged due to plot development concerning Sirius Black in the later books~\citep{sirius}.
This line is not actually hallucinated; 
it is just from an  earlier edition of the book than the one we diff with.
Of course, this difference still contributes to our quantitative measurements of similarity, even though it is not a meaningful substantive difference.

\lstset{style=mystyle}
\begin{lstlisting}[language=Python, caption=Recovery with a single seed prompt., label={app:code:hp}]
from transformers import AutoTokenizer, AutoModelForCausalLM
import torch, json
from tqdm import tqdm

# load the first line of HP 1
with open("seed.txt", "r", encoding="utf-8") as f:
    seed_text = f.read().strip()

# Load the model and tokenizer
model_name = "meta-llama/Llama-3.1-70B"
tokenizer = AutoTokenizer.from_pretrained(model_name)
model = AutoModelForCausalLM.from_pretrained(
    model_name,
    device_map="auto",
    torch_dtype=torch.bfloat16,
)
model.eval()

max_context_tokens = 3000
max_new_tokens = 50
# the book is more like 106k; 
# adding buffer for potential formatting differences
max_story_tokens = 113000

eos_token_id = tokenizer.eos_token_id

input_ids = tokenizer(seed_text, return_tensors="pt").input_ids.to(model.device)
generated_ids = input_ids

# Set up progress bar
pbar = tqdm(total=max_story_tokens, desc="Generating story tokens")
pbar.update(generated_ids.shape[1])

generation_steps = []
generation_num = 1
chapter_count = 1

chapter_nums = {1 : "One", 2 : "Two", 3 : "Three", 4 : "Four", 5 : "Five", 6 : "Six", 7 : "Seven", 8 : "Eight", 9 : "Nine", 10 : "Ten", 11 : "Eleven", 12 : "Twelve", 13 : "Thirteen", 14 : "Fourteen", 15 : "Fifteen", 16 : "Sixteen", 17 : "Seventeen"}

tokens_since_last_eos = 0

while True:
    # Get the tokens for the (up to) 3000-token prompt
    slice_start = max(0, generated_ids.shape[1] - (max_context_tokens - max_new_tokens))
    input_ids_window = generated_ids[:, slice_start:]
    attention_mask = torch.ones_like(input_ids_window)

    # call generation code
    outputs = model.generate(
        input_ids_window,
        attention_mask=attention_mask,
        max_new_tokens=max_new_tokens,
        do_sample=False,
        num_beams=8,
        early_stopping=False,
        length_penalty=1.2,
        temperature=1.0,
        top_p=1.0,
        pad_token_id=eos_token_id,
    )

    # remove the prompt from the output
    new_tokens = outputs[:, input_ids_window.shape[1]:]
    new_tokens_list = new_tokens[0].tolist()

    tokens_since_last_eos += len(new_tokens_list)

    # Some fun tricks to deal with EOS tokens
    if eos_token_id in new_tokens_list:
        if tokens_since_last_eos >= 10000:
            print("More than 10000 tokens since last EOS; probably missed a chapter break")
            print("Incrementing chapter count")
            chapter_count += 1

        print(f"EOS: replacing with chapter break")
        new_tokens_list = [t for t in new_tokens_list if t != eos_token_id]
        chapter_count += 1
        chapter_text = chapter_nums.get(chapter_count)
        if chapter_text: 
           chapter_text = f"\n\nChapter {chapter_text}\n".upper()
        else:
            chapter_text="\n"
        chapter_tokens = tokenizer(chapter_text, add_special_tokens=False, return_tensors="pt").input_ids[0].tolist()

        new_tokens_list.extend(chapter_tokens)
        new_tokens = torch.tensor([new_tokens_list], device=model.device)

        tokens_since_last_eos = 0

    # Everything below is just saving and printing progress
    prompt_tokens = input_ids_window[0].tolist()
    prompt_text = tokenizer.decode(prompt_tokens, skip_special_tokens=True)

    generated_ids = torch.cat([generated_ids, new_tokens], dim=-1)
    pbar.update(new_tokens.shape[1])

    chunk_text = tokenizer.decode(new_tokens[0], skip_special_tokens=True)

    print(f"\n=== Generated chunk ({new_tokens.shape[1]} tokens) ===\n{chunk_text}")

    generation_steps.append({
        "generation": generation_num,
        "prompt_text": prompt_text,
        "generated_text": chunk_text,
        "total_generated_tokens": generated_ids.shape[1]
    })
    generation_num += 1

    with open("generation_log.json", "w", encoding="utf-8") as f:
        json.dump(generation_steps, f, indent=2)

    with open("generated_ids.json", "w", encoding="utf-8") as f:
        json.dump(generated_ids[0].tolist(), f)

    if generated_ids.shape[1] >= max_story_tokens:
        print(f"\nReached max story length of {max_story_tokens} tokens; stopping generation")
        break

pbar.close()

# save the full story text
full_text = seed_text + "".join(step["generated_text"] for step in generation_steps)

with open("generated_story.txt", "w", encoding="utf-8") as f:
    f.write(full_text.strip())
\end{lstlisting}

\subsection{Partial recovery}\label{app:sec:reconstruct:partial}

We ran the exact same script on the essay ``The Case for Reparations,'' which is included in Ta Nehisi-Coates' book, \emph{When We Were Eight Years in Power}~\citep{We_Were_Eight_Years_in_Power}.
We observed enormous amounts of memorization for this essay in our sliding-window experiments.
(See Section~\ref{sec:validity}, highly memorized middle section; Appendix~\ref{app:sec:validity:baseline}).
This was one of the books in the \citet{kadreyamendedconsolidated} lawsuit, for which the court's judgment noted that neither party produced evidence that demonstrated sizable amounts of extraction of training data of the books in scope for the suit~\citep{kadreyjudgment}.

We summarize the first run of this experiment, where we ran the exact same code that recovered the \emph{Harry Potter} on the first line of the essay ($31$ tokens, including $\texttt{BOS}$):

\begin{quote}
\small
Clyde Ross was born in 1923, the seventh of thirteen children, near Clarksdale, Mississippi, the home of the blues.
\end{quote}

The entire essay is about $19{,}{000}$ tokens long, and running this script generated an essentially verbatim copy of the first ${\approx}68\%$ of the text (through the end of section VII).
The diff includes small punctuation/formatting differences (e.g., two consecutive \verb|\n| between paragraphs in \texttt{Books3}, with only one in the generation), generating numbers in numeric format rather than spelled out as a word (e.g., generating ``21'' instead of ``twenty-one''), and not generating the essay's section headers.
With minor text normalization (e.g., removing \_, which indicates italicized book titles in the \texttt{Books3} version), we obtain word-level greedy approximation of LCS (via \texttt{difflib}'s \textsf{SequenceMatcher}) of $0.992$ and a document-level (of the first ${\approx}68\%$) greedy approximation of LCS of $0.996$.
In word-space (not token-space), the seed prompt is $19$ words, and the generated near-exact (punctuation-normalized) text is $10{,}039$ words.

\subsection{Running recovery for books with low memorization signal}\label{app:sec:reconstruct:fail}

As a sanity check, we run the same method on \textsc{Llama 3.1 70B} for a book for which we observe effectively no extracted sequences.

For \emph{Sandman Slim}~\cite{Sandman_Slim} (Appendix~\ref{app:sec:sliding:Sandman_Slim}), we use the first paragraph as the prompt and (including this prompt, shown in \textcolor{blue}{blue}) produce a $2000$-token document.

Note that the generation gets caught in a loop (we mark where the prompt gets generated in \textcolor{blue}{blue} later on in the generation.) 
Interestingly, even though we do not  generate anything that we can identify as copied from the book, the generation does contain references to ``Sandman Slim,'' the main character ``James Stark'' (who is never referred to directly that way in the book, but as ``James Butler Hickok Stark'') as well as another character called ``Mason Faim.''

\begin{quote}
\scriptsize
\textcolor{blue}{I WAKE UP on a pile of smoldering garbage and leaves in the old Hollywood Forever cemetery behind the Paramount Studio lot on Melrose, though these last details don't come to me until later. Right now all I know is that I'm back in the world and I'm on fire. My mind hasn't quite kicked in yet, but my body knows enough to roll off the burning trash and to keep rolling until I can't feel the heat anymore.} I come to a stop against a gravestone and lie there for a moment, trying to remember how I got here and who I am. I don't always know right away. It's the amnesia. It's a hell of a way to travel, but it's better than being dead.
Slowly, like dirty water dropping out of a rusty pipe, it comes back to me. My name is James Stark. Friends call me Sandman Slim. Call me what you like. I've been called worse. I got this name because all I do is bring nightmares. True, I spend a lot of time in the City of Angels, but it's not like they're putting up a statue to me. Before this, I lived in Hell for eleven years. Even when I tried to stay out of trouble, trouble would come right to my door and make friends. That kind of thing has a way of making a man a little jaded when it comes to trusting his fellow man. Or demons. Even angels. I'm like that. I've got a bad attitude. I drink. I've been kicked out of every twelve-step program I've ever joined. I don't have many friends, so the ones I have mean a lot to me. They mean enough that if someone hurts them, I'll tear the universe down to save them. I'm telling you this so you'll know who I am and what I'm talking about when I say there are things in the dark. Things we don't always see that move among us. Some of these things are good and some aren't and sometimes they do things that make you want to rip open your chest and pull out your heart because you've seen too much and there's no going back. I know about these things because I'm one of them. I'm a monster and I'm a killer. I'm good at it and I never let emotion get in the way. When you live where I live, love can get you killed. And when it doesn't, it's worse because if you don't die, you just wish you would and every night you pray for something you lost in Hell. Pray for sweet, sweet death to come make a man out of you. But death doesn't come because if there's one thing I learned in Hell, it's that it will never, ever come for you. No matter how hard you pray. No matter what terrible things you do. But down deep in my heart, I know there's another truth, one that even Hell can't touch. And that's why, no matter what, I can never do the one simple thing that would set me free. The one thing I want more than anything else. I can never, ever kill the man who murdered my friends and started me on this journey to Hell and back. No matter how many times I try, I can't kill Mason Faim, the man I once called best friend. I wish I wasn't me sometimes, but to tell you the truth, even when this is all over and they finally let me die, I wouldn't trade this life for any other. It's been one hell of a ride and I'd do it all over again if I had to. And I probably will. Over and over and over again. Because that's all life is. One big ride that you can never get off of. You can just tighten or loosen your grip on the reins and see which way the horses take you.
\textcolor{blue}{I WAKE UP on a pile of smoldering garbage and leaves in the old Hollywood Forever cemetery behind the Paramount Studio lot on Melrose, though these last details don't come to me until later. Right now all I know is that I'm back in the world and I'm on fire. My mind hasn't quite kicked in yet, but my body knows enough to roll off the burning trash and to keep rolling until I can't feel the heat anymore.} I come to a stop against a gravestone and lie there for a moment, trying to remember how I got here and who I am. I don't always know right away. It's the amnesia. It's a hell of a way to travel, but it's better than being dead.
...
\end{quote}

We also tried much longer prompts (e.g., the first two pages of the book), and the model does not continue with text from the book. 
While these tests are not exhaustive, they are meant to illustrate that this is not behavior that we observe for all books with this approach.
\section{Discussion of extended results}\label{app:sec:book-level-discussion}

We simply reiterate our main takeaway, which is clear from the extended results we present in the Appendix.
Memorization (as measured by a specific technique that focuses on probabilistic extraction, see Appendix~\ref{app:sec:background}) varies significantly across models for the same book, and across books for the same model (Section~\ref{sec:book:compare} \& Appendix~\ref{app:sec:sliding-window}).
Average extraction rates---the typical metric in ML memorization research---do not reveal this nuance.
Instead, it is necessary to examine per-book extraction metrics to get this view.

\paragraph{We need to run more books to make general claims about overall memorization.}
We only ran experiments for a small fraction of the entire \texttt{Books3} dataset---running detailed experiments for $200$ books (Appendix~\ref{app:sec:sliding-window}).
For $100$ books, we took care to sample books from a variety of sources---in copyright, in the public domain, openly licensed. 
We included popular books, as well as more obscure ones (Table~\ref{app:tab:sliding-books}). 
We randomly sampled the other $100$. 
Overall, very popular books exhibit the most memorization.
It seems likely that these books are duplicated on different parts of the Internet; 
de-duplication is a challenging problem to implement in practice~\citep{lee2022dedup}, so it is likely that least some duplicate text persists in training datasets for LLMs.\looseness=-1

\paragraph{We do not know exactly what is going on with \textsc{Llama 3+} models.} 
Even so, it seems unlikely that (relatively small numbers of) duplicates completely explain the patterns we observe. 
\textsc{Llama 3.1 70B} exhibits a lot more memorization than any other model. 
We find higher amounts and degrees of extraction compared to \textsc{Llama 3}, \textsc{Llama 2 70B}, and \textsc{Llama 1 65B} (models of a similar size also trained by Meta).
It also exhibits more memorization than  a model that registers lower loss and superior performance on downstream benchmarks (\textsc{Qwen 2.5 72B}).
That is, as our results also show for \textsc{Llama 3.1 70B} on books in the training data that it has \emph{not} memorized, it is not the case that the model's loss is ``low everywhere,'' and that this is responsible for the high suffix probabilities we observe.
First, that is not how machine learning works;
the model cannot so-strongly fit to all of the massive dataset it was trained on.
Second, we do not observe this anyway for our results. 
And third, we do not observe the same degree of memorization in a model that exhibits lower loss on average than \textsc{LLama 3.1 70B}.

\paragraph{High-level memorization patterns.}
In general, we observe a pattern that later generations of \textsc{Llama} models memorize more than earlier ones, with respect to average extraction rates (Appendix~\ref{app:sec:rates}) and extraction coverage (Appendix~\ref{app:sec:coverage}). 
Most books we test exhibited minimal memorization, measured with respect to probabilistic extraction with top-$k$ decoding ($T\!=\!1$, $k\!=\!40$) and $100$-token sequences ($50$-token prefixes $+$ $50$-token suffixes).
Our results on longer prefixes (Appendix~\ref{app:sec:validity}), however, indicate that there is a lot more memorization to uncover---even if it is harder to extract, where hardness is loosely equated with requiring a longer prefix.\looseness=-1

Nevertheless, much of the memorization that we do observe for books in our main sliding-window experiments frequently falls into one of a few categories: 
copyright notices, lists of ordered (page) numbers (a common artifact of converting \textsc{epub} files to \textsc{txt} format, see Appendix~\ref{app:sec:validity:phi4}), publisher addresses, chapter number listings, and author biographies.
All of these are types of text that are highly duplicated (partially or exactly) in various sources, both the Internet and across books. 
And so, extraction of a copyright notice from a given book does not necessarily mean it was memorized from that book;
it was likely memorized due to the presence of numerous similar pieces of text in the training data.
The same is also true for author biographies, which are printed on websites, not just in books.\looseness=-1

Another category was text from popular sources that are quoted within books: 
quotes from the Bible~\citep{Embraced, Unglued, We_Were_Eight_Years_in_Power, emperorgladness}, philosophers like John Stuart Mill~\citep{The_Future_of_the_Internet_and_How_to_Stop_It}, classics like those by Dante Alighieri~\citep{Dante_and_the_Origins_of_Italian_Literary_Culture}, text from U.S. government documents~\citep{The_Future_of_the_Internet_and_How_to_Stop_It}, famous political speeches~\citep{We_Were_Eight_Years_in_Power}, public domain books~\citep{careless}, social media posts referenced in numerous news articles~\citep{careless}, and famous quotes posted on blogs and platforms like Goodreads~\citep{Beloved}. 
In many cases, this was the only such text we were able to extract with non-trivial probability from some books using $50$ token prompts. 
For our negative controls, we painstakingly verify that each piece of text that we extract within a sequence can be found on multiple sources on the Internet. 

With respect to the books that we tested that are within the scope of the~\citet{kadreyamendedconsolidated} class action suit, we were not able to extract much memorized training data.
There were notable exceptions (e.g., \citet{We_Were_Eight_Years_in_Power}). 
It is of course possible that another extraction technique could reveal additional memorization, but we were unable to meaningfully do so for most books using a $50$-token prefixes.
It is also important to note that suit was recently decided in favor of the defendants (Meta), and that memorization of training data was only one issue being argued.

\paragraph{Our memorization measurements do not promise that reconstruction naturally follows.}
We emphasize again that, even for highly memorized books, our sliding-window experiments are \emph{not} extracting enormous amounts of text in one go, i.e., with a single seed prompt to a model.  
That is \emph{not} what we focused on in this project. 
However, we do show that it is possible to reconstruct a highly memorized book by using an LLM autoregressively, starting with a single seed prompt of ground-truth text (Section~\ref{sec:book:seed} \& Appendix~\ref{app:sec:reconstruct}).
This type of experiment differs from the main type of results that we showcase in this paper, which use probabilistic discoverable extraction to quantify memorization (Appendix~\ref{app:sec:background}). 
We have made a significant effort throughout to make this point clear. 
If something remains unclear, please reach out to the corresponding authors. 

\begin{figure}[t!]
        \includegraphics[width=\linewidth]{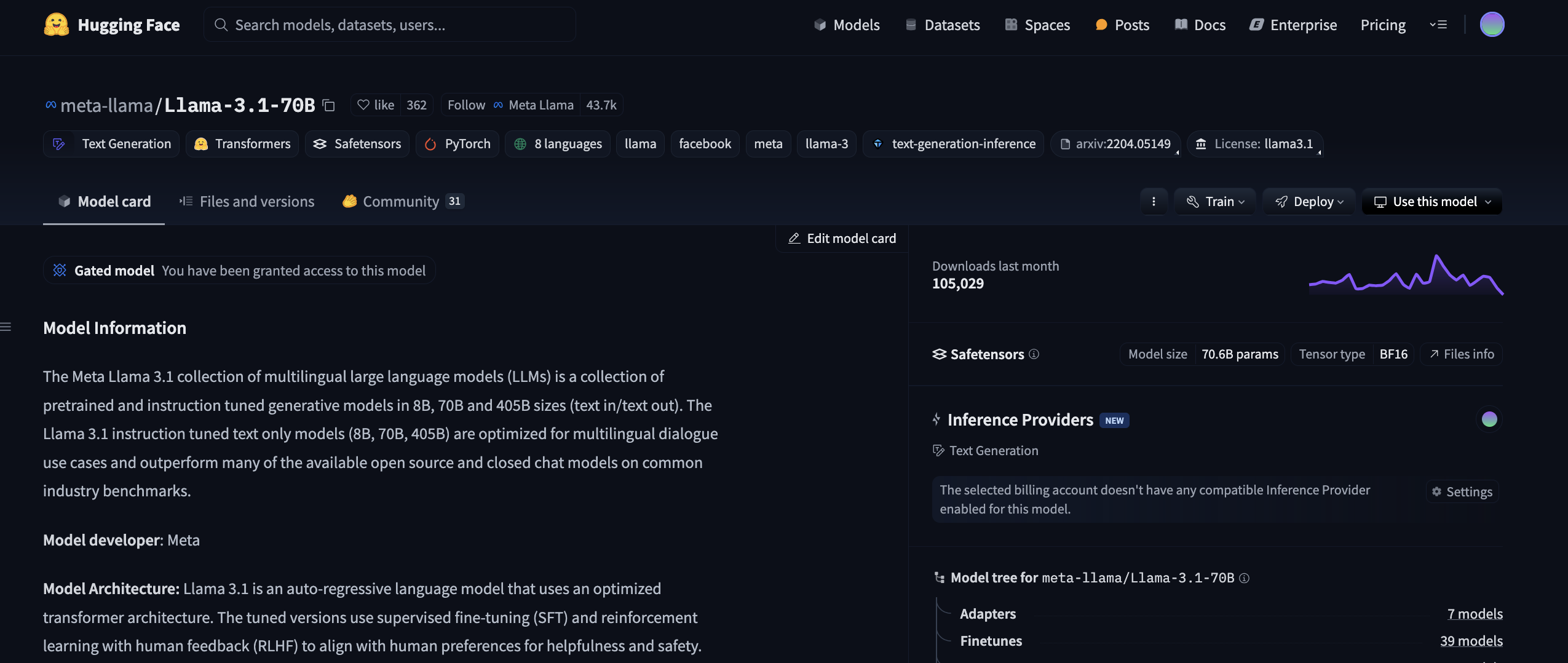}
    \caption{\textbf{Screenshot from HuggingFace.} Downloads of \textsc{Llama 3.1 70B}, May 2025.}
    \label{fig:hfllamadownloads}
\end{figure}

\paragraph{Final comments on amount of books memorization}
We were able to extract relatively enormous amounts of memorized training data in some cases, from both public domain books (e.g.,\citet{The_Great_Gatsby, Alice_s_Adventures_in_Wonderland, Ulysses}) and popular in-copyright books of all stripes (e.g., \citet{Harry_Potter_and_the_Sorcerer_s_Stone, A_Game_of_Thrones, Lean_In, The_Da_Vinci_Code, The_Hobbit, The_Myth_of_Sisyphus}). 
\textsc{Llama 3+} models exhibited an order of magnitude more memorization on some books than our worst-case estimates prior to starting this project.
It was this enormous extent of memorization that encouraged us to try to reconstruct a book starting from a single seed prompt; 
we would not have believed such an outcome was possible prior to measuring memorization in \textsc{Llama 3+} models---certainly not a flagship model from a frontier company, which continues to be downloaded and used by millions of users (Figure~\ref{fig:hfllamadownloads}).

\end{document}